\PassOptionsToPackage{table}{xcolor}
\PassOptionsToPackage{sort&compress}{natbib}
\documentclass[onecolumn]{dataflow}
\usepackage{amsmath,amssymb,tabularx,adjustbox}
\newif\ifNeurCoTableFloat
\AtBeginEnvironment{table}{\NeurCoTableFloattrue}
\AtBeginEnvironment{table*}{\NeurCoTableFloattrue}
\let\NeurCoNativeTabular\tabular
\let\NeurCoNativeEndTabular\endtabular
\newsavebox{\NeurCoZeroTableBox}
\newsavebox{\NeurCoUnitTableBox}
\newsavebox{\NeurCoFullWidthTableBox}

\ExplSyntaxOn
\tl_new:N \l__povi_table_columns_tl
\dim_new:N \l__povi_table_zero_dim
\dim_new:N \l__povi_table_step_dim

\cs_new_protected:Npn \__povi_native_table:nnn #1#2#3
  {
    \NeurCoNativeTabular[#1]{#2}
    #3
    \NeurCoNativeEndTabular
  }

\cs_new_protected:Npn \__povi_fixed_table:nnn #1#2#3
  {
    \group_begin:
      \cs_set_eq:NN \NeurCoSavedMulticolumn \multicolumn
      \cs_set:Npn \multicolumn ##1##2##3
        { \NeurCoSavedMulticolumn{##1}{##2}{} }
      \dim_zero:N \tabcolsep
      \sbox{\NeurCoZeroTableBox}{\__povi_native_table:nnn {#1}{#2}{#3}}
      \dim_set:Nn \tabcolsep {1pt}
      \sbox{\NeurCoUnitTableBox}{\__povi_native_table:nnn {#1}{#2}{#3}}
      \cs_set_eq:NN \multicolumn \NeurCoSavedMulticolumn

      \dim_set:Nn \l__povi_table_zero_dim {\box_wd:N \NeurCoZeroTableBox}
      \dim_set:Nn \l__povi_table_step_dim
        {\box_wd:N \NeurCoUnitTableBox - \l__povi_table_zero_dim}
      \dim_set:Nn \tabcolsep
        {
          \fp_to_dim:n
            {
              max(0,
                (\dim_to_fp:n{\textwidth}
                  - \dim_to_fp:n{\l__povi_table_zero_dim})
                / \dim_to_fp:n{\l__povi_table_step_dim}
              )
            }
        }
      \sbox{\NeurCoFullWidthTableBox}{\__povi_native_table:nnn {#1}{#2}{#3}}
      \typeout{NeurCo~table~\thetable\c_colon_str~width=\the\wd\NeurCoFullWidthTableBox;~textwidth=\the\textwidth;~padding=\the\tabcolsep}
      \begin{adjustbox}{max~width=\textwidth}\clipbox{0pt~0pt~0pt~0pt}{\usebox{\NeurCoFullWidthTableBox}}\end{adjustbox}
    \group_end:
  }

\NewDocumentCommand{\NeurCoFullWidthTable}{m m +m}
  {
    \tl_set:Nn \l__povi_table_columns_tl {#2}
    \tl_if_eq:nnT {#2} {@{}llcp{6.6cm}l@{}}
      {\tl_set:Nn \l__povi_table_columns_tl {llcp{6.6cm}l}}
    \tl_if_eq:nnT {#2} {@{}lcccc@{}}
      {\tl_set:Nn \l__povi_table_columns_tl {lcccc}}
    \tl_if_eq:nnT {#2} {@{}lcc@{}}
      {\tl_set:Nn \l__povi_table_columns_tl {lcc}}
    \tl_if_eq:nnT {#2} {lcccccccc@{\hskip8pt}cccc}
      {\tl_set:Nn \l__povi_table_columns_tl {lcccccccccccc}}
    \exp_args:NnV \__povi_fixed_table:nnn {#1} \l__povi_table_columns_tl {#3}
  }
\ExplSyntaxOff

\RenewDocumentEnvironment{tabular}{O{c} m +b}{%
  \ifNeurCoTableFloat
    \NeurCoFullWidthTable{#1}{#2}{#3}%
  \else
    \begin{adjustbox}{max width=\linewidth}%
      \NeurCoNativeTabular[#1]{#2}%
      #3%
      \NeurCoNativeEndTabular
    \end{adjustbox}%
  \fi
}{}
\AtBeginEnvironment{tabularx}{%
  \let\tabular\NeurCoNativeTabular
  \let\endtabular\NeurCoNativeEndTabular
}
\BeforeBeginEnvironment{tabular*}{\begin{adjustbox}{max width=\linewidth}}
\AfterEndEnvironment{tabular*}{\end{adjustbox}}

\newif\ifNeurCoSingleColumnFigure
\AtBeginEnvironment{figure}{%
  \NeurCoSingleColumnFiguretrue
  \captionsetup{width=\textwidth}%
}
\AtBeginEnvironment{figure*}{%
  \NeurCoSingleColumnFigurefalse
  \captionsetup{width=\textwidth}%
}
\let\NeurCoUnscaledIncludeGraphics\includegraphics

\NewDocumentCommand{\NeurCoDrawOriginalGraphics}{m m m}{%
  \IfBooleanTF{#1}{%
    \NeurCoUnscaledIncludeGraphics*[#2]{#3}%
  }{%
    \NeurCoUnscaledIncludeGraphics[#2]{#3}%
  }%
}

\RenewDocumentCommand{\includegraphics}{s O{} m}{%
  \ifNeurCoSingleColumnFigure
    \ifdim\linewidth=\textwidth
      \typeout{NeurCo single-column graphic: #3; width=\the\textwidth}%
      \ifstrempty{#2}{%
        \NeurCoDrawOriginalGraphics{#1}{width=\textwidth}{#3}%
      }{%
        \NeurCoDrawOriginalGraphics{#1}{#2,width=\textwidth}{#3}%
      }%
    \else
      \NeurCoDrawOriginalGraphics{#1}{#2}{#3}%
    \fi
  \else
    \NeurCoDrawOriginalGraphics{#1}{#2}{#3}%
  \fi
}

\let\NeurCoFloatFigure\figure
\let\NeurCoFloatEndFigure\endfigure
\expandafter\let\expandafter\NeurCoFloatWideFigure\csname figure*\endcsname
\expandafter\let\expandafter\NeurCoFloatEndWideFigure\csname endfigure*\endcsname
\RenewDocumentEnvironment{figure}{O{!htbp}}{%
  \NeurCoFloatFigure[!htbp]%
}{\NeurCoFloatEndFigure}
\RenewDocumentEnvironment{figure*}{O{!htbp}}{%
  \NeurCoFloatWideFigure[!htbp]%
}{\NeurCoFloatEndWideFigure}

\let\NeurCoFloatTable\table
\let\NeurCoFloatEndTable\endtable
\expandafter\let\expandafter\NeurCoFloatWideTable\csname table*\endcsname
\expandafter\let\expandafter\NeurCoFloatEndWideTable\csname endtable*\endcsname
\RenewDocumentEnvironment{table}{O{!htbp}}{%
  \NeurCoFloatTable[!htbp]%
}{\NeurCoFloatEndTable}
\RenewDocumentEnvironment{table*}{O{!htbp}}{%
  \NeurCoFloatWideTable[!htbp]%
}{\NeurCoFloatEndWideTable}

\AtBeginDocument{%
  \pretocmd{\section}{\FloatBarrier}{}{%
    \PackageError{NeurCo-layout}{Cannot attach section float barrier}{}%
  }%
  \pretocmd{\subsection}{\FloatBarrier}{}{%
    \PackageError{NeurCo-layout}{Cannot attach subsection float barrier}{}%
  }%
}

\renewcommand{\templatelogo}{%
  \includegraphics[height=9mm]{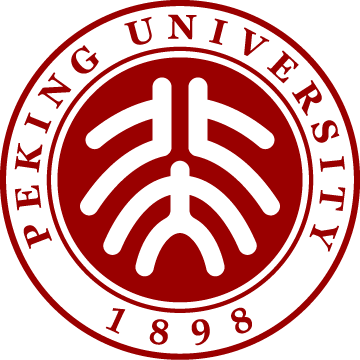}%
  \hspace{3mm}%
  \includegraphics[height=9mm]{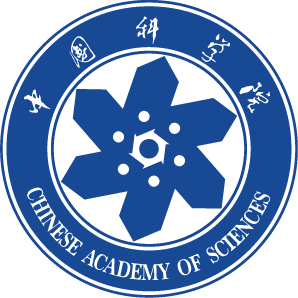}%
  \hspace{4mm}%
  \includegraphics[height=9mm]{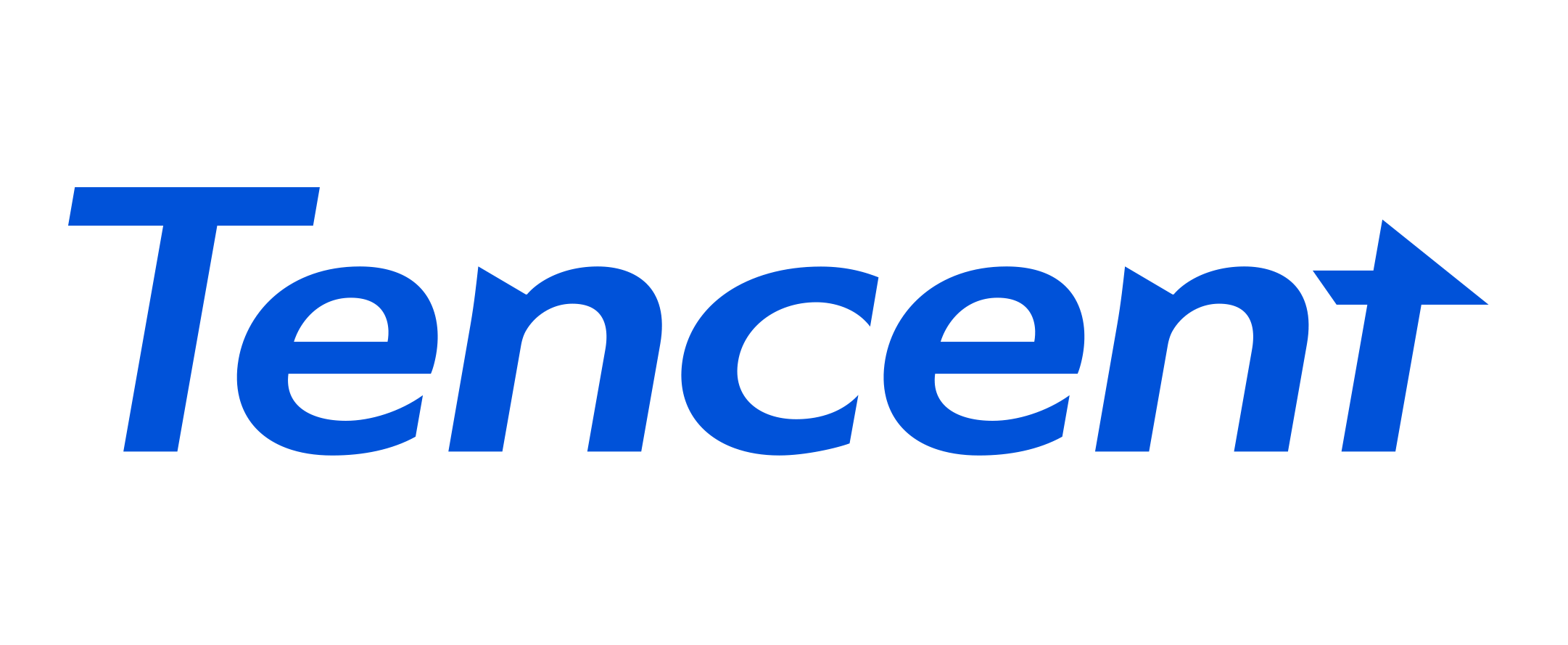}%
}
\fancypagestyle{firststyle}{%
  \fancyhf{}
  \renewcommand{\headrulewidth}{0pt}
  \fancyhead[L]{\templatelogo}
  \fancyhead[R]{%
    \raisebox{2.5mm}{\normalfont\small\textcolor[HTML]{4D4D4D}{\today}}%
  }
}
\renewcommand{\titlefont}{%
  \sffamily\fontsize{16}{20}\selectfont\raggedright
  \hyphenpenalty=10000\relax\exhyphenpenalty=10000\relax
}
\DeclareFontShape{T1}{bytesans}{m}{it}{<->ssub*lmss/m/sl}{}
\DeclareFontShape{T1}{bytesans}{b}{it}{<->ssub*lmss/bx/sl}{}
\DeclareFontShape{T1}{bytesans}{bx}{it}{<->ssub*lmss/bx/sl}{}
\setcitestyle{numbers,square,comma}
\providecommand{\thetitle}{{\sffamily\fontsize{16}{20}\selectfont\thetitletext}}

\let\NeurCoOriginalAppendix\appendix
\renewcommand{\appendix}{%
  \NeurCoOriginalAppendix
  \renewcommand{\thetitle}{{\sffamily\fontsize{16}{20}\selectfont\thetitletext}}%
  \let\NeurCoAppendixSetCounter\setcounter
  \renewcommand{\setcounter}[2]{%
    \ifstrequal{##1}{page}{}{\NeurCoAppendixSetCounter{##1}{##2}}%
  }%
  \renewcommand{\pagenumbering}[1]{}%
}

\let\NeurCoTemplateAbstract\abstract
\let\NeurCoTemplateMakeTitle\maketitle
\renewcommand{\maketitle}{}
\RenewDocumentEnvironment{abstract}{+b}{%
  \NeurCoTemplateAbstract{#1}%
  \NeurCoTemplateMakeTitle
}{}

\RenewDocumentCommand{\addtolist}{O{}+m+m+m+m}{%
  \ifdefempty{#3}{}{\gappto#3{#5}}%
  \gappto#3{#4[#1]{#2}}%
}

\renewcommand{\email}[1]{%
  \href{mailto:#1}{\textcolor{black}{\normalfont\small\detokenize{#1}}}%
}

\renewcommand{\checkdataformat}[2][]{%
  \par\noindent
  \parbox[t]{55pt}{%
    \checkdatafont\sffamily\bfseries\strut\mbox{#1:}}%
  \parbox[t]{\dimexpr\linewidth-55pt\relax}{%
    \normalfont\small\raggedright\strut #2}\par
}

\makeatletter
\renewcommand{\mymaketitle}{%
  \vspace*{-4mm}%
  \@toptitlebar
  \begin{tcolorbox}[
    enhanced,colback=boxbg,frame hidden,arc=3mm,
    left=8mm,right=8mm,top=5mm,bottom=5mm,
    before skip=0pt,after skip=0pt
  ]
    \setlength{\parskip}{0pt}
    \setlength{\parindent}{0pt}
    {\titlelist\par}
    \ifdefempty{\authorlist}{}{\vskip 3mm {\centering\authorlist\par}}
    \ifdefempty{\affiliationlist}{}{\vskip 2mm {\centering\affiliationlist\par}}
    \vskip 3mm
    \abstractlist\par
    \ifdefempty{\contributionlist}{}{%
      \vskip 2mm
      \contributionlist\par
    }
    \ifdefempty{\checkdatalist}{}{%
      \vskip 2mm
      \checkdatalist\par
    }
  \end{tcolorbox}
  \tcbset{reset}
  \FloatBarrier
  \thispagestyle{firststyle}
}
\makeatother
\renewcommand{\NeurCoTemplateMakeTitle}{\mymaketitle\vskip 4mm}

\usepackage{cuted} 

\usepackage{currfile} 

\usepackage{caption} 
\definecolor{cvprblue}{rgb}{0.21,0.49,0.74}

\usepackage{booktabs}
\usepackage{multirow}
\usepackage{amsmath}
\usepackage{tabularx}
\usepackage{placeins}

\usepackage[table]{xcolor}

\definecolor{bestblue}{RGB}{226,238,249}

\usepackage{tikz}
\newsavebox{\caseimage}
\usepackage{graphicx}
\usepackage{xcolor}
\usepackage{tikz}

\usepackage[table]{xcolor}

\definecolor{bestblue}{RGB}{226,238,249}

\usepackage[most]{tcolorbox}
\usepackage{enumitem}

\newtcolorbox{promptcard}[2][]{
  enhanced,
  breakable,
  colback=white,
  colframe=black!22,
  colbacktitle=orange!7,
  coltitle=black,
  title={#2},
  fonttitle=\bfseries\small,
  fontupper=\footnotesize,
  fontlower=\footnotesize,
  boxrule=0.45pt,
  arc=0.8mm,
  left=1.8mm,
  right=1.8mm,
  top=1.4mm,
  bottom=1.4mm,
  titlerule=0.35pt,
  borderline west={1.8pt}{0pt}{orange!65!black},
  before skip=7pt,
  after skip=9pt,
  #1
}

\title{Neural Rendering Compiler: Generative Video Models as Programmable Renderers}

\author[1,*]{Yuxia Zhu}
\author[2,*]{Xinglong Xu}
\author[1]{Hao Liang}
\author[2]{Linzhuang Sun}
\author[2]{Caijun Jia}
\author[1]{Qifeng Cai}
\author[1]{Jincen Luo}
\author[3]{Lei Li}
\author[3]{Alden Cai}
\author[3,\dagger]{Asher Qin}
\author[1,\dagger]{Wentao Zhang}
\author[3,\dagger]{Cheng Tan}

\affiliation[]{%
  $^{1}$Peking University\quad
  $^{2}$University of Chinese Academy of Sciences\quad
  $^{3}$Tencent
}
\contribution[*]{Equal contribution.}
\contribution[\dagger]{Corresponding author.}

\newcommand{\emailicon}{\raisebox{-1.5pt}{\includegraphics[height=1.05em]{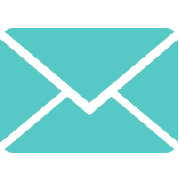}}}

\newcommand{\sourcelink}
{\raisebox{-1.5pt}{\includegraphics[height=1.05em]{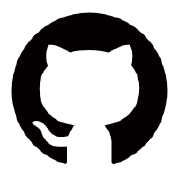}}}

\newcommand{\websitelink}
{\raisebox{-1.5pt}{\includegraphics[height=1.05em]{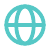}}}

\newcommand{\contactlabel}[2]{%
  \makebox[11pt][c]{#1}\hspace{3pt}#2}

\checkdata[\contactlabel{\emailicon}{Email}]{%
  \email{Zzzyxiia@outlook.com}}

\checkdata[\contactlabel{\sourcelink}{Code}]{%
  \href{https://github.com/OpenRaiser/NeurCo}
       {https://github.com/OpenRaiser/NeurCo}}

\checkdata[\contactlabel{\websitelink}{Website}]{%
  \href{https://openraiser.github.io/NeurCo}
       {https://openraiser.github.io/NeurCo}}

\makeatletter
\AtBeginDocument{%
  \if@twocolumn
    \PackageError{NeurCo-layout}{Expected single-column layout}{Compile main.tex.}%
  \fi
  \typeout{NeurCo layout: single column; columnwidth=\the\columnwidth; textwidth=\the\textwidth}%
}
\makeatother

\begin{document}
\maketitle

\begin{abstract}
Generative video models are becoming powerful renderers, yet lack the predictable control interfaces of conventional systems such as HTML/CSS engines or Blender. Their capabilities emerge implicitly from training, making it difficult to determine what they can reliably execute. We present Neural Rendering Compiler (NeurCo), which automatically probes frozen video models to construct model-specific capability manuals describing which semantic requirements they can reliably satisfy. NeurCo represents user intent in a model-agnostic intermediate representation of objects, constraints, events, and cameras, then compiles it into model-specific execution strategies. Reliable requirements remain unchanged, while unreliable ones are selectively rewritten or compiled into alternative execution strategies. We introduce VComBench, a benchmark of 380 compositional, multi-requirement tasks spanning text-to-video, image-to-video, and reference-to-video generation. NeurCo improves requirement satisfaction from 46.8\% to 52.3\% on Wan and from 66.5\% to 69.3\% on Minimax H3, and achieves further gains when combined with prompt enhancers, including on Seedance 2.5. Our analysis reveals model-specific differences in control capabilities and trade-offs between controllability and photorealism, highlighting capability-guided compilation as a promising approach to programmable generative video rendering.
\end{abstract}

\begin{figure}[t]
    \centering
    \includegraphics[width=\columnwidth]{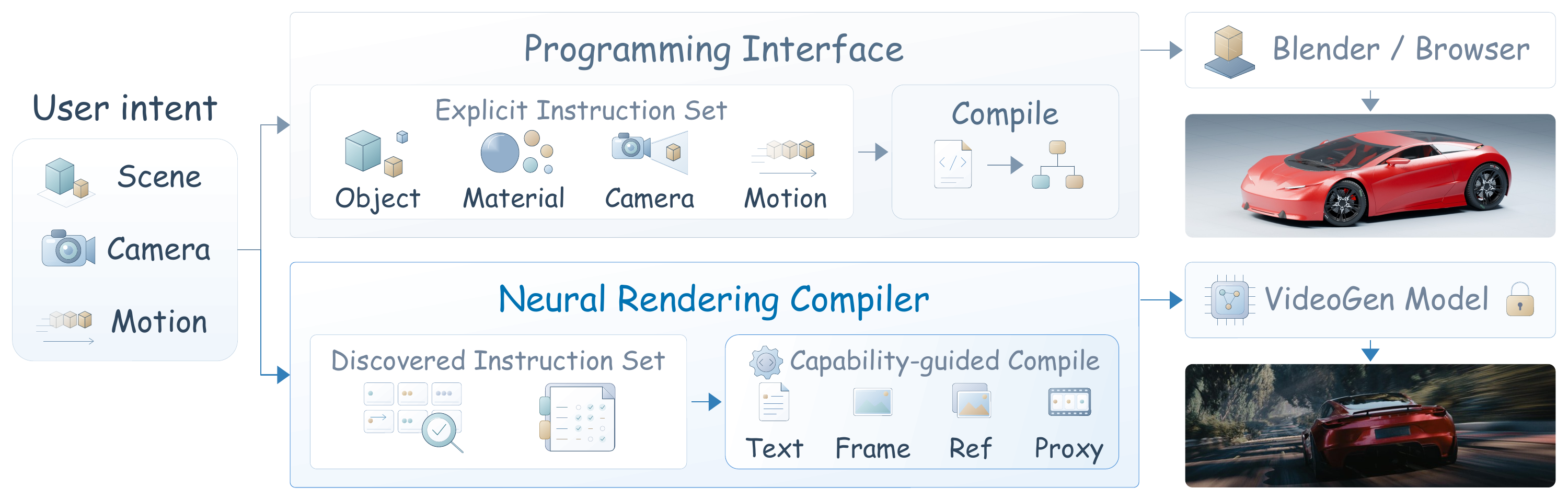}
    \caption{From conventional to generative rendering. NeurCo enables programmable control of generative video models through capability profiling and model-specific compilation.}
    \label{fig:compare}
\end{figure}

\section{Introduction}
\label{sec:intro}
Rendering is more than producing convincing visuals. It requires precise, composable control over how they are created. In conventional engines such as Blender, operations have well-defined semantics~\cite{Cai_2024_CVPR}. Moving a camera changes the viewpoint without altering scene geometry, while changing a material leaves an object's motion intact. Such predictable effects allow complex scenes to be programmed. Generative video models instead synthesize rich scenes and motion directly from text or visual references~\cite{ICLR2025_ce31378e,xing2023dynamicrafter,Jiang_2025_ICCV,team2026bernini}, but offer no comparable execution contract~\cite{he2025cameractrlenablingcameracontrol}. The effective control capabilities emerge implicitly from training and vary across models and conditioning modes~\cite{Huang_2024_CVPR,zheng2025vbench20advancingvideogeneration,huang2024vbenchcomprehensiveversatilebenchmark}. Figure~\ref{fig:compare} illustrates this contrast between explicit and empirically discovered rendering controls. This raises a fundamental question: \textit{How can we program a generative renderer whose instruction set is unknown in advance?}

\begin{figure*}[t]
    \centering
    \includegraphics[width=\textwidth]{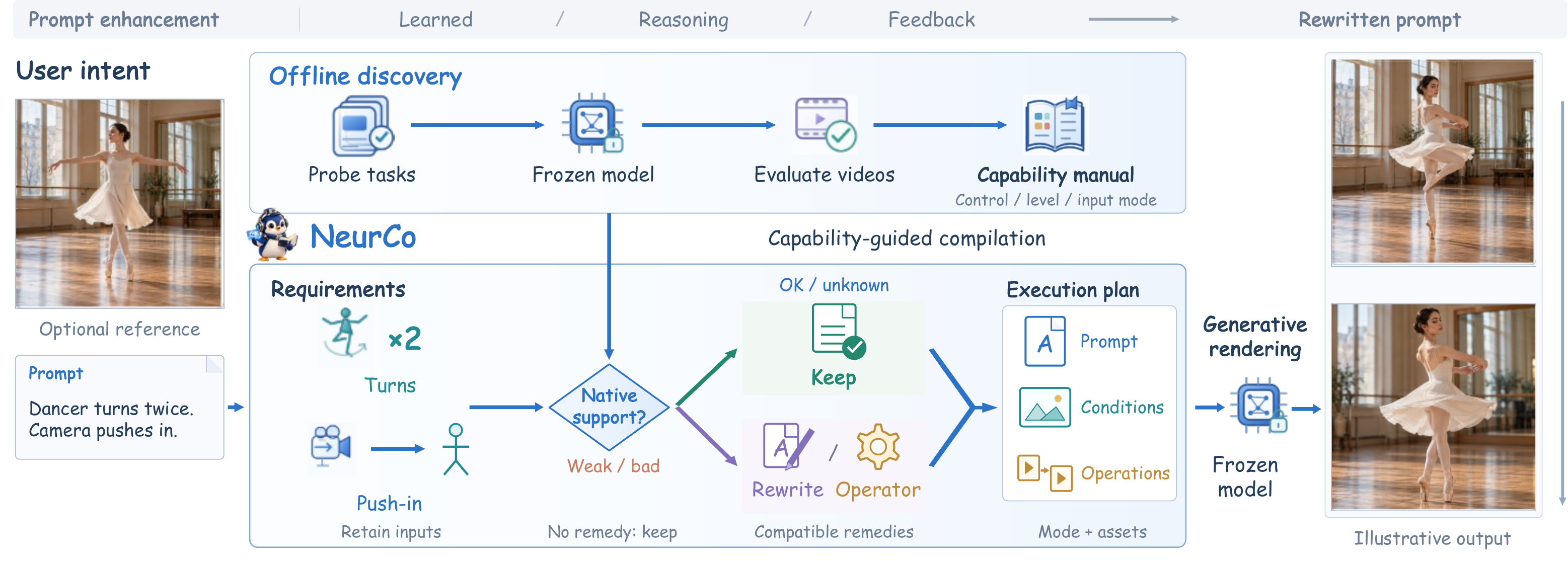}
    \caption{\textbf{Overview of NeurCo.}
    Offline profiling builds a deployment-specific capability manual.
    Online compilation uses this manual to retain or adapt
    requirements and assemble an execution plan containing prompts,
    visual conditions, and operations for a video model.
    }
    \label{fig:overview}
\end{figure*}

Existing approaches improve the controllability of video generation from different directions. Prompt enhancers refine user instructions~\cite{Ji_2025_ICCV,Cheng_2025_ICCV}, agentic pipelines~\cite{long2025vistatesttimeselfimprovingvideo,Motionagent} decompose requests into plans or tool calls, and control-conditioned models introduce explicit camera or motion controls~\cite{he2025cameractrlenablingcameracontrol,Wang_2024}. However, they do not systematically characterize which requirements a given model can reliably execute or how control strategies should adapt to its capabilities. Consequently, they may unnecessarily rewrite requirements already satisfied by the model while failing to identify when alternative execution strategies are needed.

We approach this problem from a compiler perspective, treating generative video models as rendering backends~\cite{zhang2025scene,Kizil_2026_CVPR}. Conventional compilers target instruction sets with well-defined semantics, whereas generative renderers lack explicit execution specifications and exhibit uneven control capabilities~\cite{Huang_2024_CVPR,Sun_2026_CVPR,huang2024vbenchcomprehensiveversatilebenchmark}. Rather than imposing a universal interface, we systematically probe frozen models using a shared vocabulary of semantic primitives to identify which requirements they can reliably satisfy. The resulting model-specific capability manual guides the compilation of high-level user intent into appropriate execution strategies. Reliable requirements are preserved, while challenging ones are selectively rewritten or lowered into alternative controls~\cite{he2025cameractrlenablingcameracontrol,motion_prompting,Ren_2025_CVPR}. Thus, while the primitive vocabulary is predefined, each renderer's effective instruction set is discovered rather than assumed.

We introduce Neural Rendering Compiler (NeurCo), which treats frozen video models as black-box rendering backends. As illustrated in Figure~\ref{fig:overview}, NeurCo systematically probes each model to construct a capability manual characterizing the reliability of semantic controls across conditioning modes. Given a user request, NeurCo identifies individual semantic requirements and consults the manual to determine appropriate execution strategies. Reliable requirements remain unchanged, while challenging ones are selectively rewritten or lowered into alternative controls, including reference images, first frames, and proxy videos. Unchanged inputs reuse native outputs, and the same compilation framework adapts to different rendering backends. We further introduce VComBench, comprising 380 challenging compositional tasks across text-to-video, image-to-video, and reference-to-video generation, evaluating multiple interacting requirements across conditioning modes. It measures both individual requirement satisfaction and whether all requirements are jointly satisfied, revealing potential interference between controls.


We evaluate NeurCo on three video model families: MiniMax-H3, Wan, and Seedance 2.5. On H3 and Wan, NeurCo achieves substantial improvements in requirement satisfaction, with larger gains on less capable backends. It also complements existing prompt enhancement methods, delivering further improvements on Seedance 2.5 across all three benchmarks. Rather than indiscriminately rewriting prompts, NeurCo selectively targets challenging requirements while preserving native generation when no intervention is needed. Our analysis further reveals that different execution strategies offer varying benefits across task types, while auxiliary controls can improve controllability at the cost of photorealism. 
Together, these findings highlight the value of capability-guided compilation and underscore the need for empirical specifications that characterize not only what generative renderers can reliably execute, but also the limitations and trade-offs that compilers must account for.

\section{Related Work}
\label{sec:related_work}

\subsection{Video Generation Evaluation}
\label{sec:rw_evaluation}

Video generation evaluation has shifted from distribution-level metrics such as FVD~\cite{unterthiner2019accurategenerativemodelsvideo} toward capability diagnosis.
FETV~\cite{NEURIPS2023_c481049f} categorizes prompts by content, attribute, and complexity; EvalCrafter~\cite{Liu_2024_CVPR} and VBench~\cite{Huang_2024_CVPR} assess visual quality, motion, temporal consistency, and prompt alignment, with VBench++~\cite{huang2024vbenchcomprehensiveversatilebenchmark} adding image-to-video generation and VBench-2.0~\cite{zheng2025vbench20advancingvideogeneration} covering human fidelity, controllability, physics, and commonsense.
Targeted benchmarks probe compositionality~\cite{Sun_2026_CVPR}, temporal change~\cite{feng-etal-2025-tc}, consecutive events~\cite{Wang_2025_CVPR}, physical commonsense and laws~\cite{bansal2024videophyevaluatingphysicalcommonsense,lin2026phygroundbenchmarkingphysicalreasoning}, scientific causality~\cite{zhang2026scivbenchevaluatingknowledgereasoningintensive}, and rule adherence and goal realization~\cite{luo2026evaluationenhancementbenchmarkingimproving}.
ETVA~\cite{Guan_2025_ICCV} checks fine-grained alignment, and PEBench~\cite{shao2026unifiedevaluationpromptenhancers} evaluates enhanced prompts.

\subsection{Prompt Enhancement for Video Generation}
\label{sec:rw_prompt_enhancement}

Prompt enhancement improves video generation by adapting user descriptions. Existing methods employ preference learning and reward-guided rewriting~\cite{Ji_2025_ICCV,Cheng_2025_ICCV}, retrieval-augmented optimization~\cite{Gao_2025_CVPR}, and iterative feedback from generated videos~\cite{gao2026rapocrossstagepromptoptimization,jia2026capet2vcaptioneranchoredpromptenhancement}. Recent approaches further explore cinematic planning and reasoning-based rewriting~\cite{zhu2026wanpecinematicpromptenhancement,luo2026evaluationenhancementbenchmarkingimproving}, multi-agent refinement and semantic verification~\cite{long2025vistatesttimeselfimprovingvideo,yang2026scmaprselfcorrectingmultiagentprompt}, reusable optimization rules~\cite{zhou2026tidetrajectoryinformedskilldistillation}, and reference-guided prompt refinement~\cite{Zhao_2026_CVPR}. Unlike these approaches, which primarily optimize prompt formulation, NeurCo uses empirically discovered, model-specific capabilities to selectively preserve, rewrite, or lower requirements into alternative execution strategies.

\subsection{Structured and Programmatic Video Control}
\label{sec:rw_structured_control}

Structured video control uses layouts and interaction graphs to specify scenes and relationships~\cite{lian2024llmgroundedvideodiffusionmodels,Lin2023VideoDirectorGPT,shah2026graphvid}, while image and reference conditioning improve appearance and temporal consistency~\cite{xing2023dynamicrafter,zhang2023i2vgenxlhighqualityimagetovideosynthesis,ren2024consisti2venhancingvisualconsistency,Jiang_2024_CVPR,Chen_2025_CVPR}. Explicit camera, motion, and structural controls further enable fine-grained manipulation~\cite{Wang_2024,he2025cameractrlenablingcameracontrol,yin2023dragnuwafinegrainedcontrolvideo,Yang_2024,chen2024controlavideocontrollabletexttovideodiffusion,Sun_2026_CVPR}. Agentic systems extend these capabilities through tool orchestration, planning, and feedback-driven refinement~\cite{wei2026videoweaverevaluatingevolvingskills,feng2026newtonagenticplanningphysically}. Programmatic approaches combine executable scene representations or simulations with neural rendering~\cite{li2026videocococodeascotphysicallyconsistentvideo,Liang_2025_CVPR,chi2026agentgartencodeworldsevolving}, although coding-based visual generation may sacrifice perceptual richness~\cite{zhou2026farcodingagentsvisual}. NeurCo empirically discovers model-specific control capabilities and selectively compiles requirements into textual instructions, visual conditions, or external operators without retraining the video generator.

\section{Preliminaries}
\label{sec:preliminaries}

\subsection{Task Formulation}
\label{sec:task_formulation}

A user request \(x=(p,A,m_0)\) comprises a prompt \(p\), optional visual assets \(A\), and an initial conditioning mode \(m_0\in\{\mathrm{T2V},\mathrm{I2V},\mathrm{R2V}\}\) for text-, first-frame-, and reference-to-video generation, respectively.

\begin{figure*}[ht]
    \centering
    \includegraphics[width=\textwidth]{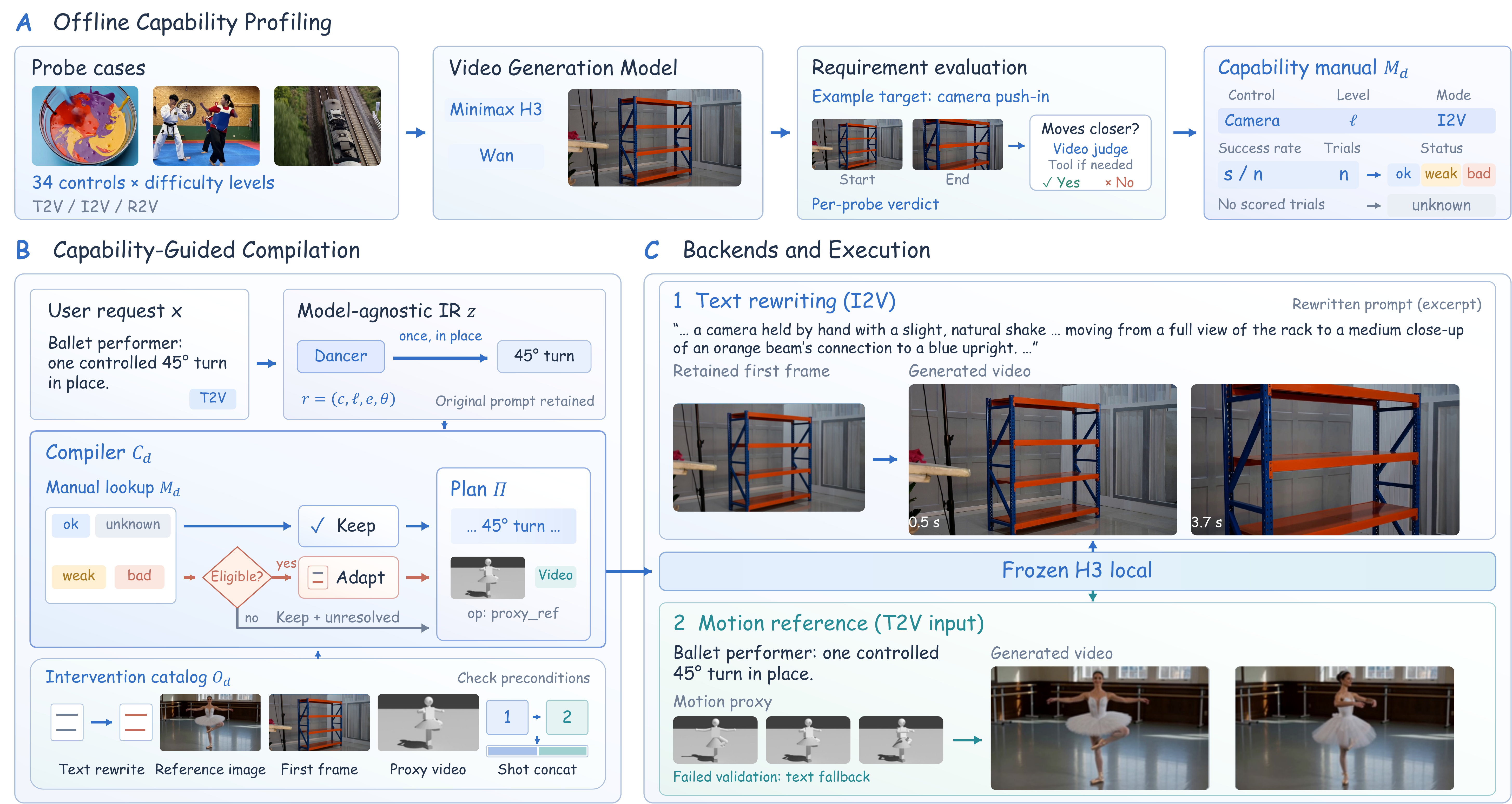}
    \caption{\textbf{NeurCo pipeline.}
    (A) Offline probing builds deployment-specific capability manuals.
    (B) Requests are parsed into an IR and compiled using the manual
    and compatible interventions.
    (C) Two examples illustrate targeted rewriting with a retained
    first frame and motion control through a proxy video,
    both using a frozen generator.}
    \label{fig:method}
\end{figure*}

NeurCo parses \(x\) into semantic requirements \(\mathcal R(x)=\{r_i\}_{i=1}^{K}\), where \(r_i=(c_i,\ell_i,e_i,\theta_i)\) records a predefined control type, difficulty level, source text span, and optional arguments (subject, count, order, or time).
Controls span object properties and counts, spatial relations, events, cameras, appearance, physics, visual text, and reference conditions.

The model-agnostic intermediate representation (IR)
\(z=\operatorname{Parse}(x)\) retains the parsed requirements \(\mathcal R(x)\), original prompt \(p\), visual assets \(A\), and initial mode \(m_0\).

A deployment \(d\) specifies the frozen checkpoint, inference configuration, and interface for each supported mode.
Compilation consults the deployment-specific capability manual \(\mathcal M_d\) and intervention catalog \(\mathcal O_d\), which defines text transformations, external operators, and their preconditions.
The resulting plan \(\Pi\) is executed to generate video \(v\):
\begin{equation}
    \Pi = \mathcal C_d(z;\mathcal M_d,\mathcal O_d),
    \qquad v = \operatorname{Execute}_d(\Pi).
    \label{eq:compilation_execution}
\end{equation}
The plan specifies prompts, visual conditions, external operations, and per-call modes, which may differ from \(m_0\) (e.g., T2V to I2V after first-frame construction).
The goal is deployment-compatible execution that preserves the original semantic requirements.

\subsection{Capability Manual}
\label{sec:capability_manual}

A capability manual \(\mathcal M_d\) records the empirical support of deployment \(d\) for predefined controls, with entries indexed by control type \(c\), difficulty level \(\ell\), and conditioning mode \(m\).
For each tested entry, successful and scored trial counts \(s_d(c,\ell,m)\) and \(n_d(c,\ell,m)\) yield the empirical success rate
\begin{equation}
    \hat p_d(c,\ell,m)
    = \frac{s_d(c,\ell,m)}{n_d(c,\ell,m)},
    \qquad n_d(c,\ell,m)>0.
    \label{eq:empirical_capability}
\end{equation}

The labels ok, weak, and bad indicate high, intermediate, and low empirical native success on the tested probes, respectively. Unknown denotes the absence of scored trials and is not evidence
of either reliability or failure. The manual characterizes native generation under the tested prompts and inputs; interventions and their preconditions are defined separately in \(\mathcal O_d\). Support for individual controls does not guarantee their joint satisfaction in compositional requests.

\section{Method}
\label{sec:method}

As shown in Figure~\ref{fig:method}, NeurCo constructs a capability manual for each deployment of a frozen video model and uses it to compile user requests into executable generation plans.
The process comprises offline capability profiling, requirement-level compilation, and backend execution.

\subsection{Offline Capability Profiling}
\label{sec:offline_profiling}

We systematically probe the shared semantic-control vocabulary at different difficulty levels under the applicable T2V, I2V, and R2V modes.
Each probe isolates a target requirement while varying the scene and subject context, and supplies the prompt and visual assets required by the mode.
A video judge, supplemented by task-specific measurements where applicable, evaluates whether the generated video successfully satisfies the target requirement.

For each \((c,\ell,m)\), we retain successful and scored trial counts and compute the empirical success rate in Eq.~\eqref{eq:empirical_capability}.
These measurements determine an ok, weak, or bad support status; entries without scored evidence are unknown.
The assignment thresholds are given in the supplementary material.
The manual records native generation performance, while the intervention catalog \(\mathcal O_d\) specifies the transformations available to the compiler.

\subsection{Capability-Guided Compilation}
\label{sec:request_compilation}

Given \(x=(p,A,m_0)\), NeurCo constructs the IR described in Sec.~\ref{sec:task_formulation}.
The parser extracts explicitly stated requirements, normalizes their control types and difficulty levels, and links them to verbatim spans in the original prompt.
The full description and user-provided assets are retained.
For VComBench, task annotations supply this IR, isolating compilation and execution from parser errors.

For each requirement \(r_i\), NeurCo retrieves the native support status \(\sigma_i\) for \((c_i,\ell_i,m_0)\).
Let \(\mathcal F_i^\lambda\) contain the remedies for that requirement permitted by configuration \(\lambda\) whose deployment, interface, structural, and input-asset preconditions are satisfied.
The decision is
\begin{equation}
    a_i=
    \begin{cases}
        \text{keep},
        & \sigma_i\in\{\text{ok},\text{unknown}\},\\
        \text{intervene},
        & \sigma_i\in\{\text{weak},\text{bad}\}
          \land \mathcal F_i^\lambda\neq\varnothing,\\
        \text{unresolved},
        & \sigma_i\in\{\text{weak},\text{bad}\}
          \land \mathcal F_i^\lambda=\varnothing.
    \end{cases}
    \label{eq:routing_decision}
\end{equation}

The intervention configuration \(\lambda\) determines which transformations and operators are available. The compiler assembles compatible interventions into an execution plan, while unresolved requirements retain their original wording. Native support is assessed in the initial mode \(m_0\), but execution may switch modes when the destination interface and conditioning requirements are satisfied.





\subsection{Backends and Execution}
\label{sec:backends_execution}

Targeted rewriting clarifies subject binding, counts, event order, camera behavior, spatial relations, or timing while preserving other stated requirements, including those sharing the edited text.

External operators provide alternative ways to express the selected controls. Reference images anchor appearance or attribute binding; first frames encode counts or visual text; layout images specify spatial placement; and proxy videos provide motion, timing, or camera guidance through compatible interfaces. Multi-shot requests can be handled within a single call when supported by the backend, or through shot concatenation when independent clips are permitted. Requests requiring a continuous shot are never split. The six operators are detailed in Appendix~\ref{sec:app_operator_catalog}.

Before execution, generated visual conditions are validated against their intended requirements. Invalid conditions are discarded, and the remaining text plan is retained. The executor then assembles the prompts, conditioning assets, and inference parameters required by the selected backend. When compilation leaves the executable inputs unchanged, the corresponding baseline output is reused if available. All video-generation checkpoints remain frozen throughout the process.

\section{VComBench}
\label{sec:vcombench}

VComBench evaluates whether multiple semantic requirements can be
satisfied within the same generated video.
It measures both individual requirement satisfaction and joint task
success, allowing us to systematically assess whether compilation improves targeted
requirements while preserving the remaining constraints.

\begin{figure}[ht]
    \centering
    \begin{minipage}[t]{0.485\textwidth}
        \vspace{0pt}\centering
        \includegraphics[width=\linewidth]{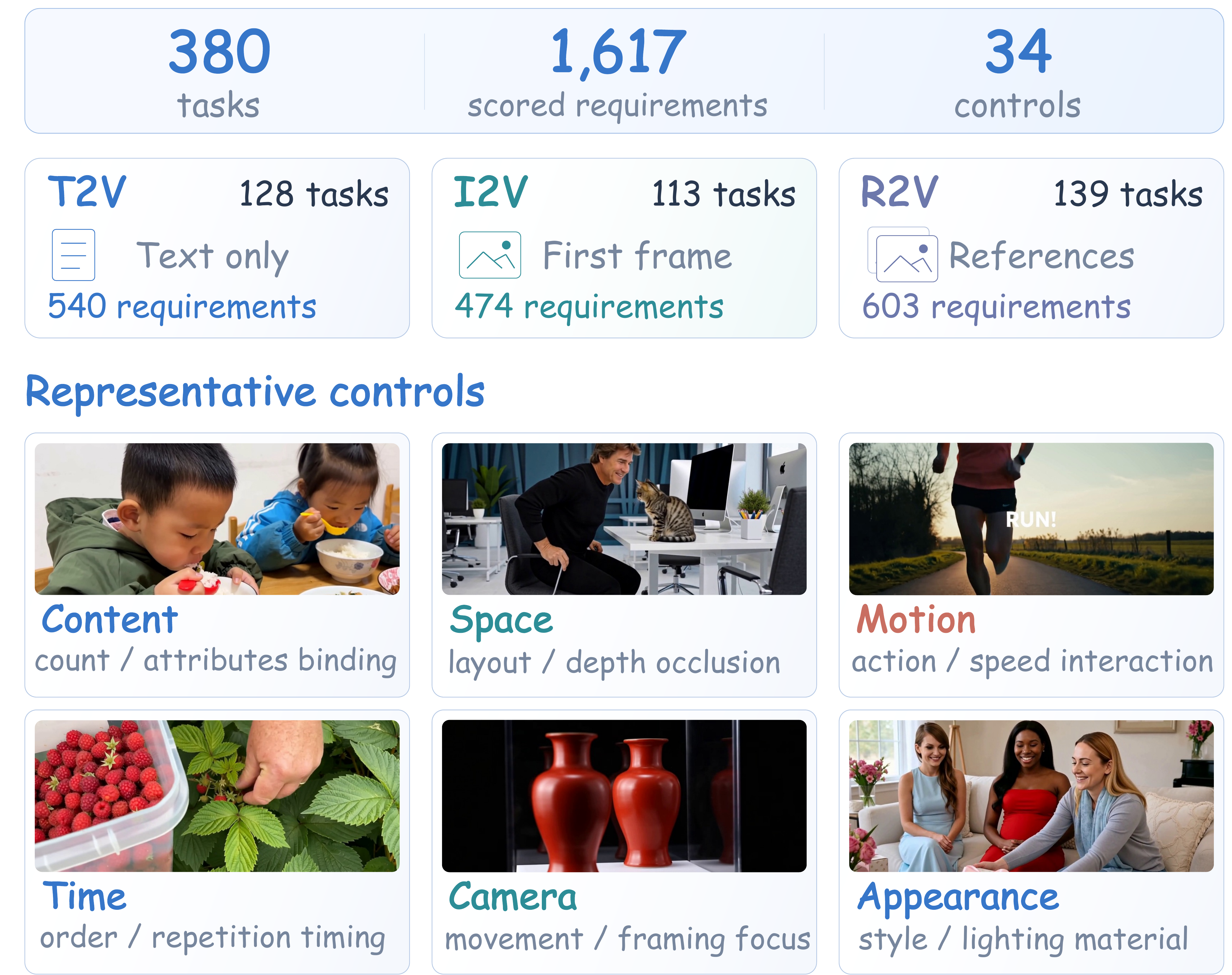}
        \captionsetup{width=\linewidth}
        \caption{\textbf{VComBench composition and representative controls.}
    Task and scored-requirement counts are shown by input mode.
    Scene thumbnails illustrate representative control families.}
        \label{fig:vcom_composition}
    \end{minipage}\hfill
    \begin{minipage}[t]{0.485\textwidth}
        \vspace{0pt}\centering
        \includegraphics[width=\linewidth]{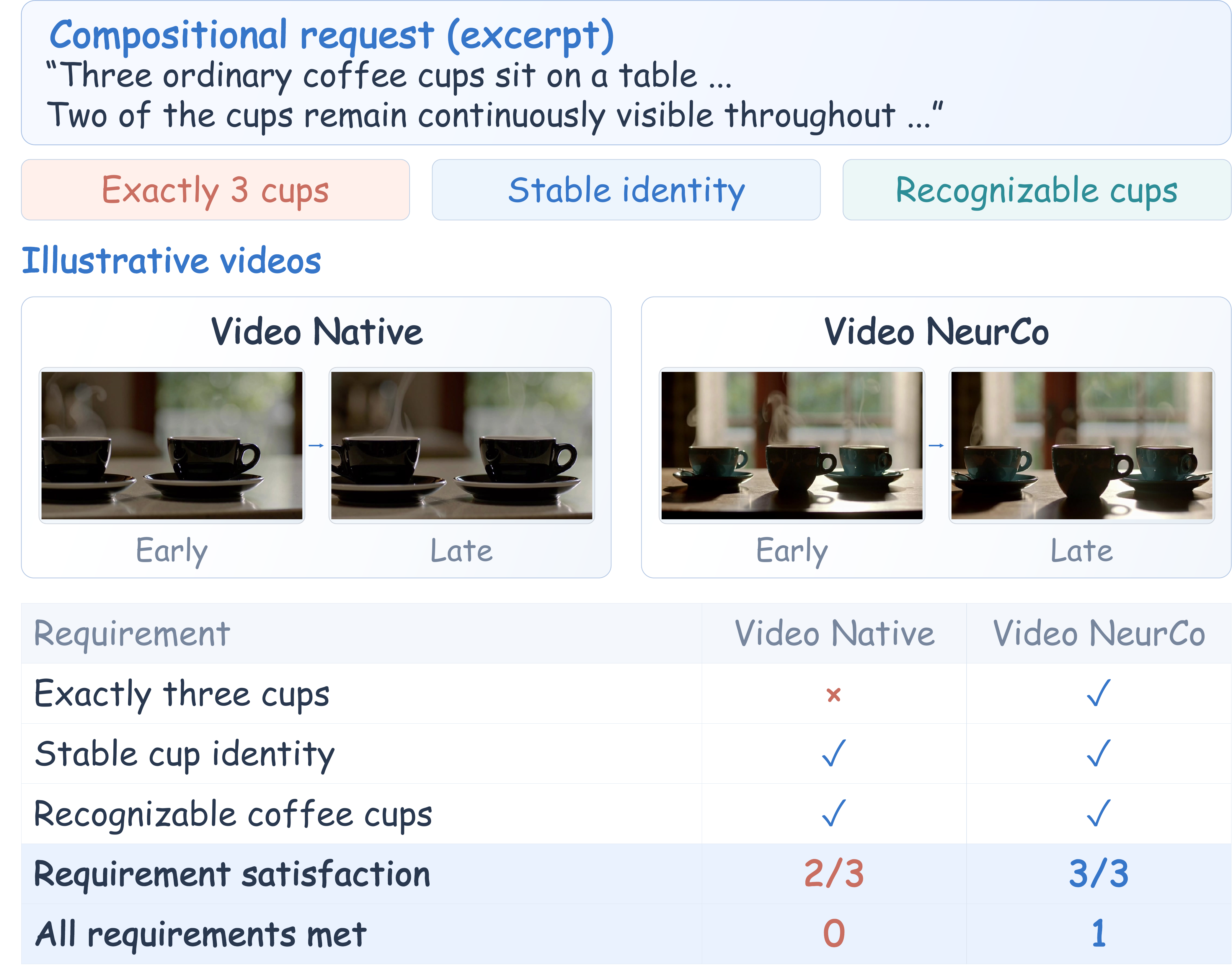}
        \captionsetup{width=\linewidth}
        \caption{\textbf{Requirement satisfaction and joint task success.}
    Video Native satisfies two of three requirements but fails the task;
    Video NeurCo satisfies all three and succeeds.}
        \label{fig:vcom_scoring}
    \end{minipage}
\end{figure}

\subsection{Task Construction}
\label{sec:vcom_task_construction}

VComBench contains 380 compositional tasks adapted from public video
benchmarks covering general video quality, image-to-video generation,
prompt enhancement, compositional generation, camera control, and
physical commonsense
\cite{Huang_2024_CVPR,huang2024vbenchcomprehensiveversatilebenchmark,
shao2026unifiedevaluationpromptenhancers,Sun_2026_CVPR,
Liu_2024_CVPR,bansal2024videophyevaluatingphysicalcommonsense}.
Each task specifies a prompt, optional visual assets, and annotated
requirements with associated checks.
The same task specifications are used across compiler variants.

Tasks span T2V, I2V, and R2V, providing text only, a first-frame image,
and reference visual inputs, respectively.
These modes describe the task's input contract; execution may use a
different conditioning mode after compilation.
Figure~\ref{fig:vcom_composition} summarizes the task and requirement
counts by input mode and illustrates representative semantic controls.

The tasks contain 1,617 scored requirements covering object properties
and counts, spatial relations, actions and interactions, temporal
order, camera behavior, appearance, physical effects, visual text,
and reference conditions.
We consolidate 50 raw annotation categories into 34 semantic-control
dimensions shared with the capability manual.
Input-validity checks are used for data validation and excluded from
scoring, which measures requirements that the generated video must
visibly realize.

Each task contains either three or six scored requirements:
221 tasks contain three, and 159 contain six.
As described in Sec.~\ref{sec:request_compilation}, the main experiments
use annotation-derived IRs to isolate compilation and execution from
parser errors.
Further construction and annotation details are provided in the
supplementary material.


\subsection{Scoring Protocol}
\label{sec:vcom_scoring}

A fixed video-judging pipeline evaluates each scored requirement,
which passes only when all associated checks pass.
Temporal, repetition, and multi-step camera requirements use denser
frame sampling.

Let $\mathcal R_q$ denote the scored requirements of task $q$, and $y_{q,r}\in\{0,1\}$ indicate whether requirement $r$ passes. Let $\mathcal E$ denote the evaluated task--requirement pairs and $\mathcal Q_{\mathrm{complete}}$ the tasks with complete judgments. Requirement satisfaction is defined as:
\begin{equation}
    S_{\mathrm{req}}
    =
    \frac{1}{|\mathcal E|}
    \sum_{(q,r)\in\mathcal E} y_{q,r}.
    \label{eq:vcom_requirement_score}
\end{equation}
Joint task success requires all scored requirements to pass:
\begin{equation}
    S_{\mathrm{all}}
    =
    \frac{1}{|\mathcal Q_{\mathrm{complete}}|}
    \sum_{q\in\mathcal Q_{\mathrm{complete}}}
    \mathbf{1}\!\left[
        \sum_{r\in\mathcal R_q}y_{q,r}=|\mathcal R_q|
    \right].
    \label{eq:vcom_all_score}
\end{equation}
As illustrated in Figure~\ref{fig:vcom_scoring}, satisfying only two of three requirements yields partial requirement satisfaction but does not count as a successful task.

We additionally report a realism rate $S_{\mathrm{real}}$, defined as
the fraction of evaluated videos judged photorealistic among the
366 tasks expected to produce photorealistic footage.
The remaining 14 tasks explicitly request non-photorealistic styles
and are excluded from this metric.

\section{Experiments}
\label{sec:experiments}


\subsection{Experimental Setup}
\label{sec:experimental_setup}


We evaluate frozen local deployments of MiniMax-H3~\cite{minimax2026h3} and Wan~\cite{wan2025}, and a hosted Seedance 2.5~\cite{seedance2026} model. Experiments cover the supported T2V, I2V, and R2V modes, with VACE used for Wan reference conditioning. For VComBench, we construct IRs from annotated task requirements to isolate compilation and execution from parsing errors. Three NeurCo configurations are used: targeted text rewriting (T), external operators without targeted rewriting (O), and their combination (T+O). We compare \emph{Native} (original inputs), \emph{NeurCo (T)}, \emph{NeurCo (T+O)}, TIDE~\cite{zhou2026tidetrajectoryinformedskilldistillation}, and TIDE combined with each NeurCo configuration. Generation seeds are shared across methods where supported. TIDE implementation details are provided in Appendix~\ref{sec:app_tide_prompt}.
External evaluation uses a VBench-2.0~\cite{zheng2025vbench20advancingvideogeneration} subset and the 150-task Sci-VBench testmini~\cite{zhang2026scivbenchevaluatingknowledgereasoningintensive}. We report the unweighted mean over 17 valid dimensions for VBench-2.0, and PG, SCC, SC, VT, and their mean for Sci-VBench. Detailed VBench-2.0 scores are provided in Table~\ref{tab:vbench2_dimensions} in the supplementary material.

\subsection{Main Results on VComBench}
\label{sec:vcom_main_results}


Table~\ref{tab:vcom_main} compares compilation with native generation and prompt enhancement. On native inputs, NeurCo (T+O) improves requirement satisfaction by 2.8 points on H3 and 5.5 on Wan, exceeding the 1.2 and 1.1 point gains from NeurCo (T). Wan also improves all-requirements success from 8.4\% to 11.8\%. On Seedance 2.5, NeurCo (T) raises requirement satisfaction from 68.5\% to 69.7\%, while NeurCo (T+O) reaches 68.2\%. For requirement satisfaction, combined compilation performs better on H3 and Wan, while text-only compilation performs better on Seedance 2.5.


\begin{table}[ht]
    \centering
    \small
    \setlength{\tabcolsep}{3.4pt}
    \renewcommand{\arraystretch}{1.08}
    \caption{VComBench results (\%). Req./All denote requirement
    satisfaction/joint success. Best and second-best values per block
    and backend are bold and underlined. Evaluated task counts
    are reported in Table~\ref{tab:vcom_coverage}.}
    \label{tab:vcom_main}
    \begin{tabular}{l*{6}{r}}
    \toprule
    & \multicolumn{2}{c}{H3}
    & \multicolumn{2}{c}{Wan}
    & \multicolumn{2}{c}{Seedance 2.5} \\
    \cmidrule(lr){2-3}\cmidrule(lr){4-5}\cmidrule(lr){6-7}
    Method & Req. & All & Req. & All & Req. & All \\
    \midrule
    Native
    & 66.5 & \textbf{28.2}
    & 46.8 & \underline{8.4}
    & \underline{68.5} & \underline{26.9} \\
    \rowcolor{bestblue}
    NeurCo (T)
    & \underline{67.7} & \underline{27.9}
    & \underline{47.9} & \underline{8.4}
    & \textbf{69.7} & \textbf{27.5} \\
    \rowcolor{bestblue}
    NeurCo (T+O)
    & \textbf{69.3} & 25.3
    & \textbf{52.3} & \textbf{11.8}
    & 68.2 & 25.8 \\
    \midrule
    TIDE
    & \underline{72.5} & 33.2
    & \underline{50.6} & \underline{12.7}
    & 71.3 & 31.1 \\
    \rowcolor{bestblue}
    TIDE + NeurCo (O)
    & \textbf{73.0} & \underline{33.7}
    & \textbf{52.5} & \textbf{14.5}
    & \textbf{72.1} & \textbf{32.2} \\
    \rowcolor{bestblue}
    TIDE + NeurCo (T)
    & 72.4 & 32.7
    & 46.3 & 9.0
    & 71.7 & \underline{32.1} \\
    \rowcolor{bestblue}
    TIDE + NeurCo (T+O)
    & 72.0 & \textbf{34.5}
    & 50.0 & 12.3
    & \underline{71.8} & 31.5 \\
    \bottomrule
    \end{tabular}
\end{table}

Wan improves both metrics, whereas H3's higher requirement satisfaction coincides with lower all-requirements success (Sec.~\ref{sec:capability_intervention_analysis}). NeurCo also complements prompt enhancement. Operator-only compilation improves TIDE's requirement satisfaction by 0.5, 1.9, and 0.8 points on H3, Wan, and Seedance 2.5, respectively, and raises all-requirements success on all three. The combined variant achieves H3's highest all-requirements success (34.5\%), despite lower requirement satisfaction than TIDE, and improves both metrics over TIDE on Seedance 2.5. On Wan, it recovers much of the text-only loss but remains below TIDE. Additional rewriting does not consistently improve on operator-only compilation.

\subsection{Mode and Quality Analysis}
\label{sec:capability_intervention_analysis}


Table~\ref{tab:vcom_modes} analyzes native-input compilation by initial input mode. On H3, NeurCo (T+O) increases requirement satisfaction in all three modes, although R2V all-requirements success drops from 33.1\% to 28.8\%. On Wan, it gives larger requirement gains on T2V and R2V, while NeurCo (T) performs better on I2V. On Seedance 2.5, combined compilation raises R2V requirement satisfaction from 66.1\% to 69.9\% and all-requirements success from 20.7\% to 28.9\%, but I2V requirement satisfaction falls. The benefits therefore depend on both the backend and the input mode.

\begin{figure}[ht]
    \centering
    \begin{minipage}[t]{0.485\textwidth}
        \vspace{0pt}\centering
        \includegraphics[width=\linewidth]{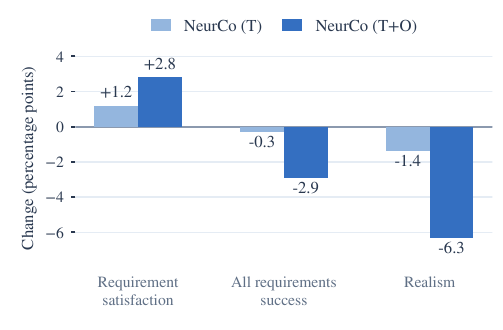}
        \captionsetup{width=\linewidth}
        \caption{Changes from Native on H3 (percentage points). Native rates: requirement satisfaction 66.5\%, all-requirements success 28.2\%, and realism 74.3\%.}
        \label{fig:control_realism_tradeoff}
    \end{minipage}\hfill
    \begin{minipage}[t]{0.485\textwidth}
        \vspace{0pt}\centering
        \includegraphics[width=\linewidth]{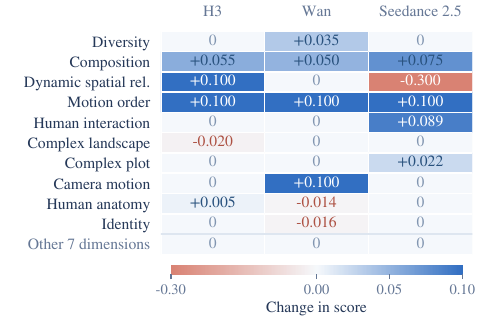}
        \captionsetup{width=\linewidth}
        \caption{VBench-2.0 dimension-wise changes from TIDE to
TIDE + NeurCo (O) on H3, Wan and Seedance2.5. Values are absolute score differences,
rounded to three decimals. Absolute dimension scores are reported
in Table~\ref{tab:vbench2_dimensions}.}
        \label{fig:vbench2_dimensions}
    \end{minipage}
\end{figure}

\begin{table}[ht]
    \centering
    \small
    \setlength{\tabcolsep}{3.5pt}
    \renewcommand{\arraystretch}{1.08}
    \caption{Native-input VComBench results by initial mode (\%).
    Req./All follow Table~\ref{tab:vcom_main}. Best and second-best
    values per mode and backend are bold and underlined, including
    ties. Seedance task counts are in Table~\ref{tab:seedance_modes}.}
    \label{tab:vcom_modes}
    \begin{tabular}{ll*{6}{r}}
    \toprule
    & & \multicolumn{2}{c}{H3} & \multicolumn{2}{c}{Wan} & \multicolumn{2}{c}{Seedance 2.5} \\
    \cmidrule(lr){3-4}\cmidrule(lr){5-6}\cmidrule(lr){7-8}
    Mode & Method & Req. & All & Req. & All & Req. & All \\
    \midrule
    T2V & Native & 66.5 & \underline{26.6} & 54.3 & 14.1 & 67.6 & \textbf{25.8} \\
    \rowcolor{bestblue}
     & NeurCo (T) & \underline{68.1} & \textbf{27.3} & \underline{56.4} & \underline{15.0} & \textbf{69.0} & \underline{25.2} \\
    \rowcolor{bestblue}
     & NeurCo (T+O) & \textbf{70.9} & 25.8 & \textbf{62.9} & \textbf{20.0} & \underline{68.2} & 24.4 \\
    \midrule
    I2V & Native & \underline{62.4} & \textbf{23.9} & 44.7 & \underline{6.2} & \underline{71.7} & \textbf{33.3} \\
    \rowcolor{bestblue}
     & NeurCo (T) & 62.2 & \underline{20.4} & \textbf{46.0} & 5.3 & \textbf{72.2} & \underline{31.4} \\
    \rowcolor{bestblue}
     & NeurCo (T+O) & \textbf{63.9} & \underline{20.4} & \underline{45.1} & \textbf{8.8} & 66.9 & 25.0 \\
    \midrule
    R2V & Native & 69.8 & \underline{33.1} & 41.6 & \underline{5.0} & 66.1 & 20.7 \\
    \rowcolor{bestblue}
     & NeurCo (T) & \underline{71.6} & \textbf{34.5} & \underline{41.8} & \underline{5.0} & \underline{67.8} & \underline{25.9} \\
    \rowcolor{bestblue}
     & NeurCo (T+O) & \textbf{72.1} & 28.8 & \textbf{49.1} & \textbf{7.2} & \textbf{69.9} & \textbf{28.9} \\
    \bottomrule
    \end{tabular}
\end{table}

Figure~\ref{fig:control_realism_tradeoff} shows that NeurCo (T+O) improves fine-grained
requirement satisfaction on H3 by 2.8 percentage points,
satisfying 45 additional requirements. However, joint
task success decreases by 2.9 points and realism by
6.3 points. The realism effect differs across backends:
it remains stable on Wan (77.5\% to 77.8\%) and improves
on Seedance~2.5 (68.3\% to 71.0\%). These results reveal
backend-dependent trade-offs between semantic control,
joint task success, and visual quality.



\begin{table*}[t]
    \centering
    \small
    \setlength{\tabcolsep}{3.4pt}
    \renewcommand{\arraystretch}{1.08}
    \caption{Sci-VBench testmini results (1--5). Avg. averages PG, SCC, SC, and VT. Best and second-best values within each block and backend are bold and underlined, including ties.}
    \label{tab:sci_results}
    \begin{tabular}{l*{15}{r}}
    \toprule
    & \multicolumn{5}{c}{H3}
    & \multicolumn{5}{c}{Wan}
    & \multicolumn{5}{c}{Seedance 2.5} \\
    \cmidrule(lr){2-6}\cmidrule(lr){7-11}\cmidrule(lr){12-16}
    Method
    & PG & SCC & SC & VT & Avg.
    & PG & SCC & SC & VT & Avg.
    & PG & SCC & SC & VT & Avg. \\
    \midrule
    Native
    & 2.340 & 1.527 & \underline{2.747} & \underline{4.056} & 2.668
    & 2.193 & 1.407 & \underline{2.513} & \textbf{4.132} & 2.561
    & 2.840 & 2.307 & \underline{2.907} & \textbf{3.913} & 2.992 \\
    \rowcolor{bestblue}
    NeurCo (T)
    & \textbf{2.413} & \underline{1.593} & 2.740 & \textbf{4.072} & \underline{2.705}
    & \underline{2.260} & \underline{1.427} & 2.473 & \underline{4.100} & \underline{2.565}
    & \underline{2.880} & \underline{2.360} & \textbf{2.913} & 3.891 & \underline{3.011} \\
    \rowcolor{bestblue}
    NeurCo (T+O)
    & \underline{2.400} & \textbf{1.620} & \textbf{2.820} & \underline{4.056} & \textbf{2.724}
    & \textbf{2.273} & \textbf{1.467} & \textbf{2.560} & 4.080 & \textbf{2.595}
    & \textbf{2.927} & \textbf{2.380} & 2.893 & \underline{3.902} & \textbf{3.025} \\
    \midrule
    TIDE
    & 3.213 & 2.253 & 3.293 & \underline{4.080} & 3.210
    & 2.740 & \underline{1.920} & 2.833 & \textbf{4.116} & 2.902
    & 3.207 & 2.353 & 3.233 & 3.932 & 3.181 \\
    \rowcolor{bestblue}
    TIDE + NeurCo (O)
    & \textbf{3.307} & \underline{2.320} & \textbf{3.347} & 4.072 & \textbf{3.262}
    & \textbf{2.780} & \textbf{1.953} & \underline{2.927} & \underline{4.108} & \textbf{2.942}
    & \underline{3.227} & 2.353 & 3.247 & 3.936 & 3.191 \\
    \rowcolor{bestblue}
    TIDE + NeurCo (T)
    & \underline{3.257} & 2.209 & 3.230 & \textbf{4.087} & 3.196
    & 2.736 & 1.804 & 2.865 & \underline{4.108} & 2.878
    & \textbf{3.287} & \underline{2.400} & \textbf{3.440} & \underline{3.944} & \textbf{3.268} \\
    \rowcolor{bestblue}
    TIDE + NeurCo (T+O)
    & 3.211 & \textbf{2.327} & \underline{3.313} & \underline{4.080} & \underline{3.233}
    & \underline{2.760} & 1.887 & \textbf{2.993} & 4.054 & \underline{2.923}
    & 3.160 & \textbf{2.493} & \underline{3.407} & \textbf{3.959} & \underline{3.255} \\
    \bottomrule
    \end{tabular}
\end{table*}

\begin{table}[t]
    \centering
    \small
    \setlength{\tabcolsep}{7pt}
    \renewcommand{\arraystretch}{1.08}
    \caption{VBench-2.0 subset results. Mean$_{17}$ averages the
    17 valid dimensions. Best and second-best values per backend
    are bold and underlined. Detailed scores are in
    Table~\ref{tab:vbench2_dimensions}.}
    \label{tab:vbench2_results}
    \begin{tabular}{lrrr}
    \toprule
    Method & H3 & Wan & Seedance 2.5 \\
    \midrule
    Native & 0.5597 & 0.5281 & 0.6520 \\
    \midrule
    TIDE
    & \underline{0.6609}
    & \underline{0.6660}
    & \underline{0.6759} \\
    \rowcolor{bestblue}
    TIDE + NeurCo (O)
    & \textbf{0.6750}
    & \textbf{0.6810}
    & 0.6750 \\
    \rowcolor{bestblue}
    TIDE + NeurCo (T)
    & 0.6566 & 0.6113 & 0.6634 \\
    \rowcolor{bestblue}
    TIDE + NeurCo (T+O)
    & 0.6593 & 0.6538 & \textbf{0.6963} \\
    \bottomrule
    \end{tabular}
\end{table}

\subsection{External Benchmark Results}
\label{sec:external_results}

On VBench-2.0 (Table~\ref{tab:vbench2_results}), text-only compilation improves
Wan's native-input Mean$_{17}$ from 0.5281 to 0.6275
(Table~13). Adding operators to TIDE improves H3 and Wan,
whereas the combined strategy performs best on Seedance~2.5
(0.6963 vs.\ 0.6759 for TIDE). These results show that
the most effective compilation strategy varies across backends.

Figure~\ref{fig:vbench2_dimensions} shows that operator-only compilation improves composition (+0.055/+0.050) and motion order (+0.100 on both) on H3 and Wan. H3 also gains in dynamic spatial relations, while Wan improves camera motion. The improvements are concentrated in specific controls, with smaller regressions in other dimensions.

Sci-VBench tests whether the improvements extend beyond VComBench (Table~\ref{tab:sci_results}). On native inputs, NeurCo (T+O) raises the mean from 2.668 to 2.724 on H3, from 2.561 to 2.595 on Wan, and from 2.992 to 3.025 on Seedance 2.5. After TIDE, operator-only compilation reaches 3.262 on H3 and 2.942 on Wan, while the combined variant reaches 3.233 and 2.923. On Seedance 2.5, the text-only and combined variants reach 3.268 and 3.255, respectively, compared with 3.181 for TIDE and 3.191 for the operator-only variant. The relative benefits of rewriting and operators therefore vary across backends.

\begin{figure*}[t]
    \centering
    \begingroup
    \small

    \definecolor{paintBlue}{HTML}{2863A6}
    \definecolor{paintFail}{HTML}{A84C46}

    \setlength{\tabcolsep}{0pt}
    \setlength{\abovecaptionskip}{2pt}
    \setlength{\belowcaptionskip}{0pt}
    \renewcommand{\arraystretch}{1.0}

    \newlength{\paintContentW}
    \newlength{\paintLabelW}
    \newlength{\paintGap}
    \newlength{\paintFrameW}
    \newlength{\paintFrameH}

    \setlength{\paintContentW}{\textwidth}
    \setlength{\paintLabelW}{.18\textwidth}
    \setlength{\paintFrameW}{.21\textwidth}
    \setlength{\paintFrameH}{.5625\paintFrameW}

    \setlength{\paintGap}{%
        \dimexpr(\paintContentW-\paintLabelW-3\paintFrameW)/3\relax}

    \newsavebox{\paintFrameBox}
    \newcommand{\caseframe}[1]{%
        \begingroup
        \sbox{\paintFrameBox}{%
            \includegraphics[width=\paintFrameW]{#1}%
        }%
        \begin{tikzpicture}[
            baseline={([yshift=-.5ex]current bounding box.center)}
        ]
            \path[use as bounding box]
                (0,0) rectangle
                (\paintFrameW,\paintFrameH);

            \begin{scope}[overlay]
                \clip (0,0) rectangle
                    (\paintFrameW,\paintFrameH);
                \node[inner sep=0,outer sep=0]
                    at (.5\paintFrameW,.5\paintFrameH) {%
                    \ifdim\ht\paintFrameBox<\paintFrameH\relax
                        \includegraphics[
                            height=\paintFrameH
                        ]{#1}%
                    \else
                        \usebox{\paintFrameBox}%
                    \fi
                };
            \end{scope}
        \end{tikzpicture}%
        \endgroup
    }

    \setlength{\fboxsep}{3pt}
    \setlength{\fboxrule}{.4pt}

    \fcolorbox{paintBlue!25}{paintBlue!4}{%
        \begin{minipage}{%
            \dimexpr\paintContentW-2\fboxsep-2\fboxrule\relax}
            \footnotesize
            \raggedright

            \textbf{Input (T2V):}
            Red, purple, and yellow cups rapidly pour equal amounts
            of matching paint into a bowl; a spoon then stirs quickly
            until the colored streaks merge into a uniform color.

            \par\vspace{2pt}
            \textbf{Scored requirements}
            \par\vspace{1pt}

            \begin{tabular*}{\linewidth}
                {@{\extracolsep{\fill}}lll@{}}
                (1) Cup--paint color binding
                & (2) Recognizable everyday cups
                & (3) Pouring before stirring
                \\[1pt]

                \textcolor{paintBlue}{%
                    \textbf{(4) Paint intermixing}}
                & (5) Rapid pouring and stirring
                & \textcolor{paintBlue}{%
                    \textbf{(6) Streaks $\rightarrow$ uniform color}}
            \end{tabular*}
        \end{minipage}%
    }
    \par\vspace{2pt}

    \begin{tabular}{%
        @{}l@{\hspace{\paintGap}}
        c@{\hspace{\paintGap}}
        c@{\hspace{\paintGap}}c@{}}

        \textbf{Method}
        & \textbf{Early}
        & \textbf{Middle}
        & \textbf{Late}
        \\[1pt]

        \parbox[c]{\paintLabelW}{\raggedright\small
            \textbf{Native}\\[1pt]
            \textbf{4/6 passed}\\[1pt]
            \textcolor{paintFail}{Fails (4), (6)}}
        & \caseframe{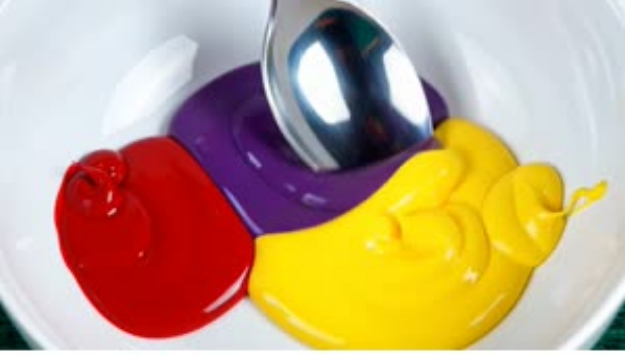}
        & \caseframe{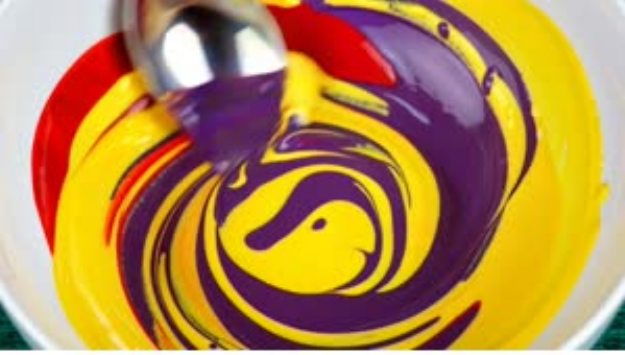}
        & \caseframe{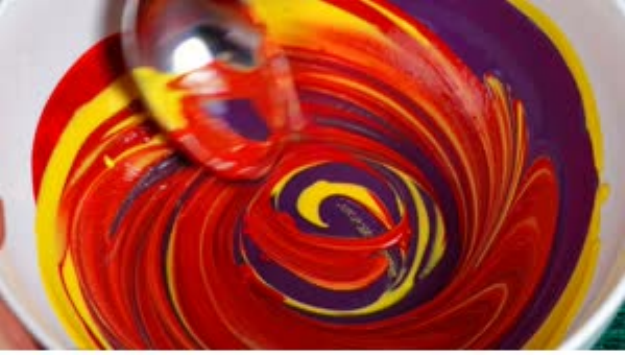}
        \\
        \noalign{\vskip 1pt}

        \parbox[c]{\paintLabelW}{\raggedright\small
            \textbf{NeurCo (T+O)}\\[1pt]
            \textcolor{paintBlue}{\textbf{6/6 passed}}\\[1pt]
            First-frame I2V}
        & \caseframe{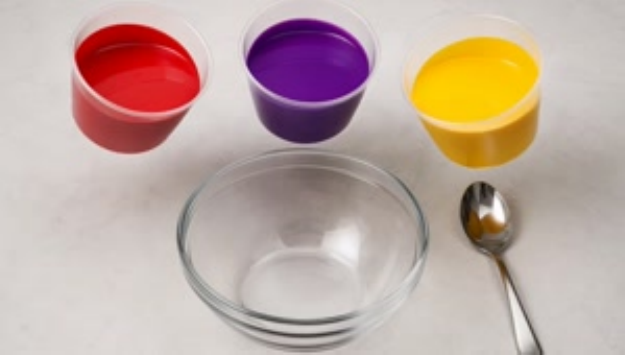}
        & \caseframe{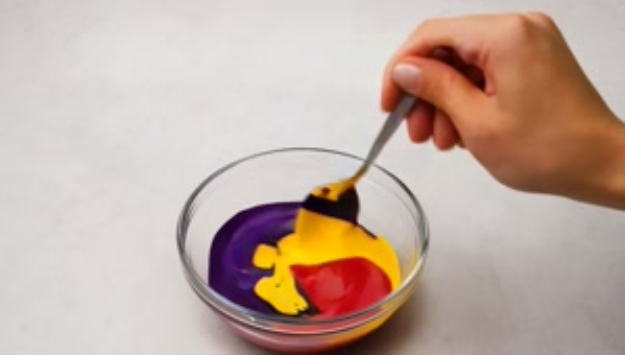}
        & \caseframe{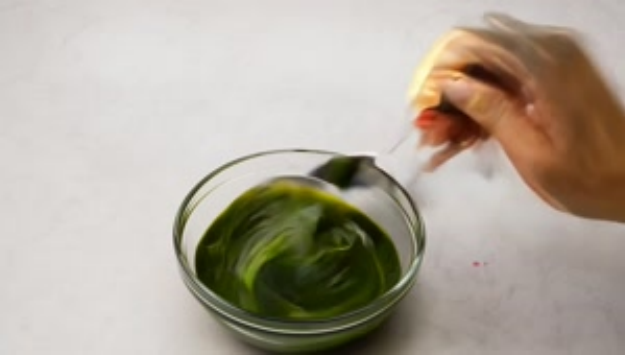}
        \\
        \noalign{\vskip 1pt}

        \parbox[c]{\paintLabelW}{\raggedright\small
            \textbf{TIDE}\\[1pt]
            \textbf{4/6 passed}\\[1pt]
            \textcolor{paintFail}{Fails (4), (6)}}
        & \caseframe{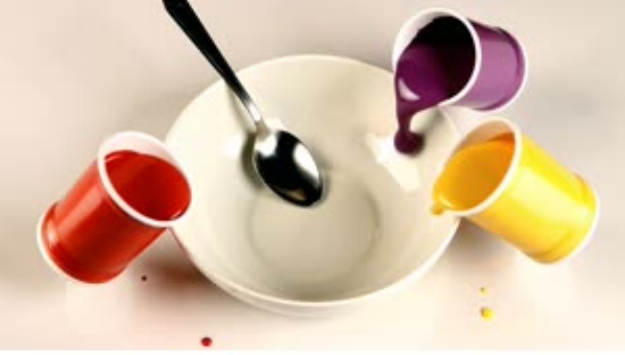}
        & \caseframe{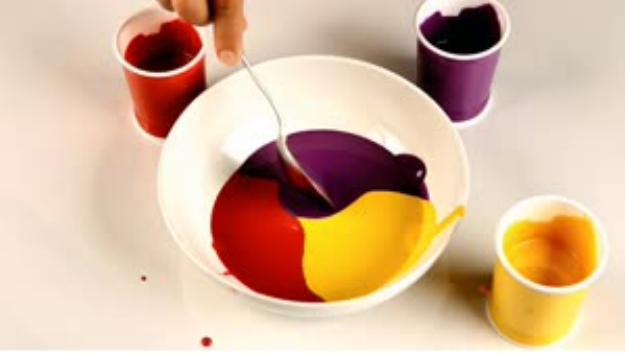}
        & \caseframe{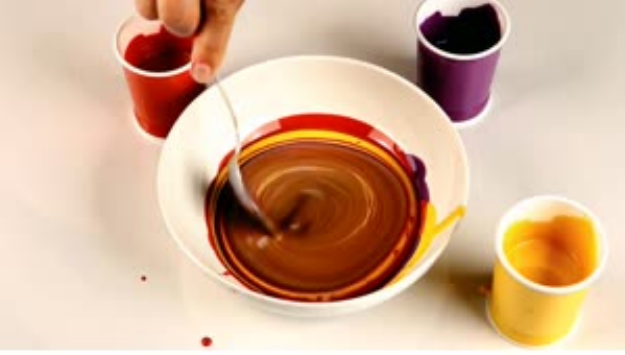}
        \\
        \noalign{\vskip 1pt}

        \parbox[c]{\paintLabelW}{\raggedright\small
            \textbf{TIDE + NeurCo (O)}\\[1pt]
            \textcolor{paintBlue}{\textbf{6/6 passed}}\\[1pt]
            Proxy control\\via VACE}
        & \caseframe{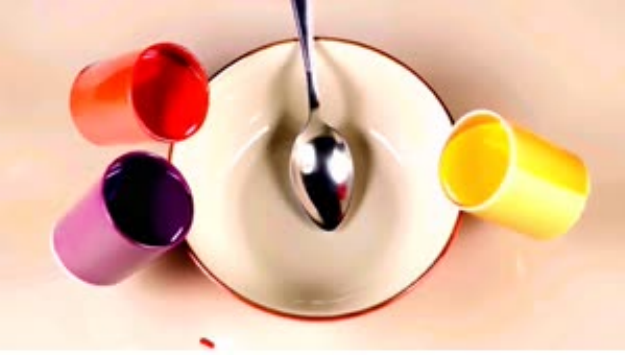}
        & \caseframe{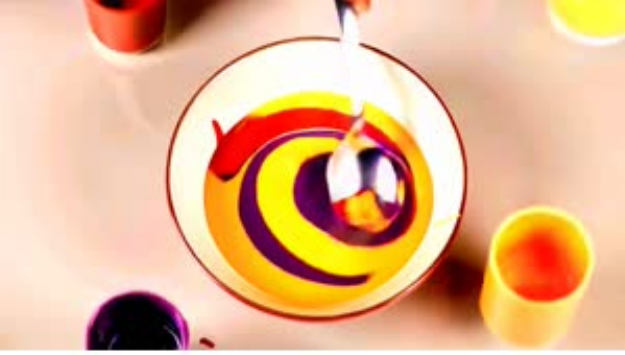}
        & \caseframe{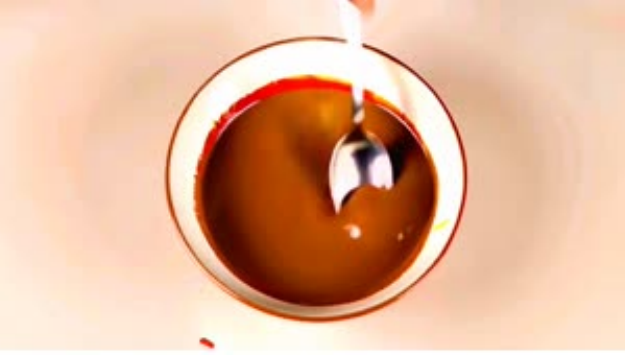}
    \end{tabular}
    \vspace{2mm}
    \caption{\textbf{Paint-mixing case study on Wan.}
    Native and TIDE fail requirements (4) and (6), while NeurCo (T+O)
    and TIDE + NeurCo (O) satisfy all six in the illustrated runs.
    Scores count requirements passed in full-video evaluation;
    the columns show three sampled times.}
    \label{fig:case_paint}

    \endgroup
\end{figure*}

\subsection{Multi-Backend Routing}
\label{sec:joint_routing}

Table~\ref{tab:joint_routing} reports routing over H3+Wan and H3+Wan+Seedance 2.5. The three-backend runs obtain higher VComBench requirement satisfaction under both raw and TIDE-enhanced inputs, reaching 69.7\% and 74.3\%, respectively. With TIDE, the Sci-VBench mean also rises from 3.128 to 3.214, while all-requirements success decreases from 33.8\% to 33.2\% and VBench-2.0 Mean$_{17}$ decreases from 0.6749 to 0.6673. The gains are therefore not uniform across metrics. Evaluated task counts are reported in Table~\ref{tab:routing_coverage}.

\begin{table}[t]
    \centering
    \small
    \setlength{\tabcolsep}{2.5pt}
    \caption{Multi-backend routing results. Req./All are VComBench
    requirement satisfaction/joint success (\%). Best and second-best
    values for each prompt setting are bold and underlined.}
    \label{tab:joint_routing}
    \begin{tabular}{llrrrr}
    \toprule
    Backend pool & Prompt & Req. & All & Sci Avg. & Mean$_{17}$ \\
    \midrule
    \rowcolor{bestblue}
    H3 + Wan & Raw & \underline{68.0} & \underline{27.9} & \underline{2.721} & \underline{0.5984} \\
    \rowcolor{bestblue}
             & TIDE & \underline{72.7} & \textbf{33.8} & \underline{3.128} & \textbf{0.6749} \\
    \midrule
    \rowcolor{bestblue}
    H3 + Wan + Seedance 2.5 & Raw & \textbf{69.7} & \textbf{32.6} & \textbf{2.766} & \textbf{0.6214} \\
    \rowcolor{bestblue}
                            & TIDE & \textbf{74.3} & \underline{33.2} & \textbf{3.214} & \underline{0.6673} \\
    \bottomrule
    \end{tabular}
\end{table}

\subsection{Case Study}
\label{sec:case_study}

Figure~\ref{fig:case_paint} illustrates two complementary
conditioning strategies for paint mixing on Wan. The task
requires three colored cups to pour matching paints into
a bowl, followed by stirring until the colored
streaks form a uniform mixture. Native and TIDE satisfy
four of six requirements, both failing to reproduce paint
intermixing and the final transition to a uniform color.
NeurCo (T+O) constructs a first frame and switches from
T2V to I2V to ground the scene, while TIDE + NeurCo (O)
uses a proxy video through VACE to guide the mixing
process. Both strategies satisfy all six requirements,
capturing the temporal progression, whereas
NeurCo (T) satisfies only three. Additional examples
and failure cases are provided in
Appendix~\ref{sec:case_study_appendix}.

\section{Conclusion}
\label{sec:conclusion}

We presented NeurCo, which treats frozen video models
as generative rendering backends and discovers their
effective control capabilities through empirical profiling.
Deployment-specific capability manuals guide the compilation
of user intent into execution strategies that preserve reliable
requirements and adapt challenging ones. Experiments on
VComBench and two external benchmarks demonstrate improved
semantic control across three backends, alongside trade-offs
in joint task success and photorealism. Beyond individual
renderers, these capability manuals provide a foundation
for capability-aware tool selection and orchestration.

\clearpage
\section*{LLM Usage Statement}

OpenAI Codex was used to assist with manuscript preparation,
including drafting and revising the conclusion, adapting the
existing \LaTeX{} project to the CVPR template, formatting
tables and figures, and resolving \LaTeX{} compatibility issues.

{\small
\bibliographystyle{ieeenat_fullname}
\bibliography{main}

\begin{thebibliography}{56}
\providecommand{\natexlab}[1]{#1}
\providecommand{\url}[1]{\texttt{#1}}
\expandafter\ifx\csname urlstyle\endcsname\relax
  \providecommand{\doi}[1]{doi: #1}\else
  \providecommand{\doi}{doi: \begingroup \urlstyle{rm}\Url}\fi

\bibitem[Bansal et~al.(2025)Bansal, Lin, Xie, Zong, Yarom, Bitton, Jiang, Sun,
  Chang, and Grover]{bansal2024videophyevaluatingphysicalcommonsense}
Hritik Bansal, Zongyu Lin, Tianyi Xie, Zeshun Zong, Michal Yarom, Yonatan
  Bitton, Chenfanfu Jiang, Yizhou Sun, Kai-Wei Chang, and Aditya Grover.
\newblock Videophy: Evaluating physical commonsense for video generation.
\newblock In \emph{International Conference on Learning Representations}, pages
  102075--102121, 2025.

\bibitem[{ByteDance Seed}(2026)]{seedance2026}
{ByteDance Seed}.
\newblock {Seedance 2.5}.
\newblock \url{https://seed.bytedance.com/en/seedance2_5}, 2026.
\newblock Accessed: November 8, 2026.

\bibitem[Cai et~al.(2024)Cai, Ceylan, Gadelha, Huang, Wang, and
  Wetzstein]{Cai_2024_CVPR}
Shengqu Cai, Duygu Ceylan, Matheus Gadelha, Chun-Hao~Paul Huang, Tuanfeng~Yang
  Wang, and Gordon Wetzstein.
\newblock Generative rendering: Controllable 4d-guided video generation with 2d
  diffusion models.
\newblock In \emph{2024 IEEE/CVF Conference on Computer Vision and Pattern
  Recognition (CVPR)}, pages 7611--7620. IEEE, 2024.

\bibitem[Chen et~al.(2025)Chen, Siarohin, Menapace, Fang, Lee, Skorokhodov,
  Aberman, Zhu, Yang, and Tulyakov]{Chen_2025_CVPR}
Tsai-Shien Chen, Aliaksandr Siarohin, Willi Menapace, Yuwei Fang, Kwot~Sin Lee,
  Ivan Skorokhodov, Kfir Aberman, Jun-Yan Zhu, Ming-Hsuan Yang, and Sergey
  Tulyakov.
\newblock Multi-subject open-set personalization in video generation.
\newblock In \emph{2025 IEEE/CVF Conference on Computer Vision and Pattern
  Recognition (CVPR)}, pages 6099--6110. IEEE, 2025.

\bibitem[Chen et~al.(2023)Chen, Ji, Wu, Wu, Xie, Li, Xia, Xiao, and
  Lin]{chen2024controlavideocontrollabletexttovideodiffusion}
Weifeng Chen, Yatai Ji, Jie Wu, Hefeng Wu, Pan Xie, Jiashi Li, Xin Xia, Xuefeng
  Xiao, and Liang Lin.
\newblock Control-a-video: Controllable text-to-video diffusion models with
  motion prior and reward feedback learning.
\newblock \emph{arXiv preprint arXiv:2305.13840}, 2023.

\bibitem[Cheng et~al.(2025)Cheng, Lyu, Gu, Liu, Xu, Lu, Teng, Yang, Dong, Tang,
  et~al.]{Cheng_2025_ICCV}
Jiale Cheng, Ruiliang Lyu, Xiaotao Gu, Xiao Liu, Jiazheng Xu, Yida Lu, Jiayan
  Teng, Zhuoyi Yang, Yuxiao Dong, Jie Tang, et~al.
\newblock Vpo: Aligning text-to-video generation models with prompt
  optimization.
\newblock In \emph{2025 IEEE/CVF International Conference on Computer Vision
  (ICCV)}, pages 15636--15645. IEEE, 2025.

\bibitem[Chi et~al.(2026)Chi, Miao, Shi, Wu, Wang, Chen, Dai, Ren, Gao, Long,
  et~al.]{chi2026agentgartencodeworldsevolving}
Jiawei Chi, Shangchen Miao, Zhiyuan Shi, Kailu Wu, Hanyang Wang, Weiliang Chen,
  Qiyu Dai, Jinshan Ren, Jun Gao, Mingsheng Long, et~al.
\newblock Agentgarten: Code worlds for evolving agents.
\newblock \emph{arXiv preprint arXiv:2610.12374}, 2026.

\bibitem[Feng et~al.(2025)Feng, Li, Saxon, Fu, Chen, and
  Wang]{feng-etal-2025-tc}
Weixi Feng, Jiachen Li, Michael Saxon, Tsu-Jui Fu, Wenhu Chen, and William~Yang
  Wang.
\newblock Tc-bench: Benchmarking temporal compositionality in conditional video
  generation.
\newblock In \emph{Findings of the Association for Computational Linguistics:
  ACL 2025}, pages 4638--4662, 2025.

\bibitem[Feng et~al.(2026)Feng, Wang, Xu, Qian, Wang, Hou, Liu, Sun, Liu, and
  Wang]{feng2026newtonagenticplanningphysically}
Yuxiang Feng, Juncheng Wang, Chao Xu, Yijie Qian, Huihan Wang, Wenlong Hou,
  Yang Liu, Baigui Sun, Yong Liu, and Shujun Wang.
\newblock Newton: Agentic planning for physically grounded video generation.
\newblock \emph{arXiv preprint arXiv:2605.18396}, 2026.

\bibitem[Gao et~al.(2025{\natexlab{a}})Gao, Gao, Wu, Zhou, Qiao, Niu, Chen, and
  Wang]{Gao_2025_CVPR}
Bingjie Gao, Xinyu Gao, Xiaoxue Wu, Yujie Zhou, Yu Qiao, Li Niu, Xinyuan Chen,
  and Yaohui Wang.
\newblock The devil is in the prompts: Retrieval-augmented prompt optimization
  for text-to-video generation.
\newblock In \emph{2025 IEEE/CVF Conference on Computer Vision and Pattern
  Recognition (CVPR)}, pages 3173--3183. IEEE, 2025{\natexlab{a}}.

\bibitem[Gao et~al.(2025{\natexlab{b}})Gao, Ma, Wu, Yang, Lan, Zhao, Chen, Liu,
  Qiao, Chen, et~al.]{gao2026rapocrossstagepromptoptimization}
Bingjie Gao, Qianli Ma, Xiaoxue Wu, Shuai Yang, Guanzhou Lan, Haonan Zhao,
  Jiaxuan Chen, Qingyang Liu, Yu Qiao, Xinyuan Chen, et~al.
\newblock Rapo++: Cross-stage prompt optimization for text-to-video generation
  via data alignment and test-time scaling.
\newblock \emph{arXiv preprint arXiv:2510.20206}, 2025{\natexlab{b}}.

\bibitem[Geng et~al.(2025)Geng, Herrmann, Hur, Cole, Zhang, Pfaff,
  Lopez-Guevara, Aytar, Rubinstein, Sun, et~al.]{motion_prompting}
Daniel Geng, Charles Herrmann, Junhwa Hur, Forrester Cole, Serena Zhang, Tobias
  Pfaff, Tatiana Lopez-Guevara, Yusuf Aytar, Michael Rubinstein, Chen Sun,
  et~al.
\newblock Motion prompting: Controlling video generation with motion
  trajectories.
\newblock In \emph{2025 IEEE/CVF Conference on Computer Vision and Pattern
  Recognition (CVPR)}, pages 1--12. IEEE, 2025.

\bibitem[Guan et~al.(2025)Guan, Lai, Sun, Zhang, Liu, Liu, Cao, and
  Song]{Guan_2025_ICCV}
Kaisi Guan, Zhengfeng Lai, Yuchong Sun, Peng Zhang, Wei Liu, Kieran Liu, Meng
  Cao, and Ruihua Song.
\newblock Etva: Evaluation of text-to-video alignment via fine-grained question
  generation and answering.
\newblock In \emph{2025 IEEE/CVF International Conference on Computer Vision
  (ICCV)}, pages 21299--21309. IEEE, 2025.

\bibitem[He et~al.(2024)He, Xu, Guo, Wetzstein, Dai, Li, and
  Yang]{he2025cameractrlenablingcameracontrol}
Hao He, Yinghao Xu, Yuwei Guo, Gordon Wetzstein, Bo Dai, Hongsheng Li, and
  Ceyuan Yang.
\newblock Cameractrl: Enabling camera control for text-to-video generation.
\newblock \emph{arXiv preprint arXiv:2404.02101}, 2024.

\bibitem[Huang et~al.(2024)Huang, He, Yu, Zhang, Si, Jiang, Zhang, Wu, Jin,
  Chanpaisit, et~al.]{Huang_2024_CVPR}
Ziqi Huang, Yinan He, Jiashuo Yu, Fan Zhang, Chenyang Si, Yuming Jiang, Yuanhan
  Zhang, Tianxing Wu, Qingyang Jin, Nattapol Chanpaisit, et~al.
\newblock Vbench: Comprehensive benchmark suite for video generative models.
\newblock In \emph{2024 IEEE/CVF Conference on Computer Vision and Pattern
  Recognition (CVPR)}, pages 21807--21818. IEEE, 2024.

\bibitem[Huang et~al.(2025)Huang, Zhang, Xu, He, Yu, Dong, Ma, Chanpaisit, Si,
  Jiang, et~al.]{huang2024vbenchcomprehensiveversatilebenchmark}
Ziqi Huang, Fan Zhang, Xiaojie Xu, Yinan He, Jiashuo Yu, Ziyue Dong, Qianli Ma,
  Nattapol Chanpaisit, Chenyang Si, Yuming Jiang, et~al.
\newblock Vbench++: Comprehensive and versatile benchmark suite for video
  generative models.
\newblock \emph{IEEE Transactions on Pattern Analysis and Machine
  Intelligence}, 2025.

\bibitem[Ji et~al.(2025)Ji, Zhang, Wu, Zhang, Chen, Ge, Sun, Chen, Shao, Xiao,
  et~al.]{Ji_2025_ICCV}
Yatai Ji, Jiacheng Zhang, Jie Wu, Shilong Zhang, Shoufa Chen, Chongjian Ge,
  Peize Sun, Weifeng Chen, Wenqi Shao, Xuefeng Xiao, et~al.
\newblock Prompt-a-video: Prompt your video diffusion model via
  preference-aligned llm.
\newblock In \emph{2025 IEEE/CVF International Conference on Computer Vision
  (ICCV)}, pages 18725--18735. IEEE, 2025.

\bibitem[Jia et~al.(2026)Jia, Hua, and
  Zhang]{jia2026capet2vcaptioneranchoredpromptenhancement}
Yizhuo Jia, Jingyun Hua, and Yuanxing Zhang.
\newblock Cape-t2v: Captioner-anchored prompt enhancement toward two-sided
  conditioning alignment in text-to-video generation.
\newblock \emph{arXiv preprint arXiv:2608.03046}, 2026.

\bibitem[Jiang et~al.(2024)Jiang, Wu, Yang, Si, Lin, Qiao, Loy, and
  Liu]{Jiang_2024_CVPR}
Yuming Jiang, Tianxing Wu, Shuai Yang, Chenyang Si, Dahua Lin, Yu Qiao,
  Chen~Change Loy, and Ziwei Liu.
\newblock Videobooth: Diffusion-based video generation with image prompts.
\newblock In \emph{2024 IEEE/CVF Conference on Computer Vision and Pattern
  Recognition (CVPR)}, pages 6689--6700. IEEE, 2024.

\bibitem[Jiang et~al.(2025)Jiang, Han, Mao, Zhang, Pan, and
  Liu]{Jiang_2025_ICCV}
Zeyinzi Jiang, Zhen Han, Chaojie Mao, Jingfeng Zhang, Yulin Pan, and Yu Liu.
\newblock Vace: All-in-one video creation and editing.
\newblock In \emph{2025 IEEE/CVF International Conference on Computer Vision
  (ICCV)}, pages 17191--17202. IEEE, 2025.

\bibitem[Kizil et~al.(2026)Kizil, Sanli, Mitra, Erdem, Erdem, and
  Ceylan]{Kizil_2026_CVPR}
Muhammed~Burak Kizil, Enes Sanli, Niloy~J Mitra, Erkut Erdem, Aykut Erdem, and
  Duygu Ceylan.
\newblock Lamp: language-assisted motion planning for controllable video
  generation.
\newblock In \emph{Proceedings of the IEEE/CVF Conference on Computer Vision
  and Pattern Recognition}, pages 12829--12838, 2026.

\bibitem[Li et~al.(2026)Li, Ren, Ma, Qing, Fang, He, Guo, Wu, Tian, Zou,
  et~al.]{li2026videocococodeascotphysicallyconsistentvideo}
Haodong Li, Tianfei Ren, Xiaoxiao Ma, Chunmei Qing, Zhen Fang, Sipeng He, Ziyu
  Guo, Haoyu Wu, Juanxi Tian, Yihang Zou, et~al.
\newblock Videococo: Code-as-cot for physically-consistent video generation via
  an agentic dual-engine system.
\newblock \emph{arXiv preprint arXiv:2607.27380}, 2026.

\bibitem[Lian et~al.(2024)Lian, Shi, Yala, Li,
  et~al.]{lian2024llmgroundedvideodiffusionmodels}
Long Lian, Baifeng Shi, Adam Yala, Boyi Li, et~al.
\newblock Llm-grounded video diffusion models.
\newblock In \emph{International Conference on Learning Representations}, pages
  50207--50227, 2024.

\bibitem[Liang et~al.(2025)Liang, Gojcic, Ling, Munkberg, Hasselgren, Lin, Gao,
  Keller, Vijaykumar, Fidler, et~al.]{Liang_2025_CVPR}
Ruofan Liang, Zan Gojcic, Huan Ling, Jacob Munkberg, Jon Hasselgren, Chih-Hao
  Lin, Jun Gao, Alexander Keller, Nandita Vijaykumar, Sanja Fidler, et~al.
\newblock Diffusionrenderer: Neural inverse and forward rendering with video
  diffusion models.
\newblock In \emph{2025 IEEE/CVF Conference on Computer Vision and Pattern
  Recognition (CVPR)}, pages 26069--26080. IEEE, 2025.

\bibitem[Liao et~al.(2025)Liao, Zeng, Wang, Yu, Lin, and Zhang]{Motionagent}
Xinyao Liao, Xianfang Zeng, Liao Wang, Gang Yu, Guosheng Lin, and Chi Zhang.
\newblock Motionagent: Fine-grained controllable video generation via motion
  field agent.
\newblock In \emph{2025 IEEE/CVF International Conference on Computer Vision
  (ICCV)}, pages 11305--11316. IEEE, 2025.

\bibitem[Lin et~al.(2023)Lin, Zala, Cho, and Bansal]{Lin2023VideoDirectorGPT}
Han Lin, Abhay Zala, Jaemin Cho, and Mohit Bansal.
\newblock Videodirectorgpt: Consistent multi-scene video generation via
  llm-guided planning.
\newblock \emph{arXiv preprint arXiv:2309.15091}, 2023.

\bibitem[Lin et~al.(2026)Lin, Akbari, He, Zhao, Zhang, Akbari, Xu, Lu, Nan,
  Deng, et~al.]{lin2026phygroundbenchmarkingphysicalreasoning}
Juyi Lin, Arash Akbari, Yumei He, Lin Zhao, Haichao Zhang, Arman Akbari,
  Xingchen Xu, Zoe~Y Lu, Enfu Nan, Hokin Deng, et~al.
\newblock Phyground: Benchmarking physical reasoning in generative world
  models.
\newblock \emph{arXiv preprint arXiv:2605.10806}, 2026.

\bibitem[Liu et~al.(2023)Liu, Li, Ren, Gao, Li, Chen, Sun, and
  Hou]{NEURIPS2023_c481049f}
Yuanxin Liu, Lei Li, Shuhuai Ren, Rundong Gao, Shicheng Li, Sishuo Chen, Xu
  Sun, and Lu Hou.
\newblock Fetv: A benchmark for fine-grained evaluation of open-domain
  text-to-video generation.
\newblock \emph{Advances in Neural Information Processing Systems},
  36:\penalty0 62352--62387, 2023.

\bibitem[Liu et~al.(2024)Liu, Cun, Liu, Wang, Zhang, Chen, Liu, Zeng, Chan, and
  Shan]{Liu_2024_CVPR}
Yaofang Liu, Xiaodong Cun, Xuebo Liu, Xintao Wang, Yong Zhang, Haoxin Chen,
  Yang Liu, Tieyong Zeng, Raymond Chan, and Ying Shan.
\newblock Evalcrafter: Benchmarking and evaluating large video generation
  models.
\newblock In \emph{2024 IEEE/CVF Conference on Computer Vision and Pattern
  Recognition (CVPR)}, pages 22139--22149. IEEE, 2024.

\bibitem[Long et~al.(2026)Long, Wan, Nakhost, Lee, Pfister, and
  Arik]{long2025vistatesttimeselfimprovingvideo}
Do~Xuan Long, Xingchen Wan, Hootan Nakhost, Chen-Yu Lee, Tomas Pfister, and
  Sercan~{\"O} Arik.
\newblock Vista: A test-time self-improving video generation agent.
\newblock In \emph{Proceedings of the IEEE/CVF conference on computer vision
  and pattern recognition}, pages 6021--6032, 2026.

\bibitem[Luo et~al.(2026)Luo, Liu, Wang, Zhang, Tao, Wan, Gai, and
  Fei]{luo2026evaluationenhancementbenchmarkingimproving}
Meng Luo, Yicheng Liu, Jiahao Wang, Yuanxing Zhang, Xin Tao, Pengfei Wan, Kun
  Gai, and Hao Fei.
\newblock From evaluation to enhancement: Benchmarking and improving
  think-with-video reasoning for video generative models.
\newblock In \emph{European Conference on Computer Vision}, pages 546--563.
  Springer, 2026.

\bibitem[{MiniMax}(2026)]{minimax2026h3}
{MiniMax}.
\newblock {MiniMax H3}: An open model breaking the boundaries between tasks and
  modalities.
\newblock \url{https://www.minimax.io/blog/minimax-h3}, 2026.

\bibitem[Ren et~al.(2024)Ren, Yang, Zhang, Wei, Du, Huang, and
  Chen]{ren2024consisti2venhancingvisualconsistency}
Weiming Ren, Huan Yang, Ge Zhang, Cong Wei, Xinrun Du, Wenhao Huang, and Wenhu
  Chen.
\newblock Consisti2v: Enhancing visual consistency for image-to-video
  generation.
\newblock \emph{arXiv preprint arXiv:2402.04324}, 2024.

\bibitem[Ren et~al.(2025)Ren, Shen, Huang, Ling, Lu, Nimier-David, M{\"u}ller,
  Keller, Fidler, and Gao]{Ren_2025_CVPR}
Xuanchi Ren, Tianchang Shen, Jiahui Huang, Huan Ling, Yifan Lu, Merlin
  Nimier-David, Thomas M{\"u}ller, Alexander Keller, Sanja Fidler, and Jun Gao.
\newblock Gen3c: 3d-informed world-consistent video generation with precise
  camera control.
\newblock In \emph{2025 IEEE/CVF Conference on Computer Vision and Pattern
  Recognition (CVPR)}, pages 6121--6132. IEEE, 2025.

\bibitem[Shah et~al.(2026)Shah, Susladkar, Prakash, Nguyen, Yu, Juvekar, Wahed,
  and Lourentzou]{shah2026graphvid}
Vedant Shah, Onkar Susladkar, Tushar Prakash, Kiet~A Nguyen, Tianjiao Yu,
  Adheesh Juvekar, Muntasir Wahed, and Ismini Lourentzou.
\newblock Graphvid: Interactive graph-controllable video generation.
\newblock In \emph{European Conference on Computer Vision}, pages 245--263.
  Springer, 2026.

\bibitem[Shao et~al.(2026)Shao, Zhu, Dai, Fang, Zhu, Jiang, Ai, Sun, Xue,
  Shang, et~al.]{shao2026unifiedevaluationpromptenhancers}
Yawen Shao, Yubo Zhu, Ziyun Dai, Zixun Fang, Kai Zhu, Zeyinzi Jiang, Yufeng Ai,
  Siyang Sun, Haolan Xue, Yu Shang, et~al.
\newblock Towards unified evaluation of prompt enhancers for video generation.
\newblock \emph{arXiv preprint arXiv:2610.11736}, 2026.

\bibitem[Sun et~al.(2026)Sun, Chen, Zhou, Li, Chen, Liang, and
  Bai]{Sun_2026_CVPR}
Zhengyang Sun, Yu Chen, Xin Zhou, Xiaofan Li, Xiwu Chen, Dingkang Liang, and
  Xiang Bai.
\newblock When numbers speak: Aligning textual numerals and visual instances in
  text-to-video diffusion models.
\newblock \emph{arXiv preprint arXiv:2604.08546}, 2026.

\bibitem[Team et~al.(2026)Team, Liu, Chen, Li, Chi, Sun, Li, Fu, Guo, Wu,
  et~al.]{team2026bernini}
Bernini Team, Chenchen Liu, Junyi Chen, Lei Li, Lu Chi, Mingzhen Sun, Zhuoying
  Li, Yi Fu, Ruoyu Guo, Yiheng Wu, et~al.
\newblock Bernini: Latent semantic planning for video diffusion.
\newblock \emph{arXiv preprint arXiv:2605.22344}, 2026.

\bibitem[Unterthiner et~al.(2018)Unterthiner, Van~Steenkiste, Kurach, Marinier,
  Michalski, and Gelly]{unterthiner2019accurategenerativemodelsvideo}
Thomas Unterthiner, Sjoerd Van~Steenkiste, Karol Kurach, Raphael Marinier,
  Marcin Michalski, and Sylvain Gelly.
\newblock Towards accurate generative models of video: A new metric \&
  challenges.
\newblock \emph{arXiv preprint arXiv:1812.01717}, 2018.

\bibitem[Wan et~al.(2025)Wan, Wang, Ai, Wen, Mao, Xie, Chen, Yu, Zhao, Yang,
  et~al.]{wan2025}
Team Wan, Ang Wang, Baole Ai, Bin Wen, Chaojie Mao, Chen-Wei Xie, Di Chen,
  Feiwu Yu, Haiming Zhao, Jianxiao Yang, et~al.
\newblock Wan: Open and advanced large-scale video generative models.
\newblock \emph{arXiv preprint arXiv:2503.20314}, 2025.

\bibitem[Wang et~al.(2025)Wang, He, Wang, Ma, Yang, Wang, Du, and
  Shen]{Wang_2025_CVPR}
Yiping Wang, Xuehai He, Kuan Wang, Luyao Ma, Jianwei Yang, Shuohang Wang,
  Simon~Shaolei Du, and Yelong Shen.
\newblock Is your world simulator a good story presenter? a consecutive
  events-based benchmark for future long video generation.
\newblock In \emph{2025 IEEE/CVF Conference on Computer Vision and Pattern
  Recognition (CVPR)}, pages 13629--13638. IEEE, 2025.

\bibitem[Wang et~al.(2024)Wang, Yuan, Wang, Li, Chen, Xia, Luo, and
  Shan]{Wang_2024}
Zhouxia Wang, Ziyang Yuan, Xintao Wang, Yaowei Li, Tianshui Chen, Menghan Xia,
  Ping Luo, and Ying Shan.
\newblock Motionctrl: A unified and flexible motion controller for video
  generation.
\newblock In \emph{ACM SIGGRAPH 2024 Conference Papers}, pages 1--11, 2024.

\bibitem[Wei et~al.(2026)Wei, Zhang, Tan, Zhu, Zhang, Chen, Zhang, Xu, Jin, and
  Liu]{wei2026videoweaverevaluatingevolvingskills}
Jianhui Wei, Yan Zhang, Jie Tan, Hengchuan Zhu, Xiaotian Zhang, Ziyi Chen,
  Daoan Zhang, Wei Xu, Yeying Jin, and Zuozhu Liu.
\newblock Videoweaver: Evaluating and evolving skills for agentic long video
  generation.
\newblock \emph{arXiv preprint arXiv:2606.08091}, 2026.

\bibitem[Xing et~al.(2024)Xing, Xia, Zhang, Chen, Yu, Liu, Liu, Wang, Shan, and
  Wong]{xing2023dynamicrafter}
Jinbo Xing, Menghan Xia, Yong Zhang, Haoxin Chen, Wangbo Yu, Hanyuan Liu,
  Gongye Liu, Xintao Wang, Ying Shan, and Tien-Tsin Wong.
\newblock Dynamicrafter: Animating open-domain images with video diffusion
  priors.
\newblock In \emph{European Conference on Computer Vision}, pages 399--417.
  Springer, 2024.

\bibitem[Yang et~al.(2026)Yang, Li, Qi, Zhou, Wu, and
  Liu]{yang2026scmaprselfcorrectingmultiagentprompt}
Chengyi Yang, Pengzhen Li, Jiayin Qi, Aimin Zhou, Ji Wu, and Ji Liu.
\newblock Scmapr: Self-correcting multi-agent prompt refinement for
  complex-scenario text-to-video generation.
\newblock In \emph{Findings of the Association for Computational Linguistics:
  ACL 2026}, pages 6942--6973, 2026.

\bibitem[Yang et~al.(2024)Yang, Hou, Huang, Ma, Wan, Zhang, Chen, and
  Liao]{Yang_2024}
Shiyuan Yang, Liang Hou, Haibin Huang, Chongyang Ma, Pengfei Wan, Di Zhang,
  Xiaodong Chen, and Jing Liao.
\newblock Direct-a-video: Customized video generation with user-directed camera
  movement and object motion.
\newblock In \emph{ACM SIGGRAPH 2024 Conference Papers}, pages 1--12, 2024.

\bibitem[Yang et~al.(2025)Yang, Teng, Zheng, Ding, Huang, Xu, Yang, Hong,
  Zhang, Feng, et~al.]{ICLR2025_ce31378e}
Zhuoyi Yang, Jiayan Teng, Wendi Zheng, Ming Ding, Shiyu Huang, Jiazheng Xu,
  Yuanming Yang, Wenyi Hong, Xiaohan Zhang, Guanyu Feng, et~al.
\newblock Cogvideox: Text-to-video diffusion models with an expert transformer.
\newblock In \emph{International Conference on Learning Representations}, pages
  83048--83077, 2025.

\bibitem[Yin et~al.(2023)Yin, Wu, Liang, Shi, Li, Ming, and
  Duan]{yin2023dragnuwafinegrainedcontrolvideo}
Shengming Yin, Chenfei Wu, Jian Liang, Jie Shi, Houqiang Li, Gong Ming, and Nan
  Duan.
\newblock Dragnuwa: Fine-grained control in video generation by integrating
  text, image, and trajectory.
\newblock \emph{arXiv preprint arXiv:2308.08089}, 2023.

\bibitem[Zhang et~al.(2026)Zhang, Song, Fu, Yang, and
  Zhao]{zhang2026scivbenchevaluatingknowledgereasoningintensive}
Diandian Zhang, Tingyu Song, Lin Fu, Zheyuan Yang, and Yilun Zhao.
\newblock Sci-vbench: Evaluating knowledge-and reasoning-intensive video
  generation in science domains.
\newblock \emph{arXiv preprint arXiv:2608.09873}, 2026.

\bibitem[Zhang et~al.(2023)Zhang, Wang, Zhang, Zhao, Yuan, Qin, Wang, Zhao, and
  Zhou]{zhang2023i2vgenxlhighqualityimagetovideosynthesis}
Shiwei Zhang, Jiayu Wang, Yingya Zhang, Kang Zhao, Hangjie Yuan, Zhiwu Qin,
  Xiang Wang, Deli Zhao, and Jingren Zhou.
\newblock I2vgen-xl: High-quality image-to-video synthesis via cascaded
  diffusion models.
\newblock \emph{arXiv preprint arXiv:2311.04145}, 2023.

\bibitem[Zhang et~al.(2025)Zhang, Li, Zhou, Wu, and Wu]{zhang2025scene}
Yunzhi Zhang, Zizhang Li, Matt Zhou, Shangzhe Wu, and Jiajun Wu.
\newblock The scene language: Representing scenes with programs, words, and
  embeddings.
\newblock In \emph{2025 IEEE/CVF Conference on Computer Vision and Pattern
  Recognition (CVPR)}, pages 24625--24634. IEEE, 2025.

\bibitem[Zhao and Shou(2026)]{Zhao_2026_CVPR}
Rui Zhao and Mike~Zheng Shou.
\newblock P-flow: Prompting visual effects generation.
\newblock \emph{arXiv preprint arXiv:2603.22091}, 2026.

\bibitem[Zheng et~al.(2025)Zheng, Huang, Liu, Zou, He, Zhang, Gu, Zhang, He,
  Zheng, et~al.]{zheng2025vbench20advancingvideogeneration}
Dian Zheng, Ziqi Huang, Hongbo Liu, Kai Zou, Yinan He, Fan Zhang, Lulu Gu,
  Yuanhan Zhang, Jingwen He, Wei-Shi Zheng, et~al.
\newblock Vbench-2.0: Advancing video generation benchmark suite for intrinsic
  faithfulness.
\newblock \emph{arXiv preprint arXiv:2503.21755}, 2025.

\bibitem[Zhou et~al.(2026{\natexlab{a}})Zhou, Huang, and
  Song]{zhou2026tidetrajectoryinformedskilldistillation}
Chengxing Zhou, Yuliang Huang, and Jingwei Song.
\newblock Tide: Trajectory-informed skill distillation for efficient
  text-to-video prompt optimization.
\newblock \emph{arXiv preprint arXiv:2610.15098}, 2026{\natexlab{a}}.

\bibitem[Zhou et~al.(2026{\natexlab{b}})Zhou, Ban, Feng, Pei, Cui, and
  Hsieh]{zhou2026farcodingagentsvisual}
Hengguang Zhou, Yuanhao Ban, Jiaqi Feng, Xiaohuan Pei, Justin Cui, and Cho-Jui
  Hsieh.
\newblock How far can coding agents go in visual generation?
\newblock \emph{arXiv preprint arXiv:2610.19025}, 2026{\natexlab{b}}.

\bibitem[Zhu et~al.(2026)Zhu, Shao, Dai, Fang, Zhu, Sun, Xue, Wang, Weng, Luo,
  et~al.]{zhu2026wanpecinematicpromptenhancement}
Yubo Zhu, Yawen Shao, Ziyun Dai, Zixun Fang, Kai Zhu, Siyang Sun, Haolan Xue,
  Chuxin Wang, Tingyu Weng, Jingming Luo, et~al.
\newblock Wanpe: Towards cinematic prompt enhancement for modern text-to-video
  generation.
\newblock \emph{arXiv preprint arXiv:2609.30221}, 2026.

\end{thebibliography}
}

\appendix






\onecolumn
\pagenumbering{arabic}
\setcounter{page}{1}
\makeatletter
\@ifundefined{c@cvprrulercount}{}{\setcounter{cvprrulercount}{1}}
\makeatother

\begin{center}
    \Large \textbf{\thetitle}\\
    \vspace{0.5em}Supplementary Material \\
    \vspace{1.0em}
\end{center}

\emergencystretch=2em

\renewcommand{\figurename}{Figure}

\definecolor{neurcoappblue}{RGB}{55,101,145}

\pdfbookmark[0]{Appendix Contents}{appendix.contents}
\section*{Appendix Contents}
\begingroup
\small
\linespread{1.16}\selectfont
\setlength{\parskip}{0pt}
\newcommand{\appcontentsmajor}[1]{%
    \par\addvspace{6pt}\noindent
    {\bfseries\hyperref[#1]{\ref*{#1}\quad\nameref*{#1}}}%
    \nobreak\dotfill\hyperref[#1]{\pageref*{#1}}\par}
\newcommand{\appcontentsminor}[1]{%
    \noindent\hspace*{1.5em}%
    \hyperref[#1]{\ref*{#1}\quad\nameref*{#1}}%
    \nobreak\dotfill\hyperref[#1]{\pageref*{#1}}\par}

\appcontentsmajor{sec:app_experimental_details}
\appcontentsminor{sec:app_generation_settings}
\appcontentsminor{sec:app_models}
\appcontentsminor{sec:app_evaluated_samples}
\appcontentsminor{sec:app_evaluation_details}
\appcontentsminor{sec:app_judge_protocol}
\appcontentsminor{sec:app_vbench2_subset}
\appcontentsminor{sec:app_sci_scoring}
\appcontentsminor{sec:app_budget}
\appcontentsminor{sec:app_vbench2_details}
\appcontentsmajor{sec:app_control_vocabulary}
\appcontentsmajor{sec:app_manual_format}
\appcontentsminor{sec:app_support_assignment}
\appcontentsminor{sec:app_manual_overview}
\appcontentsminor{sec:app_compilation_example}
\appcontentsmajor{sec:app_operator_catalog}
\appcontentsminor{sec:app_external_conditions}
\appcontentsminor{sec:app_condition_validation}
\appcontentsminor{sec:app_combining_strategies}
\appcontentsmajor{sec:case_study_appendix}
\appcontentsminor{sec:app_cases_proxy}
\appcontentsminor{sec:app_cases_text}
\appcontentsminor{sec:app_cases_visual}
\appcontentsminor{sec:app_cases_failures}
\appcontentsmajor{sec:app_neurco_templates}
\appcontentsminor{sec:app_neurco_prompt}
\appcontentsminor{sec:app_ref_image_prompts}
\appcontentsminor{sec:app_first_frame_prompts}
\appcontentsminor{sec:app_layout_frame_prompts}
\appcontentsminor{sec:app_proxy_shared_prompts}
\appcontentsminor{sec:app_shot_concat_prompts}
\appcontentsmajor{sec:app_tide_prompt}
\appcontentsminor{sec:app_tide_rule_modules}
\endgroup
\clearpage

\section{Additional Experimental Details}
\label{sec:app_experimental_details}

\subsection{Backend Deployments and Generation Settings}
\label{sec:app_generation_settings}

We evaluate MiniMax-H3 and three Wan2.2 variants with frozen weights
and BF16 precision.
Table~\ref{tab:app_generation_settings} summarizes their generation settings.
For I2V, the first-frame aspect ratio is preserved:
H3 uses a short side of 768 pixels, while Wan maintains approximately
the pixel budget of $1280\times720$.

\begin{table*}[ht]
    \centering
    \small
    \setlength{\tabcolsep}{5pt}
    \renewcommand{\arraystretch}{1.08}
    \caption{Local generation settings.
    Wan entries use A14B variants.
    Guidance pairs are ordered as (low-noise, high-noise);
    H3 shifts are video/audio.
    Resolutions are width $\times$ height.
    Durations are nominal.}
    \label{tab:app_generation_settings}
    \begin{tabular*}{\textwidth}{
        @{\extracolsep{\fill}}lcccc@{}
    }
        \toprule
        & MiniMax-H3 & Wan2.2 T2V & Wan2.2 I2V & Wan2.2 VACE-Fun \\
        \midrule
        Sampling steps
        & 50 & 40 & 40 & 50 \\

        Sampler
        & Flow matching & UniPC & UniPC & Flow-matching Euler \\

        Guidance scale
        & 1.0 & (3.0, 4.0) & (3.5, 3.5) & 5.0 \\

        Timestep shift
        & 12 / 3 & 12 & 5 & 12 \\

        Expert boundary
        & --- & 0.875 & 0.900 & 0.875 \\

        Resolution
        & $1344\times768$ & $1280\times720$
        & Adaptive & $720\times1280$ \\

        Frame rate
        & 24 fps & 16 fps & 16 fps & 16 fps \\

        Frames (5 s / 10 s)
        & 124 / 243 & 81 / 161 & 81 / 161 & 81 / 161 \\
        \bottomrule
    \end{tabular*}
\end{table*}
\paragraph{Conditioning.}
H3 uses first-frame conditioning for I2V and image or video references
for R2V.
When both are present, reference images precede the proxy video,
which is resampled to 24 fps.
Wan uses VACE for reference-image and proxy-depth conditioning,
with a control strength of 0.5 for both experts when a depth video
is supplied.
Depth-conditioned calls additionally use a photorealism instruction
and negative terms suppressing cartoon-like rendering.
Other Wan calls use the official default negative prompt.

\paragraph{Seedance 2.5.}
We evaluate a fixed version of the hosted Seedance~2.5 model at
720p and 24 fps, with audio and watermarking disabled. We access the model through the official API.
T2V and R2V use a 16:9 aspect ratio, while I2V follows the first frame.
Requested durations are 5 or 10 seconds.

\paragraph{Sampling protocol.}
VComBench uses nominal durations of 5 seconds, extended to 10 seconds
for shot-count or timed-event requirements.
Sci-VBench and VBench-2.0 use 5-second clips.
A deterministic task-specific seed is shared across methods within
each deployment.
For text-only comparisons, conditioning inputs and generation settings
are also held fixed.
Multiple samples use distinct task--sample seeds, while clips belonging
to the same concatenated task share a seed.

For VBench-2.0, native-input and raw-routing configurations use three
videos per task, and TIDE-based configurations use one.
Diversity evaluation uses 20 videos per prompt for all configurations.
When compilation leaves the generation inputs unchanged, the
corresponding baseline video is reused.

\subsection{Models Used by Each Component}
\label{sec:app_models}

Table~\ref{tab:app_models} summarizes the models specified for
compilation, condition construction, and evaluation.
For VComBench, we construct the IR directly from task annotations,
isolating compilation and execution from requirement-parsing errors.

\begin{table*}[ht]
    \centering
    \small
    \setlength{\tabcolsep}{6pt}
    \renewcommand{\arraystretch}{1.08}
    \caption{Models used by NeurCo and the benchmark evaluators.}
    \label{tab:app_models}
    \begin{tabular*}{\textwidth}{
        @{\extracolsep{\fill}}ll@{}
    }
        \toprule
        Component & Model \\
        \midrule
        Prompt rewriting
        & GPT-6.1 Sol \\

        Condition planning and validation
        & GPT-6.1 Sol \\

        Image generation: single-backend
        & ChatGPT Images 2.0 \\

        Image generation: multi-backend
        & GPT Image 1.5 \\

        VComBench evaluation
        & GPT-6.1 Sol \\

        Sci-VBench evaluation
        & Qwen3.5-397B-A17B \\

        VBench-2.0 evaluation
        & Official evaluation suite \\
        \bottomrule
    \end{tabular*}
\end{table*}

\paragraph{Compilation and condition construction.}
Targeted rewriting and TIDE enhancement use the same language model
with their respective instructions.
This model also plans reference images, layouts, proxy animations,
and shot decompositions, and validates the constructed visual conditions.
Backend and operator selection follow the capability manual and
operator eligibility rules without an additional model call.
The TIDE instructions are provided in
Appendix~\ref{sec:app_tide_prompt}.

\paragraph{Condition images.}
The image generator is configured as ChatGPT Images 2.0 for single-backend
experiments and GPT Image 1.5 for multi-backend experiments.
Images are generated at $1536\times1024$ resolution.
Layout first frames are obtained by editing rendered layout sketches.
Comparisons between single- and multi-backend configurations therefore
also reflect the difference in condition-image generators.

\paragraph{Evaluation settings.}
Language-model temperature and top-$p$ use provider defaults.
Reasoning effort is set to medium when explicitly configured.
VComBench evaluation requests specify the same judge model across
configurations; route-dependent settings and agreement checks are
reported in Sec.~\ref{sec:app_judge_protocol}.
Sci-VBench scoring follows Sec.~\ref{sec:app_sci_scoring},
and VBench-2.0 uses the official evaluation suite.

\subsection{Evaluated Samples}
\label{sec:app_evaluated_samples}

Some Seedance 2.5 requests in VComBench were rejected by the service's content moderation. Table~\ref{tab:vcom_coverage} reports completely judged tasks for each configuration. Table~\ref{tab:seedance_modes} gives the initial-mode counts used by the main mode analysis. Table~\ref{tab:routing_coverage} reports the evaluated task counts
for the multi-backend routing experiments.

\begin{table}[ht]
    \centering
    \small
    \setlength{\tabcolsep}{4pt}
    \renewcommand{\arraystretch}{1.08}
    \caption{Numbers of completely judged VComBench tasks out of 380.}
    \label{tab:vcom_coverage}
    \begin{tabularx}{\textwidth}{>{\raggedright\arraybackslash}X*{3}{>{\raggedleft\arraybackslash}X}}
    \toprule
    Method & H3 & Wan & Seedance 2.5 \\
    \midrule
    Native & 380 & 380 & 305 \\
    \rowcolor{bestblue}
    NeurCo (T) & 380 & 379 & 306 \\
    \rowcolor{bestblue}
    NeurCo (T+O) & 380 & 372 & 310 \\
    \midrule
    TIDE & 379 & 379 & 312 \\
    \rowcolor{bestblue}
    TIDE + NeurCo (O) & 380 & 379 & 307 \\
    \rowcolor{bestblue}
    TIDE + NeurCo (T) & 376 & 377 & 308 \\
    \rowcolor{bestblue}
    TIDE + NeurCo (T+O) & 377 & 373 & 317 \\
    \bottomrule
    \end{tabularx}
\end{table}

\begin{table}[!ht]
    \centering
    \small
    \setlength{\tabcolsep}{7pt}
    \caption{Completely judged Seedance 2.5 tasks by initial mode.
    VComBench contains 128 T2V, 113 I2V, and 139 R2V tasks.}
    \label{tab:seedance_modes}
    \begin{tabularx}{\textwidth}{>{\raggedright\arraybackslash}X*{4}{>{\raggedleft\arraybackslash}X}}
    \toprule
    Method & T2V & I2V & R2V & Total \\
    \midrule
    Native & 124 & 99 & 82 & 305 \\
    \rowcolor{bestblue}
    NeurCo (T) & 123 & 102 & 81 & 306 \\
    \rowcolor{bestblue}
    NeurCo (T+O) & 123 & 104 & 83 & 310 \\
    \bottomrule
    \end{tabularx}
\end{table}

\begin{table}[ht]
    \centering
    \small
    \setlength{\tabcolsep}{5pt}
    \caption{Numbers of judged routing tasks out of 380 on VComBench
    and 150 on Sci-VBench.}
    \label{tab:routing_coverage}
    \begin{tabularx}{\textwidth}{*{2}{>{\raggedright\arraybackslash}X}*{2}{>{\raggedleft\arraybackslash}X}}
    \toprule
    Backend pool & Prompt & VComBench & Sci-VBench \\
    \midrule
    H3 + Wan & Raw & 380 & 150 \\
             & TIDE & 380 & 142 \\
    \midrule
    H3 + Wan + Seedance 2.5 & Raw & 347 & 138 \\
                            & TIDE & 346 & 142 \\
    \bottomrule
    \end{tabularx}
\end{table}

\subsection{Evaluation Details}
\label{sec:app_evaluation_details}

For VBench-2.0, native-input and raw-routing configurations use three generated videos per task, whereas TIDE-based configurations use one. Diversity uses 20 videos in all configurations. Mean$_{17}$ is the unweighted mean over the 17 valid dimensions and is not an official benchmark aggregate. Multi-view consistency is excluded because it has no valid samples.

\subsection{VComBench Judging Protocol}
\label{sec:app_judge_protocol}

\paragraph{Frame sampling.}
GPT-6.1 Sol evaluates each scored video once, without repeated
scoring or voting.
The judge receives uniformly sampled, timestamped frames,
including the first and last frames, resized to a maximum side
of 448 pixels.
Tasks with nominal durations up to 5 seconds use 8 frames;
longer tasks use 12.
We use 24 frames when any scored requirement concerns
repetition count, event timing, rebound count, event order,
or a camera-move sequence.
Original I2V first frames and R2V reference images are provided
with their roles.
Concatenated clips are evaluated as one video.

\paragraph{Requirement scoring.}
Each requirement contains one to three atomic visual checks,
generated during task construction and screened for quality.
Quantitative targets, including distances, angles, and speeds,
are recorded separately in task annotations.
For each check, the judge returns \emph{yes}, \emph{no},
or \emph{unclear}, with a brief explanation.
The judge must reject checks not clearly satisfied
and reserve \emph{unclear} for cases that cannot be resolved from
the sampled frames, such as events between frames.

A requirement passes only when all checks receive \emph{yes}.
Any \emph{no} yields failure; otherwise, an \emph{unclear} or missing
answer makes the requirement unclear.
Unclear requirements count as unmet.
Across configurations, 2.3\% of check answers were unclear.
Input-validity checks are recorded separately and excluded from
requirement satisfaction.
For tasks requesting photorealistic footage, realism is assessed
in a separate evaluation using four sampled frames.
Videos without valid judgments after repeated request failures are
excluded from metric denominators; affected configurations
contain 8--12 videos.

\paragraph{Agreement checks.}
Table~\ref{tab:app_judge_agreement} summarizes cross-route stability
and preliminary human agreement.
For configurations affected by route changes, method--baseline
comparisons use baselines evaluated through the same route.

\begin{table*}[ht]
    \centering
    \small
    \setlength{\tabcolsep}{4pt}
    \renewcommand{\arraystretch}{1.08}
    \caption{Agreement checks.
    Score changes are in percentage points and are absolute
    for the TIDE baselines.}
    \label{tab:app_judge_agreement}
    \begin{tabular*}{\textwidth}{
        @{\extracolsep{\fill}}llrc@{}
    }
        \toprule
        Check & Evaluation set & Agreement (\%) & Score change \\
        \midrule
        Cross-route
        & H3 Native (40 videos, 171 requirements)
        & 93.0 & $S_{\mathrm{req}}: +1.7$ \\

        Cross-route
        & H3 TIDE baseline
        & 94.9 & $|\Delta S_{\mathrm{all}}|: 1.4$ \\

        Cross-route
        & Wan TIDE baseline
        & 95.5 & $|\Delta S_{\mathrm{all}}|: 0.5$ \\

        Human audit
        & H3 pilot (154 videos, 286 requirements)
        & 89.2 & --- \\
        \bottomrule
    \end{tabular*}
\end{table*}

Human annotations were collected without access to automatic verdicts.
Among requirements with definite labels in both sources, 255/286 agreed;
the judge rejected 23 human-accepted requirements and accepted 8
human-rejected requirements.
This earlier H3-only audit uses one annotation set, with no reported
inter-annotator agreement.
Its judge configuration was not retained, so it provides preliminary
evidence rather than validation of the final protocol.

\subsection{VBench-2.0 Subset}
\label{sec:app_vbench2_subset}

\begin{table}[!ht]
    \centering
    \footnotesize
    \setlength{\tabcolsep}{2.6pt}
    \renewcommand{\arraystretch}{1.05}
    \caption{VBench-2.0 subset results. Mean$_{17}$ averages the 17 valid
    dimensions. Best and second-best values per backend are bold
    and underlined, including ties. Native-input rows are included where available.}
    \label{tab:vbench2_dimensions}
    \begin{tabular}{ll*{9}{r}}
\toprule
    Backend & Method & Diversity & Composition & \shortstack{Motion\\rationality} & \shortstack{Instance\\preservation} & \shortstack{Dynamic\\spatial rel.} & \shortstack{Dynamic\\attribute} & \shortstack{Motion\\order} & \shortstack{Human\\interaction} & \shortstack{Complex\\landscape} \\
    \midrule
    H3 & Native & \textbf{0.5006} & 0.4450 & 0.3667 & \underline{0.9667} & \textbf{0.5000} & 0.3333 & 0.5000 & 0.7000 & \textbf{0.1933} \\
     & TIDE & 0.3445 & 0.7900 & \textbf{0.7000} & \textbf{1.0000} & 0.3000 & \textbf{0.5000} & \underline{0.6000} & \textbf{1.0000} & \underline{0.1800} \\
    \rowcolor{bestblue}
     & TIDE + NeurCo (O) & 0.3445 & \textbf{0.8450} & \textbf{0.7000} & \textbf{1.0000} & \underline{0.4000} & \textbf{0.5000} & \textbf{0.7000} & \textbf{1.0000} & 0.1600 \\
    \rowcolor{bestblue}
     & TIDE + NeurCo (T) & \underline{0.3640} & \underline{0.7950} & \underline{0.6000} & \textbf{1.0000} & 0.3000 & \underline{0.4000} & \underline{0.6000} & \textbf{1.0000} & 0.1400 \\
    \rowcolor{bestblue}
     & TIDE + NeurCo (T+O) & 0.3356 & 0.7150 & \textbf{0.7000} & \textbf{1.0000} & \textbf{0.5000} & \underline{0.4000} & \underline{0.6000} & \underline{0.8000} & 0.1400 \\
    \midrule
    Wan & Native & \underline{0.6935} & 0.4867 & \underline{0.5000} & \underline{0.9667} & 0.2000 & 0.4000 & 0.2667 & \underline{0.9000} & \underline{0.1800} \\
    \rowcolor{bestblue}
     & NeurCo (T) & \textbf{0.7254} & 0.5225 & \textbf{0.6000} & 0.9545 & \textbf{0.8000} & \underline{0.6000} & 0.3000 & 0.8333 & \textbf{0.2600} \\
     & TIDE & 0.4593 & \underline{0.7150} & \underline{0.5000} & \textbf{1.0000} & \underline{0.5000} & \textbf{0.7000} & \underline{0.4000} & \textbf{1.0000} & 0.1600 \\
    \rowcolor{bestblue}
     & TIDE + NeurCo (O) & 0.4939 & \textbf{0.7650} & \underline{0.5000} & \textbf{1.0000} & \underline{0.5000} & \textbf{0.7000} & \textbf{0.5000} & \textbf{1.0000} & 0.1600 \\
    \rowcolor{bestblue}
     & TIDE + NeurCo (T) & 0.4505 & 0.5900 & \underline{0.5000} & \textbf{1.0000} & 0.1000 & \underline{0.6000} & \underline{0.4000} & \textbf{1.0000} & 0.1600 \\
    \rowcolor{bestblue}
     & TIDE + NeurCo (T+O) & 0.4675 & 0.6150 & \textbf{0.6000} & \textbf{1.0000} & 0.4000 & \textbf{0.7000} & \textbf{0.5000} & \underline{0.9000} & \underline{0.1800} \\
    \midrule
    Seedance 2.5 & Native & 0.3619 & 0.5750 & 0.5333 & \underline{0.9655} & 0.3214 & \textbf{0.8667} & \textbf{0.6000} & 0.8571 & 0.1800 \\
    \rowcolor{bestblue}
     & NeurCo (T) & \underline{0.3645} & 0.5533 & 0.5000 & \underline{0.9655} & 0.3214 & \textbf{0.8667} & \textbf{0.6000} & 0.8214 & 0.1600 \\
    \rowcolor{bestblue}
     & NeurCo (T+O) & \textbf{0.3663} & 0.4768 & 0.5333 & 0.9643 & 0.3214 & \textbf{0.8667} & \underline{0.5333} & \underline{0.8966} & \underline{0.1867} \\
     & TIDE & 0.2532 & 0.5900 & \underline{0.6000} & 0.9000 & \textbf{0.6000} & \underline{0.8000} & 0.5000 & 0.8000 & \textbf{0.2200} \\
    \rowcolor{bestblue}
     & TIDE + NeurCo (O) & 0.2532 & \underline{0.6650} & \underline{0.6000} & 0.9000 & 0.3000 & \underline{0.8000} & \textbf{0.6000} & 0.8889 & \textbf{0.2200} \\
    \rowcolor{bestblue}
     & TIDE + NeurCo (T) & 0.2454 & 0.6400 & \underline{0.6000} & \textbf{1.0000} & \textbf{0.6000} & \underline{0.8000} & \textbf{0.6000} & 0.8000 & \textbf{0.2200} \\
    \rowcolor{bestblue}
     & TIDE + NeurCo (T+O) & 0.2363 & \textbf{0.7900} & \textbf{0.7000} & \textbf{1.0000} & \underline{0.4000} & \underline{0.8000} & 0.5000 & \textbf{0.9000} & \textbf{0.2200} \\
\midrule
    \addlinespace[4pt]
    Backend & Method & \shortstack{Complex\\plot} & \shortstack{Camera\\motion} & \shortstack{Human\\anatomy} & Identity & Clothing & Mechanics & Thermal & Material & Mean$_{17}$ \\
    \midrule
    H3 & Native & \underline{0.1800} & \underline{0.7333} & 0.9110 & 0.6434 & \underline{0.9630} & 0.6190 & 0.5714 & \underline{0.3889} & 0.5597 \\
     & TIDE & 0.1000 & \textbf{0.8000} & 0.9270 & 0.7937 & \textbf{1.0000} & \textbf{0.6667} & 0.7000 & \textbf{0.8333} & \underline{0.6609} \\
    \rowcolor{bestblue}
     & TIDE + NeurCo (O) & 0.1000 & \textbf{0.8000} & \textbf{0.9318} & 0.7937 & \textbf{1.0000} & \textbf{0.6667} & 0.7000 & \textbf{0.8333} & \textbf{0.6750} \\
    \rowcolor{bestblue}
     & TIDE + NeurCo (T) & 0.1000 & \textbf{0.8000} & \underline{0.9293} & \textbf{0.9008} & \textbf{1.0000} & 0.6000 & \textbf{0.8000} & \textbf{0.8333} & 0.6566 \\
    \rowcolor{bestblue}
     & TIDE + NeurCo (T+O) & \textbf{0.2000} & \textbf{0.8000} & 0.9199 & \underline{0.8614} & \textbf{1.0000} & \underline{0.6250} & \underline{0.7778} & \textbf{0.8333} & 0.6593 \\
    \midrule
    Wan & Native & 0.0800 & 0.4333 & 0.8272 & 0.6891 & 0.9259 & 0.5417 & 0.3571 & 0.5294 & 0.5281 \\
    \rowcolor{bestblue}
     & NeurCo (T) & 0.0889 & 0.4091 & \textbf{0.8827} & 0.8237 & \underline{0.9286} & 0.4545 & \underline{0.6667} & \underline{0.8182} & 0.6275 \\
     & TIDE & \underline{0.1000} & \underline{0.6000} & 0.8708 & \underline{0.8844} & \textbf{1.0000} & \textbf{0.9000} & \textbf{0.7000} & \textbf{0.8333} & \underline{0.6660} \\
    \rowcolor{bestblue}
     & TIDE + NeurCo (O) & \underline{0.1000} & \textbf{0.7000} & 0.8570 & 0.8684 & \textbf{1.0000} & \textbf{0.9000} & \textbf{0.7000} & \textbf{0.8333} & \textbf{0.6810} \\
    \rowcolor{bestblue}
     & TIDE + NeurCo (T) & 0.0800 & \textbf{0.7000} & 0.8030 & \textbf{0.9415} & \textbf{1.0000} & \textbf{0.9000} & 0.5000 & 0.6667 & 0.6113 \\
    \rowcolor{bestblue}
     & TIDE + NeurCo (T+O) & \textbf{0.1778} & \textbf{0.7000} & \underline{0.8812} & 0.8786 & \textbf{1.0000} & \underline{0.8000} & 0.6000 & 0.7143 & 0.6538 \\
    \midrule
    Seedance 2.5 & Native & 0.1000 & \underline{0.8333} & 0.8851 & 0.7684 & \underline{0.9000} & 0.6957 & 0.7407 & \underline{0.9000} & 0.6520 \\
    \rowcolor{bestblue}
     & NeurCo (T) & 0.0667 & \underline{0.8333} & 0.8588 & 0.7605 & 0.8966 & 0.7500 & 0.7407 & 0.8947 & 0.6444 \\
    \rowcolor{bestblue}
     & NeurCo (T+O) & 0.0846 & \underline{0.8333} & 0.8957 & \textbf{0.8761} & 0.8621 & \underline{0.8400} & 0.7407 & 0.7727 & 0.6500 \\
     & TIDE & \underline{0.1111} & \textbf{0.9000} & \textbf{0.9458} & 0.7863 & \textbf{1.0000} & 0.7500 & \textbf{0.9000} & 0.8333 & \underline{0.6759} \\
    \rowcolor{bestblue}
     & TIDE + NeurCo (O) & \textbf{0.1333} & \textbf{0.9000} & \textbf{0.9458} & 0.7863 & \textbf{1.0000} & 0.7500 & \textbf{0.9000} & 0.8333 & 0.6750 \\
    \rowcolor{bestblue}
     & TIDE + NeurCo (T) & 0.0667 & \textbf{0.9000} & 0.8326 & 0.7080 & \textbf{1.0000} & 0.7500 & \underline{0.8000} & 0.7143 & 0.6634 \\
    \rowcolor{bestblue}
     & TIDE + NeurCo (T+O) & 0.0800 & \textbf{0.9000} & \underline{0.9377} & \underline{0.8319} & \textbf{1.0000} & \textbf{0.8750} & 0.6667 & \textbf{1.0000} & \textbf{0.6963} \\
    \bottomrule
    \end{tabular}
\end{table}

We select 175 text-to-video tasks from the 18 official dimensions,
using 10 prompts per dimension except Diversity, which uses five.
For a prompt list of length $P$, the $k$ selected indices are
$\lfloor iP/k\rfloor$, $i=0,\ldots,k-1$.
All tasks use nominal 5-second clips; task identifiers are listed
in the supplementary material.
Video counts follow Sec.~\ref{sec:app_generation_settings}.

We apply the official per-dimension scorers.
Multi-view consistency yields no valid samples and is excluded,
leaving 165 tasks across 17 scored dimensions.
Mean$_{17}$ is their unweighted mean, not an official aggregate.
Scoring follows the official validity filters and excludes unavailable
outputs, so effective sample counts vary across configurations.

\subsection{Sci-VBench Scoring}
\label{sec:app_sci_scoring}

We evaluate the 150-task testmini split using one nominal 5-second
video per task.
Qwen3.5-397B-A17B scores prompt grounding (PG), scientific and causal
correctness (SCC), and spatiotemporal consistency (SC) using the
benchmark's task-specific 1--5 rubrics.
It receives video frames at 2 fps, with temperature 0.7 and top-$p$ 0.8.
We use one scoring round instead of the default three, so small
differences should be interpreted cautiously.

Perceptual quality is measured by VT, the mean of six normalized
automatic metrics: subject consistency, background consistency,
motion smoothness, dynamic degree, aesthetic quality, and imaging quality.
We rescale VT to 1--5 and report an equal-weight summary:
\begin{equation}
    \mathrm{VT}_{1\text{--}5}
    = 1+\frac{4}{6}\sum_{j=1}^{6}s_j,
    \qquad s_j\in[0,1],
\end{equation}
\begin{equation}
    \mathrm{AVG}
    = \frac{
        \mathrm{PG}+\mathrm{SCC}+\mathrm{SC}
        +\mathrm{VT}_{1\text{--}5}
    }{4}.
\end{equation}
Reported scores are averaged over the 150 tasks.
AVG is our summary statistic, not an official benchmark aggregate.
Reused videos retain their existing evaluation scores.

\subsection{Compute and Cost}
\label{sec:app_budget}

Local generation uses NVIDIA H20 GPUs, with one GPU per H3 video
and eight per Wan video.
Table~\ref{tab:app_budget} reports the video-generation expenditure of the
experiments in the main paper.
Requests whose compiled inputs match a baseline reuse that baseline's video
and are not counted.
Local counts therefore cover only generated videos, with VComBench scaled
to its 380 scored tasks.
Routing experiments are listed separately: a task routed to a backend inherits
the video of a matching single-backend run when one exists, and only the
remaining tasks are generated.
GPU hours sum the recorded generation time of each video and exclude model
loading, queueing, idle time, and failed retries;
Seedance costs are recorded charges for successful requests only.

\begin{table*}[ht]
    \centering
    \small
    \setlength{\tabcolsep}{6pt}
    \renewcommand{\arraystretch}{1.05}
    \caption{Generation budget of the main experiments, excluding reused videos.
    Local video counts include one clip per shot for shot-concatenated tasks.
    GPU hours are approximate; monetary costs are rounded to whole CNY.}
    \label{tab:app_budget}
    \begin{tabular*}{\textwidth}{
        @{\extracolsep{\fill}}llrrr@{}
    }
        \toprule
        Backend & Scope & Videos & GPU-h & Cost (CNY) \\
        \midrule
        H3 & Evaluation & 4{,}250 & 3{,}490 & --- \\
           & Routing    & 1{,}390 & 1{,}220 & --- \\
           & Profiling  & 1{,}507 & 1{,}250 & --- \\
           & Total      & 7{,}147 & 5{,}960 & --- \\
        \midrule
        Wan & Evaluation & 5{,}120 & 12{,}400 & --- \\
            & Routing    & 140 & 270 & --- \\
            & Profiling  & 1{,}522 & 3{,}490  & --- \\
            & Total      & 6{,}782 & 16{,}160 & --- \\
        \midrule
        Seedance 2.5 & Evaluation & 4{,}675 & --- & 38{,}246 \\
                     & Routing    & 530      & --- & 4{,}390 \\
                     & Profiling  & 1{,}235 & --- & 9{,}717 \\
                     & Total      & 6{,}440 & --- & 52{,}353 \\
        \bottomrule
    \end{tabular*}
\end{table*}

NeurCo adds compilation and condition-construction costs.
Shot concatenation requires additional video generations, while visual
conditioning can increase generation time.
Language-model calls, image generation, proxy rendering, and evaluation
were not fully timed or costed and are excluded from the table.
The reported budget therefore does not measure total system cost
or paired per-task overhead.

\begin{table}[!ht]
    \centering
    \footnotesize
    \setlength{\tabcolsep}{4pt}
    \renewcommand{\arraystretch}{1.09}
    \caption{The 34 semantic controls used by the capability manual.
    T, I, and R denote T2V, I2V, and R2V. Settings describe the probed
    counts or categories; mode applicability does not imply complete
    measurement coverage.}
    \label{tab:app_control_vocabulary}
    \begin{tabularx}{\textwidth}{@{}p{.045\textwidth}>{\raggedright\arraybackslash}p{.225\textwidth}>{\raggedright\arraybackslash}Xp{.06\textwidth}@{}}
        \toprule
        ID & Control & Definition and settings & Modes \\
        \midrule
        \multicolumn{4}{@{}l}{\textbf{Content (8)}} \\
        A3 & Attributes per subject
           & One subject with 1, 2, 4, or 6 attributes. & TIR \\
        A4 & Attribute binding
           & Distinct attributes assigned to 2, 3, or 4 subjects. & TIR \\
        A5 & Composite concept
           & One or two conceptual combinations in a subject. & TIR \\
        A6 & Subject stability
           & Stable shape and identity for 1, 2, or 3 subjects, without
             duplication or spontaneous appearance/disappearance;
             occlusion and frame exit are allowed. & TIR \\
        P1 & Subject count
           & Exactly 1, 3, 5, 10, 20, or 50 same-kind subjects. & TIR \\
        P7 & Text length
           & Visible text containing 4, 12, 30, or 100 characters. & TIR \\
        G2 & Text language
           & English, Chinese, or Japanese text. & TIR \\
        G3 & Text carrier
           & Text on a sign, screen, t-shirt, or subtitle. & TIR \\
        \midrule
        \multicolumn{4}{@{}l}{\textbf{Space (3)}} \\
        C1 & Spatial relation
           & Relative position, depth order, and occlusion among subjects. & TIR \\
        C2 & Changing spatial relation
           & One or two changes in the relation between subjects. & TIR \\
        C4 & Frame position
           & Subject placed in the left third, center, or right third. & TIR \\
        \midrule
        \multicolumn{4}{@{}l}{\textbf{Motion (7)}} \\
        B1 & Action type
           & Daily activity, sports, fighting, dance, or fine hand manipulation. & TIR \\
        B3 & Subject--action binding
           & Two or three subjects performing individually specified actions. & TIR \\
        B4 & Interaction
           & Handover, pushing/pulling, hugging, or chasing. & TIR \\
        B5 & Action speed
           & Slow motion, normal speed, or fast motion. & TIR \\
        A7 & Facial expression
           & A fixed expression or a change within the shot. & TIR \\
        P2 & Displacement
           & Specified direction and approximate frame-relative distance. & TIR \\
        P3 & Rotation
           & In-place rotation by \(45^\circ\), \(90^\circ\),
             \(180^\circ\), or \(360^\circ\). & TIR \\
        \midrule
        \multicolumn{4}{@{}l}{\textbf{Time (5)}} \\
        T1 & Event timing
           & Event occurrence at a specified time, or a timed event sequence. & TIR \\
        T2 & Repetition count
           & One, three, or five action repetitions; one or three bounces. & TIR \\
        B2 & Event order
           & One, two, three, or five events in a specified order. & TIR \\
        T3 & State change
           & Attribute changes, transformations, or thermal/phase changes. & TIR \\
        E1 & Shot count
           & A sequence of 2, 4, or 6 shots. & TIR \\
        \midrule
        \multicolumn{4}{@{}l}{\textbf{Camera (3)}} \\
        K1 & Camera motion
           & Motion type, tracking style, or an ordered sequence of moves. & TIR \\
        K2 & Camera language
           & Shot size, viewing angle, stability, or focus pull. & TIR \\
        P6 & Push-in scale change
           & Subject height grows from about 20\% to 30\%, 50\%, or 80\%
             of the frame. & TIR \\
        \midrule
        \multicolumn{4}{@{}l}{\textbf{Appearance and physics (6)}} \\
        F1 & Visual style
           & Photorealism, anime, oil painting, pixel art, watercolor,
             or claymation. & TR \\
        F2 & Lighting
           & Time of day, light direction, or color temperature. & TIR \\
        F3 & Color tone
           & Cool/warm tone or a specified dominant color. & TIR \\
        F4 & Weather and effects
           & Rain, snow, fog, fire, or explosion. & TIR \\
        H1 & Mechanics
           & One or two gravity, collision, or fluid phenomena. & TIR \\
        H3 & Material behavior
           & Mixing, dissolving, or bending. & TIR \\
        \midrule
        \multicolumn{4}{@{}l}{\textbf{Reference (2)}} \\
        I3 & Reference role
           & Reference used for a subject, scene, or style. & R \\
        I4 & Reference binding
           & Two, three, or four bindings across supplied references. & R \\
        \bottomrule
    \end{tabularx}
\end{table}

\subsection{Detailed VBench-2.0 Results}
\label{sec:app_vbench2_details}

Table~\ref{tab:vbench2_dimensions} expands the means in
Table~\ref{tab:vbench2_results} into individual dimensions.
It includes the available native-input text and combined runs, including
the Wan text-only result of 0.6275 discussed in the main paper.
After TIDE, operator-only compilation obtains the highest mean on H3 and
Wan, while combined compilation obtains the highest mean on Seedance 2.5.
The dimension scores show mixed effects: operator-only compilation improves
composition on all three backends, but lowers Seedance's dynamic spatial
relations from 0.6000 to 0.3000. On Seedance, combined compilation improves
composition and material properties while reducing the thermal score.


\section{Semantic Control Vocabulary}
\label{sec:app_control_manual}
\label{sec:app_control_vocabulary}

The capability manual covers 34 semantic controls in seven groups
(Table~\ref{tab:app_control_vocabulary}).
Each control is probed under the applicable T2V, I2V, and R2V modes.
The setting \(\ell\) specifies a count, angle, event structure, or
categorical choice, depending on the control.
For broader controls such as camera motion, it also retains the
subtype, so a two-step camera sequence is distinguished from an
individual camera move.
Auxiliary input checks are separate from these controls and are excluded
from benchmark scoring, as described in the main paper.


\begingroup
\setlength{\intextsep}{6pt plus 1pt minus 1pt}
\setlength{\textfloatsep}{10pt plus 1pt minus 1pt}
\setlength{\floatsep}{8pt plus 1pt minus 1pt}
\section{Capability Profiling and Compilation}
\label{sec:app_manual_format}

\subsection{Support Assignment}
\label{sec:app_support_assignment}

Each manual entry records successful and scored single-target probes for a
deployment, control, setting, and input mode.
Valid human labels override automatic judgments; dense re-evaluations replace
default judgments when available.
Invalid specifications and unclear judgments are excluded.
With \(s\) successes among \(n\) scored probes, support is
\begin{equation}
    \sigma(s,n)=
    \begin{cases}
        \text{unknown}, & n=0,\\
        \text{ok}, & n\geq1,\ s/n\geq2/3,\\
        \text{bad}, & n\geq2,\ s/n\leq1/3,\\
        \text{weak}, & \text{otherwise}.
    \end{cases}
    \label{eq:app_support_thresholds}
\end{equation}
A missing entry is unknown; a single failed trial is weak.
The labels summarize empirical native performance on the tested
settings. Their evidence is limited by the number and diversity of probes.

\subsection{Manual Coverage Across Backends}
\label{sec:app_manual_overview}

Table~\ref{tab:app_manual_overview} summarizes entries with scored evidence
for each backend and input mode.
An entry is a control--setting pair; auxiliary input-validity checks are excluded.
Each manual uses only its own backend's probes.

\begin{table}[ht]
    \centering
    \small
    \setlength{\abovecaptionskip}{4pt}
    \setlength{\belowcaptionskip}{4pt}
    \setlength{\tabcolsep}{2pt}
    \renewcommand{\arraystretch}{1.02}
    \caption{Capability-manual coverage and native success rates
    on scored probes.}
    \label{tab:app_manual_overview}
    \begin{tabular*}{\columnwidth}{
        @{\extracolsep{\fill}}llrrrr@{}
    }
        \toprule
        Backend & Mode & Controls & Entries & Probes & Success (\%) \\
        \midrule
        H3 & T2V & 32 & 155 & 544 & 61 \\
           & I2V & 31 & 150 & 444 & 61 \\
           & R2V & 34 & 160 & 470 & 64 \\
        \midrule
        Wan & T2V & 32 & 155 & 545 & 48 \\
            & I2V & 31 & 150 & 444 & 43 \\
            & R2V & 34 & 160 & 474 & 36 \\
        \midrule
        Seedance 2.5 & T2V & 32 & 155 & 526 & 63 \\
                     & I2V & 30 & 144 & 402 & 61 \\
                     & R2V & 33 & 138 & 273 & 65 \\
        \bottomrule
    \end{tabular*}
\end{table}

Seedance~2.5 profiling submitted 1{,}507 requests and obtained
1{,}235 videos; 1{,}201 scored probes
contribute to this summary.
Service refusals particularly reduce R2V coverage.
Success rates are conditional on scored probes.

\subsection{Compilation Example}
\label{sec:app_compilation_example}

Consider an H3 coffee-cup request:
\emph{Three ordinary coffee cups sit on a table, gently steaming with
fresh coffee. Two of the cups remain continuously visible throughout,
retaining their original shapes and appearances.}

Table~\ref{tab:app_compilation_example} traces two of its requirements.
The manual labels the three-cup count as bad and the stability of two
subjects as ok.
Text-only compilation changes ``Three ordinary coffee cups'' to
``Exactly three ordinary coffee cups'' and retains the remaining text.
The operator-enabled plan uses the same rewrite and constructs a first
frame, changing execution from T2V to I2V.
Both plans retain the stability requirement. Figure~\ref{fig:case_cups}
shows the corresponding videos: Native satisfies $2/3$ requirements,
whereas both compiled plans satisfy $3/3$.

\begin{table}[!ht]
    \centering
    \small
    \setlength{\abovecaptionskip}{4pt}
    \setlength{\belowcaptionskip}{4pt}
    \setlength{\tabcolsep}{3pt}
    \renewcommand{\arraystretch}{1.02}
    \caption{Two requirements from the coffee-cup task.
    Counts are successful/scored T2V probes in the H3 manual.}
    \label{tab:app_compilation_example}
    \begin{tabularx}{\columnwidth}{@{}>{\raggedright\arraybackslash}Xccc@{}}
        \toprule
        Requirement & Trials & Status & Decision \\
        \midrule
        Three cups (P1) & 1/3 & bad & Intervene \\
        Two stable subjects (A6) & 3/3 & ok & Keep \\
        \bottomrule
    \end{tabularx}
\end{table}

\section{Execution Strategies}
\label{sec:app_operators_prompts}
\label{sec:app_operator_catalog}

\subsection{External Conditions and Operations}
\label{sec:app_external_conditions}

Table~\ref{tab:app_operators} summarizes the six external operators, their construction, and applicable backends and inputs.
The compiler considers an operator for a weak or bad requirement when
its input and backend conditions are met.
Reference images encode appearance, first frames encode static content
or layout, and proxy videos encode temporal or camera requirements.
Section~\ref{sec:app_cases_proxy} shows proxy-video examples on H3 and Wan.
Wan reference and depth-conditioned calls use VACE.
All video-generation checkpoints remain frozen.

\subsection{Condition Validation}
\label{sec:app_condition_validation}

Before generation, reference images are checked against every target requirement;
first frames for subject count and quoted text (normalizing case, punctuation,
and whitespace); layouts for requested relations; and proxy videos on 24 sampled
frames for timing, repetition, or camera-motion targets and 12 otherwise.
Reference images allow up to two construction attempts; first frames, layouts,
and proxies allow up to three, with proxy retries using validation feedback.
Failed construction or an incorrect number of nonempty shot descriptions retains
the text plan; missing clips are reported as incomplete generation.
These checks assess only the auxiliary condition; generated videos follow
the benchmark judging protocol.

Some first-frame and proxy templates request photorealistic rendering, which can
conflict with stylized requests.
The main paper's realism metric covers the 366 tasks expecting photorealistic
footage and excludes the 14 explicitly stylized ones.

\subsection{Combining Strategies}
\label{sec:app_combining_strategies}

Shot concatenation is used alone.
Otherwise proxy-video and reference-image conditions can be combined through
H3's reference interface or Wan VACE, which skips first-frame construction.
For remaining T2V requests the compiler tries a generated first frame and then a
layout frame; the first success becomes the I2V input.
Original I2V inputs are preserved.
T and T+O enable targeted text rewriting; O disables it but keeps
operator-specific conditioning instructions.
H3 can express a camera-move sequence as a multi-shot prompt in one call, unlike
Wan shot concatenation, which generates separate clips; continuous-shot requests
are never split.

\begin{table}[!ht]
    \centering
    \small
    \setlength{\abovecaptionskip}{4pt}
    \setlength{\belowcaptionskip}{4pt}
    \setlength{\tabcolsep}{3pt}
    \renewcommand{\arraystretch}{1.04}
    \caption{External operators for H3, Wan, and Seedance~2.5.
    T and R denote initial T2V and R2V inputs.
    First-frame construction changes execution from T2V to I2V.}
    \label{tab:app_operators}
    \begin{tabularx}{\textwidth}{
        @{}
        >{\raggedright\arraybackslash}p{.15\textwidth}
        >{\raggedright\arraybackslash}p{.25\textwidth}
        >{\raggedright\arraybackslash}X
        >{\raggedright\arraybackslash}p{.14\textwidth}
        @{}
    }
        \toprule
        Operator & Target requirements & Construction & Backend/input \\
        \midrule

        Reference image
        & Attributes, attribute binding, composite concepts, and visual style
        & Generate a reference image expressing the selected requirements
          and specify its role alongside any existing references.
        & H3, Wan, Seedance~2.5; T/R \\
        \midrule

        Generated first frame
        & Subject count and visible text
        & Generate an opening frame with the required count or quoted text
          and use it as the I2V input.
        & H3, Wan, Seedance~2.5; T \\
        \midrule

        Layout first frame
        & Spatial relations, depth order, and frame position
        & Plan object boxes and depth order, render a layout sketch,
          and convert it to a photorealistic first frame.
        & H3, Seedance~2.5; T \\
        \midrule

        Proxy video reference
        & Displacement, rotation, repetitions, event timing,
          changing relations, and camera-move sequences
        & Program a gray proxy animation and provide it as a video
          reference. H3 is instructed to transfer motion while retaining
          the requested appearance.
        & H3; T/R \\
        \midrule

        Proxy depth control
        & Motion, timing, repetitions, spatial relations, camera motion,
          shot size, viewing angle, and push-in scale change
        & Render a proxy animation and its depth sequence. Supply the depth
          video to VACE, optionally together with reference images.
        & Wan; T/R \\
        \midrule

        Shot concatenation
        & Explicit shot count
        & Generate self-contained shot clips independently and concatenate
          them in order. Wan clips are concatenated at 16 fps.
        & Wan, Seedance~2.5; T/R \\
        \bottomrule
    \end{tabularx}
\end{table}

\paragraph{Seedance 2.5 adaptation.}
Reference images change T2V to R2V, while generated or layout first frames
change it to I2V; original I2V inputs are preserved.
Proxy-video and depth control are unavailable through the evaluated interface,
and last-frame conditioning is untested.
Shot segments last at least 4 seconds.
Moderation-rejected requests produce no video and are excluded
(Table~\ref{tab:vcom_coverage}).

\endgroup

\definecolor{appBlue}{HTML}{2863A6}
\definecolor{appFail}{HTML}{A84C46}
\newlength{\appLabelW}
\newlength{\appGap}
\newlength{\appFrameW}
\newlength{\appFrameH}
\newcommand{\appcaseinit}{%
    \small
    \setlength{\tabcolsep}{0pt}%
    \setlength{\abovecaptionskip}{3pt}%
    \renewcommand{\arraystretch}{1.05}%
    \setlength{\appLabelW}{.14\textwidth}%
    \setlength{\appGap}{6pt}%
    \setlength{\appFrameW}{%
        \dimexpr(\textwidth-\appLabelW-3\appGap)/3\relax}%
    \setlength{\appFrameH}{.5625\appFrameW}}
\newsavebox{\appFrameBox}
\newcommand{\appcaseframe}[2][\appFrameW]{%
    \begingroup
    \setlength{\appFrameW}{#1}%
    \setlength{\appFrameH}{.5625\appFrameW}%
    \sbox{\appFrameBox}{\includegraphics[width=\appFrameW]{#2}}%
    \begin{tikzpicture}[
        baseline={([yshift=-.5ex]current bounding box.center)}
    ]
        \path[use as bounding box]
            (0,0) rectangle (\appFrameW,\appFrameH);
        \begin{scope}[overlay]
            \clip (0,0) rectangle (\appFrameW,\appFrameH);
            \node[inner sep=0,outer sep=0]
                at (.5\appFrameW,.5\appFrameH) {%
                \ifdim\ht\appFrameBox<\appFrameH\relax
                    \includegraphics[height=\appFrameH]{#2}%
                \else
                    \usebox{\appFrameBox}%
                \fi
            };
        \end{scope}
    \end{tikzpicture}%
    \endgroup
}
\newcommand{\appcaseinput}[2]{%
    \setlength{\fboxsep}{4pt}%
    \setlength{\fboxrule}{.4pt}%
    \fcolorbox{appBlue!25}{appBlue!4}{%
        \begin{minipage}{%
            \dimexpr\textwidth-2\fboxsep-2\fboxrule\relax}
            \raggedright\small
            \textbf{Input (T2V).} #1%
            \if\relax\detokenize{#2}\relax\else
                \par\vspace{4pt}
                \textbf{Scored requirements}
                \par\vspace{2pt}#2%
            \fi
        \end{minipage}}%
    \par\vspace{3pt}}
\newcommand{\appcaselabel}[3]{%
    \parbox[c]{\appLabelW}{\centering\small
        \textbf{#1}\\[2pt]#2%
        \if\relax\detokenize{#3}\relax\else\\[2pt]#3\fi}}
\newcommand{\appassetgroup}[2]{%
    {\centering\strut\textbf{#1}\par\vspace{3pt}#2\par}}
\newcommand{\appcaseinputas}[5]{%
    \setlength{\fboxsep}{4pt}%
    \setlength{\fboxrule}{.4pt}%
    \fcolorbox{appBlue!25}{appBlue!4}{%
        \begin{minipage}{\dimexpr\textwidth-2\fboxsep-2\fboxrule\relax}
            \raggedright\small
            \textbf{Input (#1).} #2%
            \if\relax\detokenize{#3}\relax
                \if\relax\detokenize{#4}\relax\else
                    \par\vspace{3pt}\appassetgroup{Conditions constructed by NeurCo}{#4}%
                \fi
            \else
                \par\vspace{3pt}%
                \if\relax\detokenize{#4}\relax
                    \appassetgroup{User-provided images}{#3}%
                \else
                    \noindent\begin{minipage}[t]{.48\linewidth}
                        \vspace{0pt}\appassetgroup{User-provided images}{#3}%
                    \end{minipage}\hfill
                    \begin{minipage}[t]{.49\linewidth}
                        \vspace{0pt}\appassetgroup{Conditions constructed by NeurCo}{#4}%
                    \end{minipage}\par
                \fi
            \fi
            \if\relax\detokenize{#5}\relax\else
                \par\vspace{3pt}\textbf{Scored requirements}\par\vspace{2pt}#5%
            \fi
        \end{minipage}}%
    \par\vspace{3pt}}
\newcommand{\appinputimg}[2]{%
    \hspace{3pt}\begin{minipage}[t]{.43\linewidth}
        \centering\includegraphics[height=1.55cm]{#1}\par\vspace{2pt}
        {\scriptsize #2\par}%
    \end{minipage}\hspace{3pt}}
\newcommand{\appteachar}{%
    \raisebox{-.1ex}{%
    \begin{tikzpicture}[x=1.35ex,y=1.35ex,baseline=0pt]
    \path[fill=black,draw=none]
        (0.02174,0.91304)
        -- (0.24348,0.91304)
        -- (0.24348,1.00000)
        -- (0.33478,1.00000)
        -- (0.33478,0.91304)
        -- (0.66522,0.91304)
        -- (0.66522,1.00000)
        -- (0.75652,1.00000)
        -- (0.75652,0.91304)
        -- (0.97826,0.91304)
        -- (0.97826,0.83043)
        -- (0.75652,0.83043)
        -- (0.75652,0.73043)
        -- (0.66522,0.73043)
        -- (0.66522,0.83043)
        -- (0.33478,0.83043)
        -- (0.33478,0.72609)
        -- (0.24348,0.72609)
        -- (0.24348,0.83043)
        -- (0.02174,0.83043)
        -- cycle
        (0.50000,0.79130)
        -- (0.59130,0.76087)
        -- (0.56522,0.73478)
        .. controls (0.70725,0.64783) and (0.85072,0.58261) .. (0.99565,0.53913)
        -- (0.94348,0.45217)
        .. controls (0.78986,0.50435) and (0.64203,0.58116) .. (0.50000,0.68261)
        .. controls (0.36087,0.58116) and (0.21014,0.50290) .. (0.04783,0.44783)
        -- (0.00000,0.52609)
        .. controls (0.23188,0.61014) and (0.39855,0.69855) .. (0.50000,0.79130)
        -- cycle
        (0.11304,0.44348)
        -- (0.45652,0.44348)
        -- (0.45652,0.57391)
        -- (0.54783,0.57391)
        -- (0.54783,0.44348)
        -- (0.88696,0.44348)
        -- (0.88696,0.36087)
        -- (0.54783,0.36087)
        -- (0.54783,0.09565)
        .. controls (0.54493,0.03188) and (0.50580,0.00000) .. (0.43043,0.00000)
        -- (0.34348,0.01304)
        -- (0.32609,0.09565)
        -- (0.41739,0.08261)
        .. controls (0.44348,0.08261) and (0.45652,0.09710) .. (0.45652,0.12609)
        -- (0.45652,0.36087)
        -- (0.11304,0.36087)
        -- cycle
        (0.23913,0.32174)
        -- (0.32174,0.27826)
        .. controls (0.24928,0.17681) and (0.16957,0.10000) .. (0.08261,0.04783)
        -- (0.00870,0.10870)
        .. controls (0.10435,0.17246) and (0.18116,0.24348) .. (0.23913,0.32174)
        -- cycle
        (0.73913,0.31739)
        -- (0.95652,0.10435)
        -- (0.87826,0.03913)
        -- (0.66957,0.26957)
        -- cycle;
    \end{tikzpicture}}}

\makeatletter
\newenvironment{appcaseblock}{%
    \par\addvspace{4pt}\begingroup\def\@captype{figure}%
}{%
    \par\endgroup\addvspace{3pt}}
\makeatother

\newcommand{\appgallerycard}[6]{%
    \begin{minipage}[t]{.32\textwidth}
        \vspace{0pt}\footnotesize\raggedright
        \parbox[t][18pt][t]{\linewidth}{\textbf{#1}}\par\vspace{2pt}
        \parbox[t][90pt][t]{\linewidth}{\textbf{Input.} #2}\par\vspace{2pt}
        \parbox[t][45pt][t]{\linewidth}{\textbf{Scored requirements.} #3}\par\vspace{2pt}
        {\centering\appcaseframe[\linewidth]{#4}\par}\vspace{3pt}
        \parbox[t][23pt][t]{\linewidth}{#5 #6}
    \end{minipage}}

\section{Additional Case Studies}
\label{sec:case_study_appendix}

NeurCo uses deployment-specific capabilities to select and combine
textual and visual controls. The cases below pair these compilation
choices with their video outcomes, spanning rendered motion guidance,
selective rewriting, visual conditioning, and failure modes. Scores
come from full-video judgments; the frames illustrate the results.

\subsection{Rendered Proxy Controls}
\label{sec:app_cases_proxy}

Rendered motion conditions provide a complementary route to controlling
camera movement and physical events. H3 uses a video reference and Wan
uses a depth-based motion guide
(Appendix~\ref{sec:app_proxy_shared_prompts}). Among tasks that invoke a
proxy, T+O improves upon both Native and T on 17 of 51 H3 tasks, with
21 ties and 13 declines; on Wan it improves on 45 of 113 tasks, with
43 ties and 25 declines. Satisfaction of proxy-targeted requirements
rises from 41.2\% to 70.6\% on H3 (17 requirements), and from
37.7\% to 69.8\% on Wan (53 requirements); the latter reaches
45.3\% with text rewriting alone.

\clearpage
\begin{appcaseblock}
\noindent\textbf{Count correction with preserved subject stability.}
The H3 coffee-cup example links manual-guided compilation to the generated
result (Table~\ref{tab:app_compilation_example} and
Figure~\ref{fig:case_cups}). The manual marks the three-cup count as
\emph{bad} ($1/3$ successful probes) and two-subject stability as
\emph{ok} ($3/3$). NeurCo revises the count requirement and leaves the
supported stability requirement unchanged. Native satisfies $2/3$
requirements, while both the text-only and first-frame plans reach
$3/3$, correcting the count while preserving subject stability.

\par\smallskip\noindent\begingroup\centering
    \centering
    \begingroup
    \appcaseinit
    \appcaseinput{Three ordinary coffee cups sit on a table, gently steaming
    with fresh coffee. Two of the cups remain continuously visible
    throughout, retaining their original shapes and appearances.}{%
        \begin{tabularx}{\linewidth}{@{}XXX@{}}
            (1) Stable cup identities
            & \textcolor{appBlue}{\textbf{(2) Exactly three cups}}
            & (3) Recognizable coffee cups
        \end{tabularx}}
    \begin{tabular}{@{}c@{\hspace{\appGap}}c@{\hspace{\appGap}}c@{\hspace{\appGap}}c@{}}
        \parbox[c]{\appLabelW}{\centering\textbf{Method}} & \textbf{Early} & \textbf{Middle} & \textbf{Late} \\[2pt]
        \appcaselabel{Native}{\textbf{2/3 passed}}{\textcolor{appFail}{Fails (2)}}
        & \appcaseframe{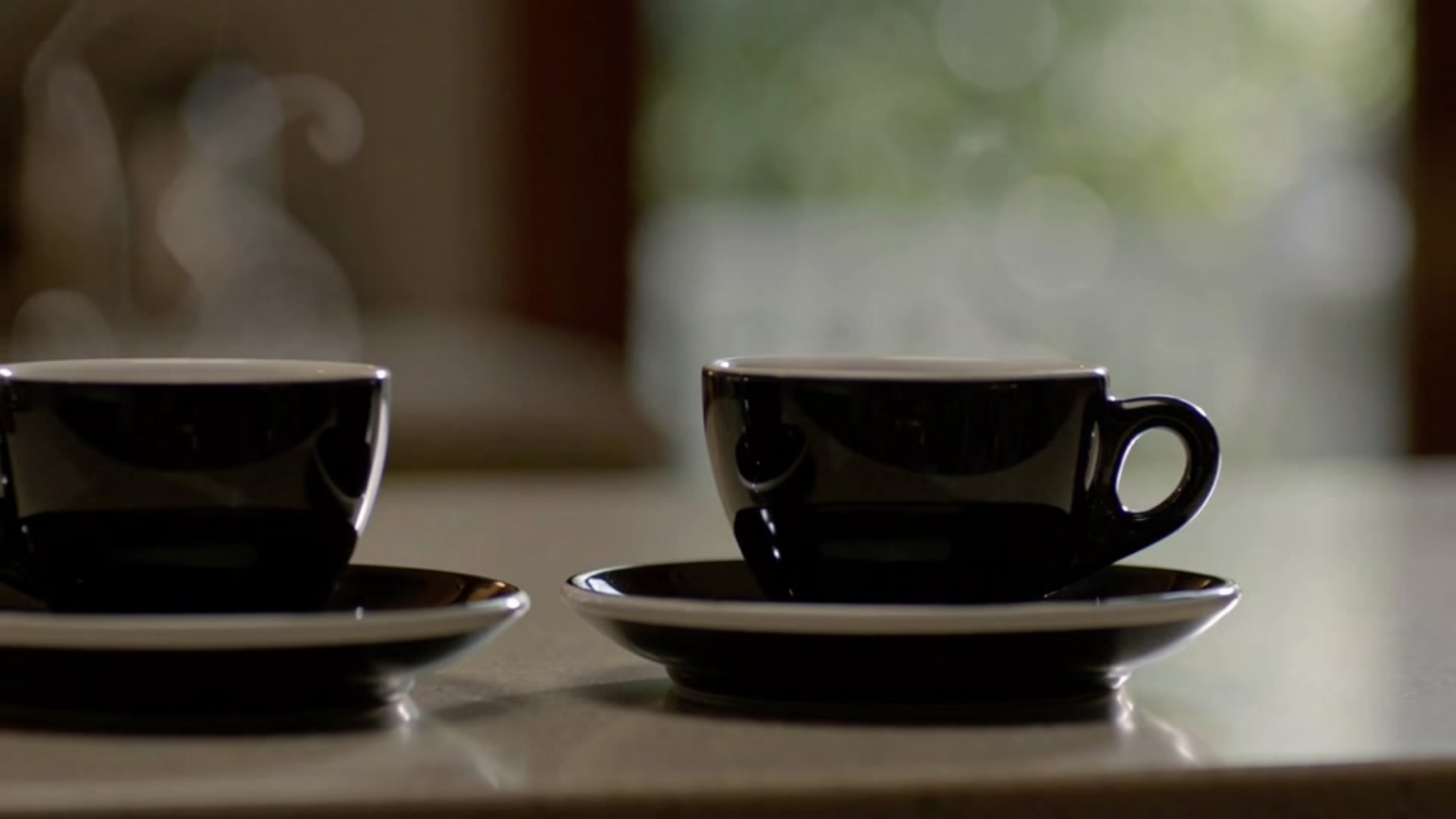}
        & \appcaseframe{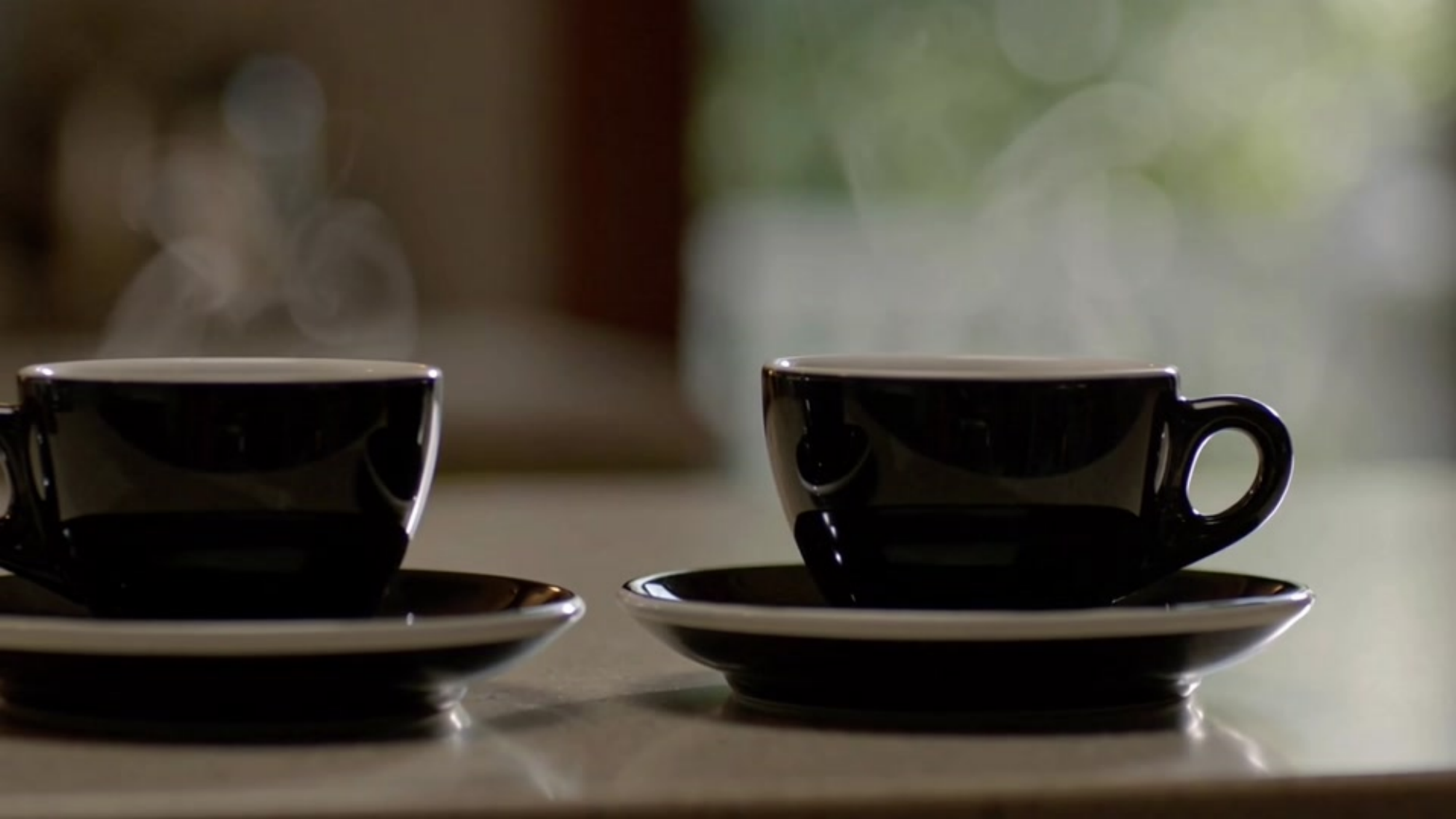}
        & \appcaseframe{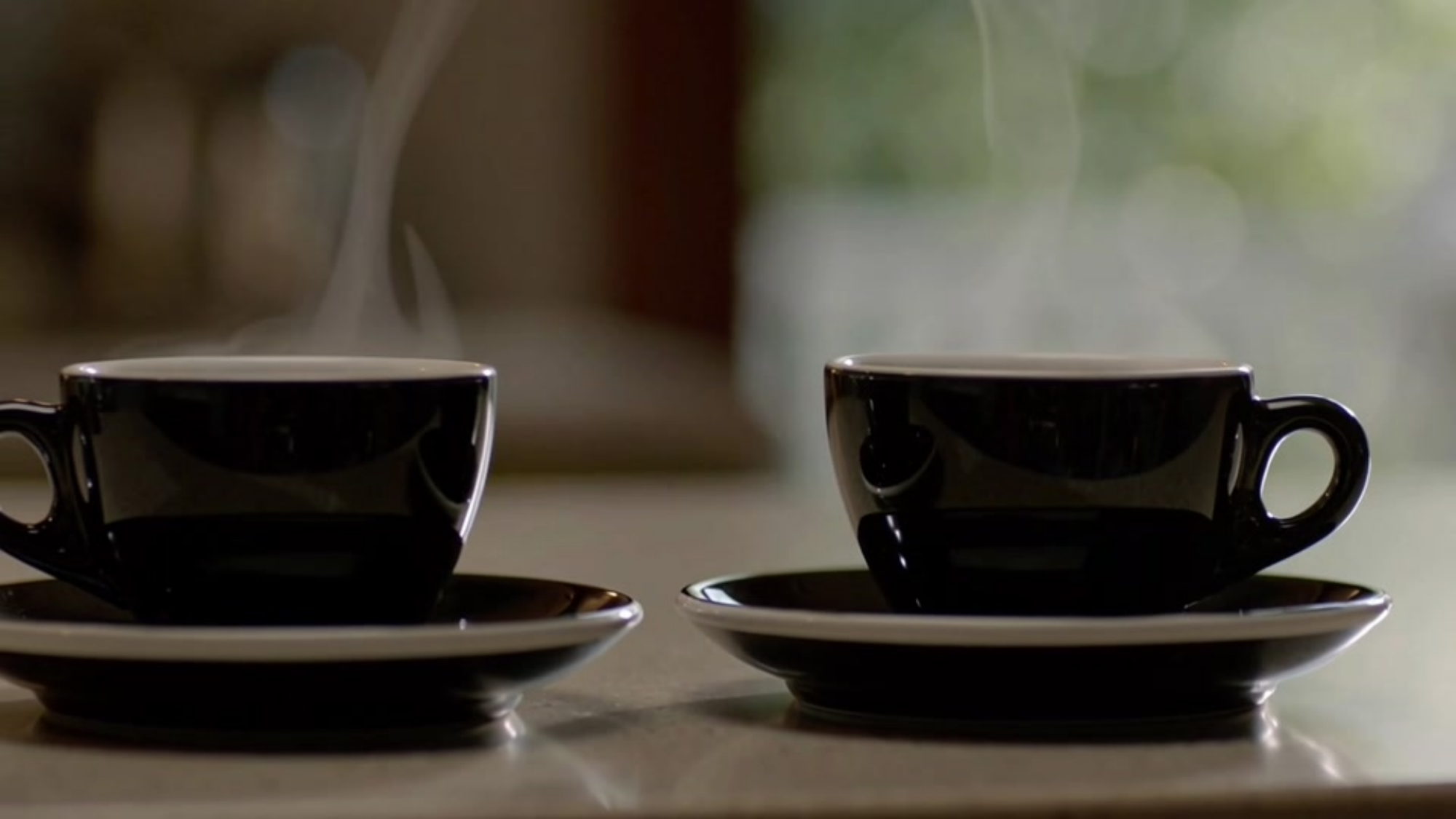}
        \end{tabular}
\par\vspace{3pt}\noindent\begin{tabular}{@{}c@{\hspace{\appGap}}c@{\hspace{\appGap}}c@{\hspace{\appGap}}c@{}}

        \appcaselabel{NeurCo (T)}{\textcolor{appBlue}{\textbf{3/3 passed}}}{Text adaptation}
        & \appcaseframe{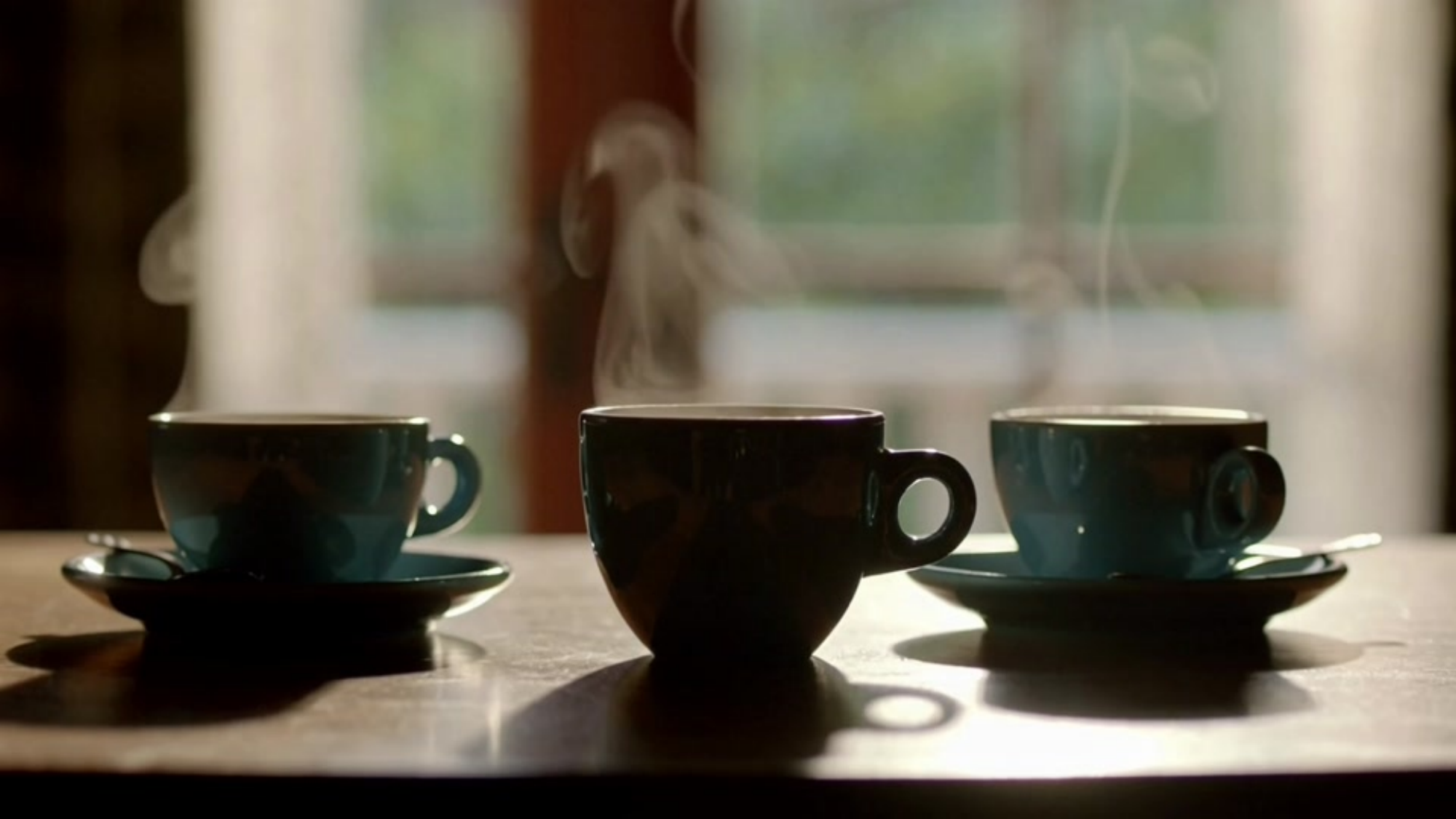}
        & \appcaseframe{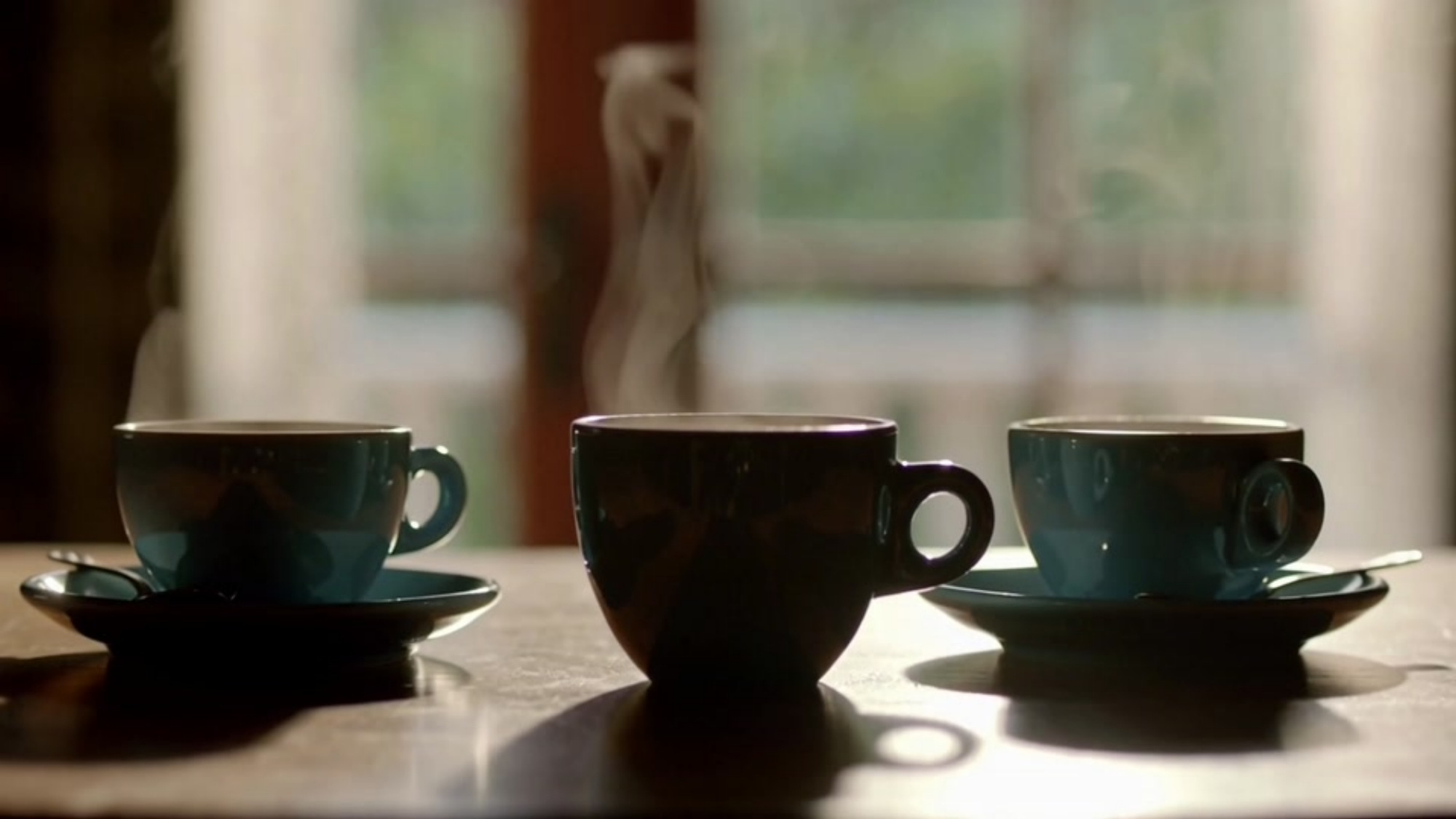}
        & \appcaseframe{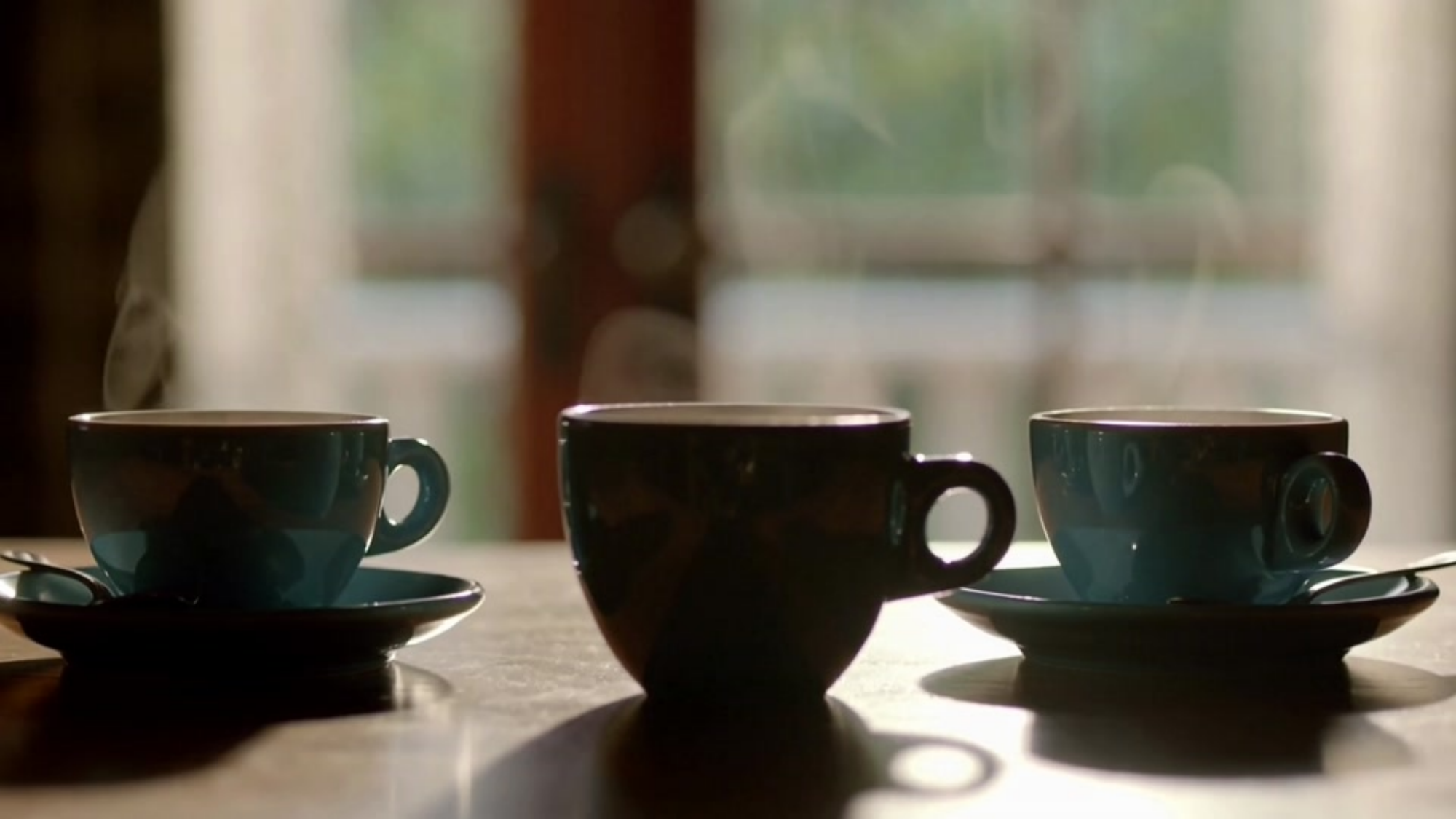}
        \end{tabular}
\par\vspace{3pt}\noindent\begin{tabular}{@{}c@{\hspace{\appGap}}c@{\hspace{\appGap}}c@{\hspace{\appGap}}c@{}}

        \appcaselabel{NeurCo (T+O)}{\textcolor{appBlue}{\textbf{3/3 passed}}}{First-frame I2V}
        & \appcaseframe{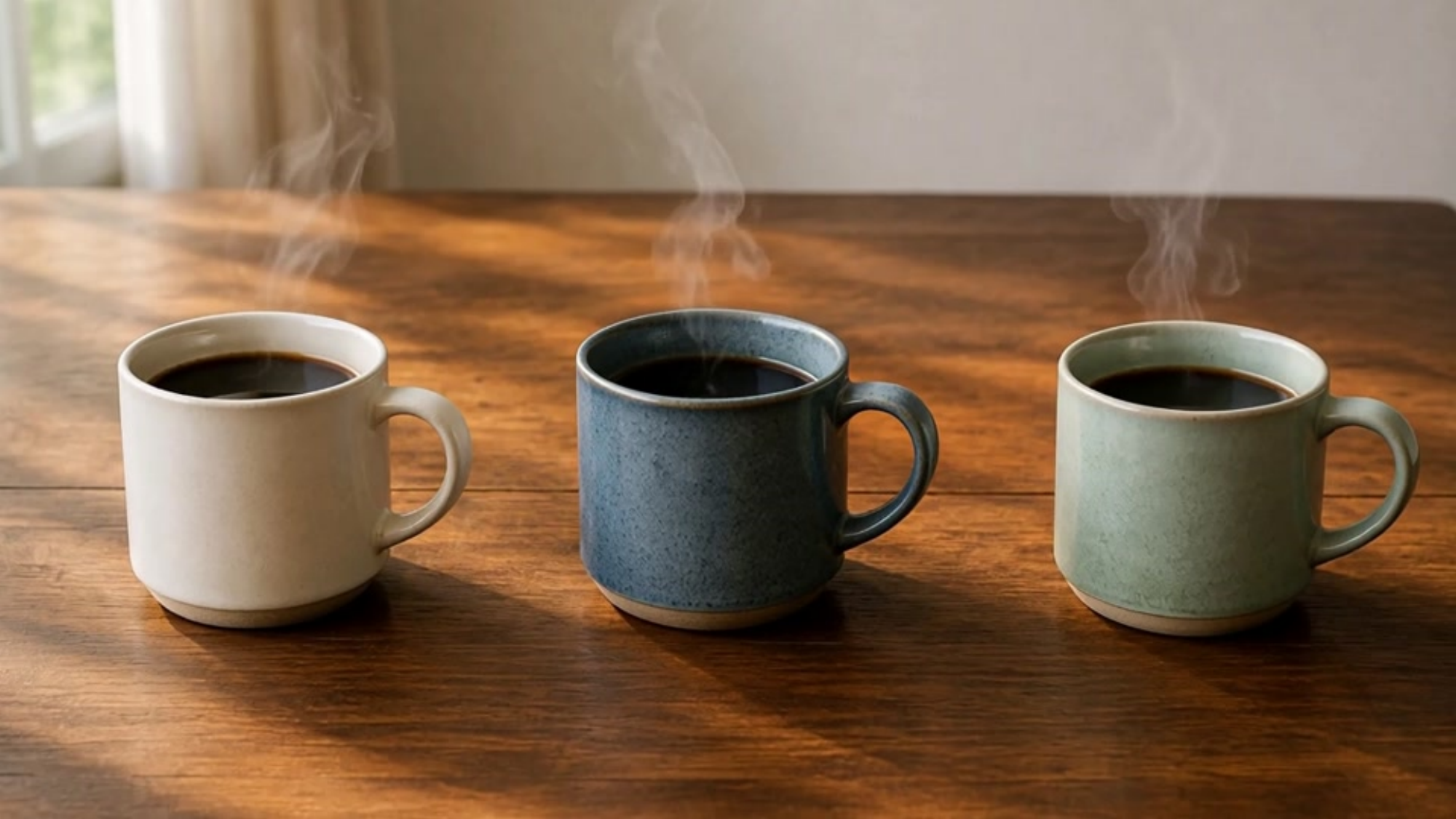}
        & \appcaseframe{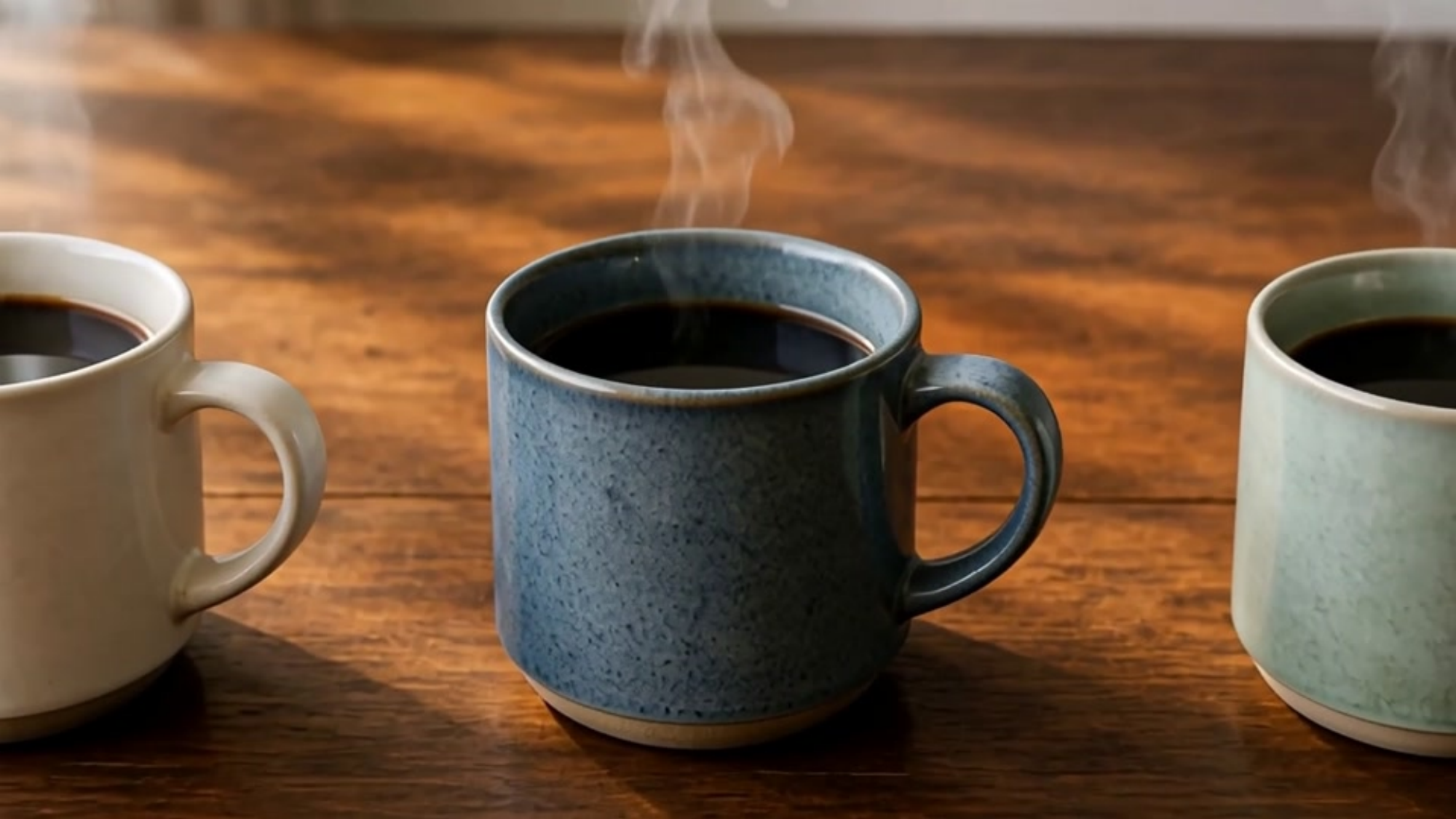}
        & \appcaseframe{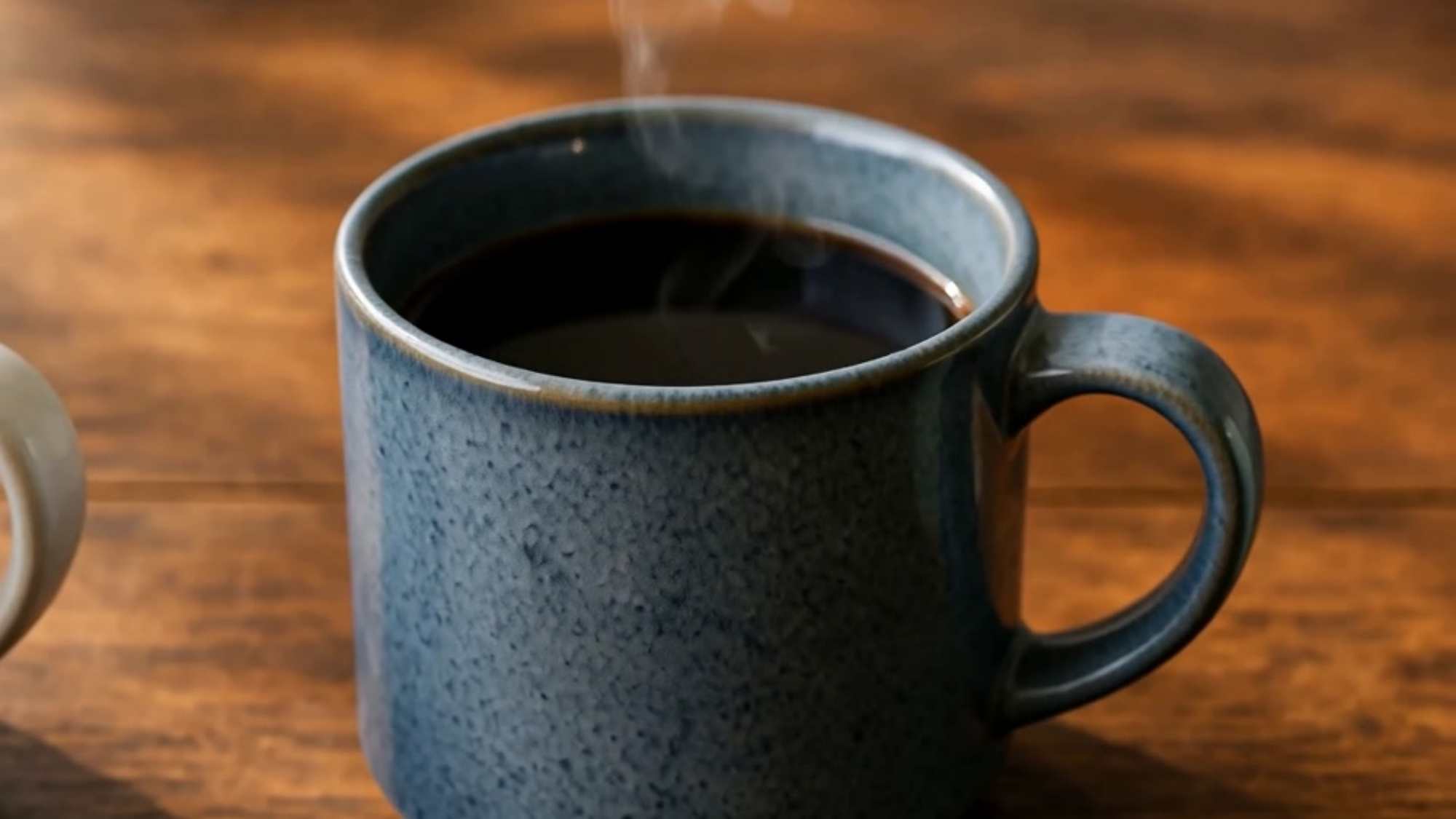}
    \end{tabular}
    \nopagebreak[3]\caption{\textbf{Selective count correction on H3.}
    Both compiled plans satisfy all three requirements, compared with
    $2/3$ for Native; the stability requirement is preserved.}
    \label{fig:case_cups}
    \endgroup
\par\endgroup
\end{appcaseblock}

\clearpage
\begin{appcaseblock}
\noindent\textbf{Camera control with depth guidance.}
The depth-guided plan recovers the low viewpoint and camera push-in on
Wan, raising requirement satisfaction from $2/6$ to $5/6$
(Figure~\ref{fig:case_proxy_hydrant}). The hydrant grows in the
frame as requested, and the lettering appears at the prescribed time;
the full message remains unreadable.

\par\smallskip\noindent\begingroup\centering
    \centering
    \begingroup
    \appcaseinit
    \appcaseinputas{T2V}{A fire hydrant stands beside a stop sign in golden-hour sunlight. Frame both from a low angle,
    looking upward. Slowly push toward the hydrant, increasing its height from about 20\% to about 30\% of the frame
    while keeping the sign visible. The sign's red face begins blank. At second 1, its lettering appears and remains
    clearly readable: ``STOP. Keep this fire hydrant clear at all times. Leave space for emergency crews to reach it
    safely.''}{}{%
        \appinputimg{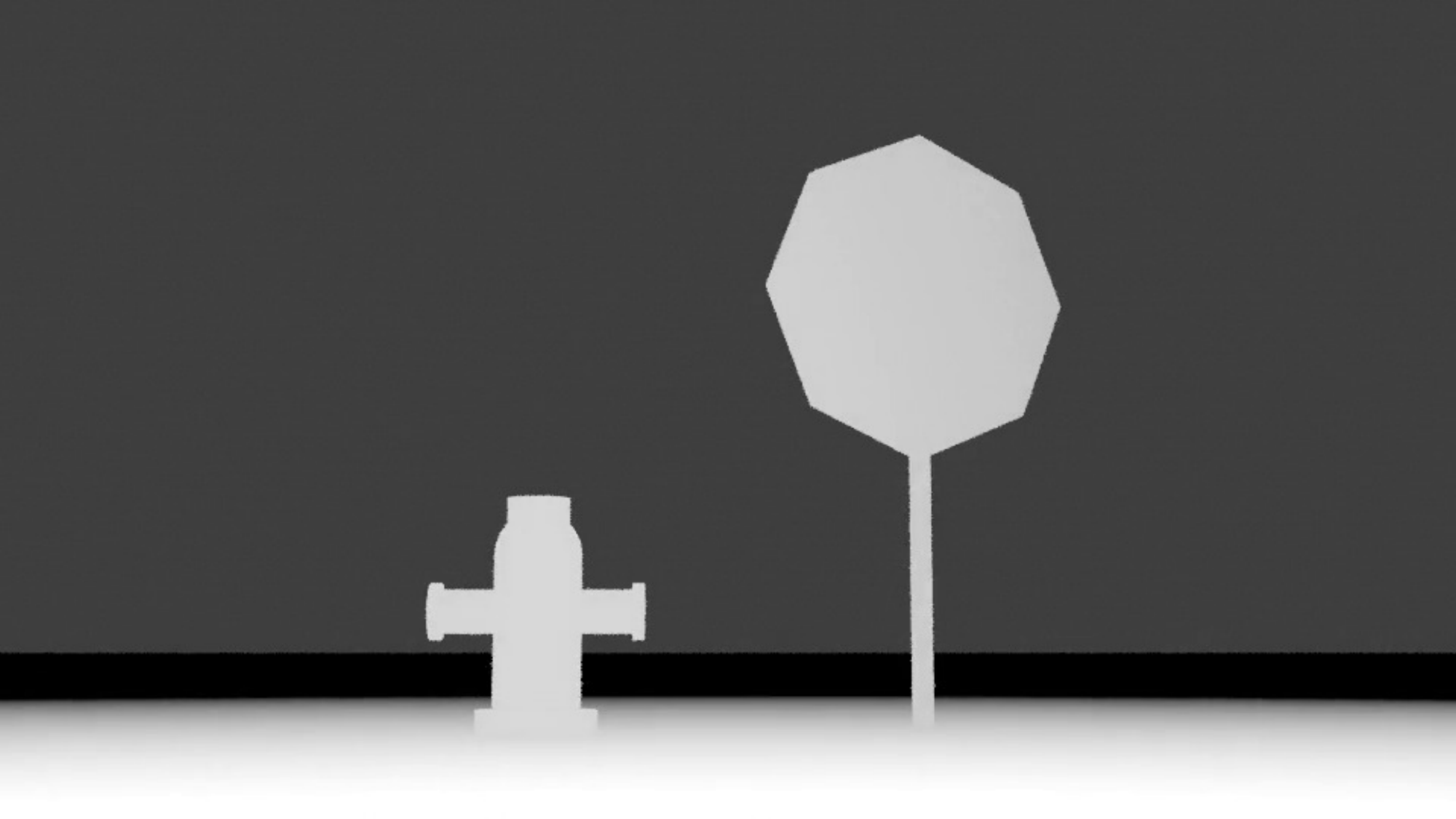}{Wan depth control (middle frame)}%
    }{%
        \begin{tabularx}{\linewidth}{@{}XXX@{}}
            \textcolor{appFail}{\textbf{(1) Complete readable message}} & (2) Hydrant and sign visible & \textcolor{appBlue}{\textbf{(3) Low angle}} \\
            \textcolor{appBlue}{\textbf{(4) Push-in, 20\% to 30\%}} & \textcolor{appBlue}{\textbf{(5) Lettering appears at second 1}} & (6) Golden hour
        \end{tabularx}}
    \begin{tabular}{@{}c@{\hspace{\appGap}}c@{\hspace{\appGap}}c@{\hspace{\appGap}}c@{}}
        \parbox[c]{\appLabelW}{\centering\textbf{Method}} & \textbf{Early} & \textbf{Middle} & \textbf{Late} \\[2pt]
        \appcaselabel{Wan Native}{\textbf{2/6 passed}}{\textcolor{appFail}{Fails (1), (3)--(5)}}
        & \appcaseframe{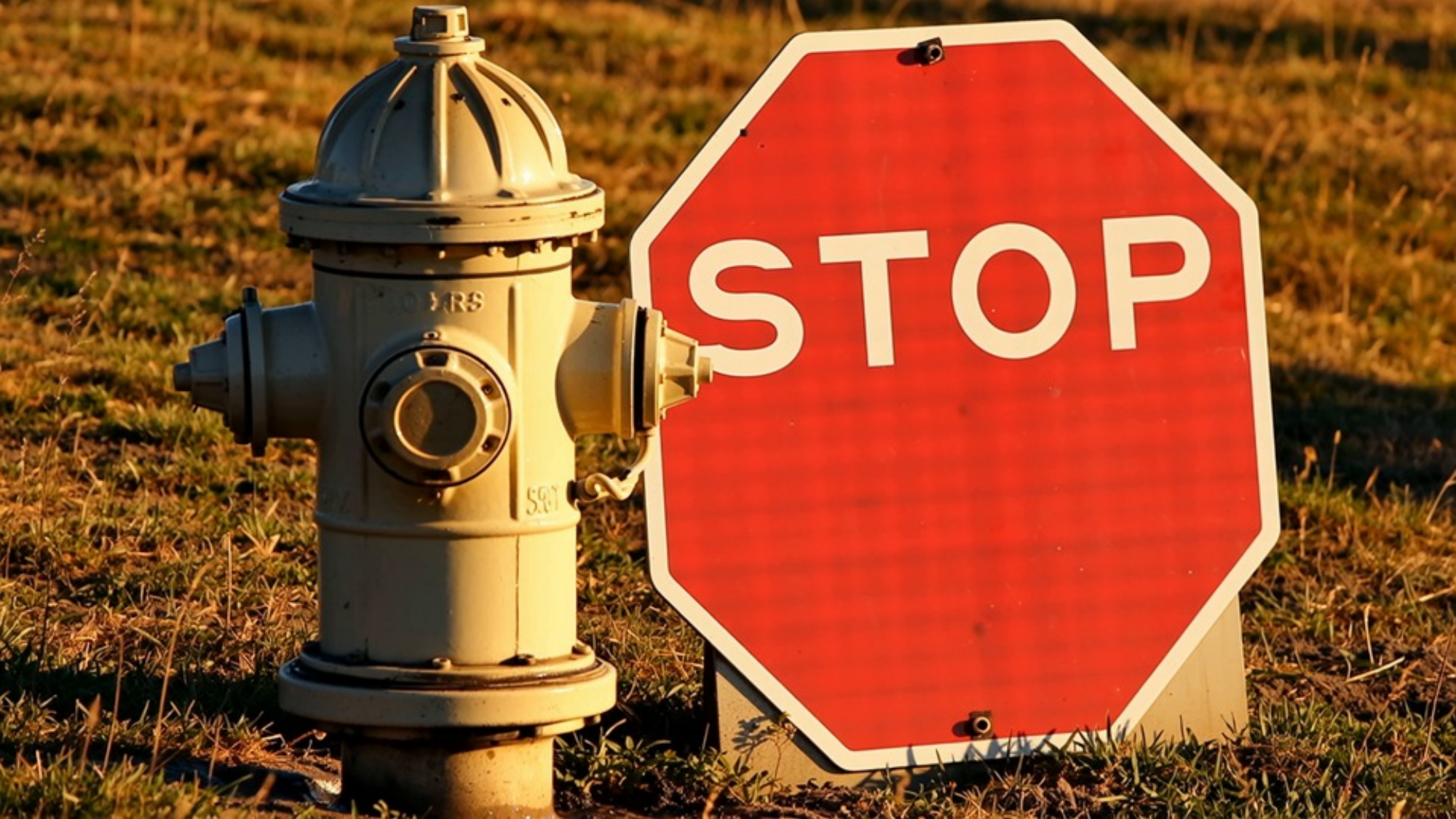}
        & \appcaseframe{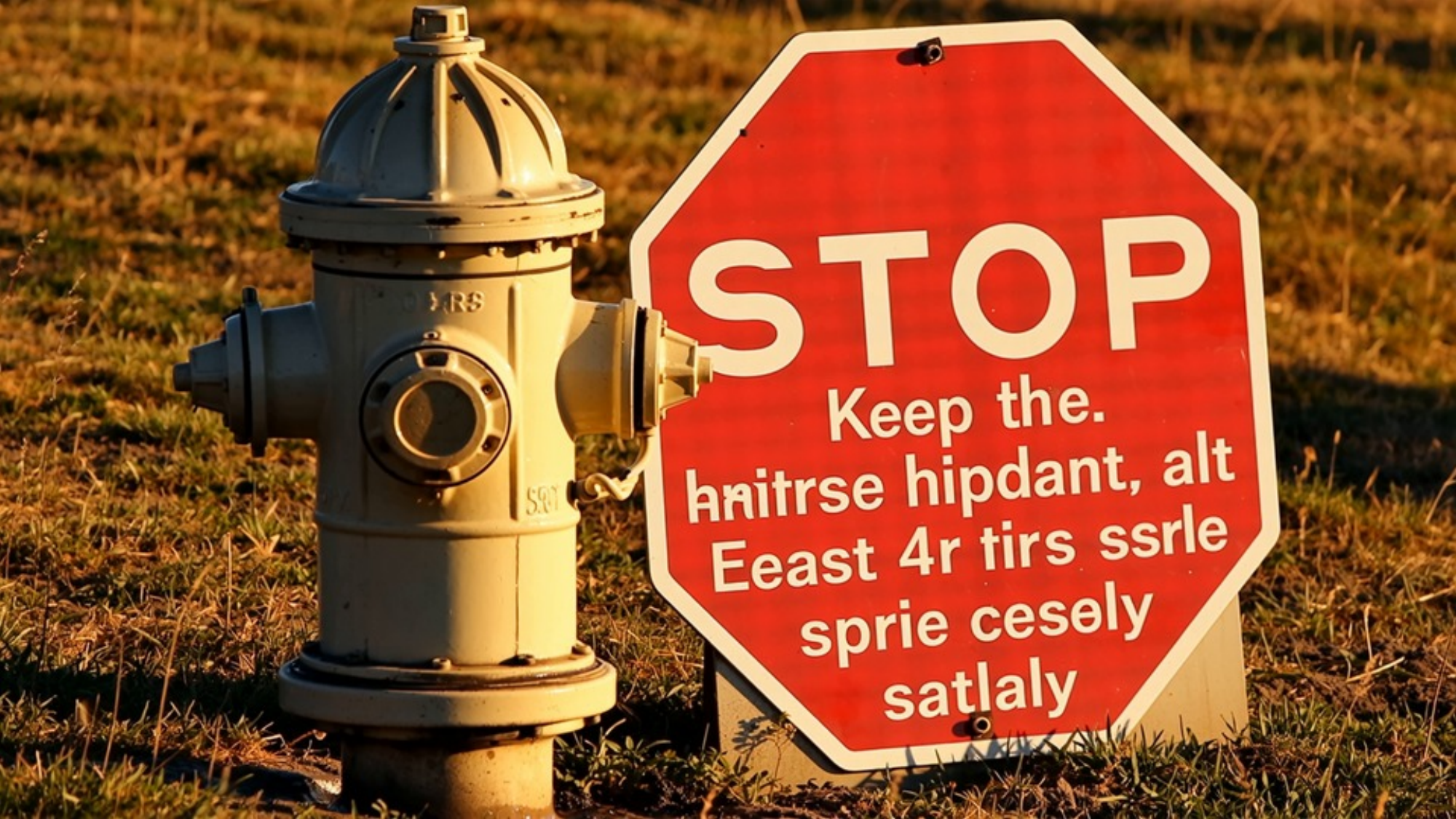}
        & \appcaseframe{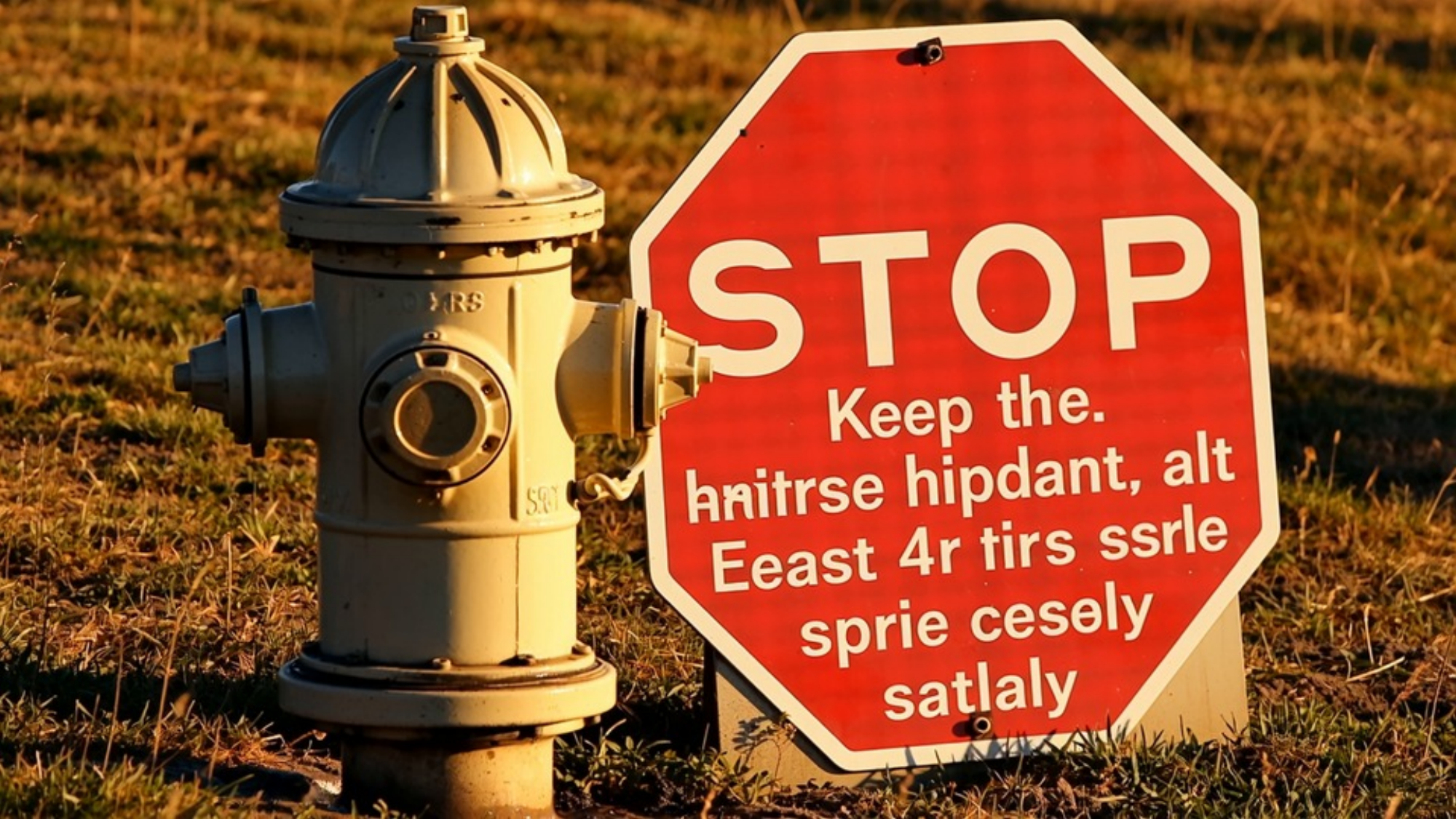}
        \end{tabular}
\par\vspace{3pt}\noindent\begin{tabular}{@{}c@{\hspace{\appGap}}c@{\hspace{\appGap}}c@{\hspace{\appGap}}c@{}}

        \appcaselabel{Wan NeurCo (T)}{\textbf{2/6 passed}}{\textcolor{appFail}{Fails (1), (3)--(5)}}
        & \appcaseframe{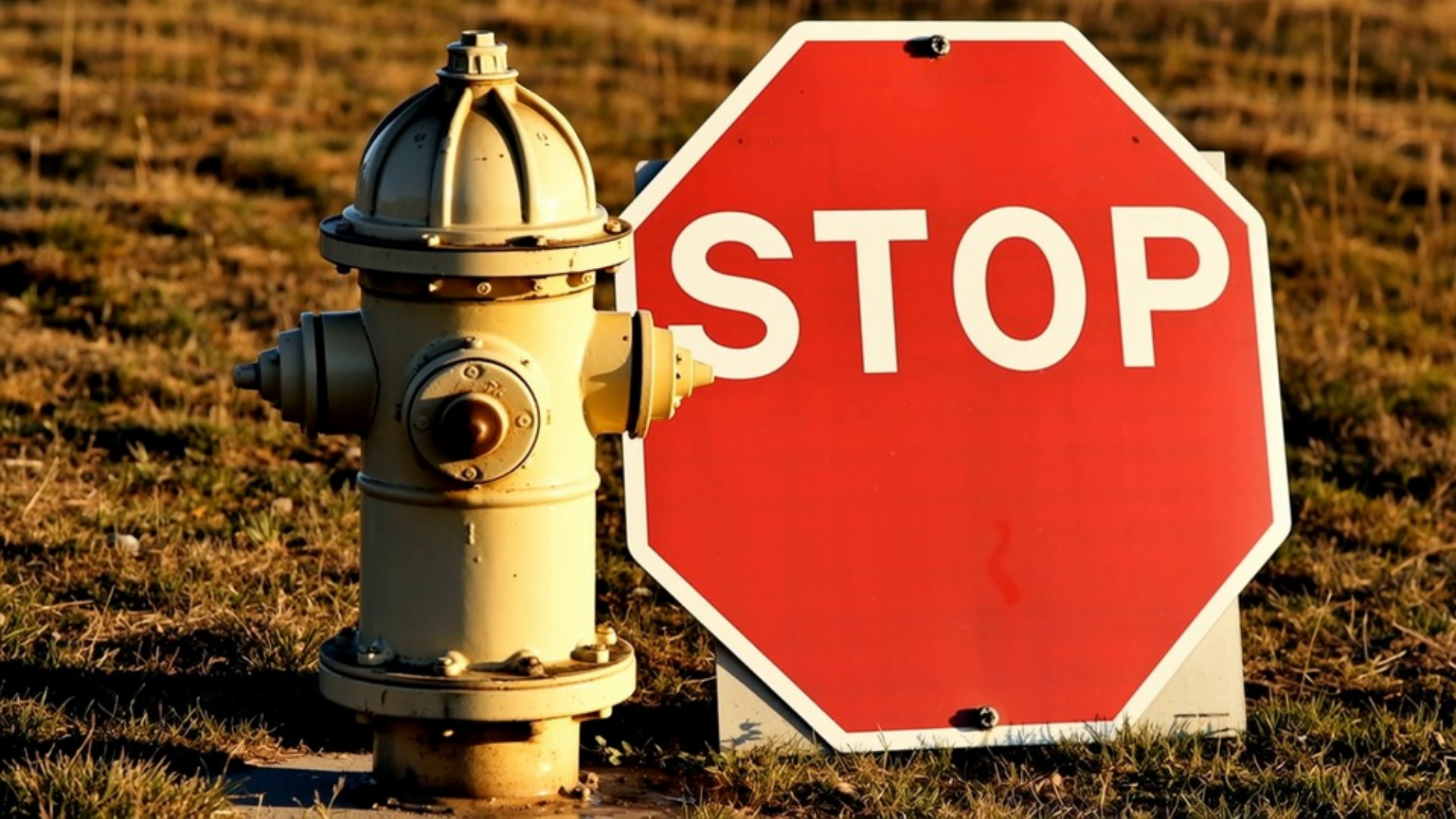}
        & \appcaseframe{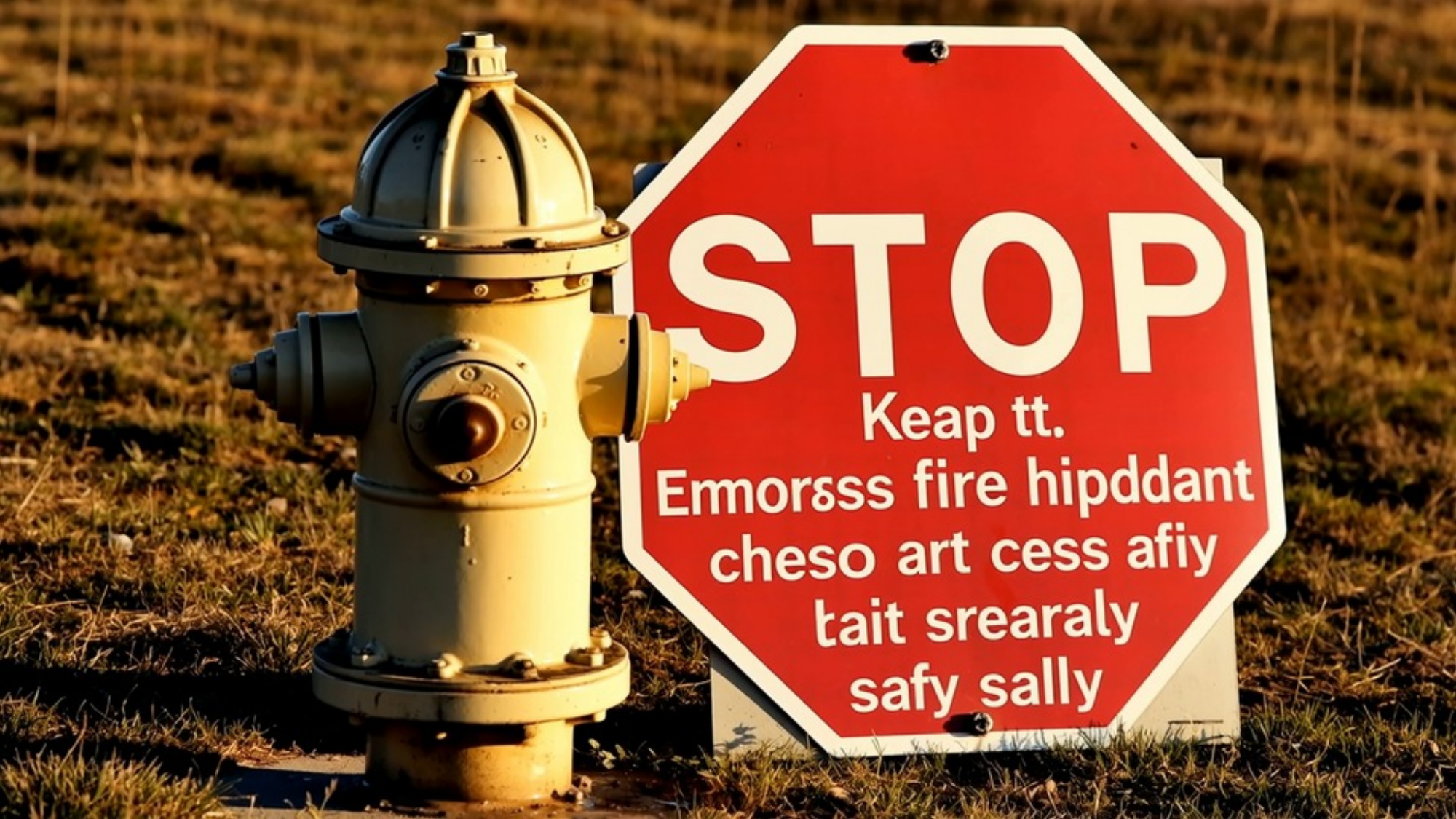}
        & \appcaseframe{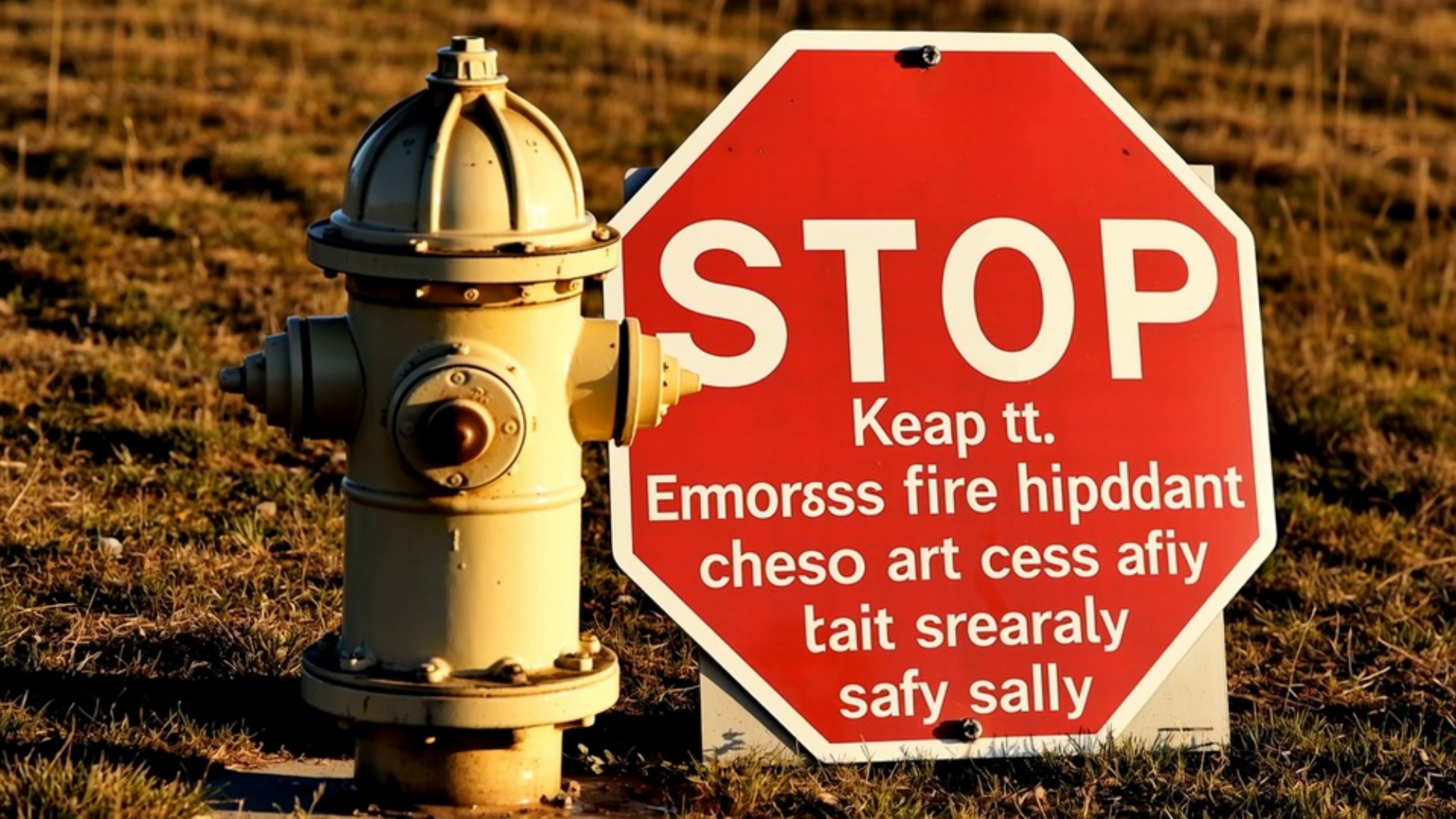}
        \end{tabular}
\par\vspace{3pt}\noindent\begin{tabular}{@{}c@{\hspace{\appGap}}c@{\hspace{\appGap}}c@{\hspace{\appGap}}c@{}}

        \appcaselabel{Wan NeurCo (T+O)}{\textcolor{appBlue}{\textbf{5/6 passed}}}{Depth control + text; \textcolor{appFail}{fails (1)}}
        & \appcaseframe{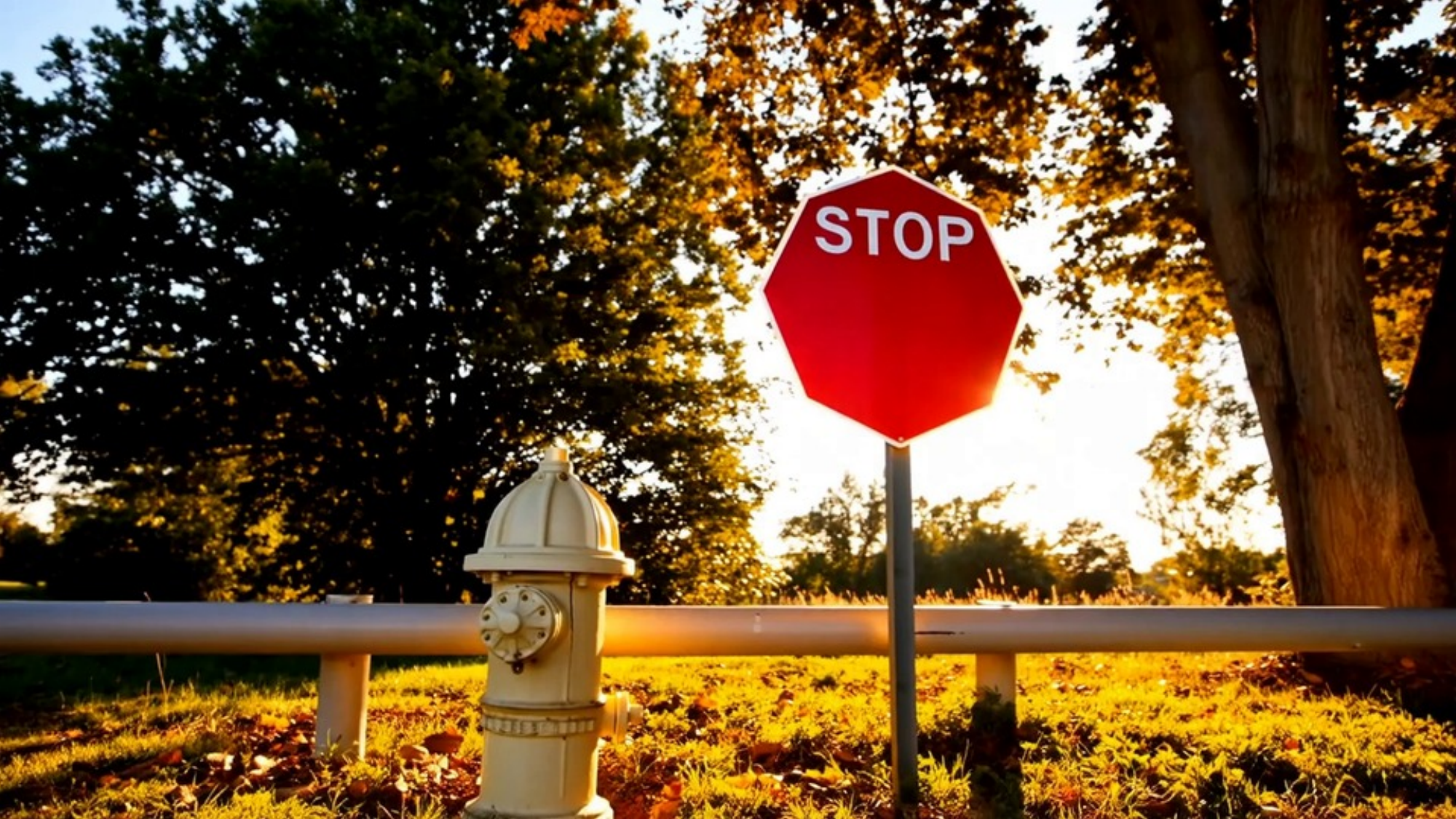}
        & \appcaseframe{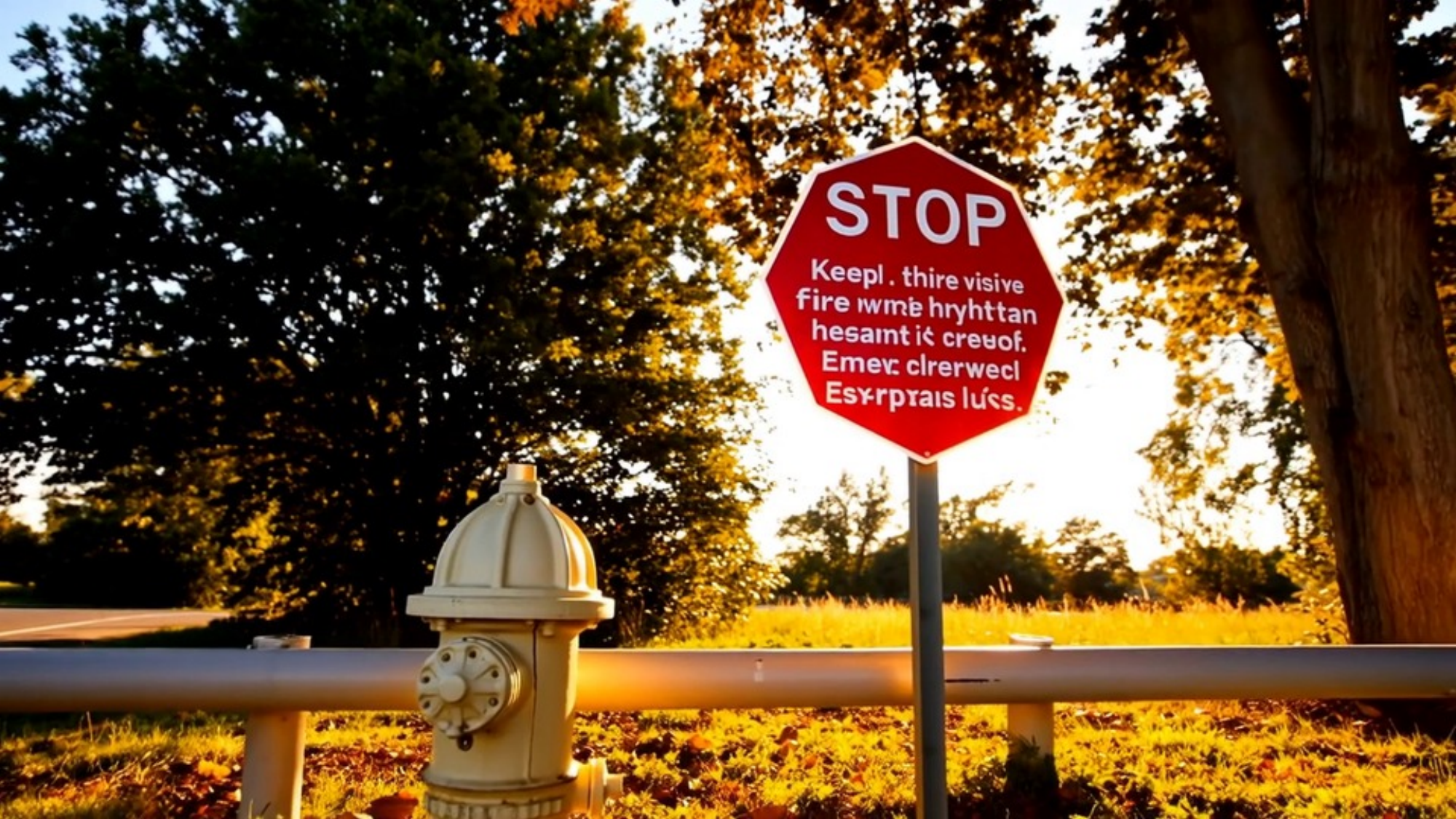}
        & \appcaseframe{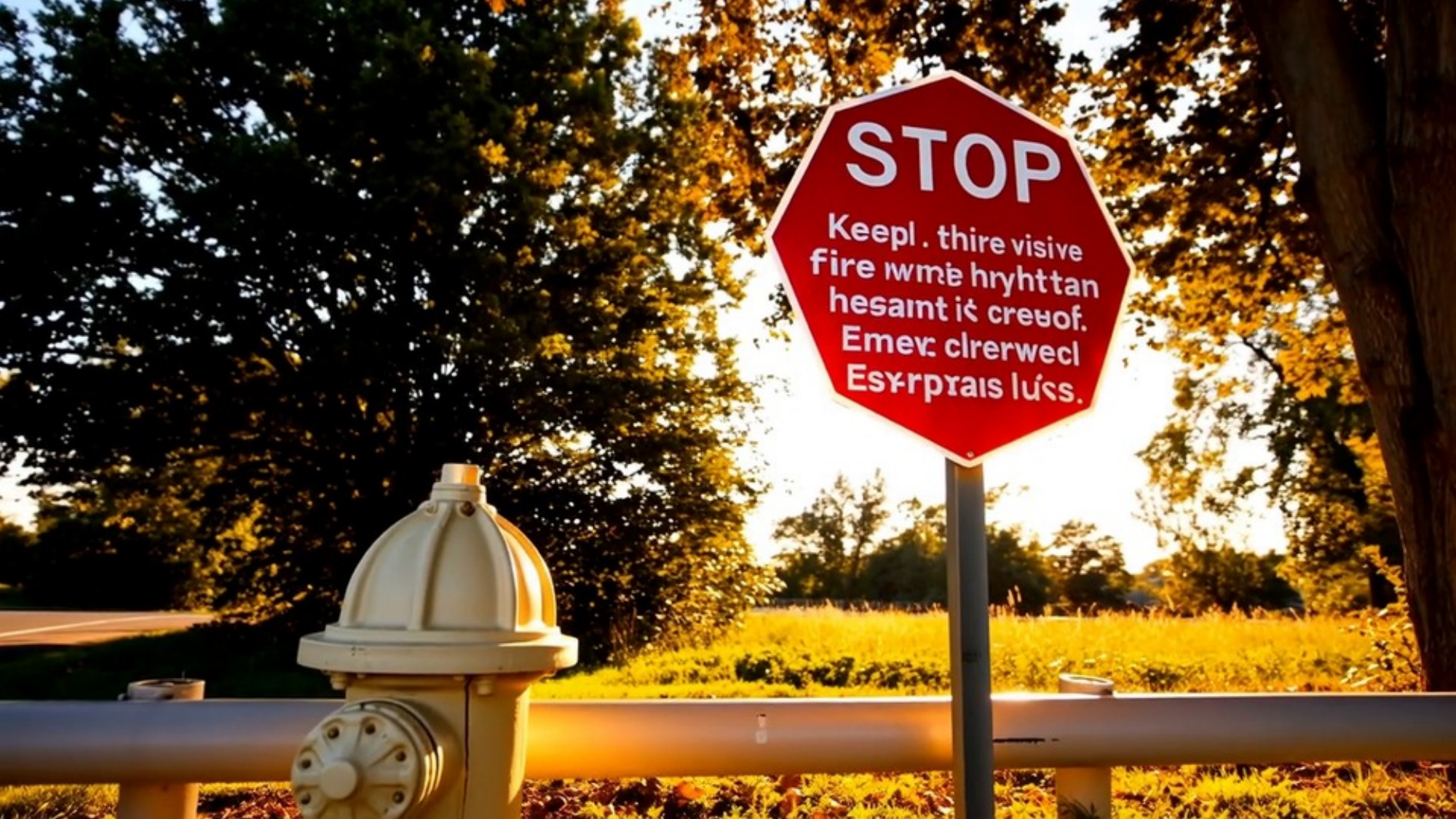}
    \end{tabular}
    \nopagebreak[3]\caption{\textbf{Depth-guided compilation improves camera control.}
    NeurCo (T+O) combines depth guidance with text adaptation, satisfying
    $5/6$ requirements on Wan; the long sign message remains illegible.}
    \label{fig:case_proxy_hydrant}
    \endgroup
\par\endgroup
\end{appcaseblock}

\clearpage
\begin{appcaseblock}
\noindent\textbf{Contact and rebound.}
Adding a rendered motion reference to targeted rewriting completes the
pen-drop sequence on H3 (Figure~\ref{fig:case_h3_r2v}). Native
satisfies $4/6$ requirements and text rewriting alone reaches $5/6$;
the combined plan reaches $6/6$, reproducing the single rebound and
subsequent rest.

\par\smallskip\noindent\begingroup\centering
    \centering
    \begingroup
    \appcaseinit
    \appcaseinputas{R2V}{The young woman in Image 1 sits at a wooden desk wearing glasses, maintaining a focused
    expression throughout. A sheet of paper lies in front of a stack of documents. She
    writes on the sheet with small, precise finger movements. Her work is interrupted when
    the pen slips from her fingers, falls onto the wooden desktop, bounces once, and
    settles.}{\appinputimg{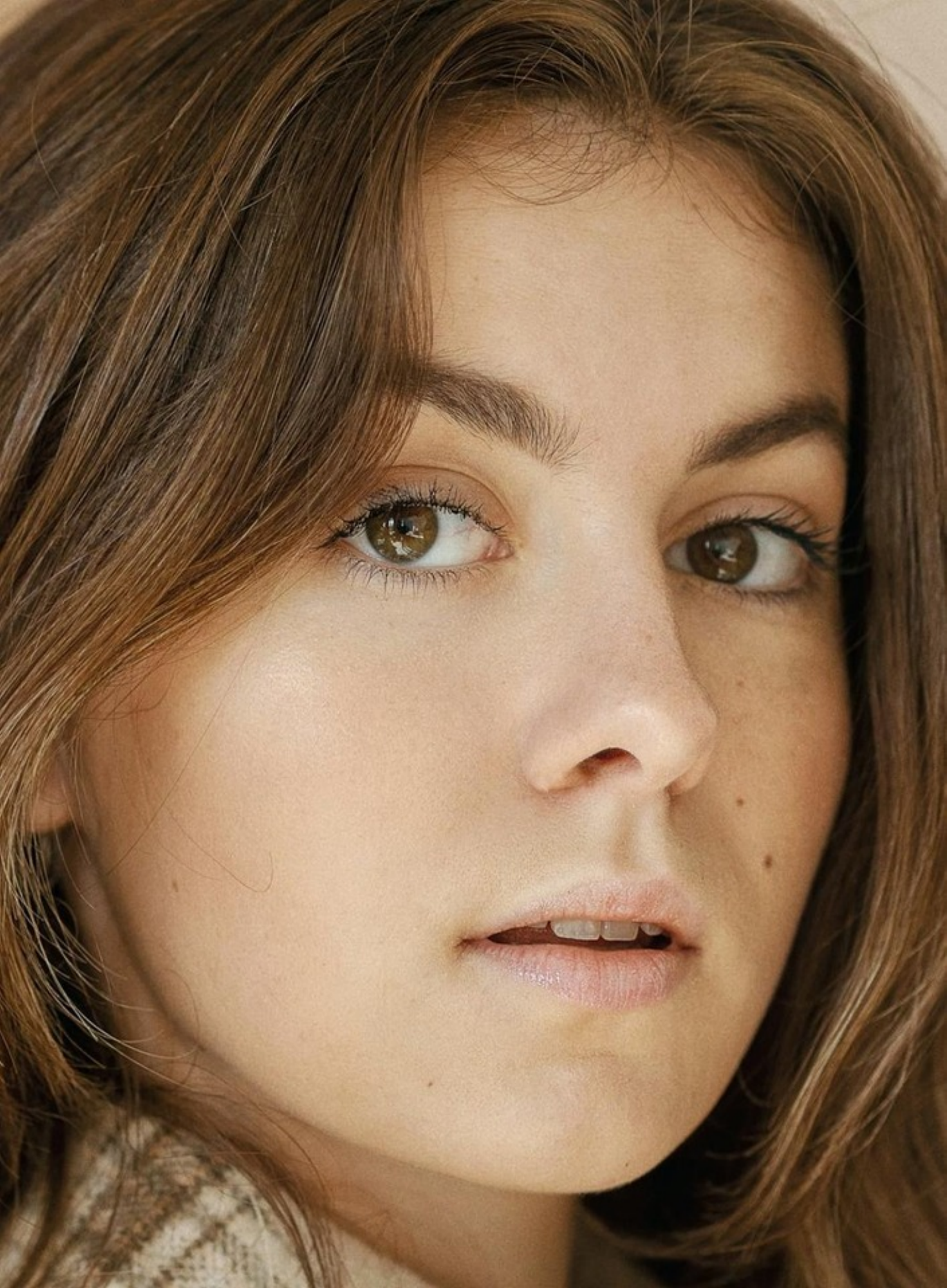}{Image 1}%
    }{\appinputimg{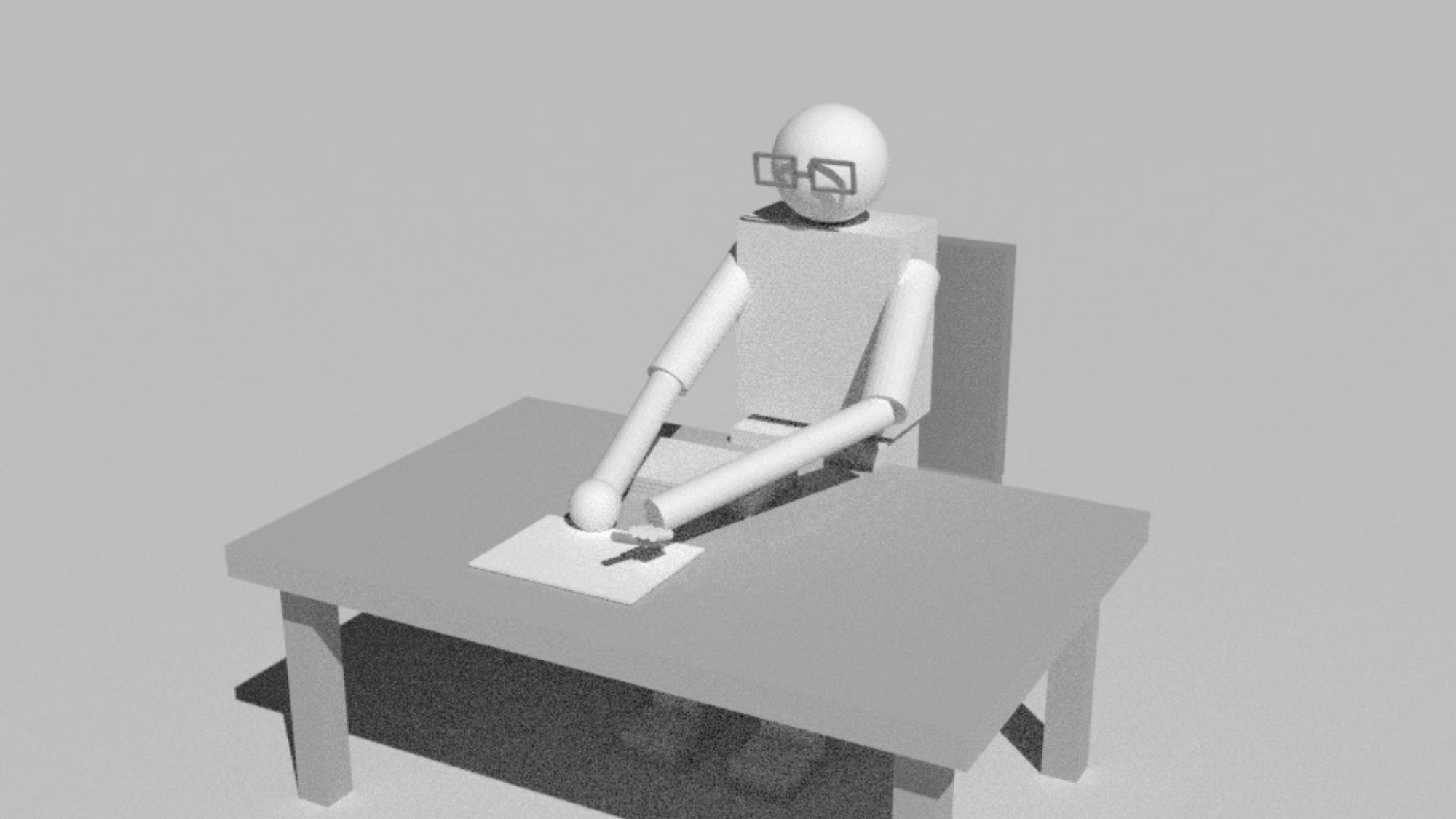}{Proxy video reference (NeurCo, middle frame)}%
    }{%
        \begin{tabularx}{\linewidth}{@{}XXX@{}}
            \textcolor{appBlue}{\textbf{(1) Pen falls, bounces once}} & (2) Focused expression & (3) Sheet in front of stack \\
            (4) Glasses & \textcolor{appBlue}{\textbf{(5) One pen-drop interruption}} & (6) Fine finger motion
        \end{tabularx}}
    \begin{tabular}{@{}c@{\hspace{\appGap}}c@{\hspace{\appGap}}c@{\hspace{\appGap}}c@{}}
        \parbox[c]{\appLabelW}{\centering\textbf{Method}} & \textbf{Early} & \textbf{Middle} & \textbf{Late} \\[2pt]
        \appcaselabel{Native}{\textbf{4/6 passed}}{\textcolor{appFail}{Fails (1), (5)}}
        & \appcaseframe{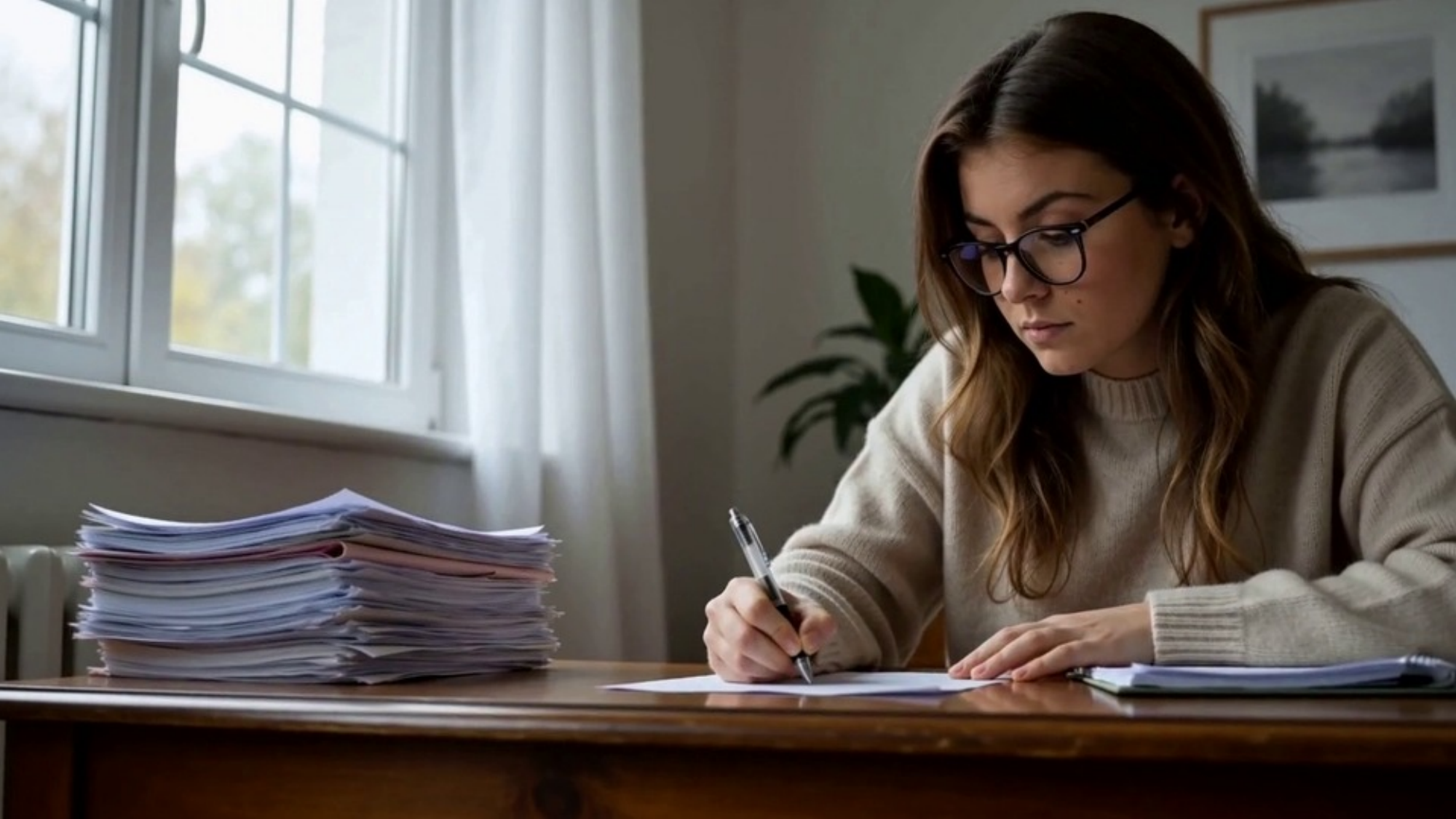}
        & \appcaseframe{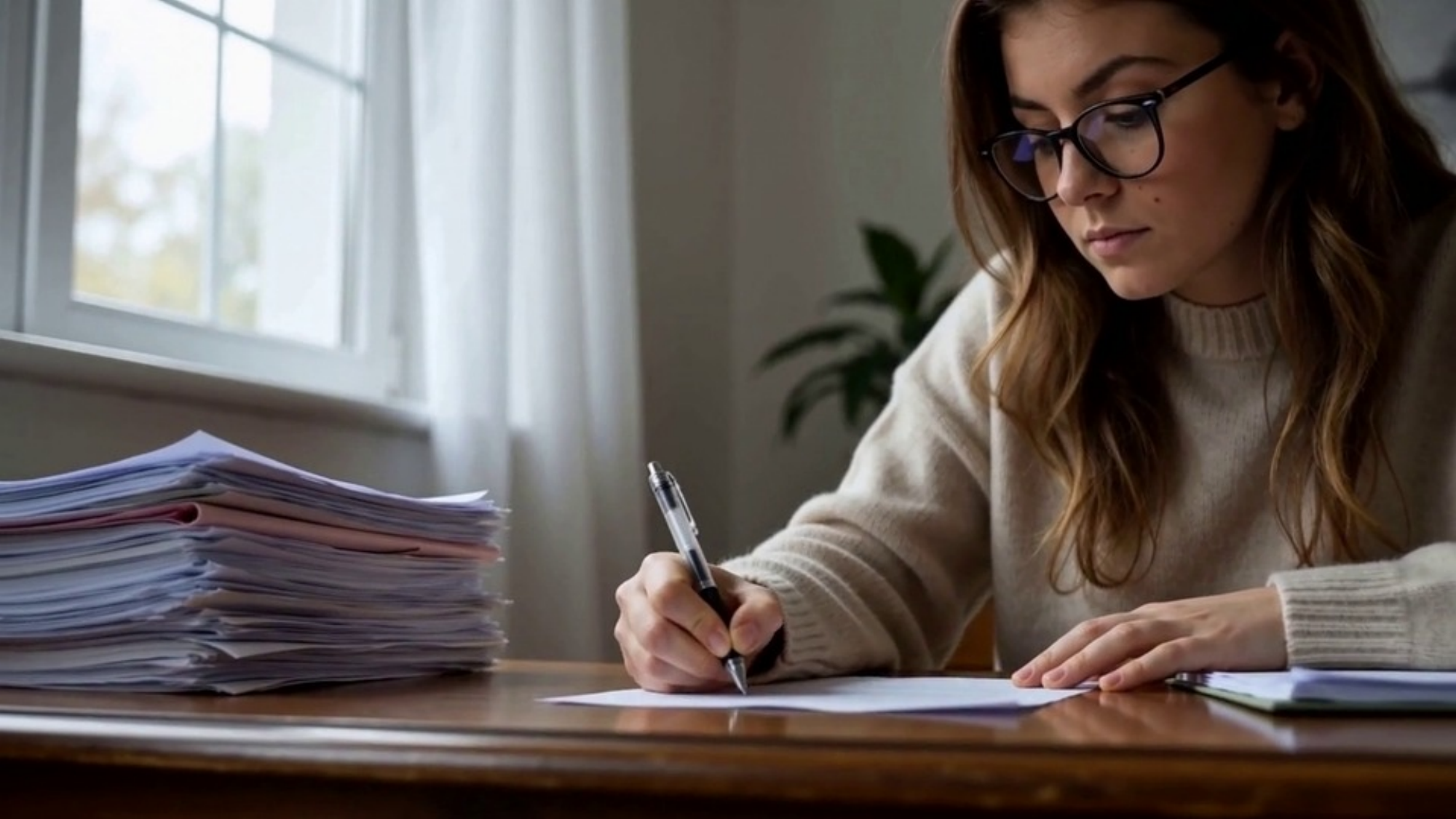}
        & \appcaseframe{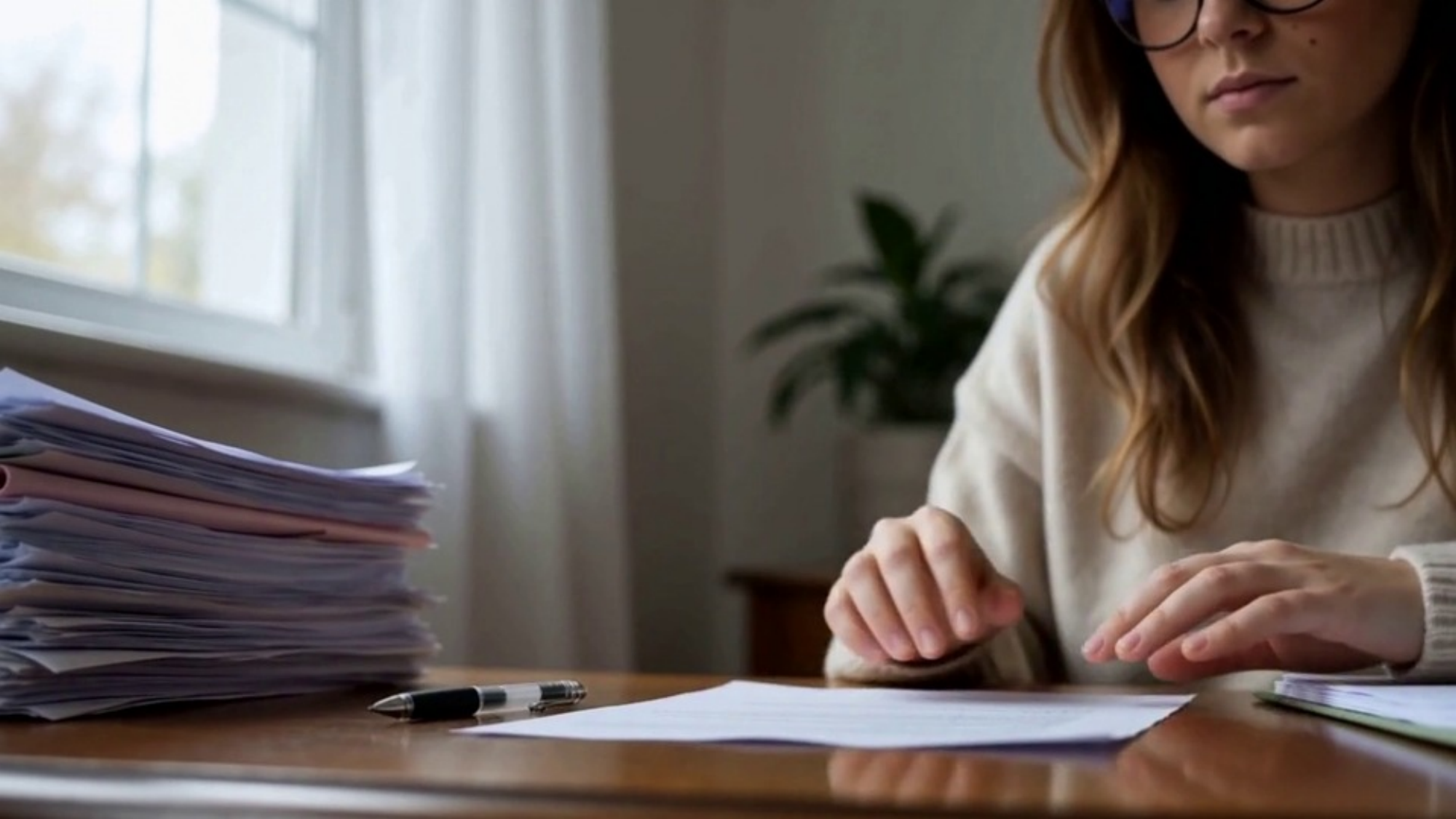}
        \end{tabular}
\par\vspace{3pt}\noindent\begin{tabular}{@{}c@{\hspace{\appGap}}c@{\hspace{\appGap}}c@{\hspace{\appGap}}c@{}}

        \appcaselabel{NeurCo (T)}{\textbf{5/6 passed}}{\textcolor{appFail}{(1) unclear}}
        & \appcaseframe{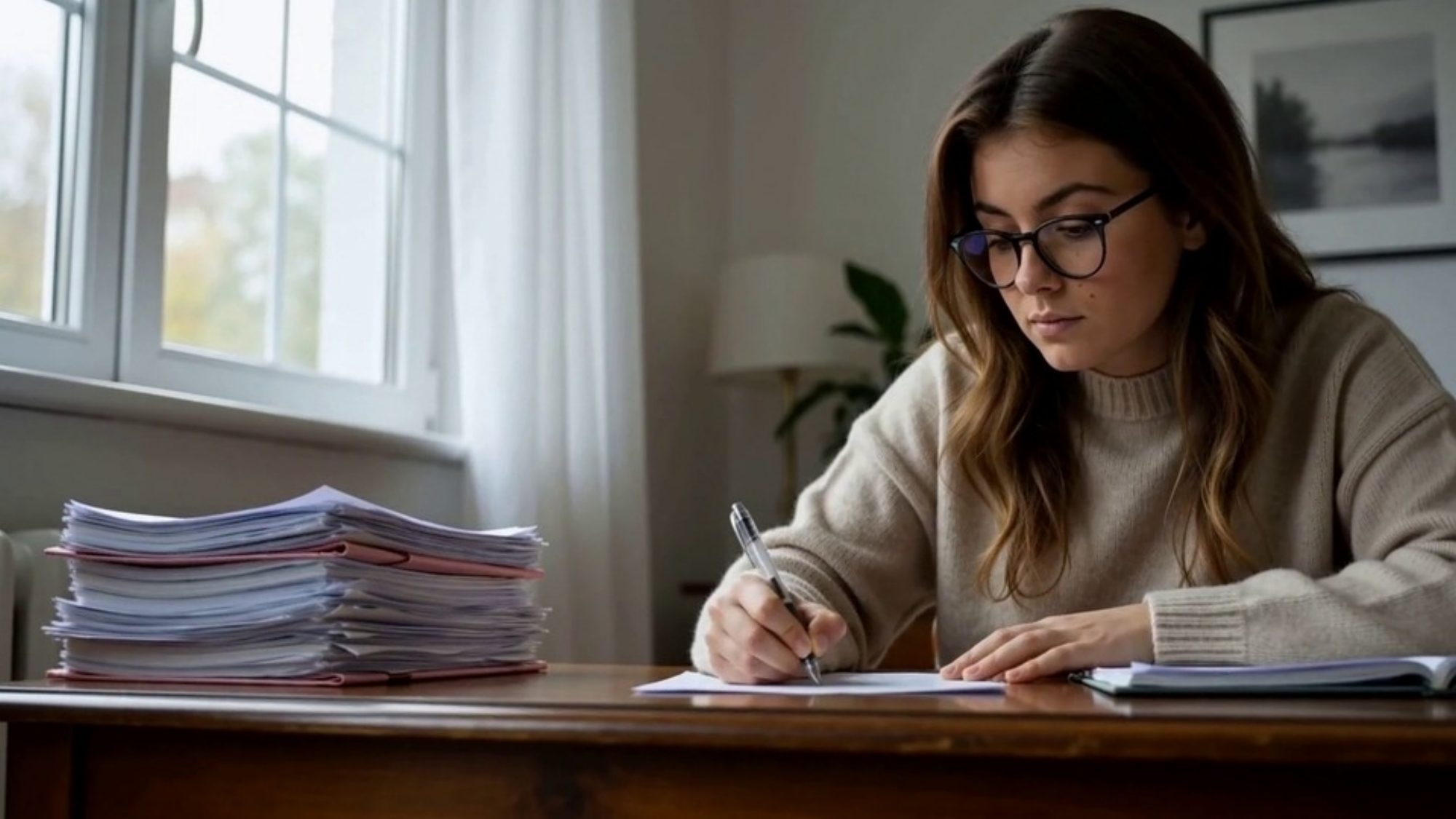}
        & \appcaseframe{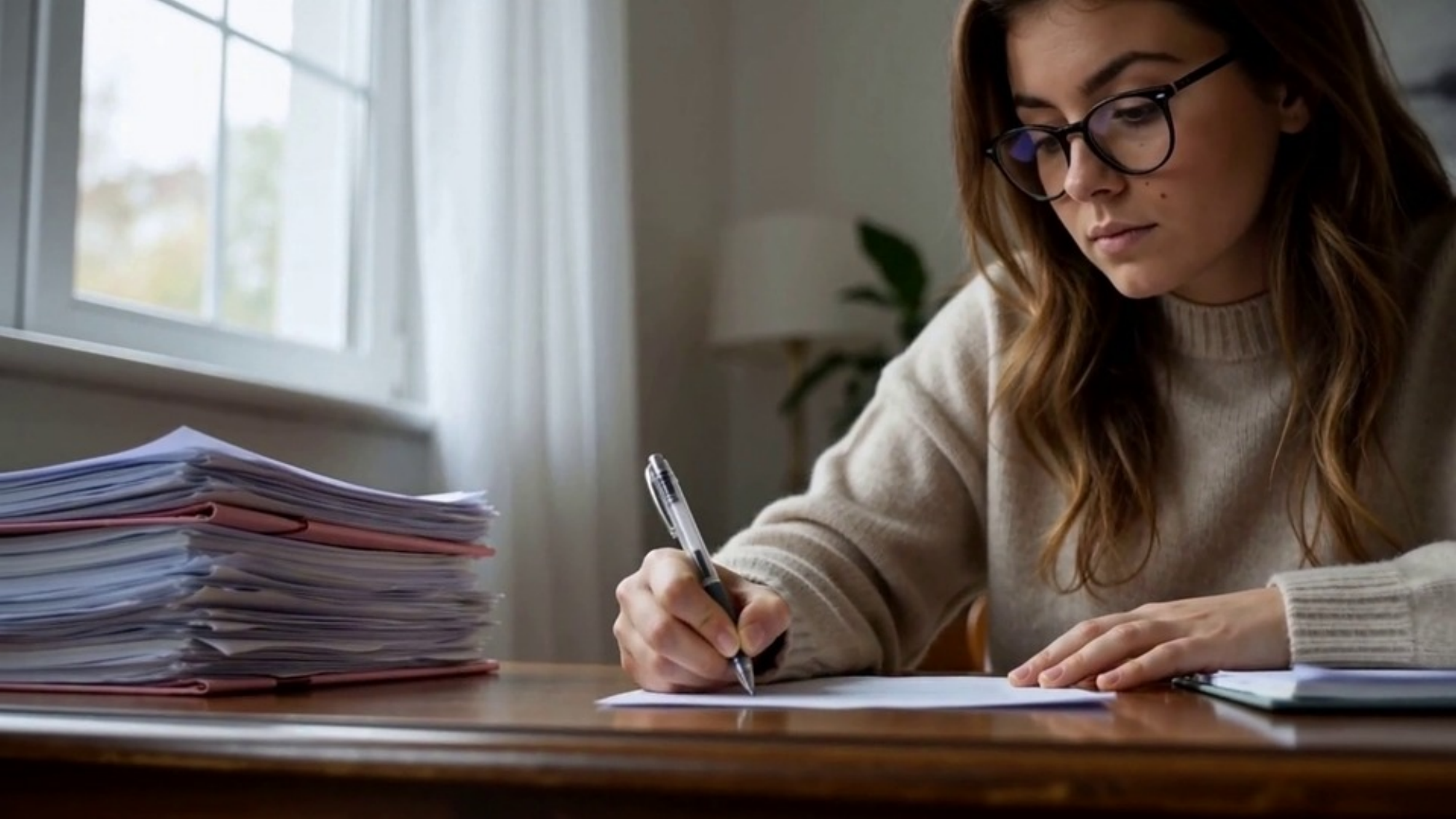}
        & \appcaseframe{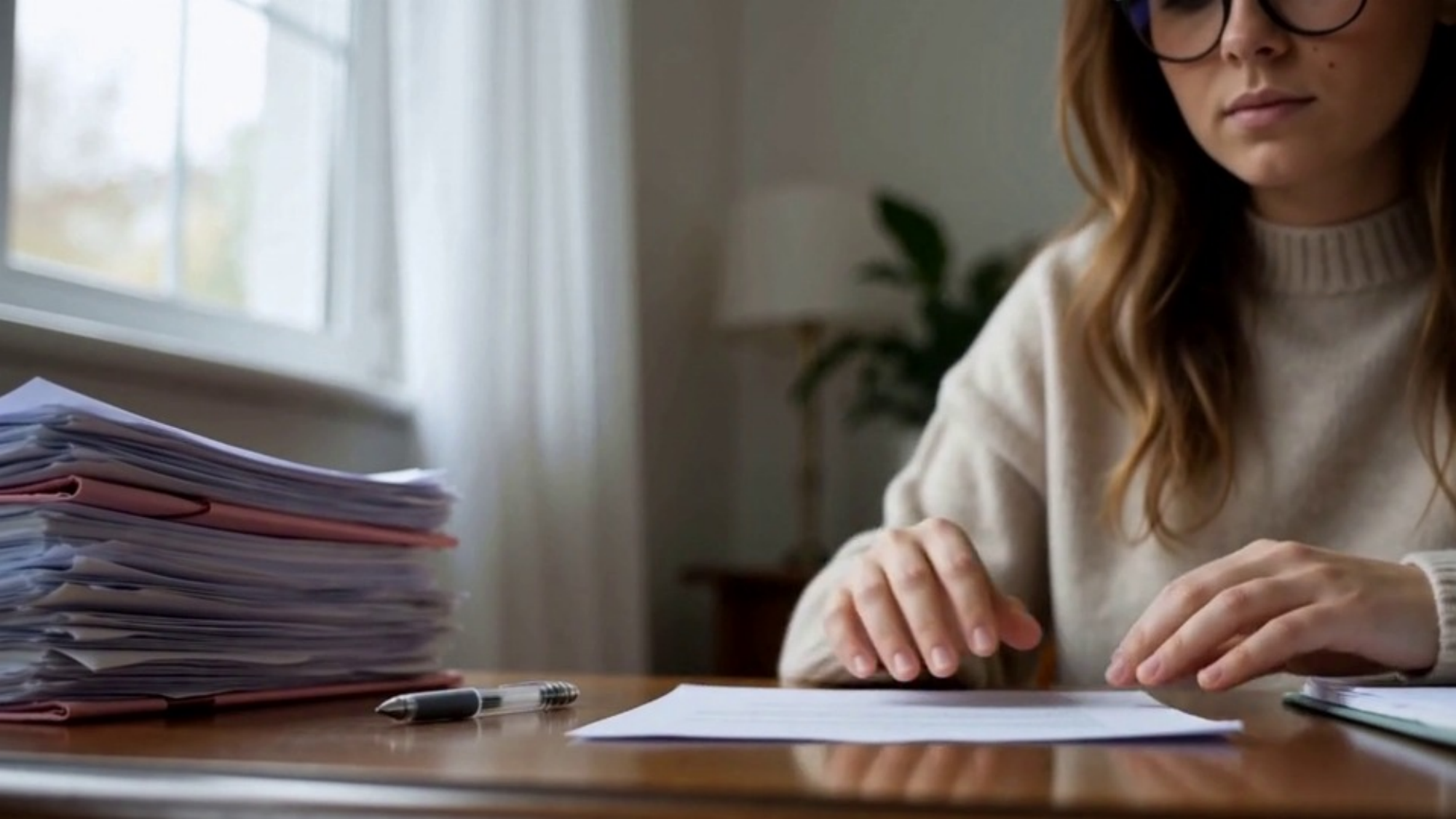}
        \end{tabular}
\par\vspace{3pt}\noindent\begin{tabular}{@{}c@{\hspace{\appGap}}c@{\hspace{\appGap}}c@{\hspace{\appGap}}c@{}}

        \appcaselabel{NeurCo (T+O)}{\textcolor{appBlue}{\textbf{6/6 passed}}}{Video reference + text}
        & \appcaseframe{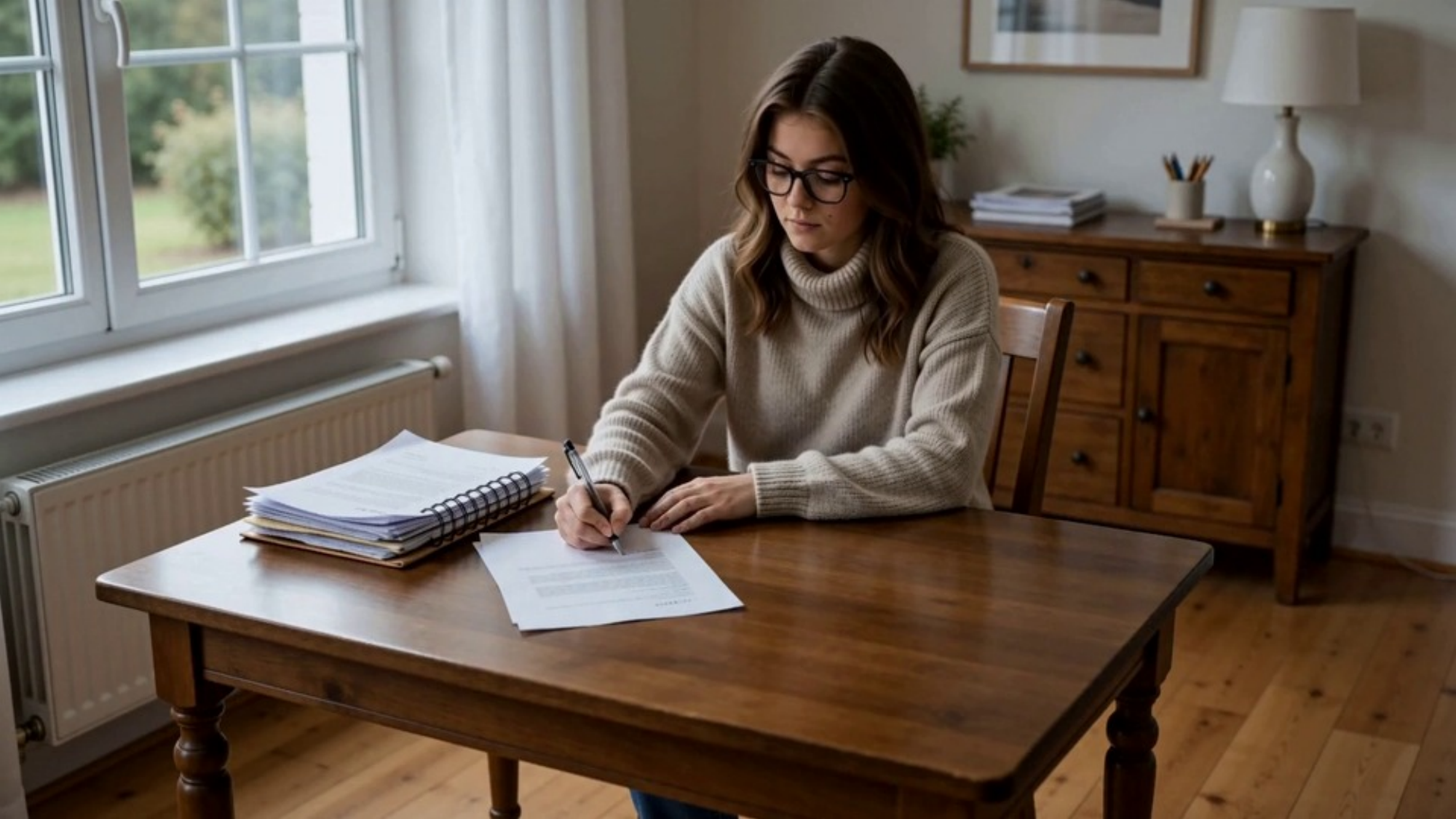}
        & \appcaseframe{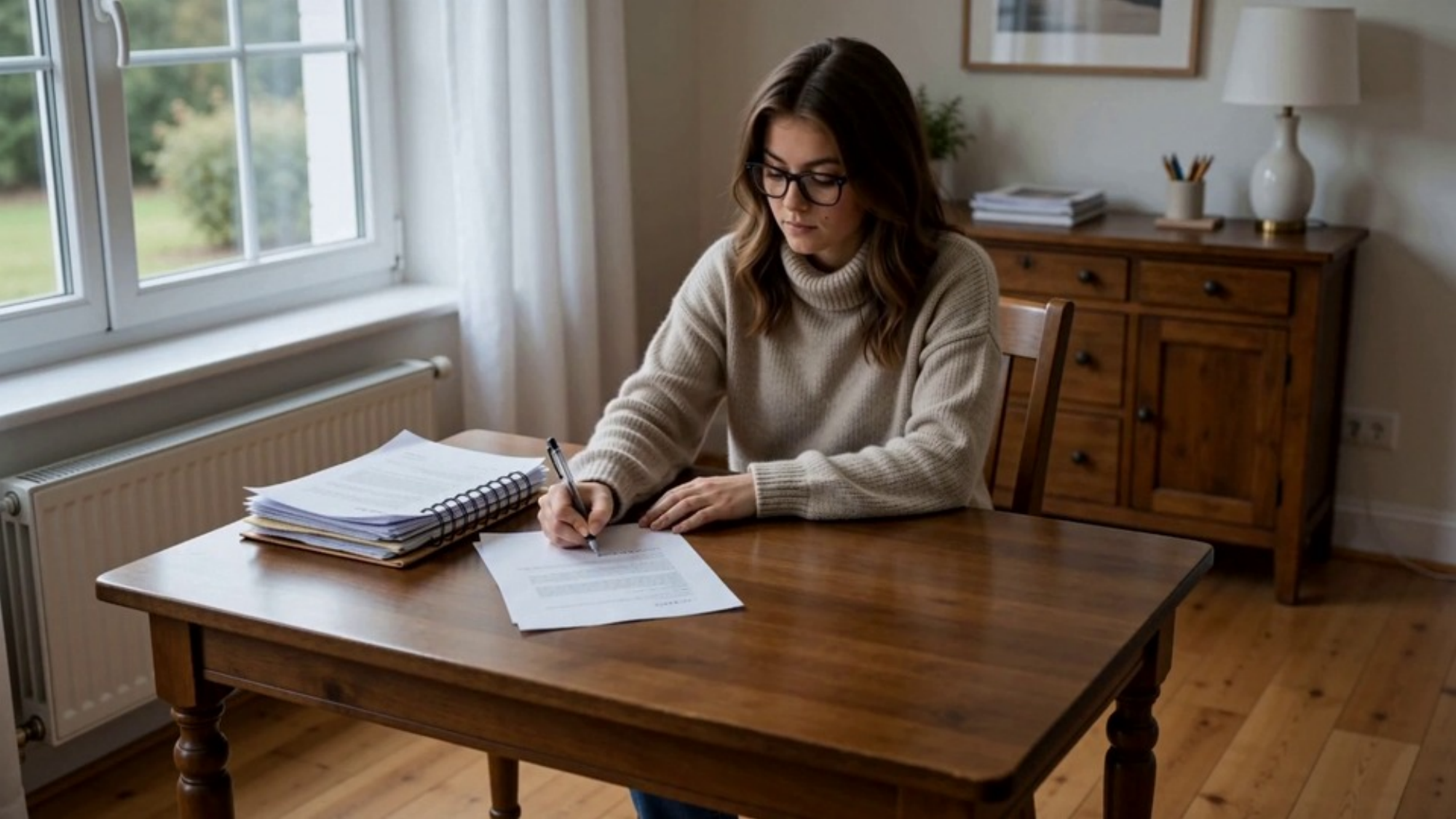}
        & \appcaseframe{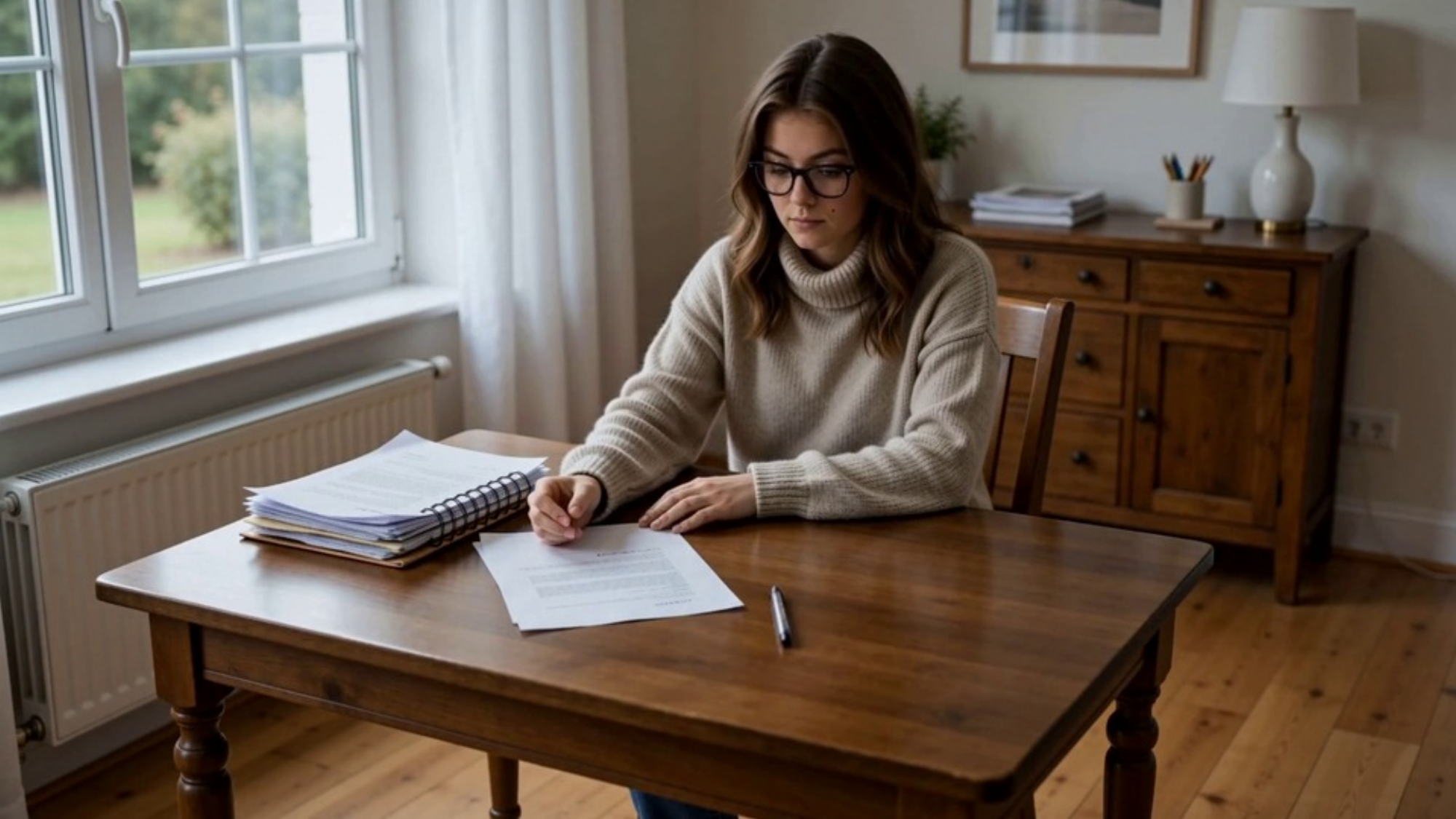}
    \end{tabular}
    \nopagebreak[3]\caption{\textbf{A rendered reference completes the pen-drop sequence.}
    On H3, Native, text-only, and combined compilation reach $4/6$,
    $5/6$, and $6/6$ requirements, respectively.}
    \label{fig:case_h3_r2v}
    \endgroup
\par\endgroup
\end{appcaseblock}

\clearpage
\begin{appcaseblock}
\noindent\textbf{Coupled camera and object motion.}
Rendered proxies recover the lamp rotation and lateral camera motion on
both backends (Figure~\ref{fig:case_proxy_lamp}). Requirement
satisfaction rises from $1/3$ to $3/3$ on H3 and from $0/3$ to $3/3$
on Wan, while text-only compilation reaches $1/3$ on each. Both
proxy-conditioned videos reproduce minor objects from the coarse scene;
the H3 result also fails the realism check.

\par\smallskip\noindent\begingroup\centering
    \centering
    \begingroup
    \appcaseinit
    \appcaseinputas{T2V}{In a plain room, a curved-neck floor lamp rotates in place through one full 360-degree
    revolution, returning to its starting orientation. Its base stays planted at the same spot. Meanwhile, the camera
    trucks six meters to the left, maintaining a fixed viewing direction and keeping the lamp visible throughout.}{}{%
        \appinputimg{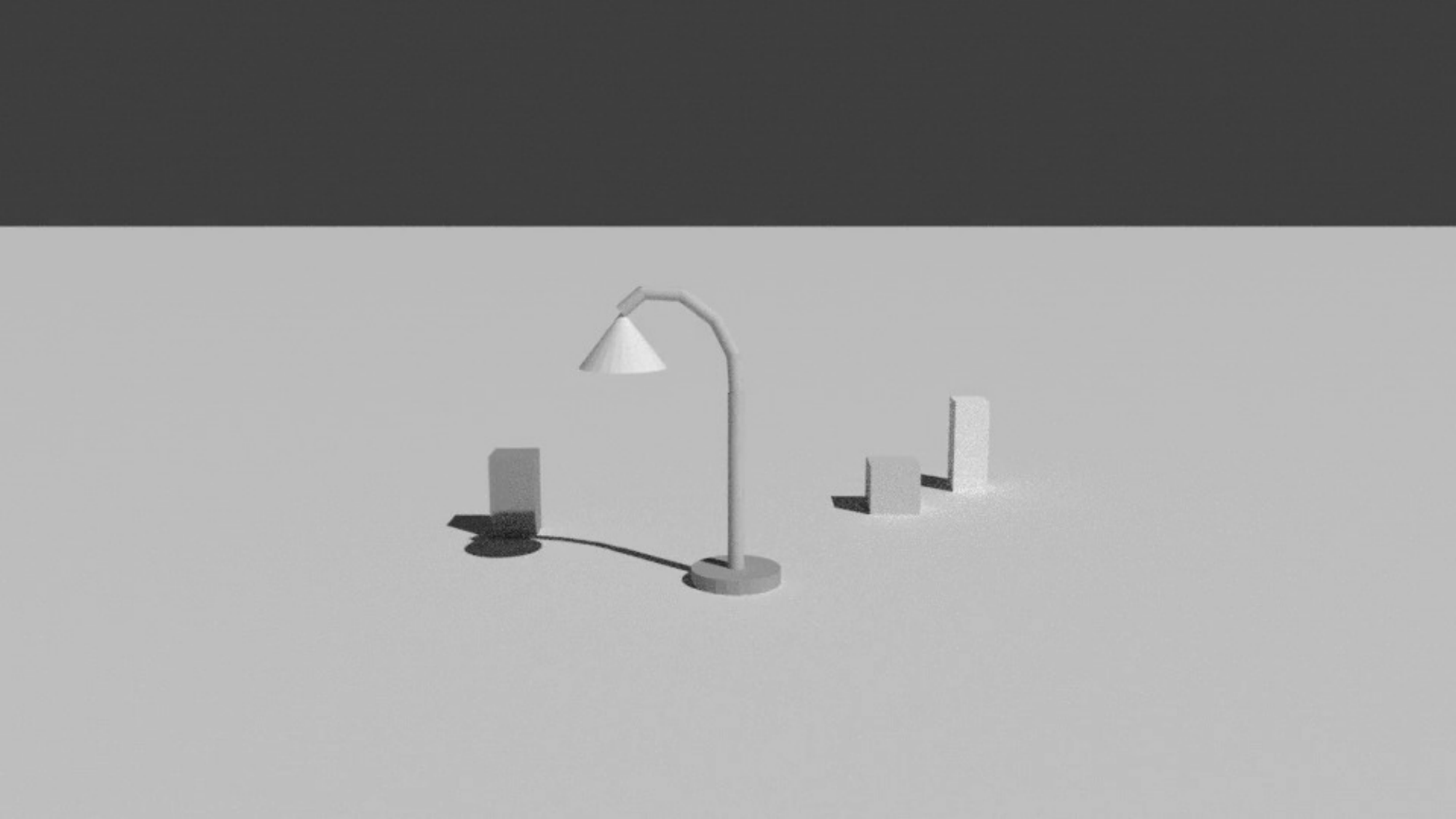}{H3 proxy video reference (middle frame)}%
        \appinputimg{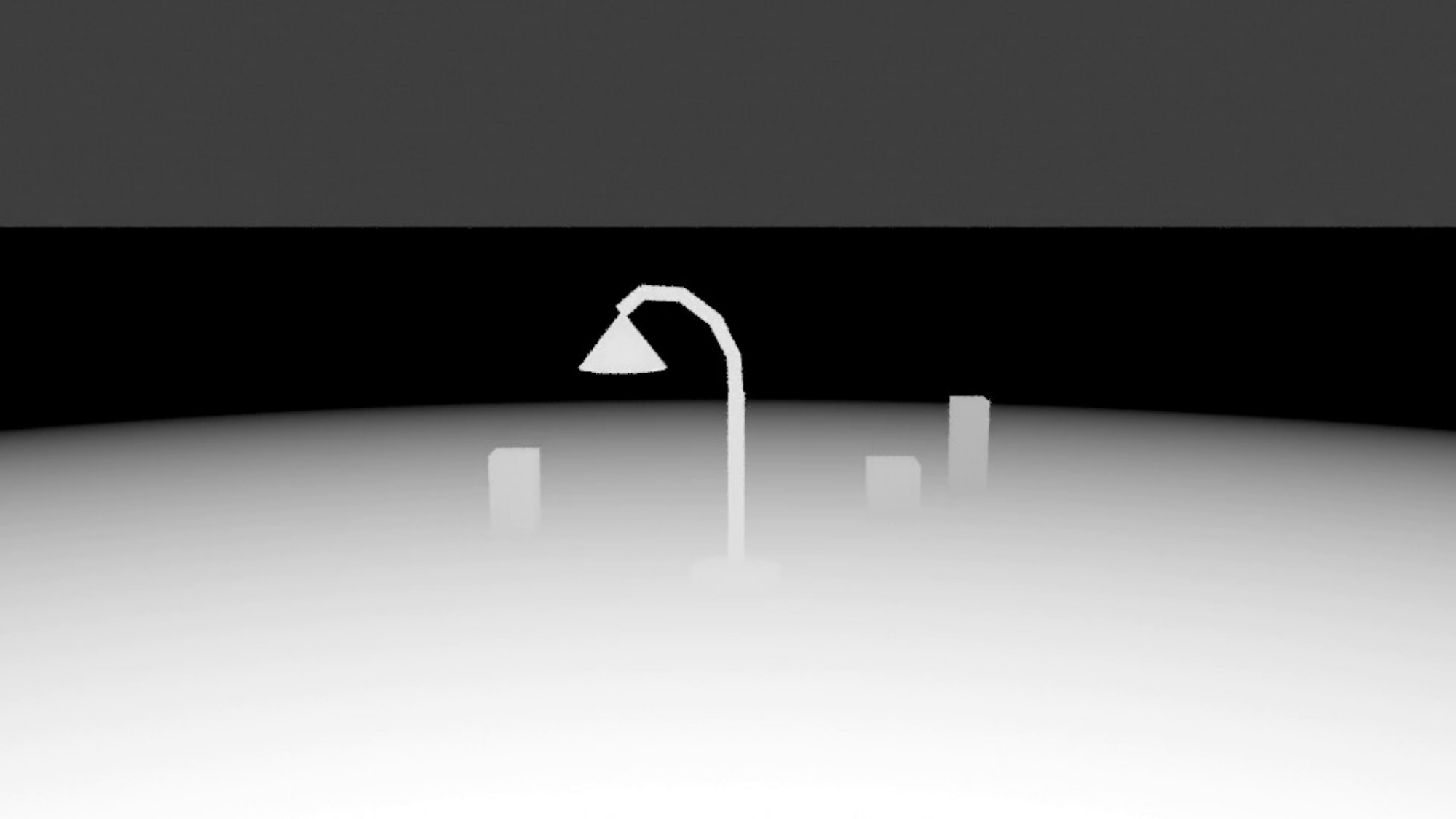}{Wan depth control (middle frame)}%
    }{%
        \begin{tabularx}{\linewidth}{@{}XXX@{}}
            \textcolor{appBlue}{\textbf{(1) Full revolution in place}}
            & \textcolor{appBlue}{\textbf{(2) Broad lateral traverse}}
            & \textcolor{appBlue}{\textbf{(3) Truck left without panning}}
        \end{tabularx}}
    \begin{tabular}{@{}c@{\hspace{\appGap}}c@{\hspace{\appGap}}c@{\hspace{\appGap}}c@{}}
        \parbox[c]{\appLabelW}{\centering\textbf{Method}} & \textbf{Early} & \textbf{Middle} & \textbf{Late} \\[2pt]
        \appcaselabel{H3 Native}{\textbf{1/3 passed}}{\textcolor{appFail}{Fails (1), (2); realism fails}}
        & \appcaseframe{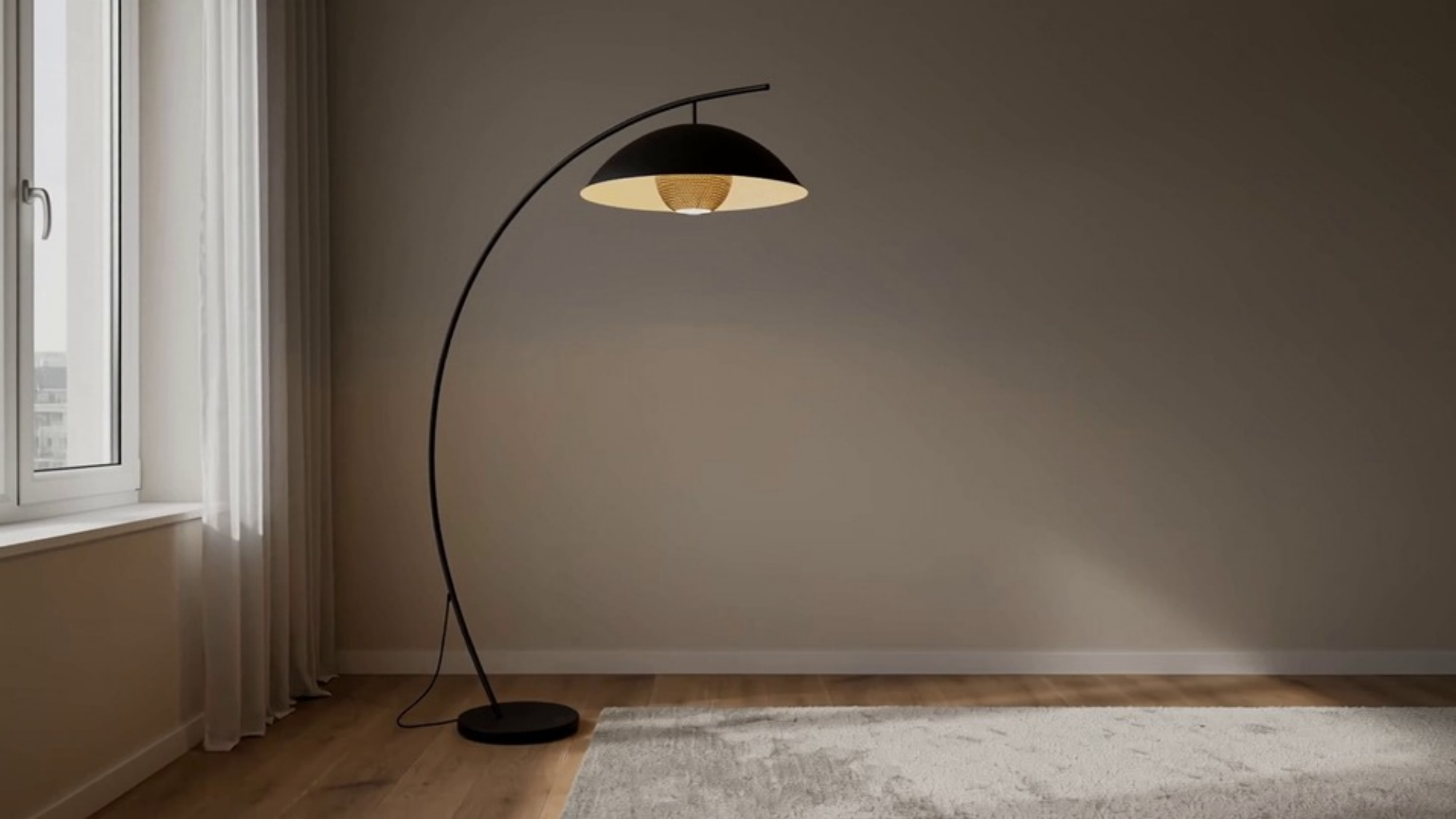}
        & \appcaseframe{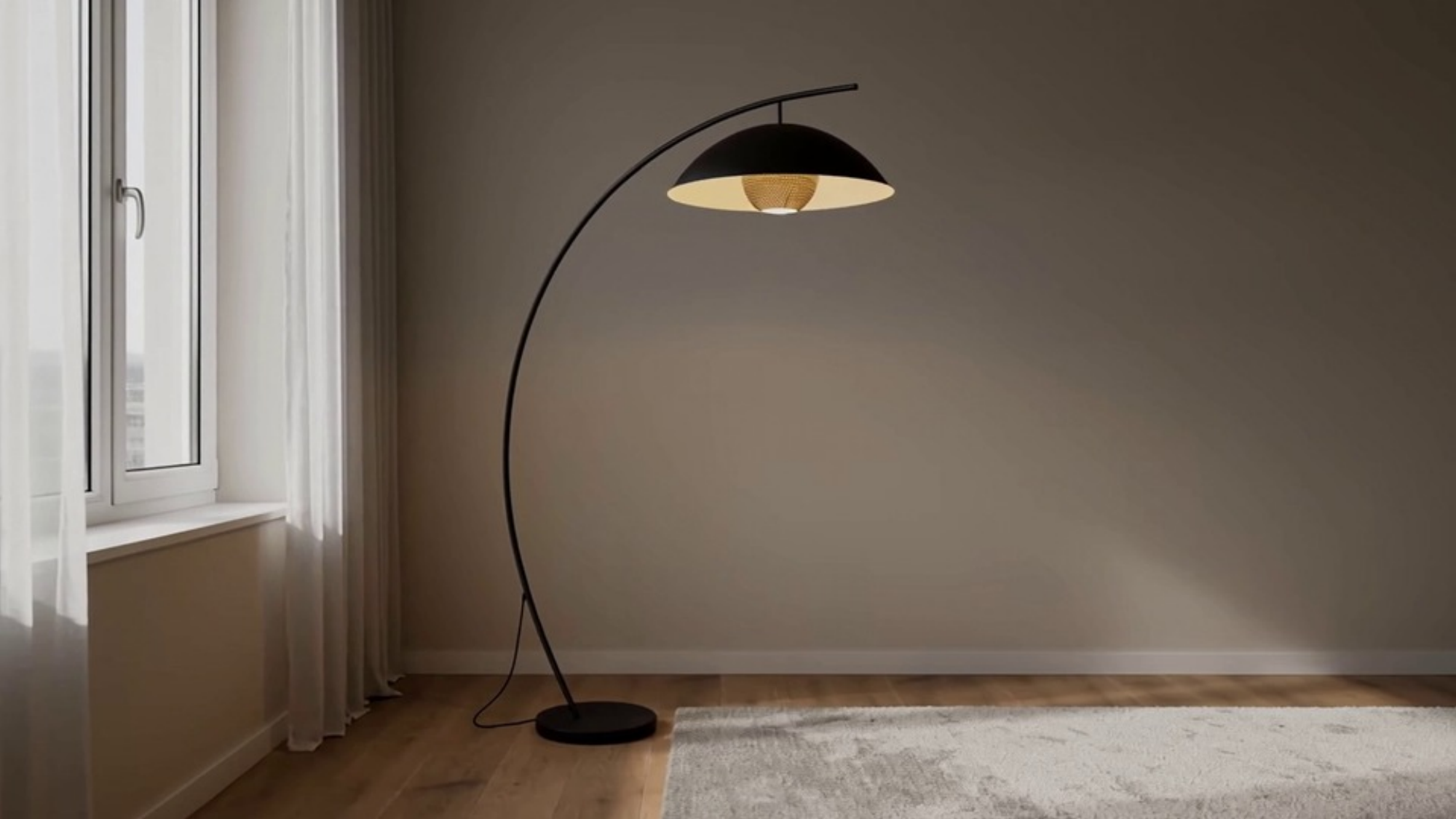}
        & \appcaseframe{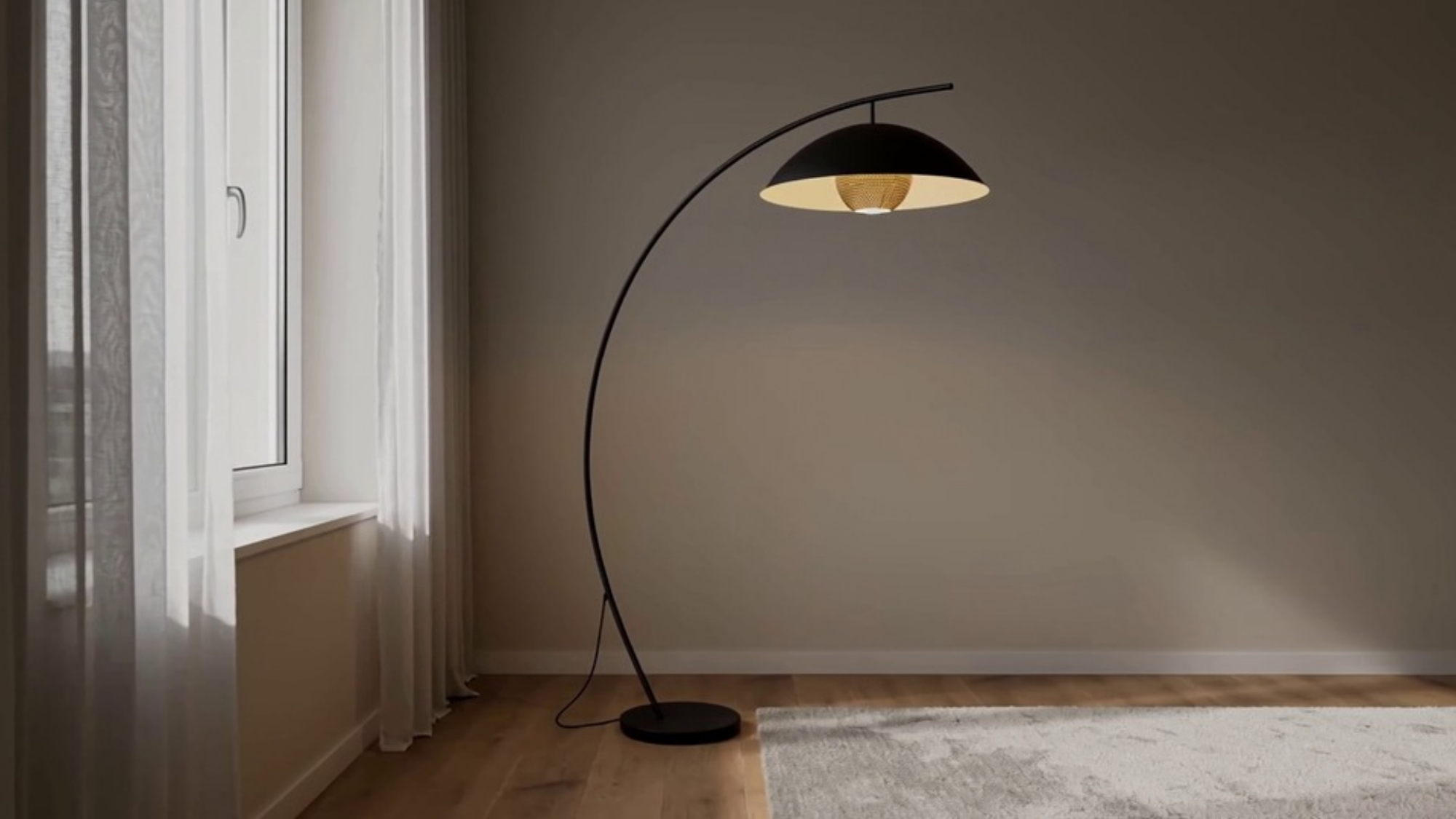}
        \end{tabular}
\par\vspace{3pt}\noindent\begin{tabular}{@{}c@{\hspace{\appGap}}c@{\hspace{\appGap}}c@{\hspace{\appGap}}c@{}}

        \appcaselabel{H3 NeurCo (T)}{\textbf{1/3 passed}}{\textcolor{appFail}{Fails (1), (2)}}
        & \appcaseframe{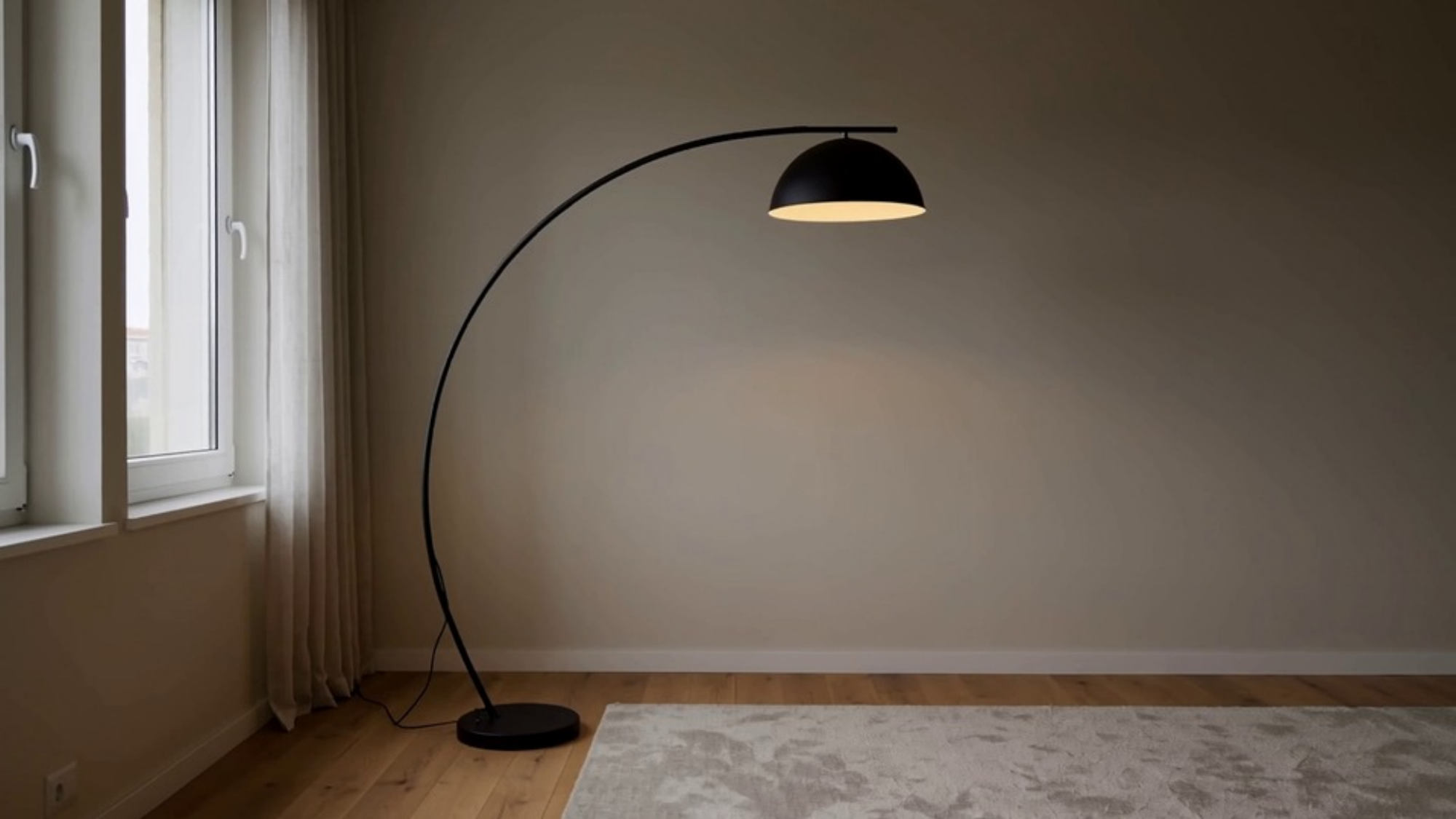}
        & \appcaseframe{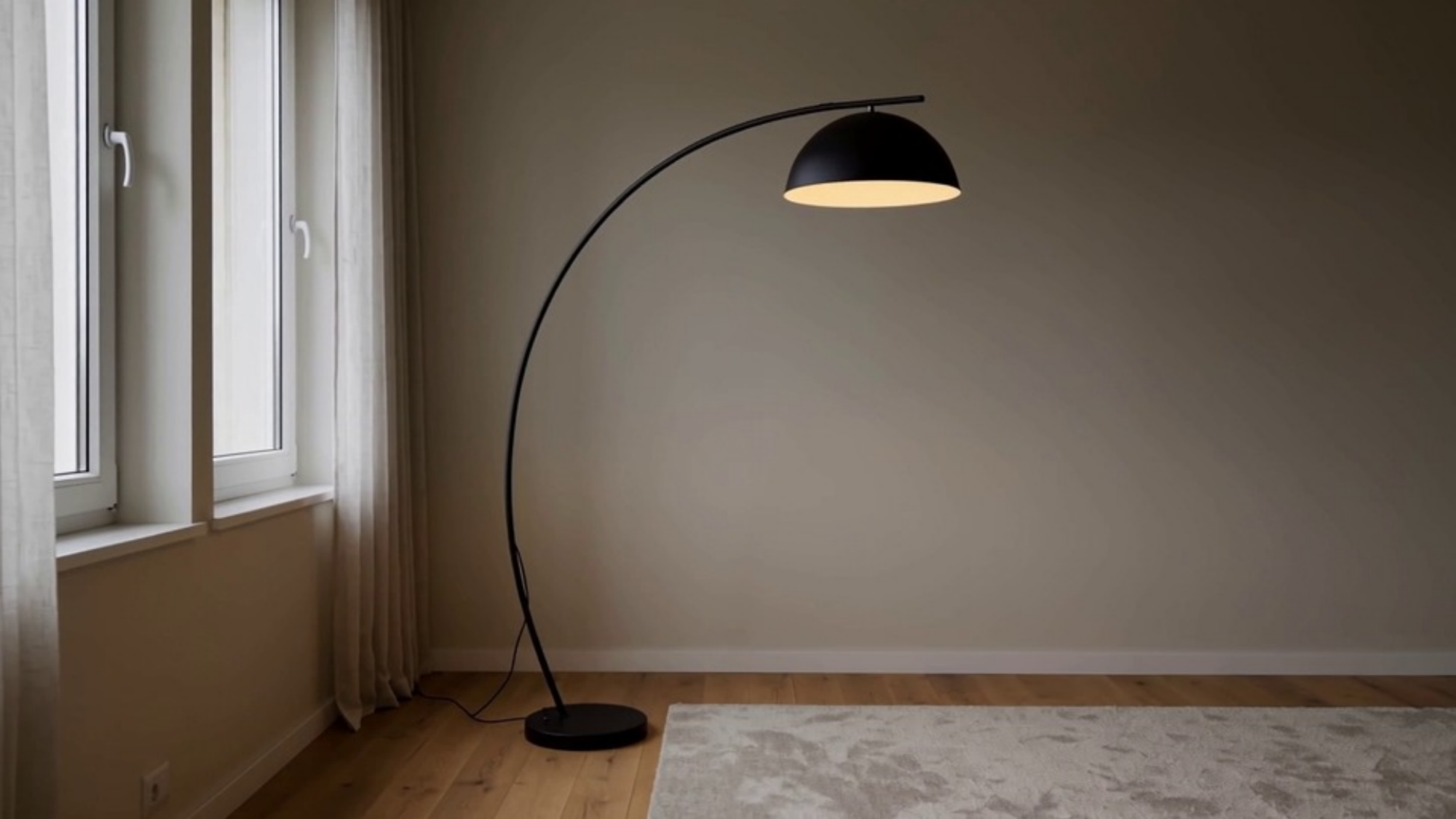}
        & \appcaseframe{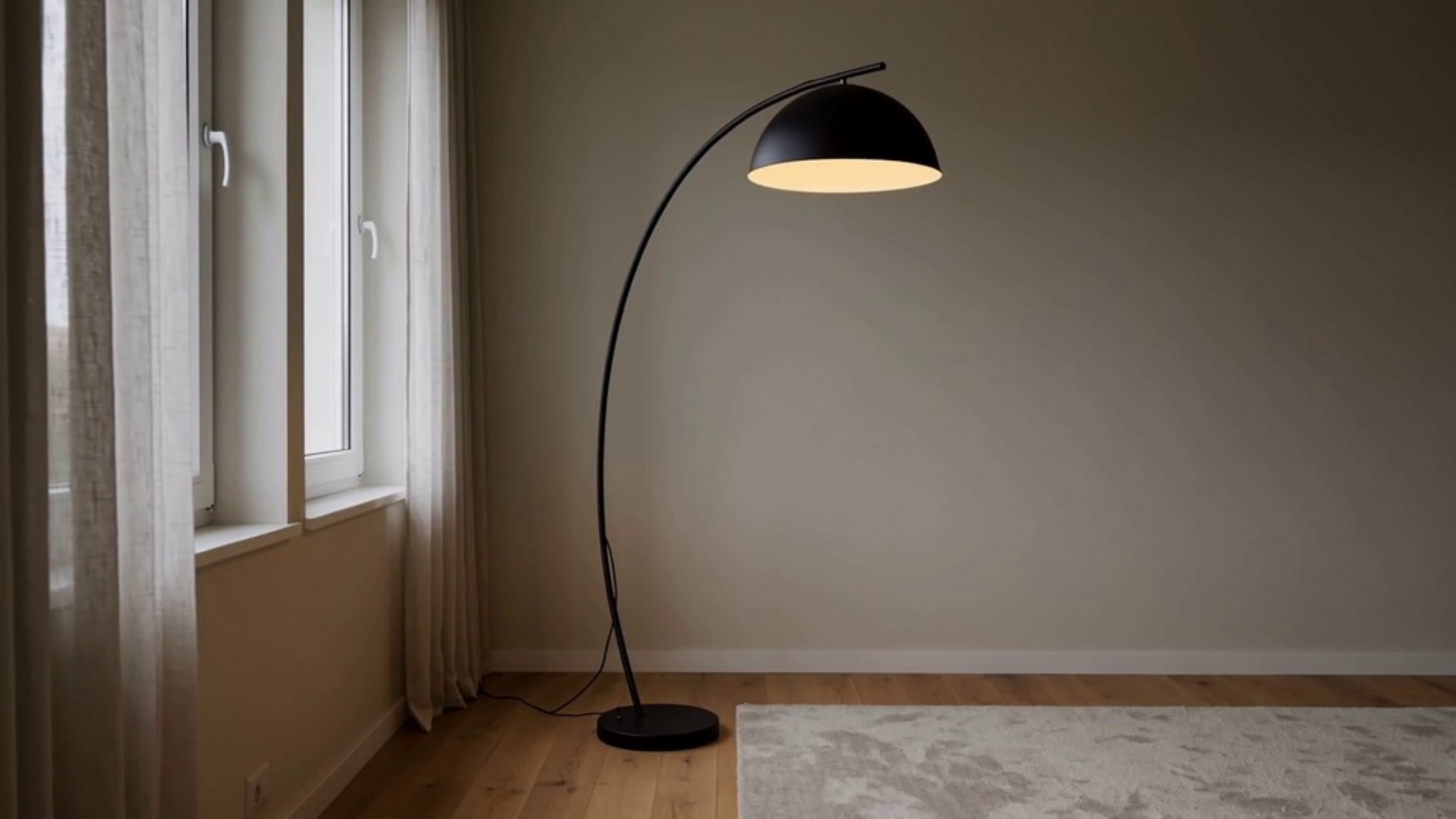}
        \end{tabular}
\par\vspace{3pt}\noindent\begin{tabular}{@{}c@{\hspace{\appGap}}c@{\hspace{\appGap}}c@{\hspace{\appGap}}c@{}}

        \appcaselabel{H3 NeurCo (T+O)}{\textcolor{appBlue}{\textbf{3/3 passed}}}{Video reference only; \textcolor{appFail}{realism fails}}
        & \appcaseframe{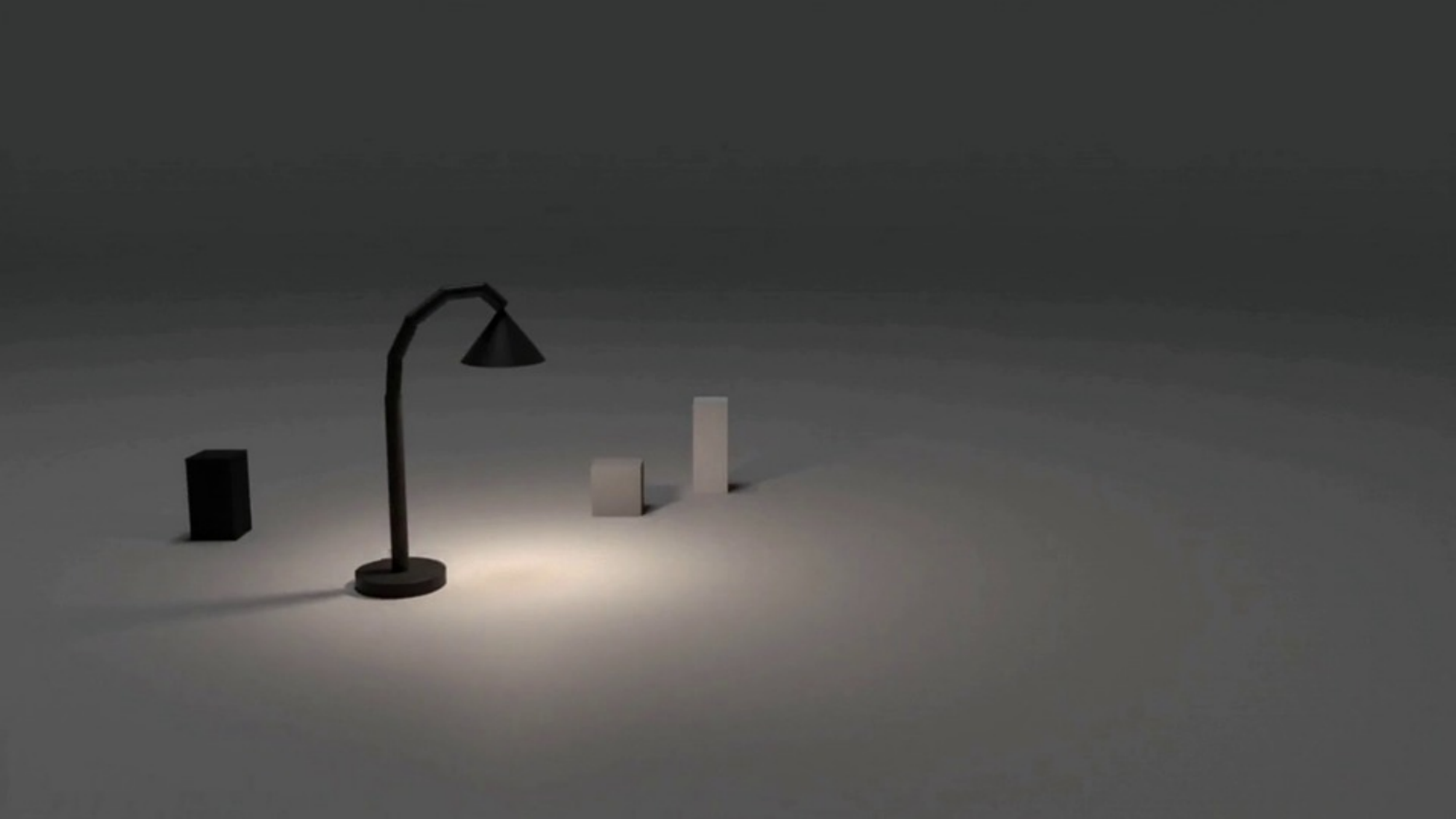}
        & \appcaseframe{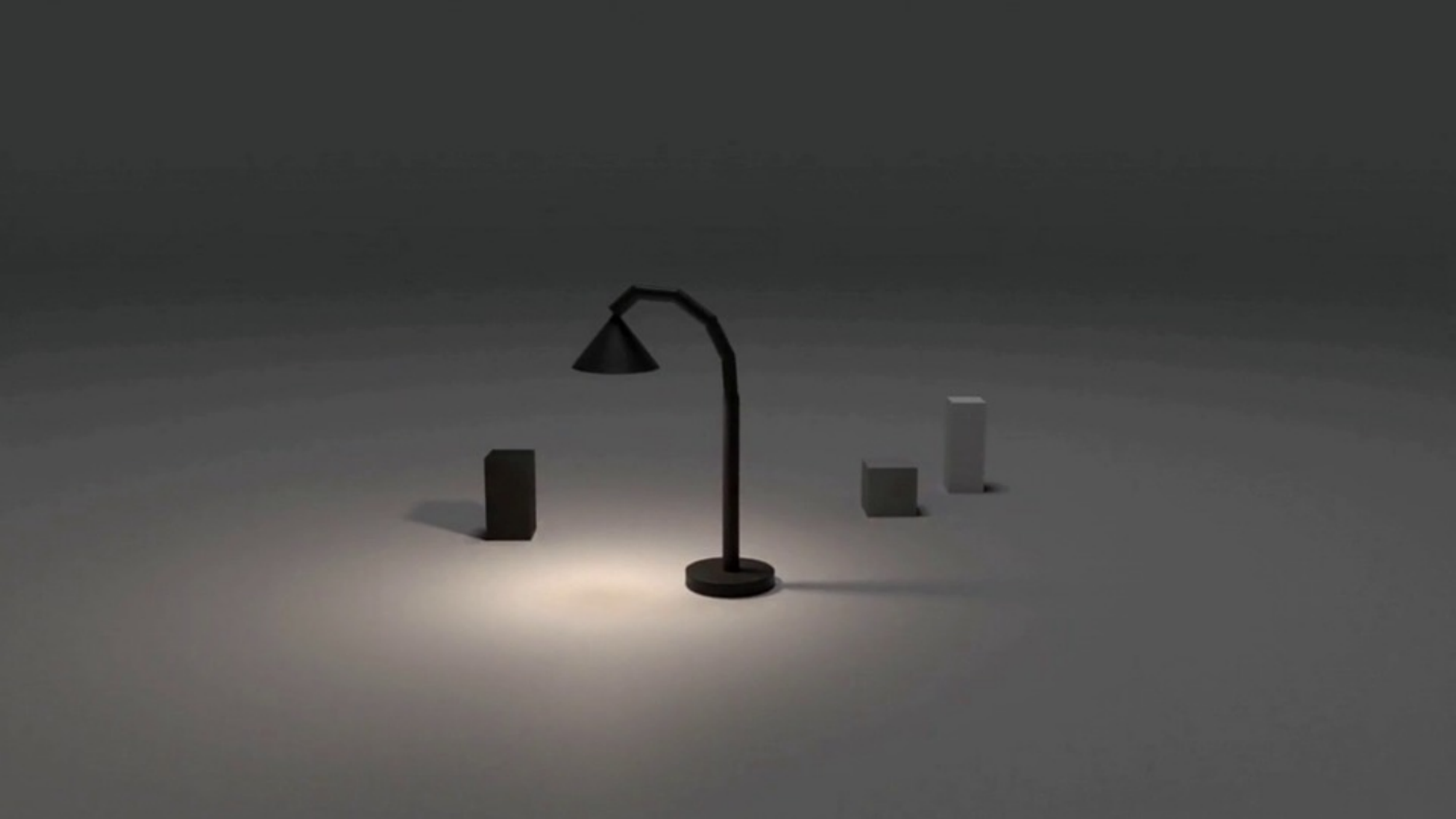}
        & \appcaseframe{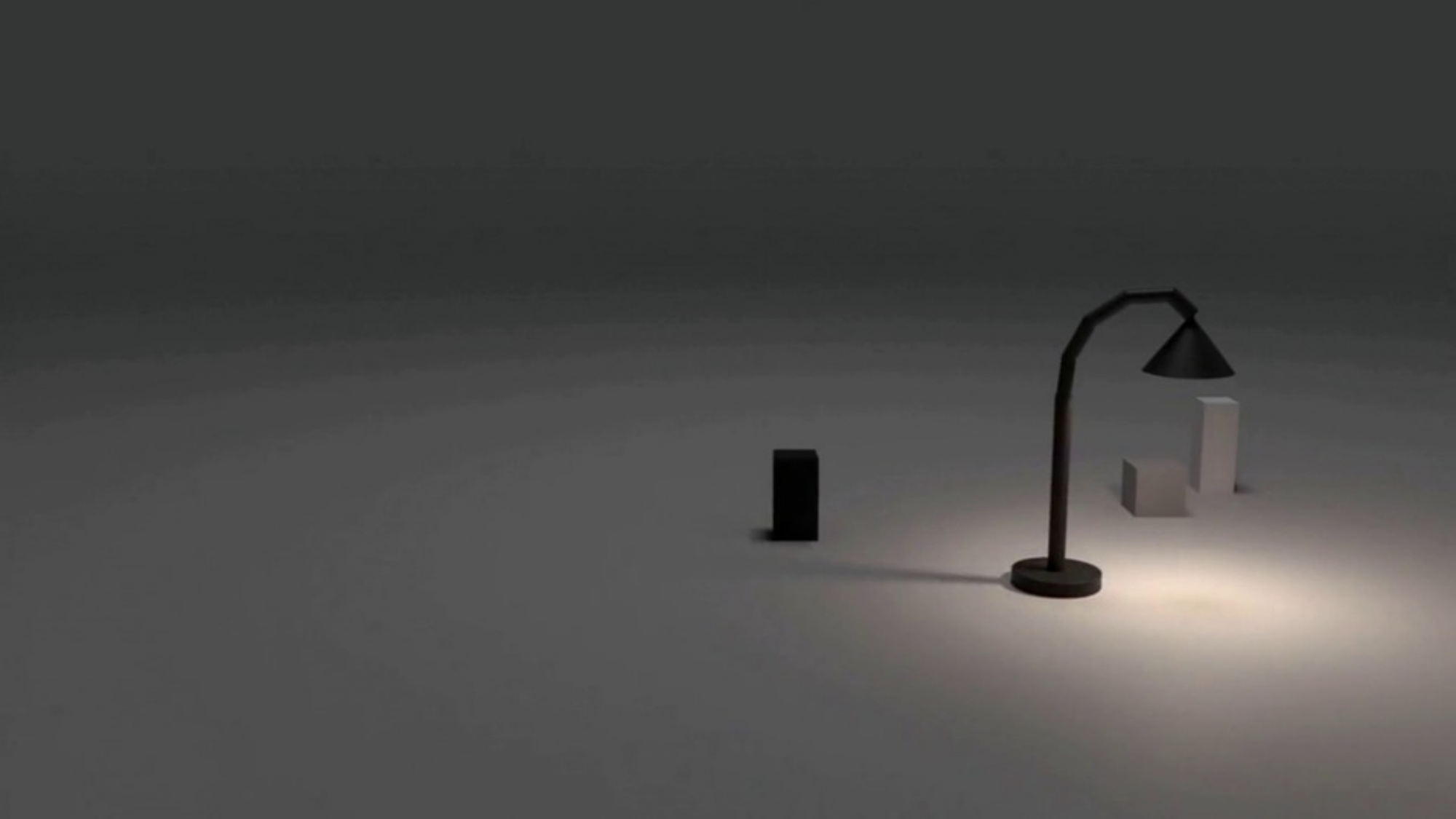}
        \end{tabular}
\par\vspace{3pt}\noindent\begin{tabular}{@{}c@{\hspace{\appGap}}c@{\hspace{\appGap}}c@{\hspace{\appGap}}c@{}}

        \appcaselabel{Wan Native}{\textbf{0/3 passed}}{\textcolor{appFail}{Fails (1)--(3)}}
        & \appcaseframe{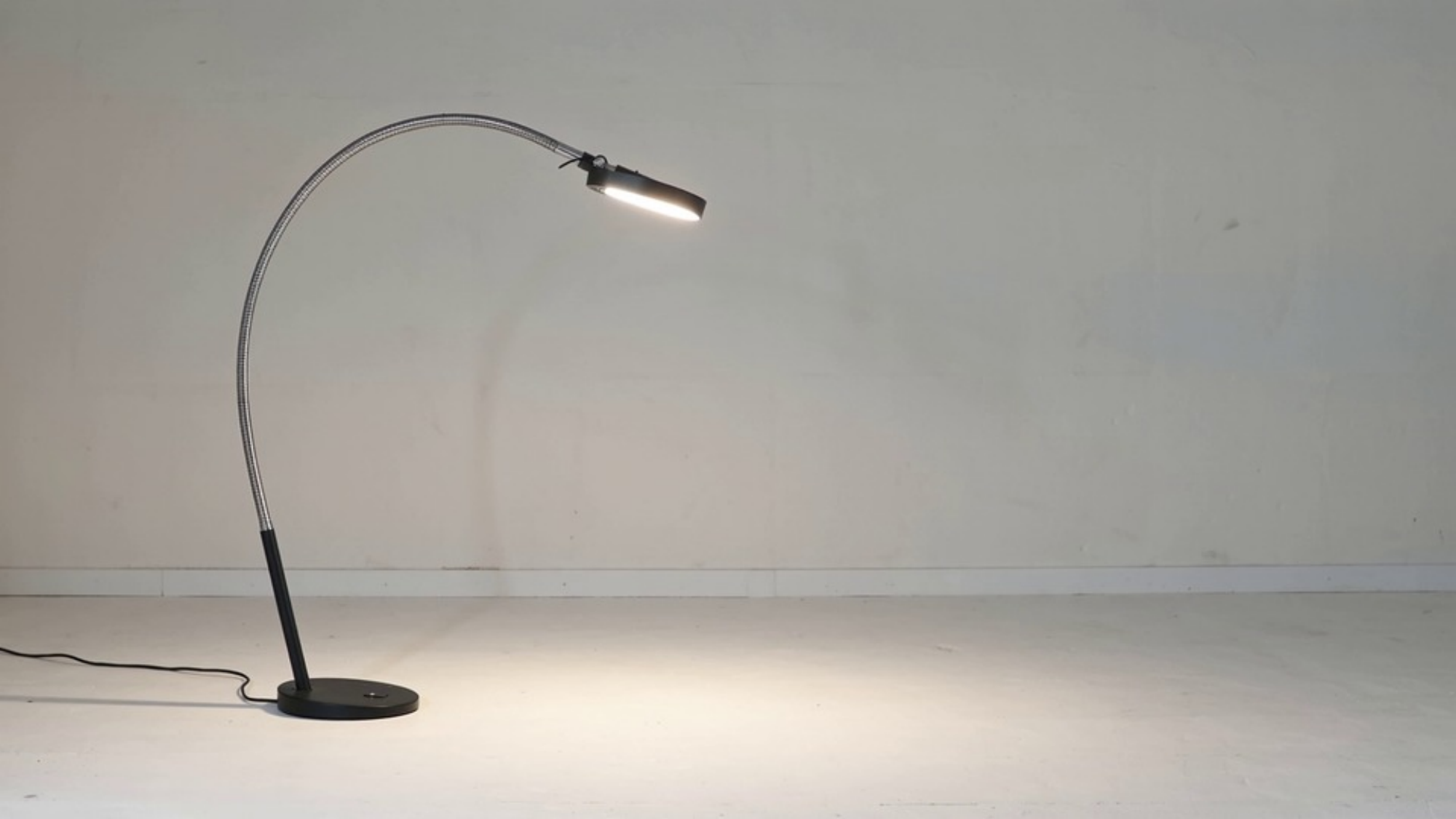}
        & \appcaseframe{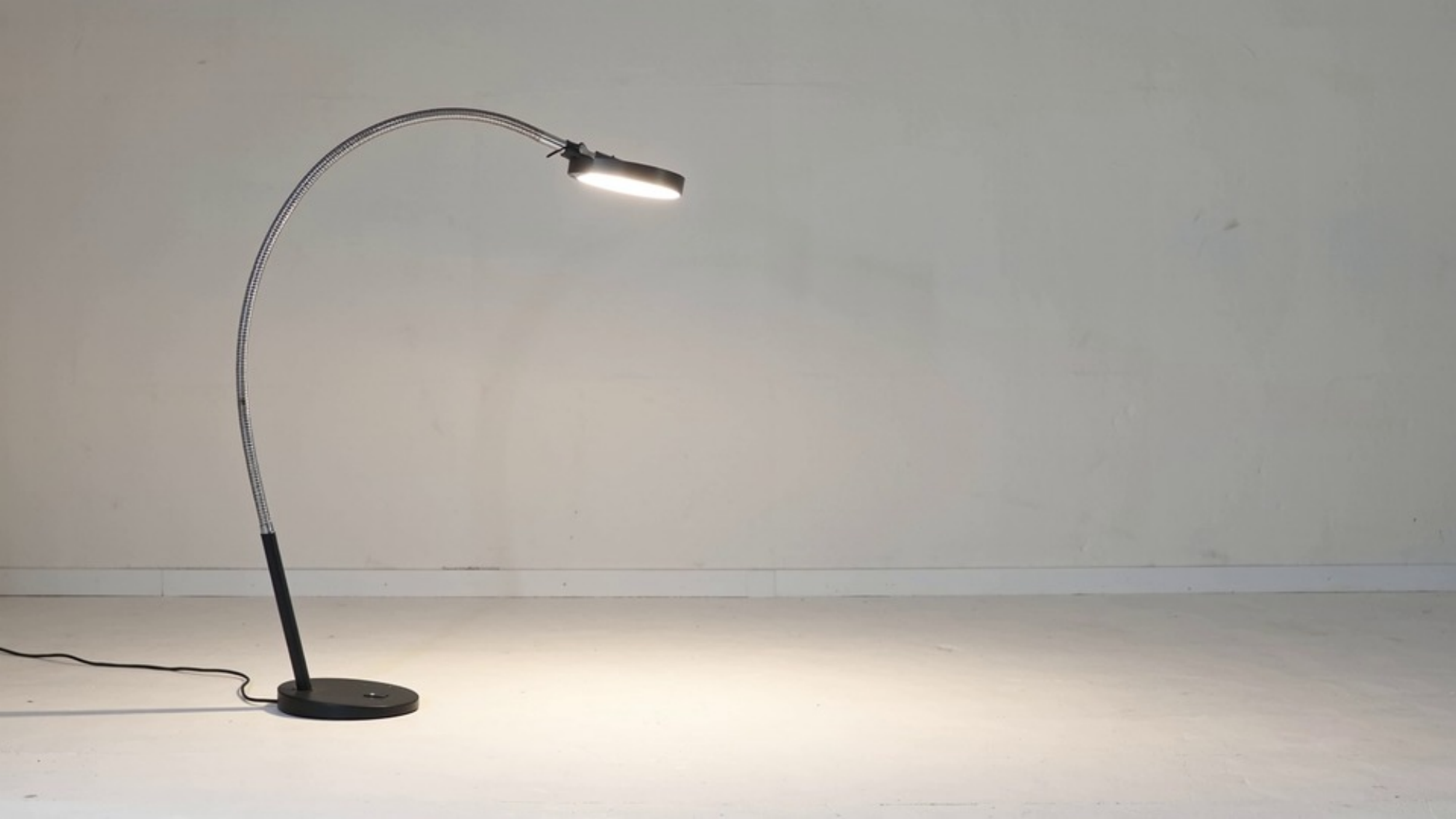}
        & \appcaseframe{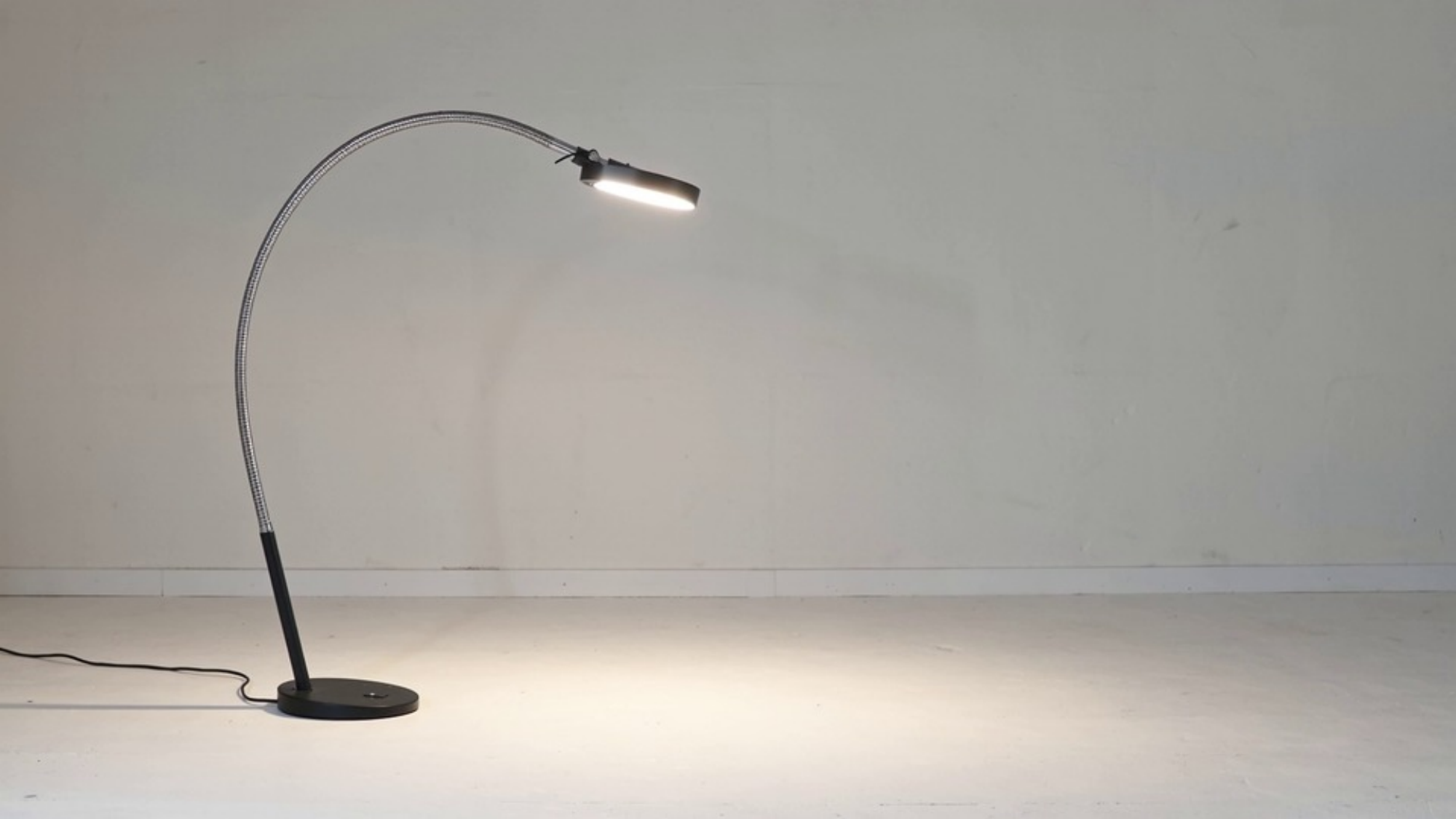}
        \end{tabular}
\par\vspace{3pt}\noindent\begin{tabular}{@{}c@{\hspace{\appGap}}c@{\hspace{\appGap}}c@{\hspace{\appGap}}c@{}}

        \appcaselabel{Wan NeurCo (T)}{\textbf{1/3 passed}}{\textcolor{appFail}{Fails (2), (3)}}
        & \appcaseframe{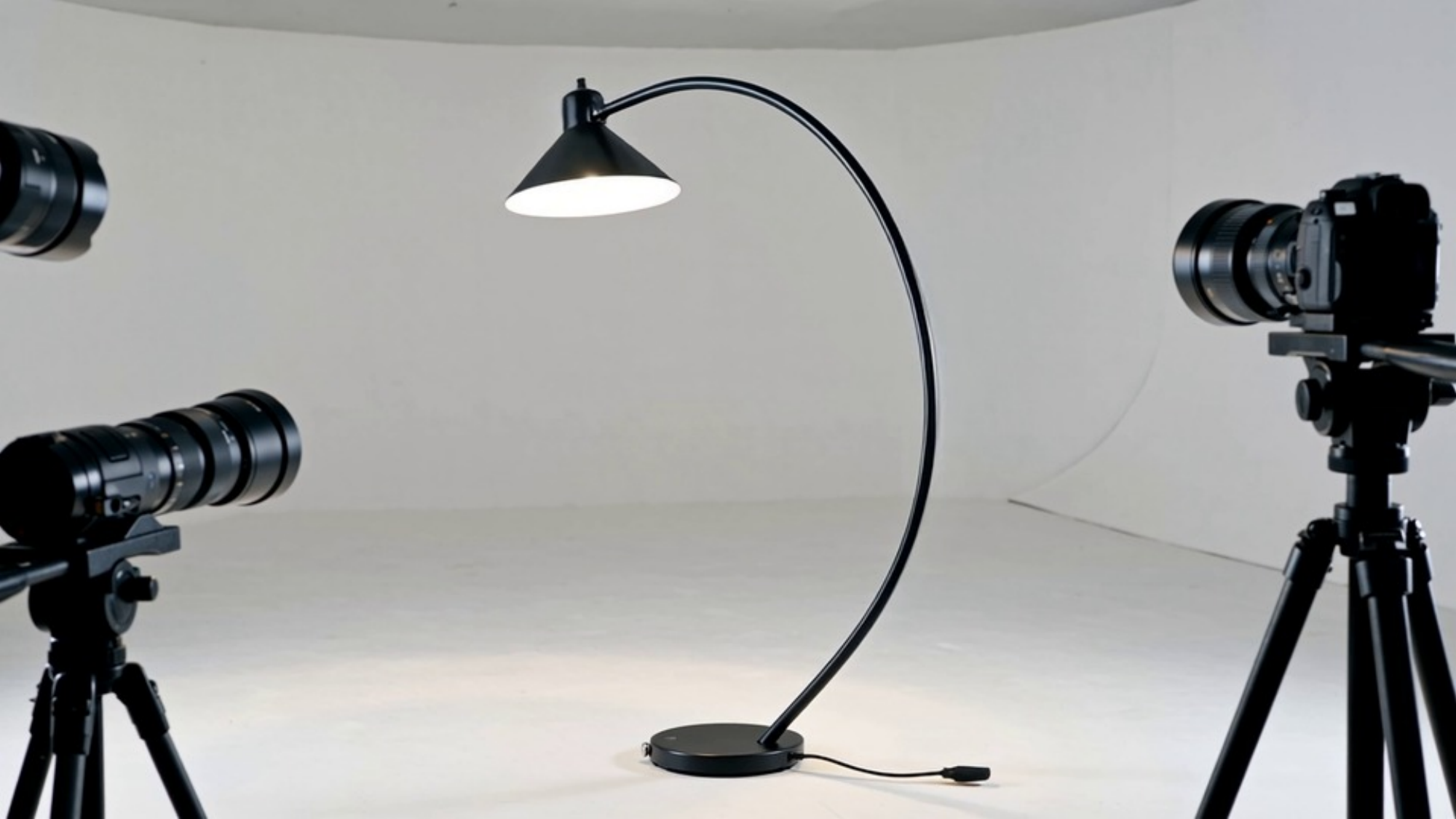}
        & \appcaseframe{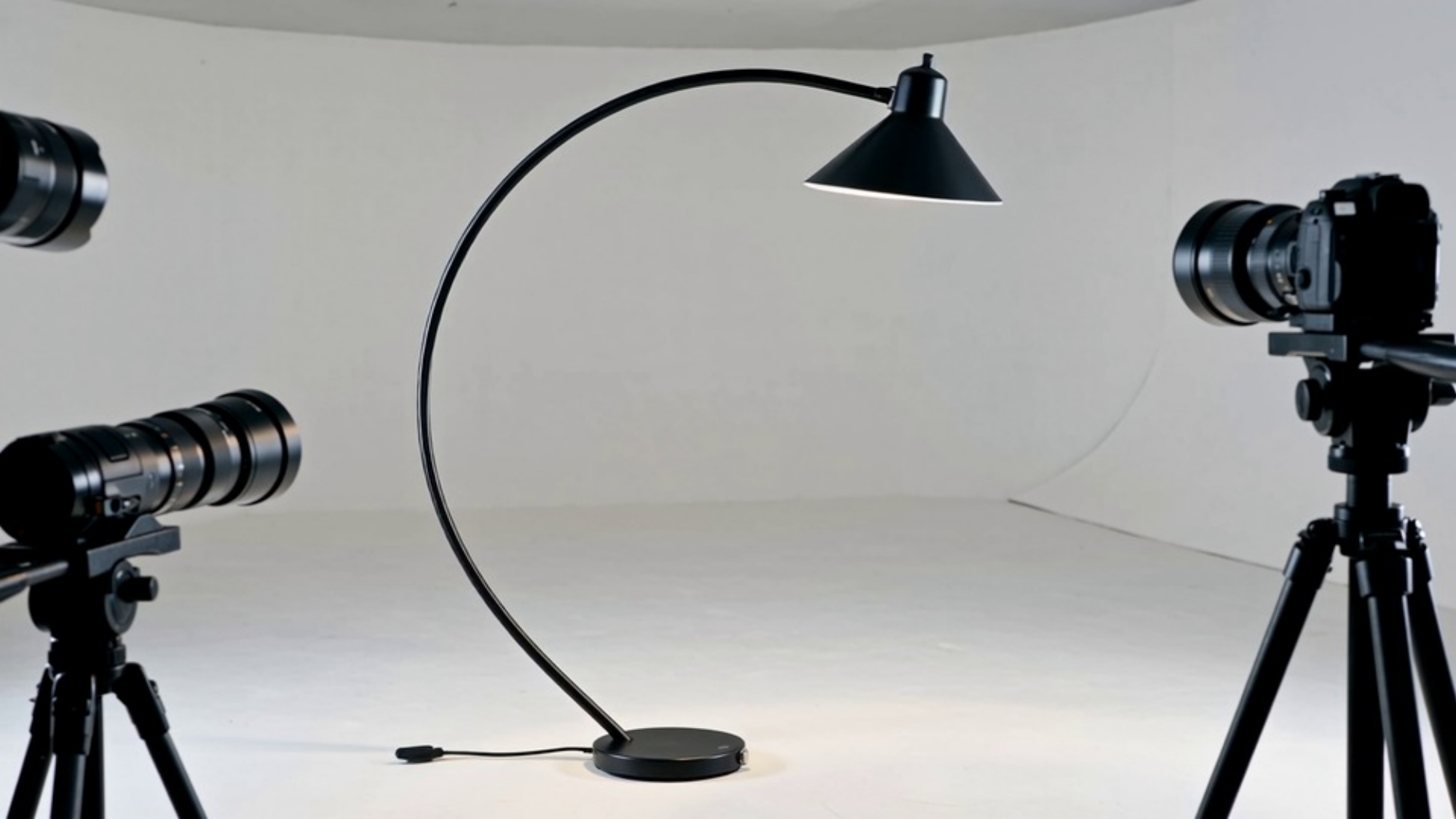}
        & \appcaseframe{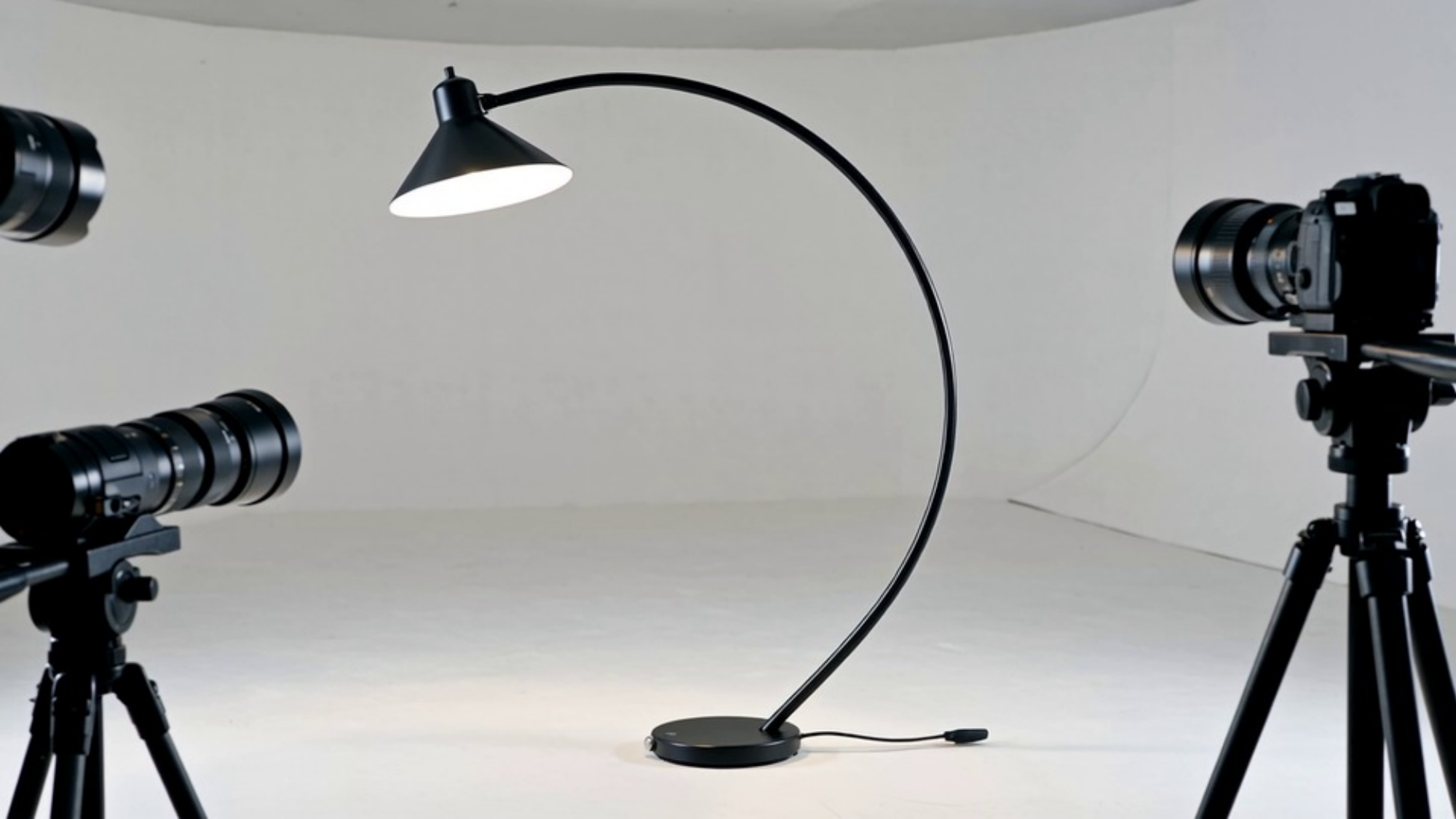}
        \end{tabular}
\par\vspace{3pt}\noindent\begin{tabular}{@{}c@{\hspace{\appGap}}c@{\hspace{\appGap}}c@{\hspace{\appGap}}c@{}}

        \appcaselabel{Wan NeurCo (T+O)}{\textcolor{appBlue}{\textbf{3/3 passed}}}{Depth control only}
        & \appcaseframe{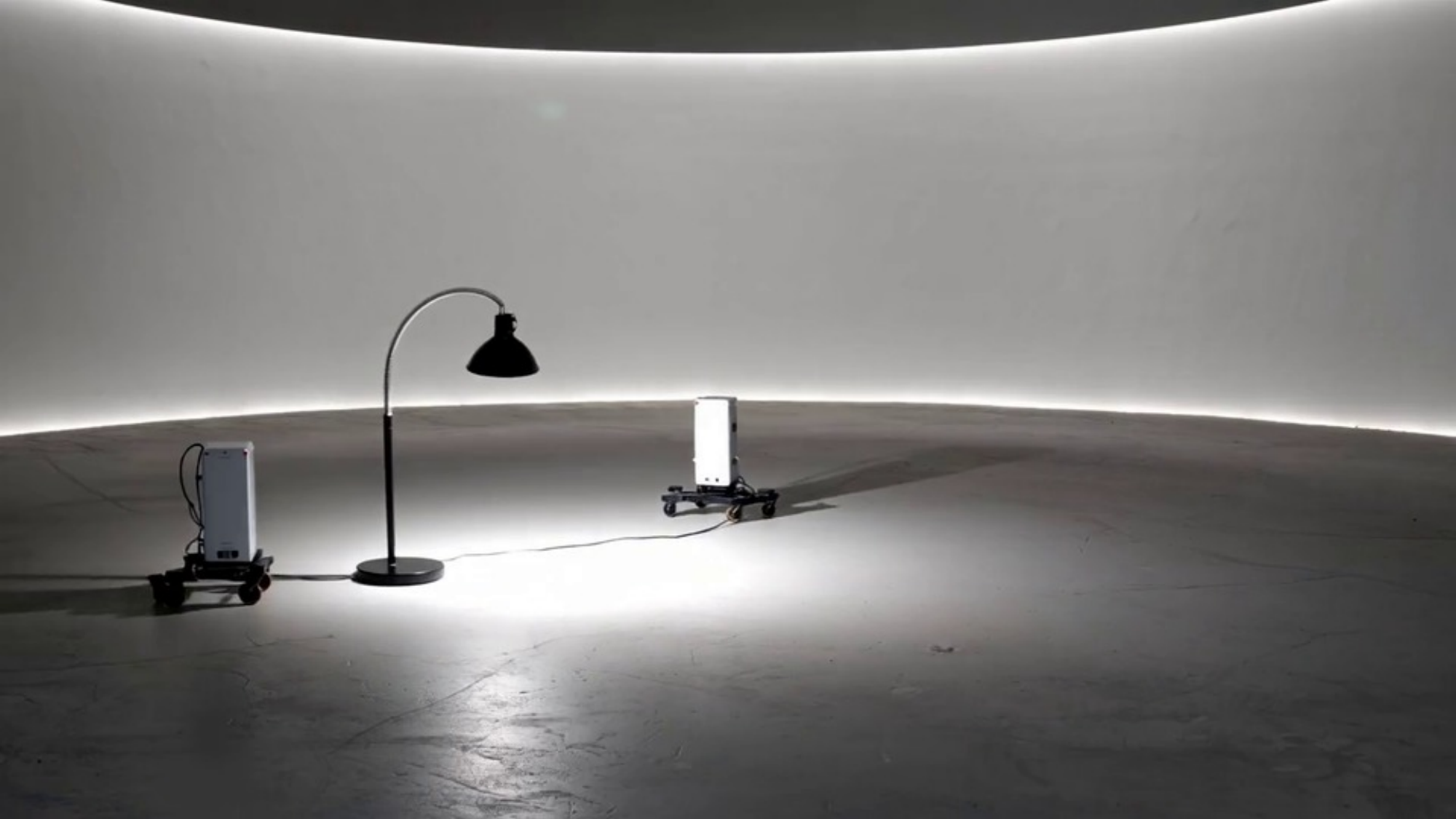}
        & \appcaseframe{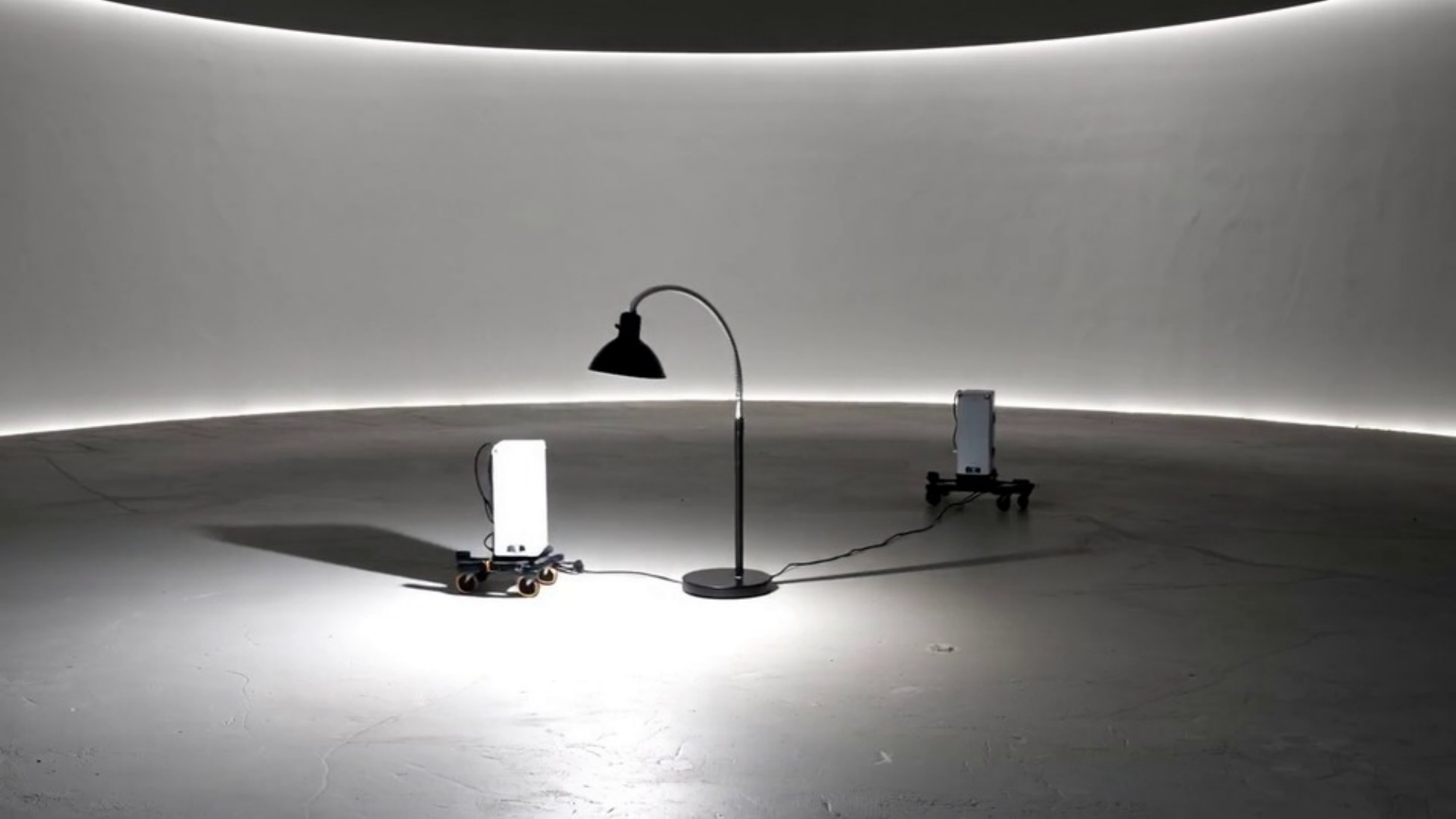}
        & \appcaseframe{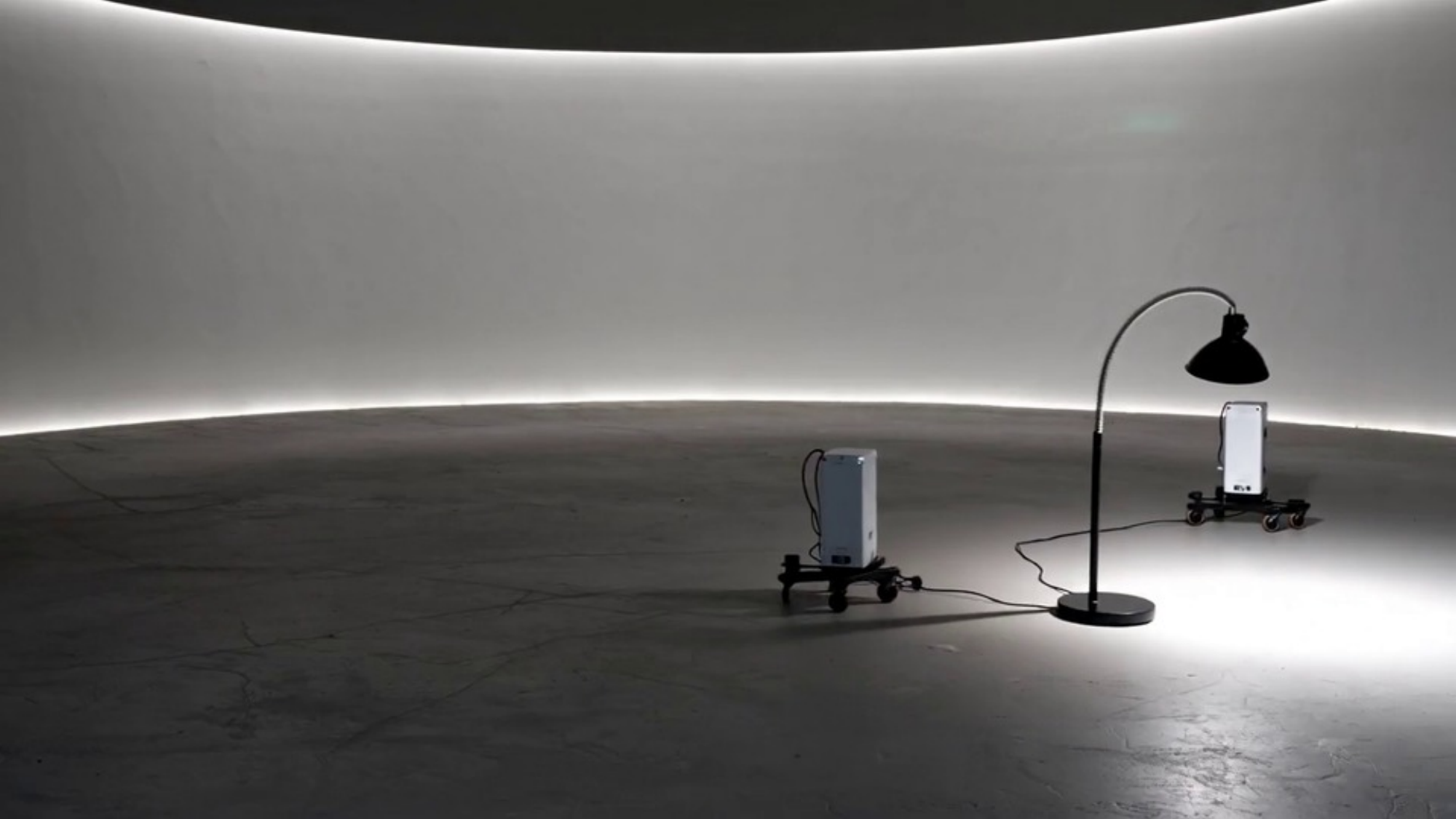}
    \end{tabular}
    \nopagebreak[3]\caption{\textbf{Rendered proxies recover coupled motion.}
    H3 uses a video reference and Wan uses depth guidance, without targeted
    rewriting in either T+O row. Both outputs copy minor proxy-scene props.}
    \label{fig:case_proxy_lamp}
    \endgroup
\par\endgroup
\end{appcaseblock}

\subsection{Capability-Guided Text Compilation}
\label{sec:app_cases_text}

Selective rewriting strengthens requirements that a backend handles
poorly while retaining supported controls. The coffee-cup example links
the capability manual to the resulting video; the subsequent image-to-video
examples illustrate compositional actions and scene transformations.
Two Seedance~2.5 examples show text rewriting alone repairing a dissolving
process and a three-step action sequence.

\begin{appcaseblock}
\noindent\textbf{Selective text rewriting.}
Text-only compilation recovers the missing actions and camera tilt in
the H3 pagoda scene (Figure~\ref{fig:case_h3_i2v}). NeurCo (T)
raises satisfaction from $3/6$ to $6/6$, including boiling water,
stirring, sweeping, and the upward camera move. The T+O configuration
also uses text rewriting alone and reaches $5/6$.

\par\smallskip\noindent\begingroup\centering
    \centering
    \begingroup
    \appcaseinit
    \appcaseinputas{I2V}{Begin with the snowy Chinese pagoda as shown. At second 1, two attendants enter the
    lower foreground: one wears a red coat and pushes a tea cart; the other wears a blue
    coat and carries a broom. The cart holds an open pot of vigorously boiling water, with
    continuous rolling bubbles and rising steam. Its label reads "\appteachar{}" in
    Japanese. The red-coated attendant stops and stirs the water while the blue-coated
    attendant sweeps snow. Slowly tilt the camera upward toward the pagoda's upper roof and
    snowy sky.}{\appinputimg{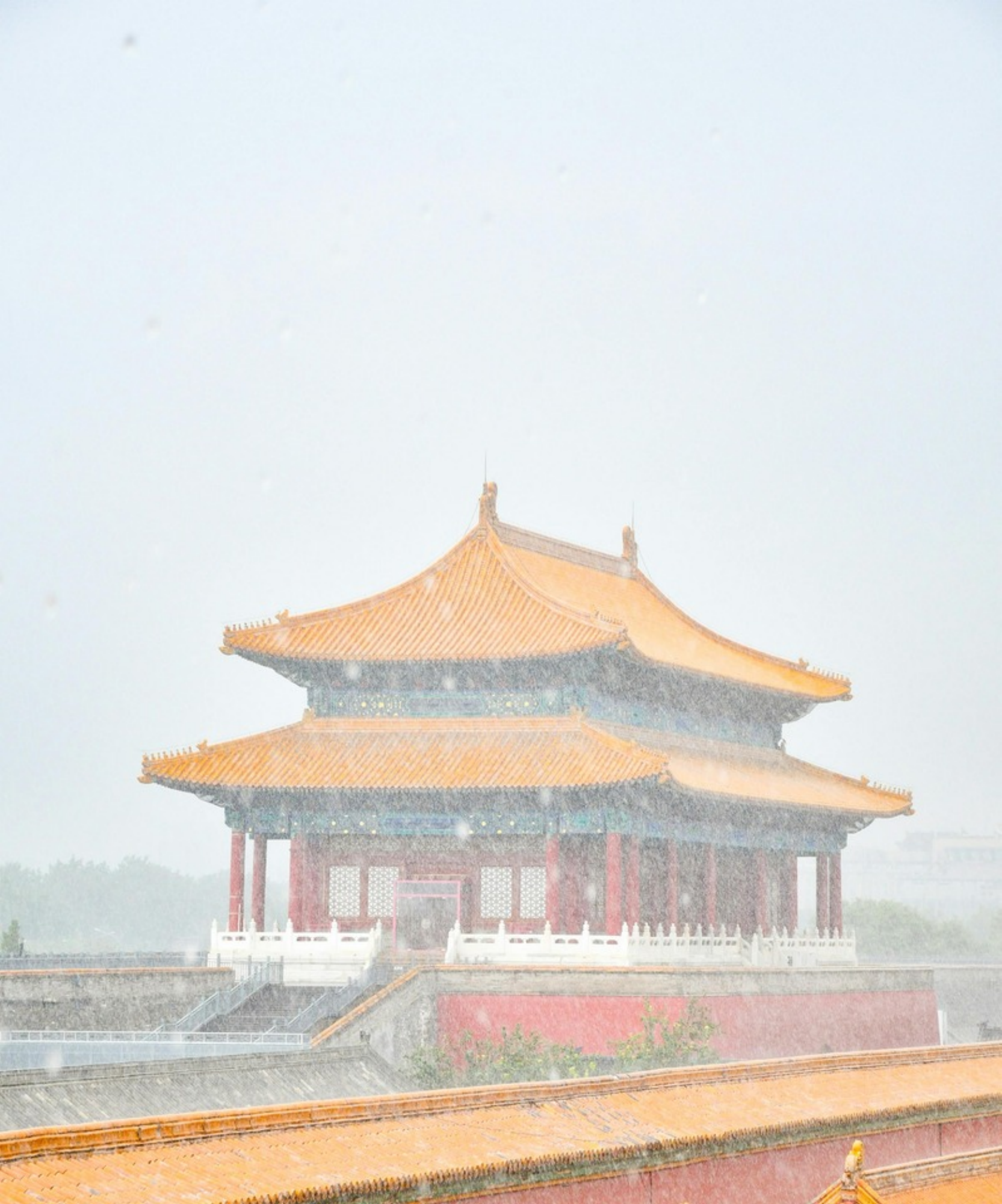}{First frame}%
    }{}{%
        \begin{tabularx}{\linewidth}{@{}XXX@{}}
            \textcolor{appBlue}{\textbf{(1) Boiling water}} & \textcolor{appBlue}{\textbf{(2) Camera tilts up}} & (3) Red and blue coats \\
            \textcolor{appBlue}{\textbf{(4) Stirring and sweeping}} & (5) Label text & (6) Entry at second 1
        \end{tabularx}}
    \begin{tabular}{@{}c@{\hspace{\appGap}}c@{\hspace{\appGap}}c@{\hspace{\appGap}}c@{}}
        \parbox[c]{\appLabelW}{\centering\textbf{Method}} & \textbf{Early} & \textbf{Middle} & \textbf{Late} \\[2pt]
        \appcaselabel{Native}{\textbf{3/6 passed}}{\textcolor{appFail}{Fails (1), (2), (4)}}
        & \appcaseframe{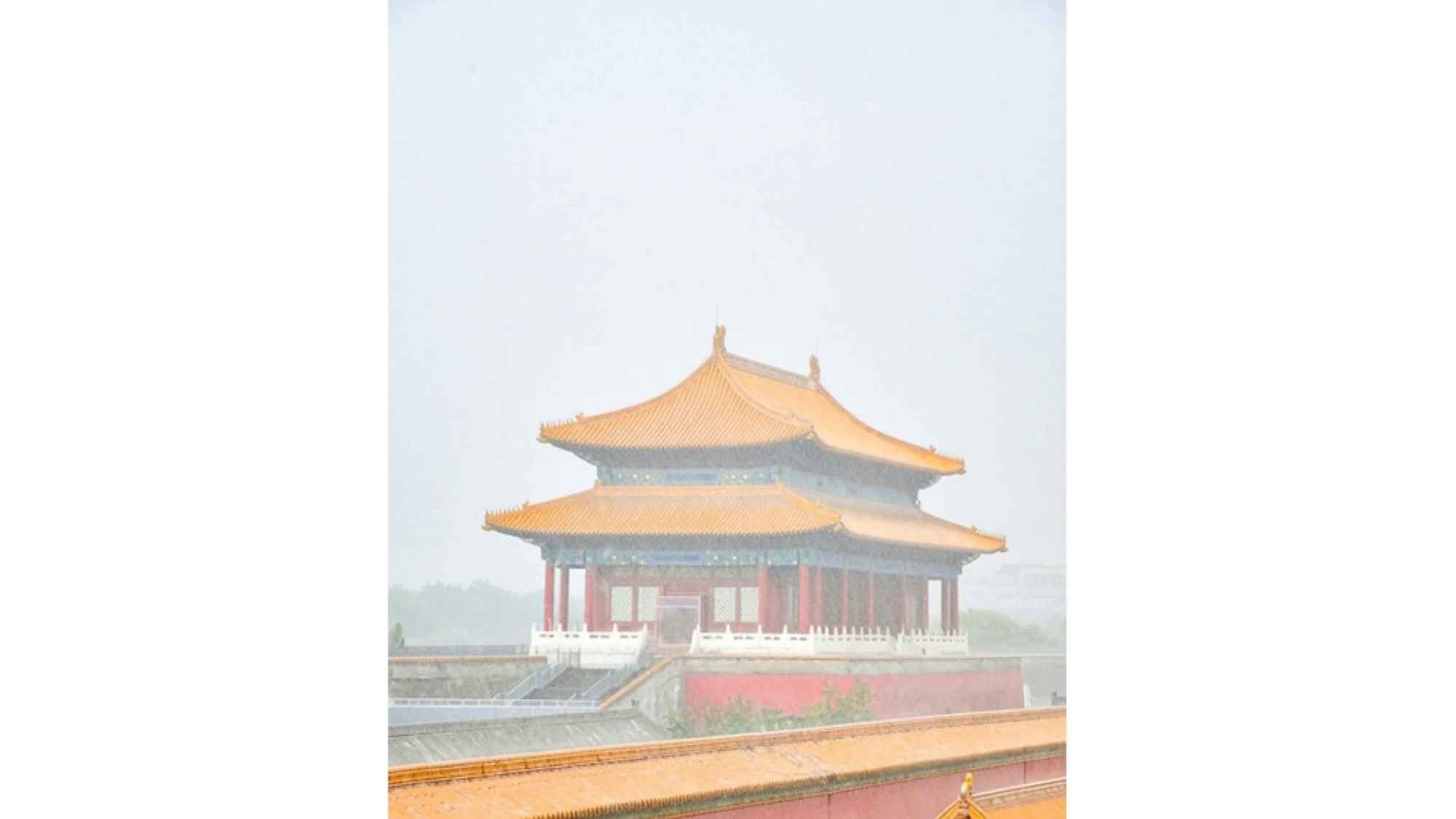}
        & \appcaseframe{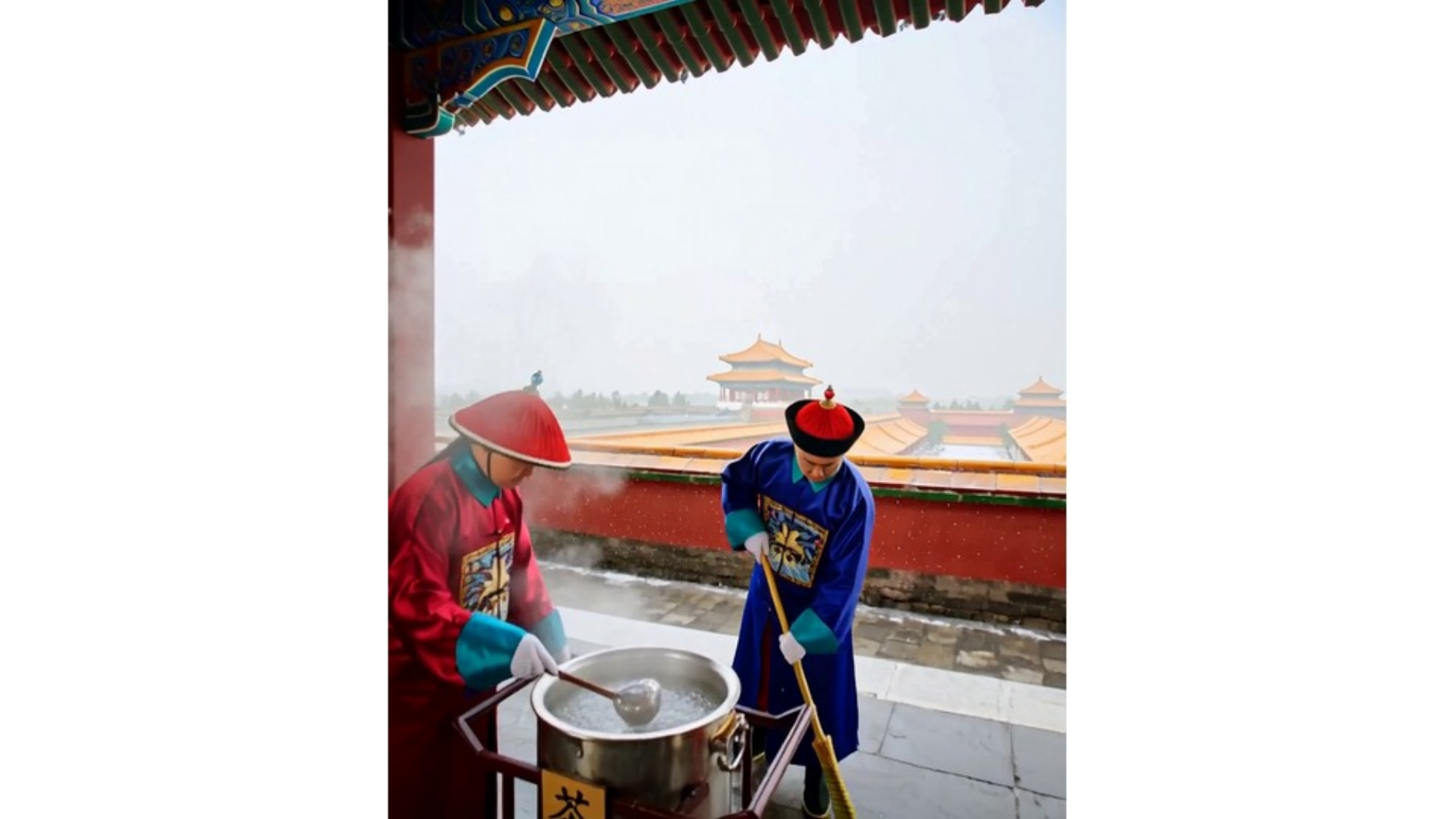}
        & \appcaseframe{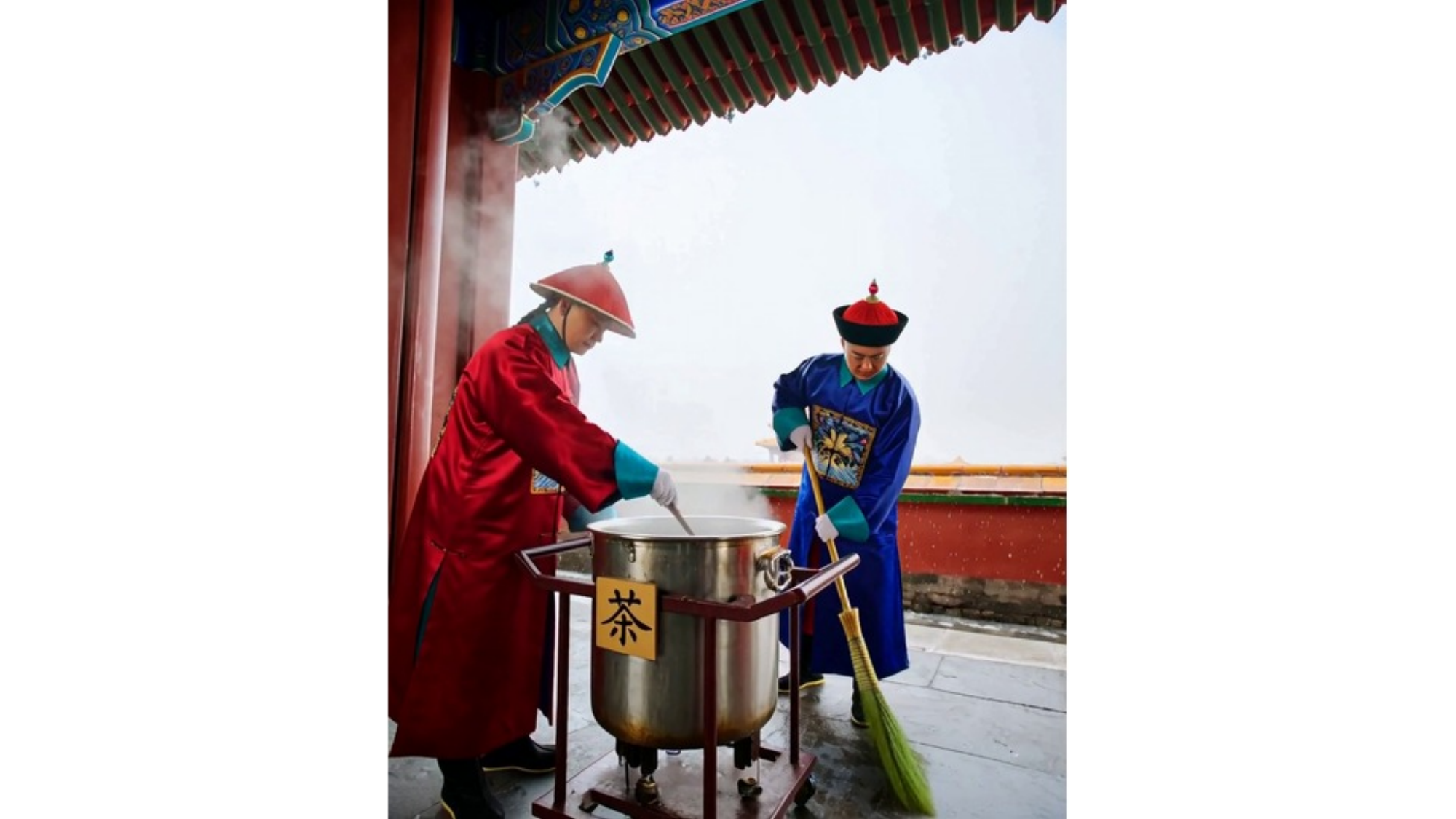}
        \end{tabular}
\par\vspace{3pt}\noindent\begin{tabular}{@{}c@{\hspace{\appGap}}c@{\hspace{\appGap}}c@{\hspace{\appGap}}c@{}}

        \appcaselabel{NeurCo (T)}{\textcolor{appBlue}{\textbf{6/6 passed}}}{Text adaptation}
        & \appcaseframe{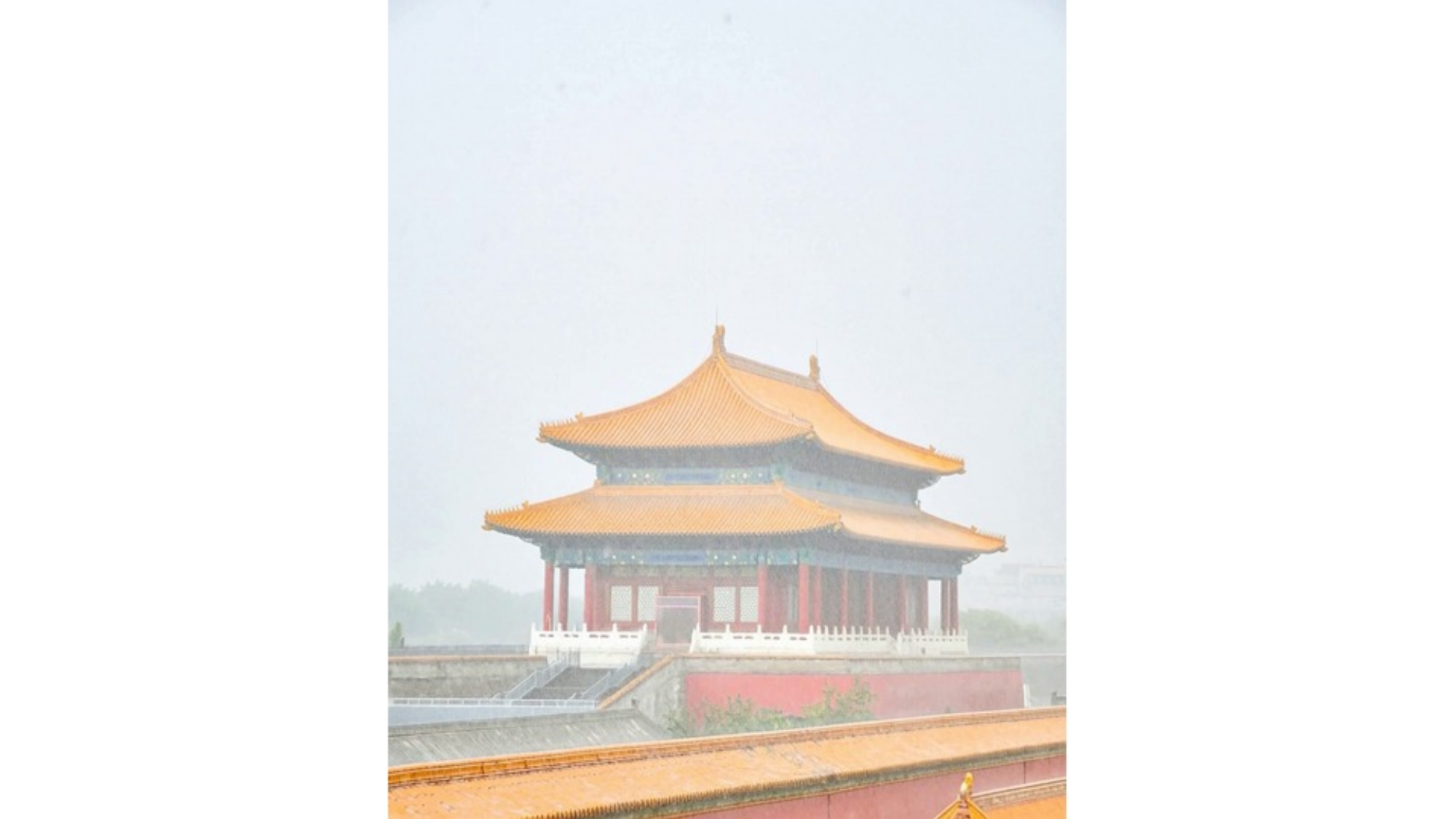}
        & \appcaseframe{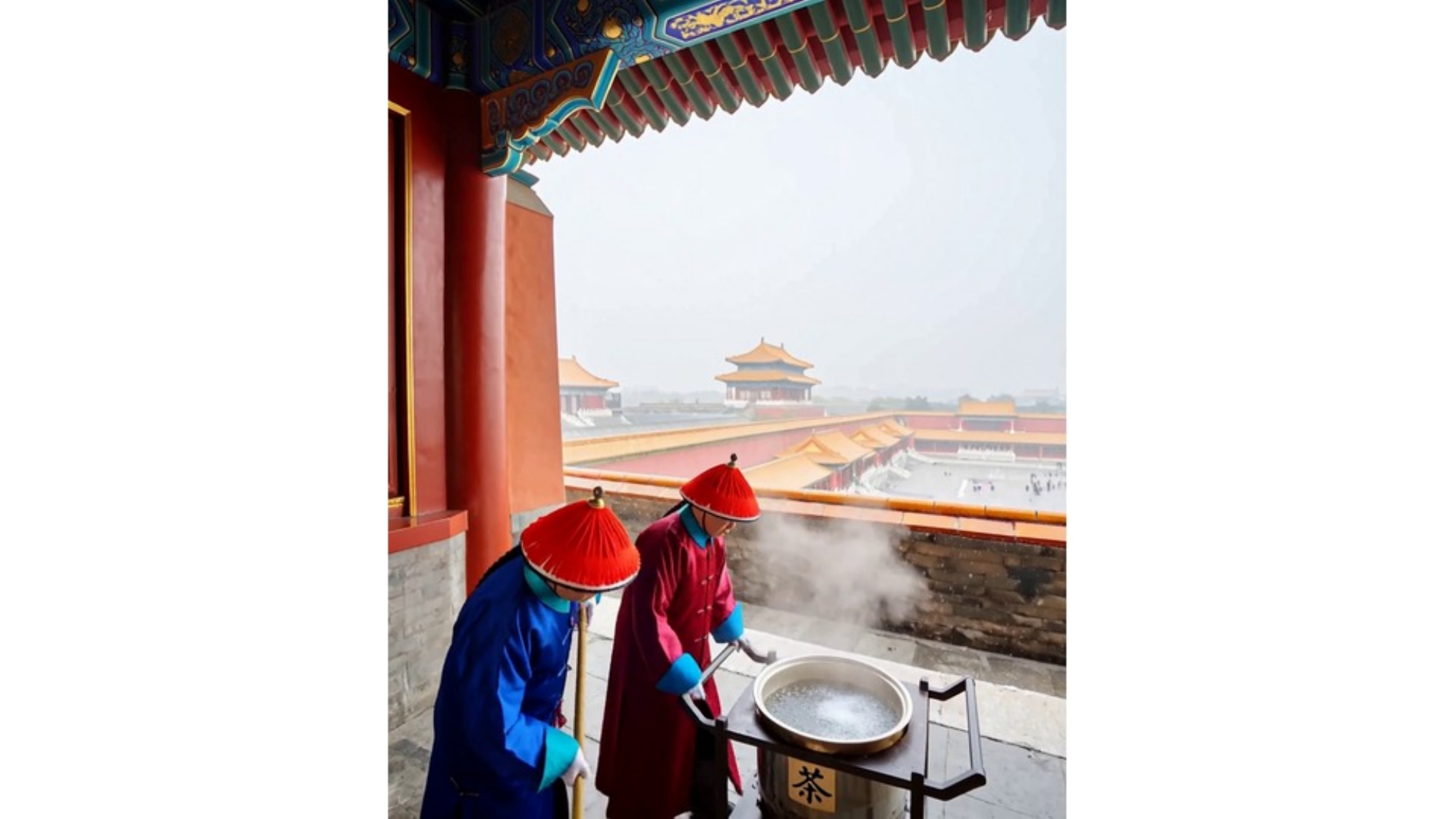}
        & \appcaseframe{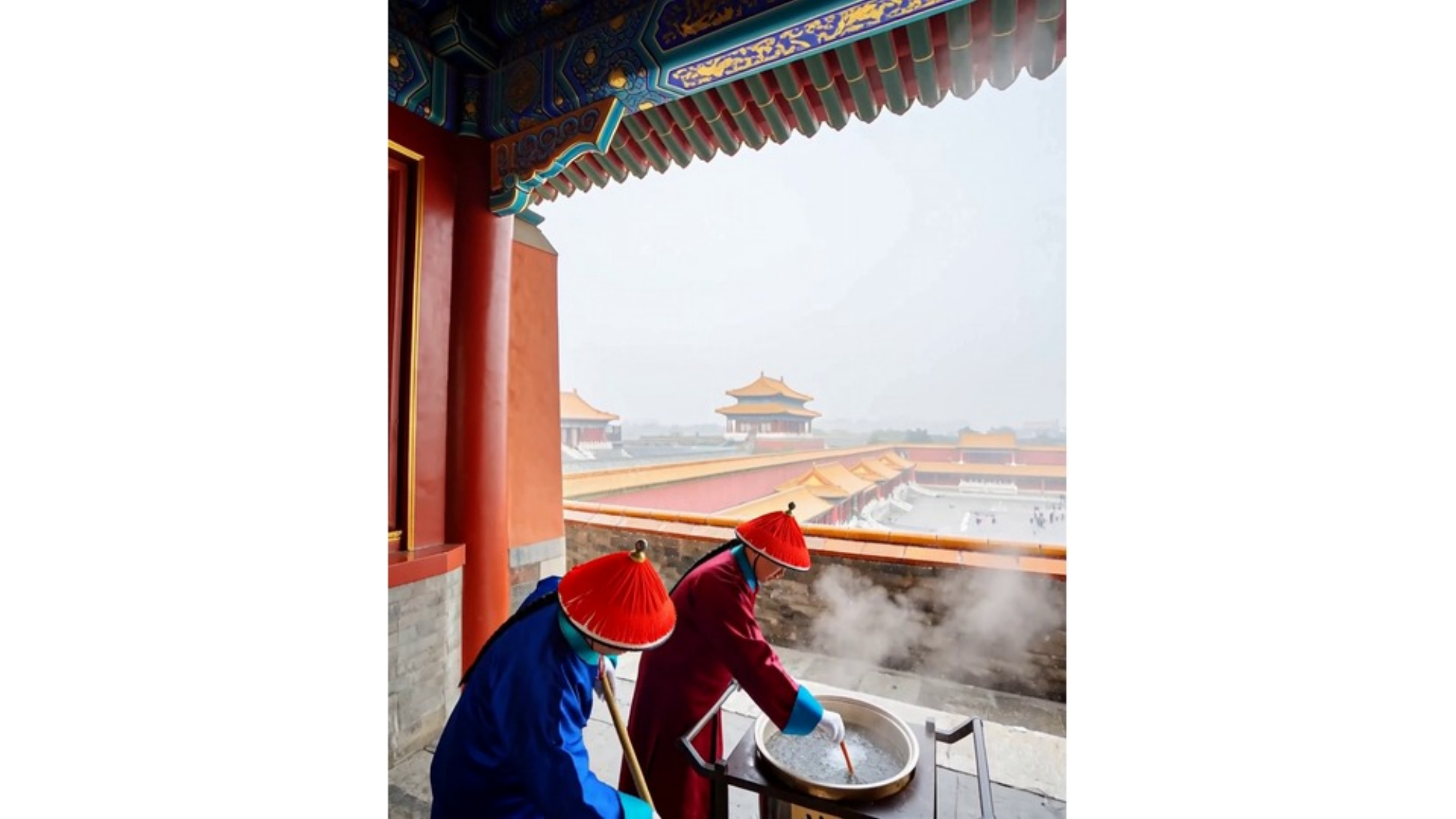}
        \end{tabular}
\par\vspace{3pt}\noindent\begin{tabular}{@{}c@{\hspace{\appGap}}c@{\hspace{\appGap}}c@{\hspace{\appGap}}c@{}}

        \appcaselabel{NeurCo (T+O)}{\textbf{5/6 passed}}{Text only}
        & \appcaseframe{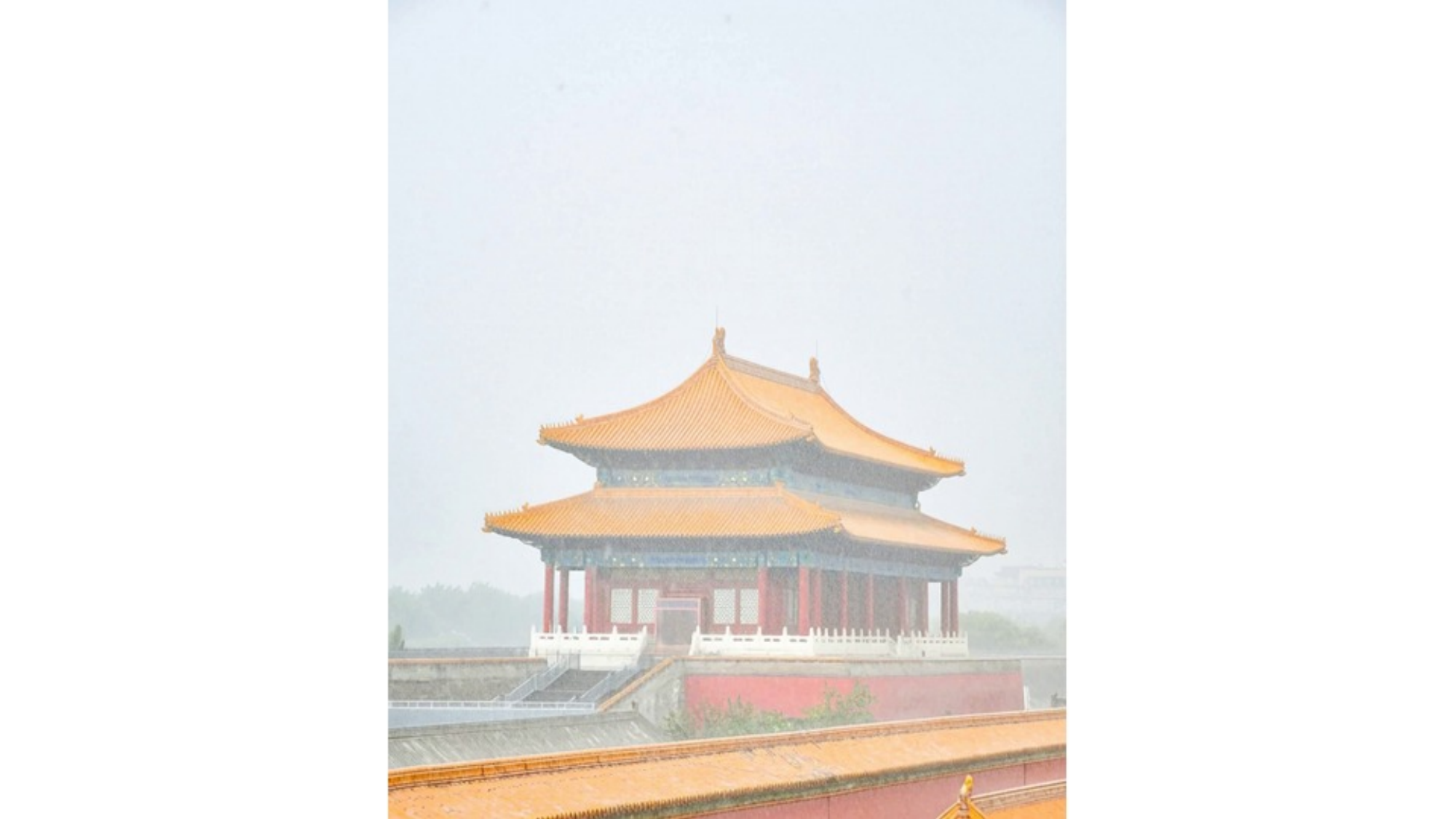}
        & \appcaseframe{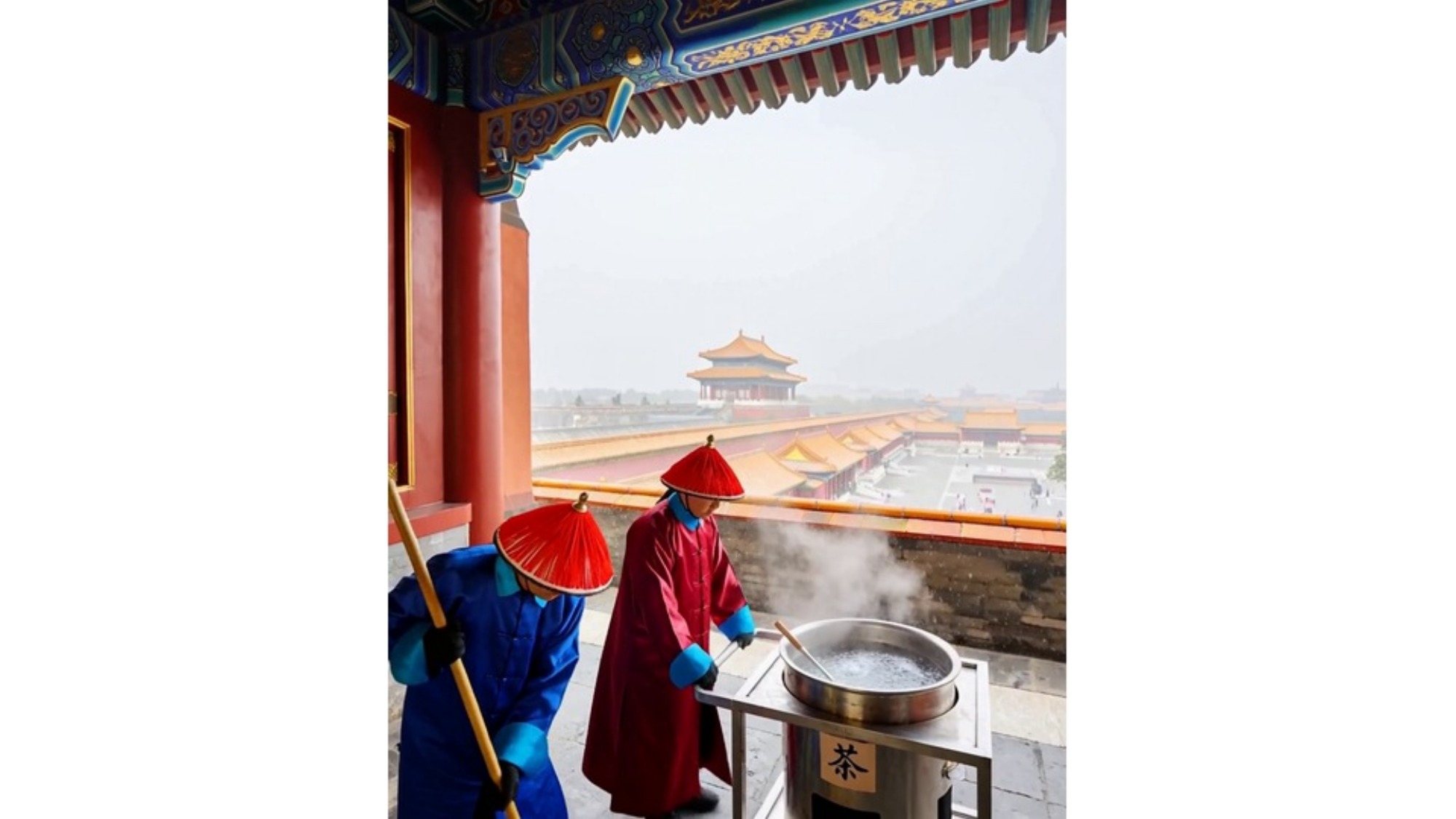}
        & \appcaseframe{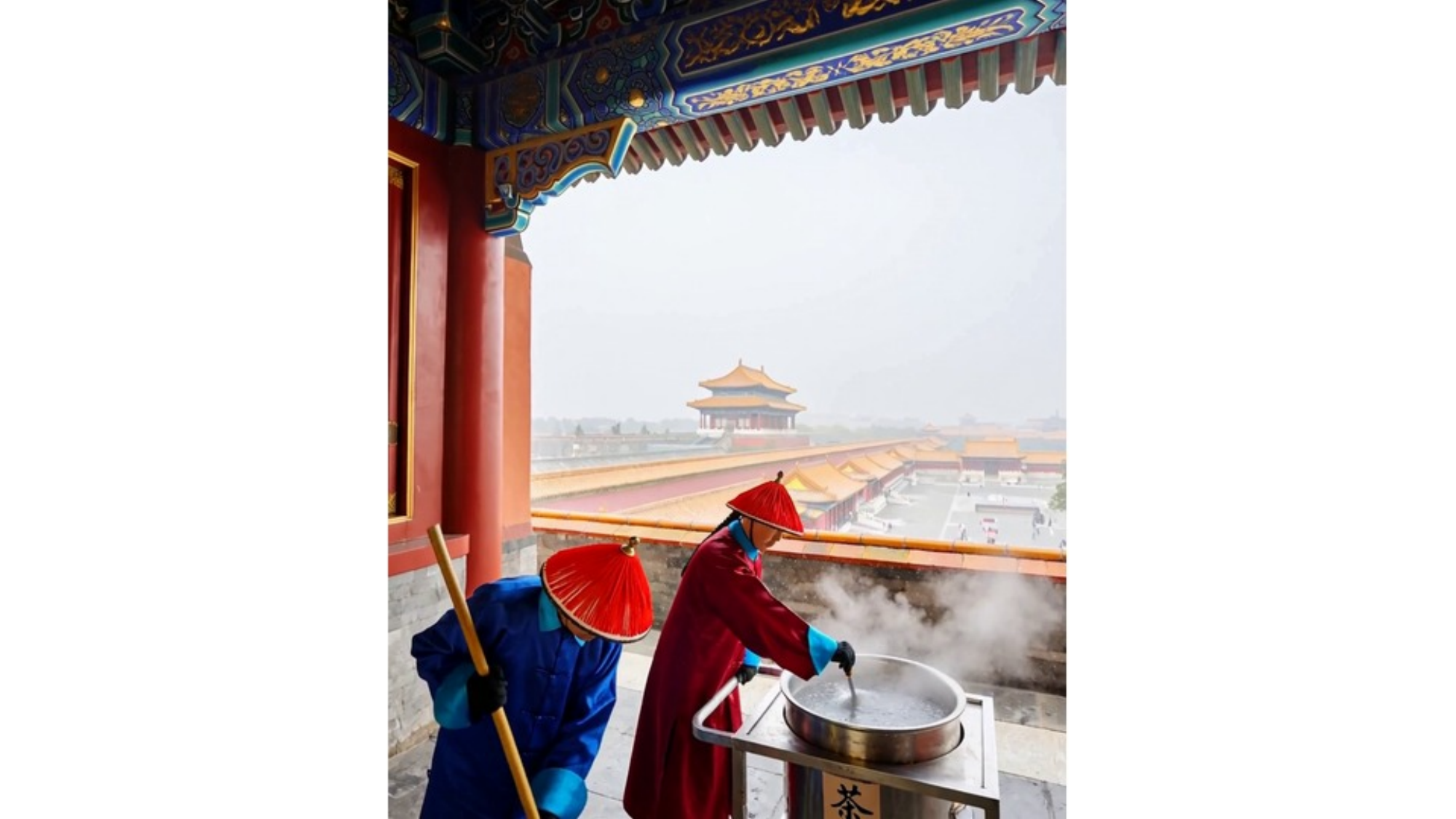}
    \end{tabular}
    \nopagebreak[3]\caption{\textbf{Text rewriting recovers compositional actions on H3.}
    NeurCo (T) satisfies all six requirements without external conditioning.
    The source frames are padded to a consistent display size.}
    \label{fig:case_h3_i2v}
    \endgroup
\par\endgroup
\end{appcaseblock}

\begin{appcaseblock}
\noindent\textbf{Sugar dissolving in iced tea.}
In Figure~\ref{fig:case_sd_tea}, Native generation releases the sugar, but the spoon never enters the tea and a mound of sugar stays on the surface ($2/6$). Targeted rewriting states that the spoon starts above the glass, enters once at second 2, and stays submerged while the crystals dissolve. Text-only compilation satisfies five of six requirements, with the spoon submerged and the sugar dissolved by the final frame. The T+O plan also uses text rewriting alone and reaches $4/6$; neither rewrite achieves the focus pull.

\par\smallskip\noindent\begingroup\centering
    \centering
    \begingroup
    \appcaseinit
    \appcaseinputas{T2V}{A spoon holding sugar crystals hovers above a glass of iced tea. Begin focused on the glass behind the foreground spoon, then pull focus onto the sugar crystals. Push the camera toward the glass, increasing its height from about 20\% to about 80\% of the frame. At second 2, lower the spoon from above the glass into the tea and release the sugar. Keep the spoon submerged as the crystals shrink and dissolve completely, forming a sugar solution that sweetens the iced tea.}{}{}{%
        \begin{tabularx}{\linewidth}{@{}XXX@{}}
            \textcolor{appBlue}{\textbf{(1) Sugar crystals dissolve}} & (2) Iced-tea glass central & (3) Focus pull glass$\to$sugar \\
            \textcolor{appBlue}{\textbf{(4) Spoon enters at second 2}} & (5) Push-in, 20\% to 80\% & \textcolor{appBlue}{\textbf{(6) Spoon starts above, stays submerged}}
        \end{tabularx}}
    \begin{tabular}{@{}c@{\hspace{\appGap}}c@{\hspace{\appGap}}c@{\hspace{\appGap}}c@{}}
        \parbox[c]{\appLabelW}{\centering\textbf{Method}} & \textbf{Early} & \textbf{Middle} & \textbf{Late} \\[2pt]
        \appcaselabel{Native}{\textbf{2/6 passed}}{\textcolor{appFail}{Fails (1), (3), (4), (6)}}
        & \appcaseframe{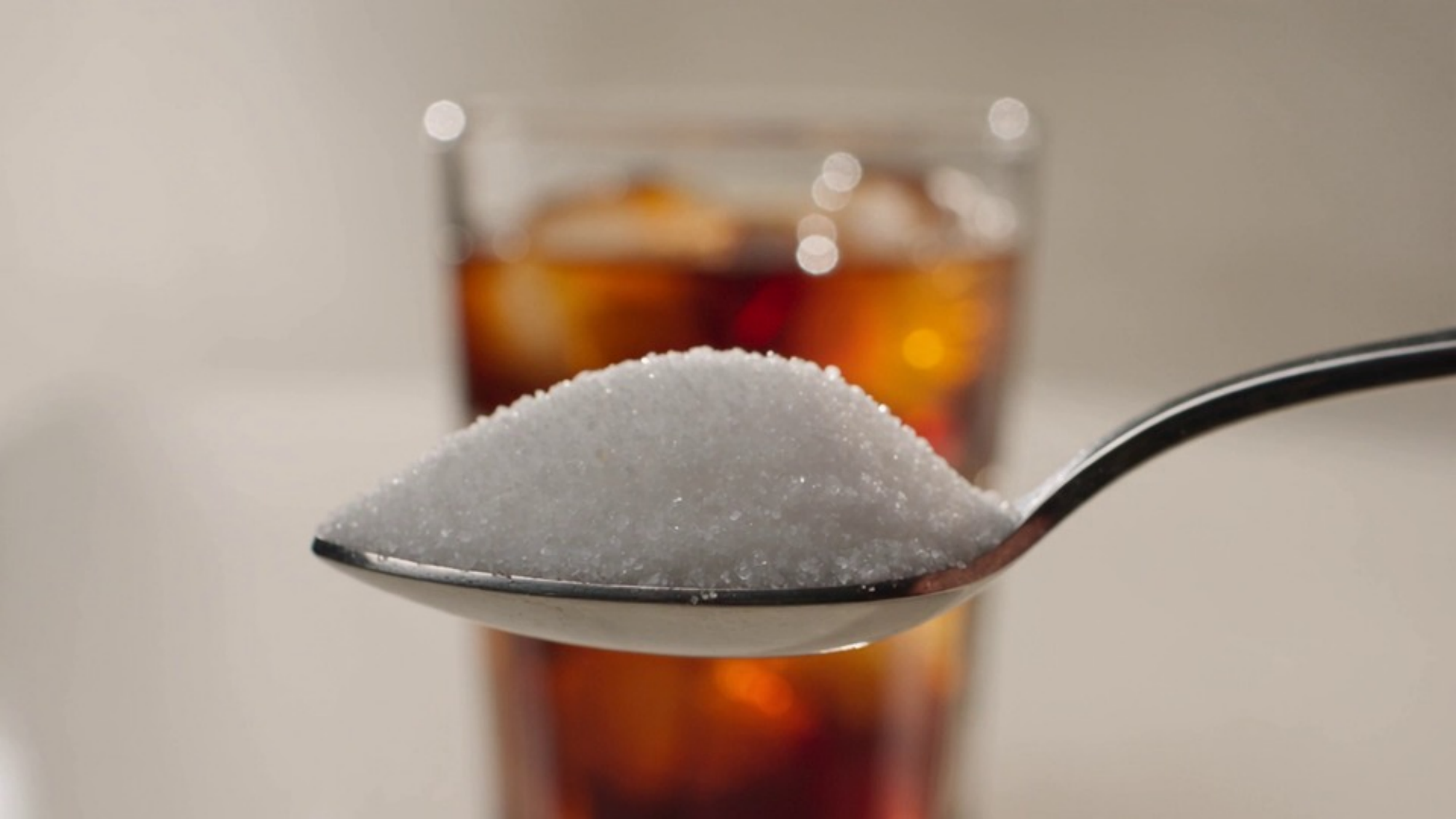}
        & \appcaseframe{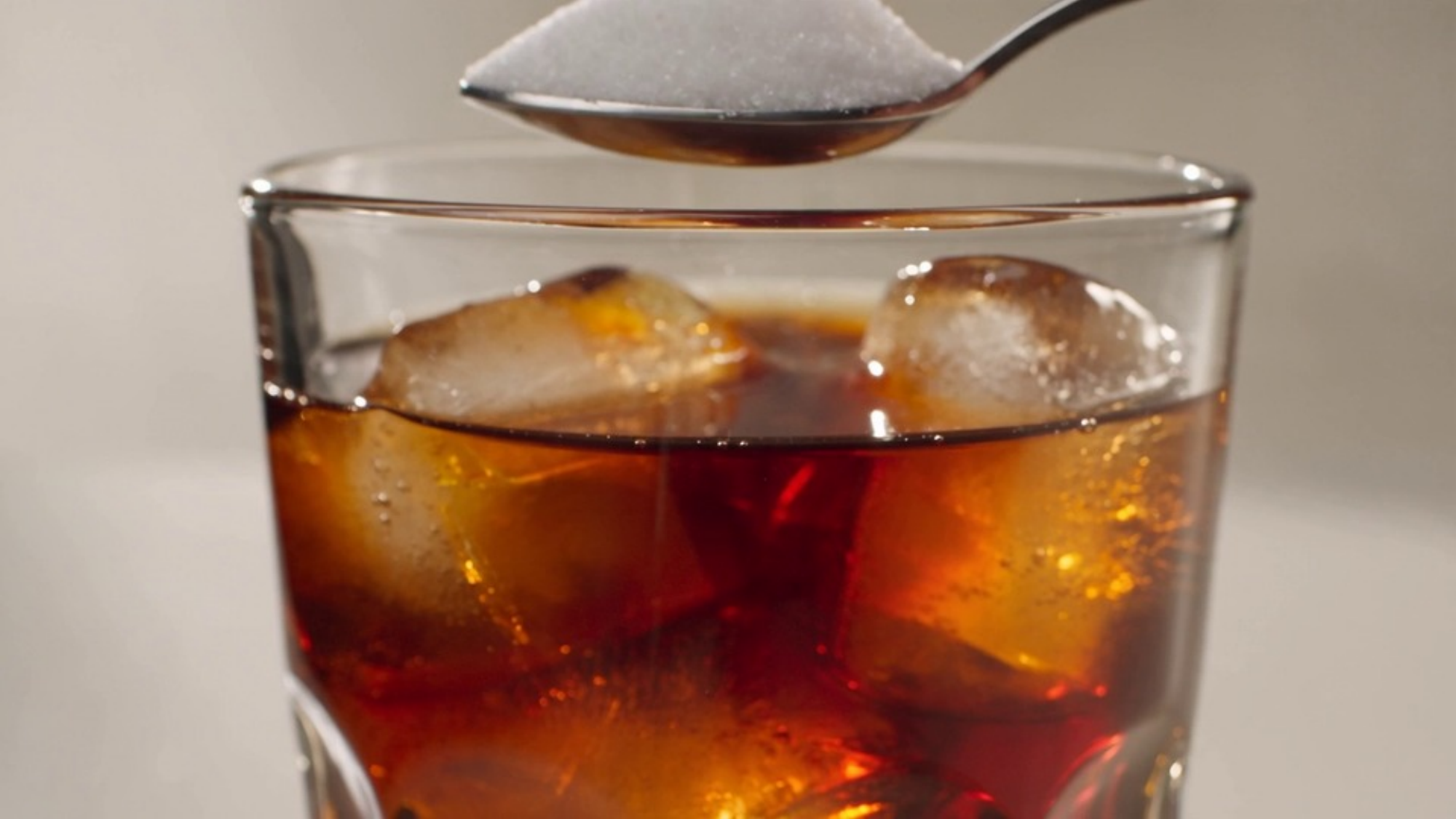}
        & \appcaseframe{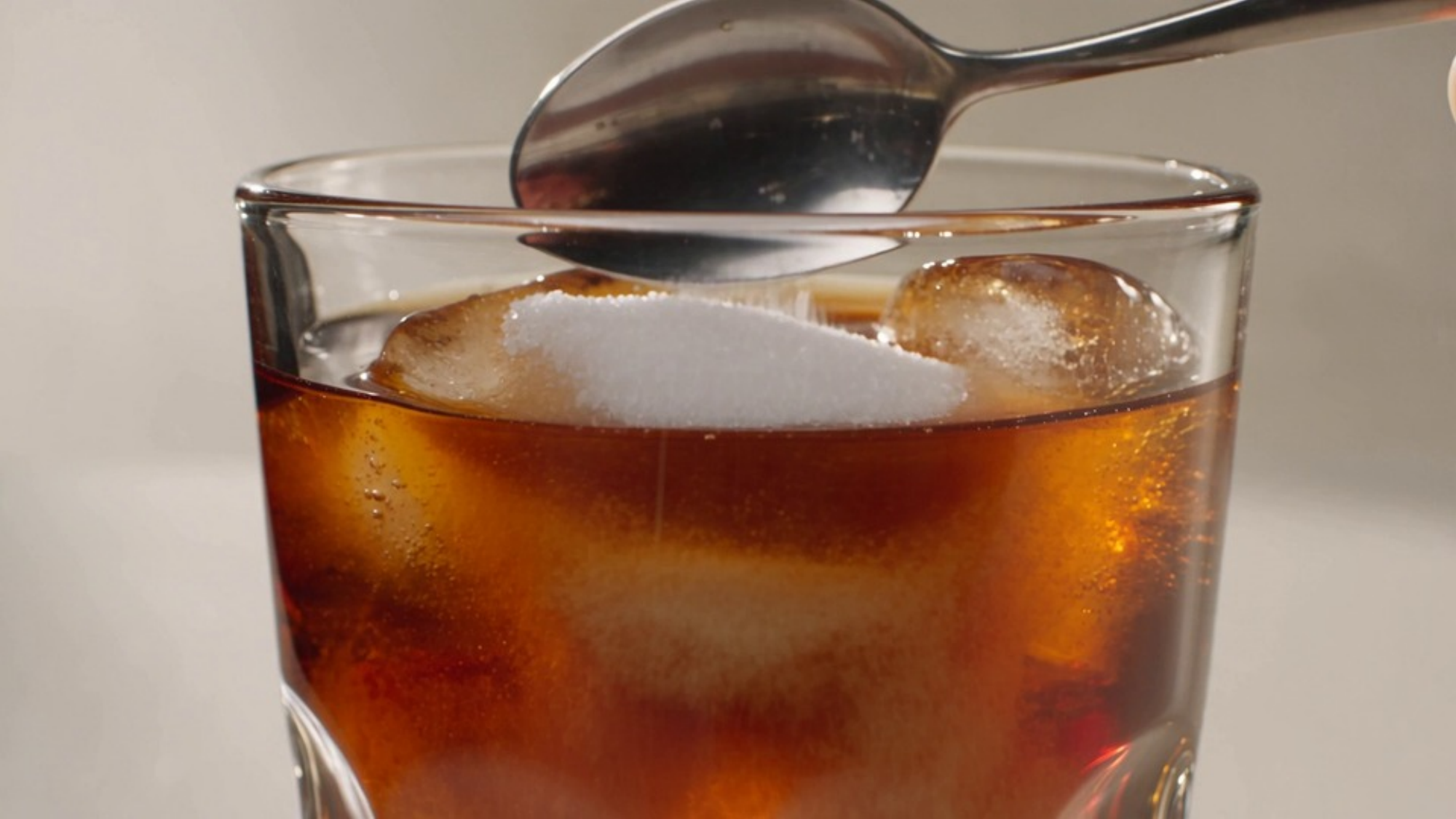}
    \end{tabular}
\par\vspace{3pt}\noindent\begin{tabular}{@{}c@{\hspace{\appGap}}c@{\hspace{\appGap}}c@{\hspace{\appGap}}c@{}}
        \appcaselabel{NeurCo (T)}{\textcolor{appBlue}{\textbf{5/6 passed}}}{\textcolor{appFail}{Fails (3)}}
        & \appcaseframe{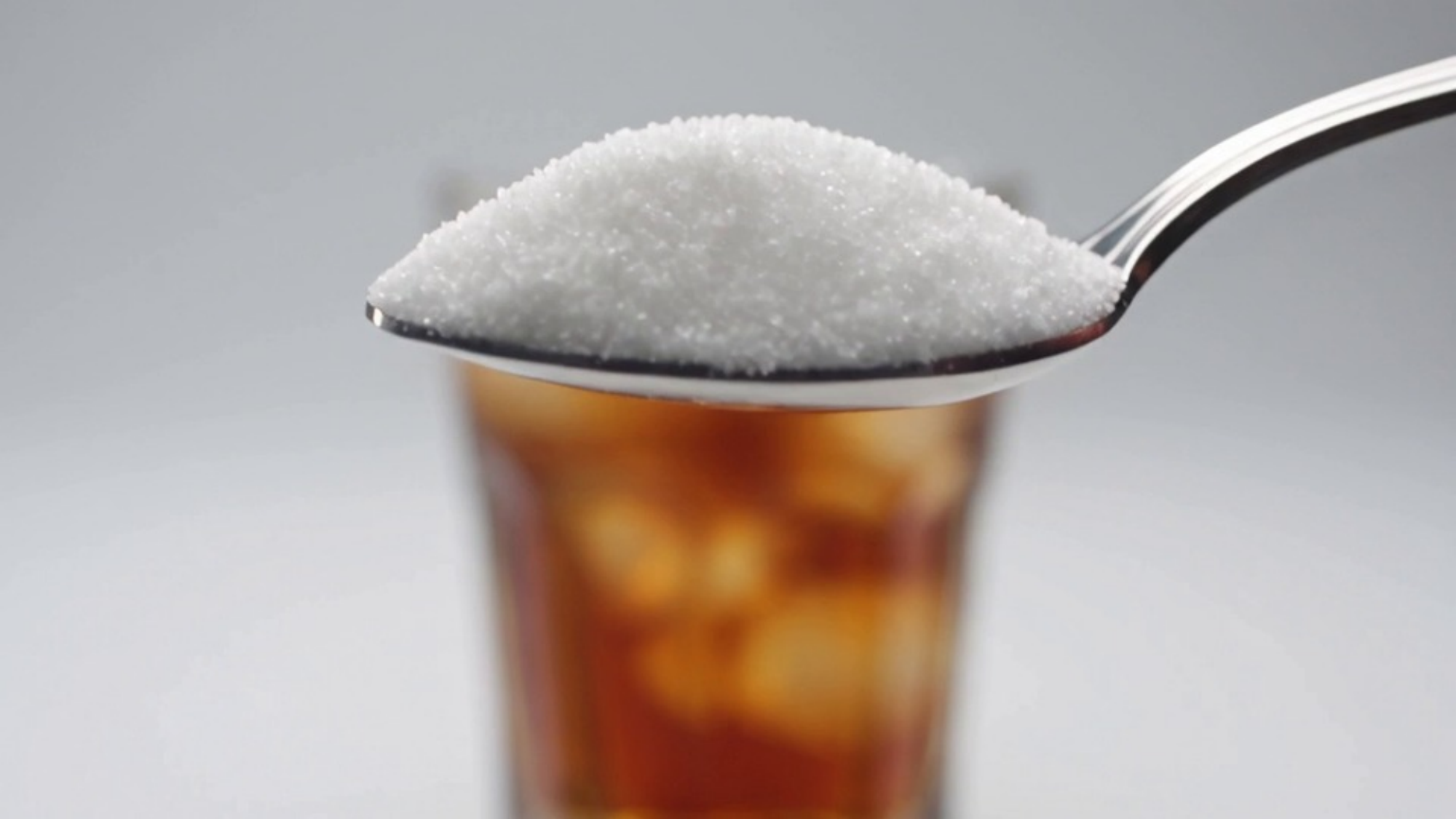}
        & \appcaseframe{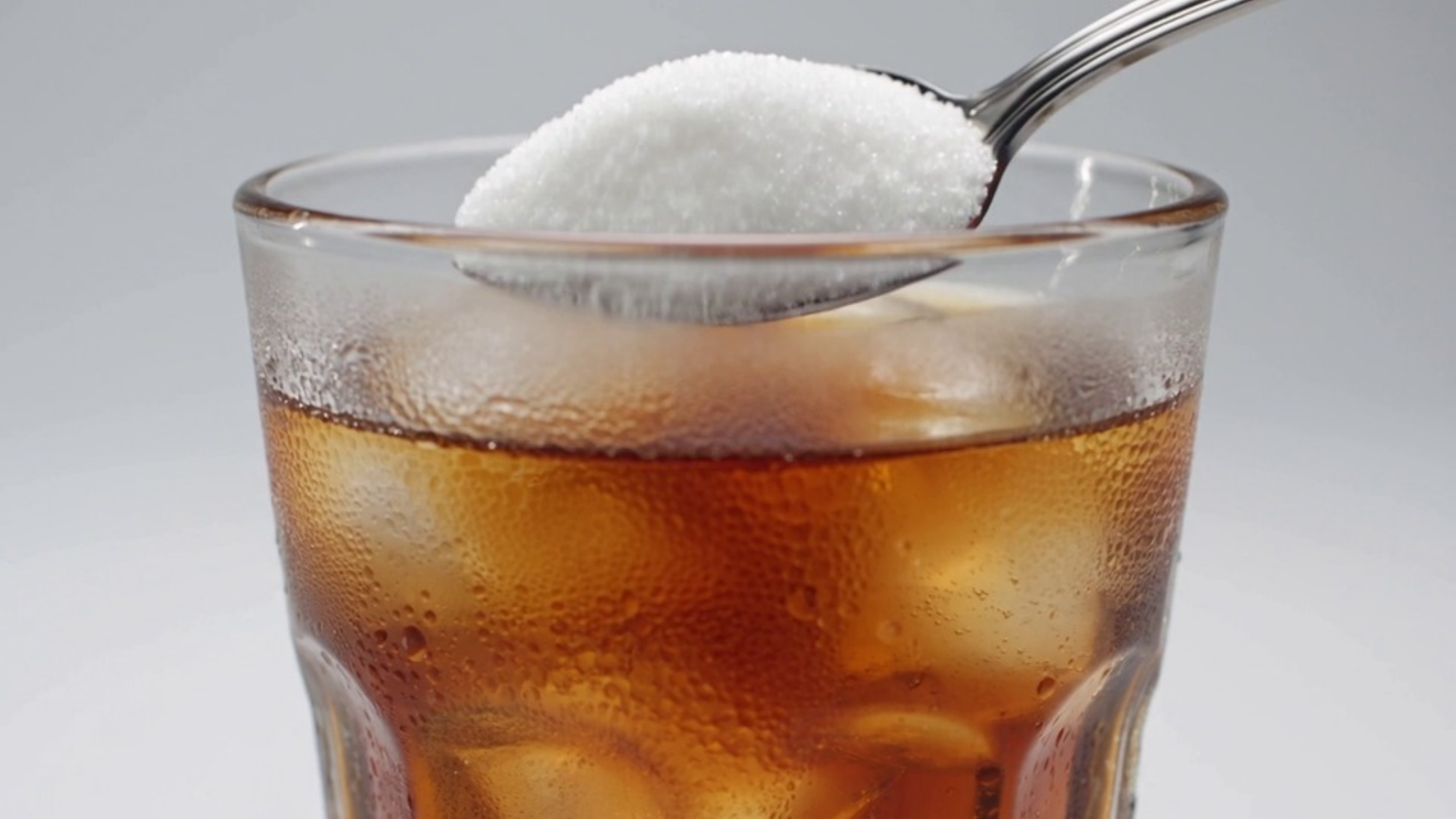}
        & \appcaseframe{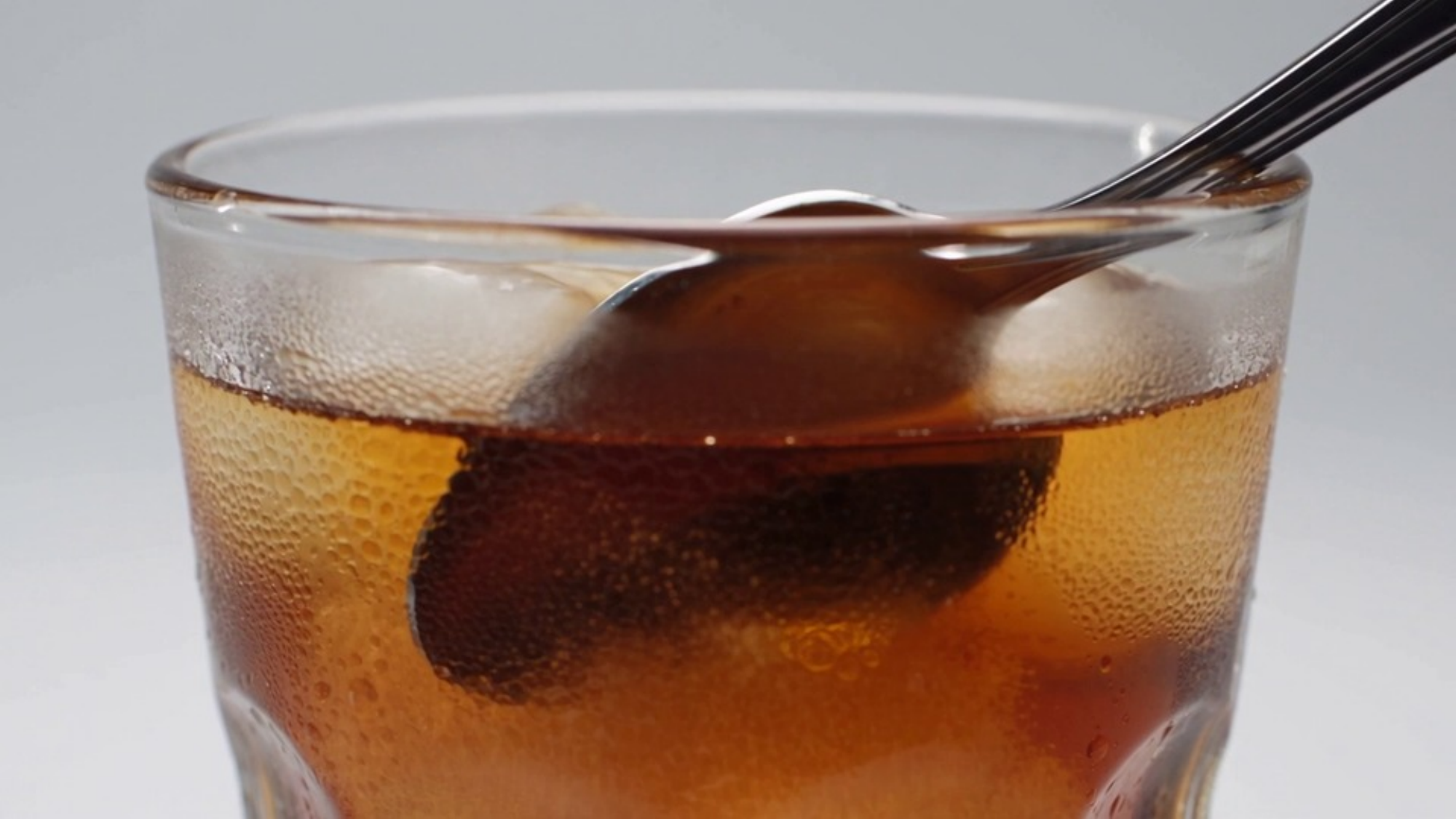}
    \end{tabular}
\par\vspace{3pt}\noindent\begin{tabular}{@{}c@{\hspace{\appGap}}c@{\hspace{\appGap}}c@{\hspace{\appGap}}c@{}}
        \appcaselabel{NeurCo (T+O)}{\textbf{4/6 passed}}{Text only; \textcolor{appFail}{Fails (3), (6)}}
        & \appcaseframe{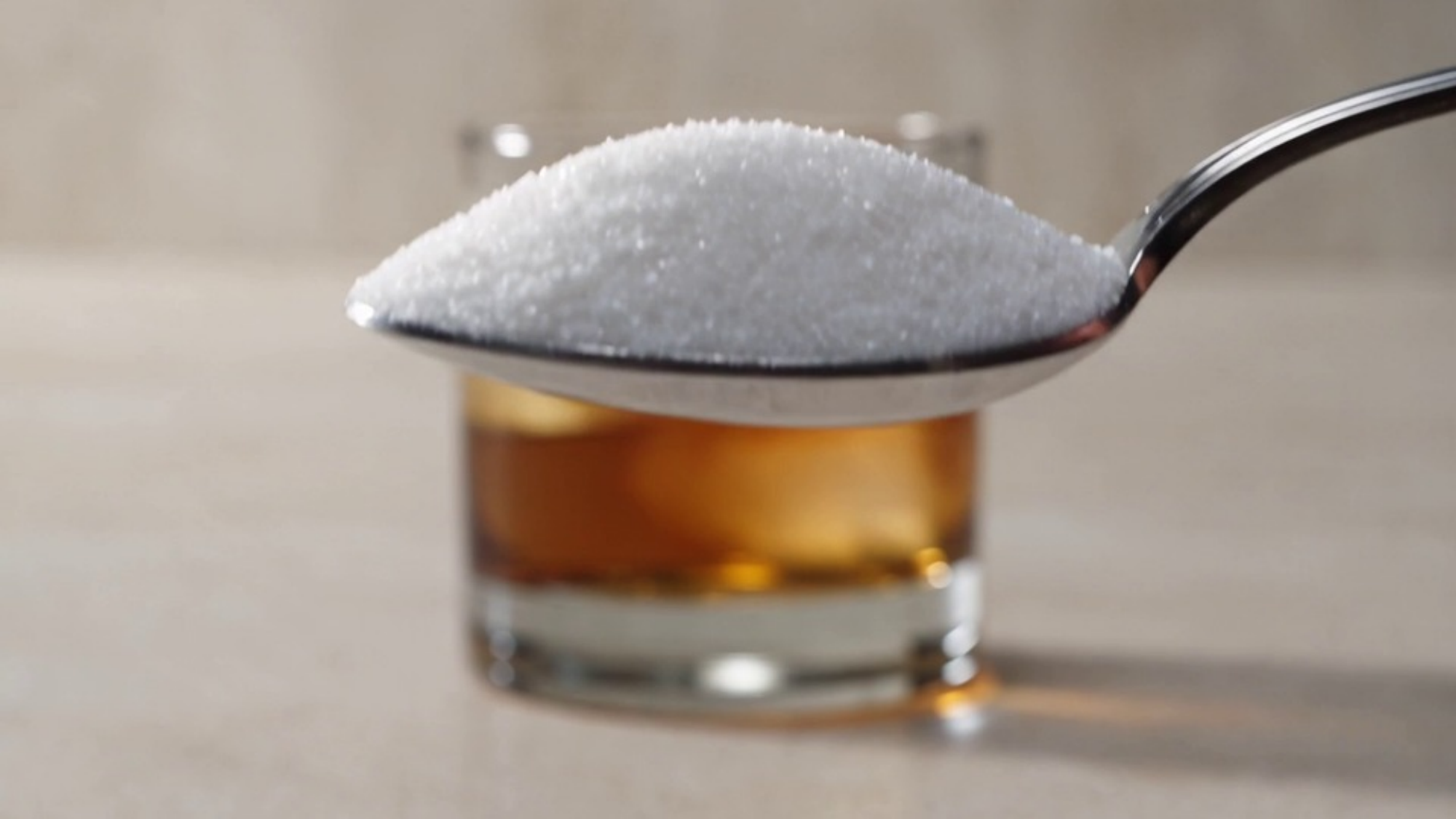}
        & \appcaseframe{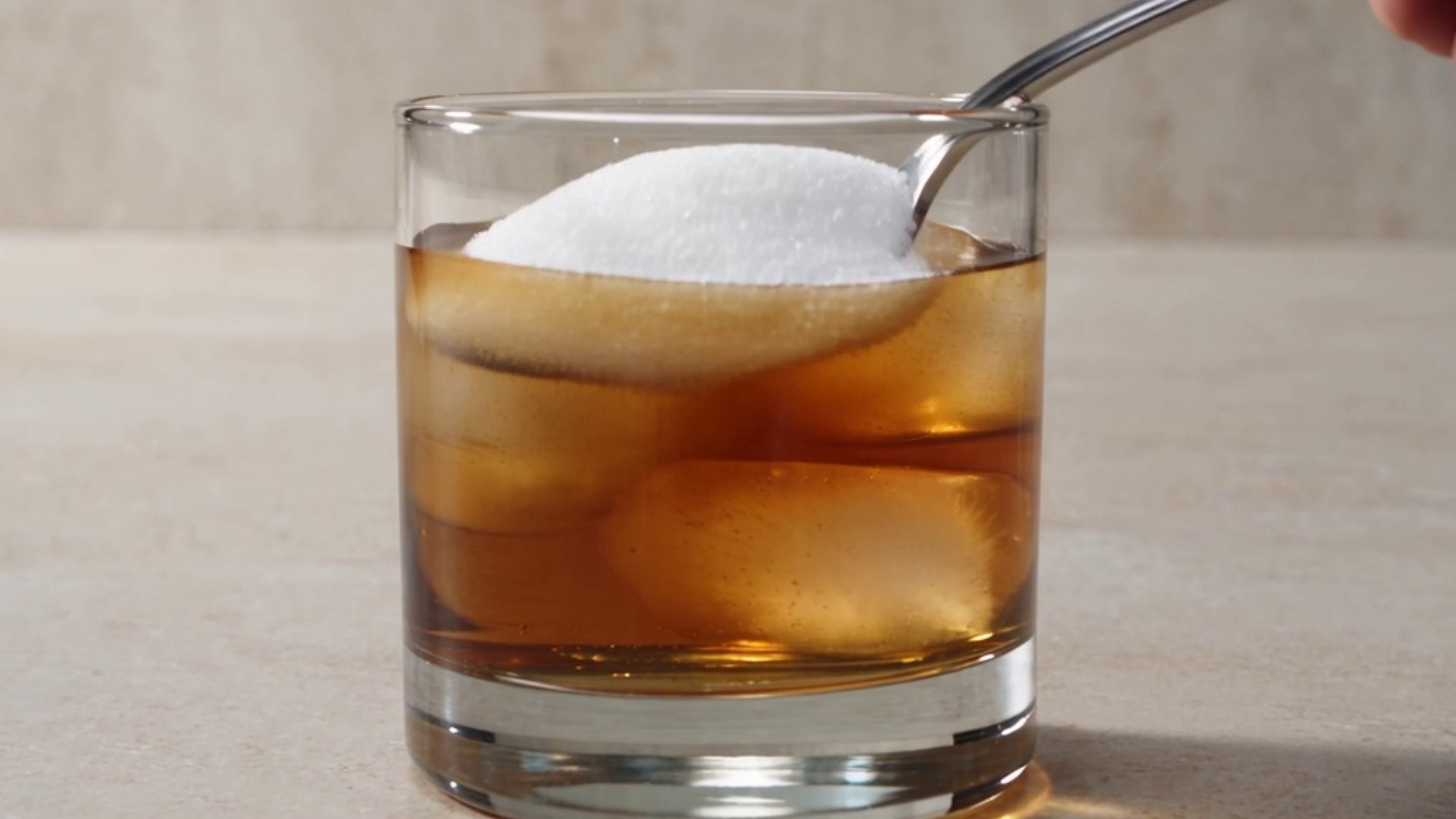}
        & \appcaseframe{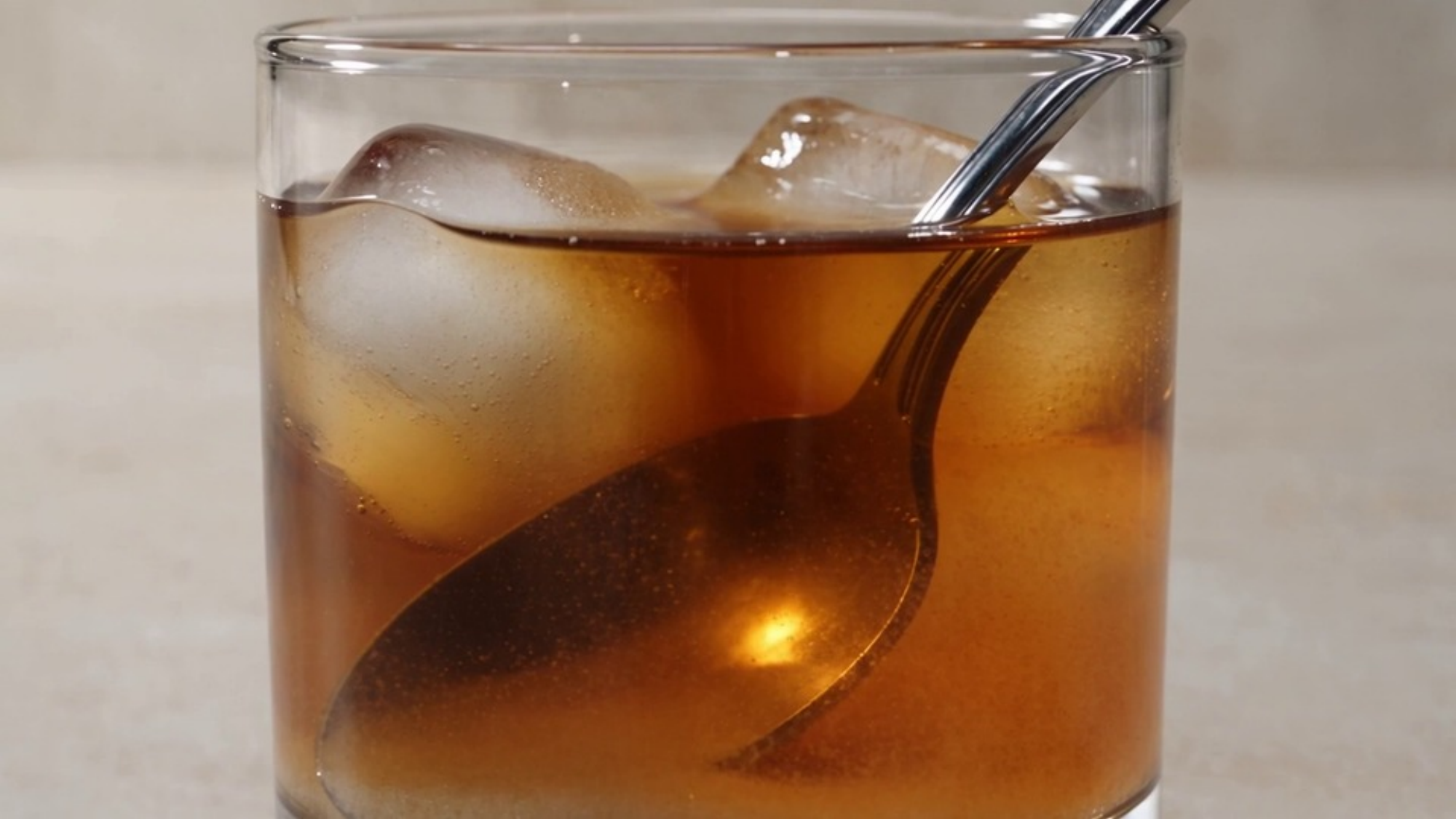}
    \end{tabular}
    \nopagebreak[3]\caption{\textbf{Sugar dissolving in iced tea on Seedance 2.5.} Rewriting makes the spoon enter the tea and the crystals dissolve; the focus pull fails in every row. In the T+O row the compiler selected text rewriting only.}
    \label{fig:case_sd_tea}
    \endgroup
\par\endgroup
\end{appcaseblock}

\clearpage
\begin{appcaseblock}
\noindent\textbf{Coordinated scene transformation.}
Text adaptation helps Wan realize two linked transformations: a building
becomes a manta-ray spacecraft, and a walkway becomes a serpent-like
bridge (Figure~\ref{fig:case_wan_i2v}). Native, T, and T+O satisfy
$0/3$, $2/3$, and $3/3$ requirements, respectively. Both compiled plans
use text alone; the $3/3$ output also fails the realism check.

\par\smallskip\noindent\begingroup\centering
    \centering
    \begingroup
    \appcaseinit
    \appcaseinputas{I2V}{Begin with the pictured aerial view of the futuristic cliffside building, curved orange
    walkway, waterfront plaza, and distant islands. The building lifts from its supports and
    morphs into a manta-ray spacecraft, retaining its white architectural shell while
    unfolding broad organic fins. The orange walkway becomes a serpent-bridge, retaining its
    railings and walking surface while developing a serpentine head and flexing body. With
    both hybrid forms visible, the manta-ray spacecraft glides toward the camera, traveling
    about 30\% of the frame width. The cliff, plaza, water, and islands remain recognizable.}{\appinputimg{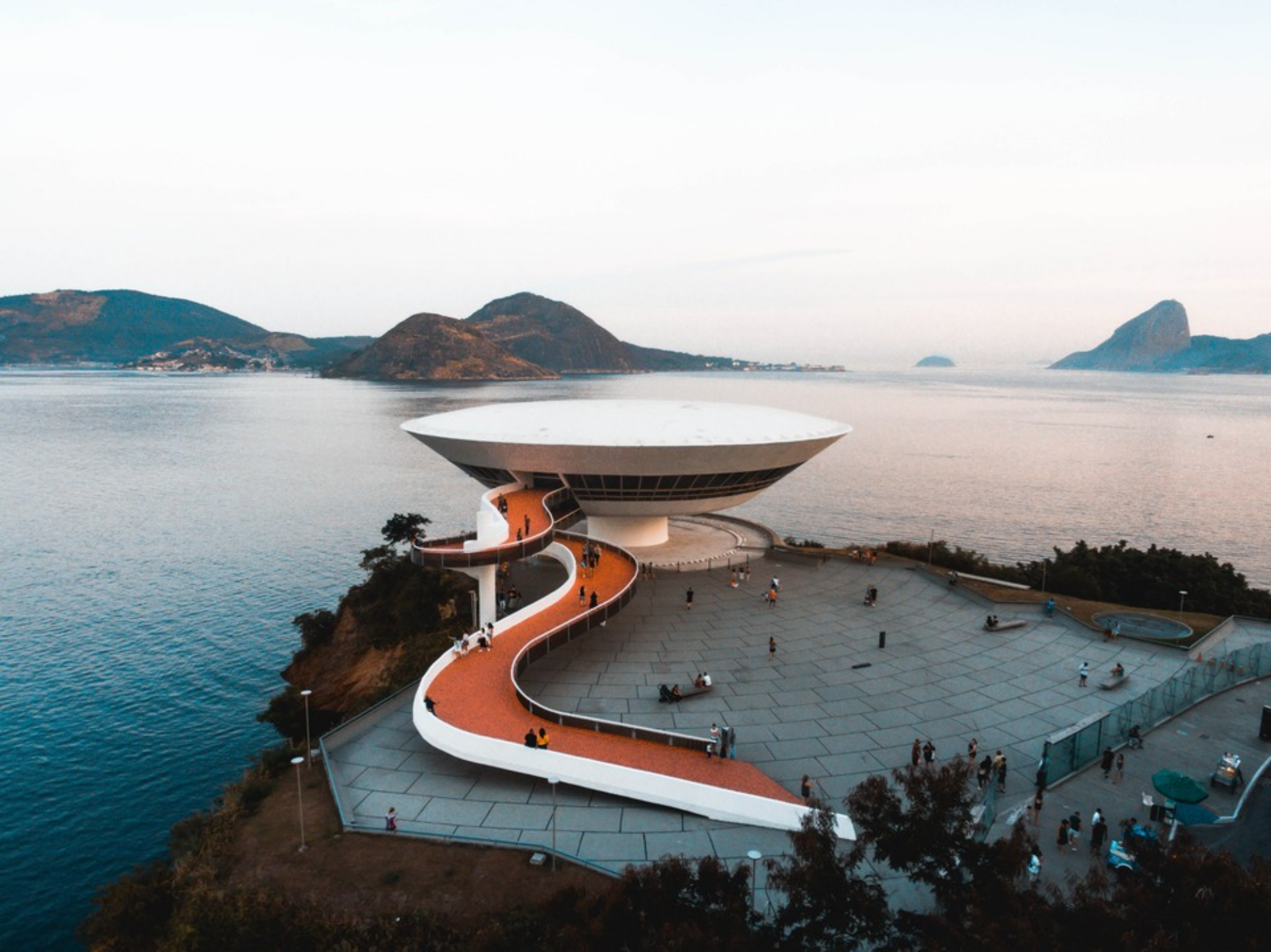}{First frame}%
    }{}{%
        \begin{tabularx}{\linewidth}{@{}XXX@{}}
            \textcolor{appBlue}{\textbf{(1) Both hybrid forms}} & (2) Building becomes the spacecraft & (3) Approach after transforming
        \end{tabularx}}
    \begin{tabular}{@{}c@{\hspace{\appGap}}c@{\hspace{\appGap}}c@{\hspace{\appGap}}c@{}}
        \parbox[c]{\appLabelW}{\centering\textbf{Method}} & \textbf{Early} & \textbf{Middle} & \textbf{Late} \\[2pt]
        \appcaselabel{Native}{\textbf{0/3 passed}}{\textcolor{appFail}{Fails (1)--(3)}}
        & \appcaseframe{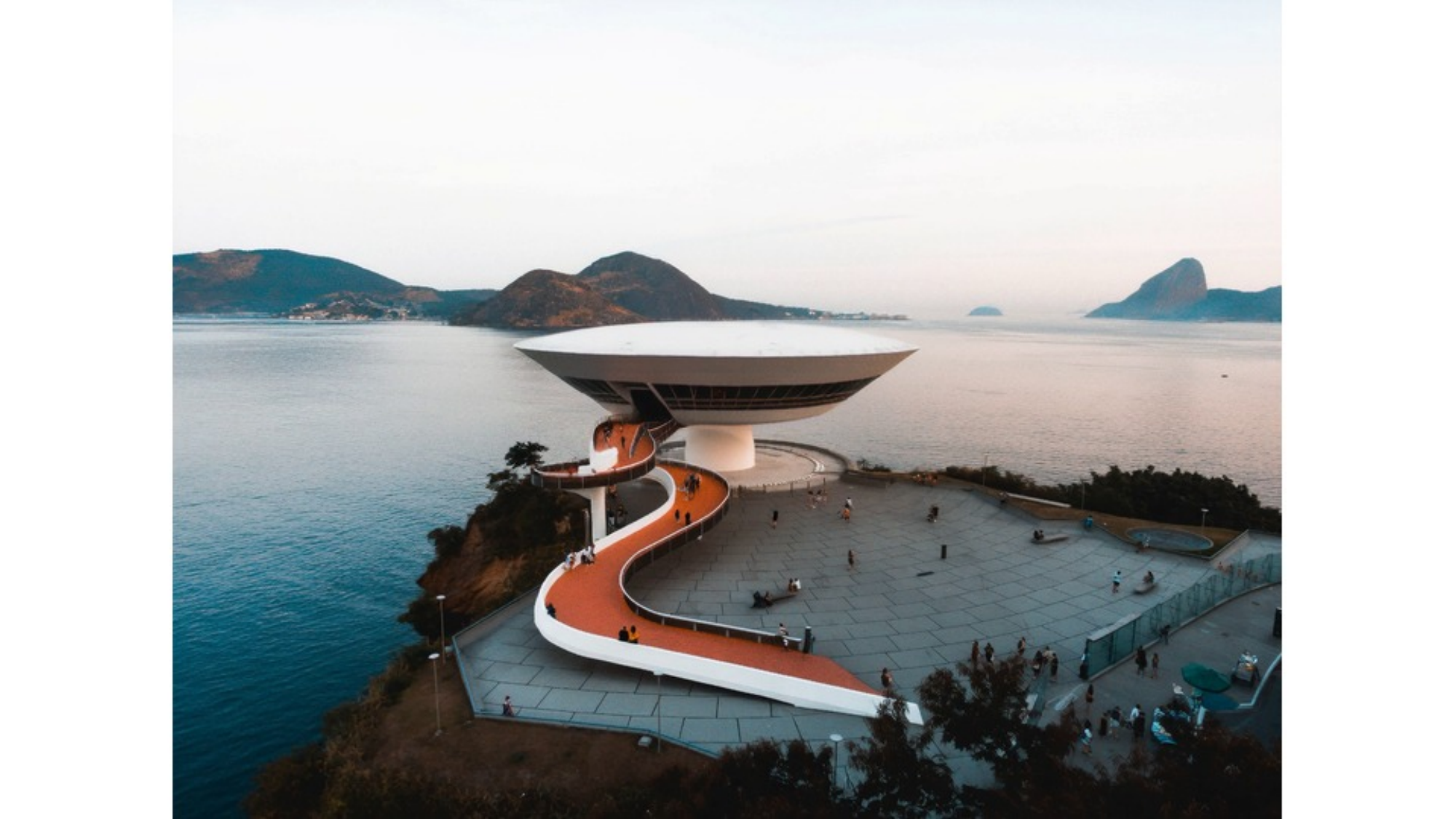}
        & \appcaseframe{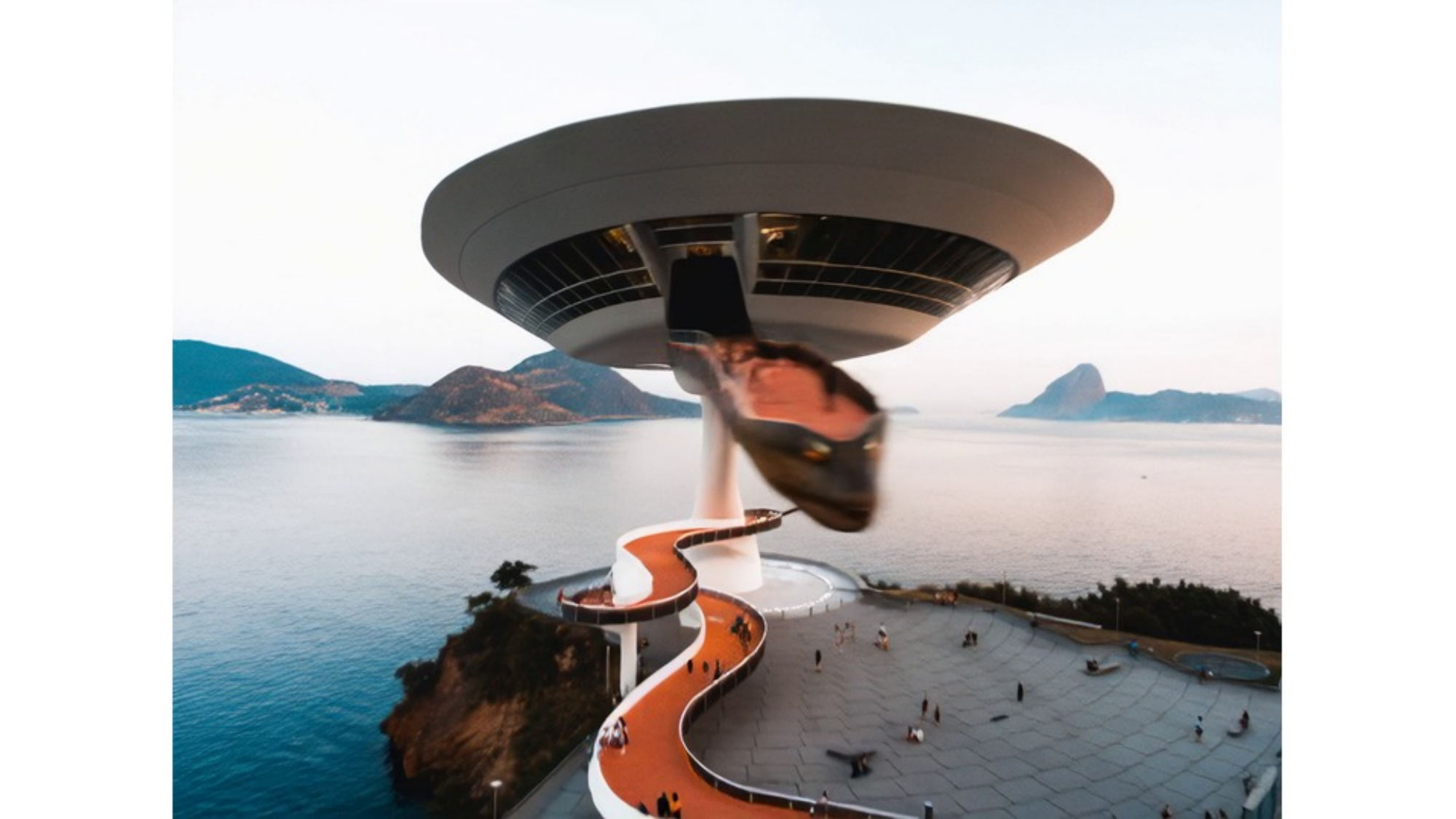}
        & \appcaseframe{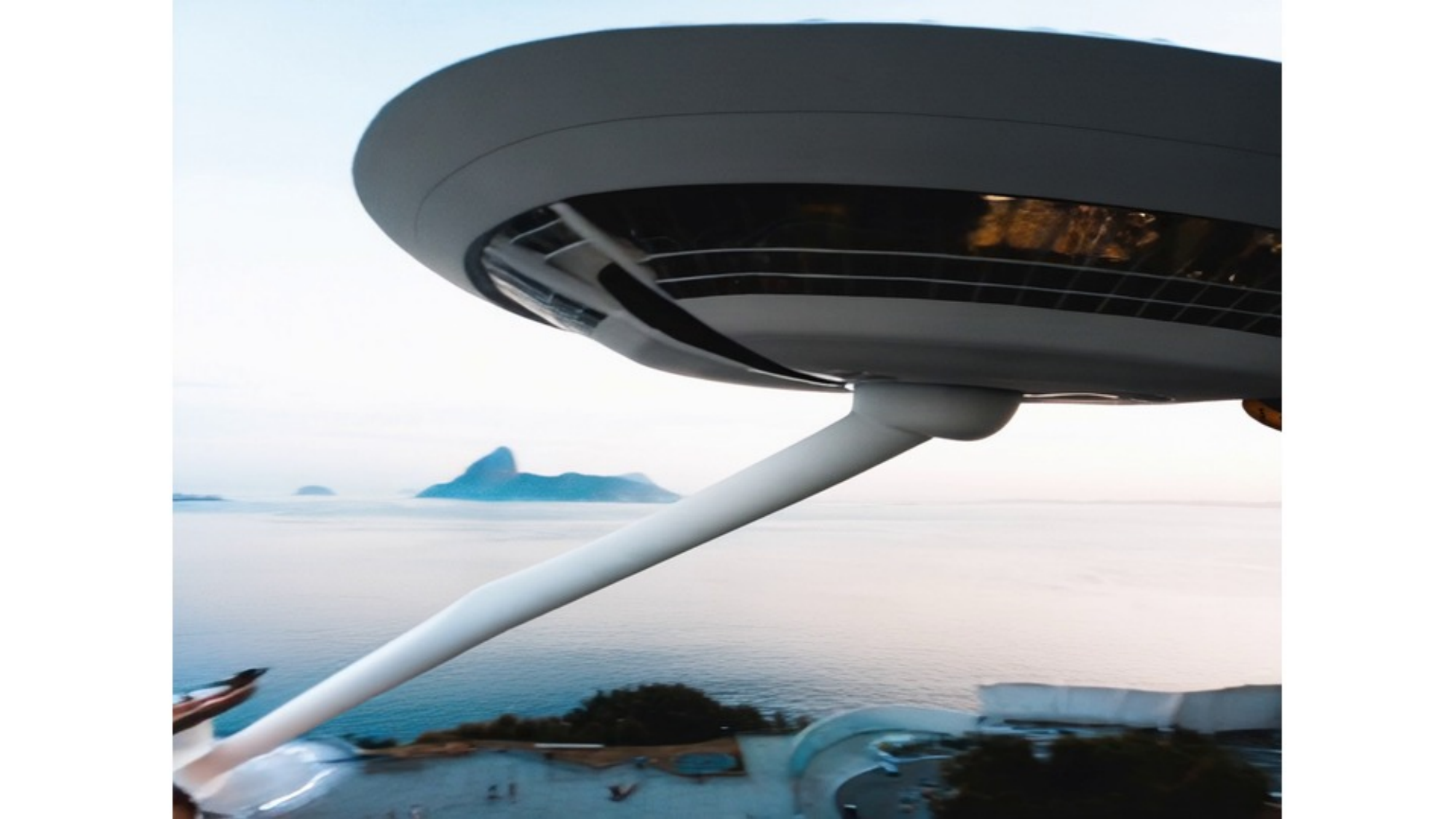}
        \end{tabular}
\par\vspace{3pt}\noindent\begin{tabular}{@{}c@{\hspace{\appGap}}c@{\hspace{\appGap}}c@{\hspace{\appGap}}c@{}}

        \appcaselabel{NeurCo (T)}{\textbf{2/3 passed}}{\textcolor{appFail}{Fails (1)}}
        & \appcaseframe{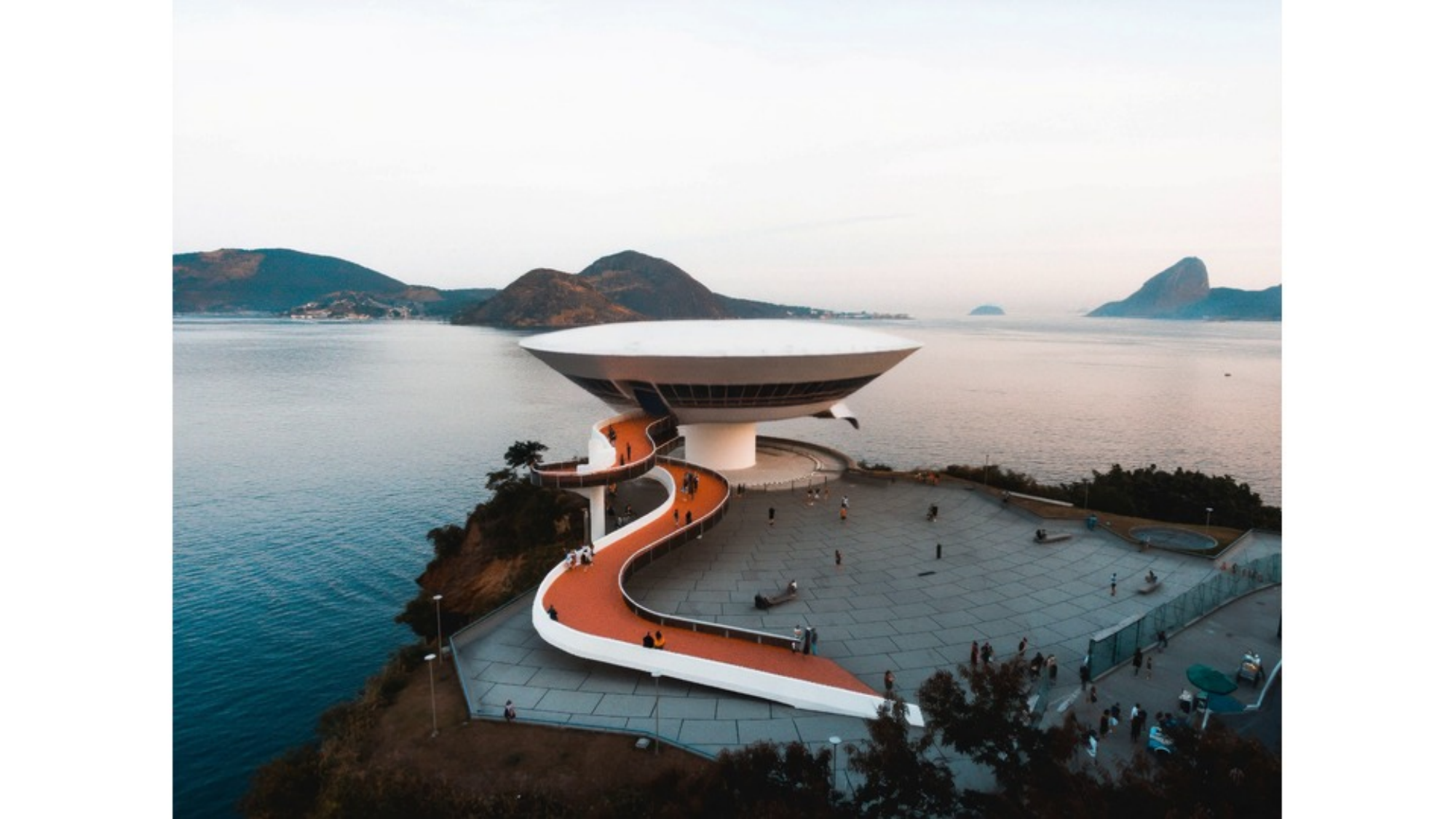}
        & \appcaseframe{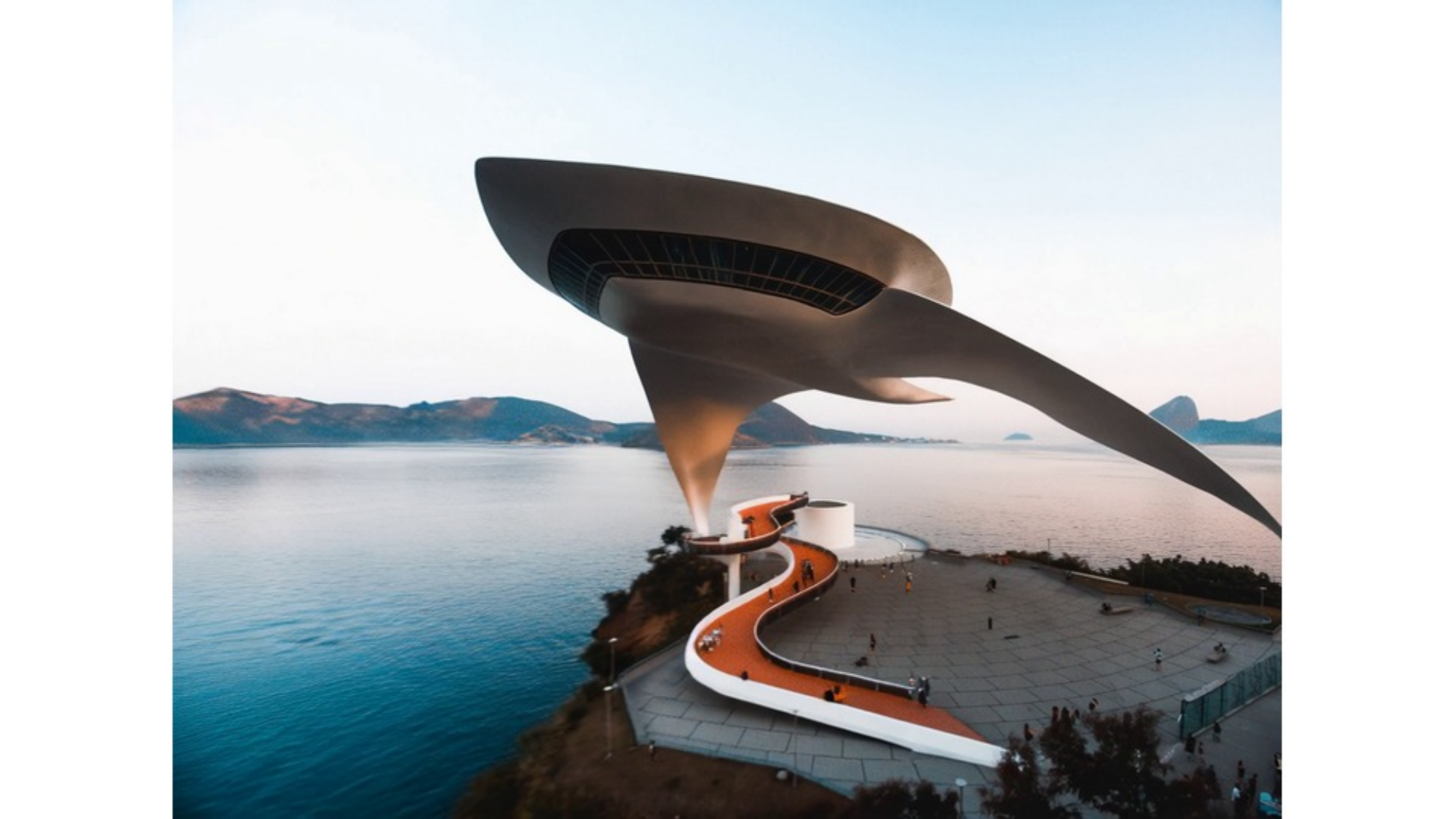}
        & \appcaseframe{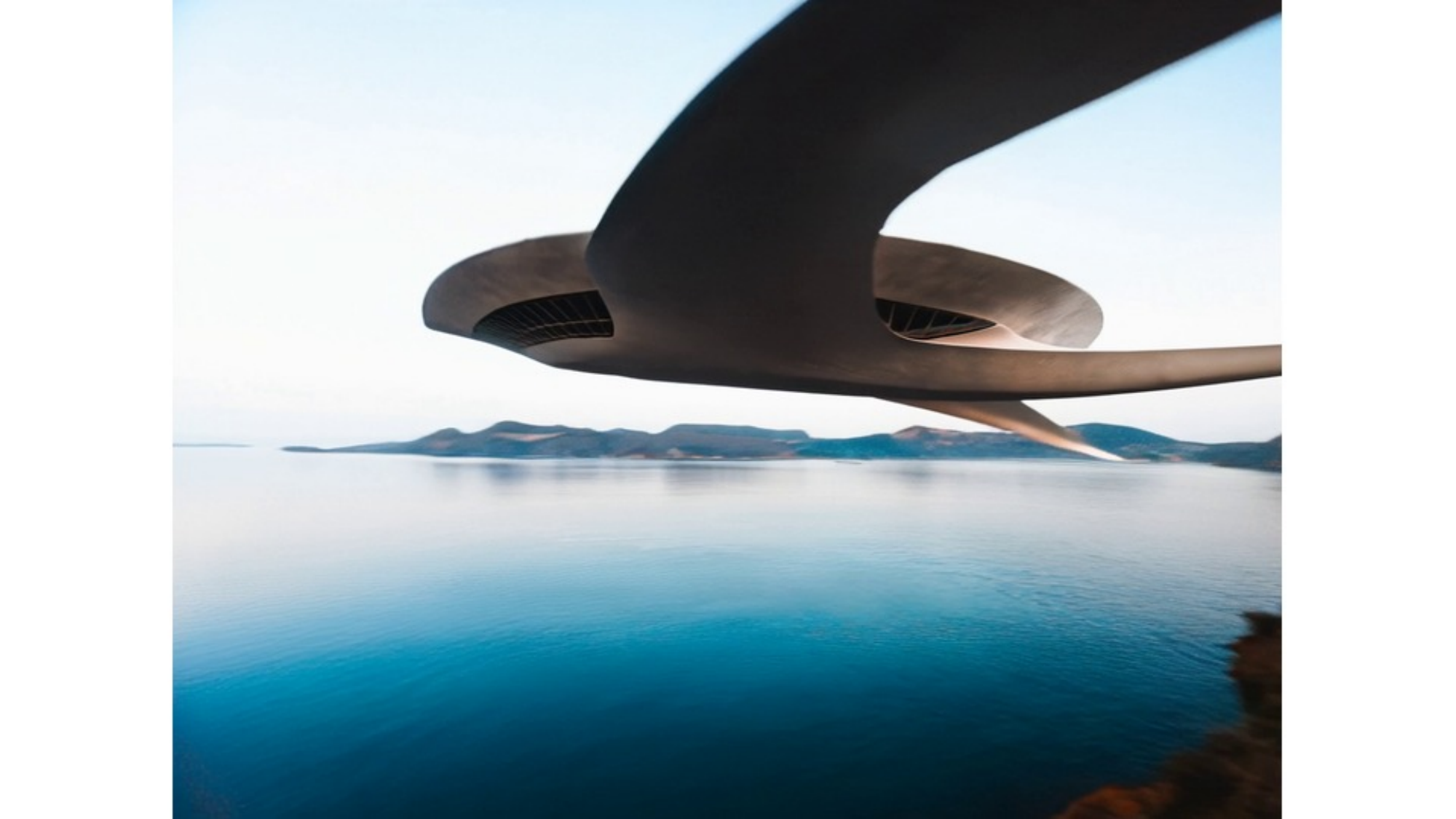}
        \end{tabular}
\par\vspace{3pt}\noindent\begin{tabular}{@{}c@{\hspace{\appGap}}c@{\hspace{\appGap}}c@{\hspace{\appGap}}c@{}}

        \appcaselabel{NeurCo (T+O)}{\textcolor{appBlue}{\textbf{3/3 passed}}}{\textcolor{appFail}{\textbf{Realism fails}}}
        & \appcaseframe{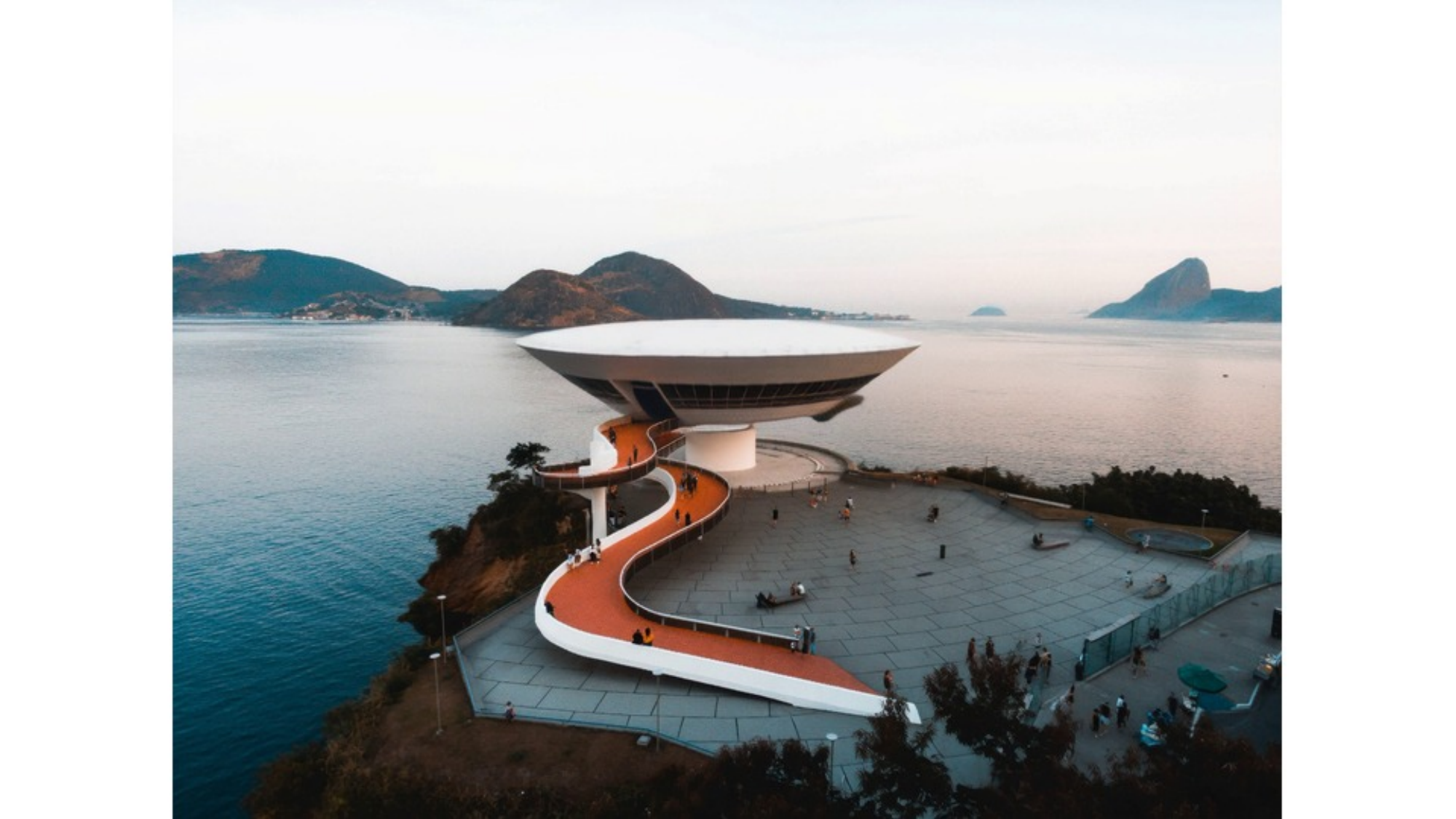}
        & \appcaseframe{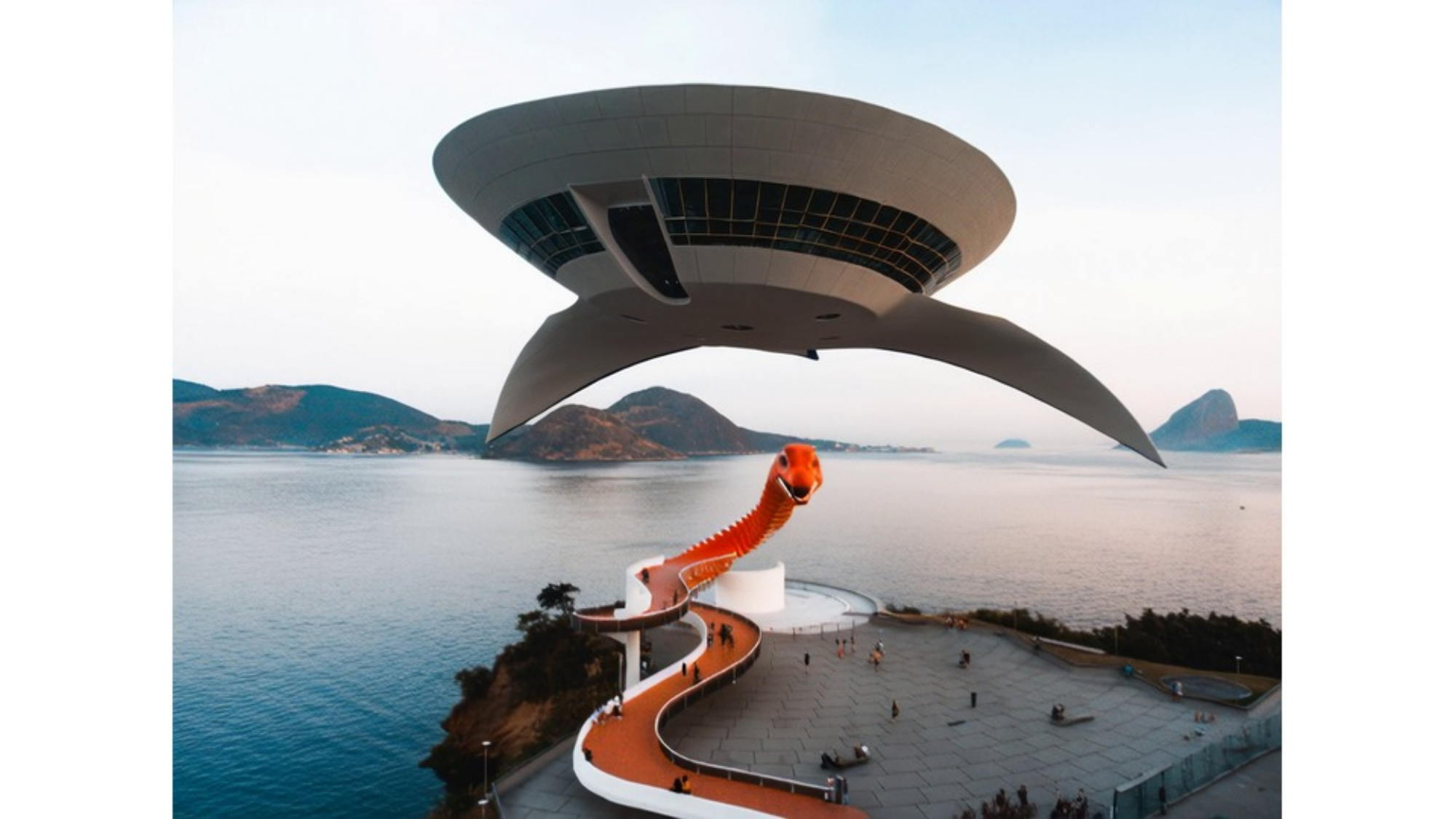}
        & \appcaseframe{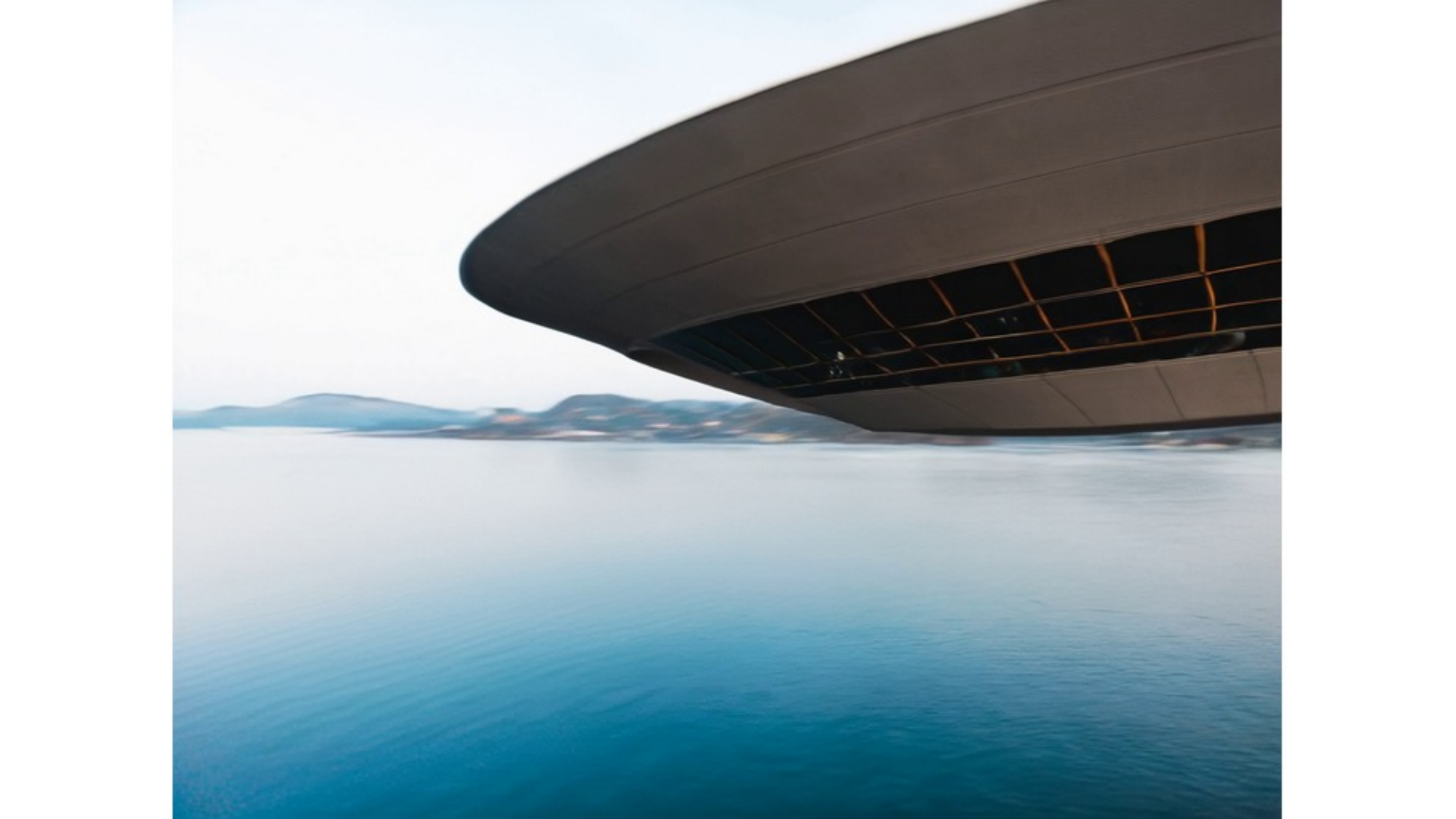}
    \end{tabular}
    \nopagebreak[3]\caption{\textbf{Text rewriting recovers the hybrid scene.}
    NeurCo (T+O) satisfies all three requirements but fails the realism
    check. Both compiled variants use text rewriting without an operator.}
    \label{fig:case_wan_i2v}
    \endgroup
\par\endgroup
\end{appcaseblock}

\clearpage
\begin{appcaseblock}
\noindent\textbf{Creature transformation with a scissors trim.}
Figure~\ref{fig:case_sd_creature} shows how text compilation improves a multi-stage transformation on Seedance~2.5. Native produces the final shoe without the scissor trim ($1/6$). NeurCo (T) recovers the trim and event order but misses the falling tuft ($4/6$). NeurCo (T+O) also selects text rewriting alone; its output shows the creature's eyes and the falling tuft, satisfying $5/6$ requirements.

\par\smallskip\noindent\begingroup\centering
    \centering
    \begingroup
    \appcaseinit
    \appcaseinputas{R2V}{Among the flowers beneath pink clouds, the green shoe-shaped plant in Image 1 is a living plant creature, with two small eyes nestled in its moss, resting on the pedestal. In slow motion, three moments unfold in sequence: the girl carefully closes her scissors to trim one tuft; the severed tuft falls onto the pedestal; then the creature transforms into the pale pink satin shoe in Image 2. Its eyes merge into the surface as its green covering becomes pale pink satin, leaving the finished shoe standing steadily.}{\appinputimg{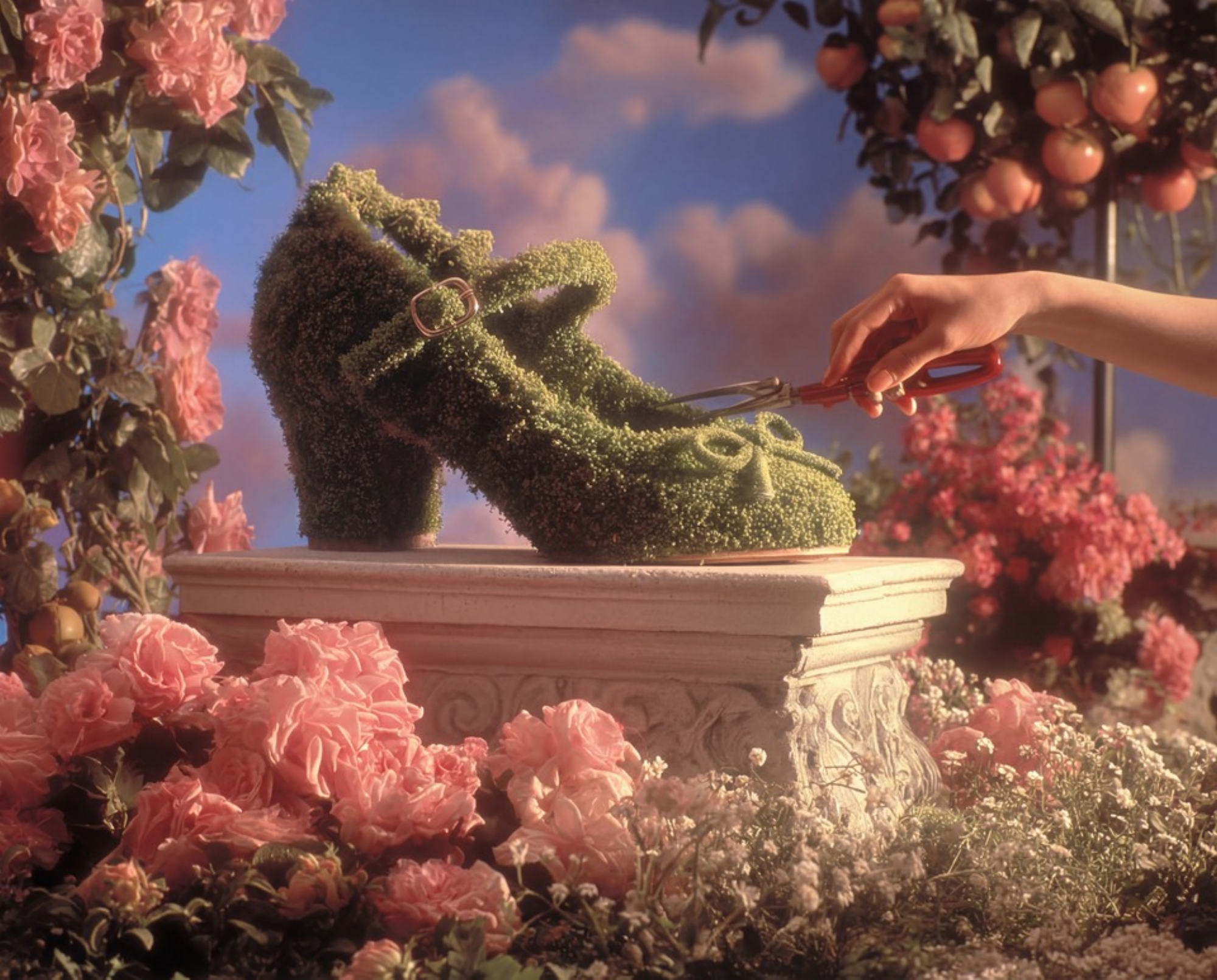}{Image 1}%
        \appinputimg{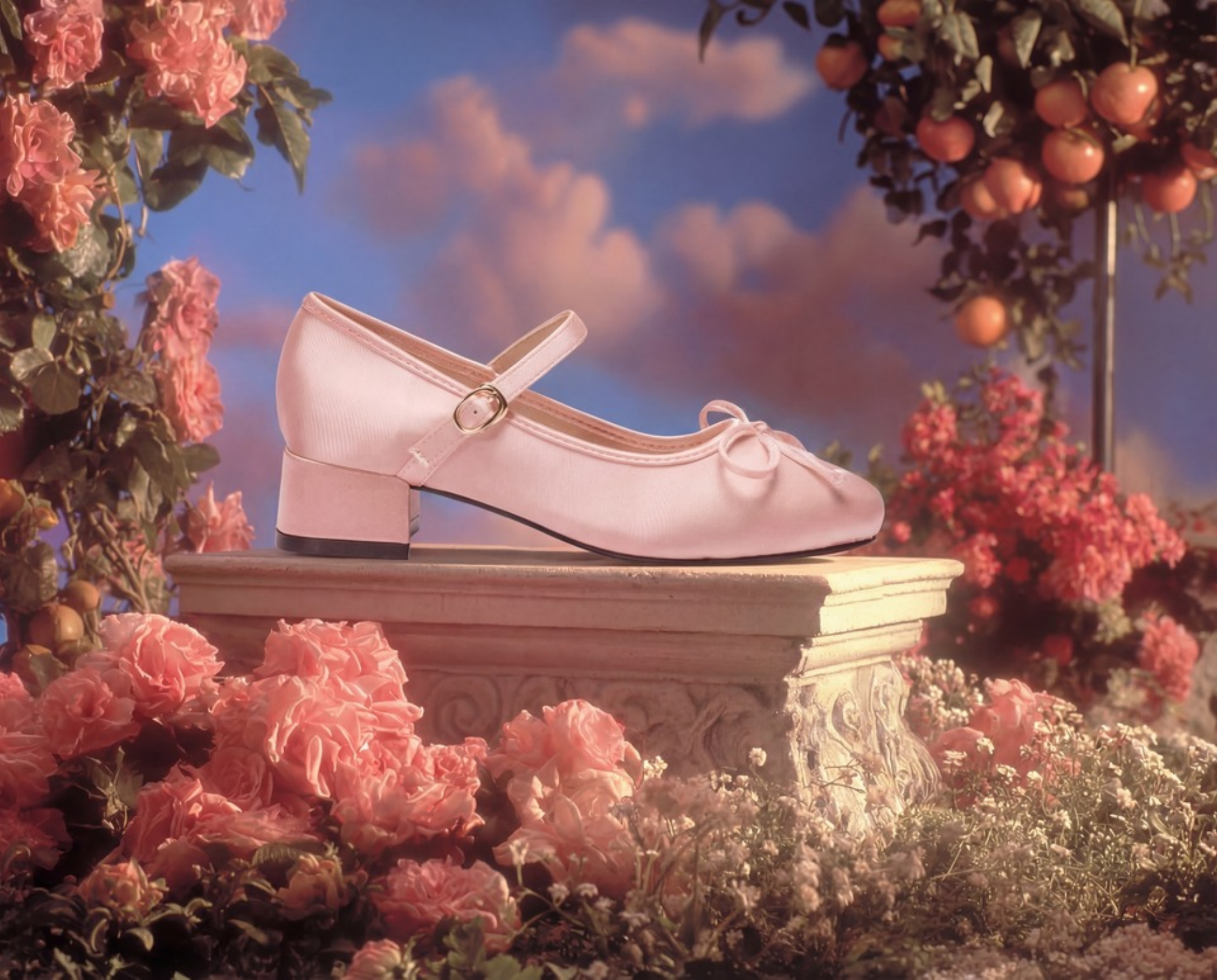}{Image 2}%
        }{}{%
        \begin{tabularx}{\linewidth}{@{}XXX@{}}
            (1) Creature transforms & \textcolor{appBlue}{\textbf{(2) Careful scissor trim}} & (3) Tuft falls onto pedestal \\
            \textcolor{appBlue}{\textbf{(4) Trim, fall, transform in order}} & \textcolor{appBlue}{\textbf{(5) Slow motion}} & (6) Green to pale pink
        \end{tabularx}}
    \begin{tabular}{@{}c@{\hspace{\appGap}}c@{\hspace{\appGap}}c@{\hspace{\appGap}}c@{}}
        \parbox[c]{\appLabelW}{\centering\textbf{Method}} & \textbf{Early} & \textbf{Middle} & \textbf{Late} \\[2pt]
        \appcaselabel{Native}{\textbf{1/6 passed}}{\textcolor{appFail}{Fails (1), (2), (3), (4), (5)}}
        & \appcaseframe{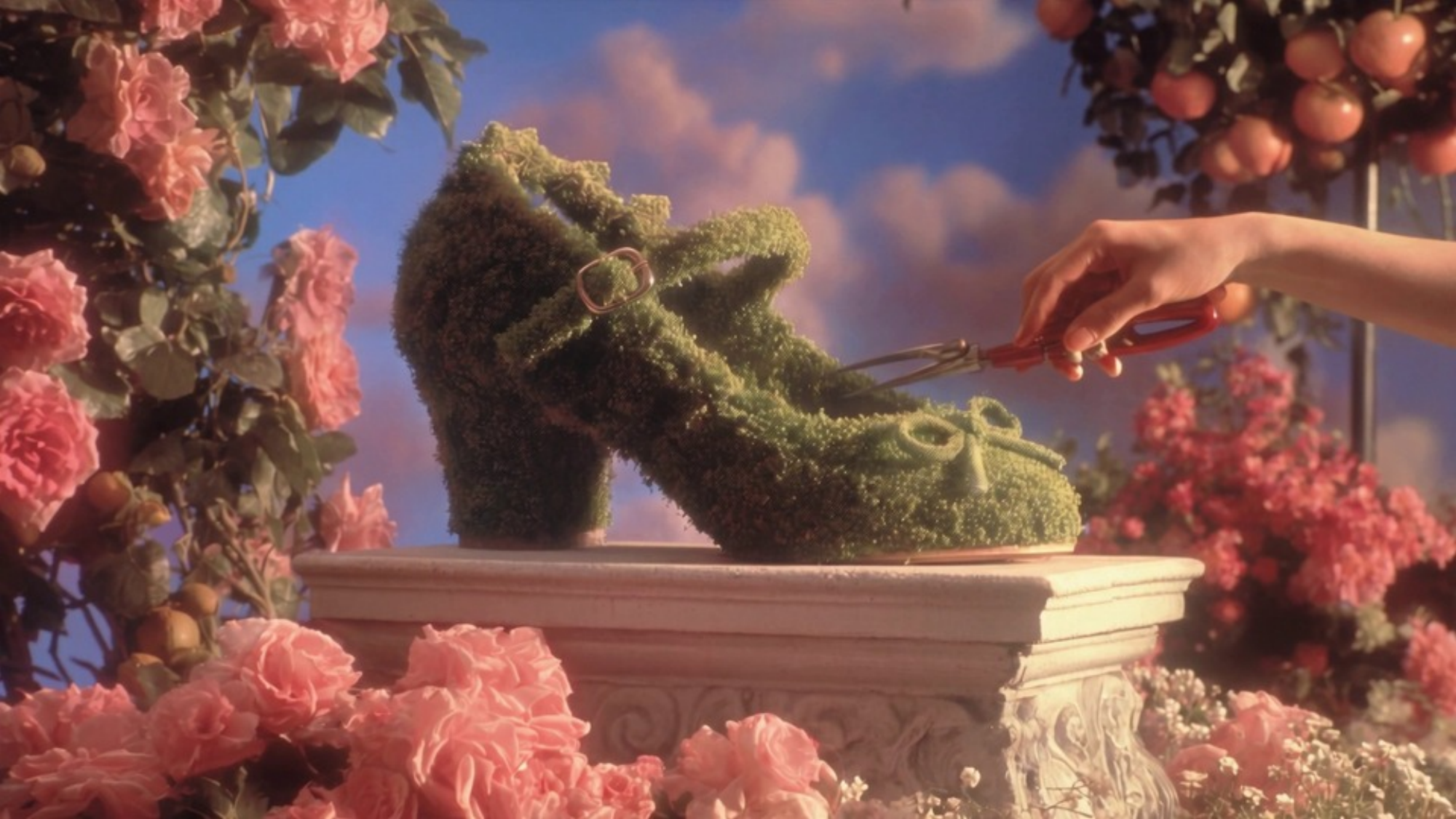}
        & \appcaseframe{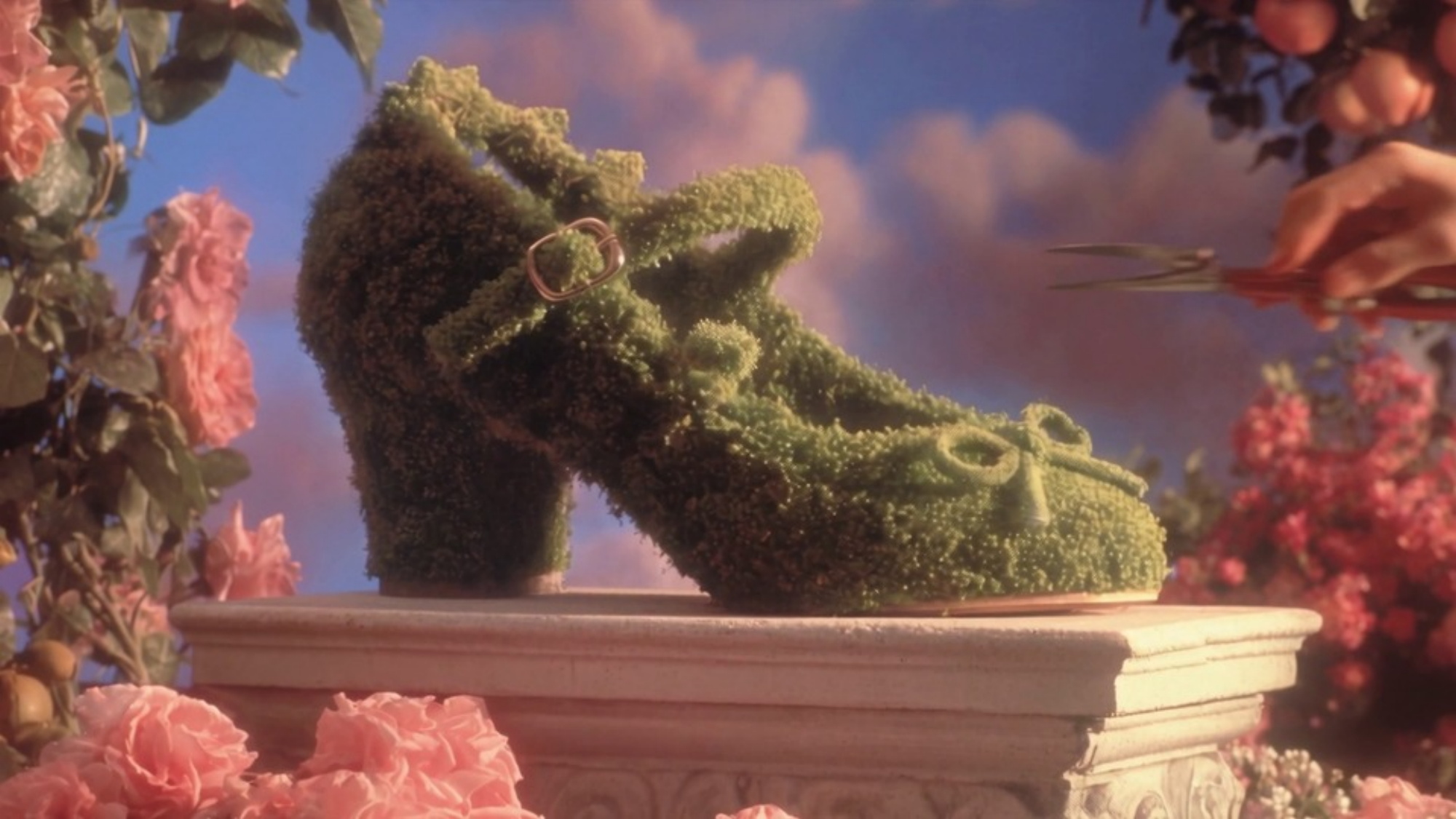}
        & \appcaseframe{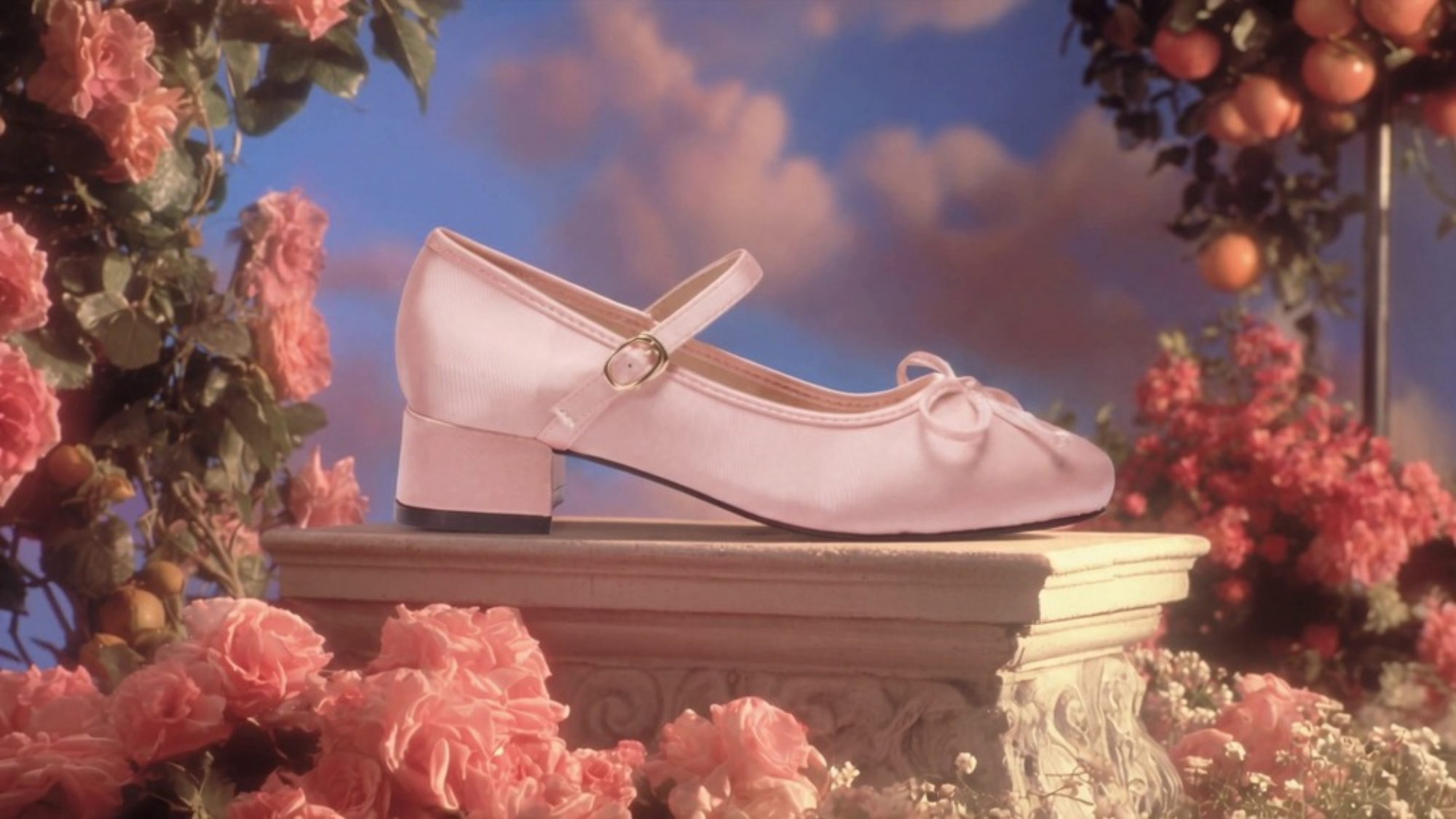}
    \end{tabular}
\par\vspace{3pt}\noindent\begin{tabular}{@{}c@{\hspace{\appGap}}c@{\hspace{\appGap}}c@{\hspace{\appGap}}c@{}}
        \appcaselabel{NeurCo (T)}{\textbf{4/6 passed}}{\textcolor{appFail}{Fails (1), (3)}}
        & \appcaseframe{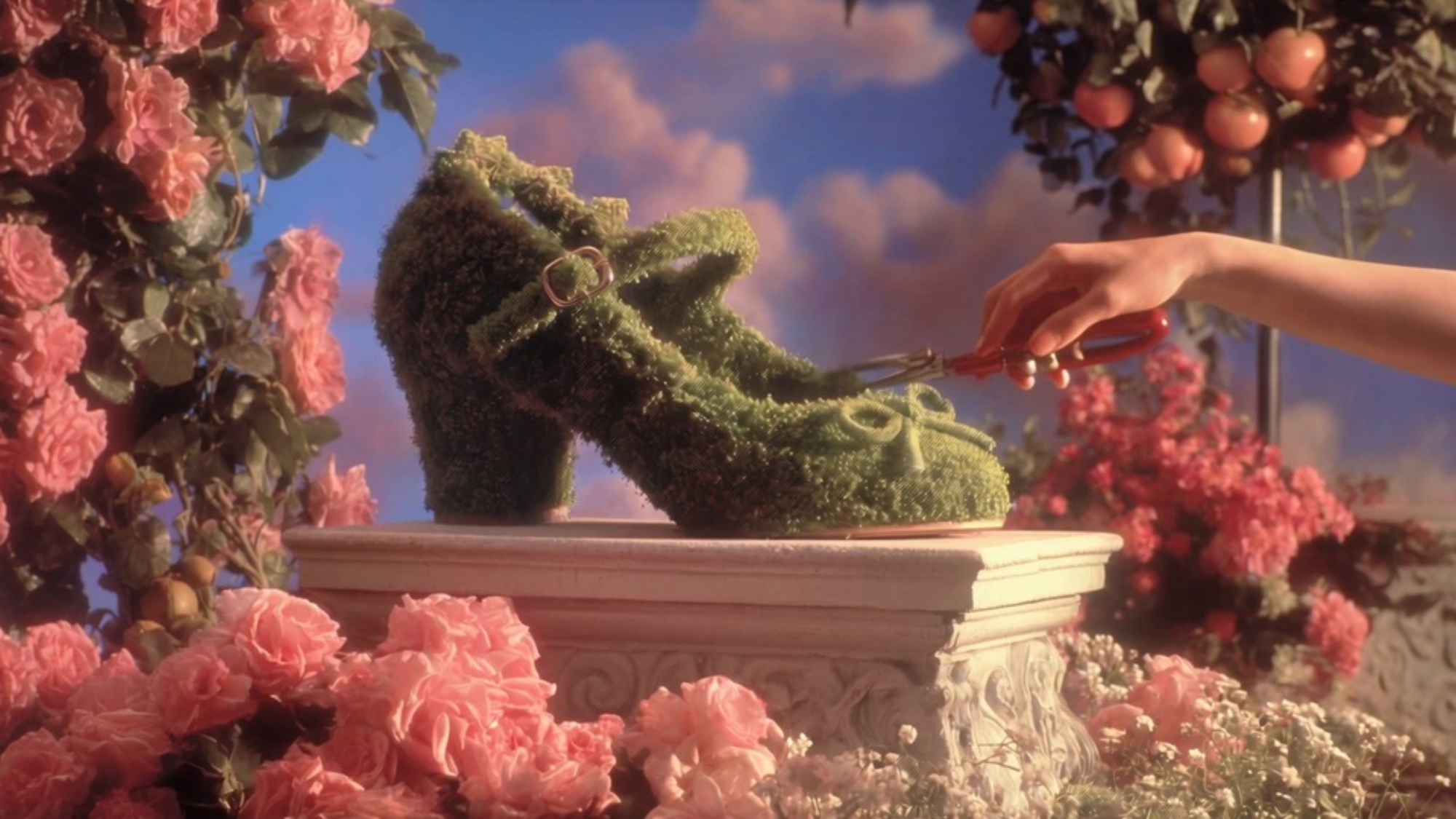}
        & \appcaseframe{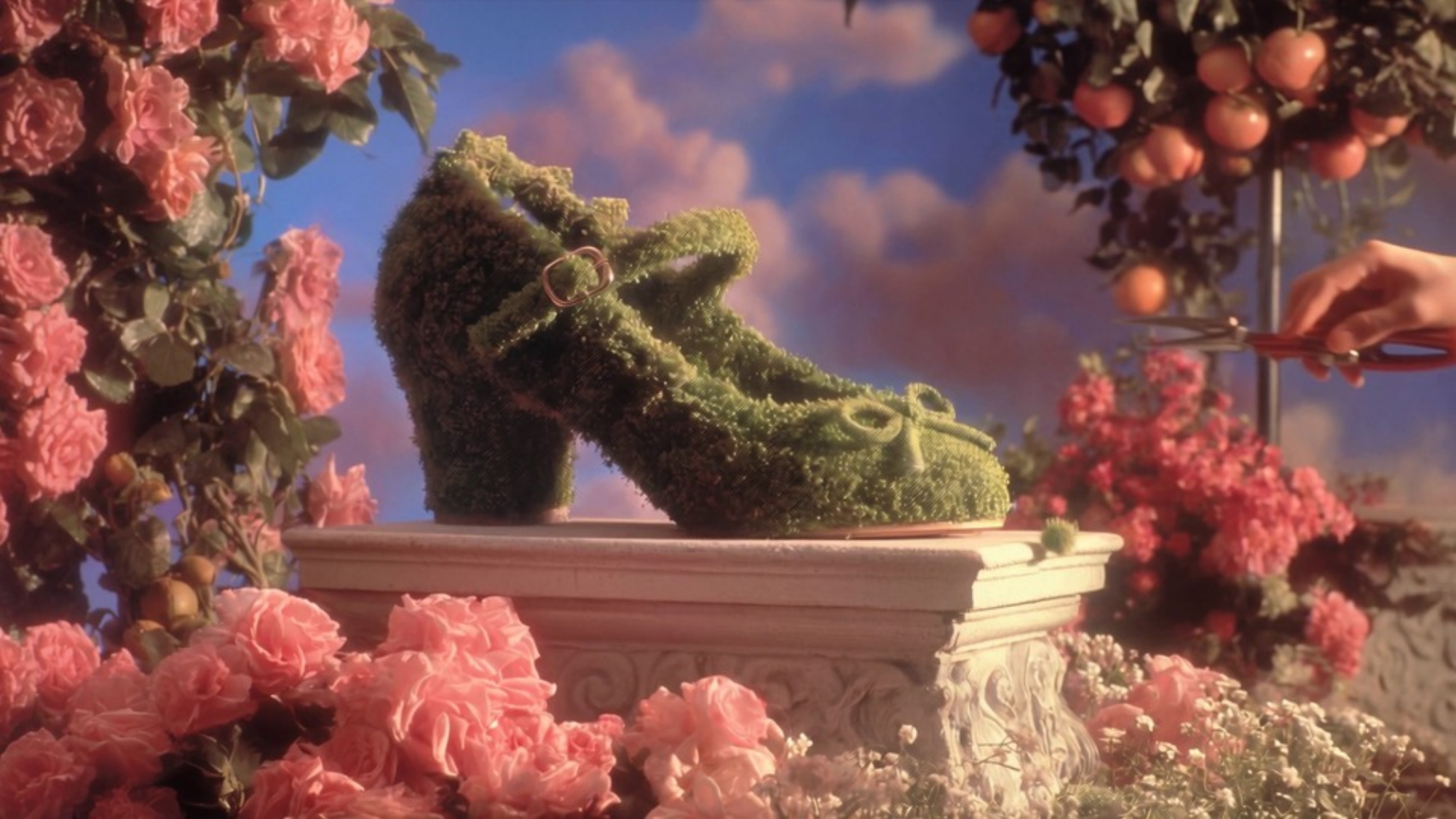}
        & \appcaseframe{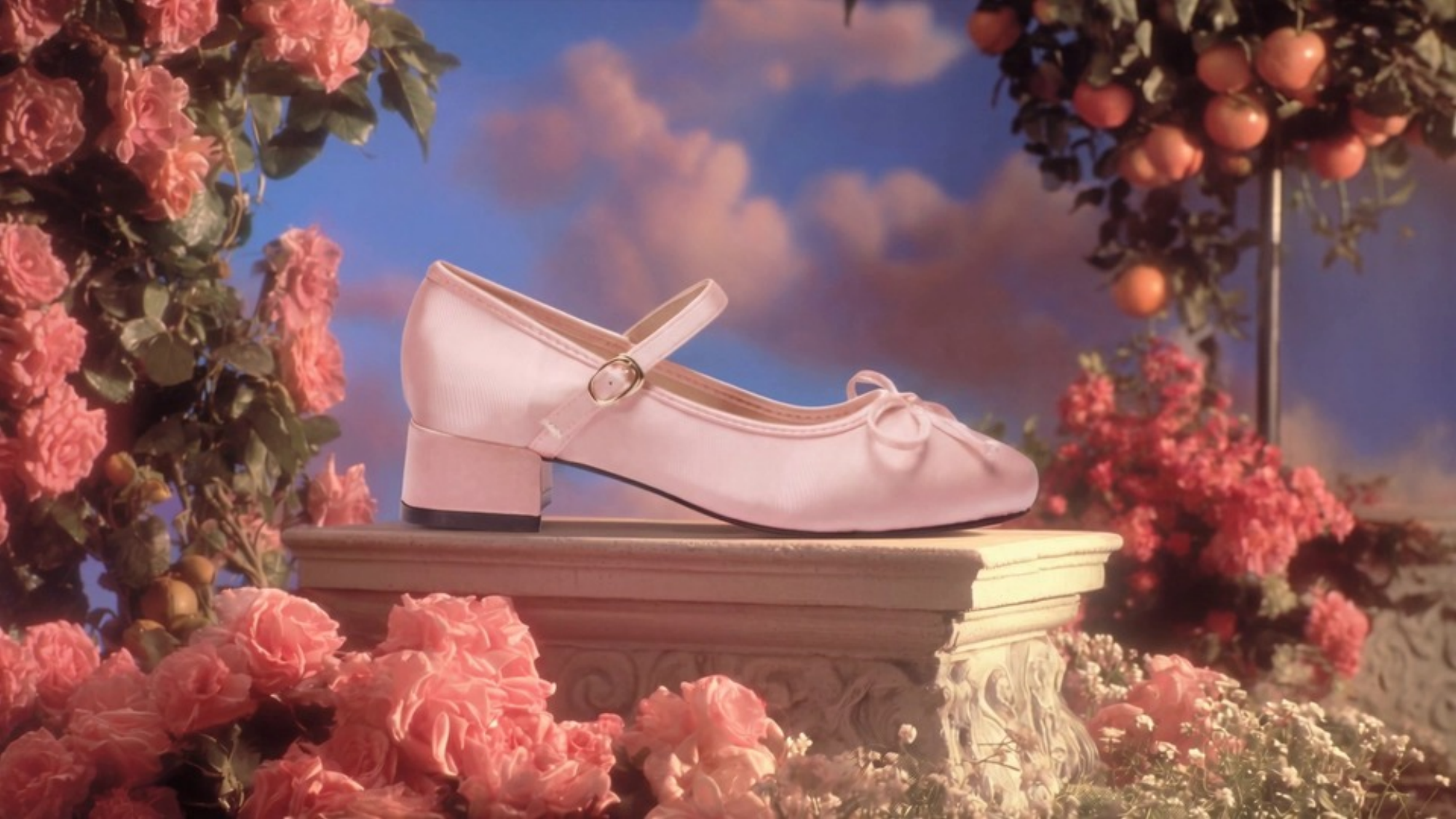}
    \end{tabular}
\par\vspace{3pt}\noindent\begin{tabular}{@{}c@{\hspace{\appGap}}c@{\hspace{\appGap}}c@{\hspace{\appGap}}c@{}}
        \appcaselabel{NeurCo (T+O)}{\textcolor{appBlue}{\textbf{5/6 passed}}}{Text only; (5) unclear}
        & \appcaseframe{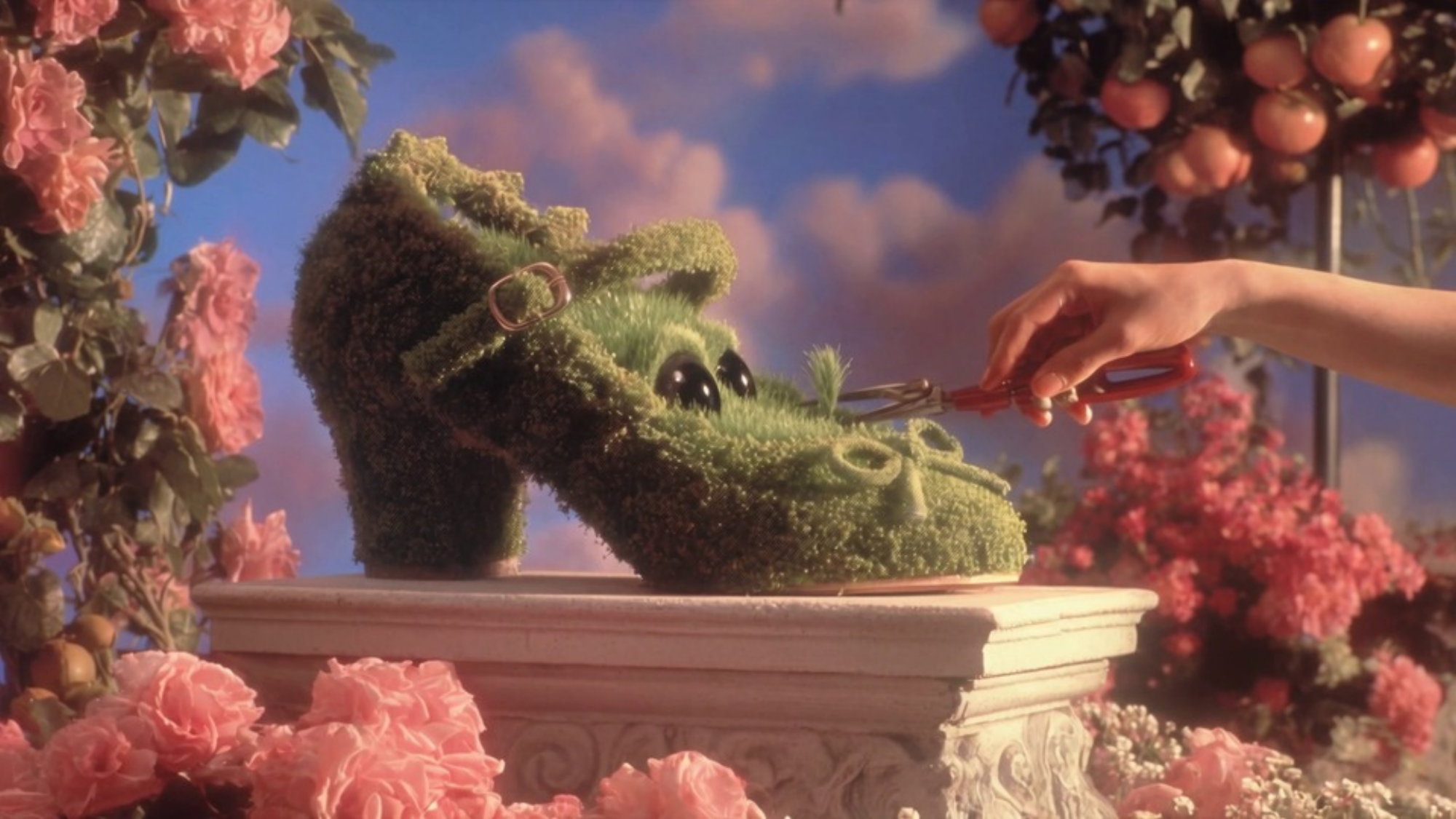}
        & \appcaseframe{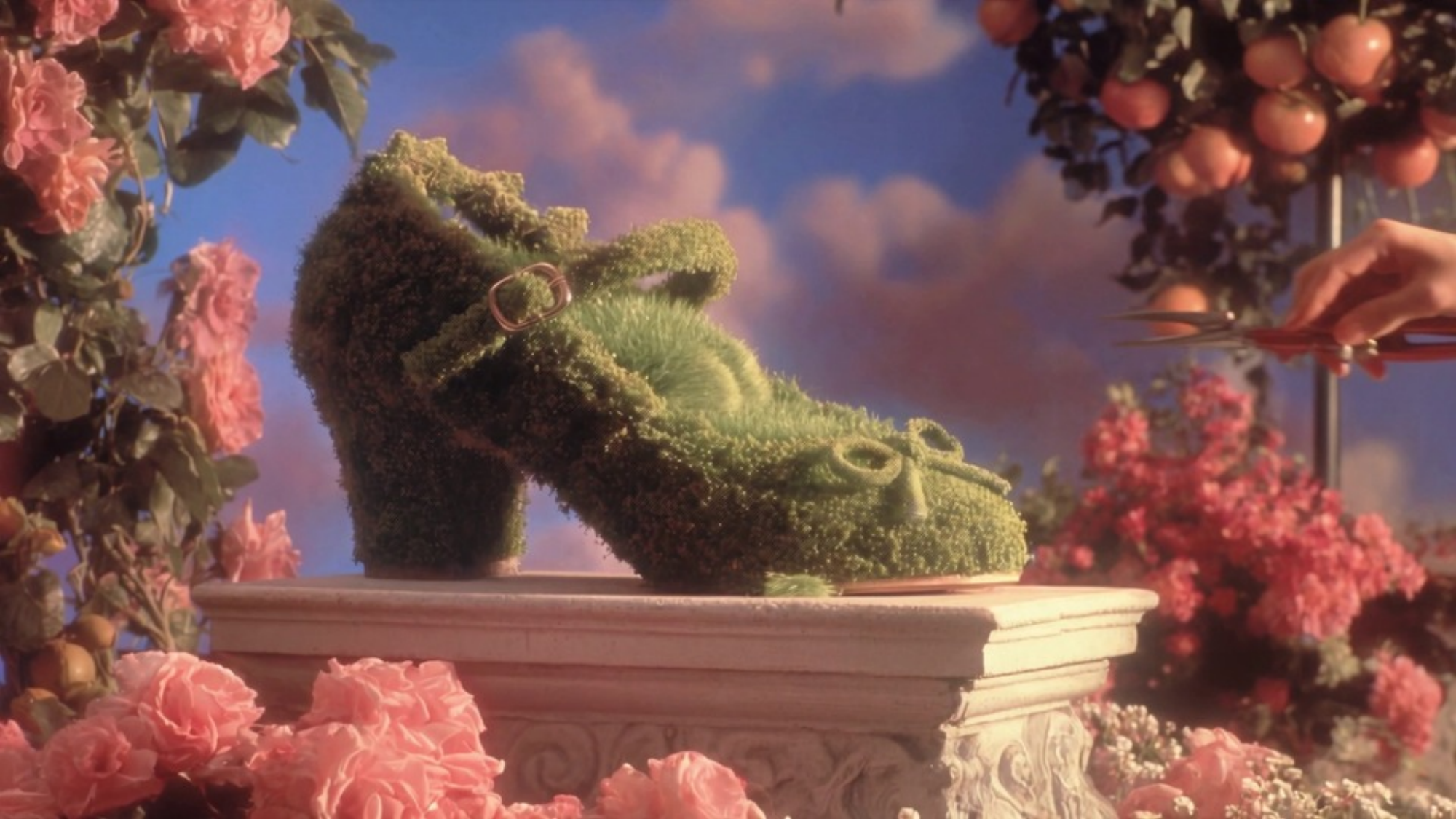}
        & \appcaseframe{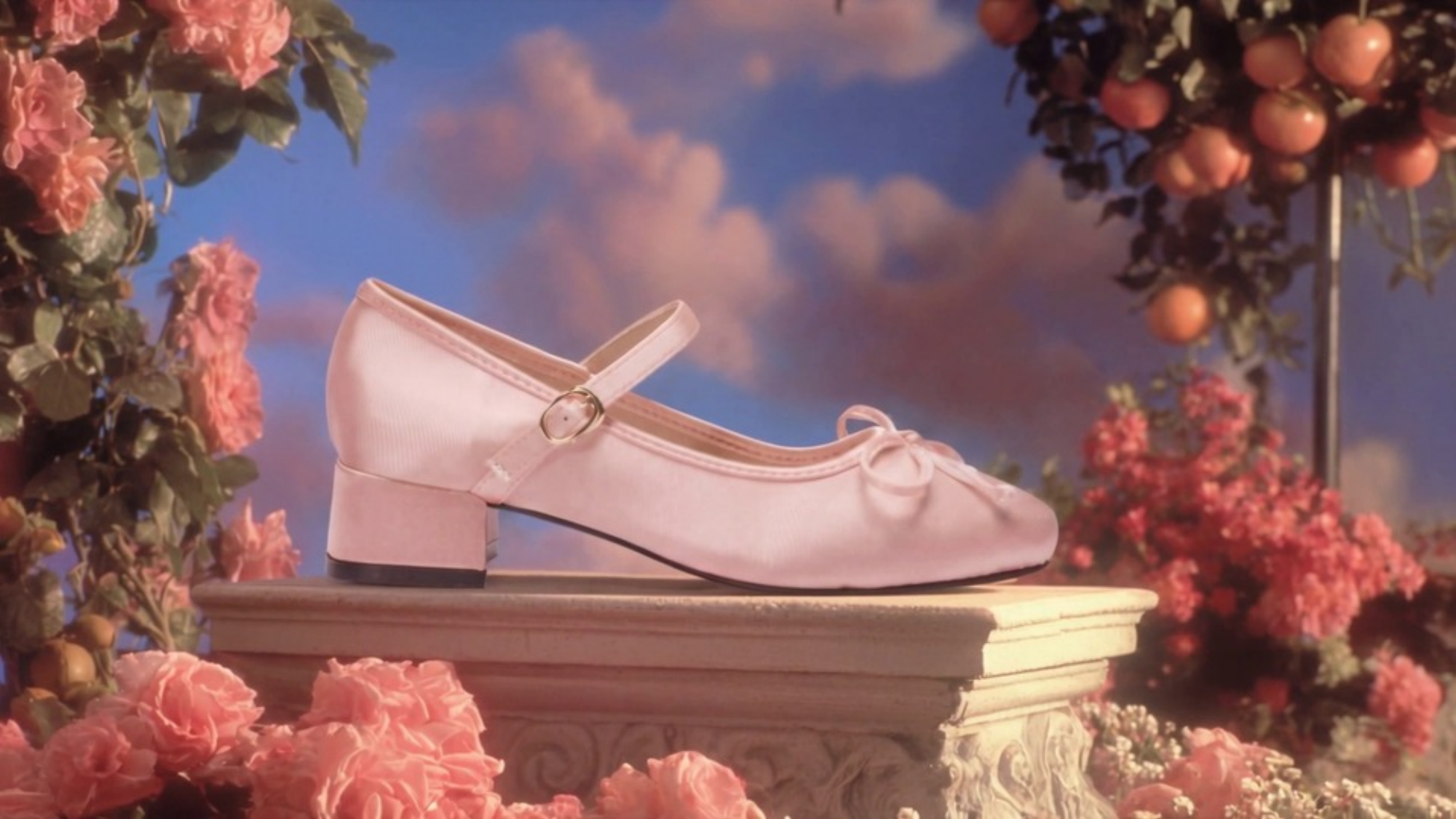}
    \end{tabular}
    \nopagebreak[3]\caption{\textbf{Text compilation improves the transformation sequence on Seedance 2.5.} NeurCo (T) recovers the trim and event order ($4/6$), while the T+O text-only plan also captures the falling tuft and visible eyes ($5/6$).}
    \label{fig:case_sd_creature}
    \endgroup
\par\endgroup
\end{appcaseblock}

\clearpage
\subsection{Visual Conditioning and Compositional Control}
\label{sec:app_cases_visual}

Visual references strengthen character and scene grounding when text
alone leaves attributes ambiguous. The Seedance~2.5 cases cover character binding, event sequencing,
attribute control, and visible text through reference images or first frames.

\begin{appcaseblock}
\noindent\textbf{Reference-guided attribute binding.}
An added visual reference strengthens character-attribute binding in
the Seedance~2.5 battle scene (Figure~\ref{fig:case_sd_r2v}).
Native and text-only compilation satisfy $4/6$ requirements, while the
reference-conditioned plan reaches $5/6$ by resolving the four heroes'
attributes. The five-step action sequence remains incorrect.

\par\smallskip\noindent\begingroup\centering
    \centering
    \begingroup
    \appcaseinit
    \appcaseinputas{R2V}{Anime in underwater stone ruins. The boy in Image 1 has blue hair and eagle wings; the
    girl in Image 2 has red buns. The blonde girl wears purple armor; the burgundy-haired
    girl wears a cactus shirt. The mermaid queen has pale-gray skin, grass-green hair, a
    teal helmet, a green forehead jewel, teal armor, and a fish tail. First, she thrusts her
    trident. The boy parries with his wings. The red-haired girl punches the queen. The
    queen swings her trident, trapping all four heroes inside a pale-blue energy bubble.
    Finally, all four strike the bubble together.}{\appinputimg{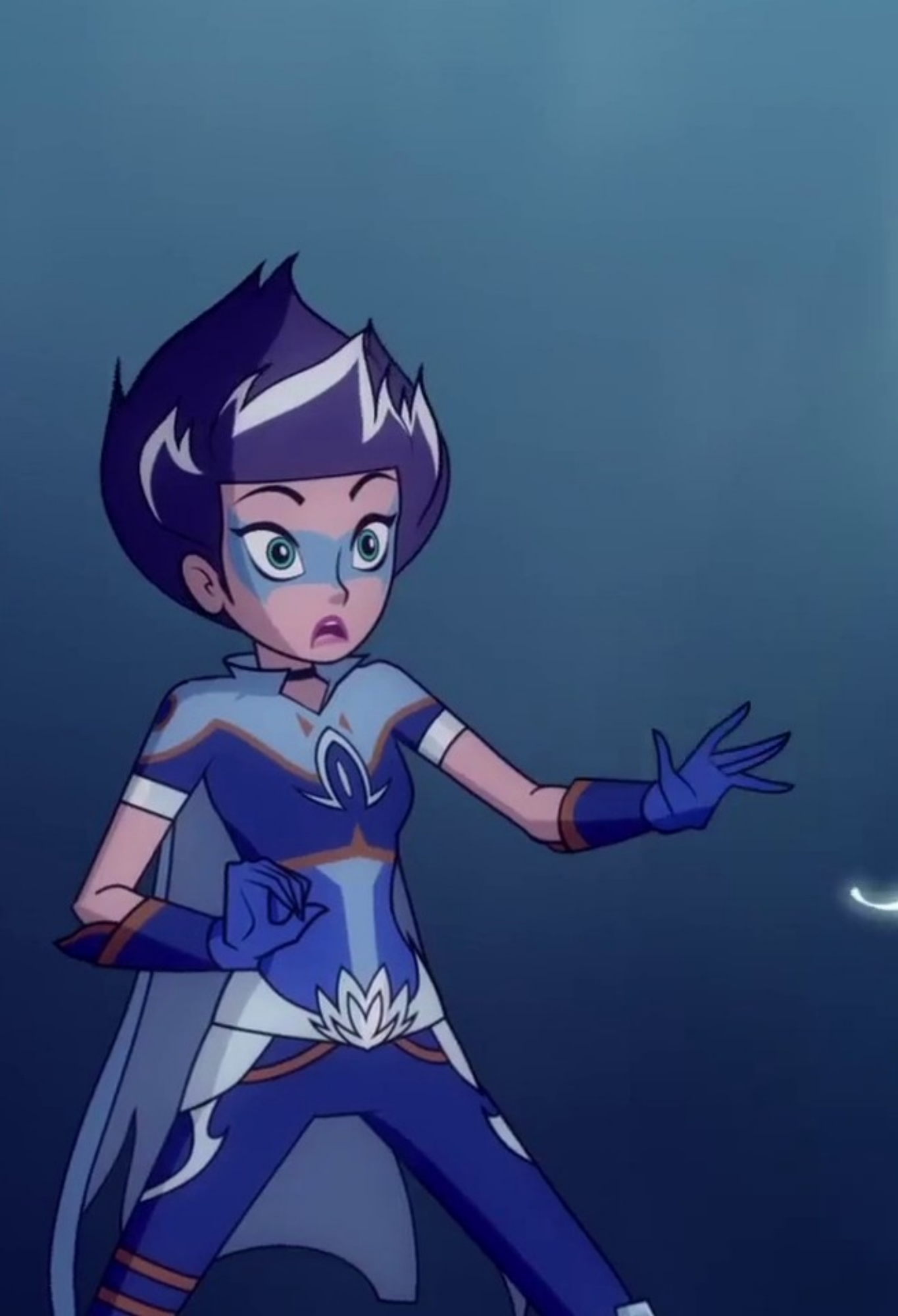}{Image 1}%
        \appinputimg{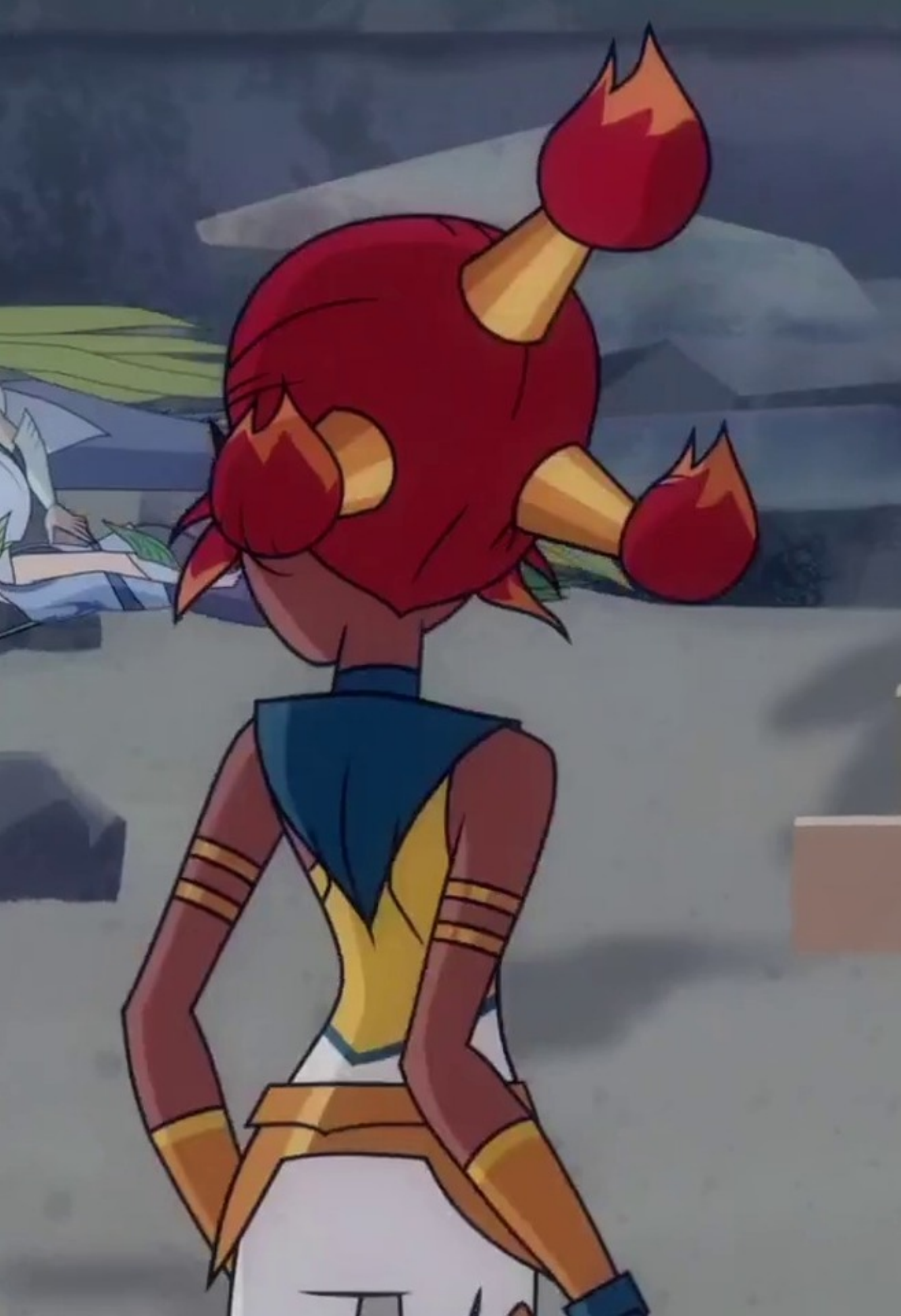}{Image 2}%
    }{\appinputimg{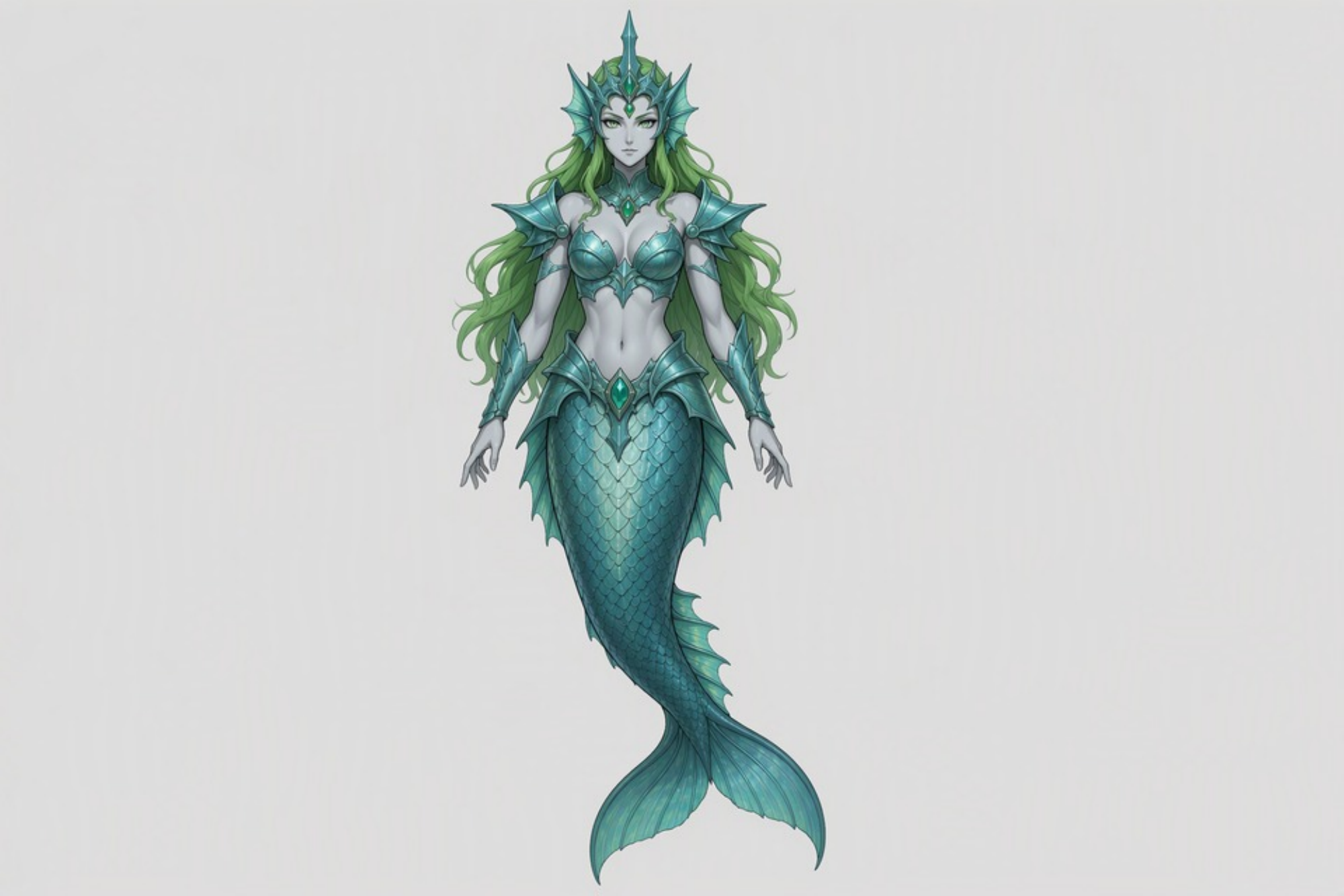}{Constructed by NeurCo (T+O row only)}%
    }{%
        \begin{tabularx}{\linewidth}{@{}XXX@{}}
            (1) Hybrid anatomy & (2) Queen appearance & (3) Anime style \\
            \textcolor{appBlue}{\textbf{(4) Four heroes' attributes}} & (5) Fighting & (6) Five-step order
        \end{tabularx}}
    \begin{tabular}{@{}c@{\hspace{\appGap}}c@{\hspace{\appGap}}c@{\hspace{\appGap}}c@{}}
        \parbox[c]{\appLabelW}{\centering\textbf{Method}} & \textbf{Early} & \textbf{Middle} & \textbf{Late} \\[2pt]
        \appcaselabel{Native}{\textbf{4/6 passed}}{\textcolor{appFail}{Fails (4), (6)}}
        & \appcaseframe{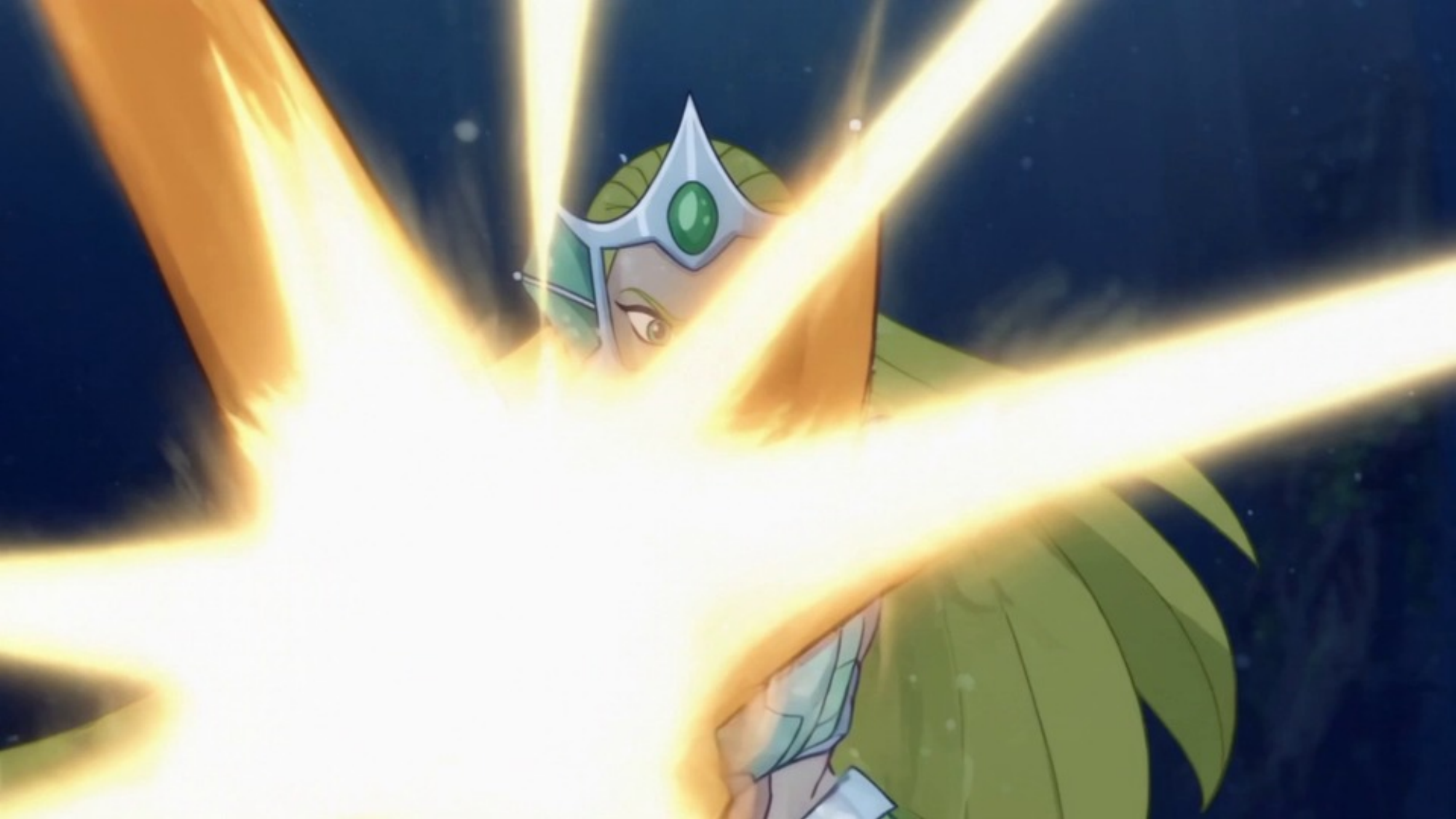}
        & \appcaseframe{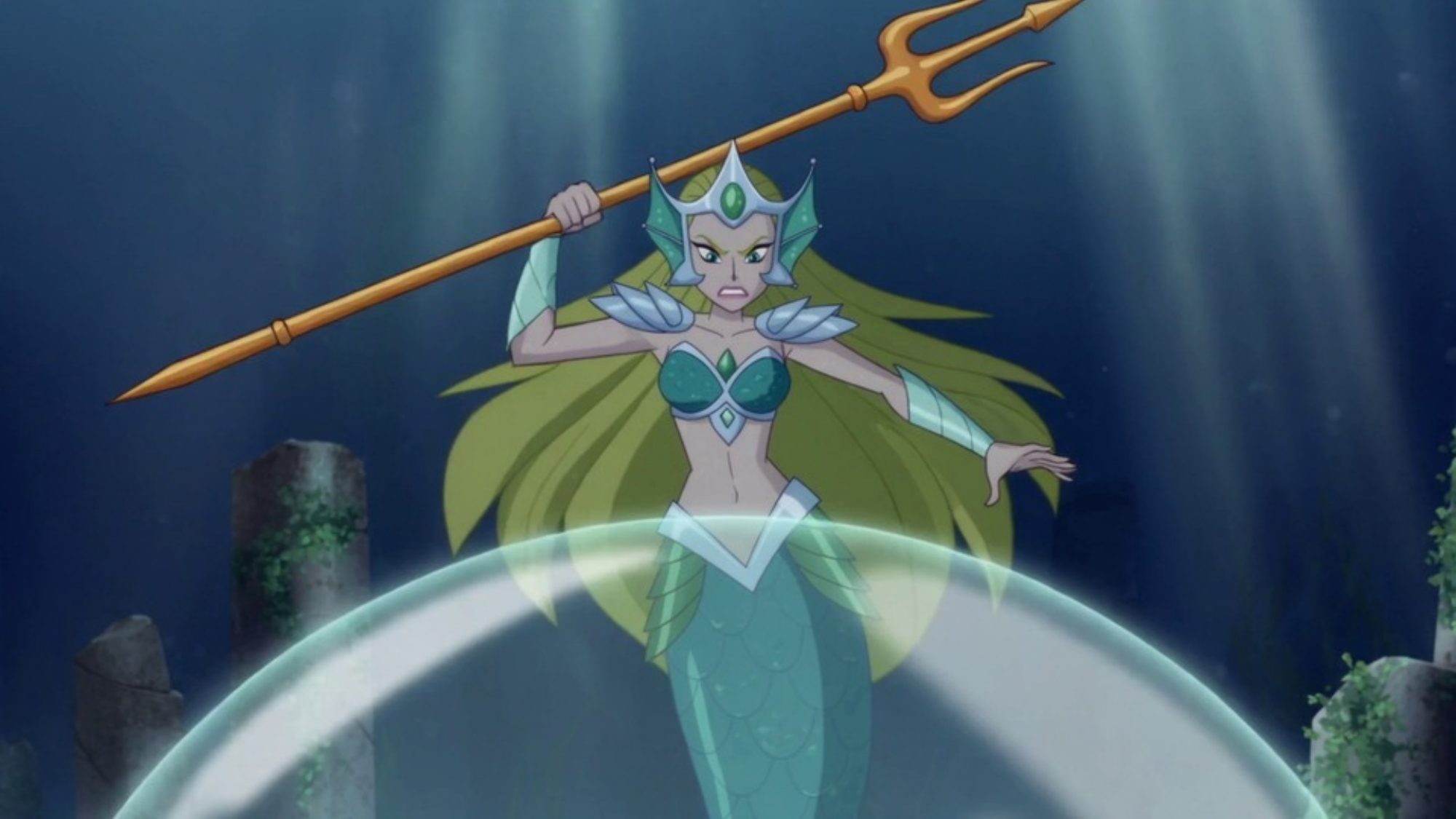}
        & \appcaseframe{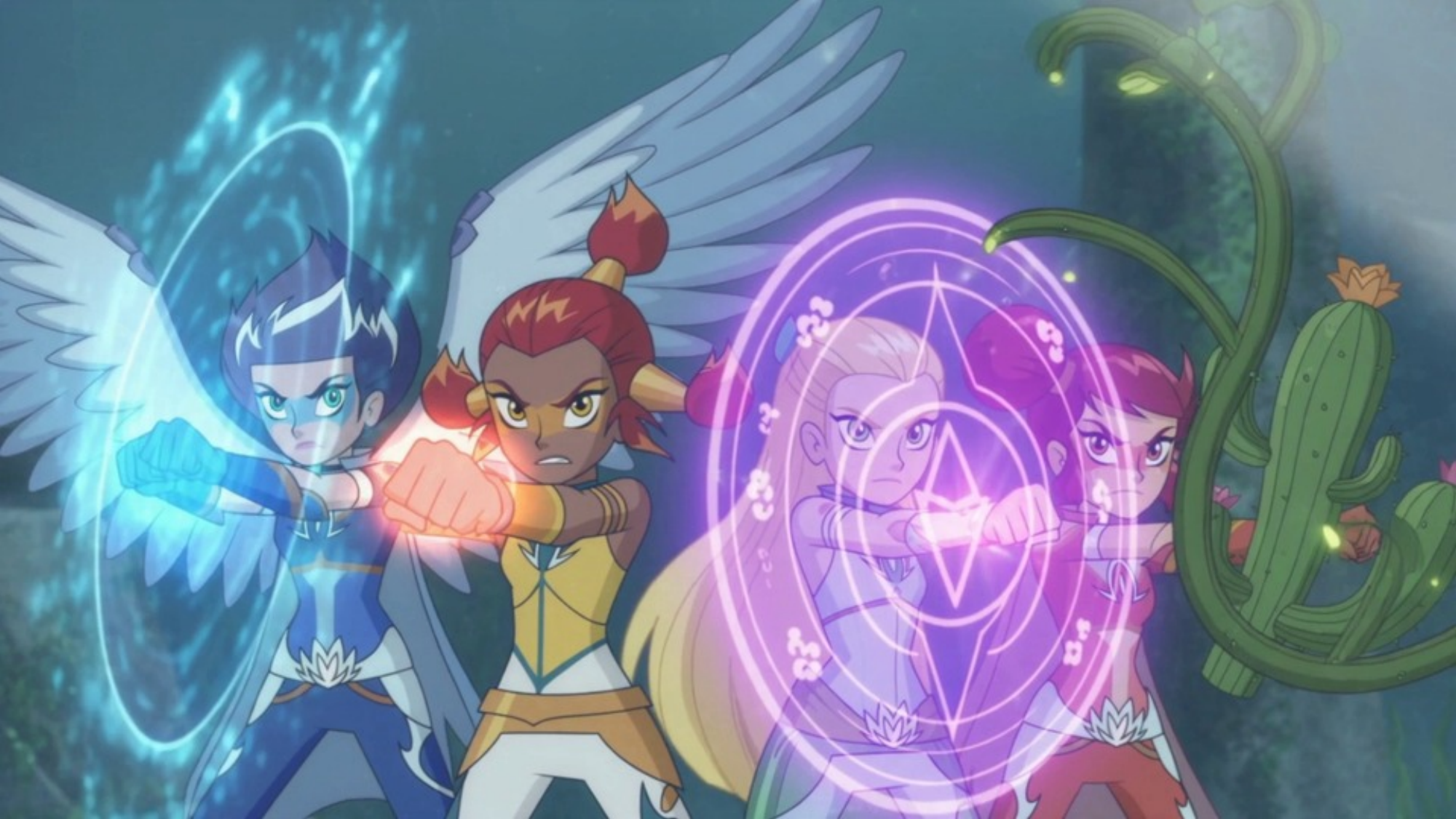}
        \end{tabular}
\par\vspace{3pt}\noindent\begin{tabular}{@{}c@{\hspace{\appGap}}c@{\hspace{\appGap}}c@{\hspace{\appGap}}c@{}}

        \appcaselabel{NeurCo (T)}{\textbf{4/6 passed}}{\textcolor{appFail}{Fails (4), (6)}}
        & \appcaseframe{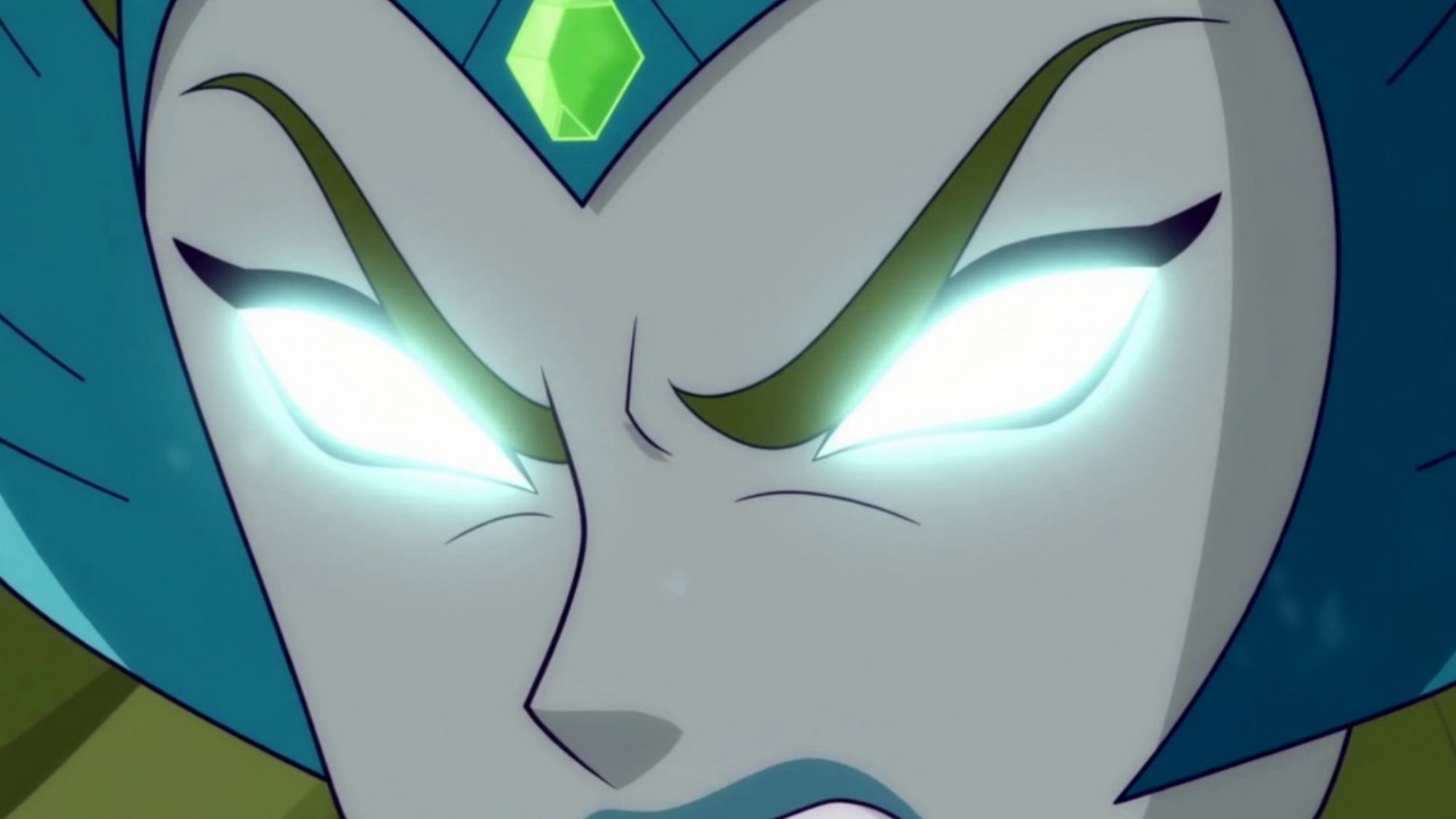}
        & \appcaseframe{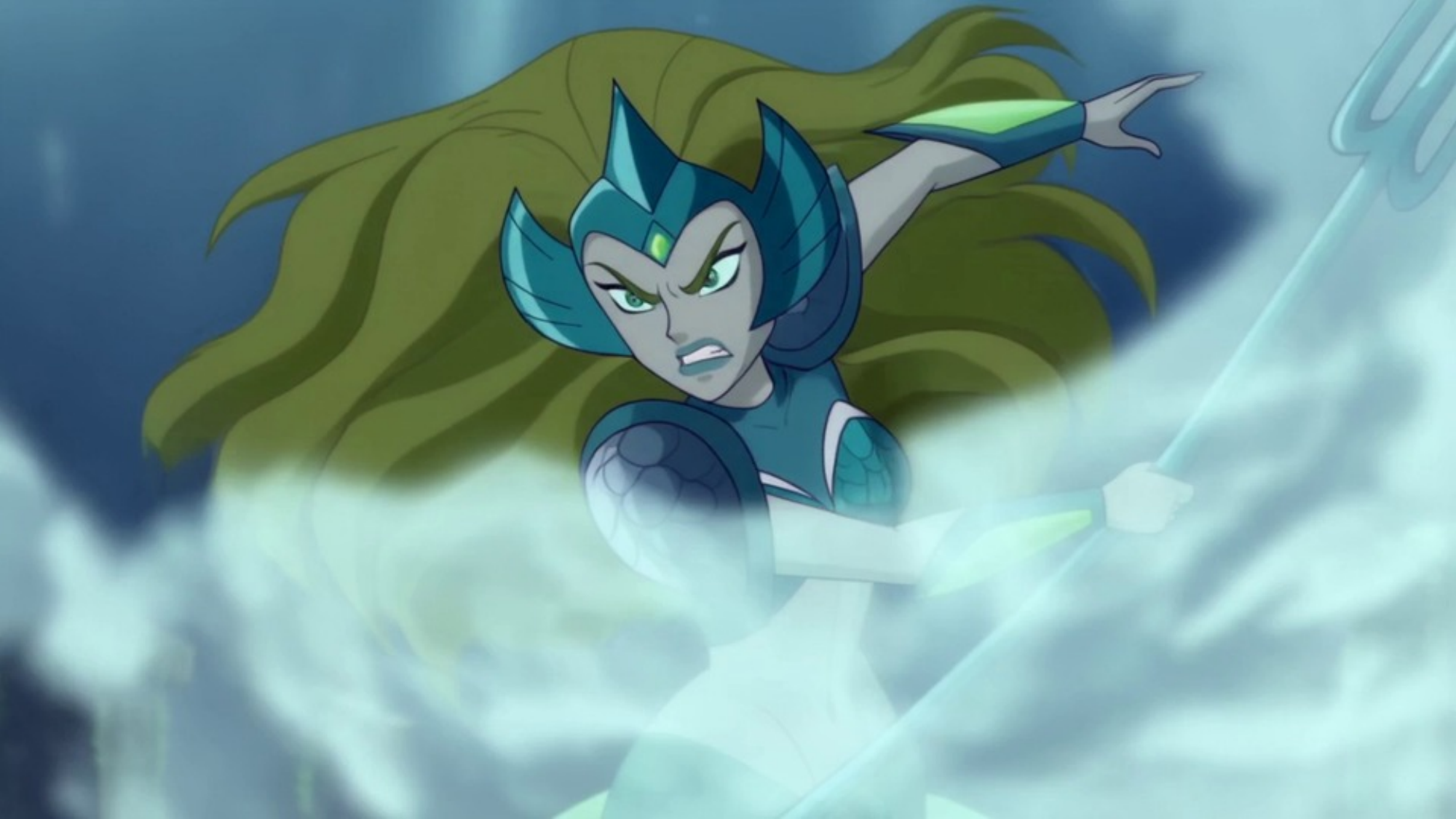}
        & \appcaseframe{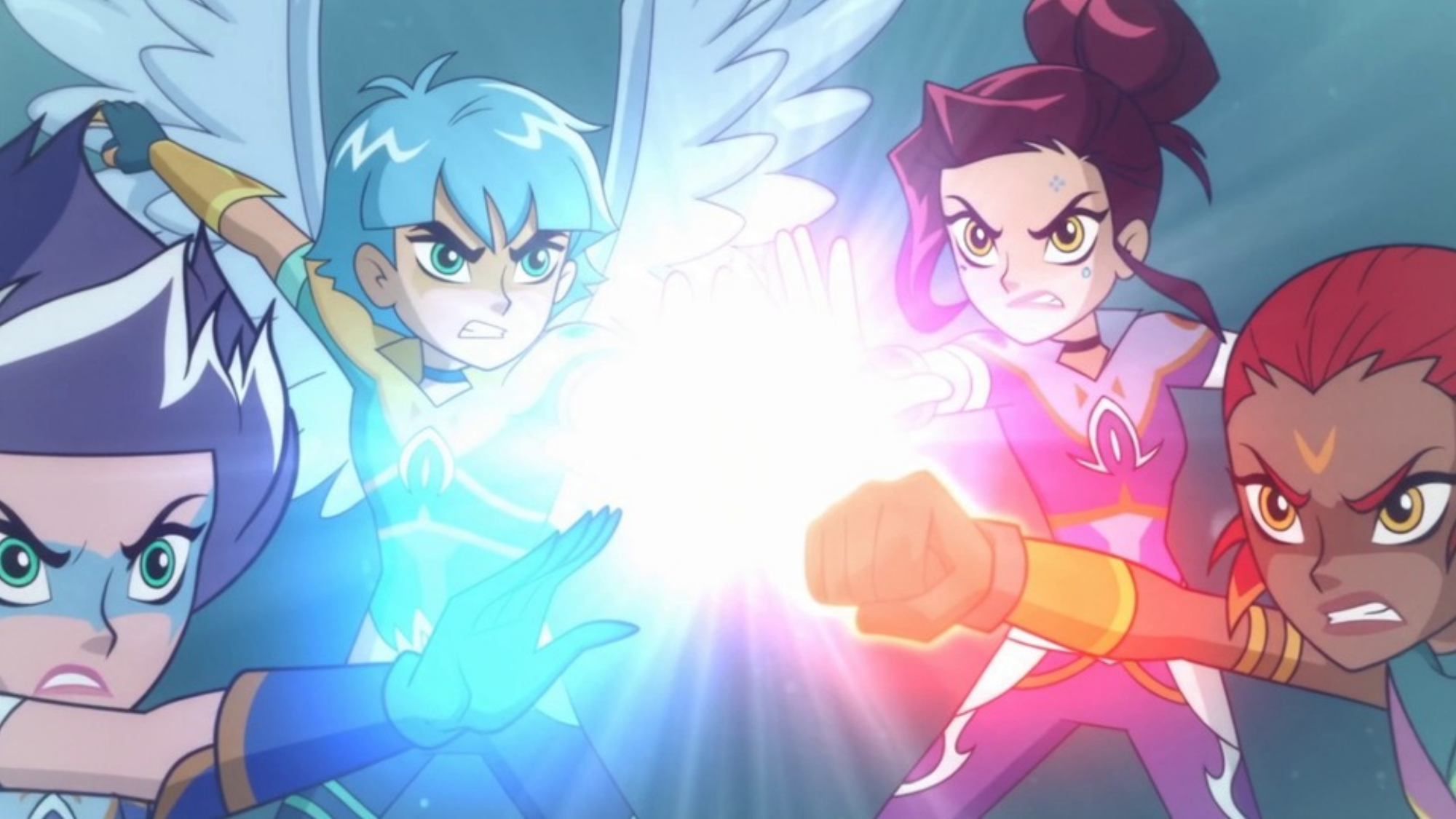}
        \end{tabular}
\par\vspace{3pt}\noindent\begin{tabular}{@{}c@{\hspace{\appGap}}c@{\hspace{\appGap}}c@{\hspace{\appGap}}c@{}}

        \appcaselabel{NeurCo (T+O)}{\textcolor{appBlue}{\textbf{5/6 passed}}}{Added reference only}
        & \appcaseframe{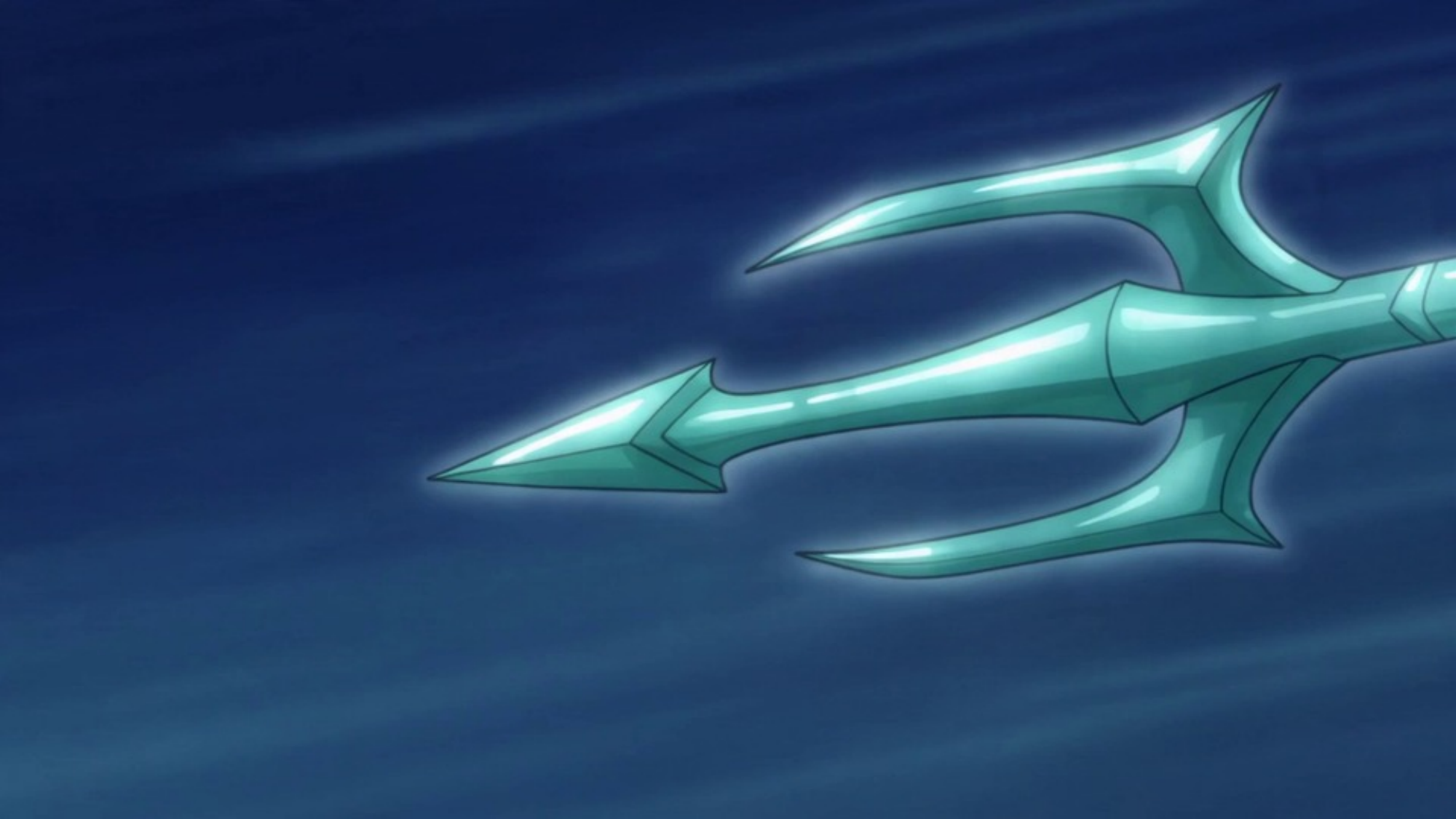}
        & \appcaseframe{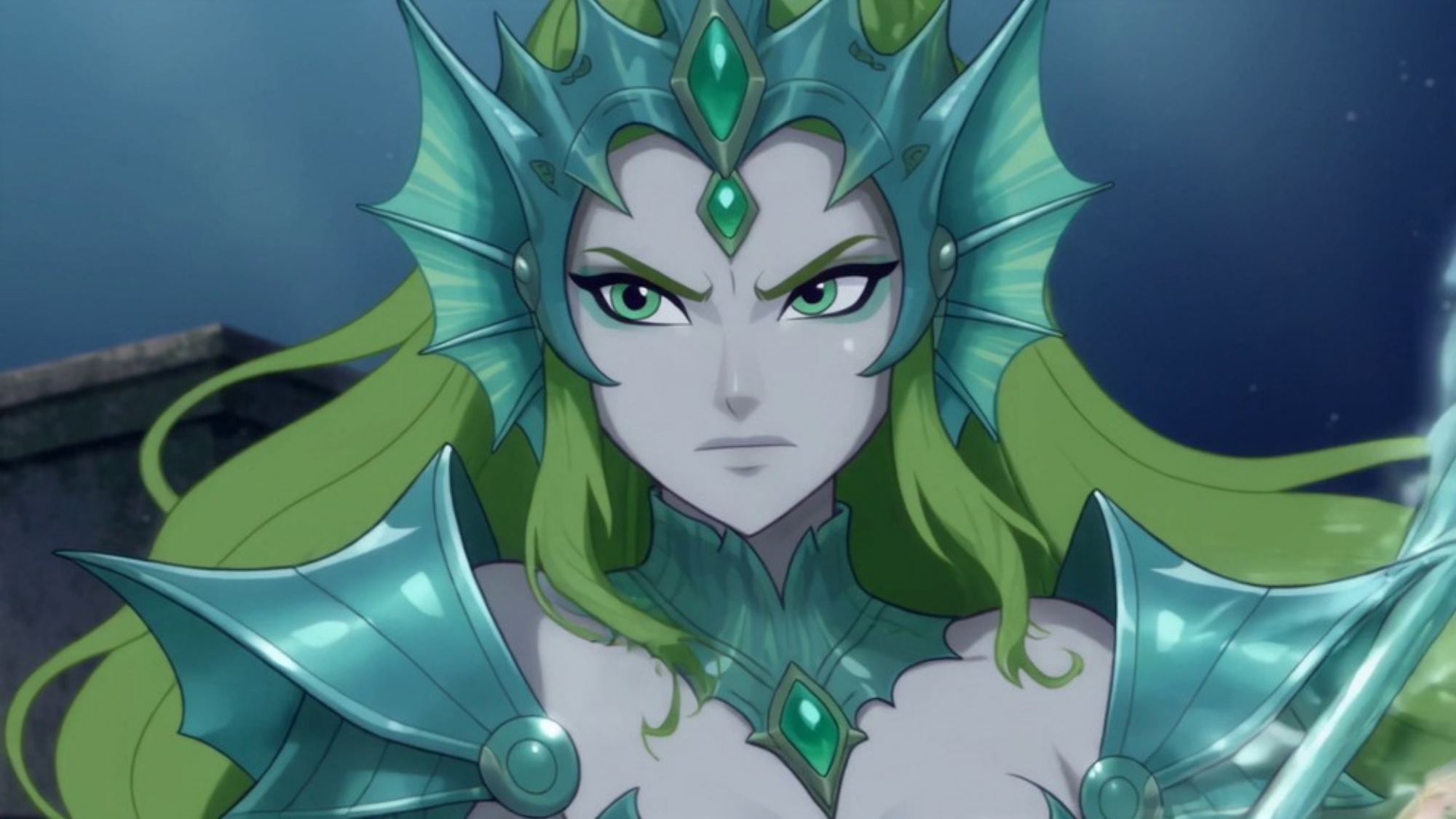}
        & \appcaseframe{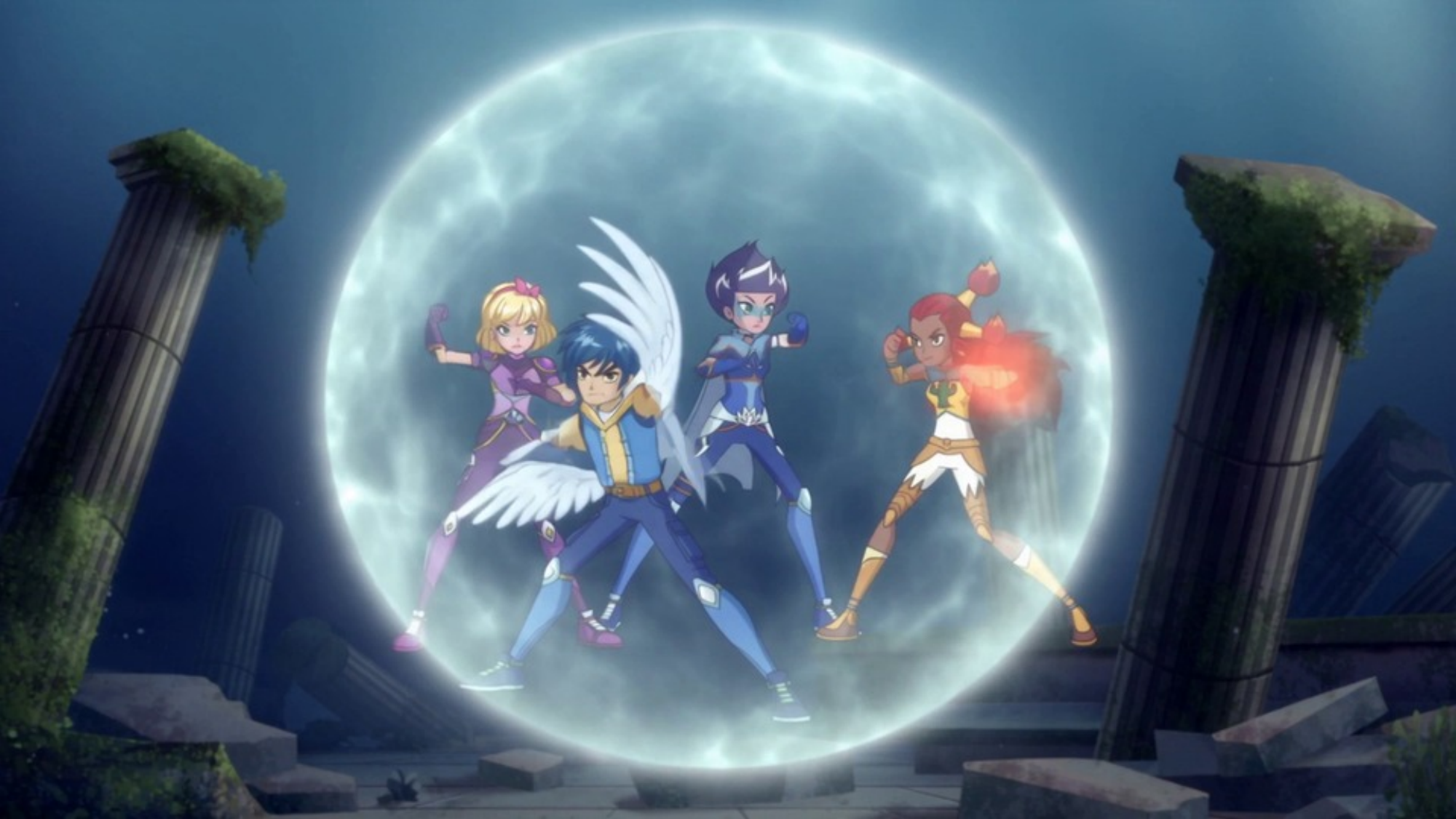}
    \end{tabular}
    \nopagebreak[3]\caption{\textbf{An added reference improves character binding.}
    On Seedance~2.5, NeurCo (T+O) raises satisfaction to $5/6$ without
    targeted rewriting; the five-step event order remains incorrect.}
    \label{fig:case_sd_r2v}
    \endgroup
\par\endgroup
\end{appcaseblock}

\clearpage
\begin{appcaseblock}
\noindent\textbf{Appearance conditioning and event order.}
Reference-conditioned generation also recovers the scored event
sequence in the Seedance~2.5 cat example
(Figure~\ref{fig:case_sd_cat}). TIDE and its text-only extension
satisfy $2/3$ requirements, whereas the O and T+O plans reach $3/3$.
The reference-conditioned clips include the intermediate pause before
the cat moves beneath the table.

\par\smallskip\noindent\begingroup\centering
    \centering
    \begingroup
    \appcaseinit
    \appcaseinputas{T2V}{A cat starts on top of a table. It hops down and runs to the front of the table, pausing
    there. Then it walks beneath the table and settles underneath it.}{}{\appinputimg{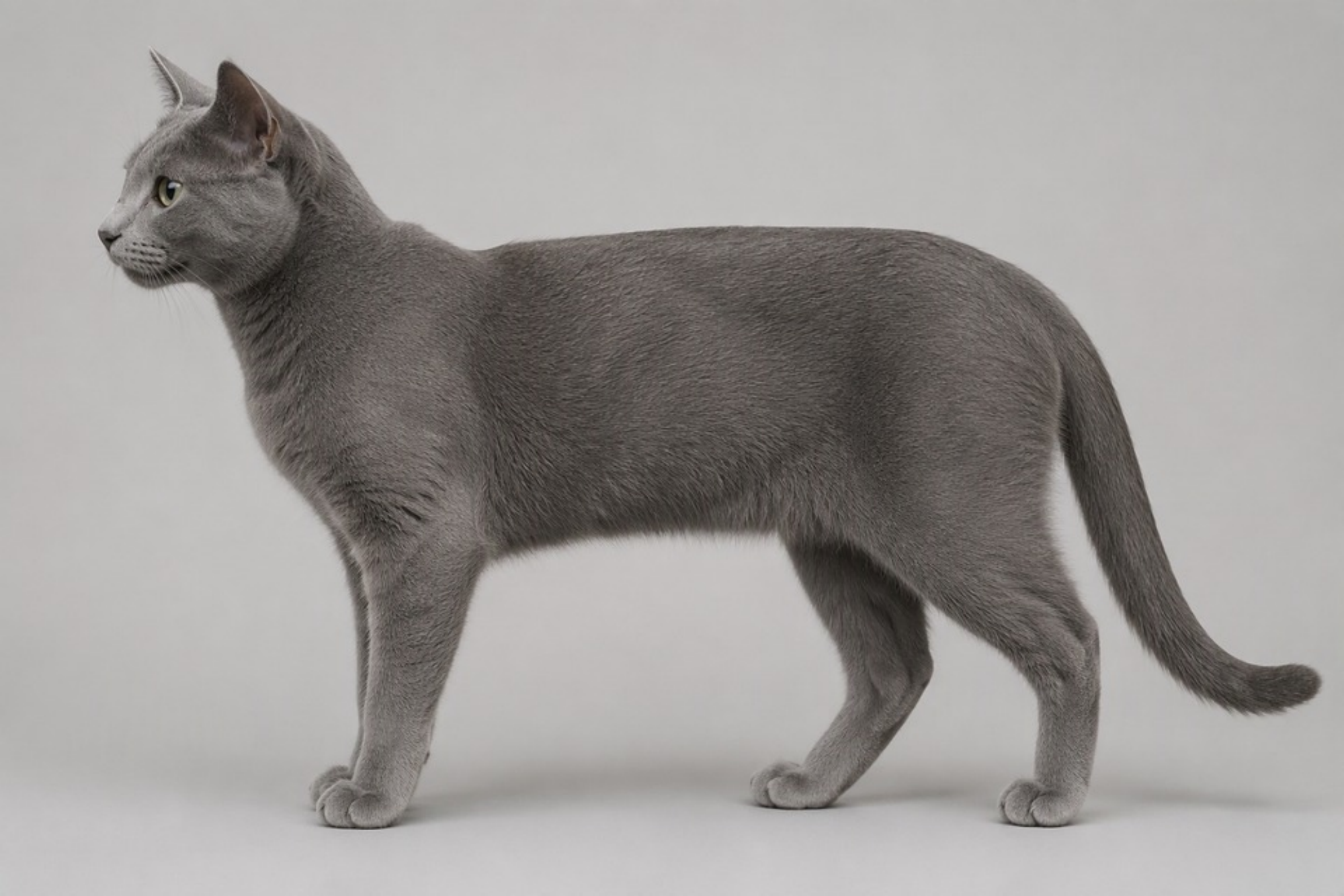}{Appearance reference (O)}%
    }{%
        \begin{tabularx}{\linewidth}{@{}XXX@{}}
            \textcolor{appBlue}{\textbf{(1) Table, pause in front, then underneath}}
            & (2) The subject is a cat
            & (3) Ordinary locomotion and settling
        \end{tabularx}}
    \begin{tabular}{@{}c@{\hspace{\appGap}}c@{\hspace{\appGap}}c@{\hspace{\appGap}}c@{}}
        \parbox[c]{\appLabelW}{\centering\textbf{Method}} & \textbf{Early} & \textbf{Middle} & \textbf{Late} \\[2pt]
        \appcaselabel{TIDE}{\textbf{2/3 passed}}{\textcolor{appFail}{(1) unclear}}
        & \appcaseframe{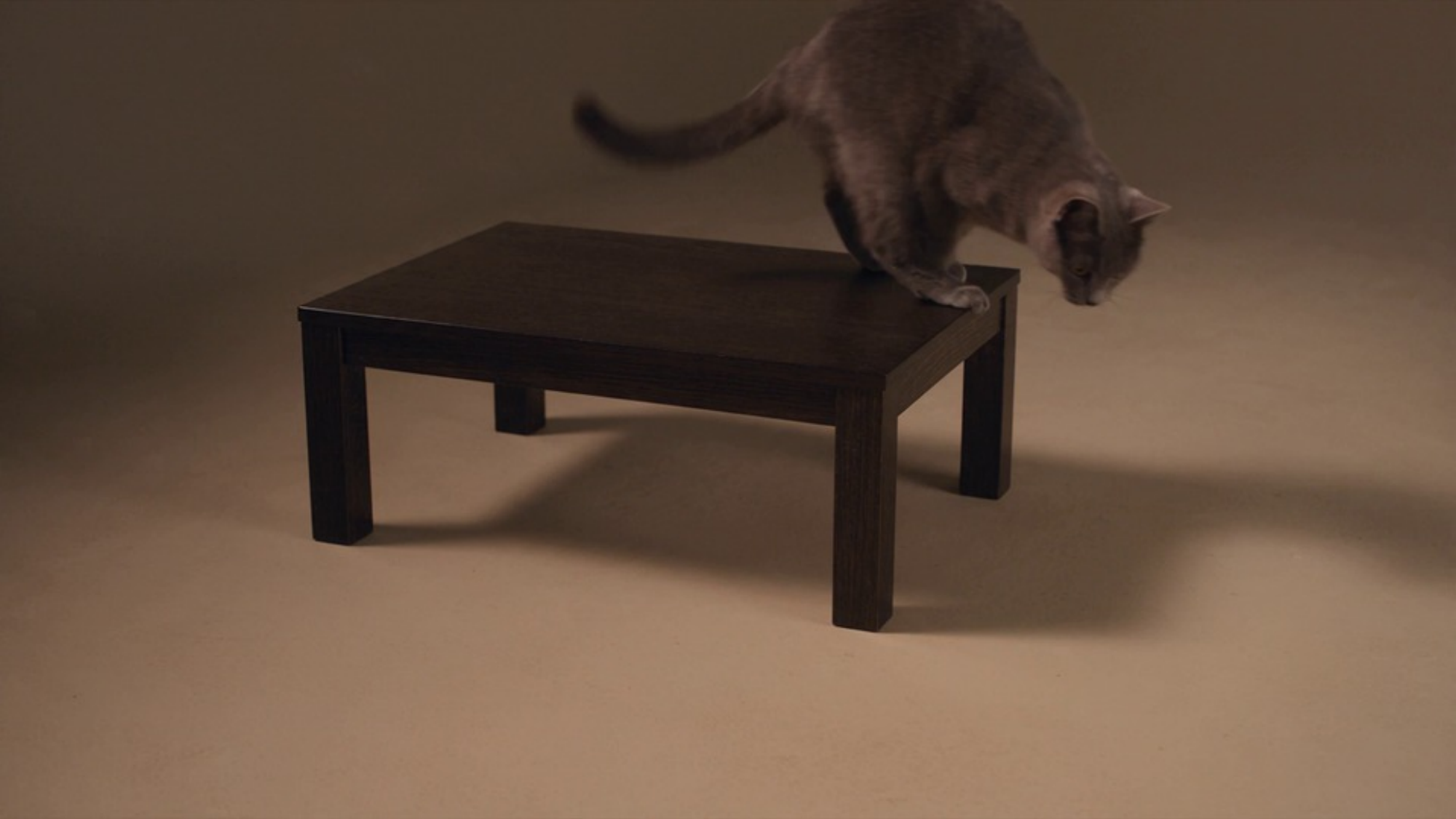}
        & \appcaseframe{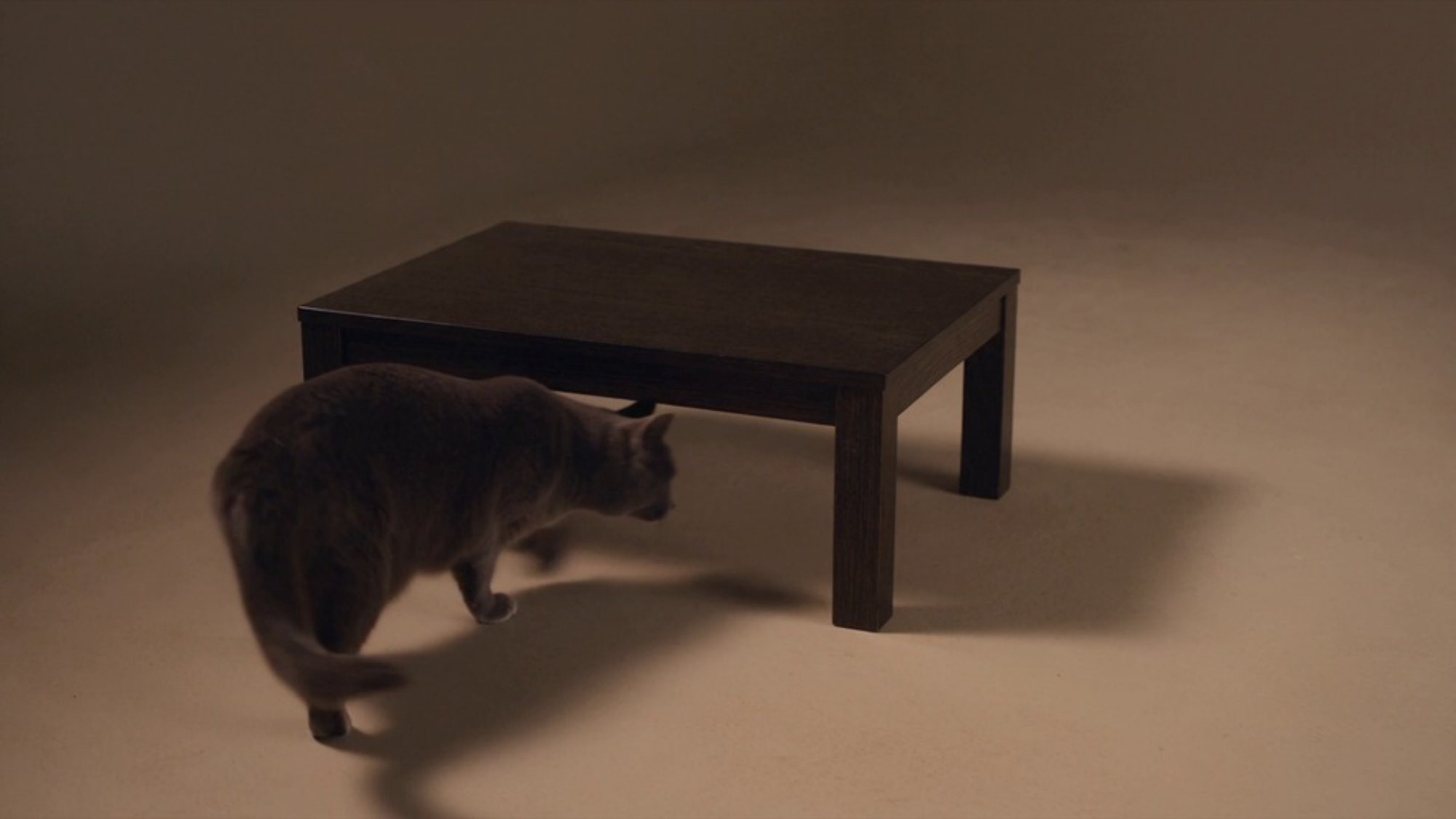}
        & \appcaseframe{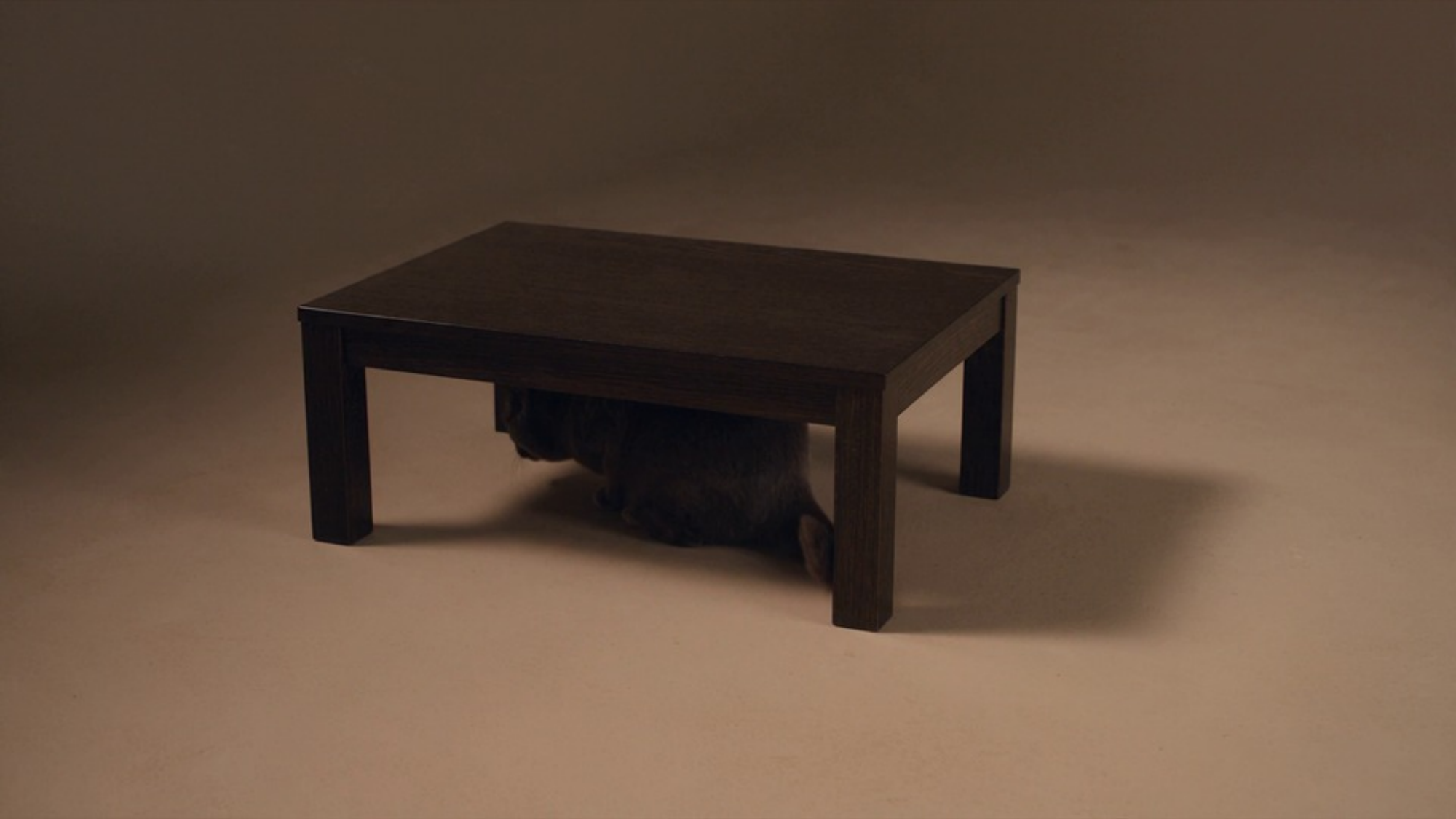}
        \end{tabular}
\par\vspace{3pt}\noindent\begin{tabular}{@{}c@{\hspace{\appGap}}c@{\hspace{\appGap}}c@{\hspace{\appGap}}c@{}}

        \appcaselabel{TIDE + NeurCo (T)}{\textbf{2/3 passed}}{\textcolor{appFail}{Fails (1)}}
        & \appcaseframe{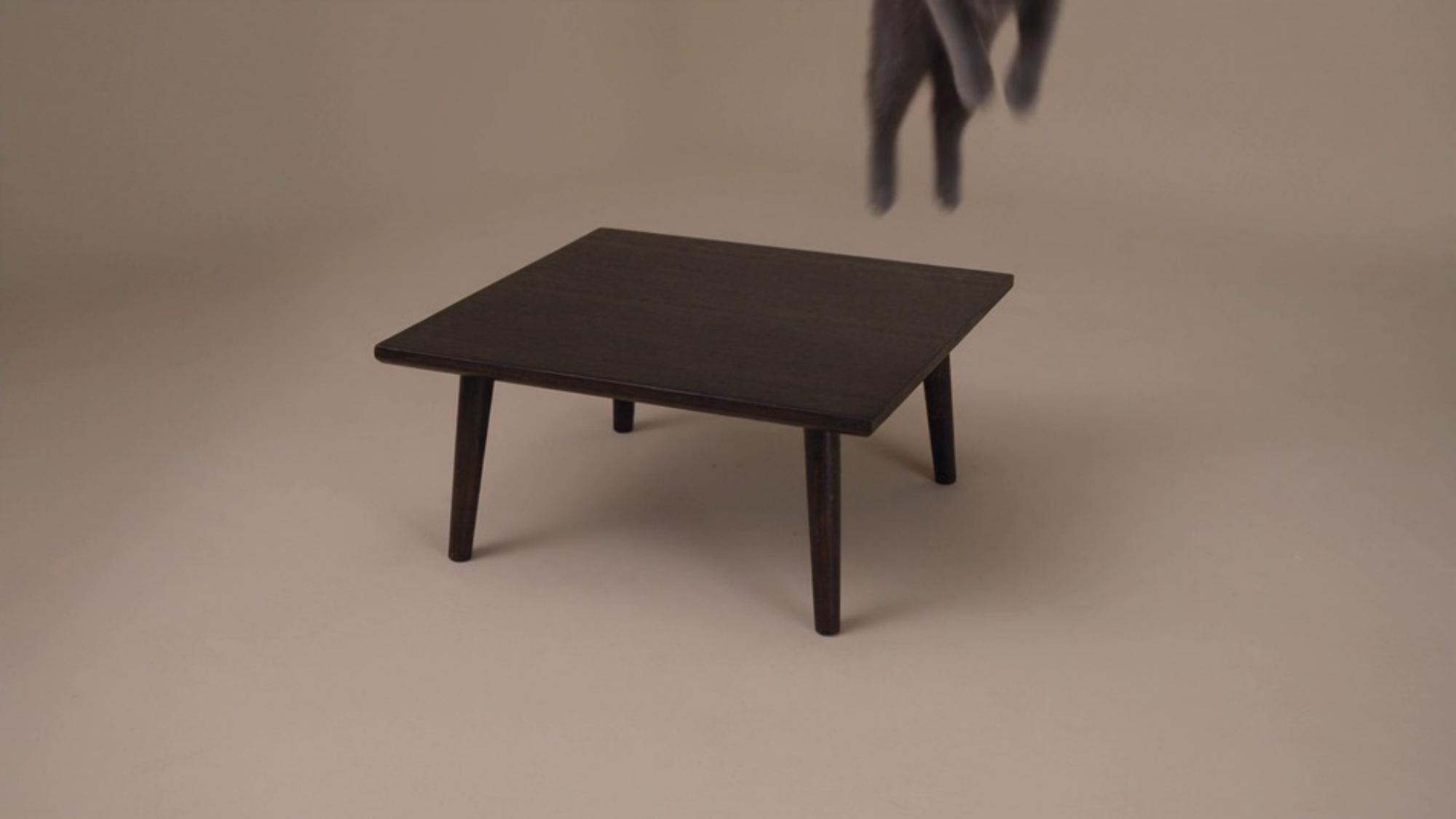}
        & \appcaseframe{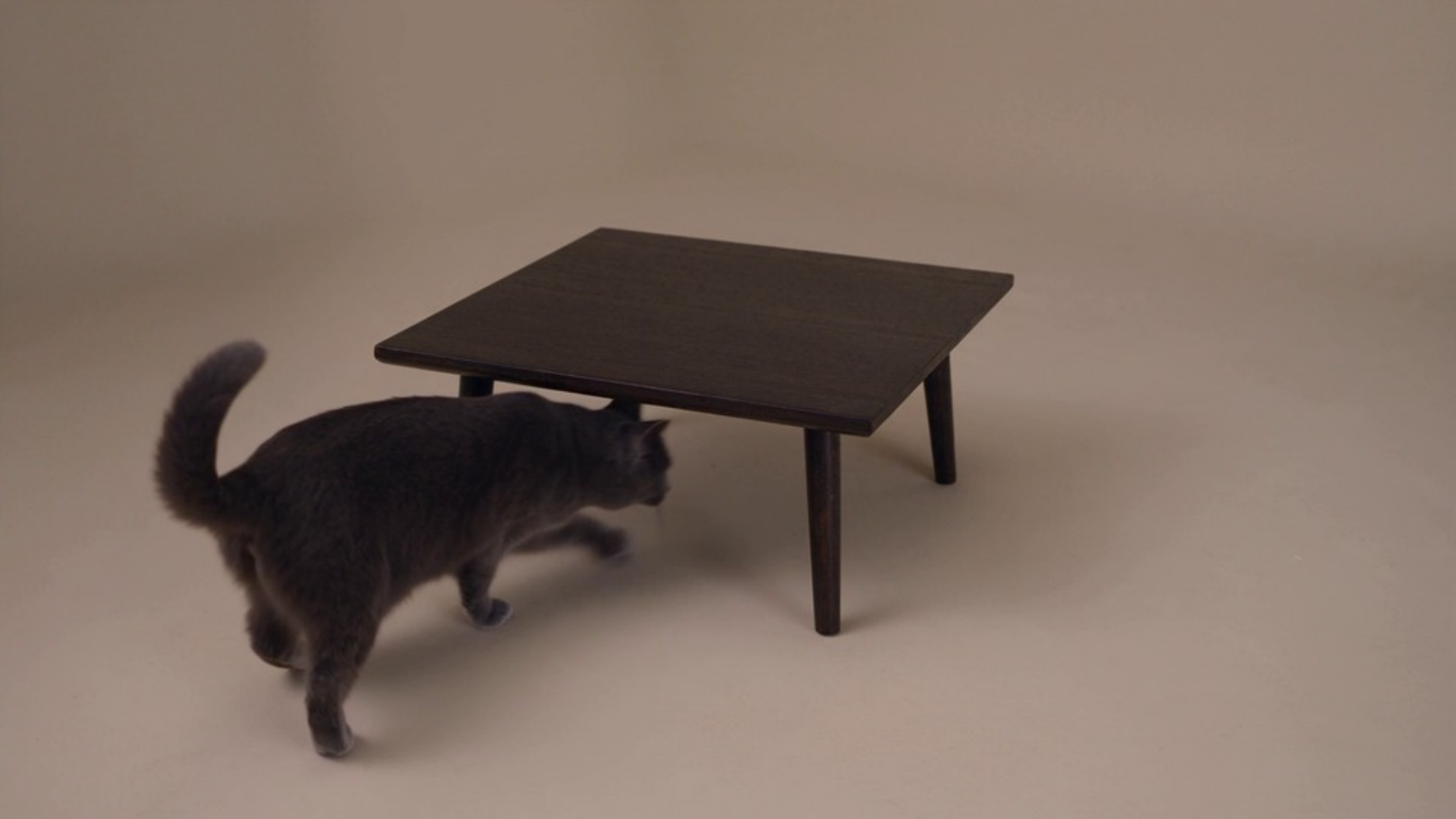}
        & \appcaseframe{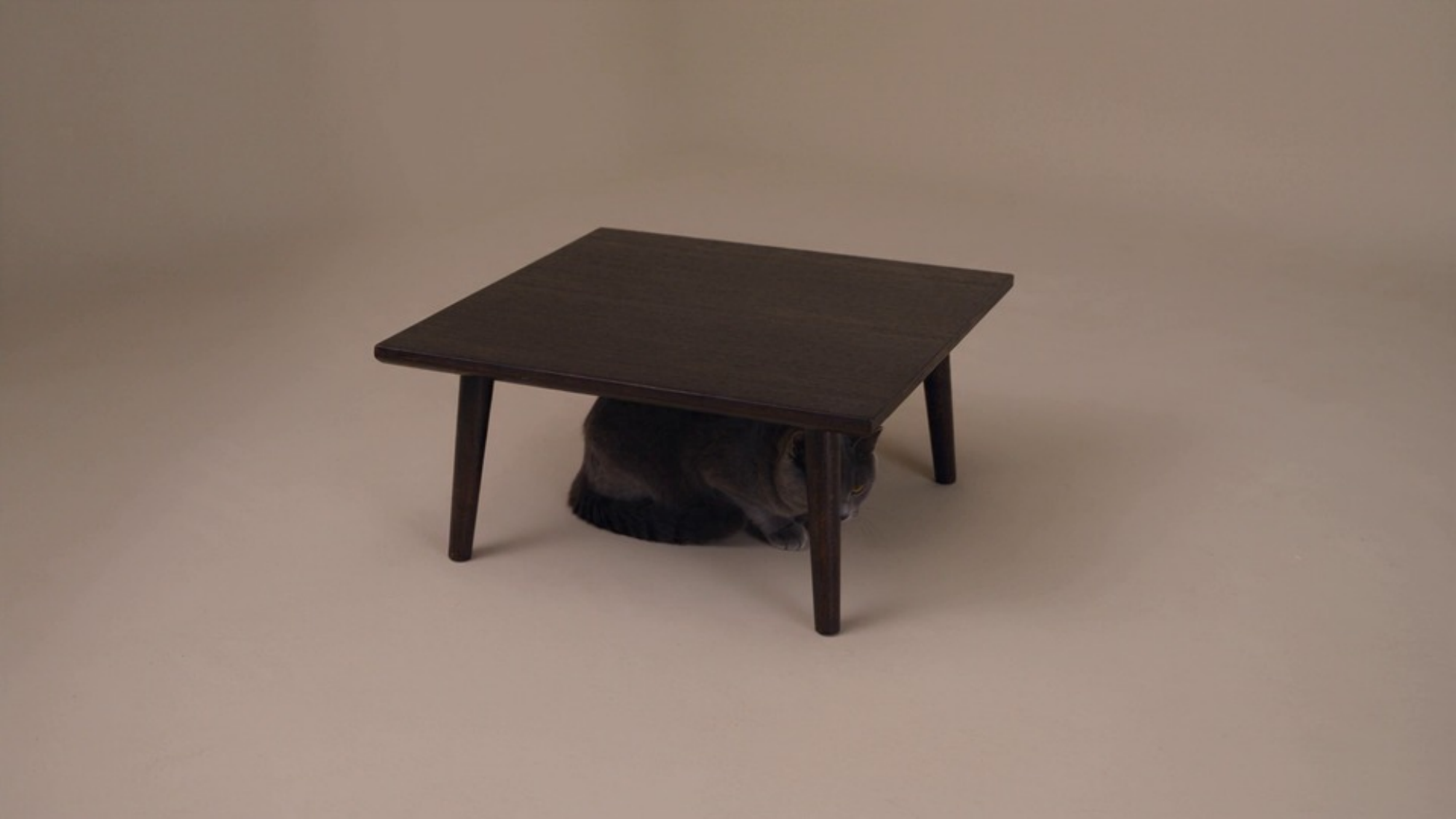}
        \end{tabular}
\par\vspace{3pt}\noindent\begin{tabular}{@{}c@{\hspace{\appGap}}c@{\hspace{\appGap}}c@{\hspace{\appGap}}c@{}}

        \appcaselabel{TIDE + NeurCo (O)}{\textcolor{appBlue}{\textbf{3/3 passed}}}{Appearance reference}
        & \appcaseframe{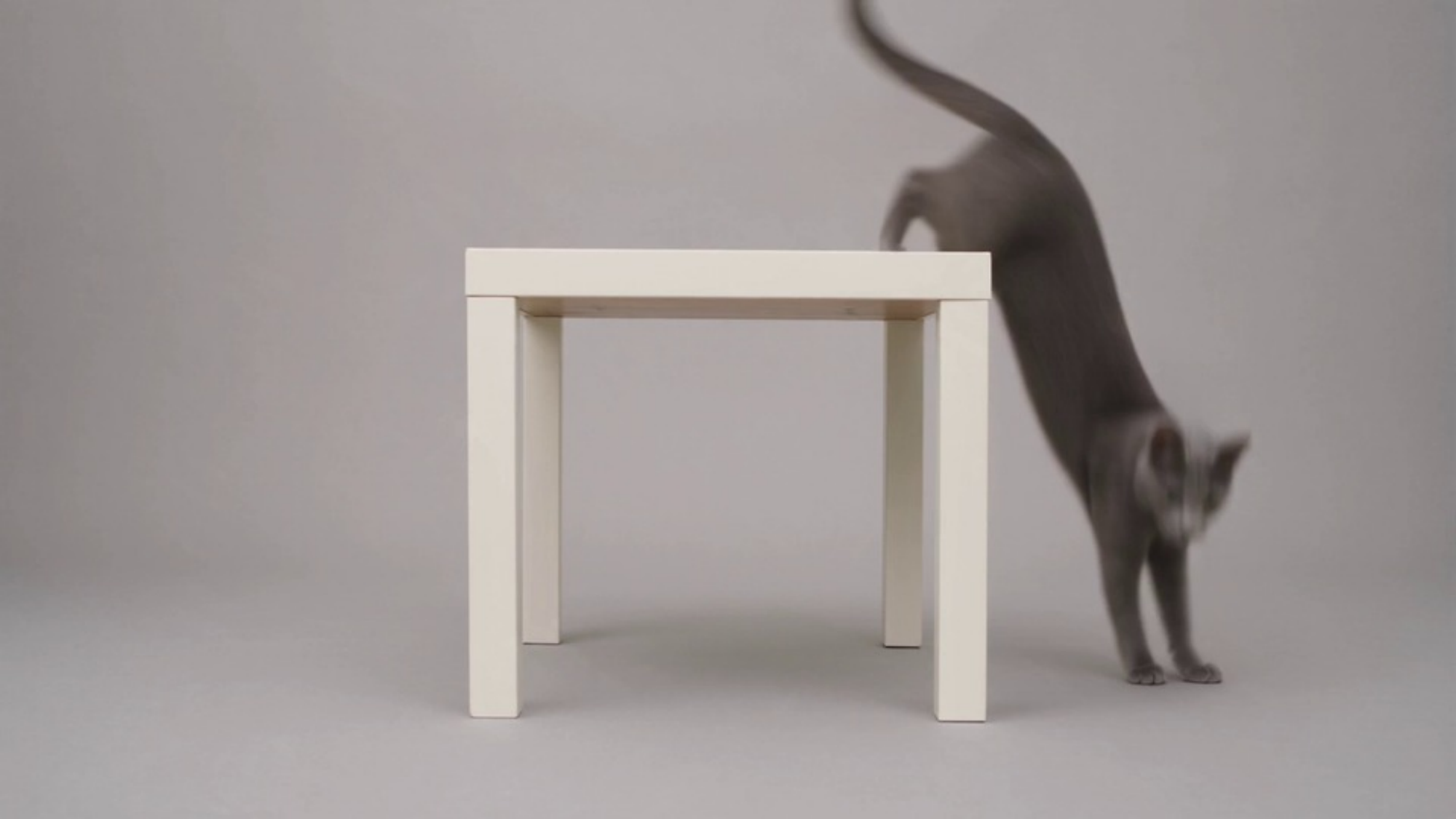}
        & \appcaseframe{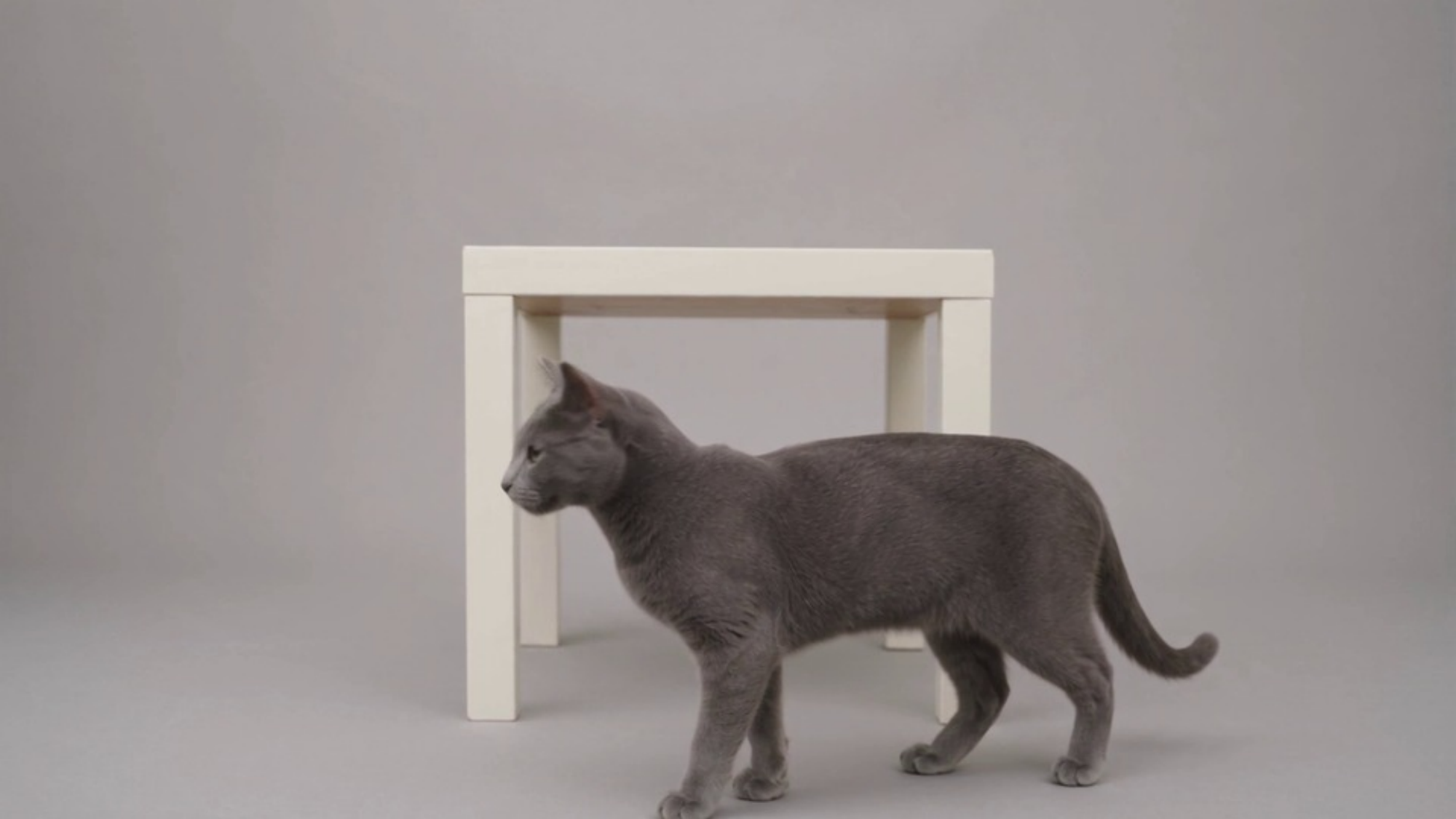}
        & \appcaseframe{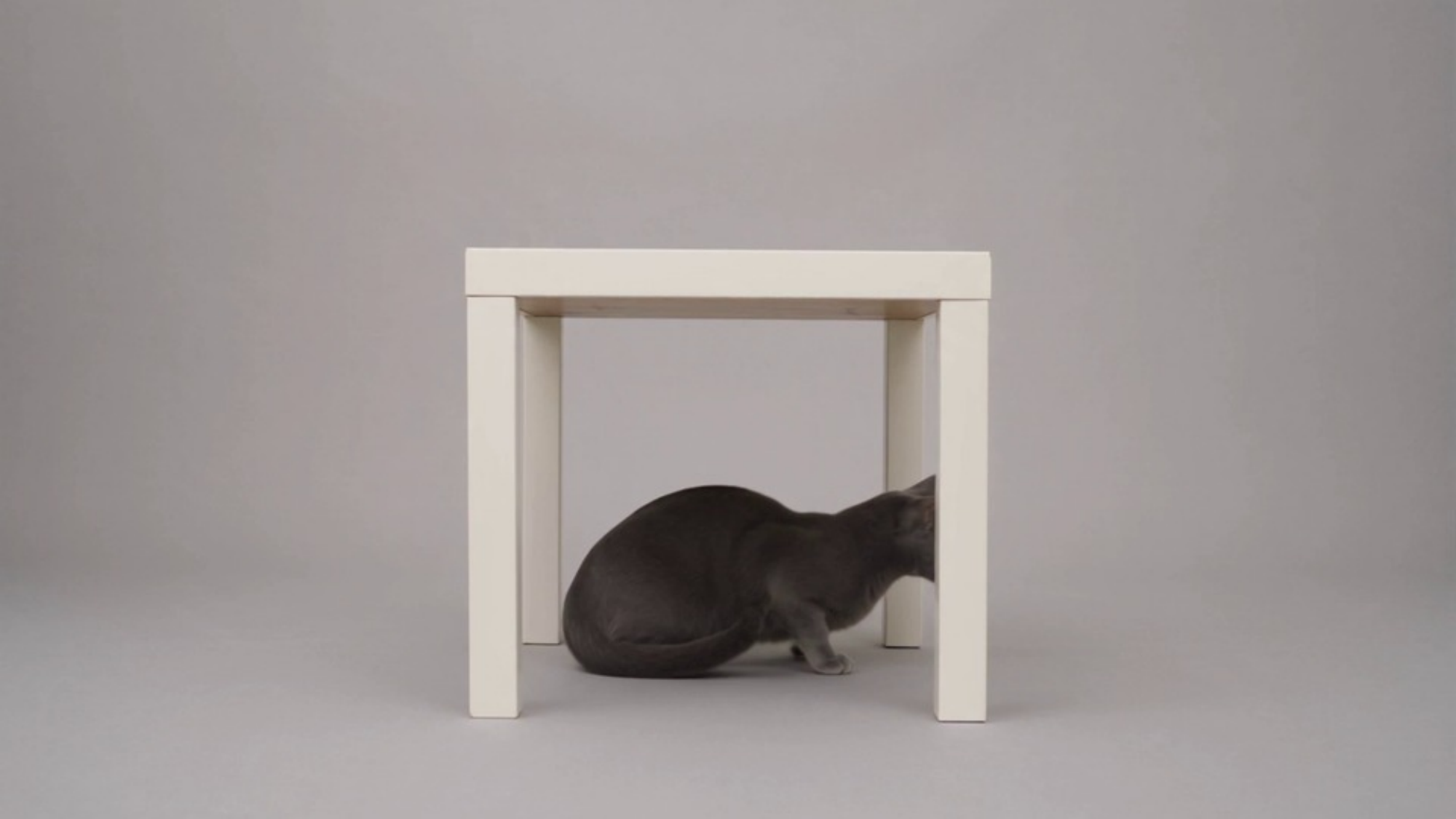}
        \end{tabular}
\par\vspace{3pt}\noindent\begin{tabular}{@{}c@{\hspace{\appGap}}c@{\hspace{\appGap}}c@{\hspace{\appGap}}c@{}}

        \appcaselabel{TIDE + NeurCo (T+O)}{\textcolor{appBlue}{\textbf{3/3 passed}}}{Reference + text}
        & \appcaseframe{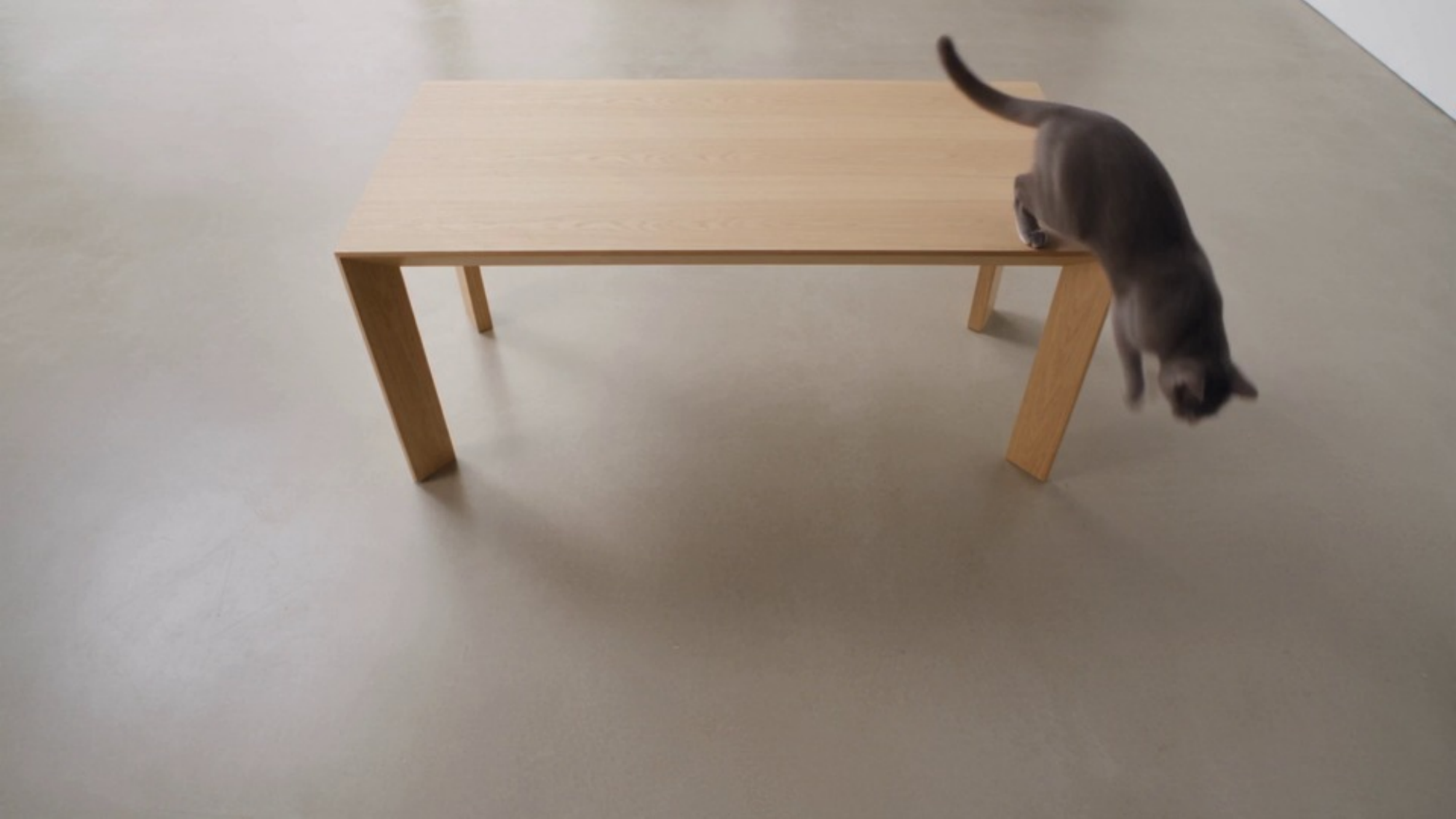}
        & \appcaseframe{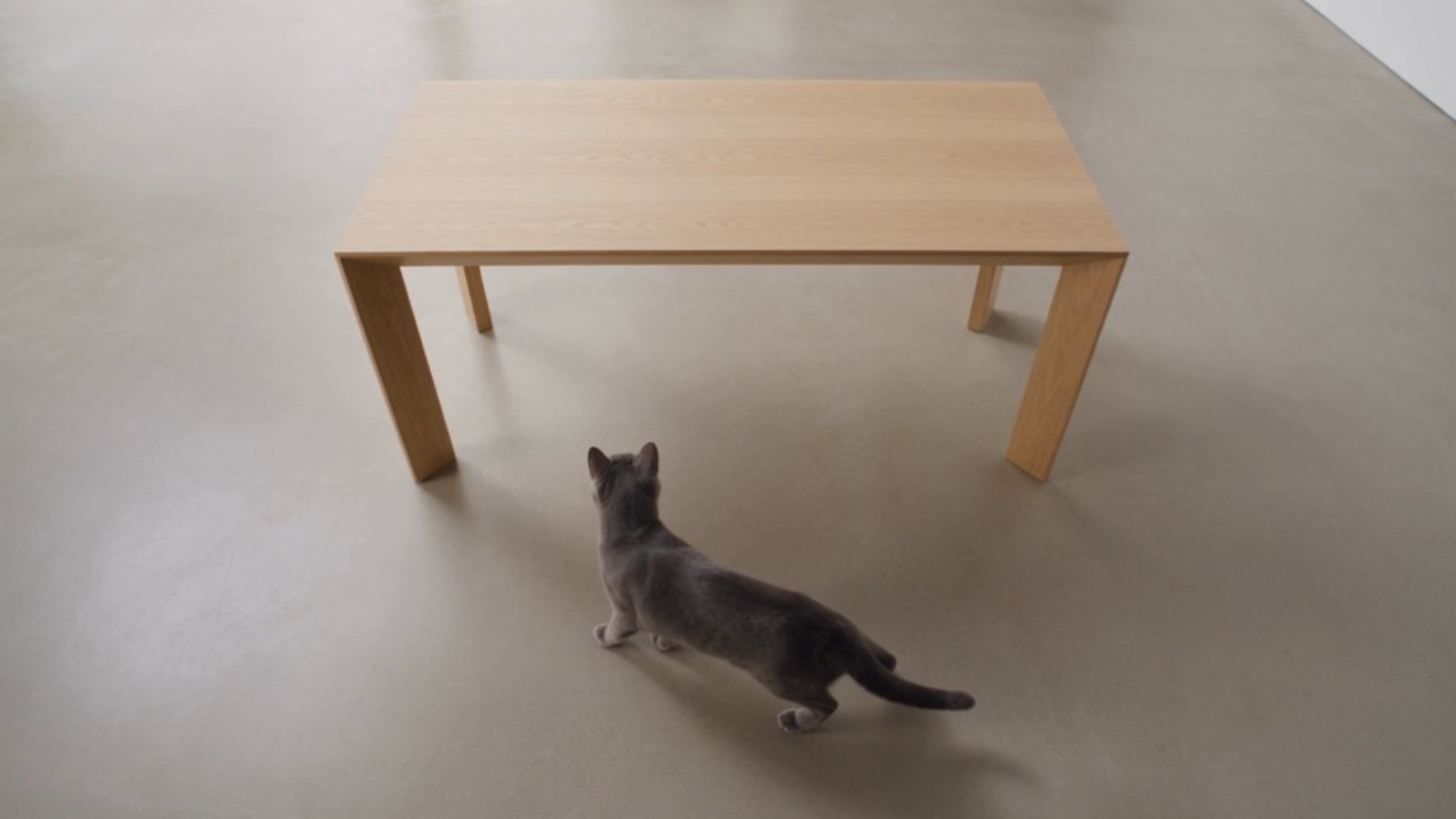}
        & \appcaseframe{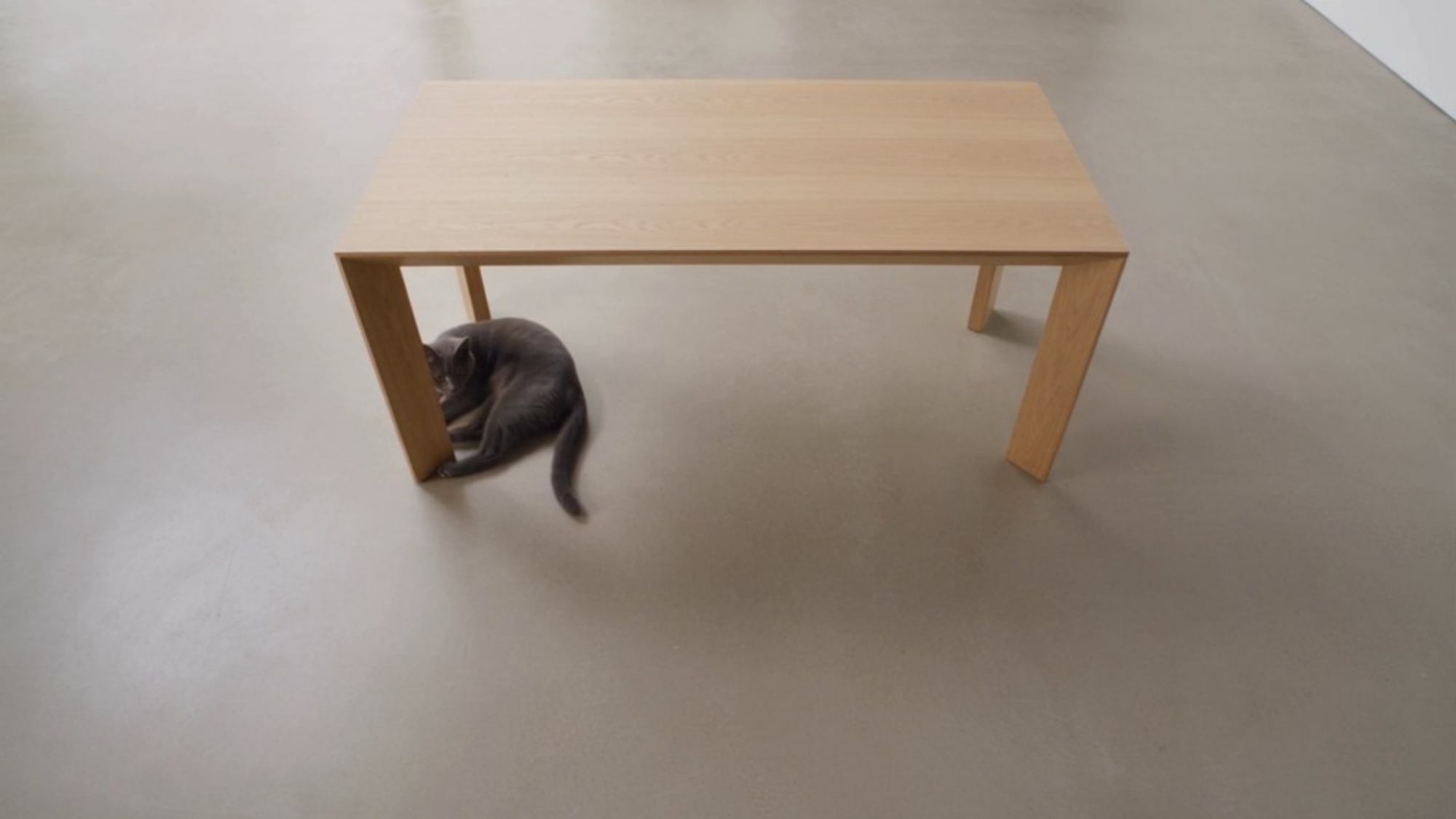}
    \end{tabular}
    \nopagebreak[3]\caption{\textbf{Appearance conditioning accompanies better event order.}
    Both reference-conditioned variants satisfy $3/3$ requirements on
    Seedance~2.5; the TIDE and text-only variants satisfy $2/3$.}
    \label{fig:case_sd_cat}
    \endgroup
\par\endgroup
\end{appcaseblock}

\clearpage
\begin{appcaseblock}
\noindent\textbf{Four glove colors from an appearance reference.}
Figure~\ref{fig:case_sd_gloves} shows how reference conditioning restores attribute binding in a TIDE-enhanced request. Native satisfies the four-glove-color requirement but misses the camera and speed requirements ($3/6$); TIDE and TIDE + NeurCo (T) miss the glove colors as well ($2/6$). A constructed reference fixes the four workers' glove colors in R2V, while the combined plan also satisfies the steady pan and rapid bar movement ($5/6$).

\par\smallskip\noindent\begingroup\centering
    \centering
    \begingroup
    \appcaseinit
    \appcaseinputas{T2V}{Four workers stand shoulder to shoulder at a workbench, one wearing red gloves, one blue gloves, one green gloves, and one yellow gloves. In close-up, their gloved hands rapidly push and pull a sliding bar back and forth together. The camera stays in a fixed position and pans slowly and steadily left from start to finish, maintaining tight framing on their hands and the bar.}{}{\appinputimg{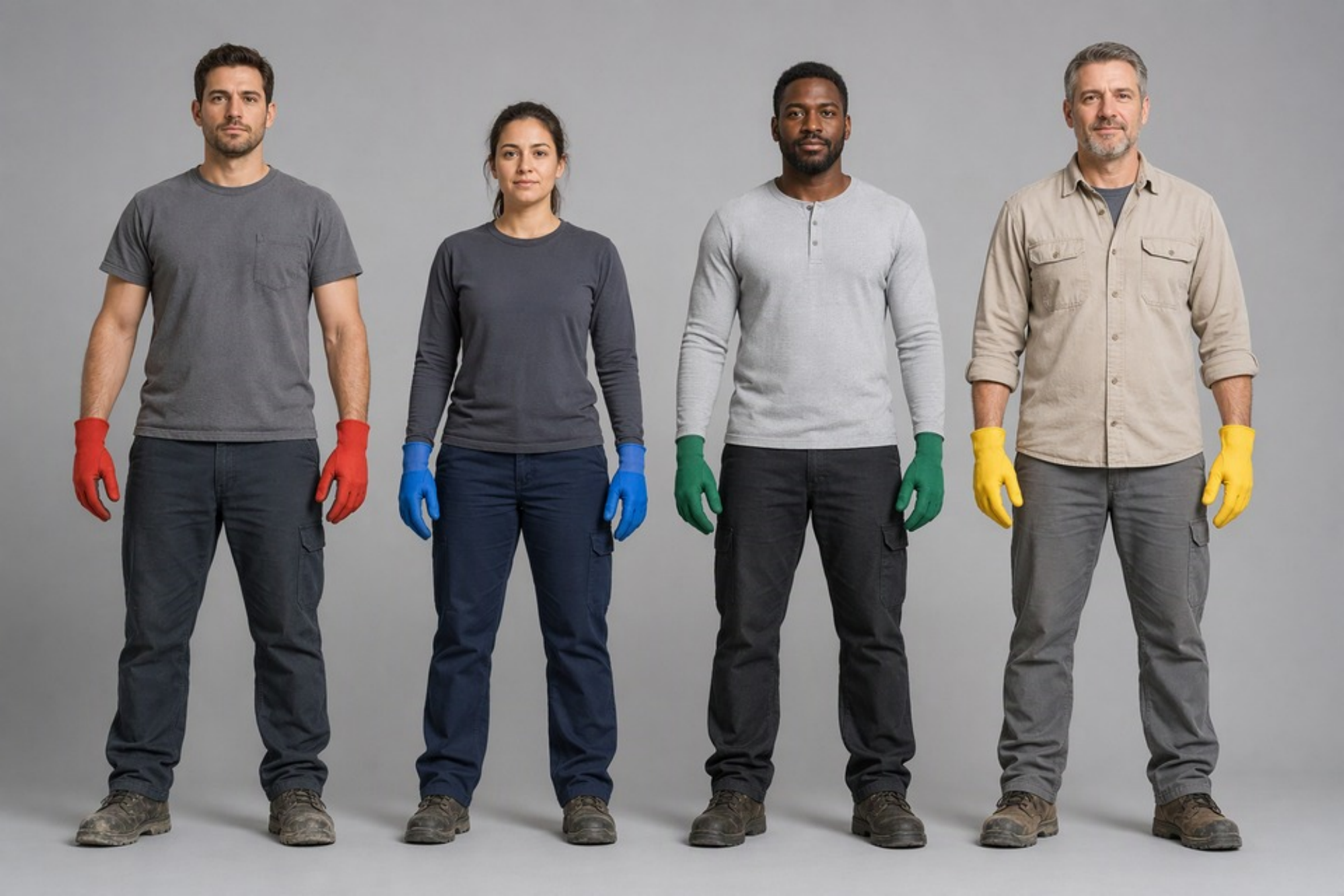}{Appearance reference (T+O row)}%
    }{%
        \begin{tabularx}{\linewidth}{@{}XXX@{}}
            (1) Steady slow pan & (2) Pan left, camera in place & \textcolor{appBlue}{\textbf{(3) Four workers, four glove colors}} \\
            (4) Push and pull the bar together & (5) Close-up on hands & (6) Fast back-and-forth
        \end{tabularx}}
    \begin{tabular}{@{}c@{\hspace{\appGap}}c@{\hspace{\appGap}}c@{\hspace{\appGap}}c@{}}
        \parbox[c]{\appLabelW}{\centering\textbf{Method}} & \textbf{Early} & \textbf{Middle} & \textbf{Late} \\[2pt]
        \appcaselabel{Native}{\textbf{3/6 passed}}{\textcolor{appFail}{Fails (1), (2), (6)}}
        & \appcaseframe{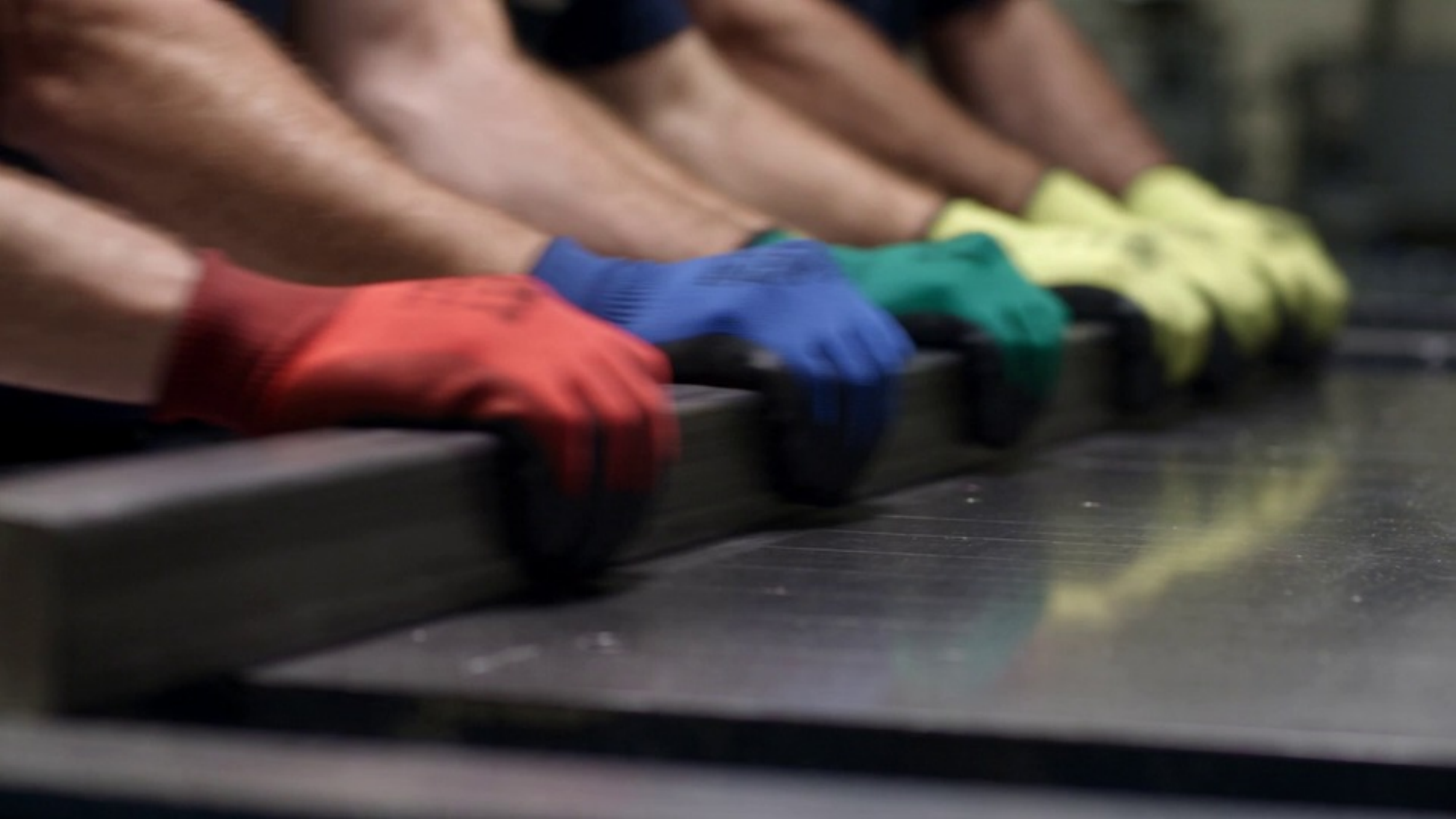}
        & \appcaseframe{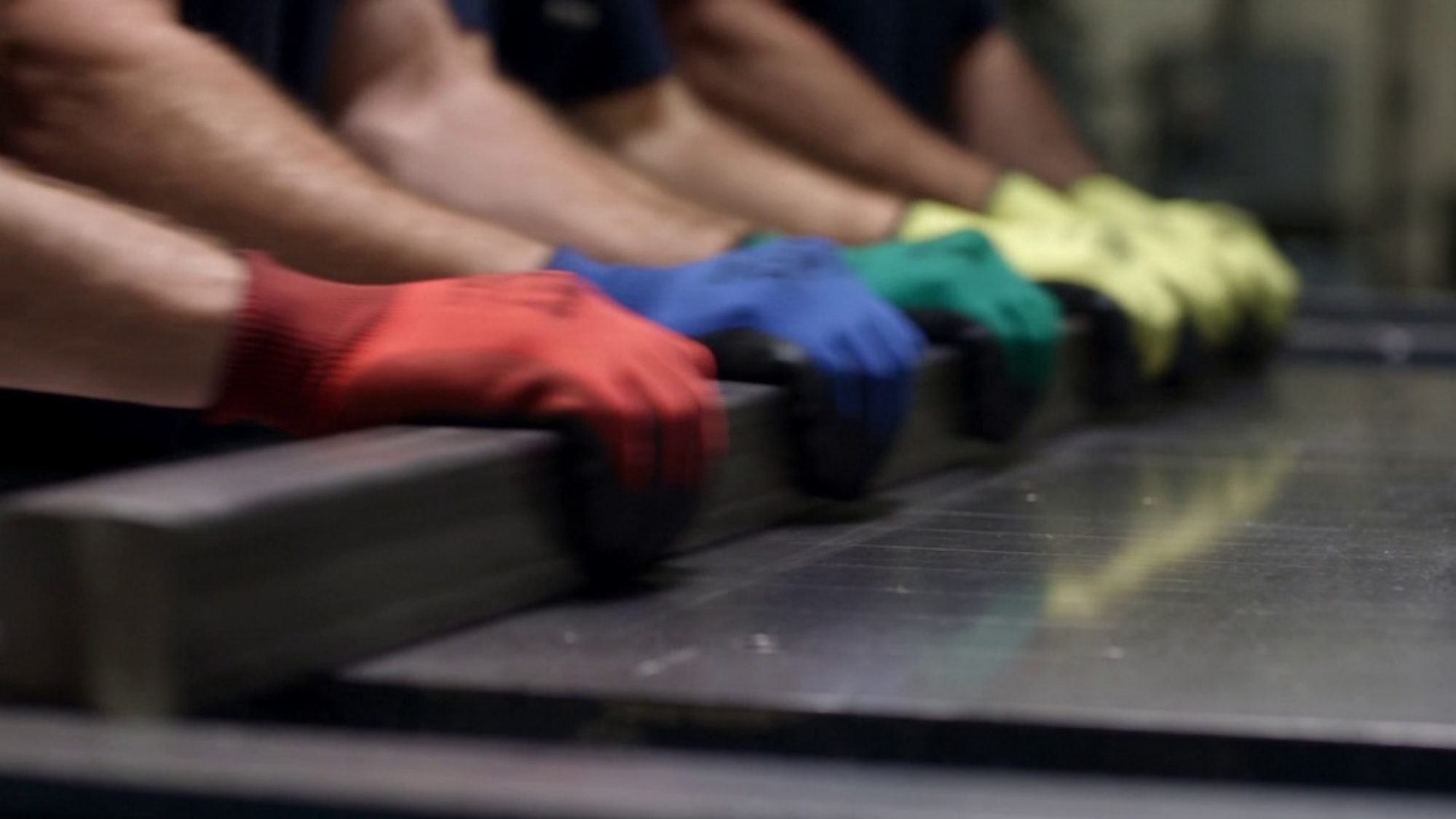}
        & \appcaseframe{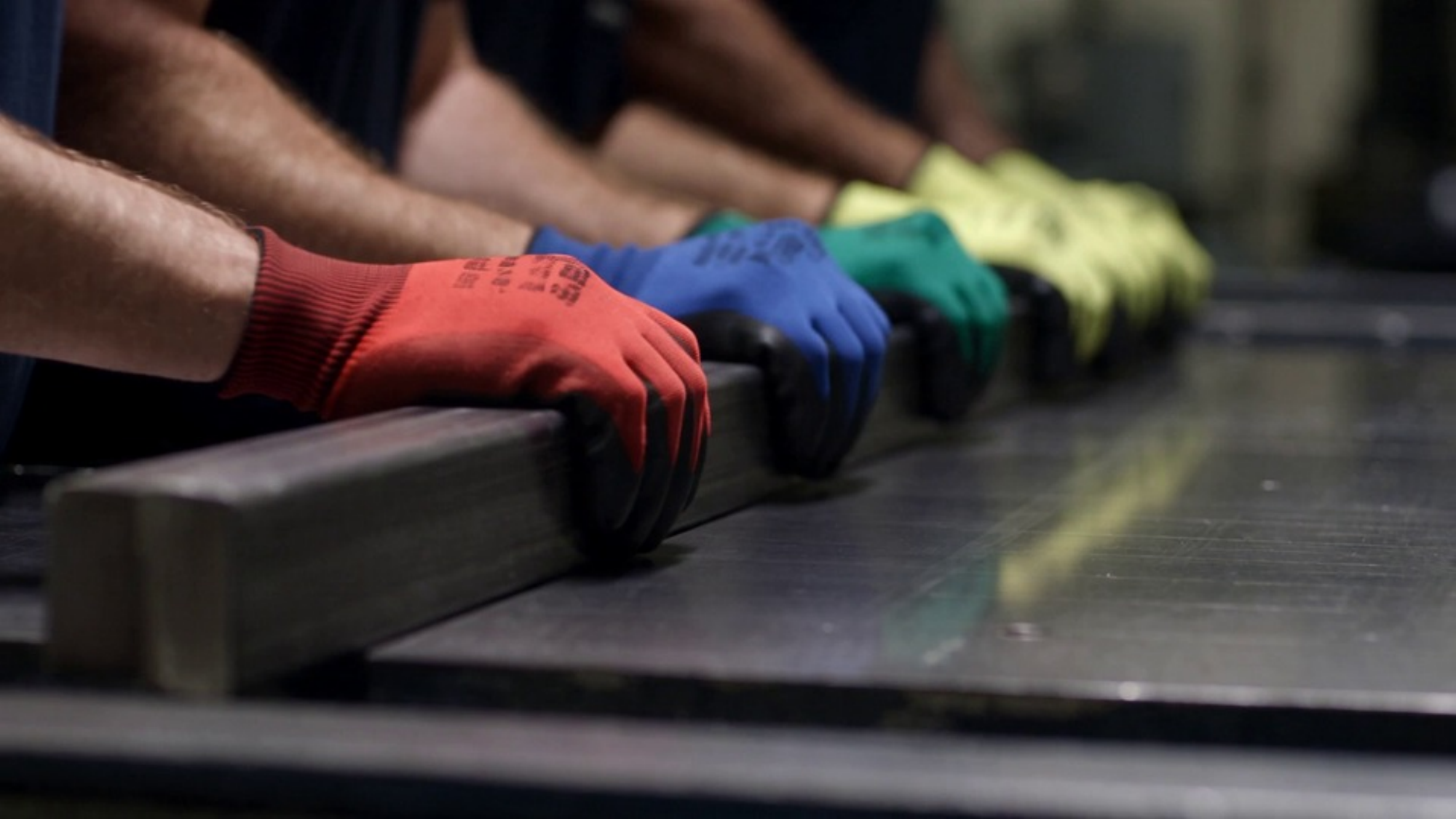}
    \end{tabular}
\par\vspace{3pt}\noindent\begin{tabular}{@{}c@{\hspace{\appGap}}c@{\hspace{\appGap}}c@{\hspace{\appGap}}c@{}}
        \appcaselabel{TIDE}{\textbf{2/6 passed}}{\textcolor{appFail}{Fails (1), (2), (3), (6)}}
        & \appcaseframe{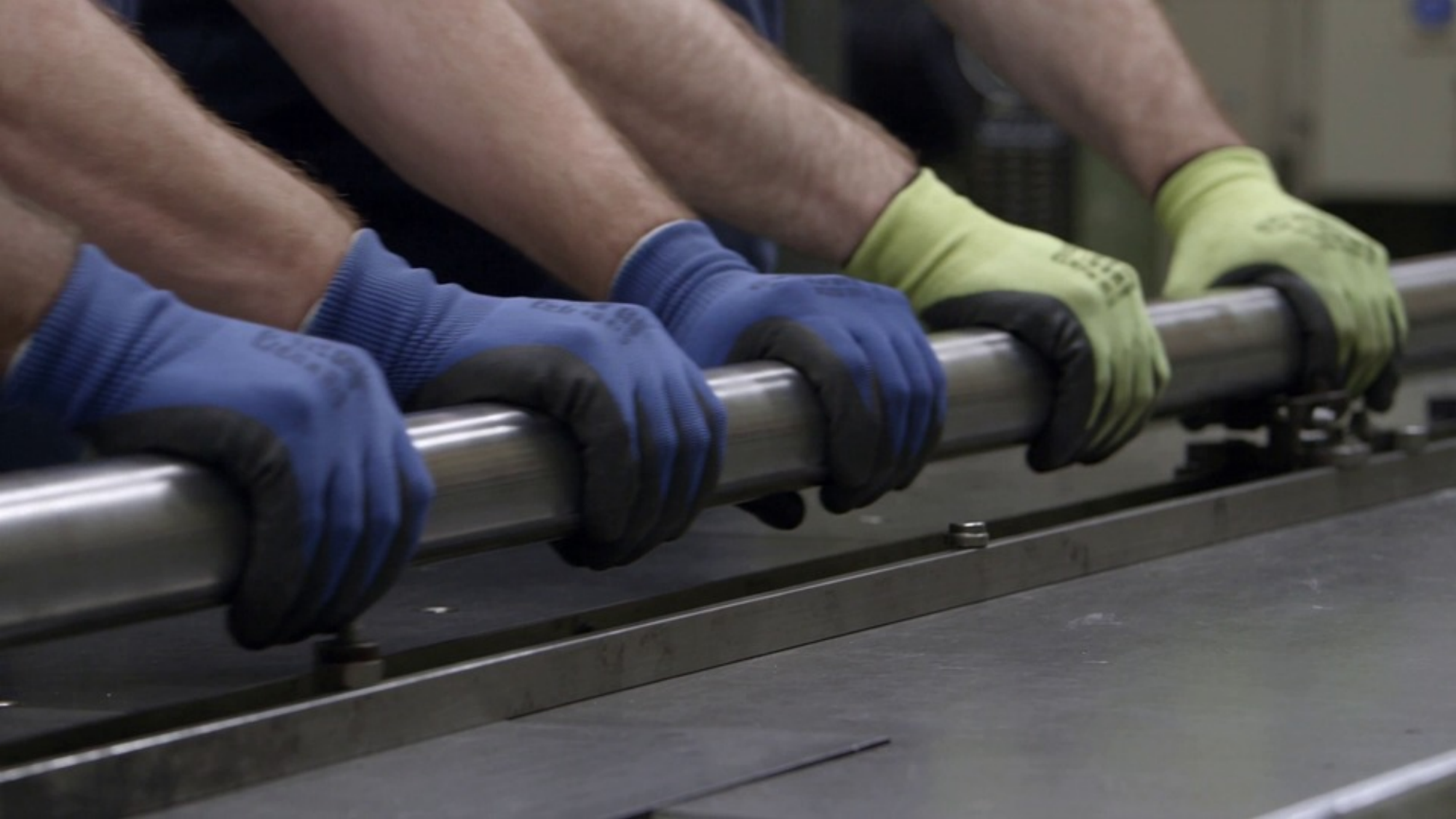}
        & \appcaseframe{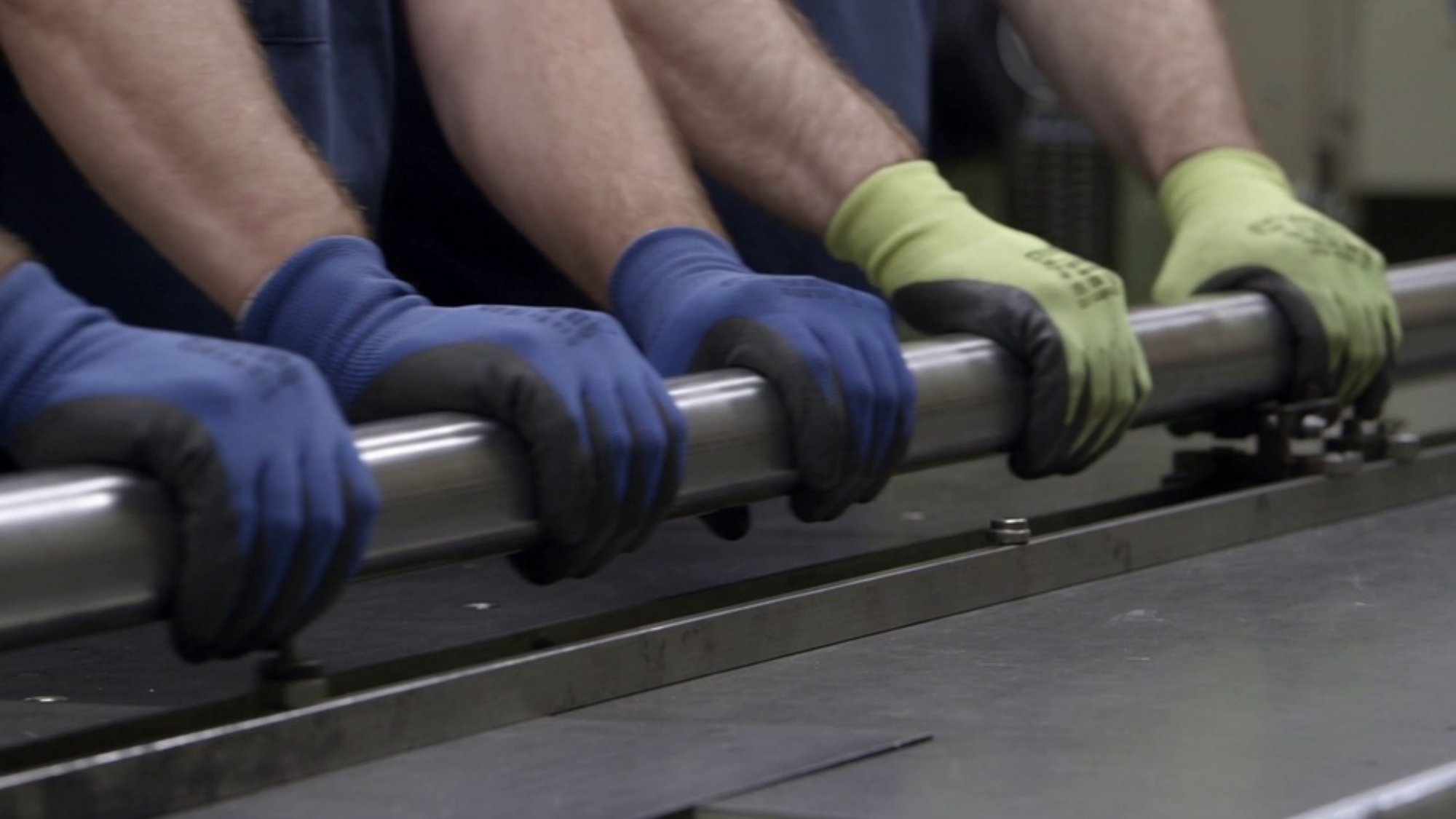}
        & \appcaseframe{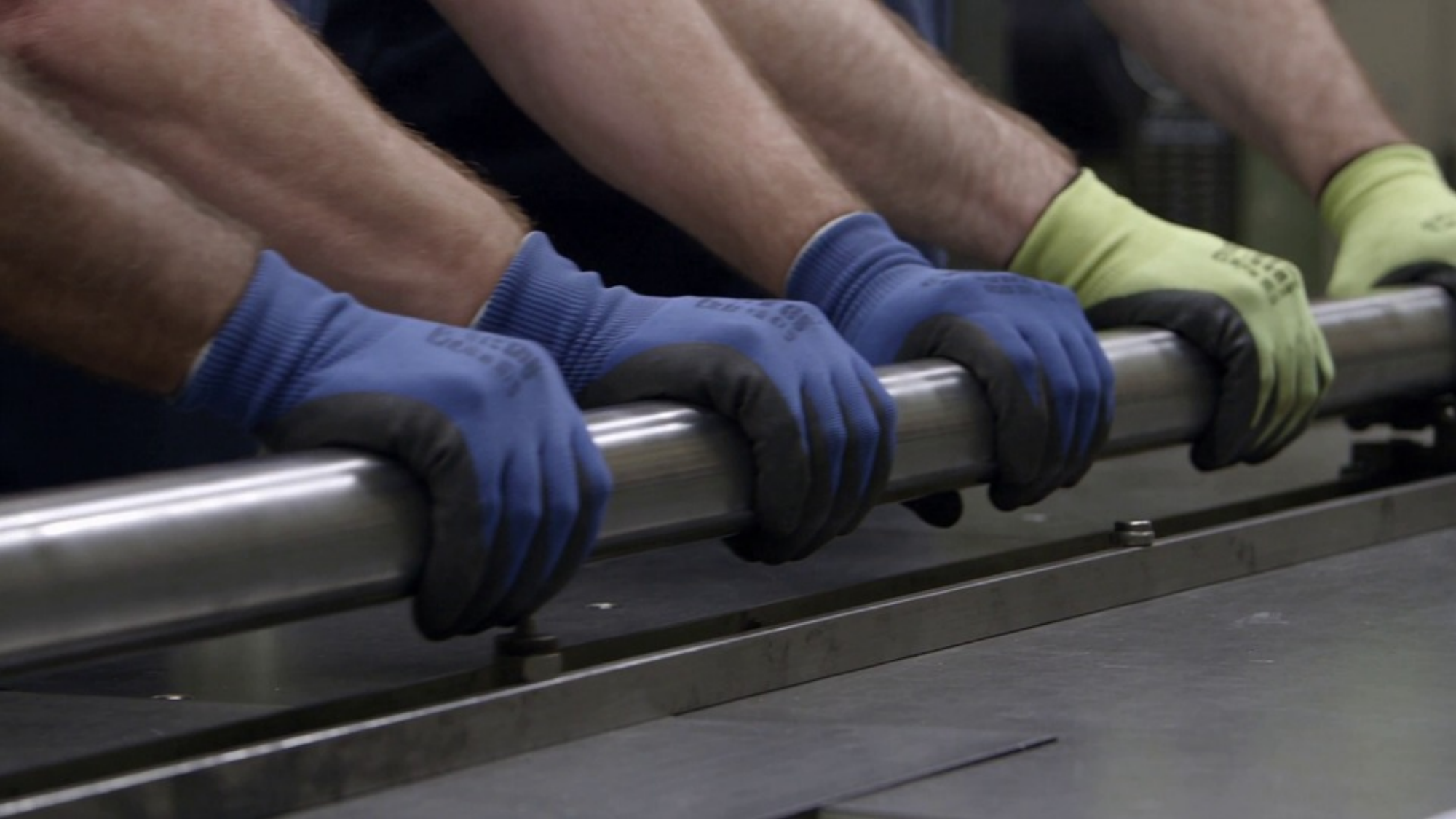}
    \end{tabular}
\par\vspace{3pt}\noindent\begin{tabular}{@{}c@{\hspace{\appGap}}c@{\hspace{\appGap}}c@{\hspace{\appGap}}c@{}}
        \appcaselabel{TIDE + NeurCo (T)}{\textbf{2/6 passed}}{\textcolor{appFail}{Fails (1), (2), (3), (6)}}
        & \appcaseframe{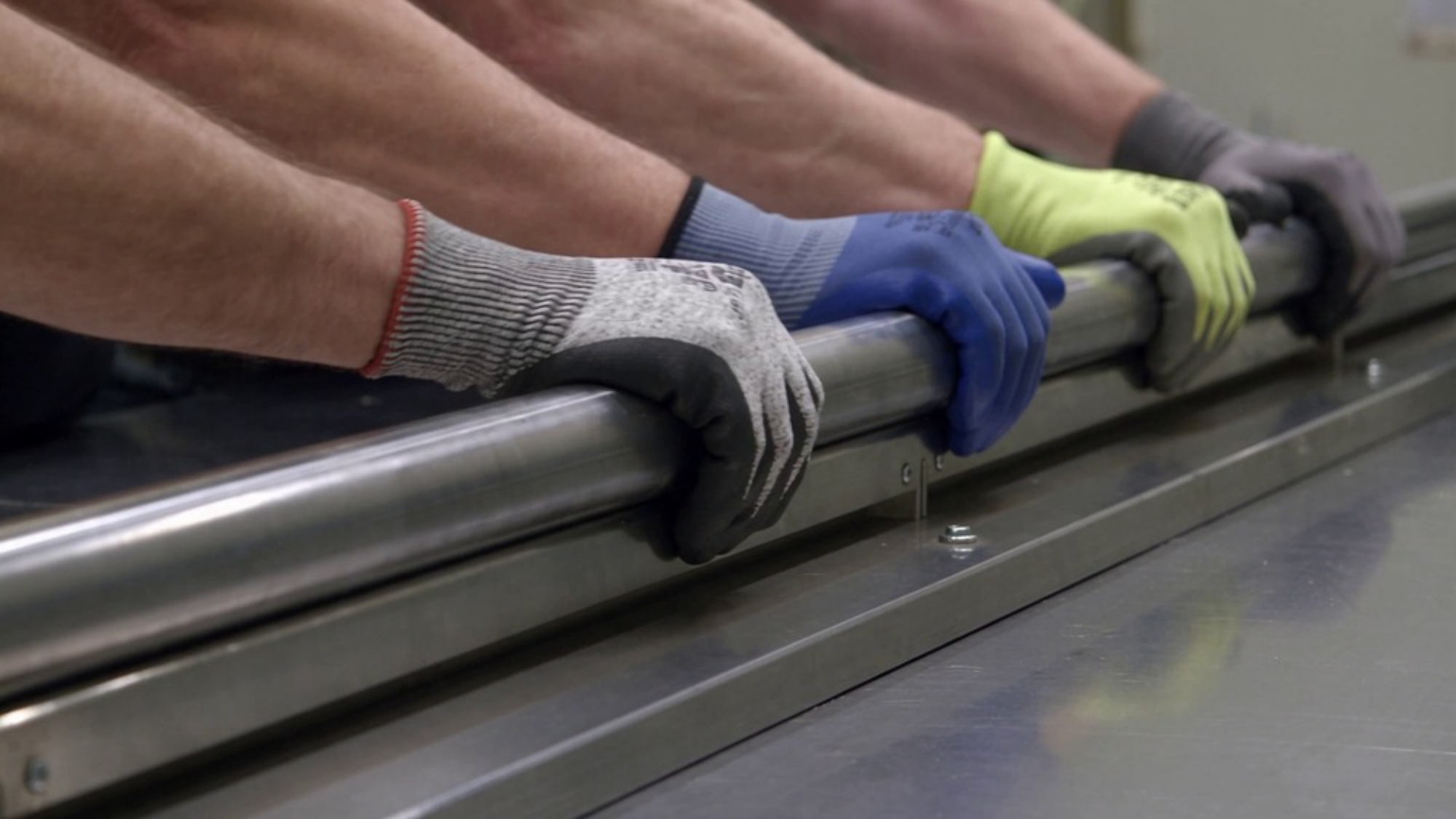}
        & \appcaseframe{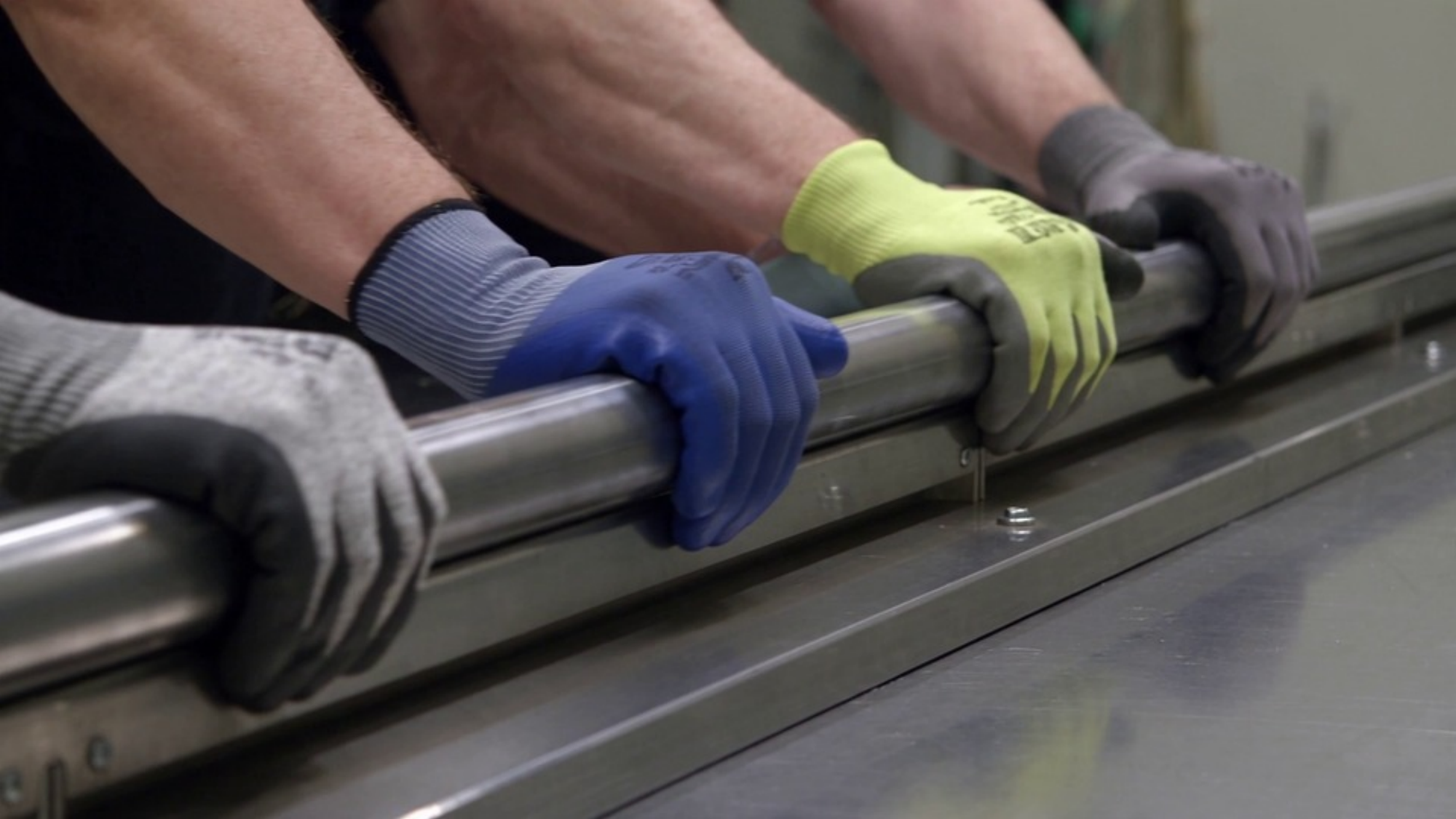}
        & \appcaseframe{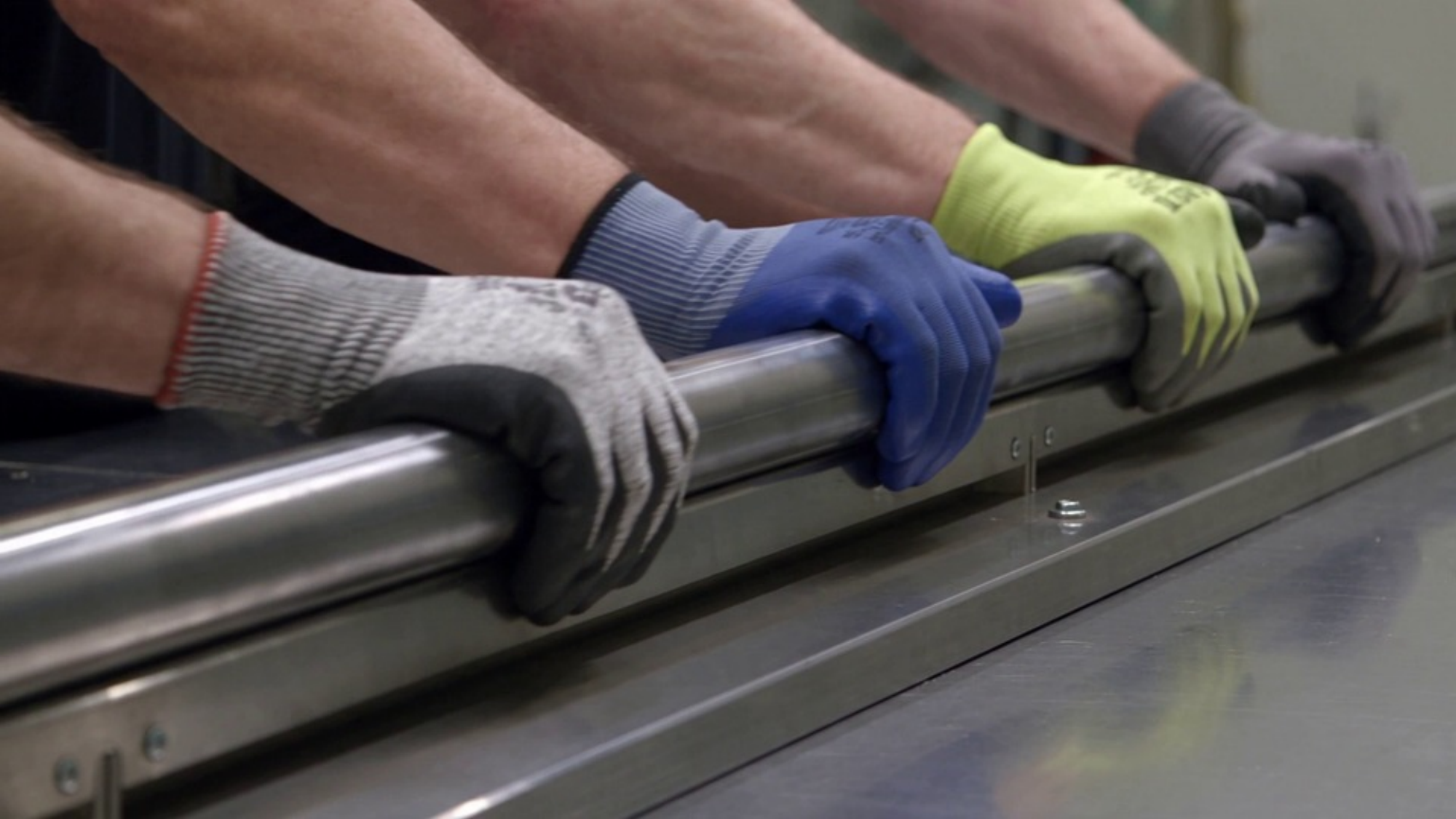}
    \end{tabular}
\par\vspace{3pt}\noindent\begin{tabular}{@{}c@{\hspace{\appGap}}c@{\hspace{\appGap}}c@{\hspace{\appGap}}c@{}}
        \appcaselabel{TIDE + NeurCo (T+O)}{\textcolor{appBlue}{\textbf{5/6 passed}}}{Reference + text; (2) unclear}
        & \appcaseframe{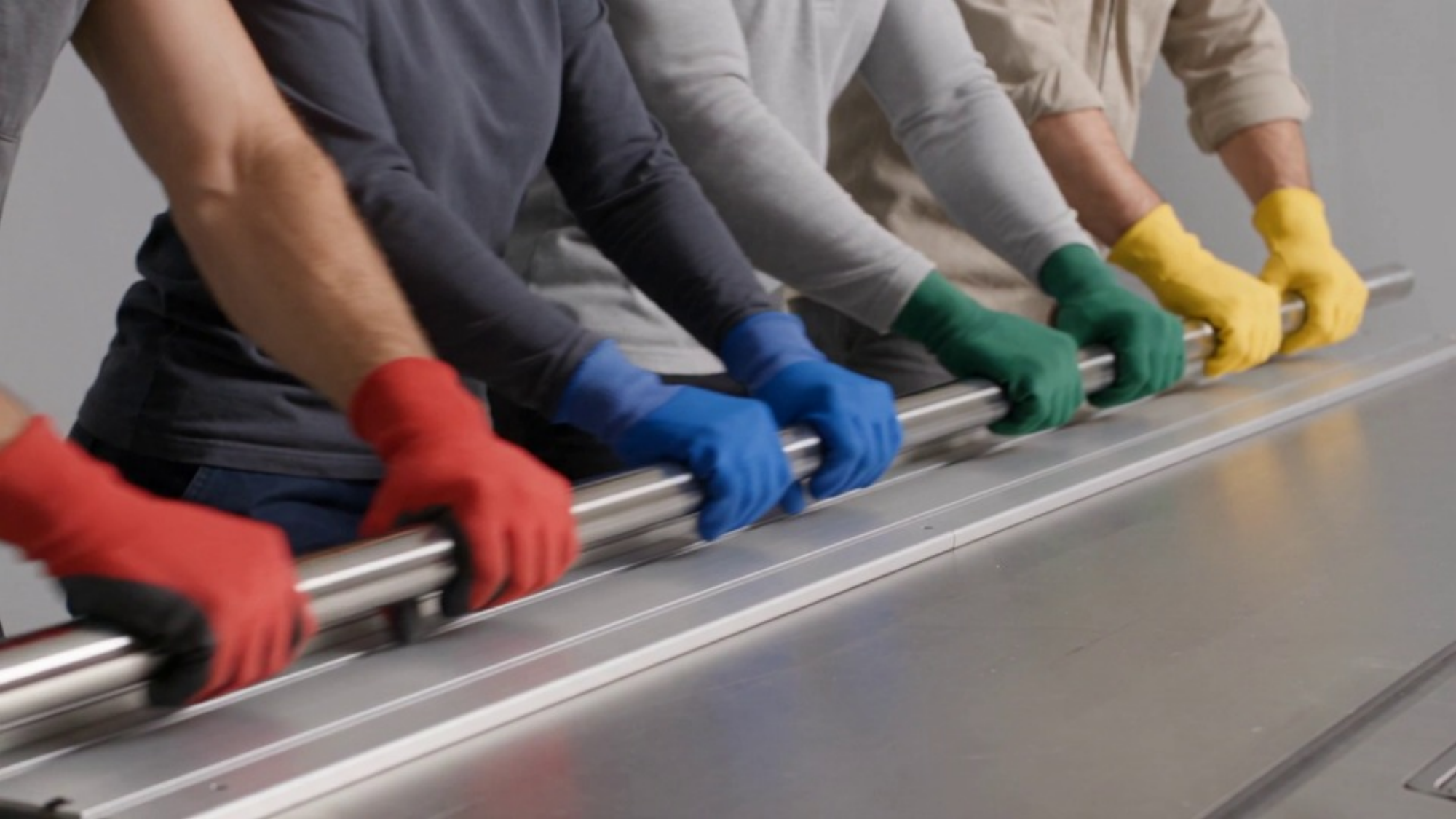}
        & \appcaseframe{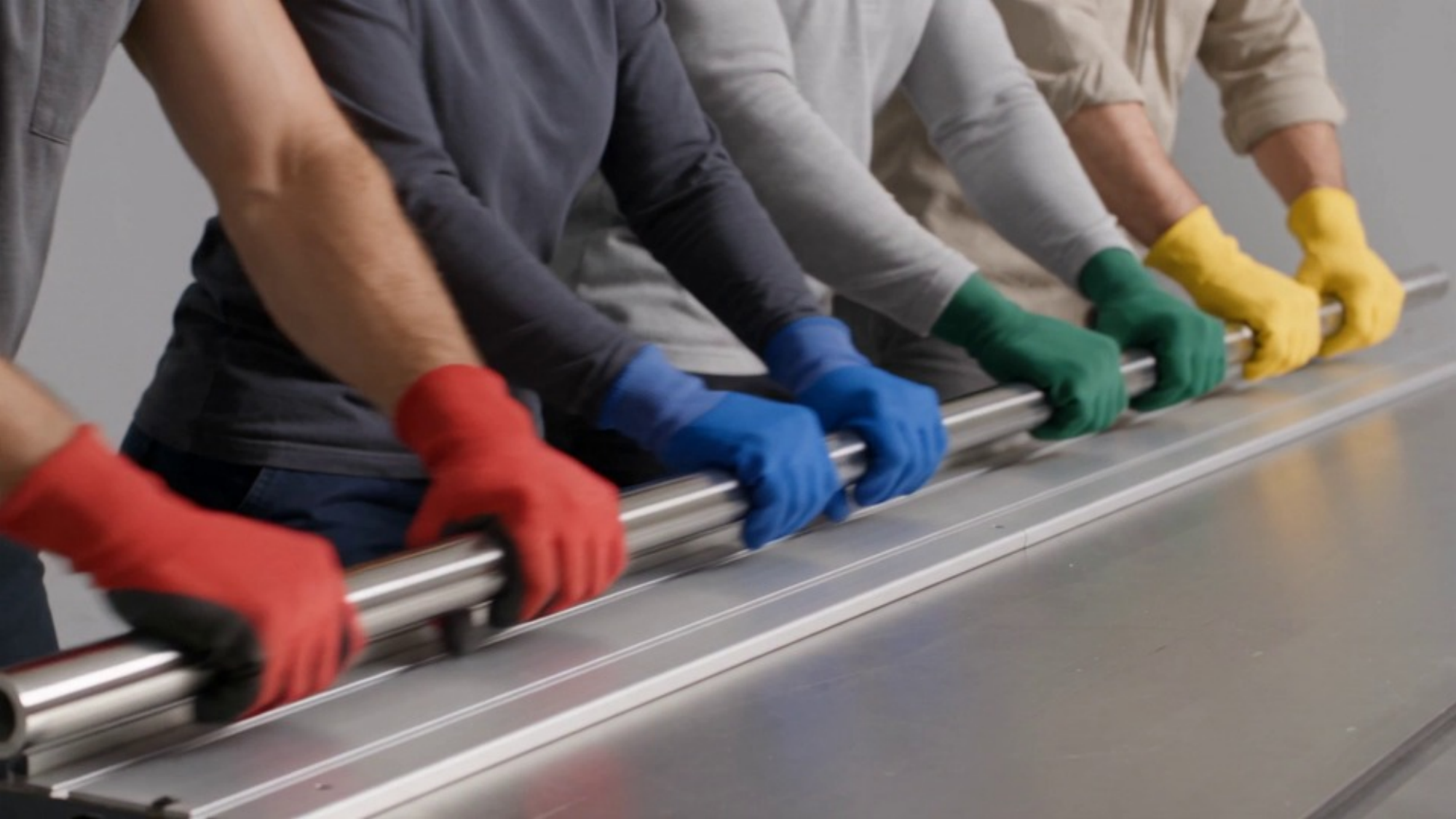}
        & \appcaseframe{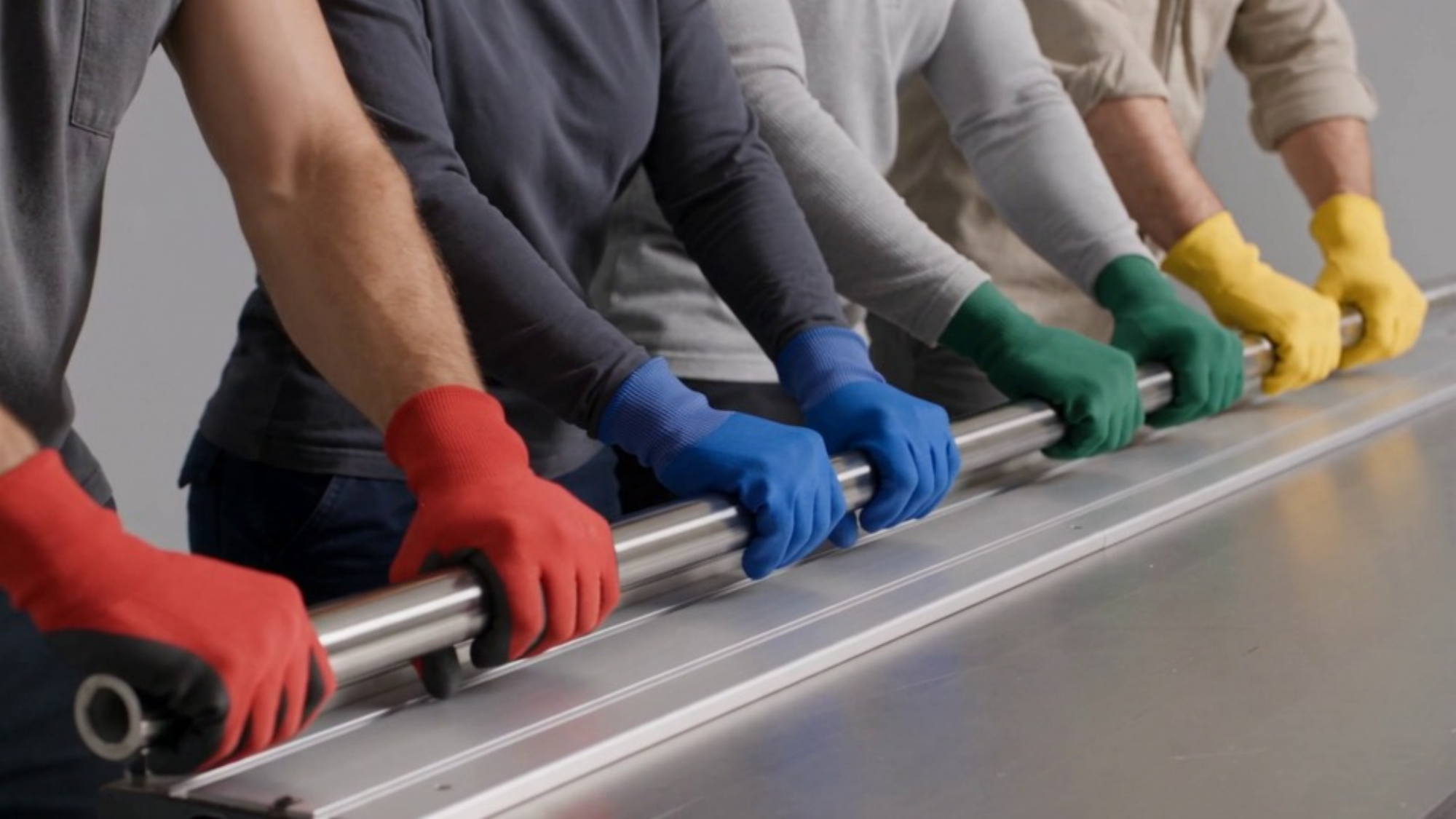}
    \end{tabular}
    \nopagebreak[3]\caption{\textbf{Four glove colors on Seedance 2.5 with TIDE.} The constructed reference switches execution from T2V to R2V; the output shows all four glove colors, and the pan and speed requirements also pass.}
    \label{fig:case_sd_gloves}
    \endgroup
\par\endgroup
\end{appcaseblock}

\clearpage
\begin{appcaseblock}
\noindent\textbf{Handwritten sentence from a generated first frame.}
Figure~\ref{fig:case_sd_pen} shows that a generated first frame preserves a long handwritten sentence on Seedance~2.5. Native and text-only generation produce malformed or cropped text ($0/3$). NeurCo renders the target sentence into the opening frame and switches from T2V to I2V; the video preserves the full sentence while the pen adds the final word ($2/3$; the initial tip contact remains unclear).

\par\smallskip\noindent\begingroup\centering
    \centering
    \begingroup
    \appcaseinit
    \appcaseinputas{T2V}{A single pink gel pen bumps tip-first against rough, textured fabric, briefly indenting it, then drags across the fibers to write the final word of a handwritten sentence. Pink marks remain behind the moving tip. The completed sentence is clearly readable: ``Pink ink flows across the rough fabric as the pen traces a fine line, leaving a bright lasting mark.''}{}{\appinputimg{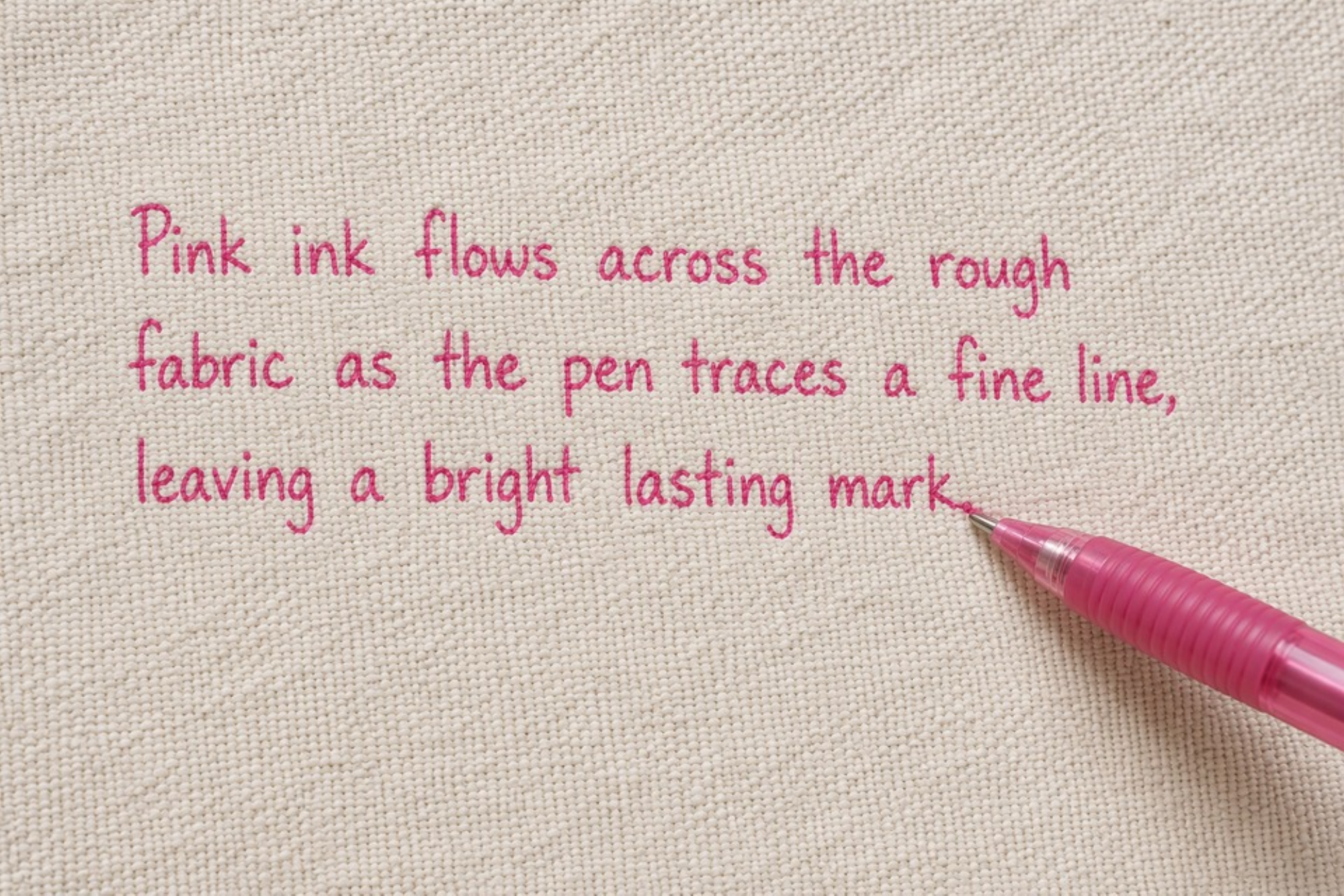}{Generated first frame (T+O row)}%
    }{%
        \begin{tabularx}{\linewidth}{@{}XXX@{}}
            \textcolor{appBlue}{\textbf{(1) Full sentence, readable and correct}} & (2) Tip bumps and indents fabric & (3) One pink gel pen writing
        \end{tabularx}}
    \begin{tabular}{@{}c@{\hspace{\appGap}}c@{\hspace{\appGap}}c@{\hspace{\appGap}}c@{}}
        \parbox[c]{\appLabelW}{\centering\textbf{Method}} & \textbf{Early} & \textbf{Middle} & \textbf{Late} \\[2pt]
        \appcaselabel{Native}{\textbf{0/3 passed}}{\textcolor{appFail}{Fails (1), (2), (3)}}
        & \appcaseframe{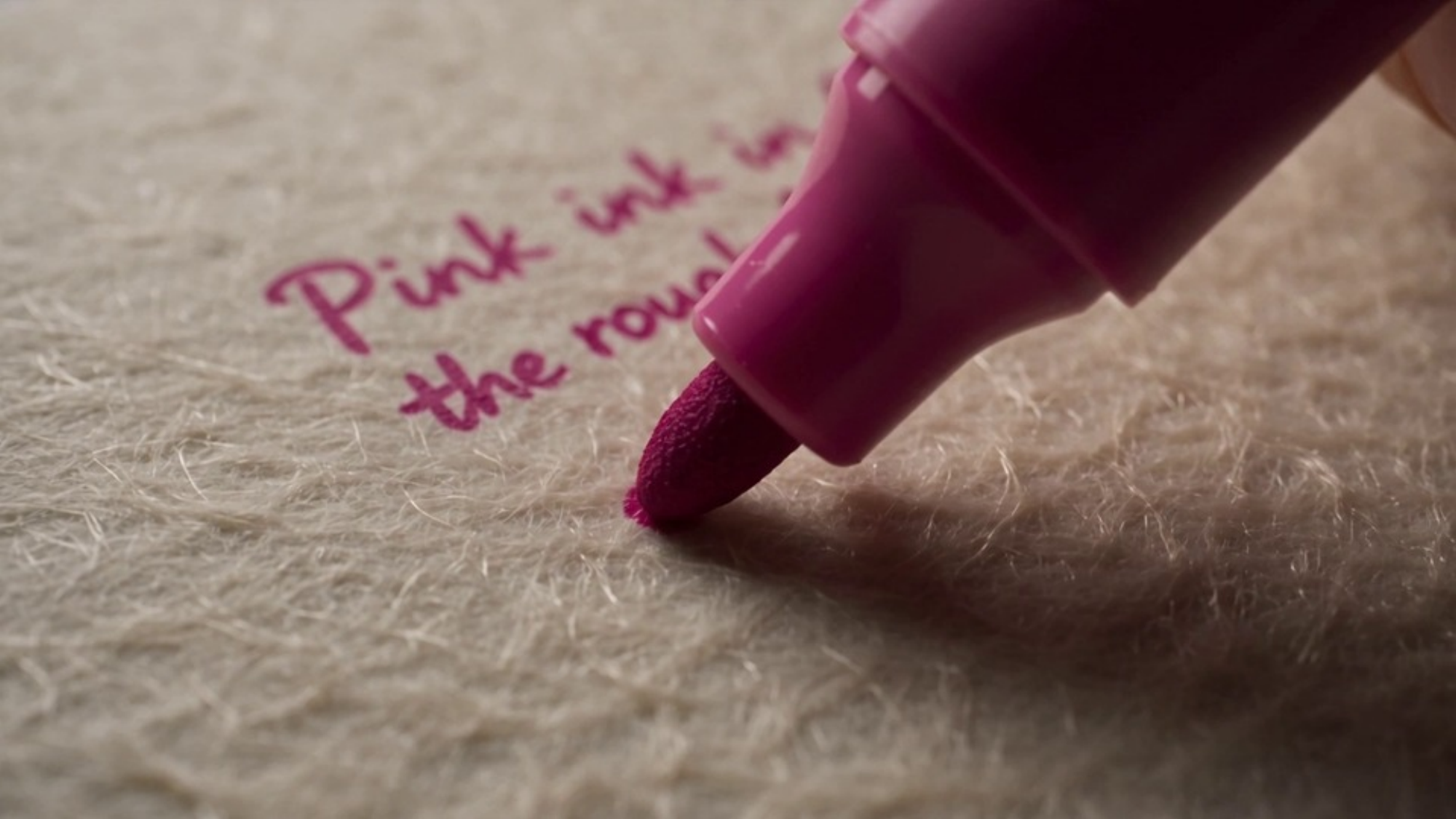}
        & \appcaseframe{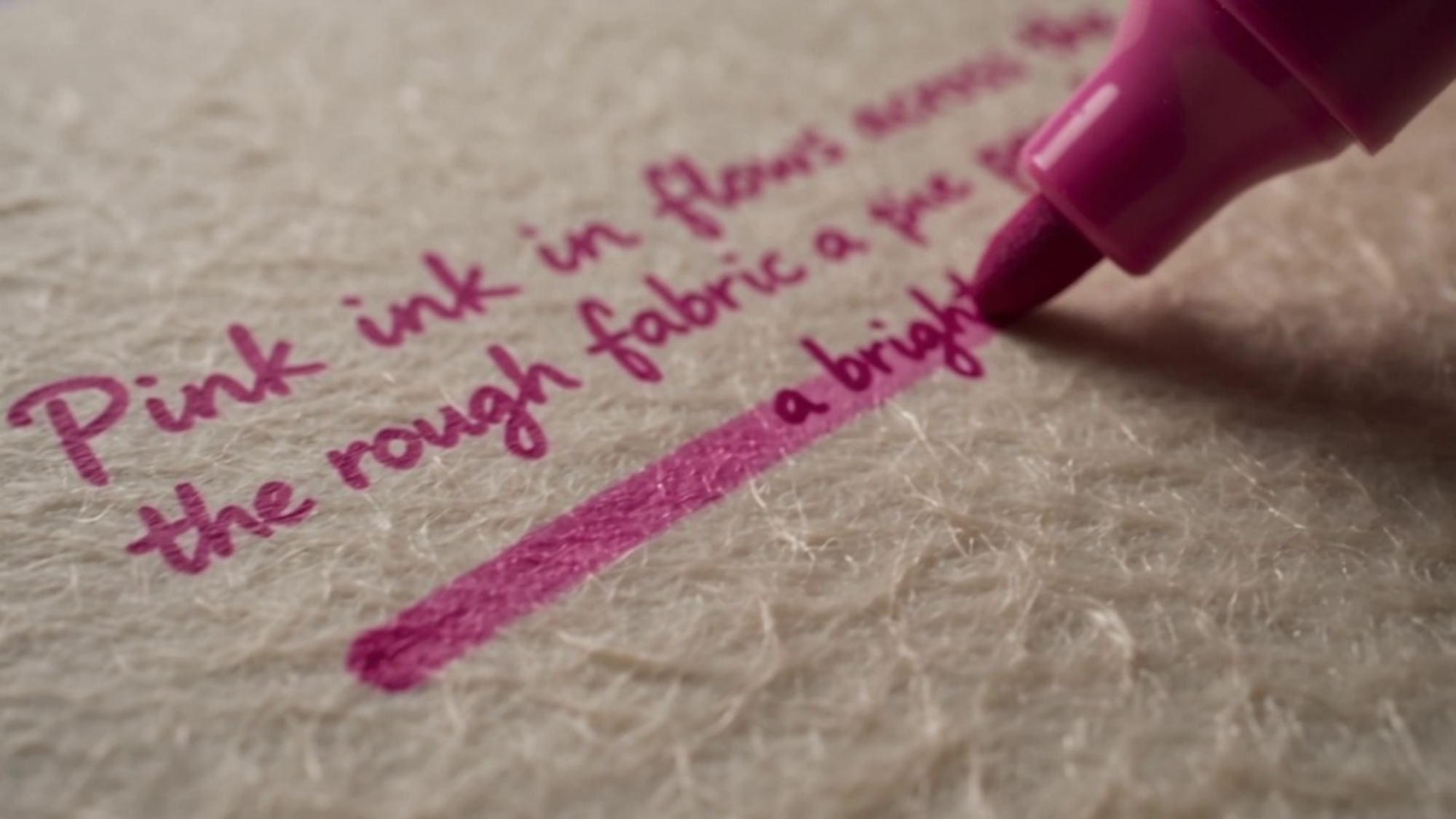}
        & \appcaseframe{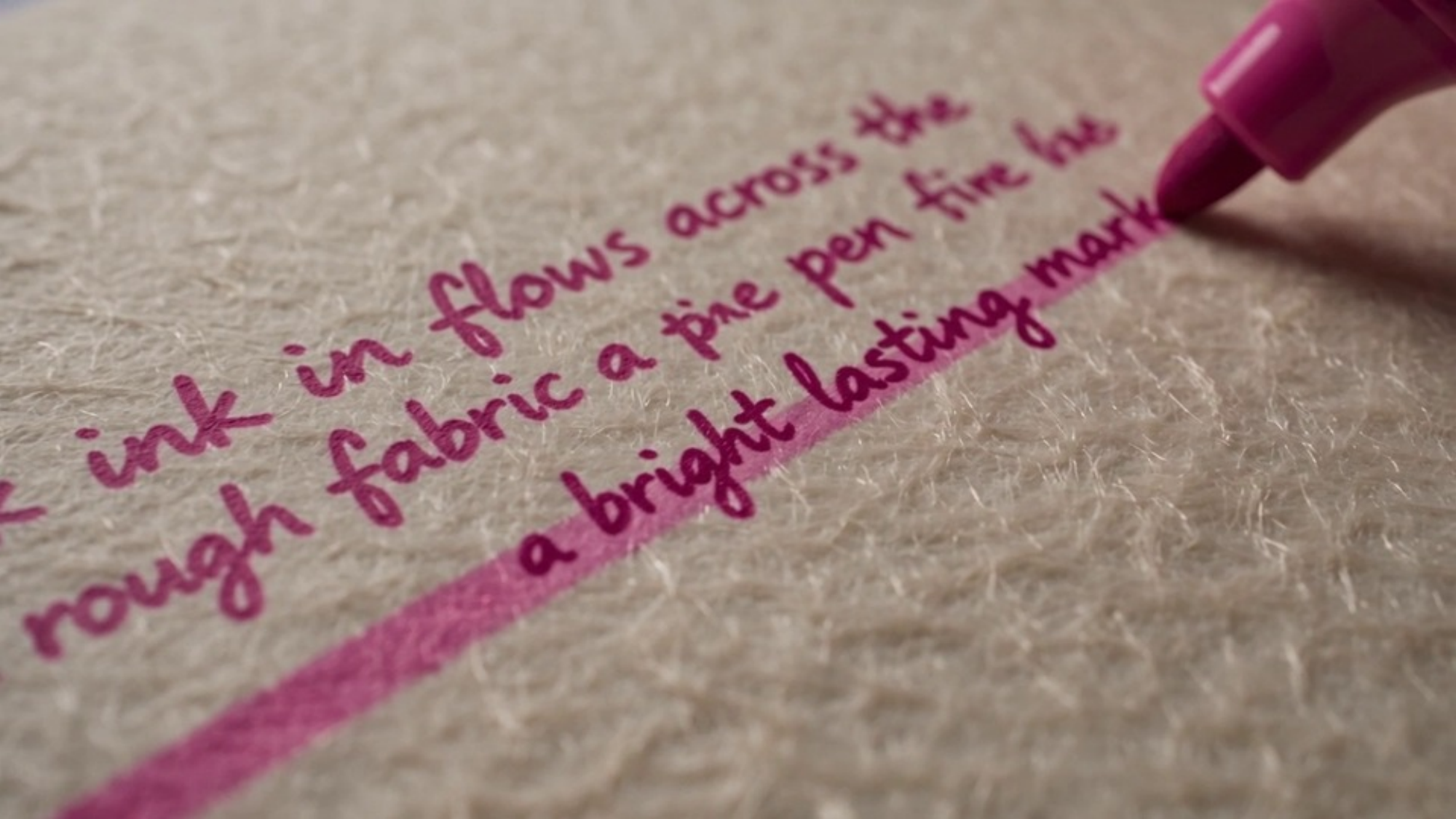}
    \end{tabular}
\par\vspace{3pt}\noindent\begin{tabular}{@{}c@{\hspace{\appGap}}c@{\hspace{\appGap}}c@{\hspace{\appGap}}c@{}}
        \appcaselabel{NeurCo (T)}{\textbf{0/3 passed}}{\textcolor{appFail}{Fails (1), (2), (3)}}
        & \appcaseframe{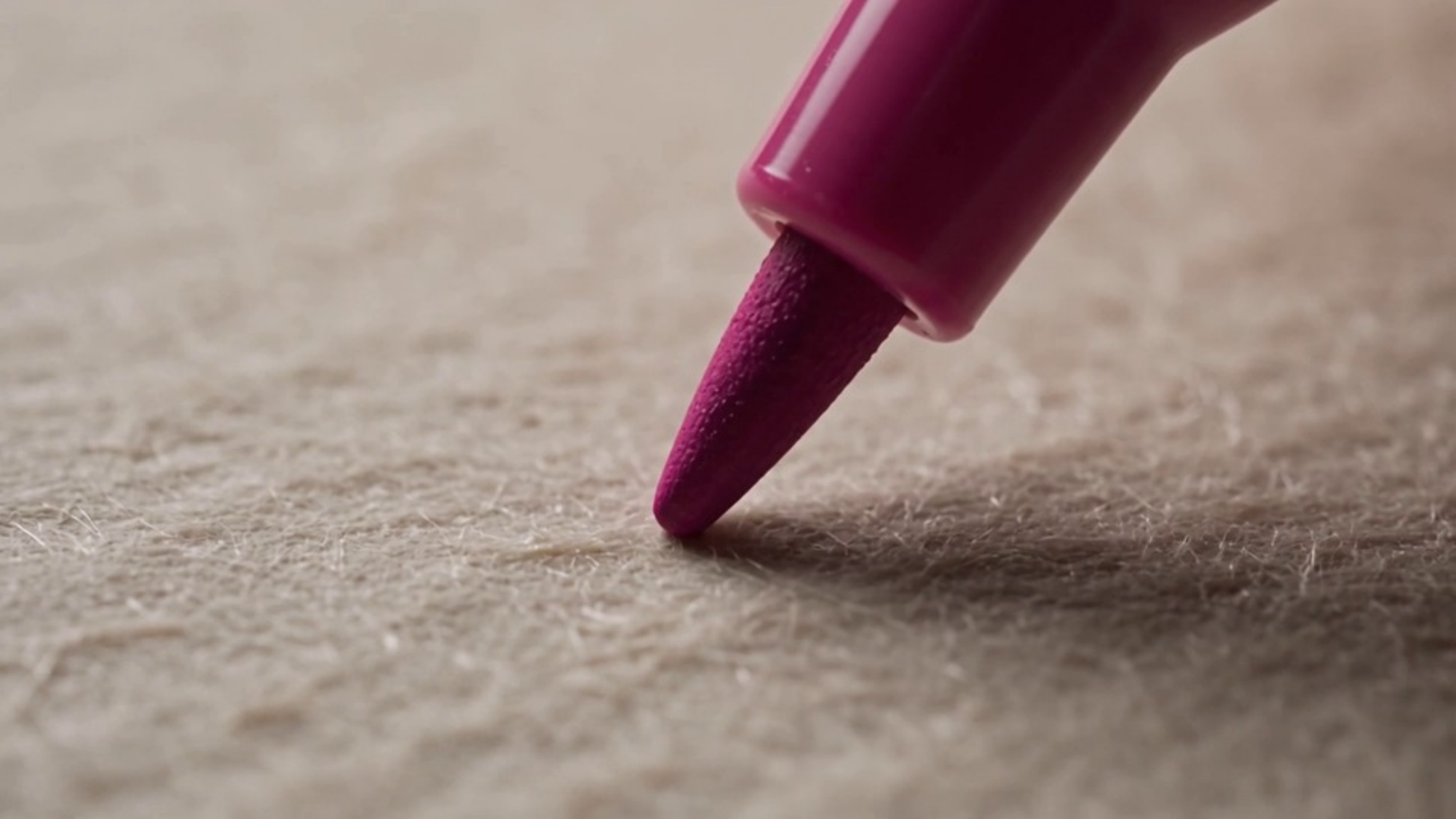}
        & \appcaseframe{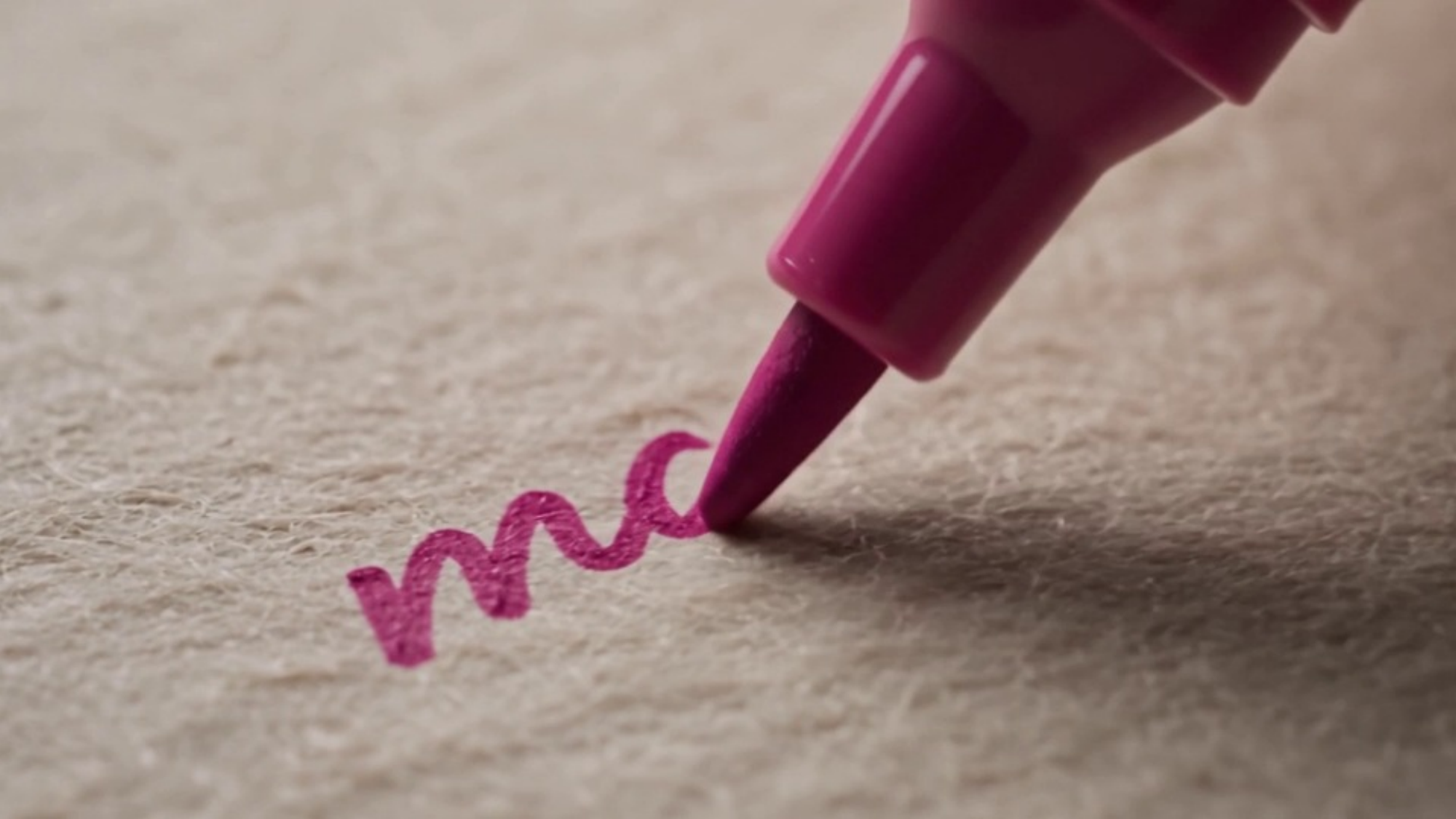}
        & \appcaseframe{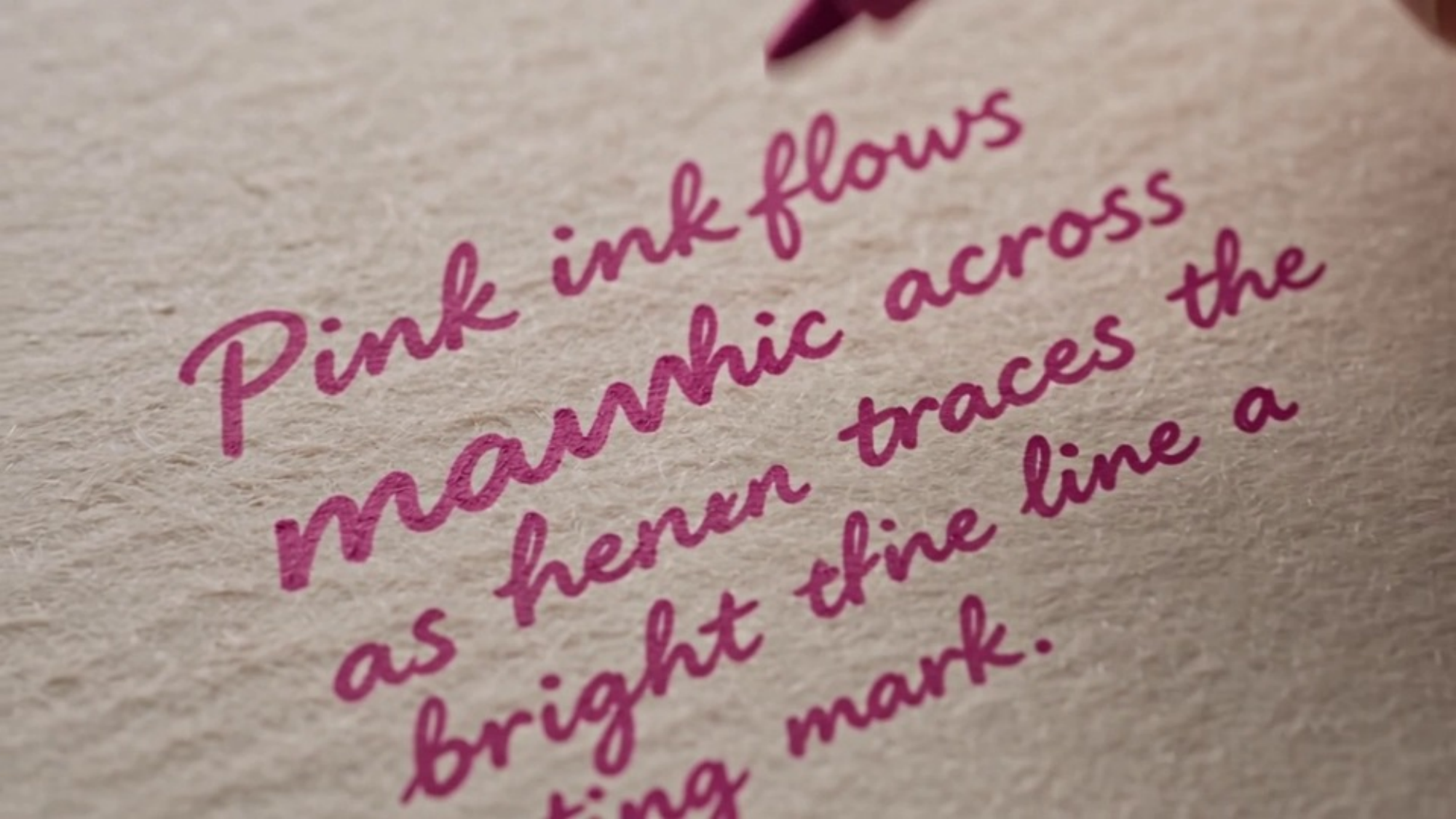}
    \end{tabular}
\par\vspace{3pt}\noindent\begin{tabular}{@{}c@{\hspace{\appGap}}c@{\hspace{\appGap}}c@{\hspace{\appGap}}c@{}}
        \appcaselabel{NeurCo (T+O)}{\textcolor{appBlue}{\textbf{2/3 passed}}}{First frame + text; (2) unclear}
        & \appcaseframe{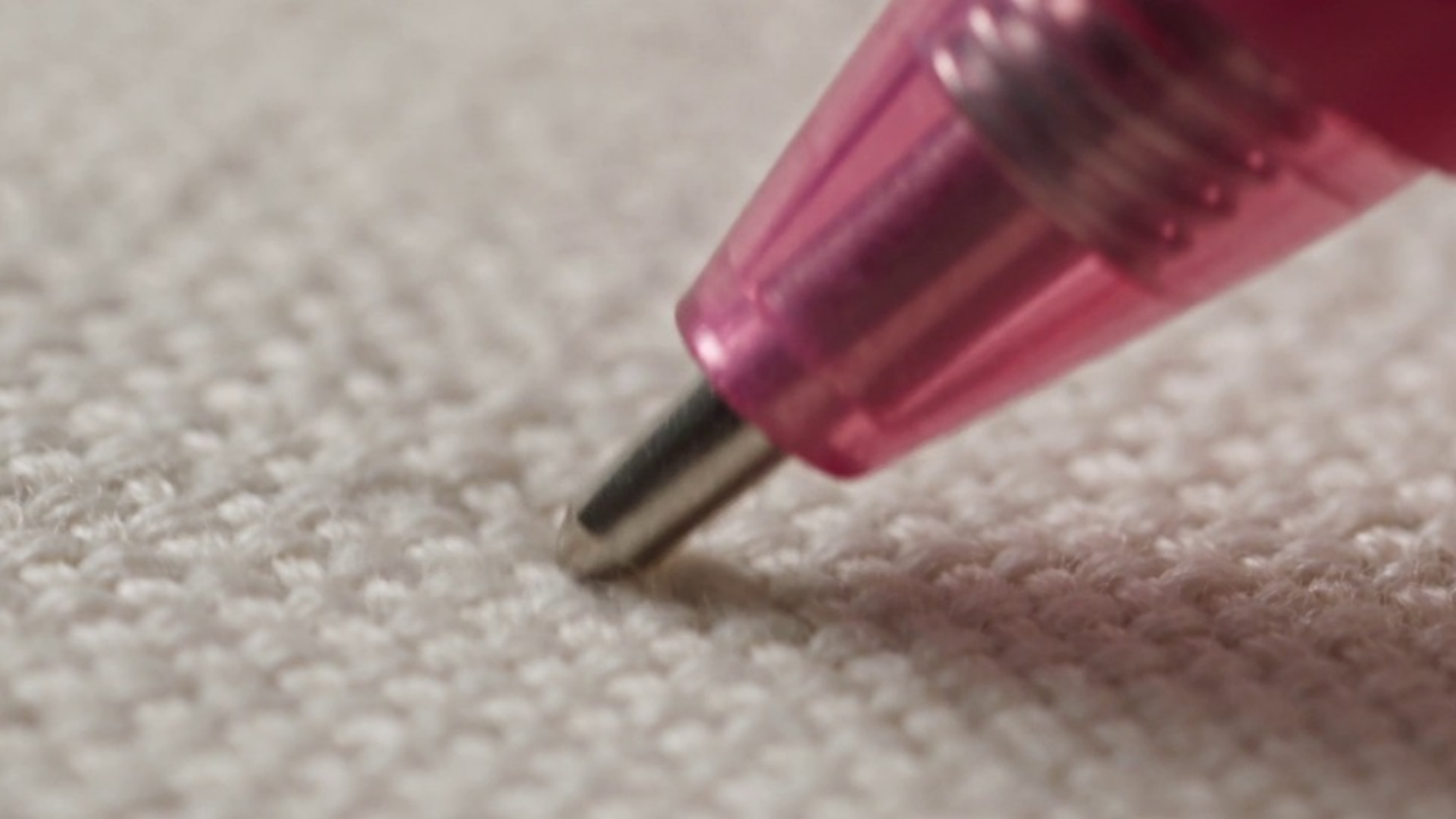}
        & \appcaseframe{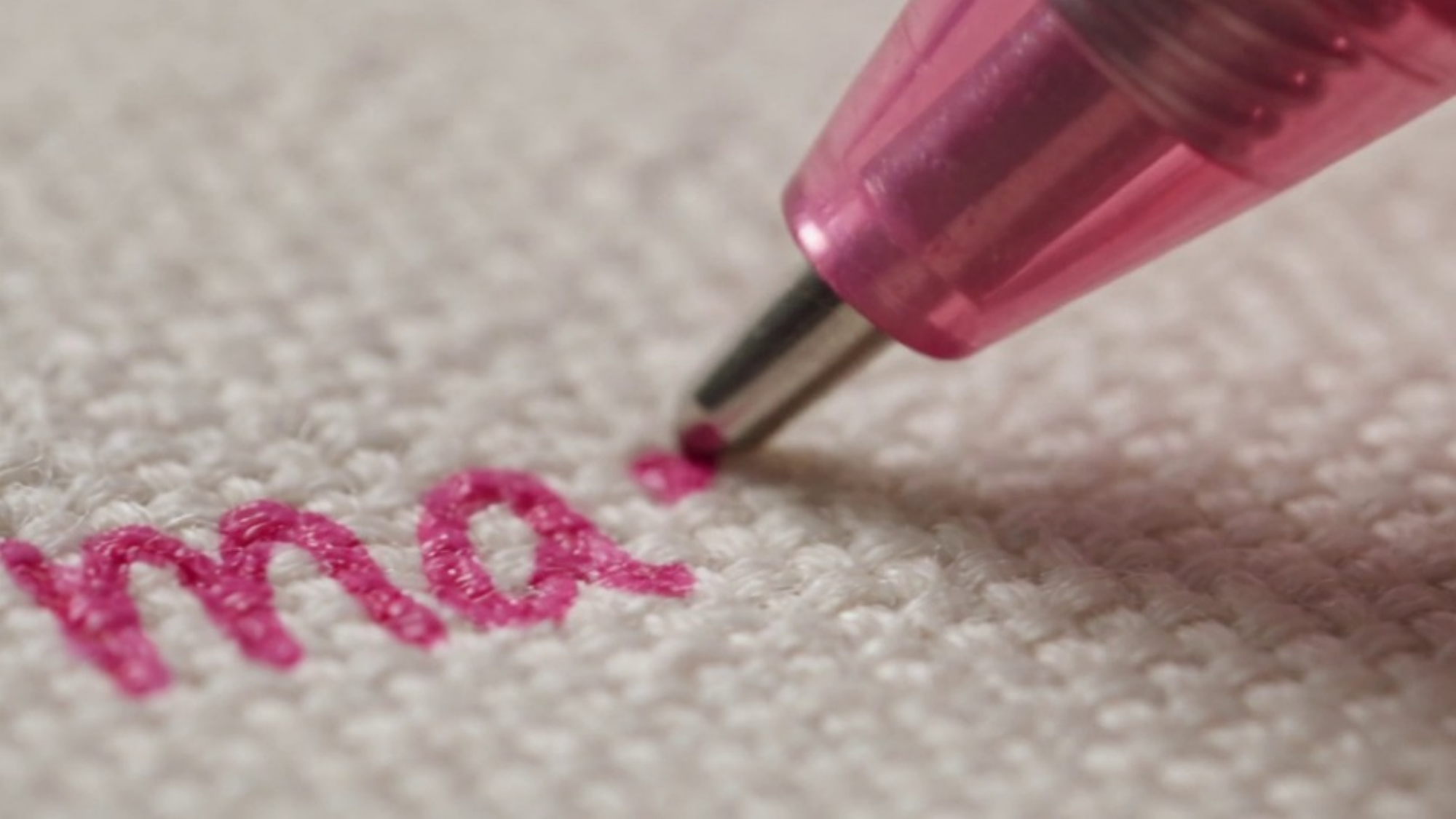}
        & \appcaseframe{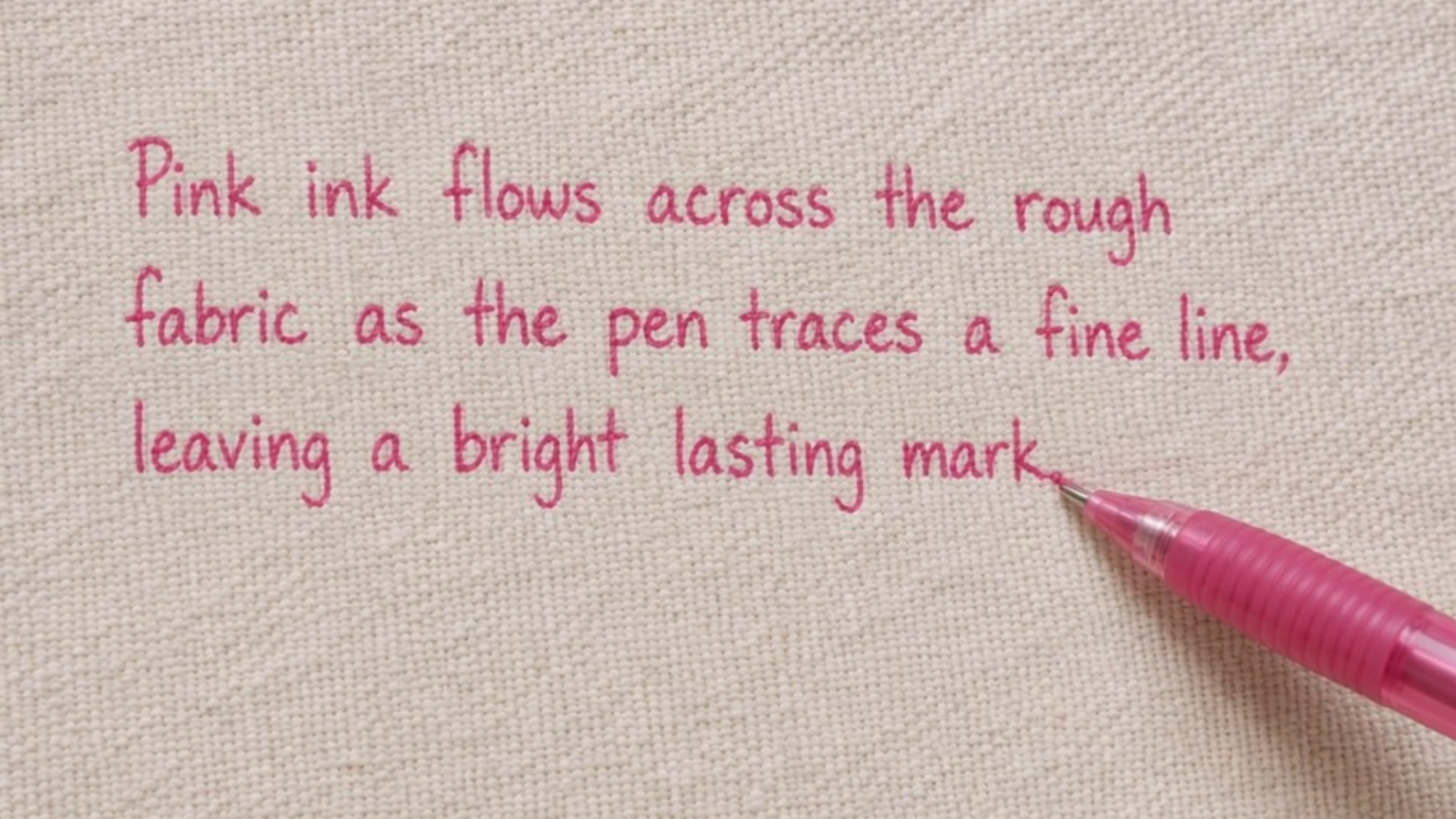}
    \end{tabular}
    \nopagebreak[3]\caption{\textbf{First-frame conditioning preserves readable handwriting on Seedance 2.5.} T2V-to-I2V compilation retains the complete target sentence while a single pen continues writing the final word ($2/3$).}
    \label{fig:case_sd_pen}
    \endgroup
\par\endgroup
\end{appcaseblock}

\clearpage
\subsection{Failure Cases and Trade-offs}
\label{sec:app_cases_failures}

Compilation also reveals trade-offs in contact geometry, temporal
progression, and scene composition. The following cases show an
incorrect text revision, a first-frame condition that disturbs a
procedure, and alternative plans for a multi-event coastline.

\begin{appcaseblock}
\noindent\textbf{Disrupted contact geometry.}
Targeted rewriting damages a correct contact event in the
Seedance~2.5 coin sequence (Figure~\ref{fig:case_sd_coin}). Native
satisfies $5/6$ requirements, including the coin rim striking a stop
pin. Both text-compiled outputs instead show the rod intersecting the
coin and satisfy $3/6$, with rotation speed still incorrect. Neither
compiled plan uses an external condition.

\par\smallskip\noindent\begingroup\centering
    \centering
    \begingroup
    \appcaseinit
    \appcaseinputas{I2V}{Begin with the antique gold coin facing forward against the dark textured backdrop,
    matching the first frame. With the camera locked off, the coin rapidly turns 90 degrees
    around its vertical center axis. Its rim strikes a small fixed stop pin, previously
    concealed behind the coin, abruptly arresting the rotation. The coin remains upright,
    motionless and edge-on for the remainder of the shot.}{\appinputimg{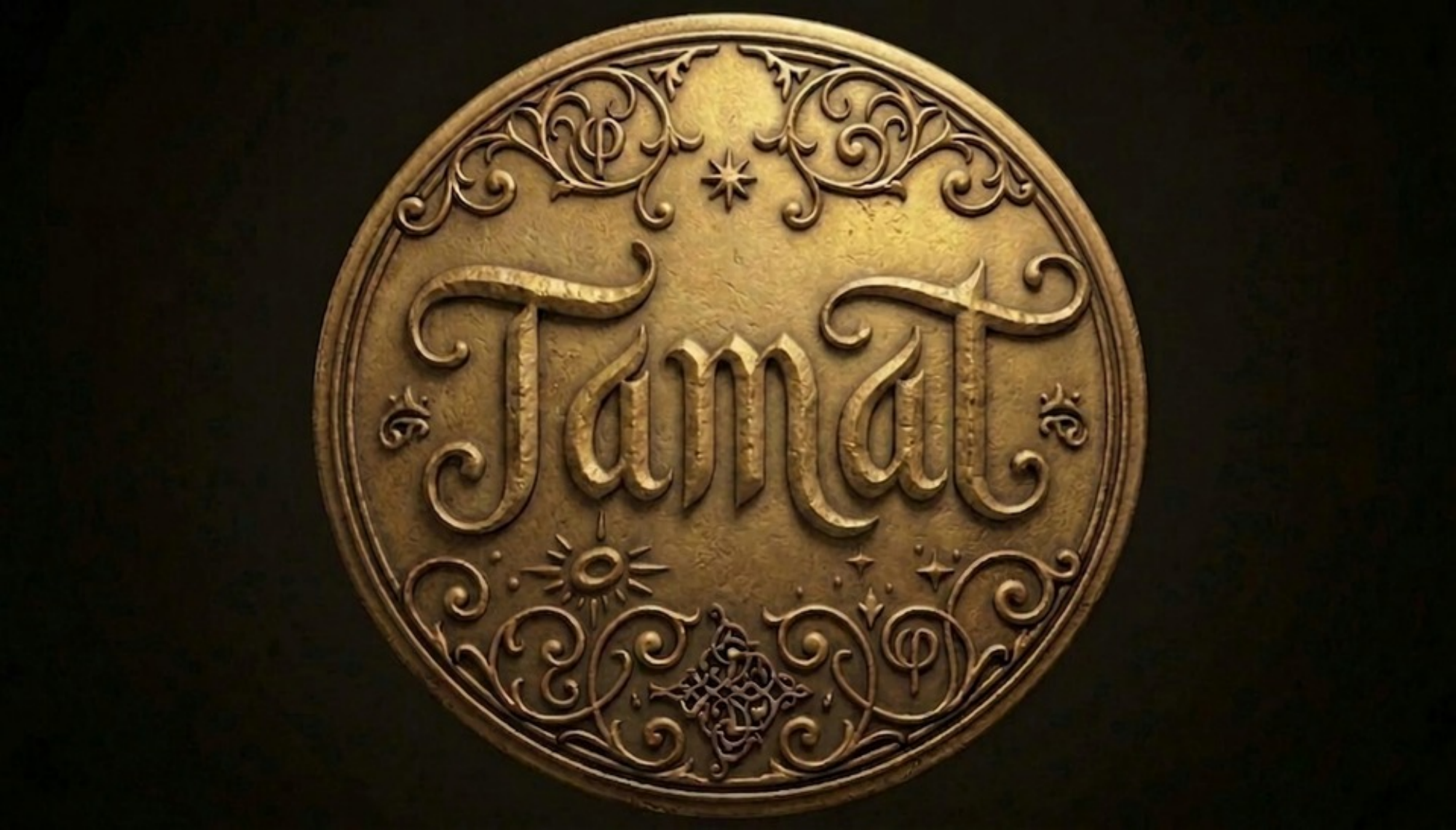}{First frame}%
    }{}{%
        \begin{tabularx}{\linewidth}{@{}XXX@{}}
            (1) Quarter turn about the vertical axis & \textcolor{appFail}{\textbf{(2) Stops on contact with the pin}} & (3) Fast turn \\
            (4) Steady framing & \textcolor{appFail}{\textbf{(5) Rim contacts the pin}} & (6) Static camera
        \end{tabularx}}
    \begin{tabular}{@{}c@{\hspace{\appGap}}c@{\hspace{\appGap}}c@{\hspace{\appGap}}c@{}}
        \parbox[c]{\appLabelW}{\centering\textbf{Method}} & \textbf{Early} & \textbf{Middle} & \textbf{Late} \\[2pt]
        \appcaselabel{Native}{\textcolor{appBlue}{\textbf{5/6 passed}}}{\textcolor{appFail}{Fails (3)}}
        & \appcaseframe{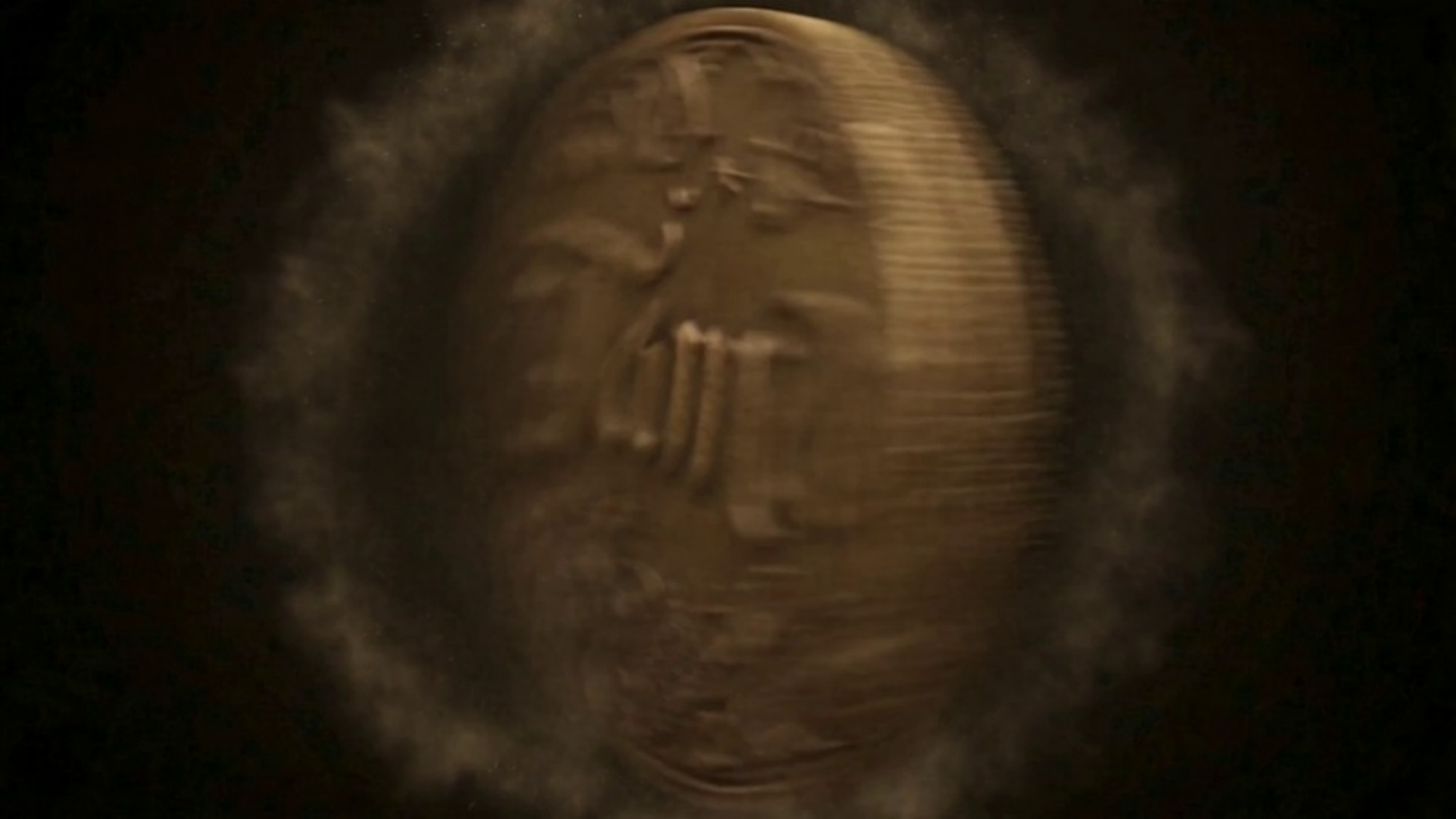}
        & \appcaseframe{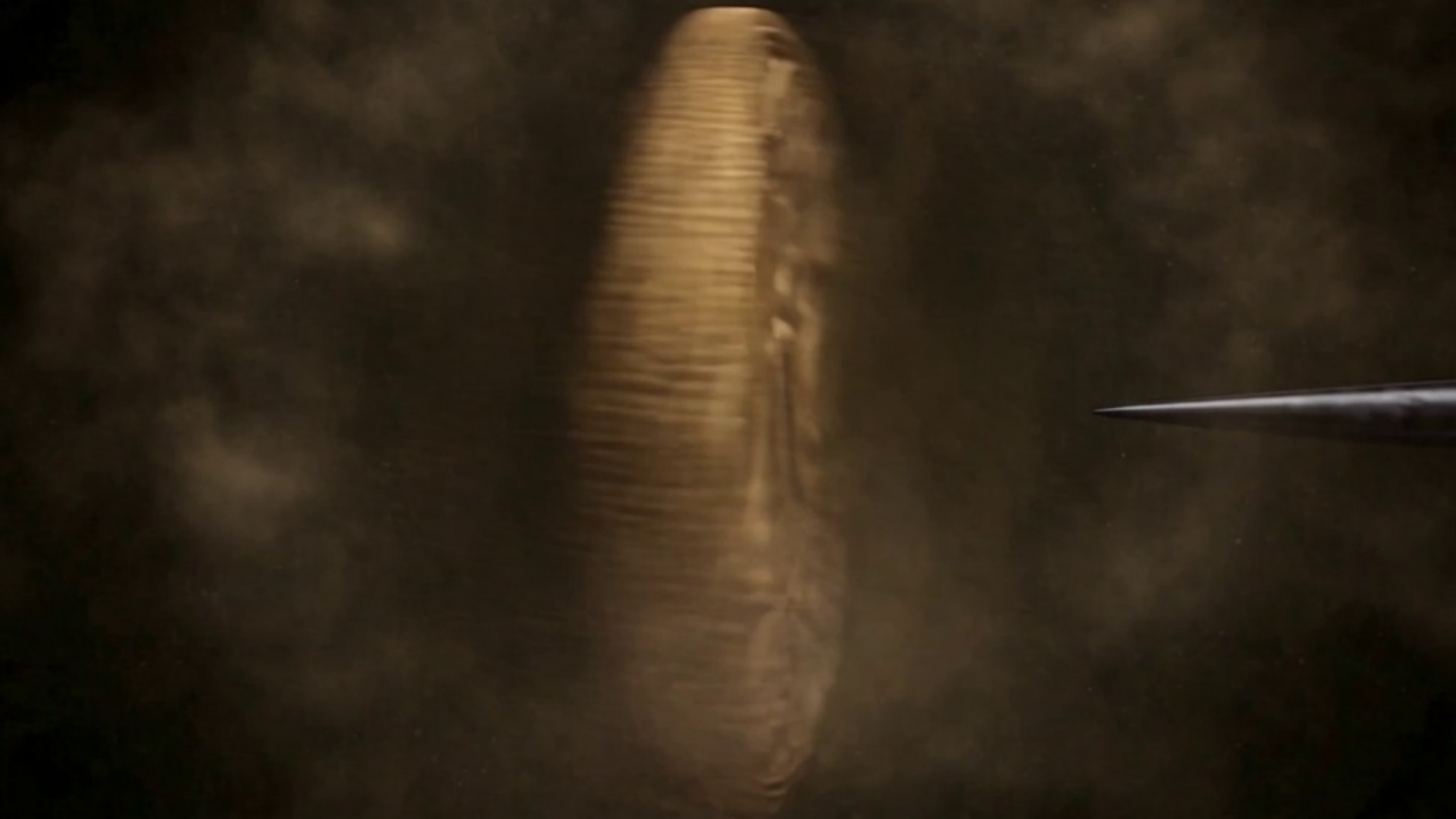}
        & \appcaseframe{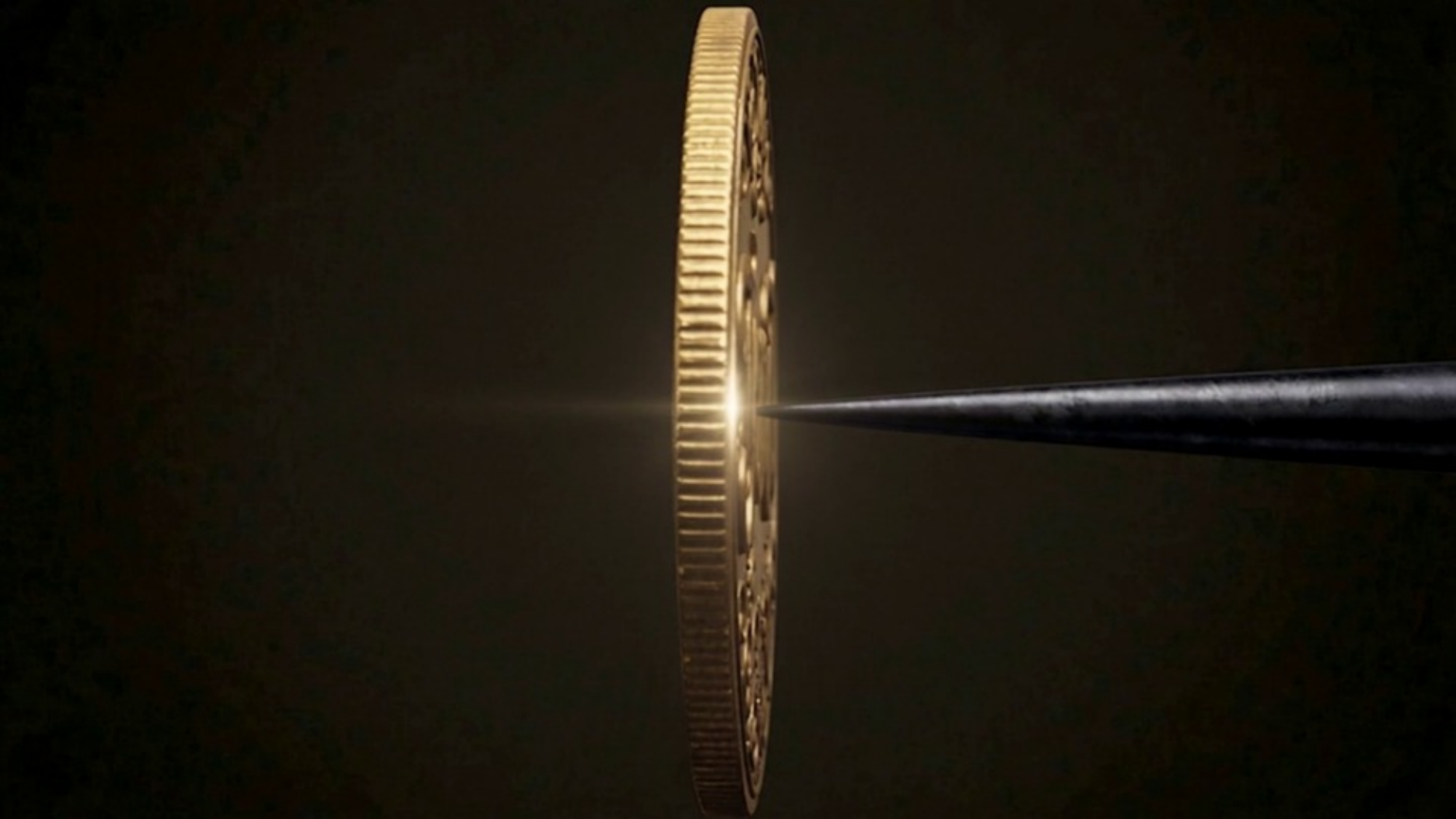}
        \end{tabular}
\par\vspace{3pt}\noindent\begin{tabular}{@{}c@{\hspace{\appGap}}c@{\hspace{\appGap}}c@{\hspace{\appGap}}c@{}}

        \appcaselabel{NeurCo (T)}{\textbf{3/6 passed}}{\textcolor{appFail}{Fails (2), (3), (5)}}
        & \appcaseframe{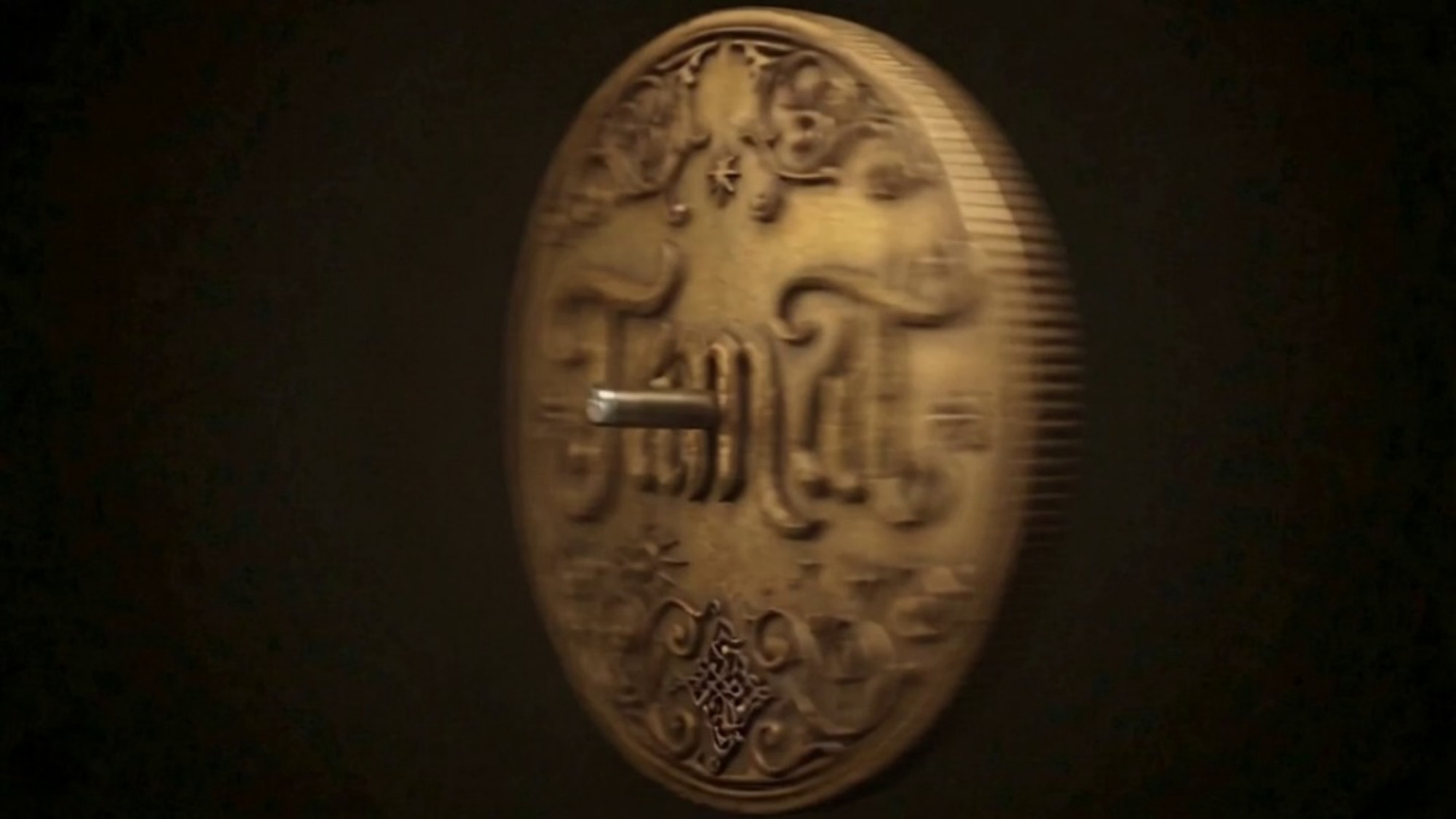}
        & \appcaseframe{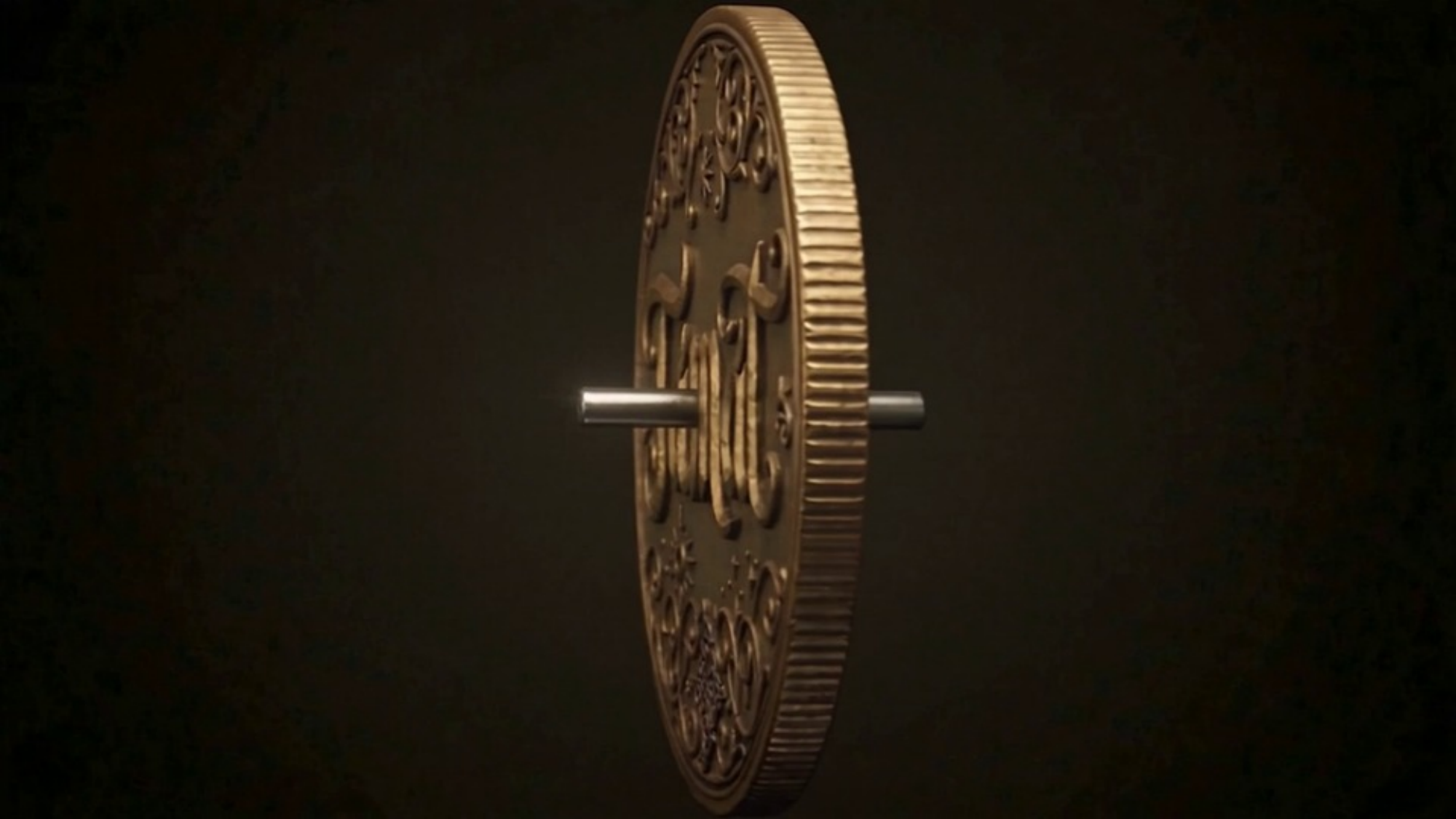}
        & \appcaseframe{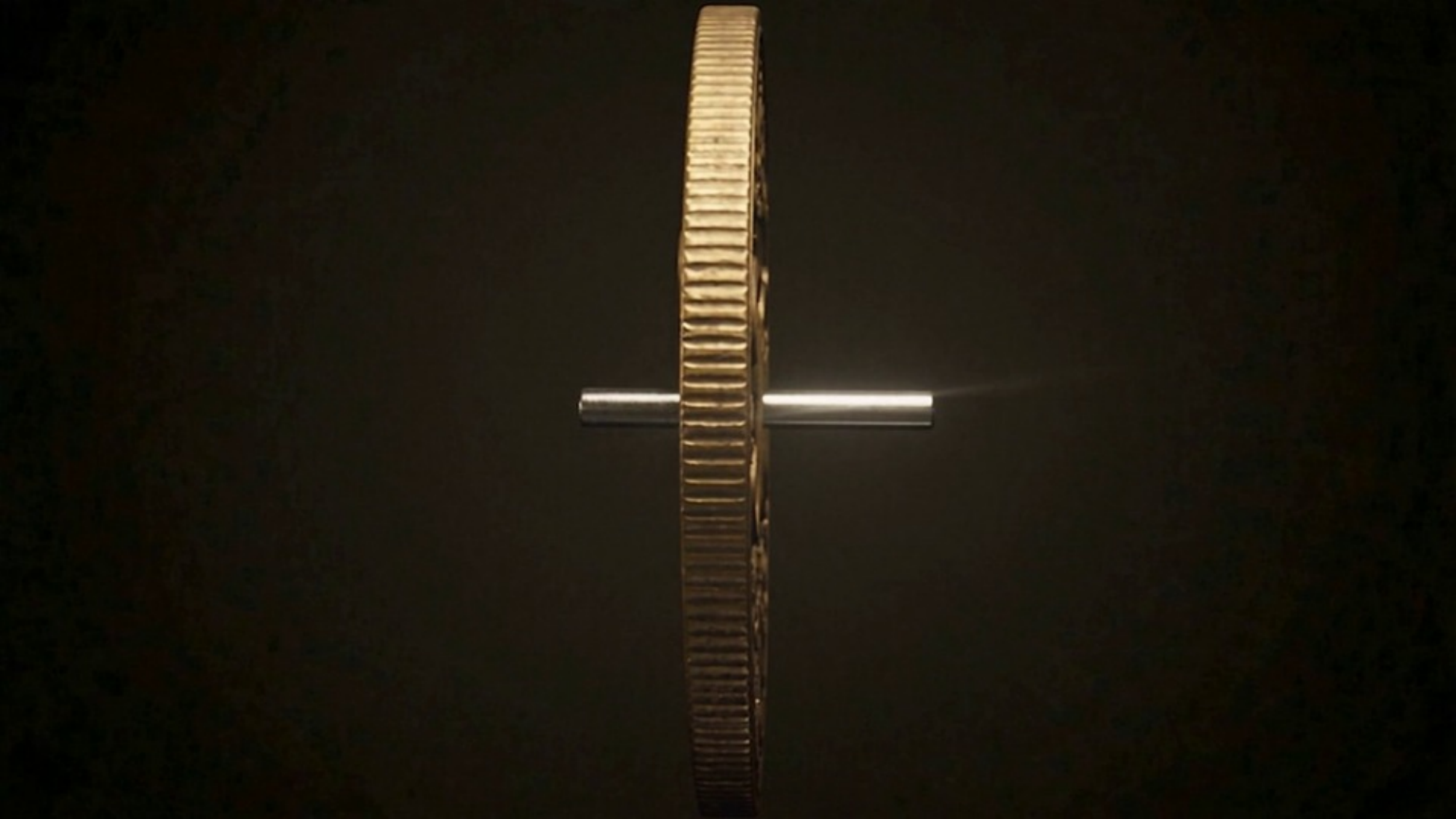}
        \end{tabular}
\par\vspace{3pt}\noindent\begin{tabular}{@{}c@{\hspace{\appGap}}c@{\hspace{\appGap}}c@{\hspace{\appGap}}c@{}}

        \appcaselabel{NeurCo (T+O)}{\textbf{3/6 passed}}{Text only}
        & \appcaseframe{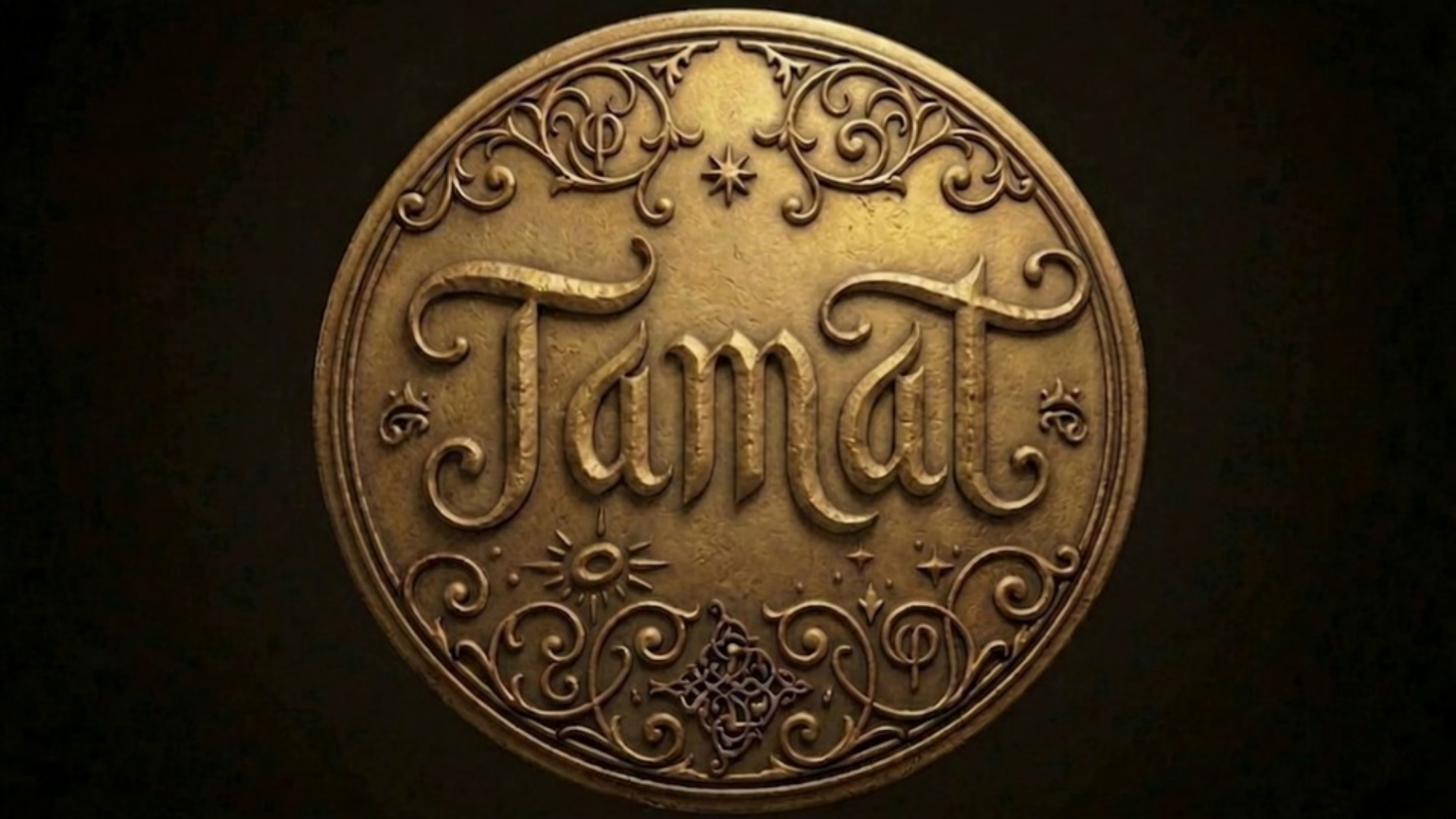}
        & \appcaseframe{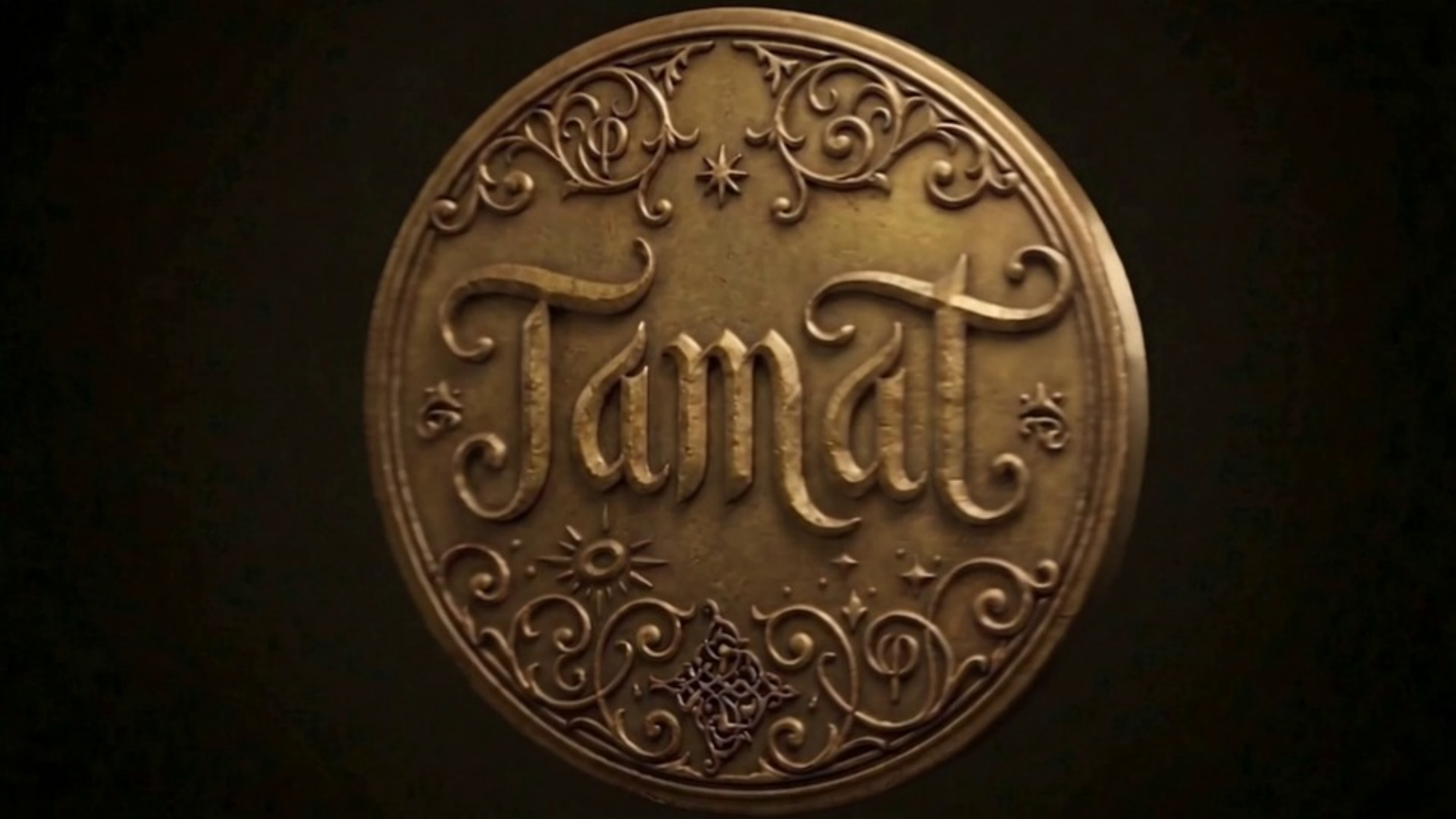}
        & \appcaseframe{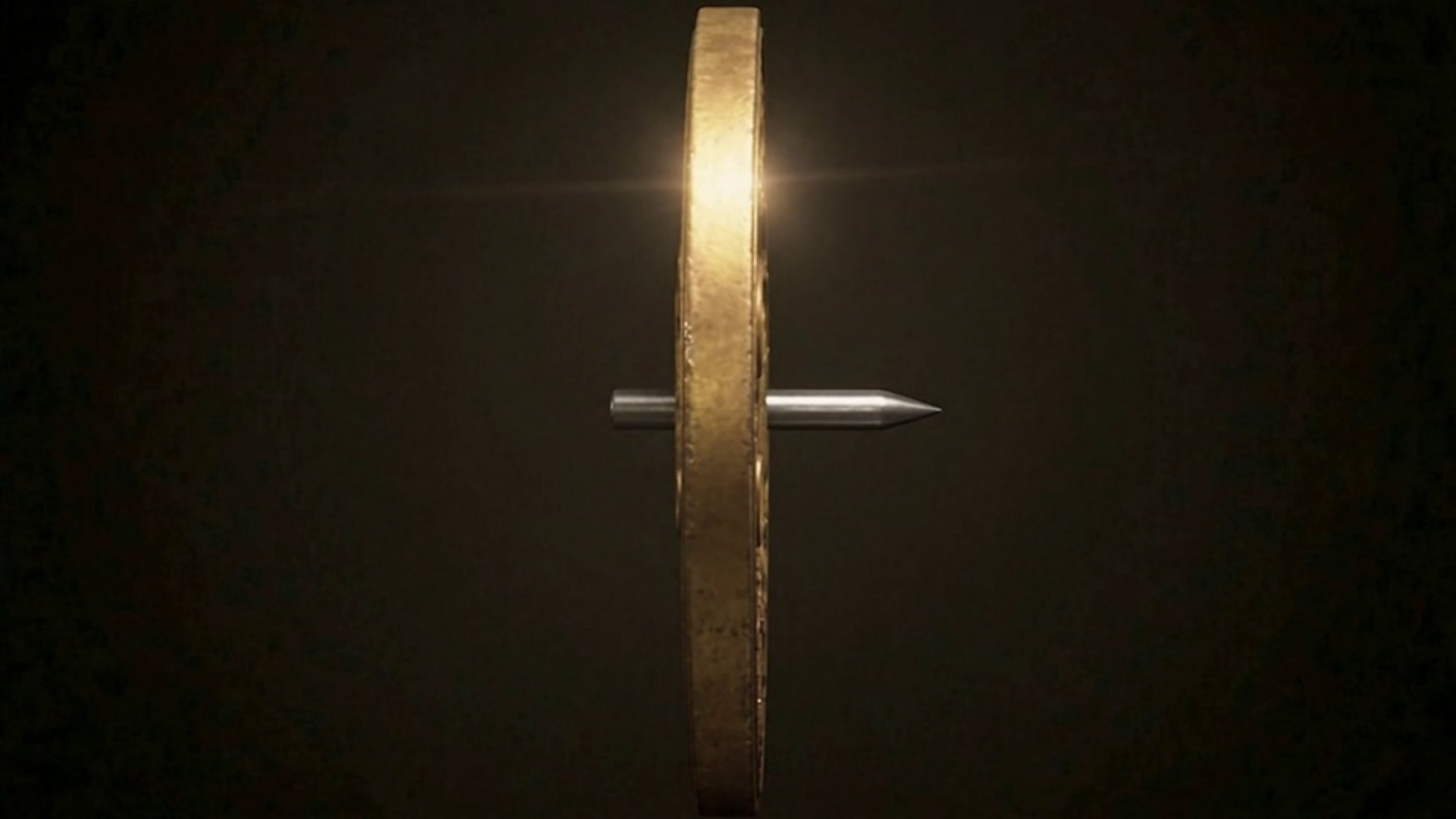}
    \end{tabular}
    \nopagebreak[3]\caption{\textbf{Text rewriting disrupts an existing contact constraint.}
    Both compiled variants fail the stop-pin contact checks on Seedance~2.5,
    reducing satisfaction from $5/6$ to $3/6$ without external operators.}
    \label{fig:case_sd_coin}
    \endgroup
\par\endgroup
\end{appcaseblock}

\clearpage
\begin{appcaseblock}
\noindent\textbf{Temporal distortion under first-frame conditioning.}
A constructed first frame organizes the starting scene but disrupts the
fermentation sequence on Seedance~2.5
(Figure~\ref{fig:case_sd_tubes}). The text-only variant records a
per-video Sci-VBench average of 3.91, compared with 2.21 for T+O and
2.35 for O. In the conditioned clip, the indicator changes color before
the intended progression.

\par\smallskip\noindent\begingroup\centering
    \centering
    \begingroup
    \appcaseinit
    \appcaseinputas{T2V}{Take three test tubes, and add 2 mL of unfermented milk, fermenting milk, and fermented
    milk to each, respectively. Add two drops of bromothymol blue indicator to each tube.}{}{\appinputimg{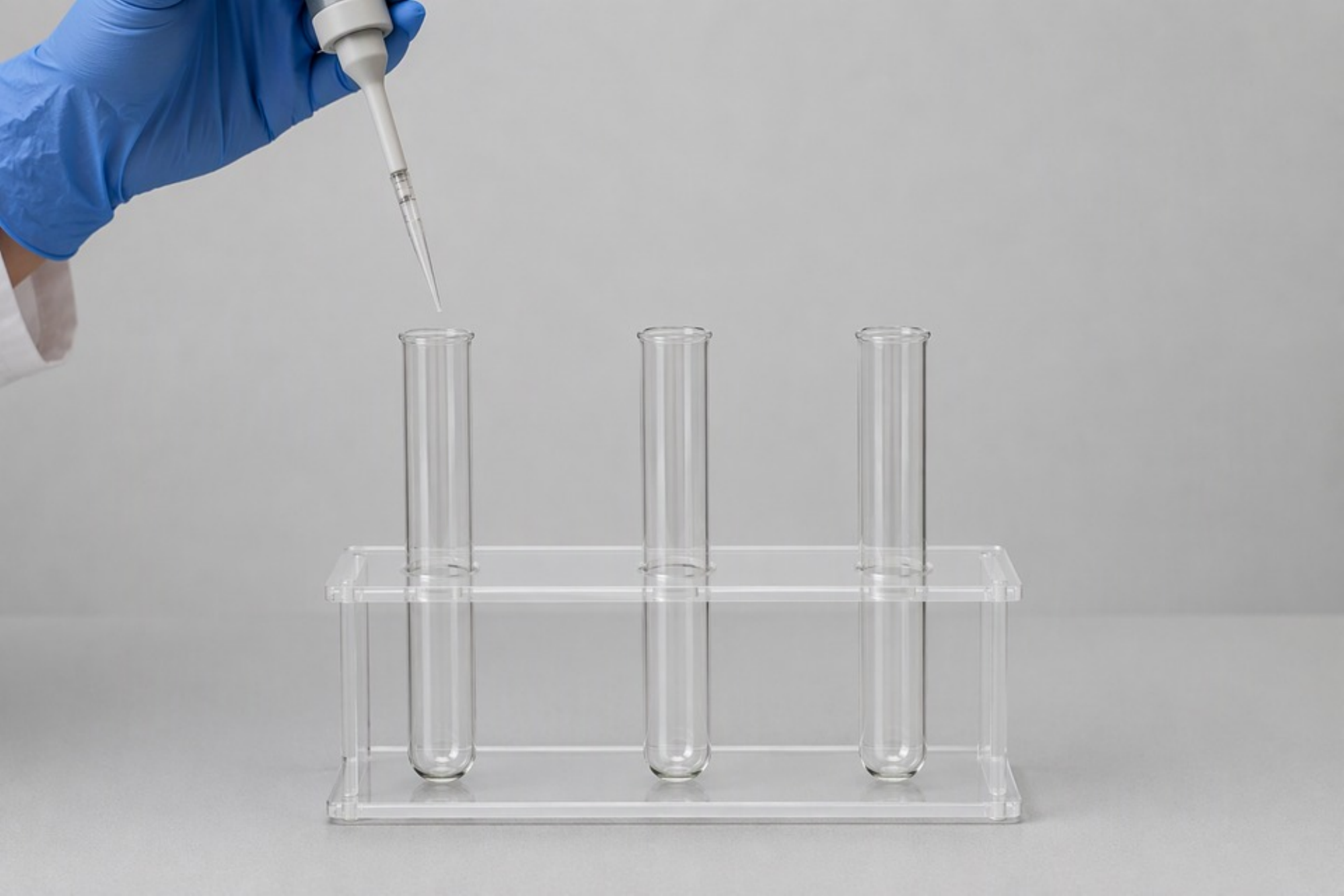}{First frame (O)}%
        \appinputimg{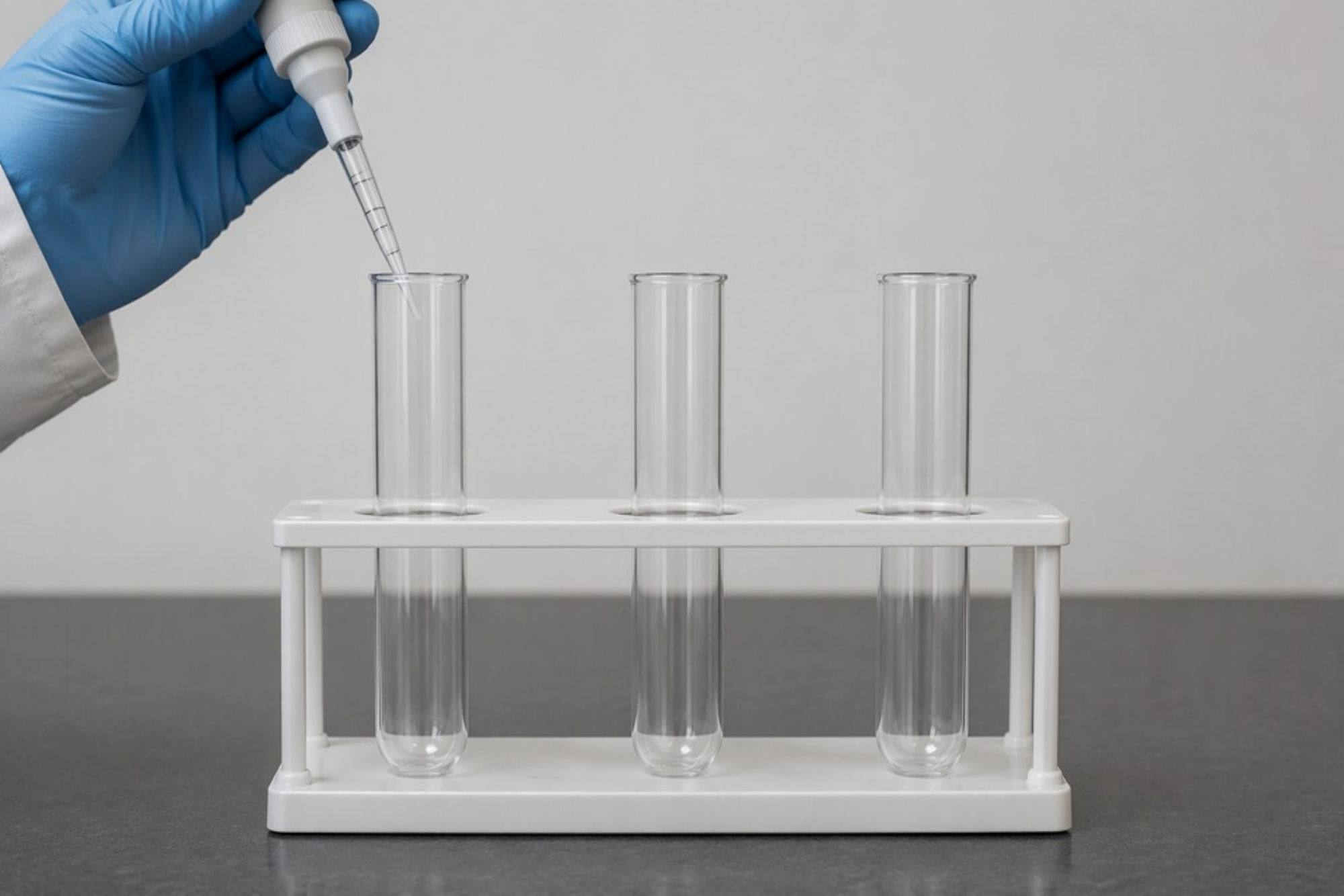}{First frame (T+O)}%
    }{}
    \begin{tabular}{@{}c@{\hspace{\appGap}}c@{\hspace{\appGap}}c@{\hspace{\appGap}}c@{}}
        \parbox[c]{\appLabelW}{\centering\textbf{Method}} & \textbf{Early} & \textbf{Middle} & \textbf{Late} \\[2pt]
        \appcaselabel{TIDE}{Avg.\ 2.58}{T2V}
        & \appcaseframe{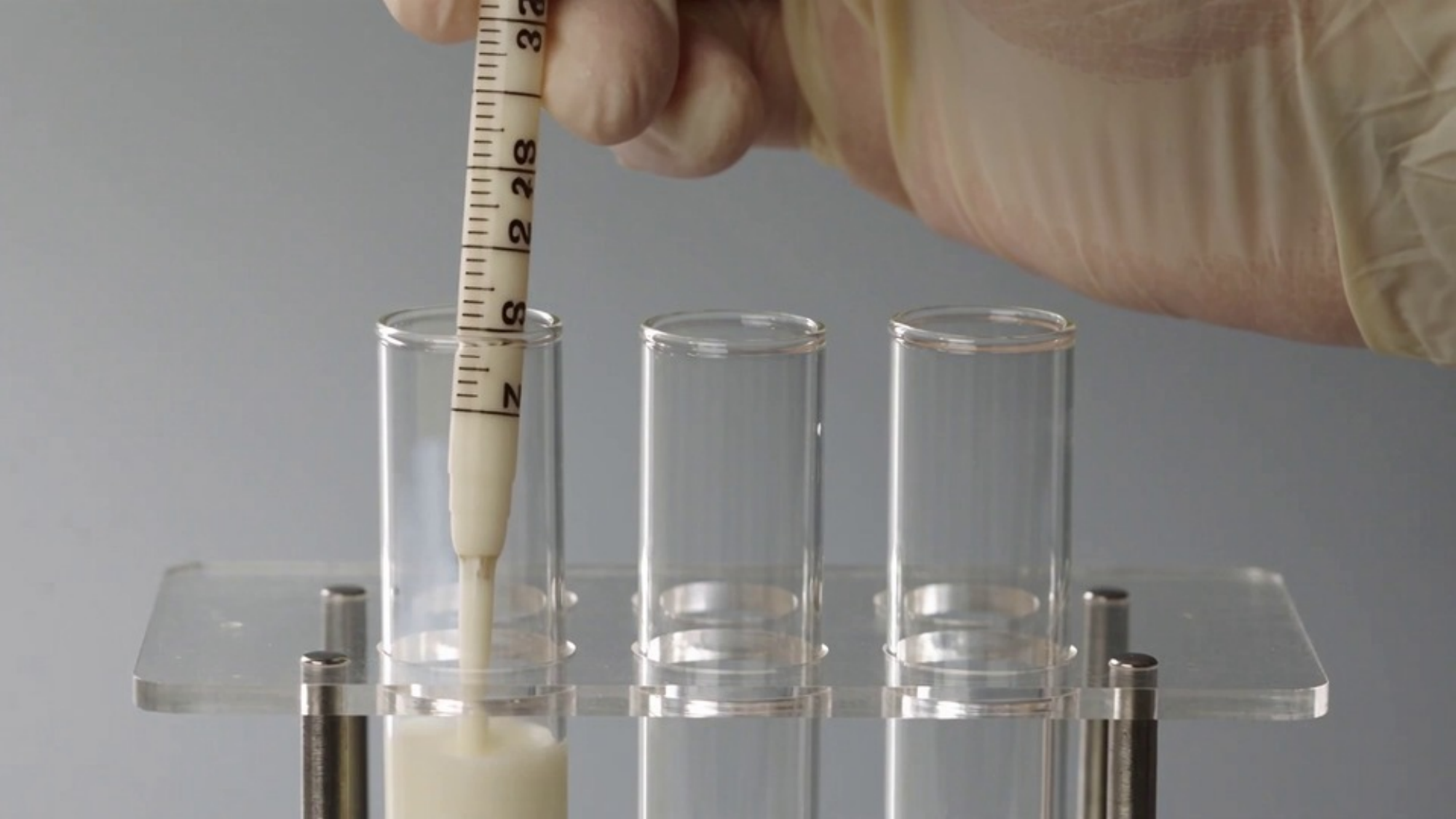}
        & \appcaseframe{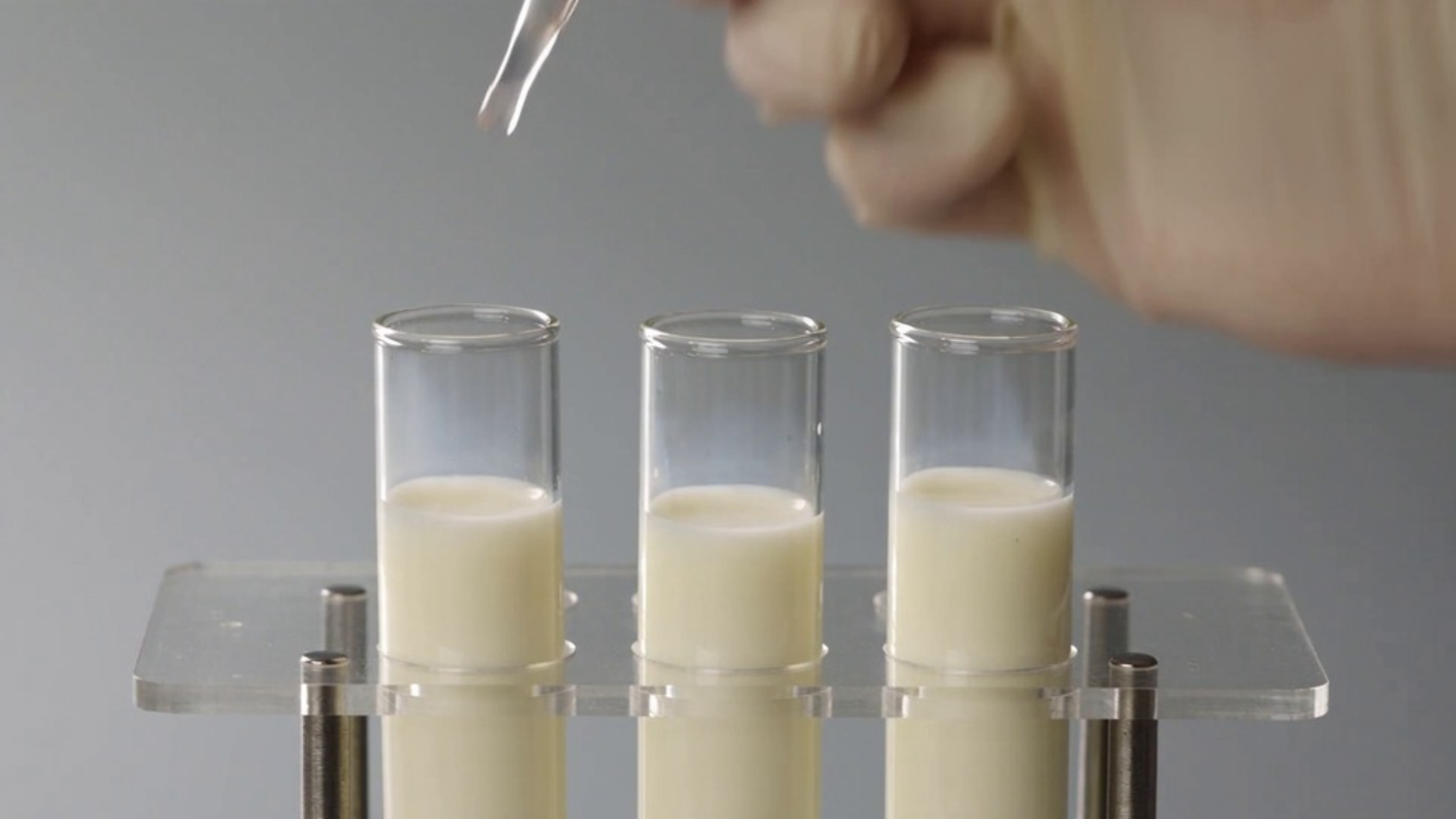}
        & \appcaseframe{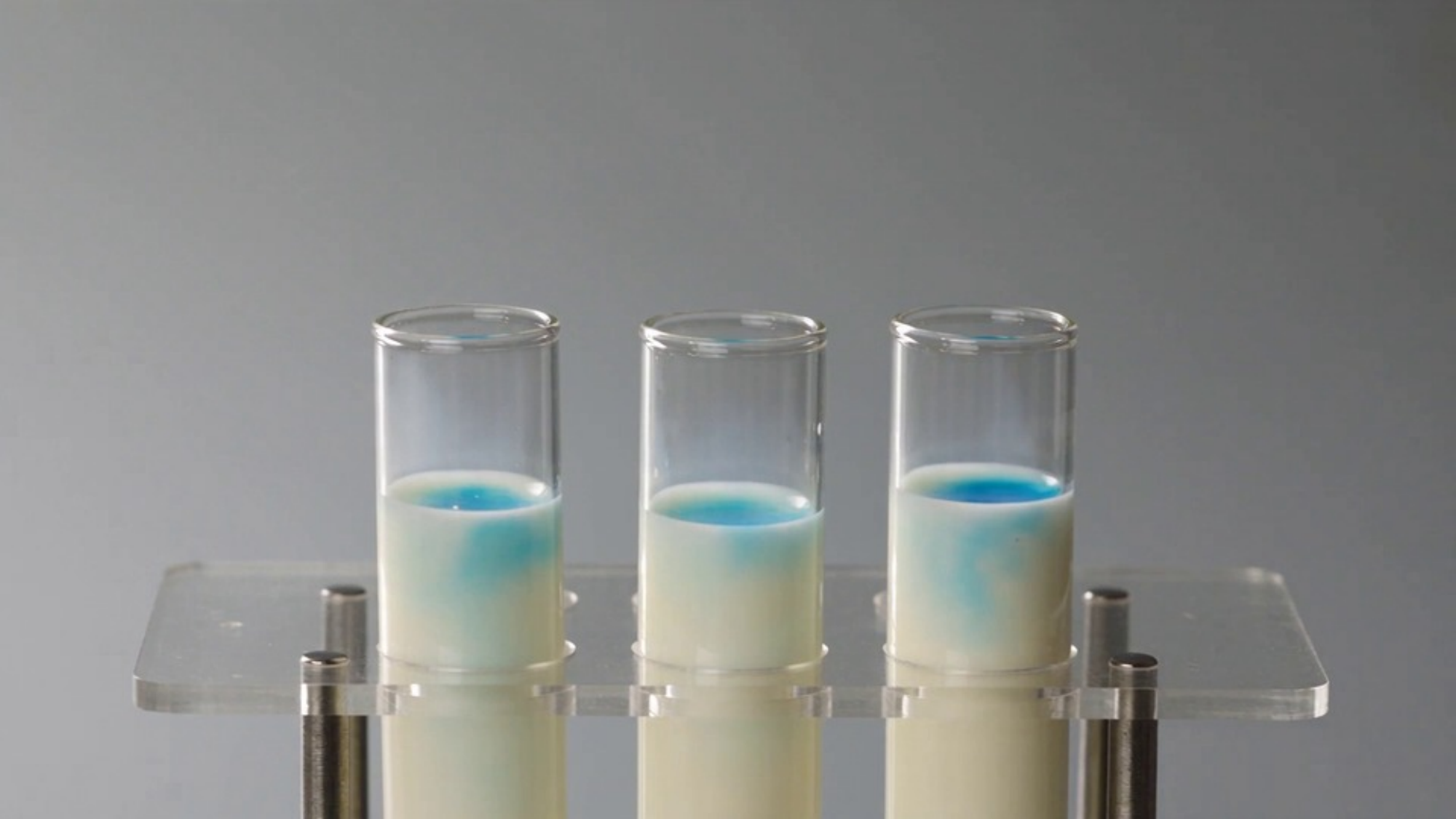}
        \end{tabular}
\par\vspace{3pt}\noindent\begin{tabular}{@{}c@{\hspace{\appGap}}c@{\hspace{\appGap}}c@{\hspace{\appGap}}c@{}}

        \appcaselabel{TIDE + NeurCo (T)}{\textcolor{appBlue}{\textbf{Avg.\ 3.91}}}{Text adaptation}
        & \appcaseframe{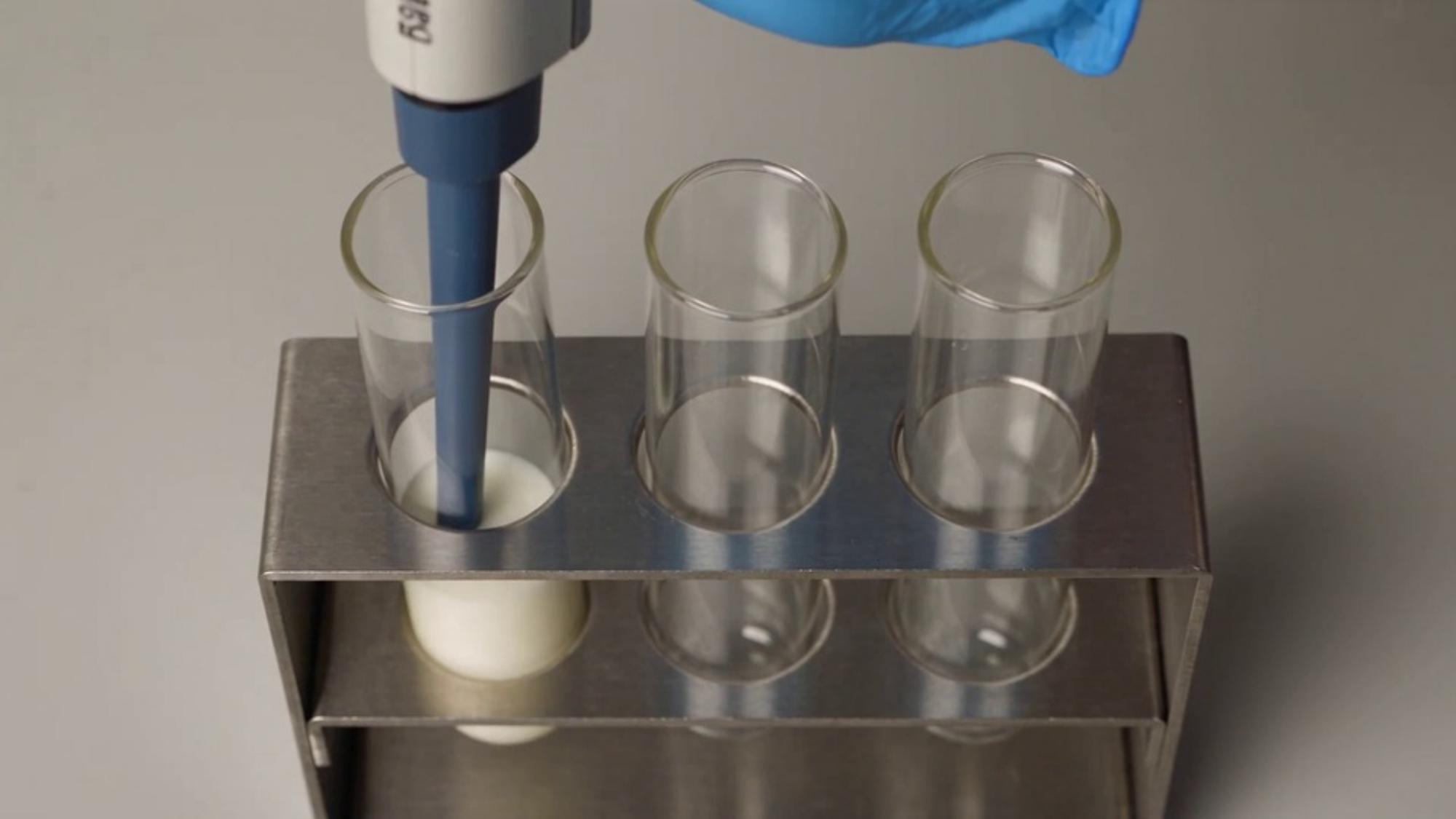}
        & \appcaseframe{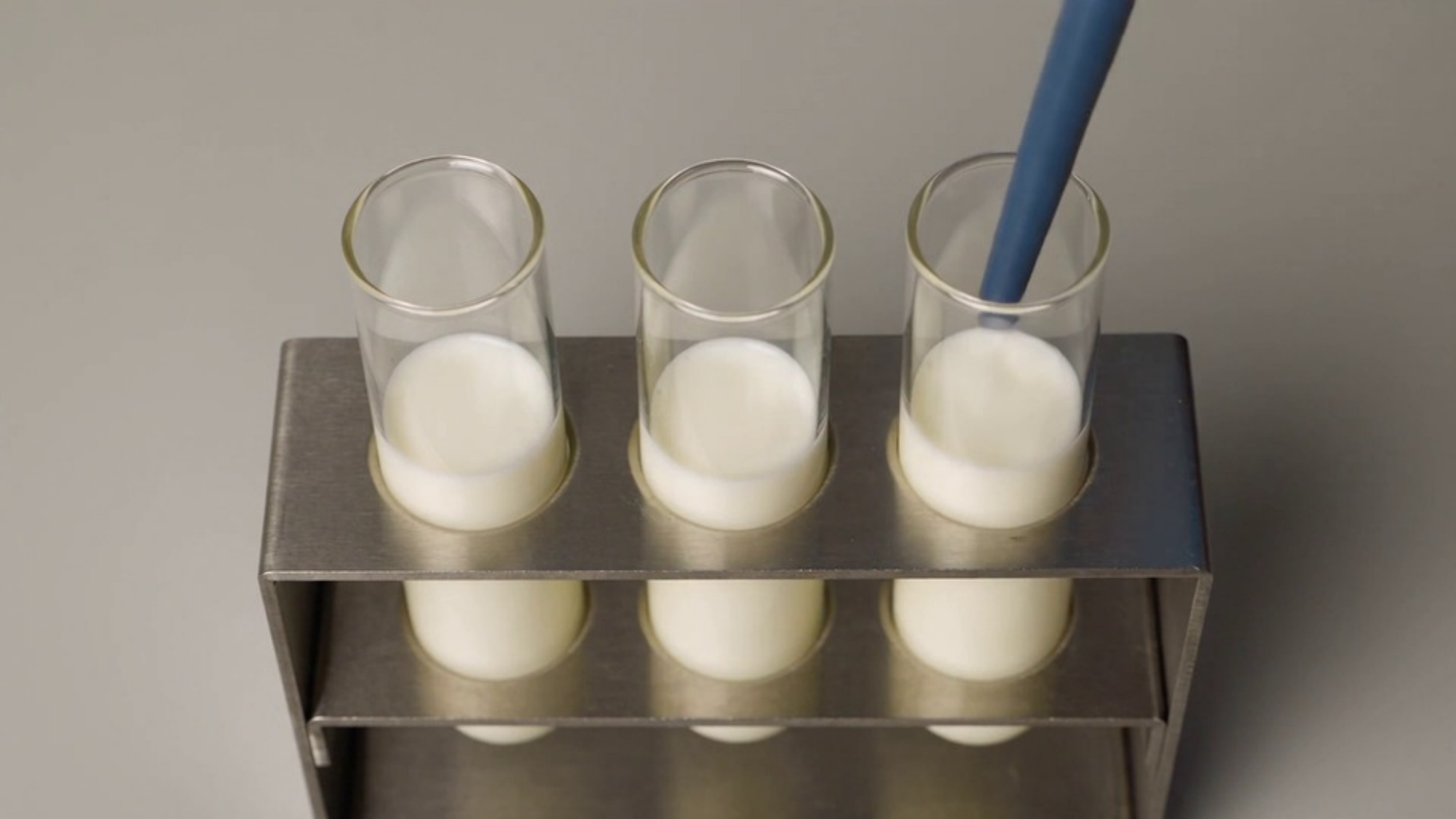}
        & \appcaseframe{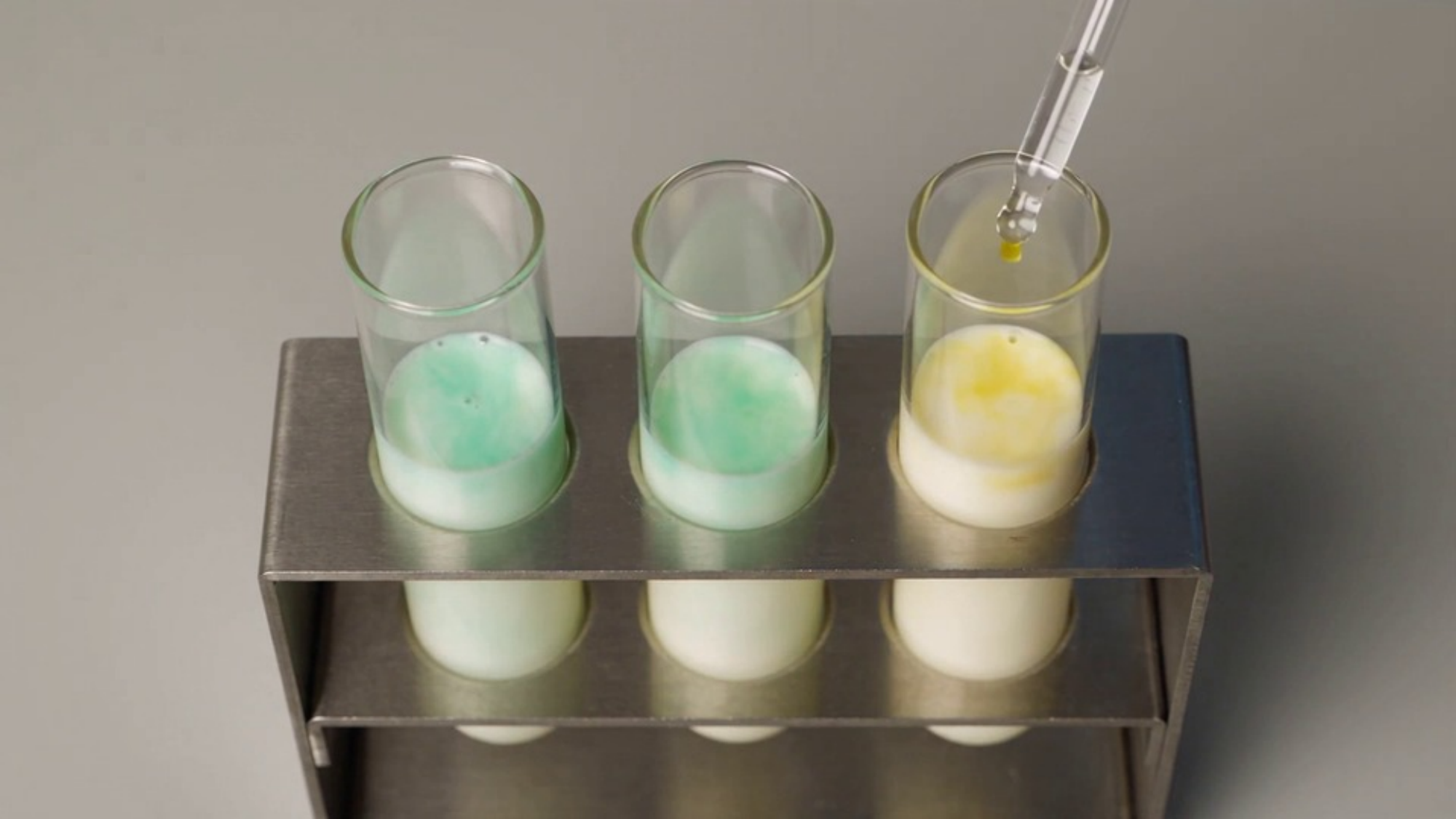}
        \end{tabular}
\par\vspace{3pt}\noindent\begin{tabular}{@{}c@{\hspace{\appGap}}c@{\hspace{\appGap}}c@{\hspace{\appGap}}c@{}}

        \appcaselabel{TIDE + NeurCo (T+O)}{Avg.\ 2.21}{First frame + text}
        & \appcaseframe{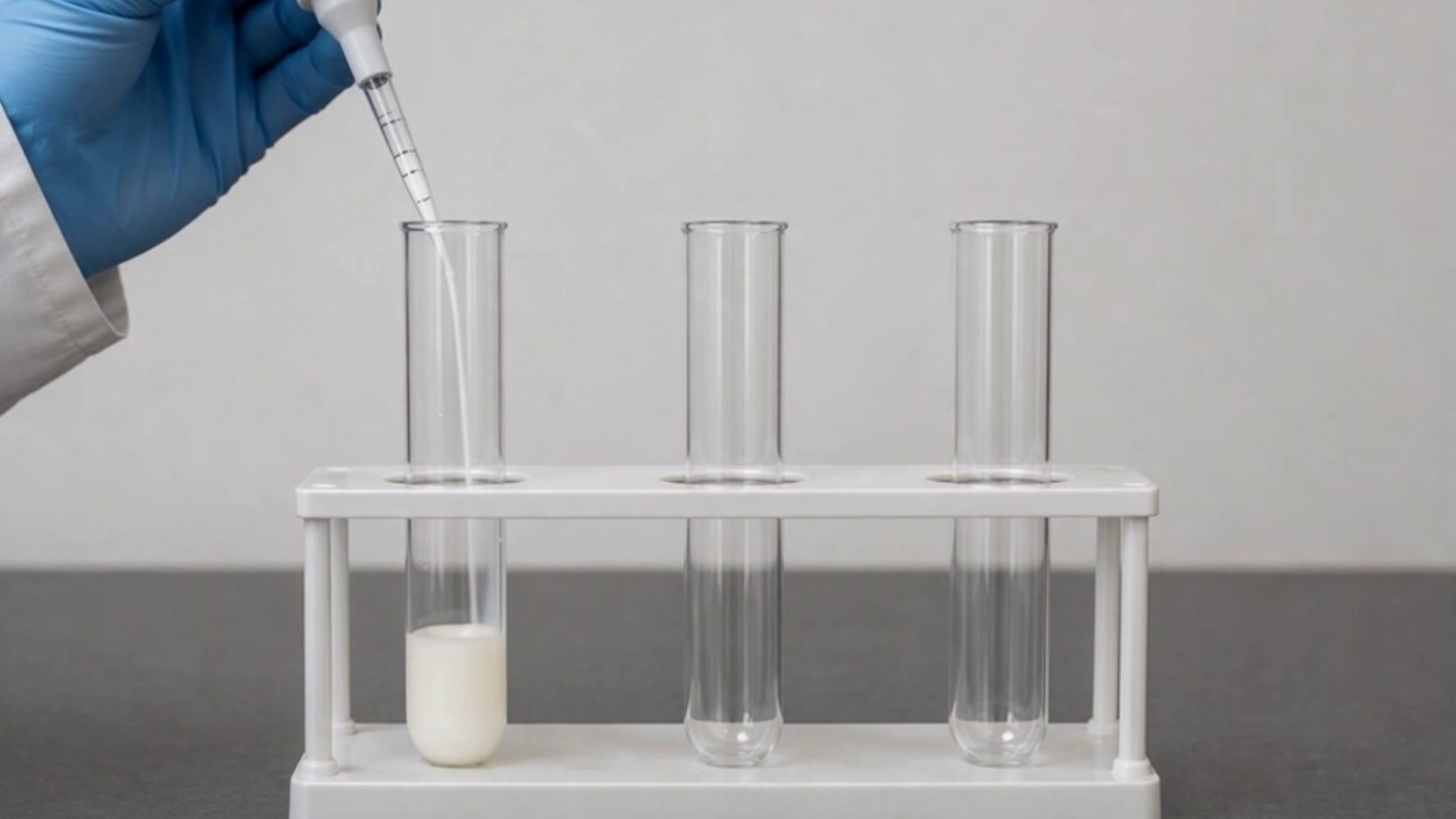}
        & \appcaseframe{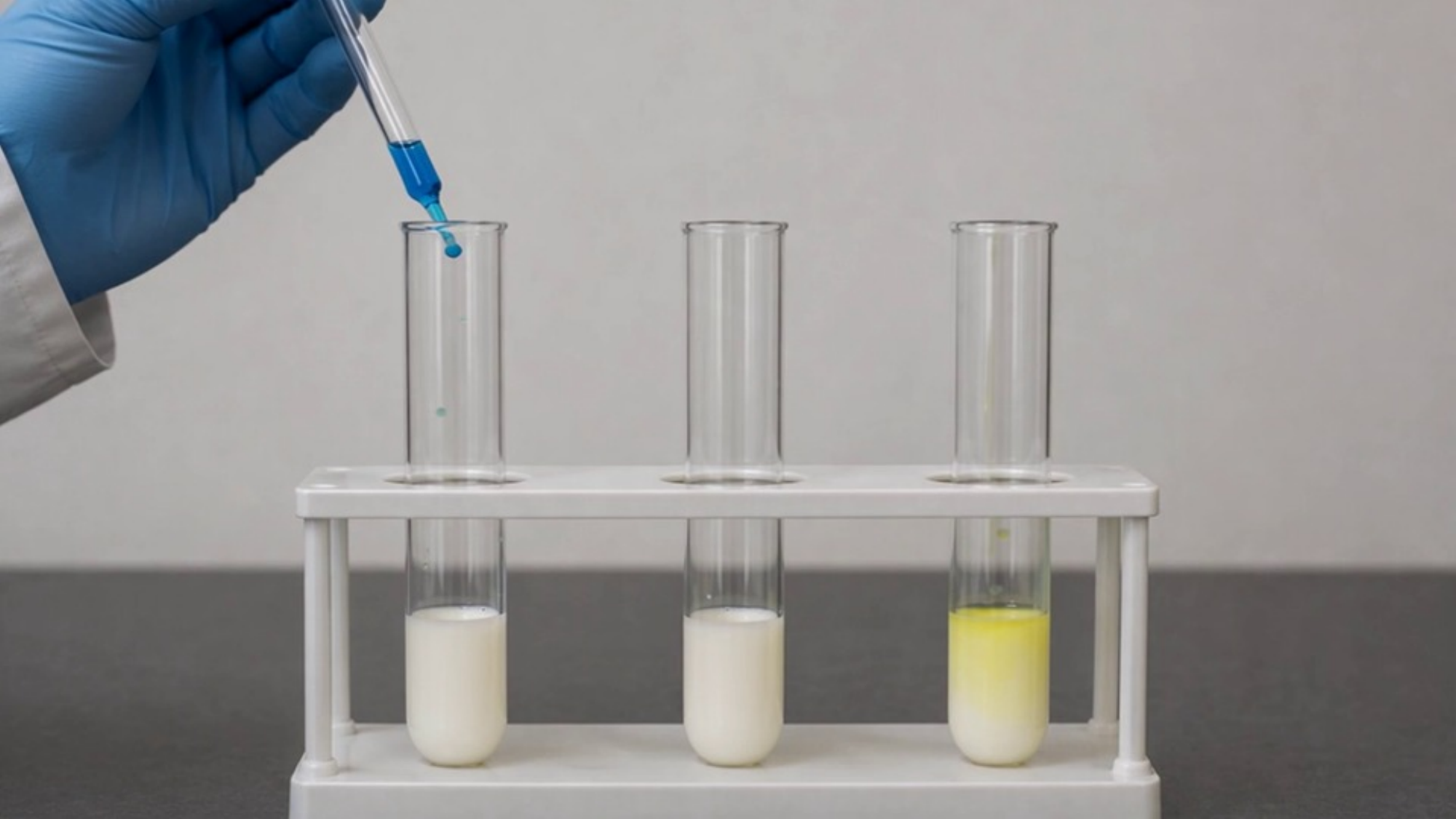}
        & \appcaseframe{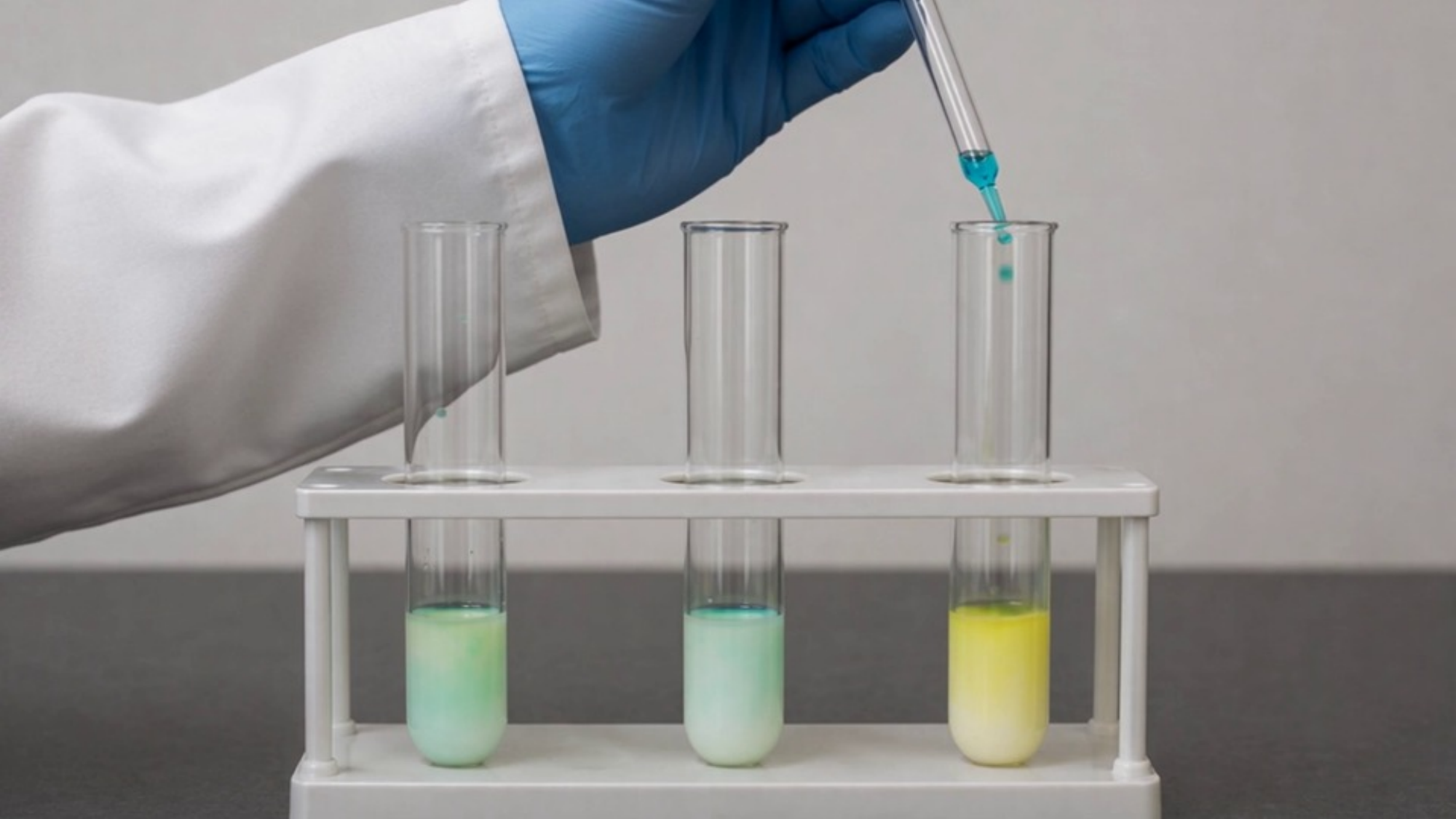}
        \end{tabular}
\par\vspace{3pt}\noindent\begin{tabular}{@{}c@{\hspace{\appGap}}c@{\hspace{\appGap}}c@{\hspace{\appGap}}c@{}}

        \appcaselabel{TIDE + NeurCo (O)}{Avg.\ 2.35}{First frame}
        & \appcaseframe{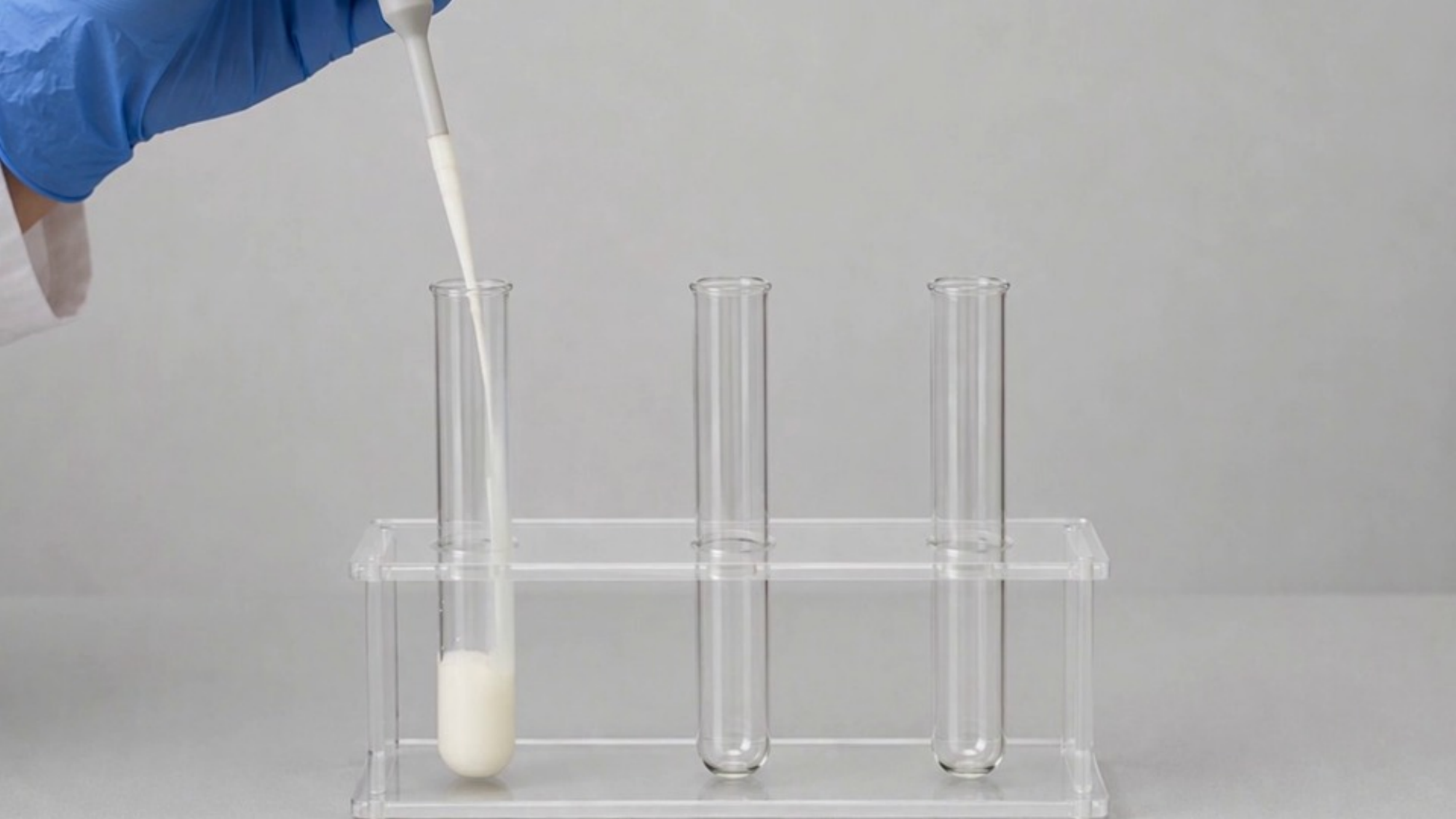}
        & \appcaseframe{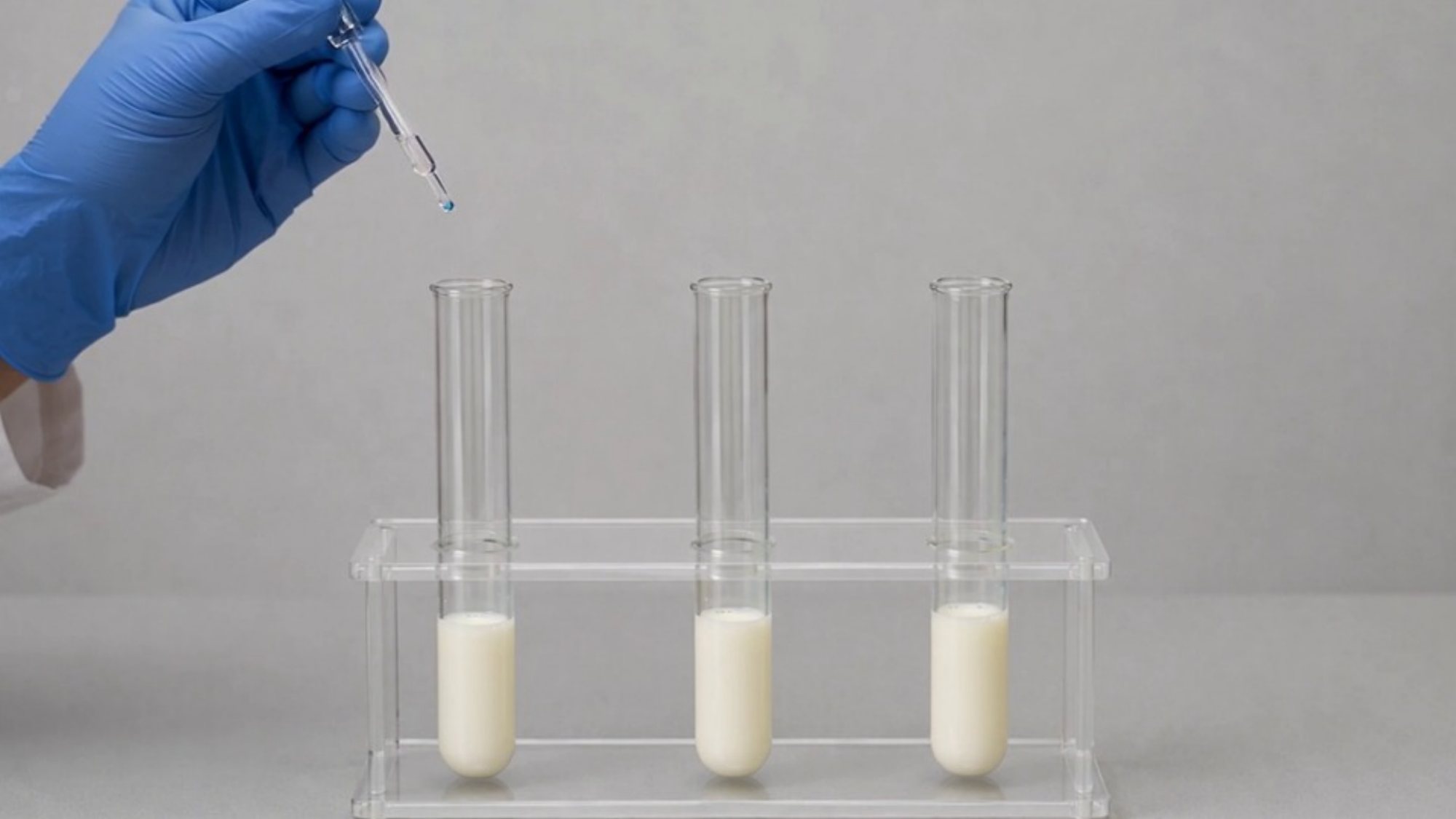}
        & \appcaseframe{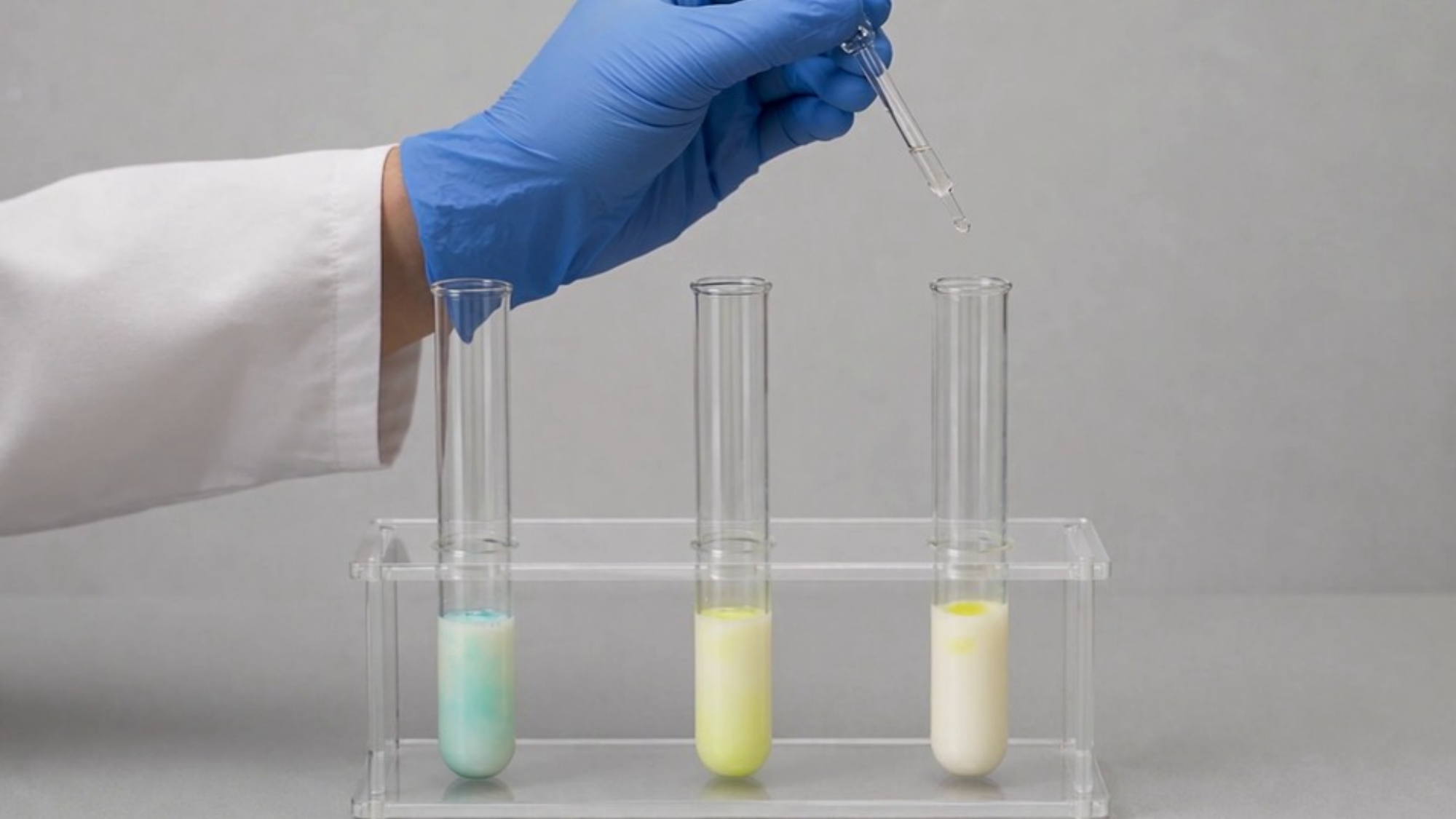}
    \end{tabular}
    \nopagebreak[3]\caption{\textbf{First-frame conditioning weakens procedure fidelity.}
    The TIDE + NeurCo (T) variant records the highest per-video average
    (3.91); scores average PG, SCC, SC, and VT as in Table~\ref{tab:sci_results}.}
    \label{fig:case_sd_tubes}
    \endgroup
\par\endgroup
\end{appcaseblock}

\clearpage
\begin{appcaseblock}
\noindent\textbf{Camera motion in a multi-event scene.}
The coastline example compares how Native, TIDE, and TIDE + NeurCo (O)
render a long camera and event sequence on H3 and Wan
(Figure~\ref{fig:case_coast}). The panels show differences in framing
and camera progression across the plans. Composition and motion-order
scores for the broader VBench-2.0 subset are reported separately in
Table~\ref{tab:vbench2_dimensions}.

\par\smallskip\noindent\begingroup\centering
    \centering
    \begingroup
    \appcaseinit
    \appcaseinput{The camera starts at a tranquil coastline, where the waves
    gently crash against the rocks, the salty sea breeze fills the air, and
    the atmosphere feels fresh and alive. As the camera moves forward, the
    surface of the water glistens with golden light from the setting sun,
    the sound of the waves becoming more distinct as the tide retreats,
    revealing a moist sandy shore. The scene shifts to the beach, where a
    few seagulls peck at the sand, occasionally flying up and breaking the
    stillness of the sky. The sun slowly sinks below the horizon, painting
    the sky in vibrant shades of orange and red, as the golden light on the
    water fades, and the waves begin to intensify. Finally, the camera pulls
    back, and the entire coastline fades into twilight, with the sea and sky
    blending together in a serene, solitary embrace.}{}
    \noindent{\small\bfseries Native generations (context)}\par\vspace{3pt}
\begin{tabular}{@{}c@{\hspace{\appGap}}c@{\hspace{\appGap}}c@{\hspace{\appGap}}c@{}}
        \parbox[c]{\appLabelW}{\centering\textbf{Method}} & \textbf{Early} & \textbf{Middle} & \textbf{Late} \\[2pt]
        \appcaselabel{H3 Native}{T2V}{}
        & \appcaseframe{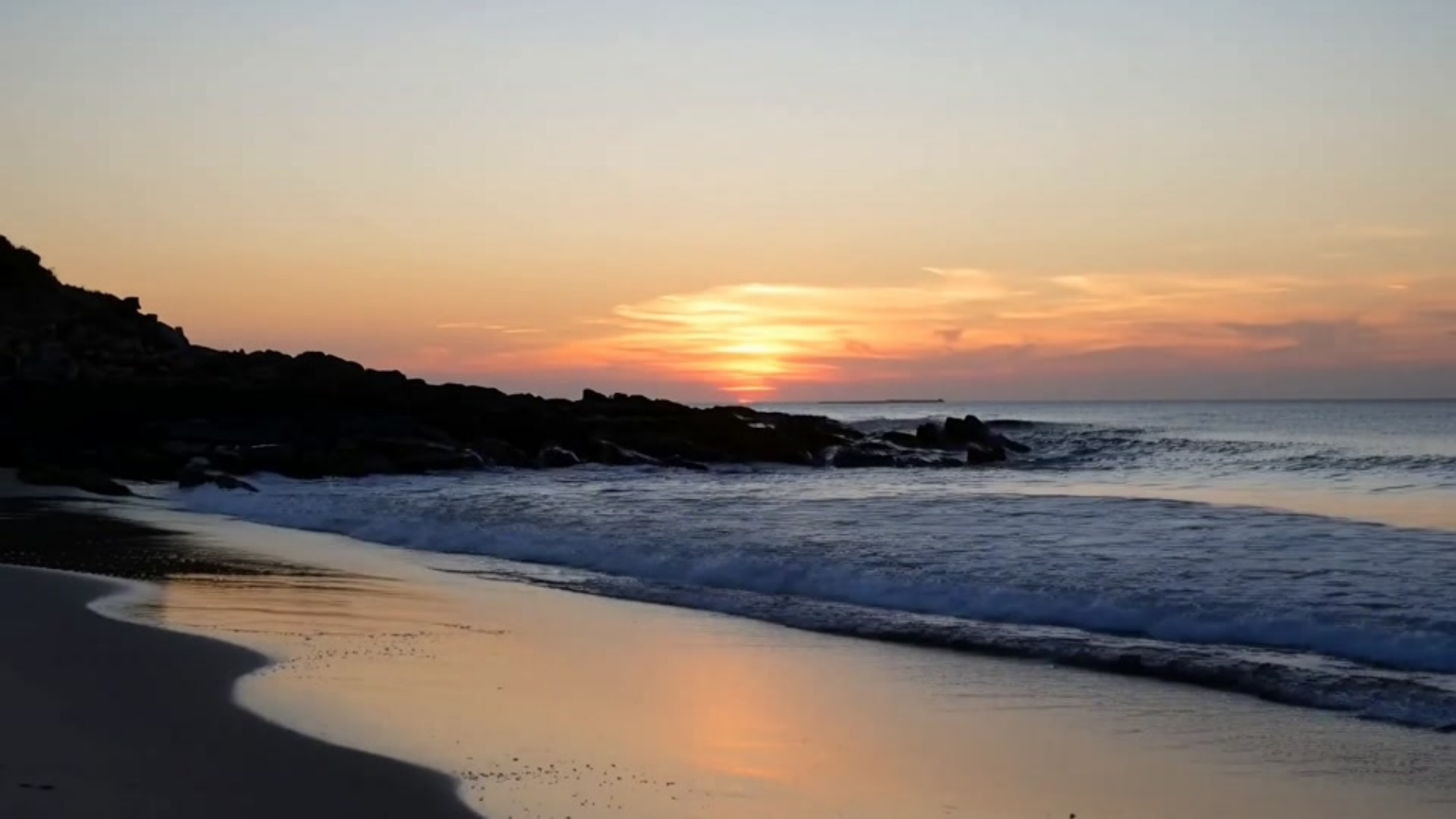}
        & \appcaseframe{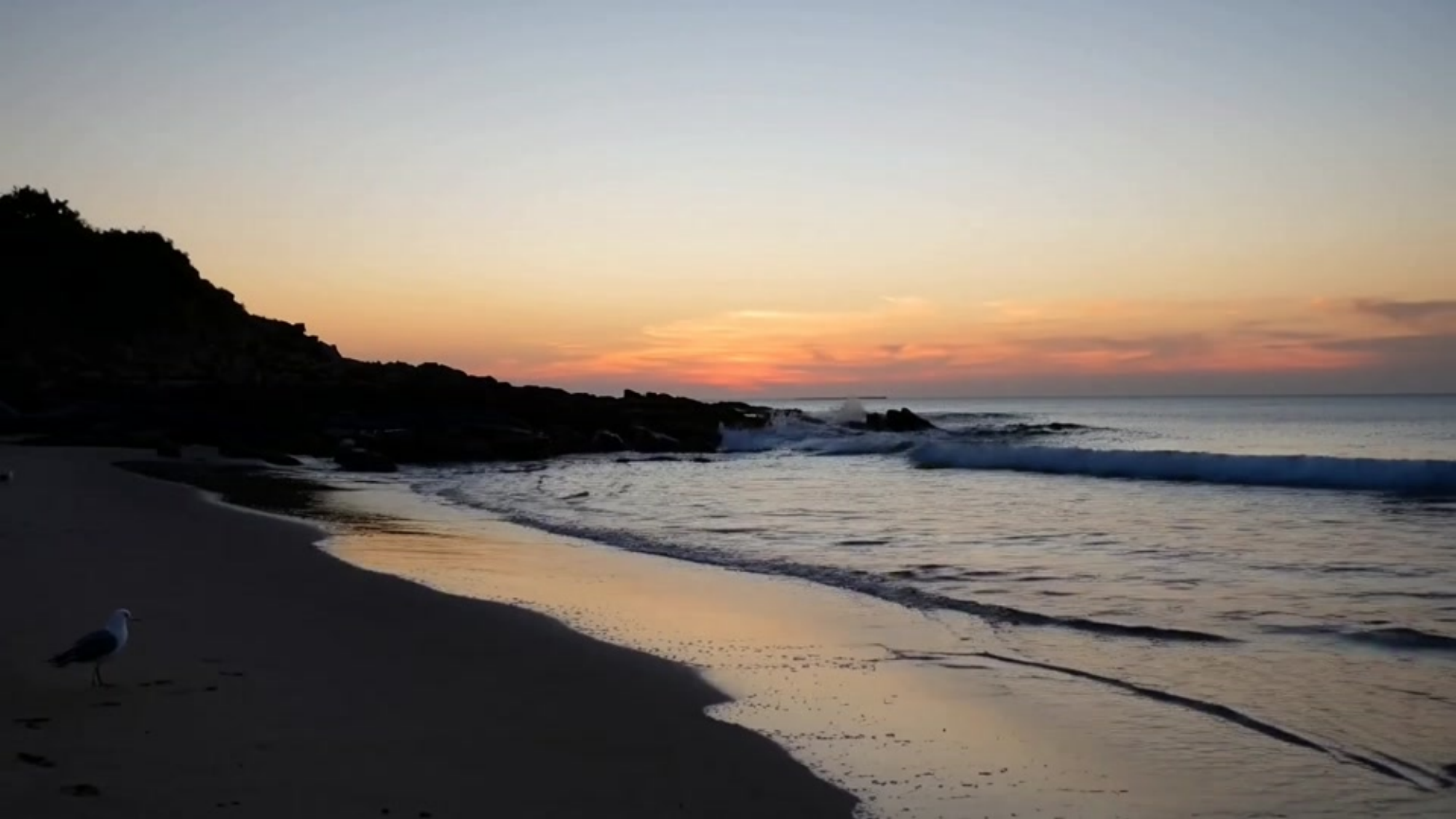}
        & \appcaseframe{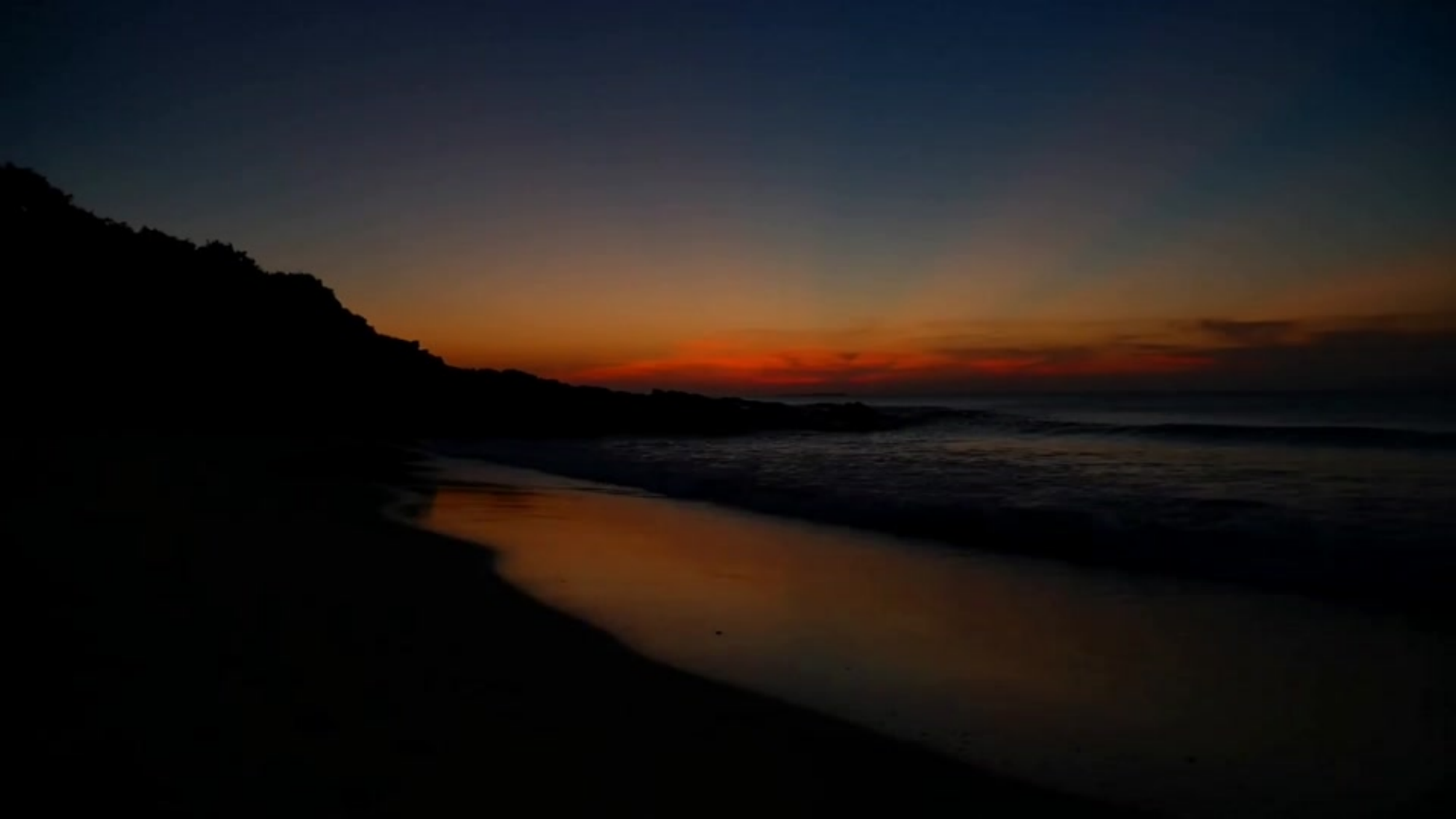}
        \end{tabular}
\par\vspace{3pt}\noindent\begin{tabular}{@{}c@{\hspace{\appGap}}c@{\hspace{\appGap}}c@{\hspace{\appGap}}c@{}}

        \appcaselabel{Wan Native}{T2V}{}
        & \appcaseframe{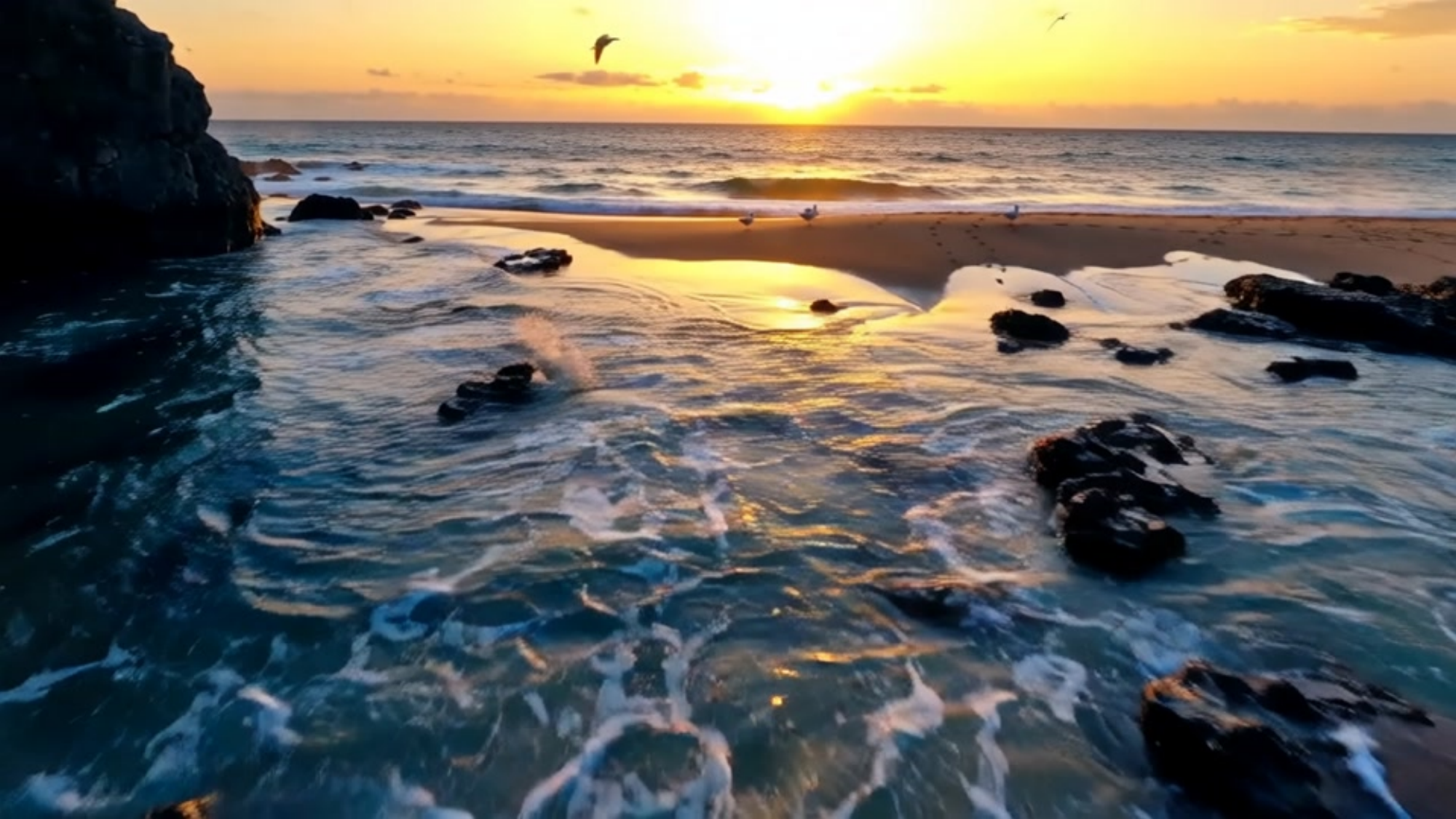}
        & \appcaseframe{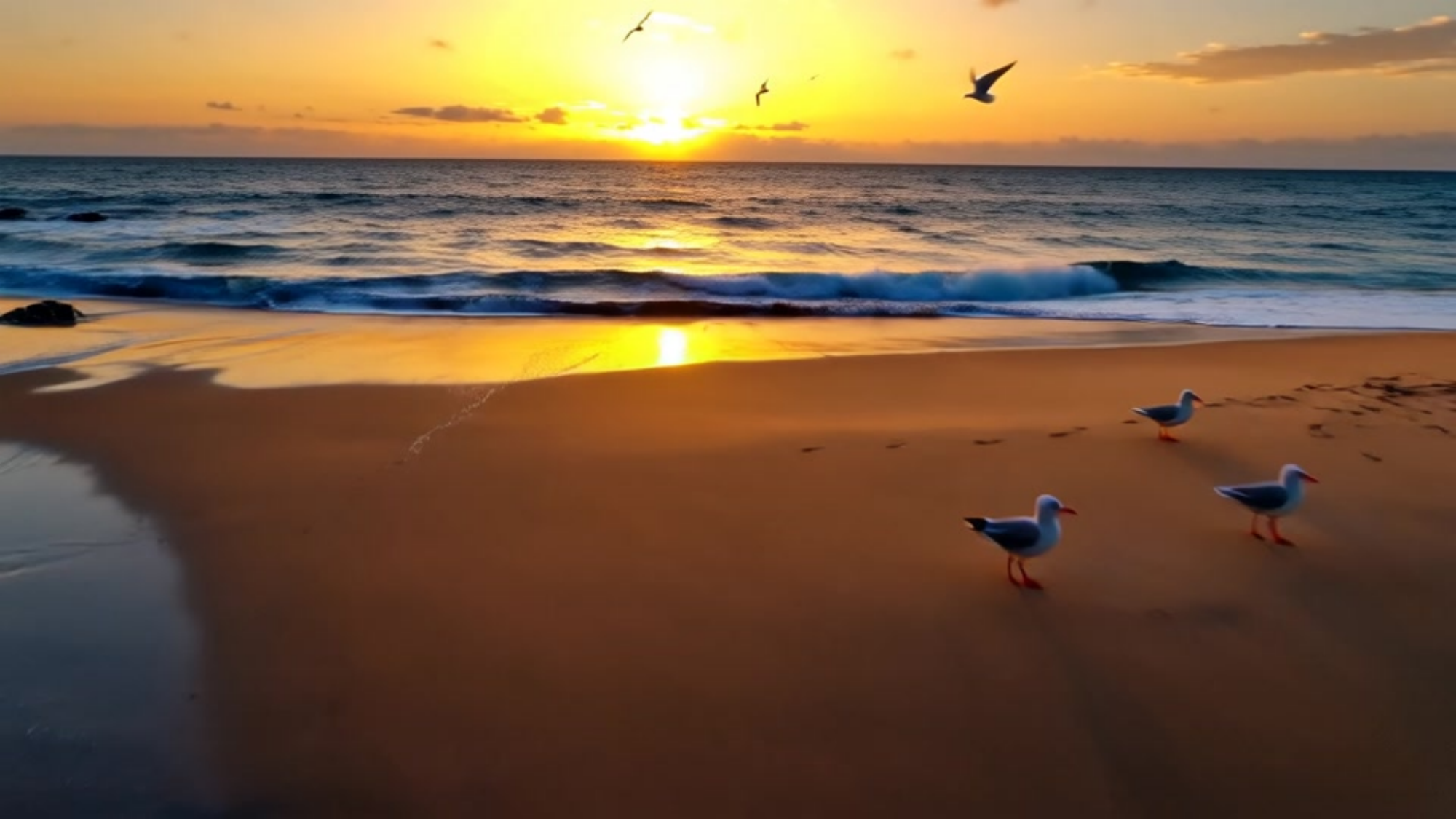}
        & \appcaseframe{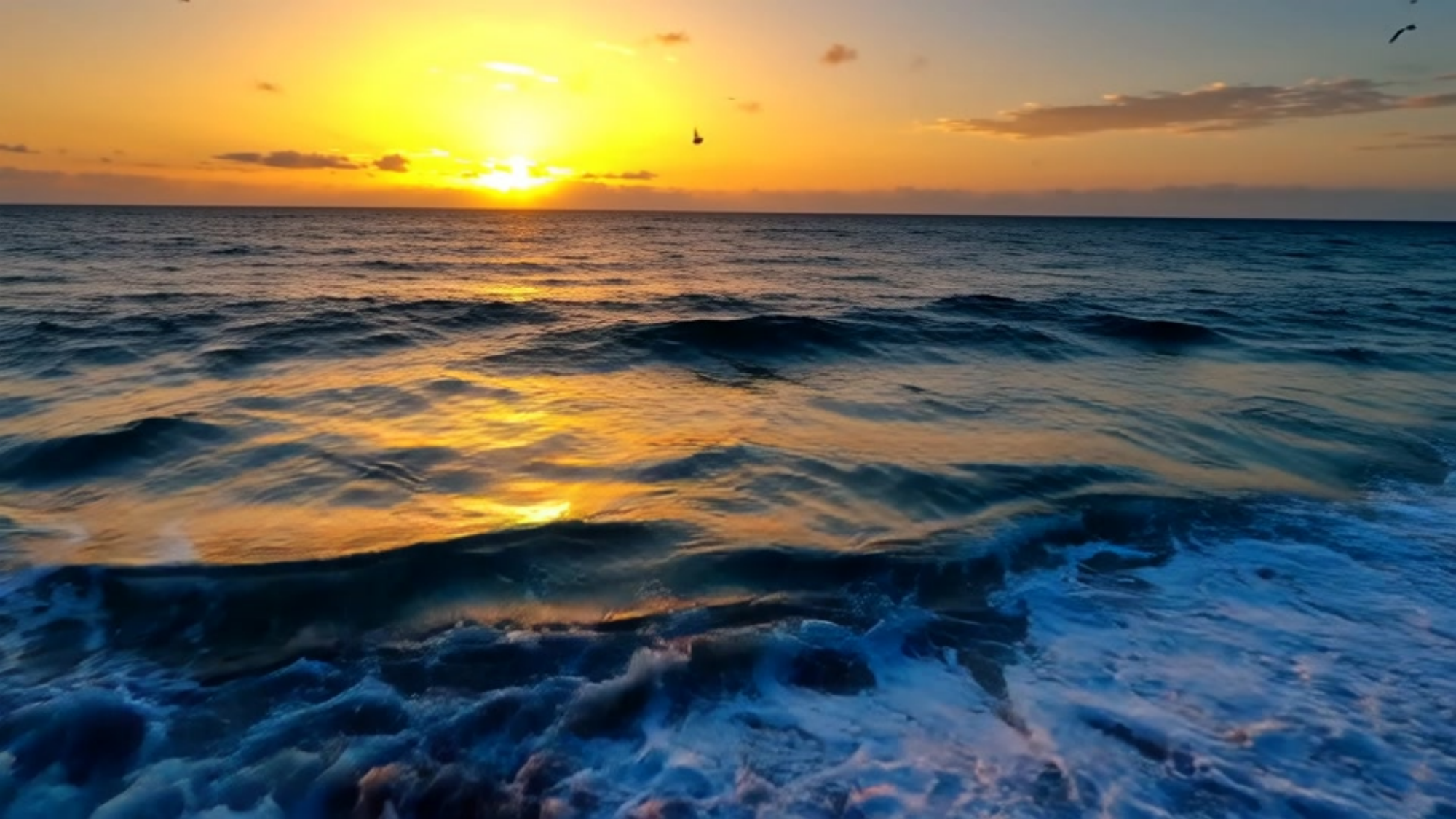}
    \end{tabular}
\par\vspace{5pt}\noindent{\small\bfseries TIDE comparisons}\par\vspace{3pt}
    \begin{tabular}{@{}c@{\hspace{\appGap}}c@{\hspace{\appGap}}c@{\hspace{\appGap}}c@{}}
        \parbox[c]{\appLabelW}{\centering\textbf{Method}} & \textbf{Early} & \textbf{Middle} & \textbf{Late} \\[2pt]
        \appcaselabel{H3 + TIDE}{T2V}{}
        & \appcaseframe{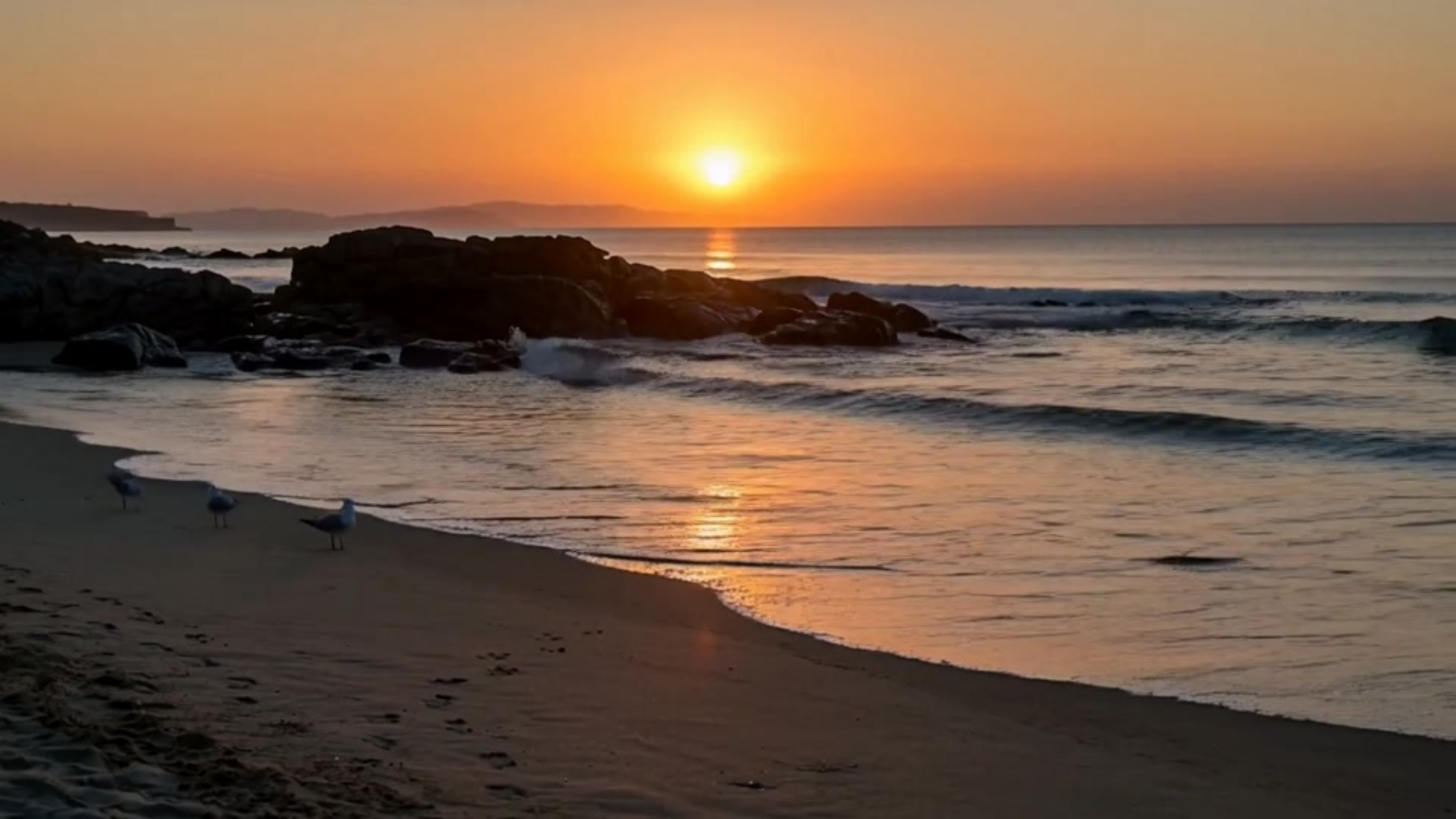}
        & \appcaseframe{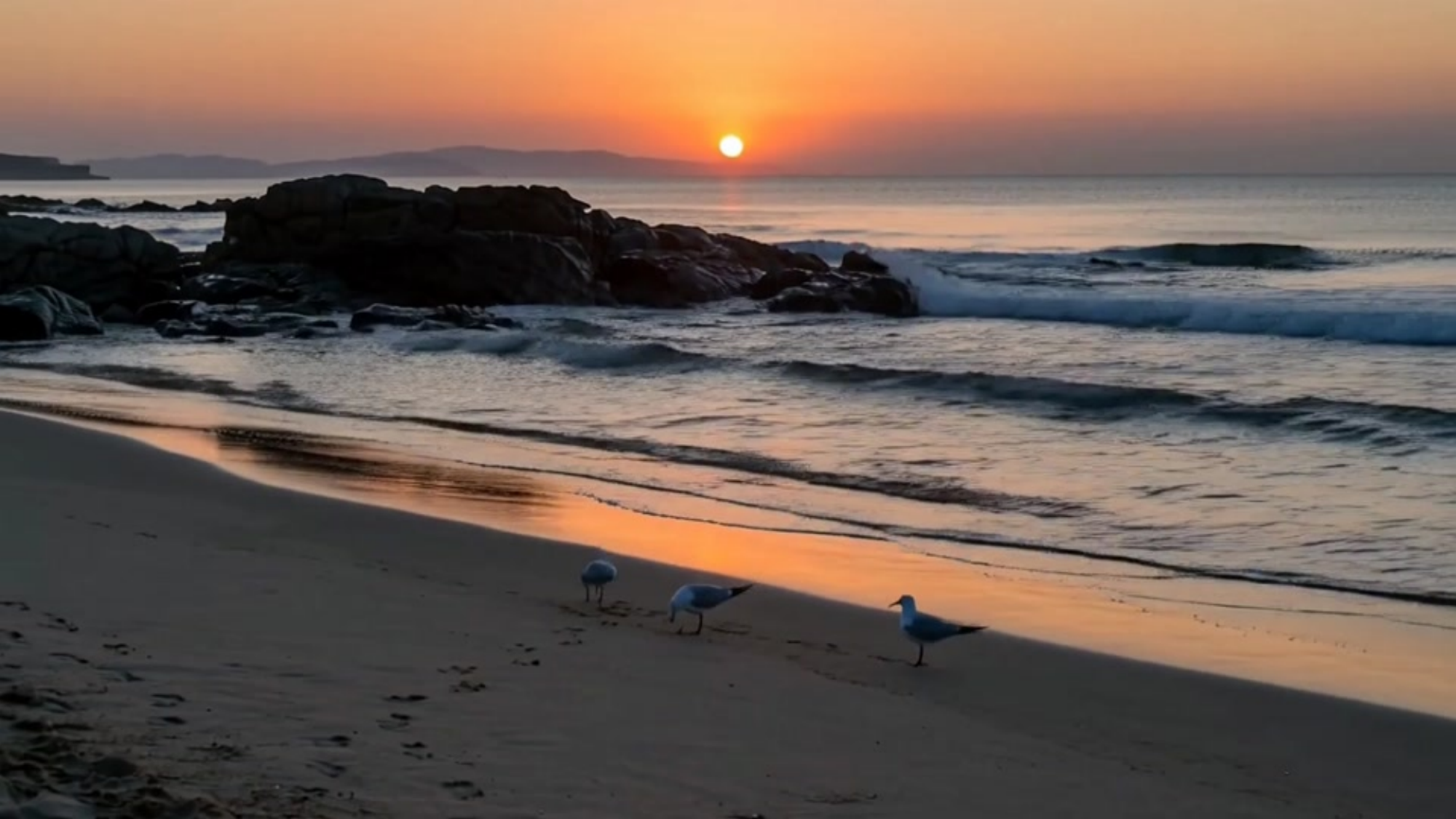}
        & \appcaseframe{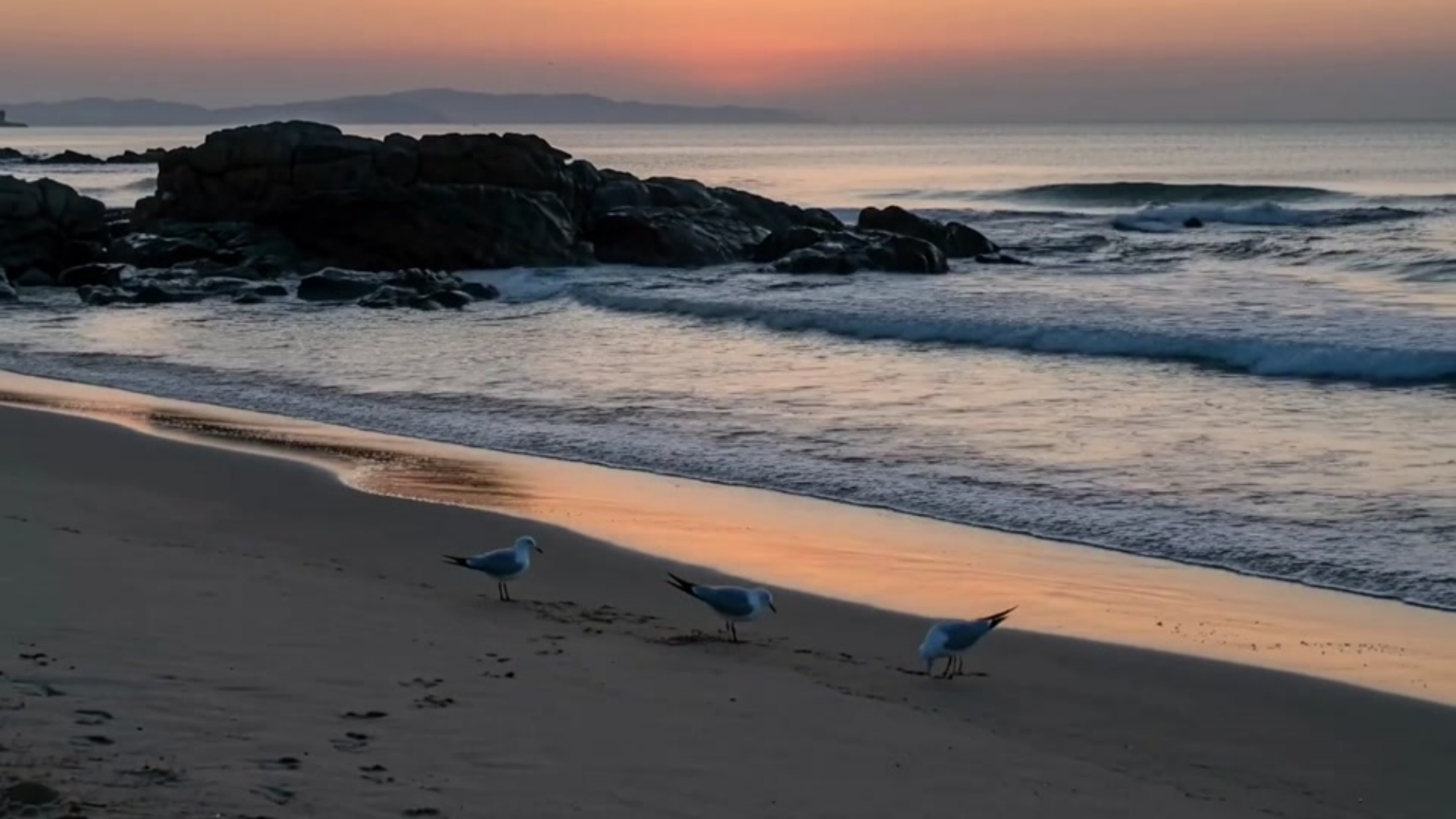}
        \end{tabular}
\par\vspace{3pt}\noindent\begin{tabular}{@{}c@{\hspace{\appGap}}c@{\hspace{\appGap}}c@{\hspace{\appGap}}c@{}}

        \appcaselabel{H3 + TIDE\\+ NeurCo (O)}{Video reference}{}
        & \appcaseframe{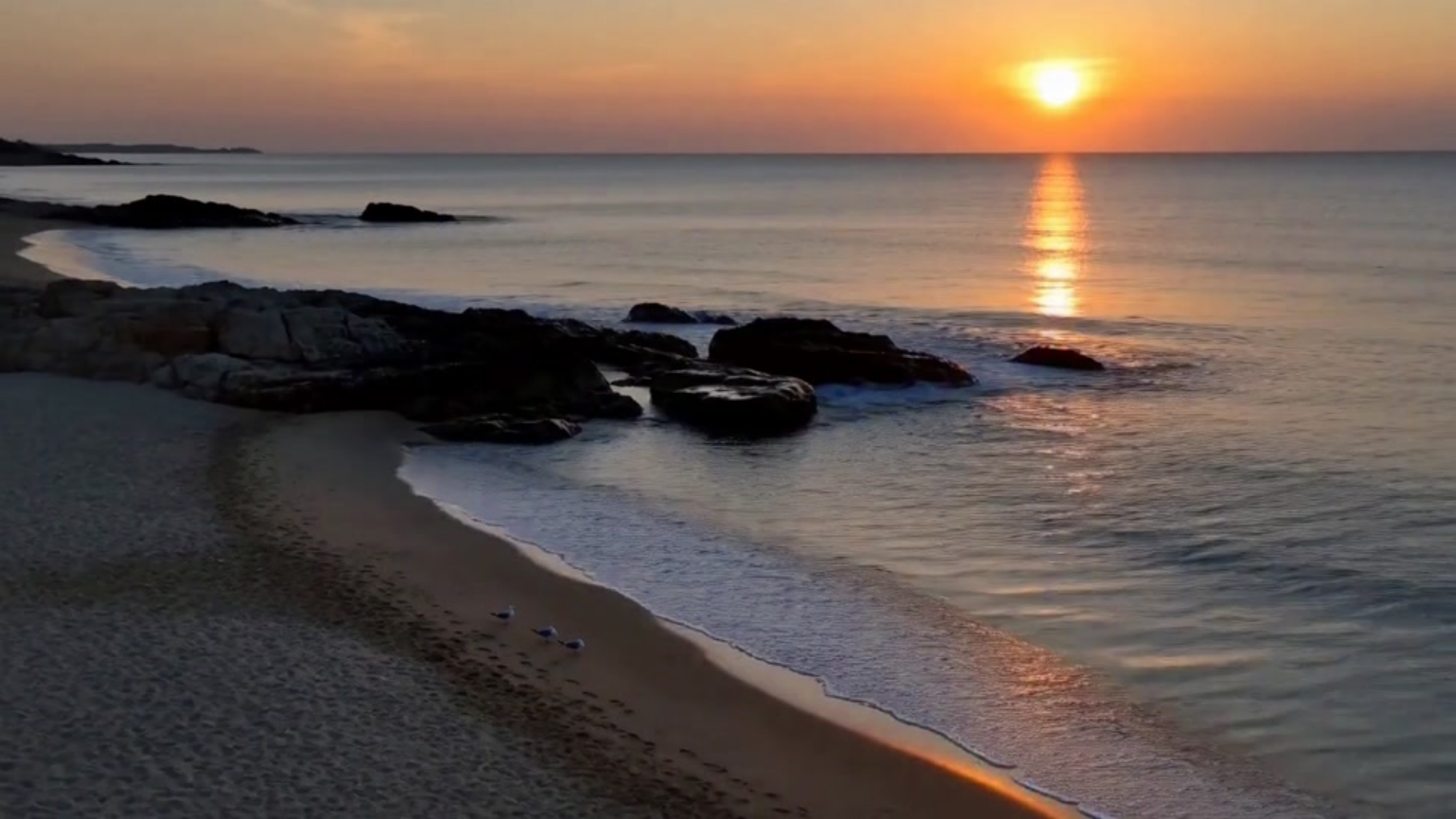}
        & \appcaseframe{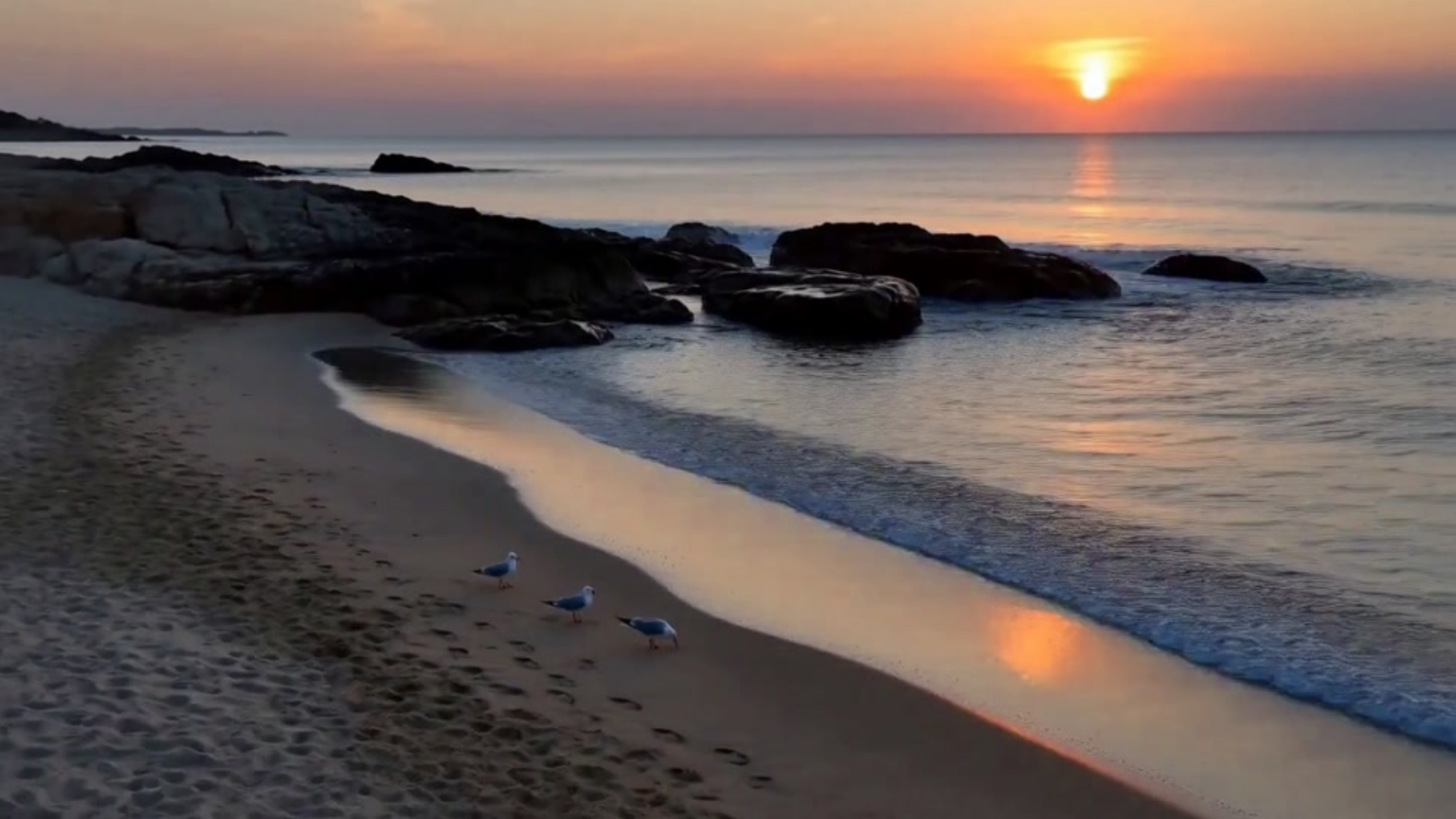}
        & \appcaseframe{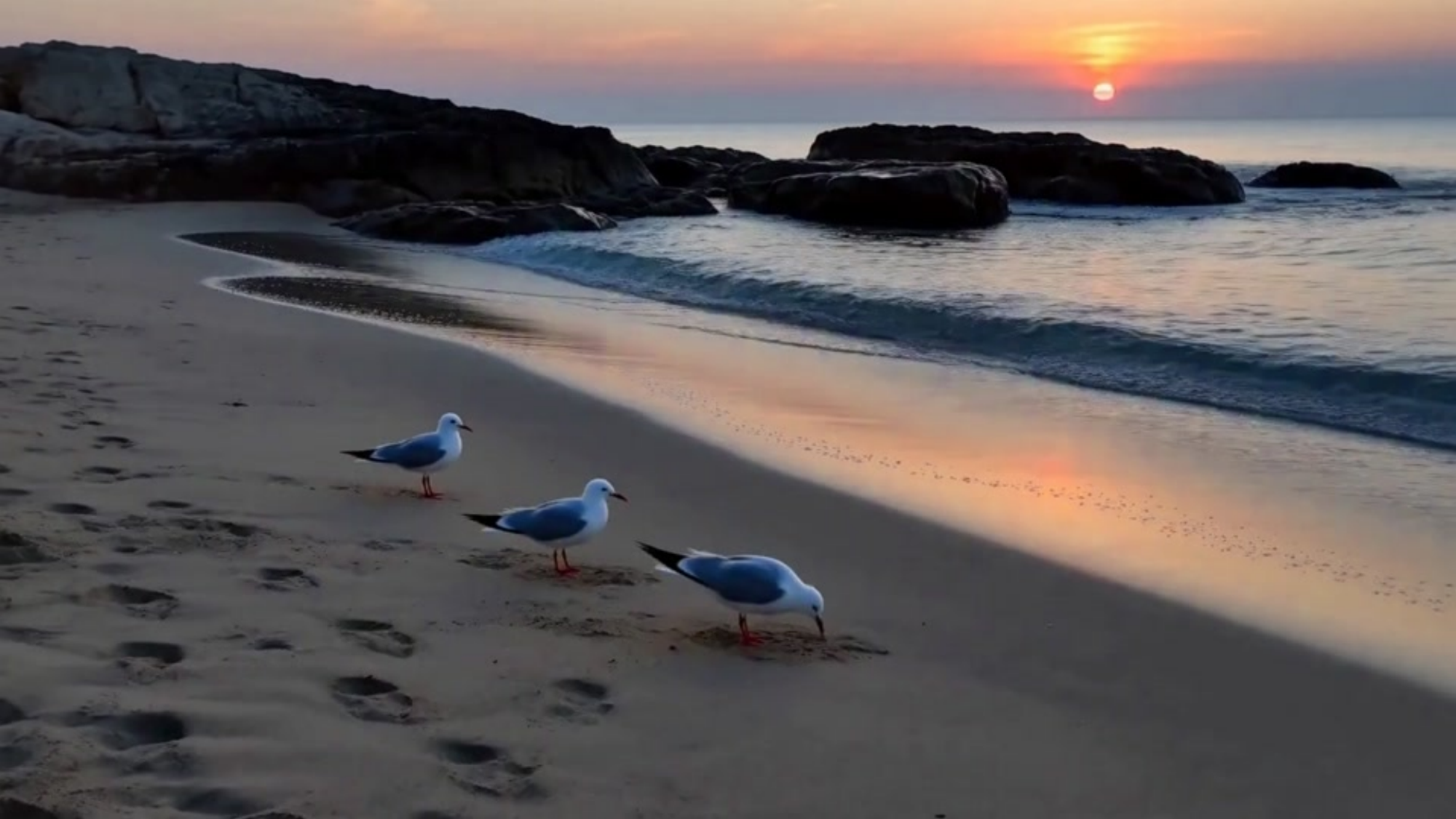}
        \end{tabular}
\par\vspace{3pt}\noindent\begin{tabular}{@{}c@{\hspace{\appGap}}c@{\hspace{\appGap}}c@{\hspace{\appGap}}c@{}}

        \appcaselabel{Wan + TIDE}{T2V}{}
        & \appcaseframe{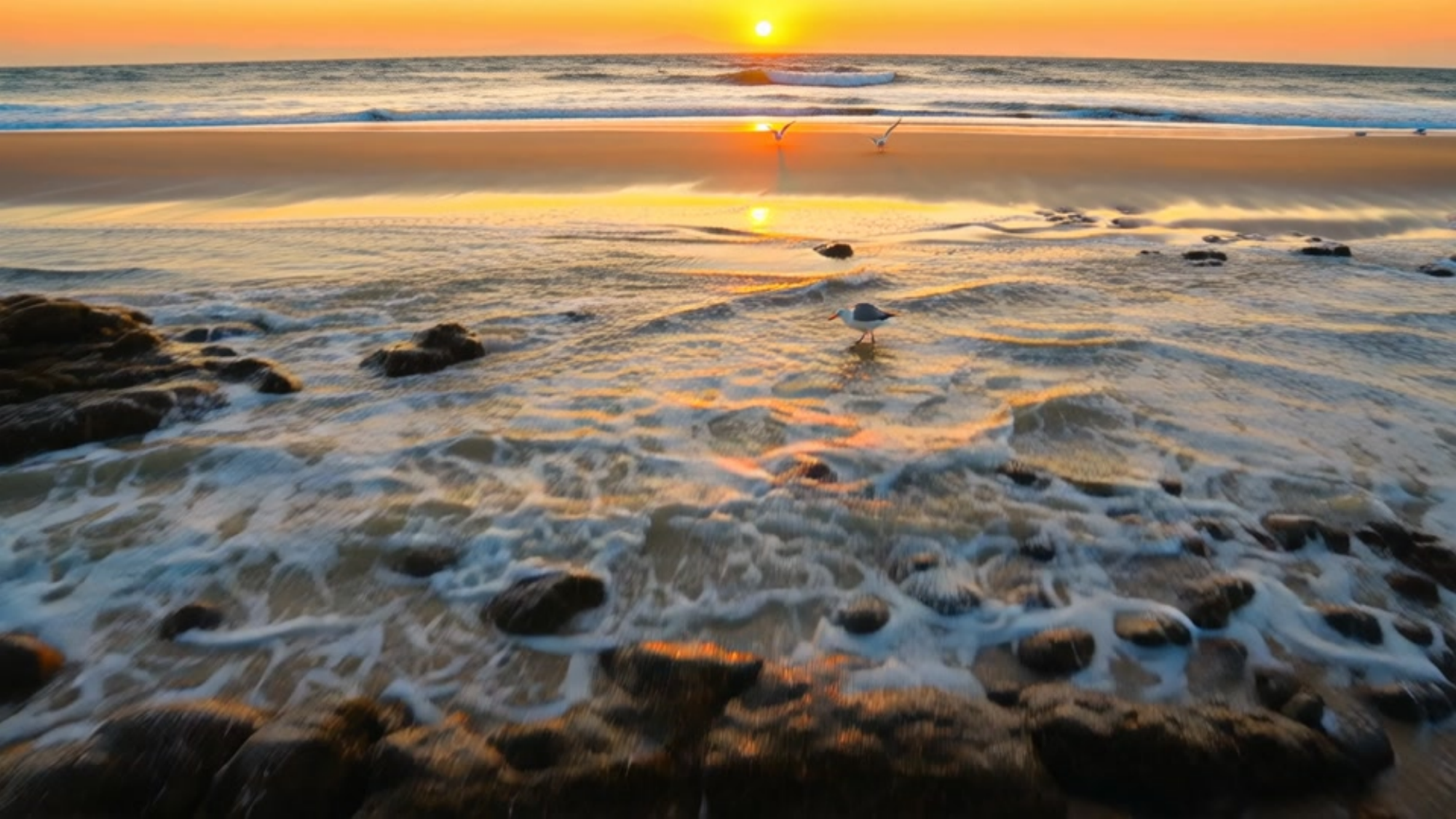}
        & \appcaseframe{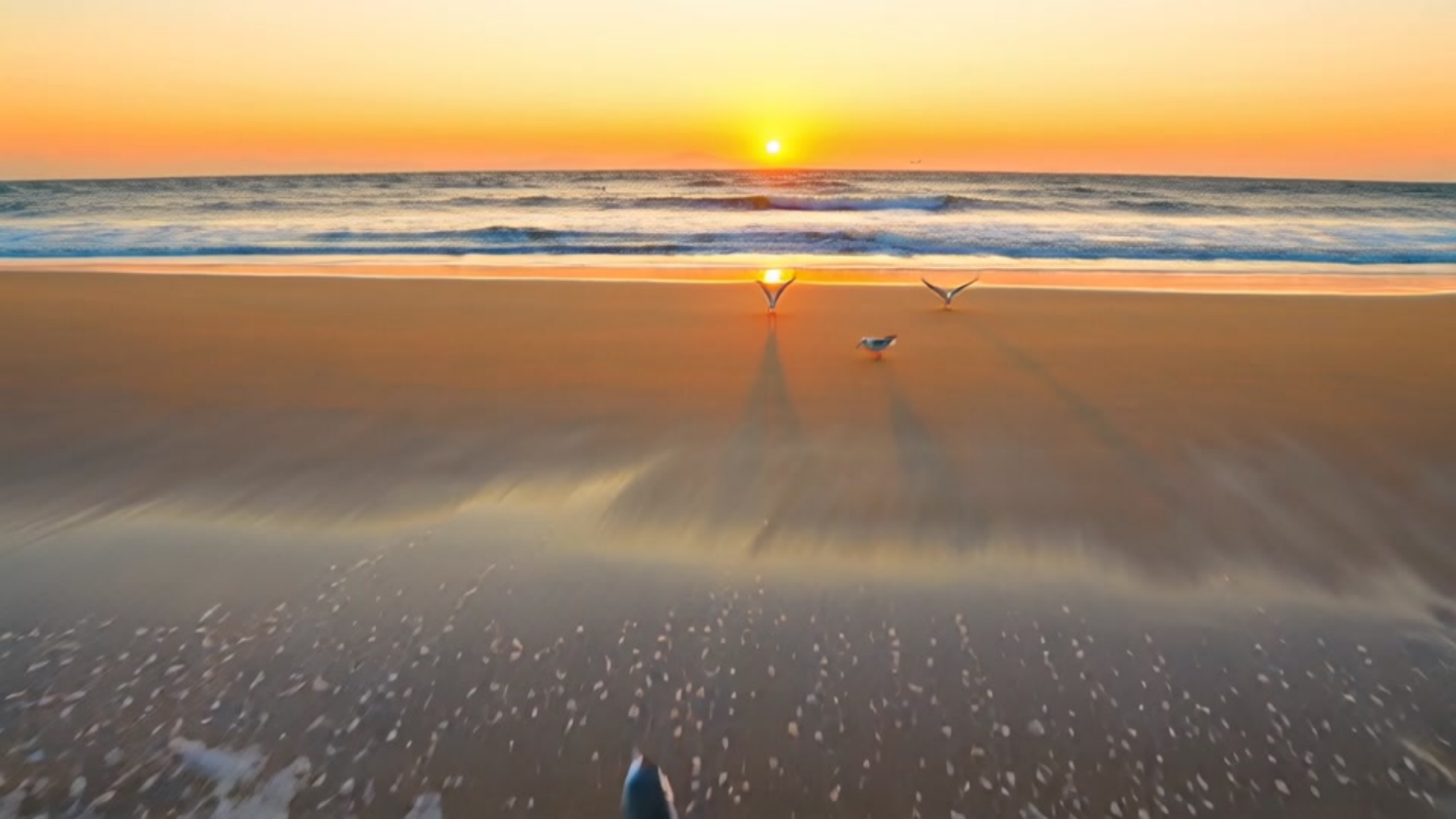}
        & \appcaseframe{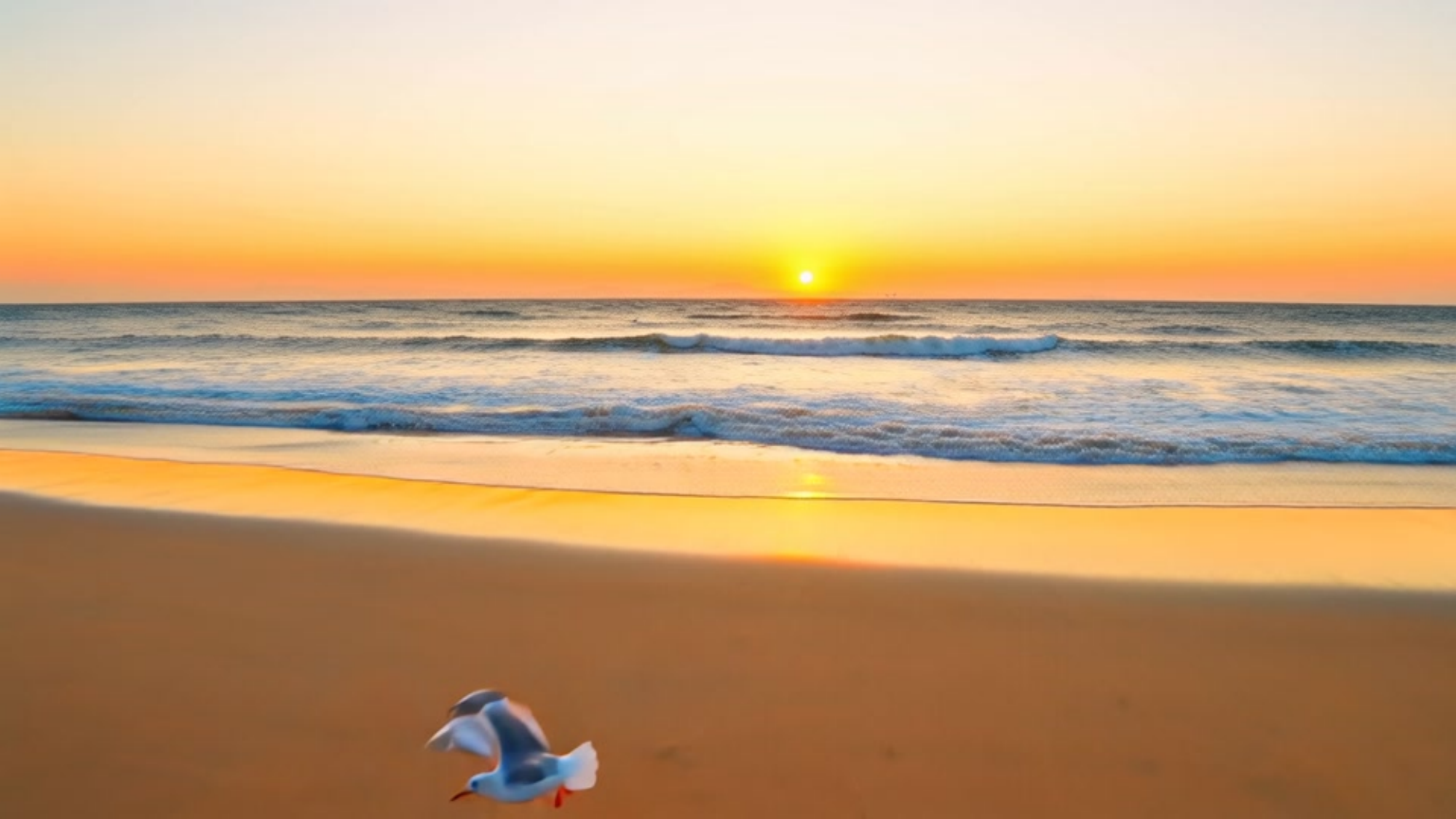}
        \end{tabular}
\par\vspace{3pt}\noindent\begin{tabular}{@{}c@{\hspace{\appGap}}c@{\hspace{\appGap}}c@{\hspace{\appGap}}c@{}}

        \appcaselabel{Wan + TIDE\\+ NeurCo (O)}{Motion guidance}{}
        & \appcaseframe{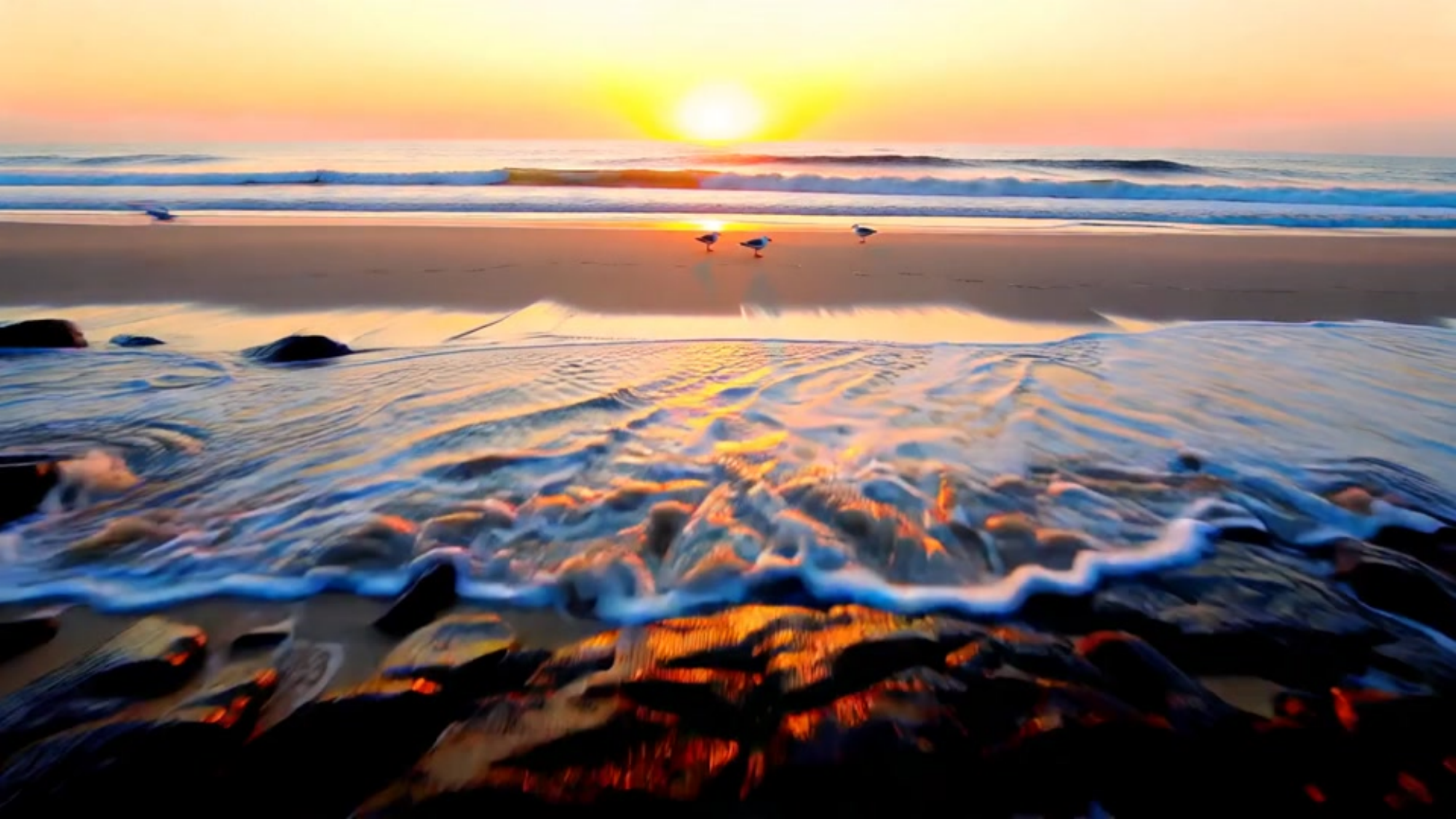}
        & \appcaseframe{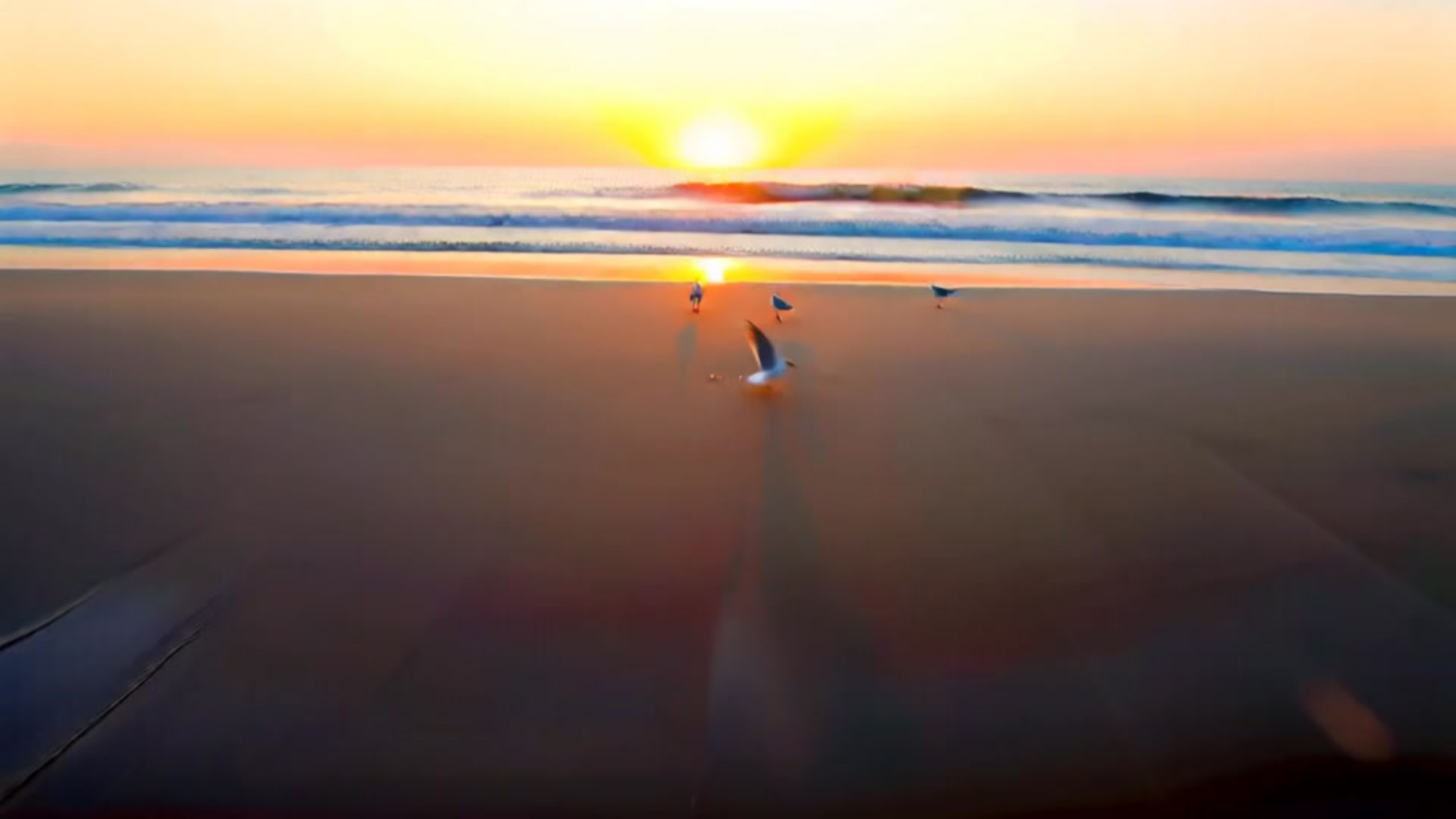}
        & \appcaseframe{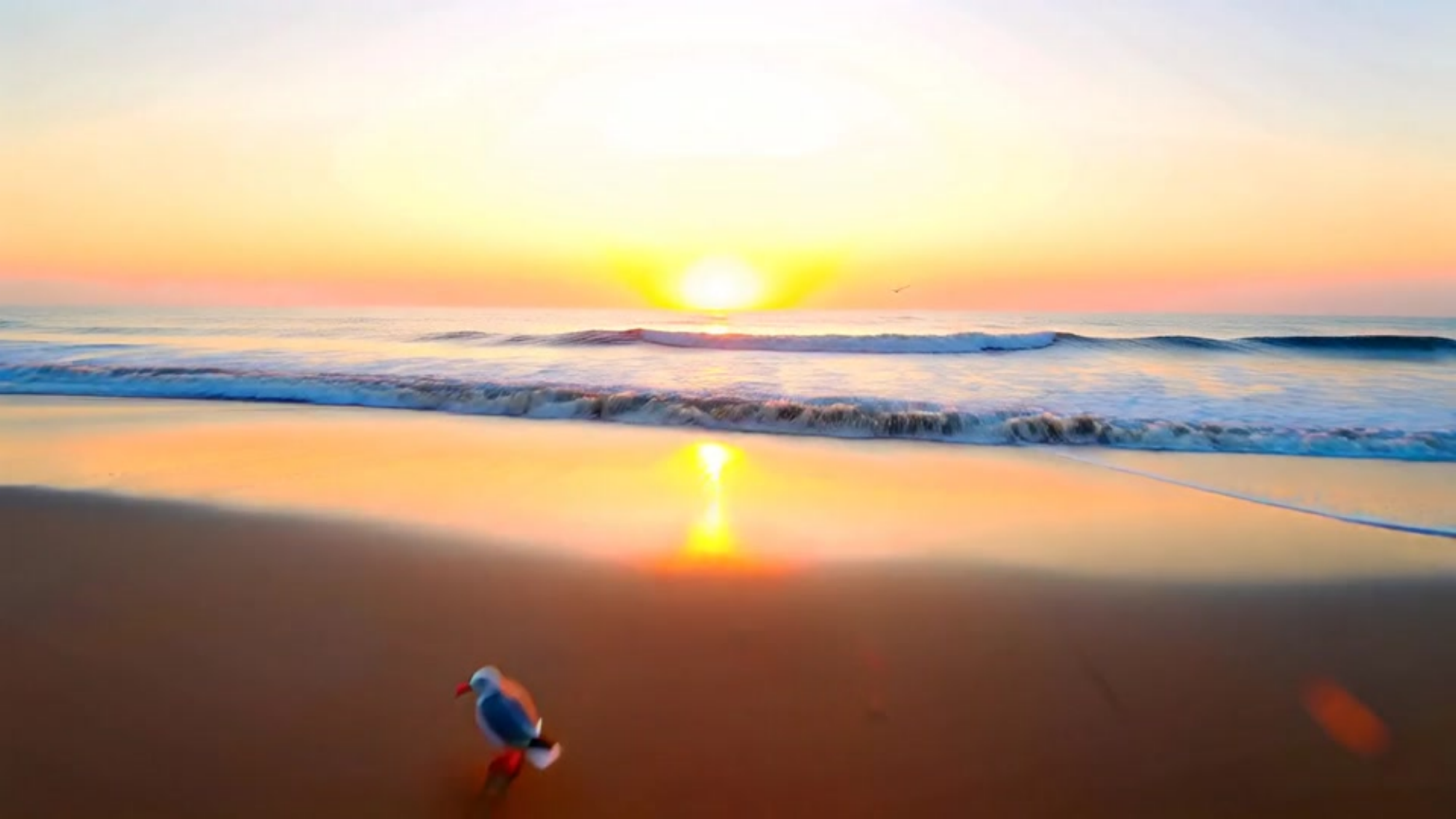}
    \end{tabular}
    \nopagebreak[3]\caption{\textbf{Alternative plans produce different camera trajectories.}
    Native, TIDE, and TIDE + NeurCo (O) are shown for the same coastline
    request on H3 and Wan. Native generations provide visual context.}
    \label{fig:case_coast}
    \endgroup
\par\endgroup
\end{appcaseblock}

\section{NeurCo Prompt Templates}
\label{sec:app_neurco_templates}

The following listings specify text rewriting, condition construction,
validation, and video conditioning for NeurCo.
Angle-bracketed fields are filled for each request; notes identify the
images or video frames supplied alongside the text.
The two proxy operators share construction and validation prompts.

\providecommand{\neurcoappcode}[1]{{\fontencoding{T1}\selectfont\detokenize{#1}}}
\providecommand{\neurcoappmono}[1]{{\fontencoding{T1}\ttfamily\footnotesize\detokenize{#1}}}
\newtcolorbox{neurcoapppromptbox}[1]{%
    enhanced, breakable,
    colback=white, colframe=neurcoappblue!40,
    colbacktitle=neurcoappblue!6, coltitle=black,
    title={#1}, title after break={#1 (continued)},
    fonttitle=\bfseries\small, fontupper=\small,
    boxrule=.4pt, arc=.5mm,
    left=1.5mm, right=1.5mm, top=1.2mm, bottom=1.2mm,
    before skip=6pt, after skip=7pt
}

\subsection{Targeted Rewriting}
\label{sec:app_neurco_prompt}

The rewriter receives the full prompt and selected source spans,
each paired with a control name and setting.
Text-only and operator-enabled compilation share the system message;
the selected targets and optional Wan vocabulary vary between calls.
A corrupted word in the repetition example is corrected for
readability.\footnote{The original source form is retained in the accompanying
prompt-template archive.}

\begin{neurcoapppromptbox}{Text rewriting: system message}
    \raggedright
    \neurcoappcode{You edit a video-generation prompt. Rewrite ONLY the TARGET sentences so the requirement each one states is unmistakable and executable for a video model.}

    \neurcoappcode{How to make a target explicit -- spell out, for the thing it asks about:}

    \neurcoappcode{- subject: bind each action to the entity that performs it;}

    \neurcoappcode{- numbers and repetitions: exact count, each occurrence in order ("jumps once, lands, jumps a second time, lands, jumps a third time"), never more times than the original says;}

    \neurcoappcode{- camera: standard move names, one after another, in the stated order; keep a static camera when the source is static, do not add a second move;}

    \neurcoappcode{- space: position relative to named objects and to the camera, screen position, initial pose, motion direction/path, occlusion or depth order, entry and exit edges with identity continuity;}

    \neurcoappcode{- motion and physics: contact, force, material response, and the end state of the action;}

    \neurcoappcode{- time: order of events as beginning -> transition -> observable result; timing in seconds when the target gives it;}

    \neurcoappcode{- text to appear on screen: exact wording, place, and carrier.}

    \neurcoappcode{Prefer detailed, observable specifications over vague brevity. Detail must still fit the clip duration.}

    \neurcoappcode{Hard limits:}

    \neurcoappcode{- Make each target say exactly what it already asks, only more concretely. Do not add events, subjects, style, mood or new content; do not invent entities; do not delete or weaken an explicit source fact.}

    \neurcoappcode{- Keep every other word of the prompt exactly as it is, in the same order.}

    \neurcoappcode{Answer only with JSON {"text": "<the full prompt after editing>"}.}
\end{neurcoapppromptbox}

\begin{neurcoapppromptbox}{Text rewriting: user message}
    \raggedright
    \neurcoappcode{PROMPT:}

    \neurcoappcode{<full input prompt>}

    \neurcoappcode{TARGET sentences (rewrite only these):}

    \neurcoappcode{- "<verbatim target span 1>"  (what it must achieve -- <control name>: <setting>)}

    \neurcoappcode{- "<verbatim target span 2>"  (what it must achieve -- <control name>: <setting>)}

    \normalfont\small
    Repeat the target line for every selected requirement.
    Target spans may be phrases rather than complete sentences.
    For Wan's configured rewriting style, append the message below.
\end{neurcoapppromptbox}

\begin{neurcoapppromptbox}{Text rewriting: Wan vocabulary}
    \raggedright
    \neurcoappcode{Where it fits, use the vocabulary of Wan's official prompt guide: shot size (Medium shot, Medium close-up shot, Wide shot, Medium wide shot, Close-up shot, Extreme close-up shot, Extreme wide shot), camera angle (Over-the-shoulder shot, Low angle shot, High angle shot, Dutch angle shot, Aerial shot, Overhead shot), composition (Center / Balanced / Right-heavy / Left-heavy / Symmetrical composition); describe an action as the step-by-step process of how it unfolds.}
\end{neurcoapppromptbox}

The response contains the full rewritten prompt in a JSON
\texttt{text} field.
If the call fails or the returned text is shorter than 60\% of the
original character length, the original prompt is retained.
This length check detects truncation; it does not verify semantic fidelity.
Text adaptation precedes condition construction, allowing a failed
visual strategy to retain its textual fallback.

\subsection{Reference Image}
\label{sec:app_ref_image_prompts}

The planner receives the video description and selected requirements.
It uses the style variant when any target specifies visual style and
the appearance variant otherwise.
The returned image prompt is submitted directly to the image generator.

\begin{neurcoapppromptbox}{Appearance reference: planning}
    \raggedright
    \neurcoappcode{VIDEO DESCRIPTION:}

    \neurcoappcode{<original video description>}

    \neurcoappcode{Write one prompt for a single still REFERENCE image that a video model will use as the appearance reference. It must show exactly:}

    \neurcoappcode{- "<target span 1>" (<control description 1>)}

    \neurcoappcode{- "<target span 2>" (<control description 2>)}

    \neurcoappcode{Show only the subject(s) these requirements are about, fully visible, side by side if several, on a plain neutral background, no other objects and no text.}

    \neurcoappcode{Return JSON {"image_prompt": str, "what": "short noun phrase naming what the reference shows"}.}
\end{neurcoapppromptbox}

\begin{neurcoapppromptbox}{Style reference: planning}
    \raggedright
    \neurcoappcode{VIDEO DESCRIPTION:}

    \neurcoappcode{<original video description>}

    \neurcoappcode{Write one prompt for a single still REFERENCE image that a video model will use as the style reference. It must show exactly:}

    \neurcoappcode{- "<target span 1>" (<control description 1>)}

    \neurcoappcode{- "<target span 2>" (<control description 2>)}

    \neurcoappcode{Show the scene's main subject in that visual style.}

    \neurcoappcode{Return JSON {"image_prompt": str, "what": "short noun phrase naming what the reference shows"}.}
\end{neurcoapppromptbox}

Repeat one target line per selected requirement.
The generated image is then attached to the following multimodal
validation request; all listed answers must be affirmative.

\begin{neurcoapppromptbox}{Reference image: validation}
    \raggedright
    \neurcoappcode{Answer each question about the image. Questions:}

    \neurcoappcode{[0] Is this clearly true in the image: "<target span 1>"?}

    \neurcoappcode{[1] Is this clearly true in the image: "<target span 2>"?}

    \neurcoappcode{Return JSON {"answers": [{"i": int, "answer": "yes"|"no"}]}}

    \normalfont\small
    Attach the generated reference image.
    Add one numbered question for every selected requirement.
\end{neurcoapppromptbox}

After validation, the relevant instruction is appended to the video prompt.
Image index \(k\) follows the existing references; the appearance phrase
uses the planner's description, defaulting to ``the subject.

\begin{neurcoapppromptbox}{Reference image: video instruction}
    \raggedright
    \textbf{Appearance variant}
\par\nobreak
    \neurcoappcode{Image <k> shows <what>; keep its appearance exactly.}

    \textbf{Style variant}
\par\nobreak
    \neurcoappcode{Use Image <k> as the visual style reference for the whole video.}
\end{neurcoapppromptbox}

\subsection{Generated First Frame}
\label{sec:app_first_frame_prompts}

The planner converts the opening scene and selected count/text targets
into one image-generation prompt.
The following template shows one target of each kind; only lines for
the selected kinds are included, and multiple target lines are allowed.
The requested text is the quoted string extracted from the original
request, not an invented string of the requested length.

\begin{neurcoapppromptbox}{First frame: planning}
    \raggedright
    \neurcoappcode{VIDEO DESCRIPTION:}

    \neurcoappcode{<original video description>}

    \neurcoappcode{Write one prompt for a photorealistic still image that will be the FIRST FRAME of this video. It must show the opening moment of the scene and contain:}

    \neurcoappcode{- exactly <count> of the subjects described as: <count target span> (all fully visible and clearly separated)}

    \neurcoappcode{- the exact on-screen text "<quoted text>", legible, spelled exactly, in a natural place for it}

    \neurcoappcode{Return JSON {"image_prompt": str, "nouns": [the plural noun to count, one per counted subject, in the same order]}.}
\end{neurcoapppromptbox}

The returned \texttt{image\_prompt} is sent directly to the image
generator.
The validator receives the generated frame and the noun phrases returned
by the planner.
It uses the appropriate variant below depending on the selected targets.

\begin{neurcoapppromptbox}{First frame: validation variants}
    \raggedright
    \textbf{Count only}
\par\nobreak
    \neurcoappcode{Count, one by one, each of these in the image: <noun 1>; <noun 2>. Return JSON {"counts": [int, ...]}.}

    \textbf{Text only}
\par\nobreak
    \neurcoappcode{Transcribe exactly all text visible in the image. Return JSON {"text": str}.}

    \textbf{Count and text}
\par\nobreak
    \neurcoappcode{Count, one by one, each of these in the image: <noun 1>; <noun 2>. Transcribe exactly all text visible in the image. Return JSON {"counts": [int, ...], "text": str}.}

    \normalfont\small
    Attach the generated first frame.
    The semicolon-separated noun list has one entry per counted target,
    in the same order as the targets.
    Only one variant is sent.
\end{neurcoapppromptbox}

The returned counts and normalized transcription are compared with the
request. On success, the frame and compiled text are submitted as the
I2V input.

\subsection{Layout First Frame}
\label{sec:app_layout_frame_prompts}

The planner returns object boxes and depth order. A fixed renderer draws
the layout sketch, which an image-editing model converts to a
photorealistic frame. The final image is checked against the spatial targets.

\begin{neurcoapppromptbox}{Layout planning: system message}
    \raggedright
    \neurcoappcode{You plan the exact spatial layout of the FIRST FRAME of a video, for a layout renderer. Answer only with JSON.}

    \neurcoappcode{Return {"background": str, "camera": "front"|"high"|"low", "objects": [{"name": str, "box": [x0, y0, x1, y1], "depth": int, "occluded_by": [names]}]}}

    \neurcoappcode{- One entry per individual object instance (ten cats = ten entries, named cat_1 ... cat_10).}

    \neurcoappcode{- box is normalized to the frame (0--1, origin top-left); keep every required object at least partly visible.}

    \neurcoappcode{- depth: 0 = nearest to the camera; larger = farther. Nearer objects must overlap farther ones exactly where the requirement asks for occlusion (list who covers whom in occluded_by).}

    \neurcoappcode{- Respect every count, position and front/back relation in the REQUIREMENTS exactly; invent nothing else.}
\end{neurcoapppromptbox}

\begin{neurcoapppromptbox}{Layout planning: user message}
    \raggedright
    \neurcoappcode{VIDEO DESCRIPTION:}

    \neurcoappcode{<original video description>}

    \neurcoappcode{REQUIREMENTS:}

    \neurcoappcode{- <spatial target span 1>}

    \neurcoappcode{- <spatial target span 2>}

    \normalfont\small
    Repeat one line per selected spatial requirement.
    On a later planning attempt, append:

    \neurcoappcode{(variant <attempt>: an independent alternative plan)}
\end{neurcoapppromptbox}

\begin{neurcoapppromptbox}{Layout: image-editing instruction}
    \raggedright
    \neurcoappcode{Turn this layout sketch into a photorealistic photograph that is the opening frame of this video: <original video description>}

    \neurcoappcode{Each colored block is one object; keep every object's position, size, count and front/back overlap exactly as in the sketch. Requirements: <spatial target span 1>; <spatial target span 2>. No labels, boxes or text from the sketch may remain.}

    \normalfont\small
    Supply the rendered sketch as the image-edit input.
    Include all selected spatial targets in the semicolon-separated list.
\end{neurcoapppromptbox}

\begin{neurcoapppromptbox}{Layout image: validation}
    \raggedright
    \neurcoappcode{Does this image satisfy every requirement exactly? Count carefully.}

    \neurcoappcode{- <spatial target span 1>}

    \neurcoappcode{- <spatial target span 2>}

    \neurcoappcode{Return JSON {"ok": bool, "problems": [str]}}

    \normalfont\small
    Attach the converted photorealistic frame.
    Repeat one line per selected spatial requirement.
\end{neurcoapppromptbox}

A validated frame is submitted with the compiled text as the I2V input.

\subsection{Shared Proxy Construction and Audit}
\label{sec:app_proxy_shared_prompts}

Both proxy operators use the same Blender planner and gray-video
validator. Selected targets and one check per target are combined into
a single request. The planner returns key states and code, from which
a fixed renderer produces the gray animation and depth sequence.

\begin{neurcoapppromptbox}{Proxy construction: system message}
    \raggedright
    \neurcoappcode{You build a gray clay PROXY animation in Blender that a video model will follow for layout, motion and camera.}

    \neurcoappcode{Only geometry, positions, motion, timing and camera matter; appearance (colors, textures, faces, clothing) does not.}

    \neurcoappcode{Answer only with JSON {"keyframes": [{"t": seconds, "state": str}], "code": str}.}

    \neurcoappcode{First plan "keyframes": 3-6 moments (t in seconds) saying where every relevant object is, which way it faces, and where the camera is. Make every requirement in REQUIREMENT and every CHECK visibly true at the right moment, in the right order.}

    \neurcoappcode{Then write "code": Python executed inside Blender with these already defined (do not import bpy or redefine them):}

    \neurcoappmono{N (frame count), FPS, sec(t) -> frame number of time t}

    \neurcoappmono{box(name, size=(x,y,z), loc, gray=0.8, parent=None, rot=(deg,deg,deg)), sphere(name, radius, loc, gray, parent),}

    \neurcoappmono{cylinder(name, radius, depth, loc, gray, parent, rot), cone(name, radius, depth, loc, gray, parent, rot)}

    \neurcoappmono{group(name, loc) -> empty to parent parts to (children use coordinates relative to the parent)}

    \neurcoappmono{person(name, loc, height=1.7, gray=0.85, facing=deg) -> root of a simple humanoid; facing=0 looks toward -Y}

    \neurcoappmono{key(ob, frame, loc=None, rot=(deg,deg,deg), scale=None) -> keyframe; cam_key(frame, loc, look_at=None, rot=None, lens=None)}

    \neurcoappmono{ground (a large plane at z=0), cam (default at (0,-12,3) looking at (0,0,1); +X is screen right, +Y is away from camera)}

    \neurcoappcode{Rules:}

    \neurcoappcode{- Units are meters; objects stand on z=0. Build each required object from these primitives (animals = body box + head sphere + legs; vehicles = boxes; buildings = tall boxes). Give each object a distinct gray (0.3-0.95) so they are distinguishable; never use text labels.}

    \neurcoappcode{- Animate with key() at sec(...) frames. Counted actions (jumps, bounces, claps) must be separate, clearly visible repetitions: exactly the stated number, each with a visible rise and return. Rotations must reach the stated angle.}

    \neurcoappcode{- Camera moves must be large and unambiguous (pan/tilt >= 25 degrees, dolly/truck >= 3 m), one move after another in the stated order, each move finishing before the next begins. If the prompt implies no camera move, keep cam_key fixed and frame every object fully inside the view.}

    \neurcoappcode{- Occlusion / depth order: place nearer objects closer to the camera along its view direction so they partly cover the farther ones exactly as required, while every object stays partly visible.}

    \neurcoappcode{- Build ONLY the objects that the REQUIREMENT and CHECKS involve, plus the ground. Do NOT add decorative scenery (trees, buildings, furniture, plants, rocks, sky objects): the video model draws the background from the prompt, and anything you build is copied literally into the video. Exception: for camera-movement requirements, add 3-6 plain boxes as static reference points so the camera motion is visible.}

    \neurcoappcode{- Frame the main subjects large (together they should span at least a third of the frame height).}

    \neurcoappcode{- Keep everything inside the frame unless the prompt says it leaves. The whole action must fit in N frames.}
\end{neurcoapppromptbox}

\begin{neurcoapppromptbox}{Proxy construction: user message}
    \raggedright
    \neurcoappcode{PROMPT:}

    \neurcoappcode{<original video description>}

    \neurcoappcode{REQUIREMENT (<joined control IDs> = <joined levels>): <joined target spans>}

    \neurcoappcode{CHECKS (each must be visibly true in the proxy):}

    \neurcoappcode{- Is this clearly and fully true in the animation: "<target span 1>"?}

    \neurcoappcode{- Is this clearly and fully true in the animation: "<target span 2>"?}

    \neurcoappcode{Duration: <duration> s, N = <frame count> frames at 16 fps.}

    \normalfont\small
    Repeat one check per target.
    The frame count is \(16\,\text{duration}+1\).
\end{neurcoapppromptbox}

\begin{neurcoapppromptbox}{Proxy validation: system message}
    \raggedright
    \neurcoappcode{You check a gray clay PROXY animation: simple gray shapes stand in for the subjects of a video prompt (a figure made of cylinders = a person, a box with legs = an animal, etc.). Ignore appearance, materials and realism entirely. Judge only positions, depth order, occlusion, motion direction, counts, rotation angles, order of events and camera movement, using the sampled frames and their timestamps. Answer JSON {"answers": [{"qid": str, "answer": "yes"|"no", "reason": str}], "problems": [str]} where problems tells the animator concretely what to change (empty if every answer is yes).}
\end{neurcoapppromptbox}

\begin{neurcoapppromptbox}{Proxy validation: user message}
    \raggedright
    \neurcoappcode{<full construction User prompt above>}

    \neurcoappcode{Sampled proxy frames (<measured duration to one decimal> s):}

    \neurcoappcode{t = <timestamp to one decimal> s}

    \normalfont\small
    Attach the corresponding gray-video frame, then repeat the timestamp
    and image pair for every sampled frame.
    End the message with the following numbered questions:

    \neurcoappcode{Questions:}

    \neurcoappcode{[q0] Is this clearly and fully true in the animation: "<target span 1>"?}

    \neurcoappcode{[q1] Is this clearly and fully true in the animation: "<target span 2>"?}
\end{neurcoapppromptbox}

Every check must receive an affirmative answer.
If rendering or validation fails, the next construction request appends
the relevant feedback below to revise the proxy code.

\begin{neurcoapppromptbox}{Proxy construction: retry feedback}
    \raggedright
    \textbf{Blender rendering failure}
\par\nobreak
    \neurcoappcode{YOUR PREVIOUS CODE:}

    \neurcoappcode{<previous generated code>}

    \neurcoappcode{IT FAILED IN BLENDER:}

    \neurcoappcode{<render error>}

    \neurcoappcode{Fix it.}

    \textbf{Rendered proxy fails the audit}
\par\nobreak
    \neurcoappcode{YOUR PREVIOUS CODE:}

    \neurcoappcode{<previous generated code>}

    \neurcoappcode{A REVIEWER WATCHED THE RENDER AND FOUND:}

    \neurcoappcode{- <problem 1>}

    \neurcoappcode{- <problem 2>}

    \neurcoappcode{Rewrite the code so every check is clearly satisfied.}
\end{neurcoapppromptbox}

\subsubsection{Proxy Video Reference: H3 Conditioning}
\label{sec:app_proxy_ref_condition}

The validated gray video is attached through H3's video-reference
interface, with the following instruction appended to the compiled prompt.

\begin{neurcoapppromptbox}{H3 proxy: video instruction}
    \raggedright
    \neurcoappcode{<Video 1> is a gray clay proxy animation used only as a motion reference: follow its object positions, movements, number of repetitions, rotation, timing and camera movement exactly, but render the scene photorealistically as described above. Do not copy its gray, untextured look.}
\end{neurcoapppromptbox}

\subsubsection{Proxy Depth Control: Wan Conditioning}
\label{sec:app_proxy_control_condition}

The validated depth sequence is supplied to VACE as its control video.
When realistic rendering is enabled, the executor appends the positive
instruction and extends the base negative prompt with the suffix below.

\begin{neurcoapppromptbox}{Wan depth control: positive instruction}
    \raggedright
    \neurcoappcode{Photorealistic live-action footage with natural lighting and real-world textures.}
\end{neurcoapppromptbox}

\begin{neurcoapppromptbox}{Wan depth control: negative suffix (translation)}
    \raggedright
    Cartoons, animation, 3D rendering, CG, flat illustration, simple line
    drawing, paper cutting, plastic texture, plain-color backgrounds,
    striped ground.
\end{neurcoapppromptbox}

The box translates the Chinese suffix. The original string is retained
in the prompt-template archive and is appended only when realistic
rendering is enabled.

\subsection{Shot Concatenation}
\label{sec:app_shot_concat_prompts}

The splitter returns the requested number of self-contained shot
prompts. Each is submitted separately to the video model.

\begin{neurcoapppromptbox}{Shot splitting: user message}
    \raggedright
    \neurcoappcode{PROMPT:}

    \neurcoappcode{<compiled video description>}

    \neurcoappcode{This prompt asks for a video with <shot count> shots. Split it into exactly <shot count> prompts, one per shot, in order. Each shot prompt must be self-contained: repeat the full description of every recurring subject and of the setting word for word so they look identical across shots, then say what happens and how the camera frames it in that shot only. Do not mention cuts or other shots. Return JSON {"shots": [str, ...]}.}
\end{neurcoapppromptbox}

The split is accepted only if it contains the requested number of nonempty
shot prompts. Once all clips are available, they are concatenated in order
at 16 fps. Repeated descriptions encourage, but do not guarantee,
cross-shot identity consistency.

\section{TIDE Implementation and Prompts}
\label{sec:app_tide_prompt}

We adapt TIDE to return one positive prompt in a single language-model
call, using its rule triggers and instructions.
The adaptation omits the separate negative-prompt output and performs
no candidate-video search or video-feedback loop.
For TIDE-based variants, NeurCo compiles the enhanced positive prompt using the same deployment-specific manual and intervention catalog. T enables targeted rewriting, O enables compatible external conditions and operations without targeted rewriting, and T+O enables both. Operator-specific conditioning instructions remain available in O. Unchanged executable inputs reuse the corresponding TIDE video.

The listed local rules are version 1.1, revision 16.
The module placeholder is replaced by the seven rules in
their listed order. Other rule-file metadata is not included in the
system message.

\begin{neurcoapppromptbox}{TIDE: system message}
    \raggedright
    \neurcoappcode{You rewrite a user request into ONE generation-ready text-to-video prompt.}

    \neurcoappcode{Procedure (do this privately; expose no reasoning):}

    \neurcoappcode{1. Build a fidelity ledger: every explicit entity, count, attribute, relation, action, event order, camera instruction, style, quoted text, duration, and delivery constraint. Mark each immutable.}

    \neurcoappcode{2. Compile a feasible action graph for a short clip: at most one main subject and one coherent action arc unless the source requires more. Express ordered events as beginning -> transition -> observable result. Do not add an event merely to make the prompt sound cinematic.}

    \neurcoappcode{3. Activate only source-triggered modules from the list below. Multiple modules may activate when separate source facts satisfy separate triggers. Do not activate a module because another fired.}

    \neurcoappcode{4. Compile the prompt: prefer detailed observable specifications over vague brevity -- bind each action to its subject; specify source-triggered screen position, initial pose, motion direction/path, contact, force, material response, and end state. Detail must remain executable within the clip duration and may not introduce a competing event.}

    \neurcoappcode{5. Run a silent verifier: source coverage, entity counts, action order, spatial relations, duration/action budget, camera consistency, physics. If any edit fails, revert it.}

    \neurcoappcode{Non-negotiable:}

    \neurcoappcode{- Do not delete, weaken, reassign, reorder, or contradict an explicit source fact.}

    \neurcoappcode{- Do not invent people, objects, actions, dialogue, captions, music, sound effects, voice-over, logos, or scene cuts unless requested or strictly necessary to express the source.}

    \neurcoappcode{- Do not turn a requested action into a static pose or a sequence that cannot fit the duration.}

    \neurcoappcode{- Preserve intent; preserve a static camera when the source is static.}

    \neurcoappcode{MODULES:}

    \neurcoappcode{<insert all seven module blocks below, in order>}

    \neurcoappcode{Answer only with JSON {"prompt": "<the rewritten prompt>"}}
\end{neurcoapppromptbox}

\begin{neurcoapppromptbox}{TIDE: user message}
    \raggedright
    \neurcoappcode{USER REQUEST:}

    \neurcoappcode{<full source prompt>}
\end{neurcoapppromptbox}

\subsection{Rule Modules}
\label{sec:app_tide_rule_modules}

Each rule supplies its ID, trigger, and instruction. All seven are sent
to the rewriter, which is asked to apply only source-triggered rules.

\begin{neurcoapppromptbox}{TIDE rule 1: intent preservation}
    \raggedright
    \neurcoappcode{- [vista.intent_preservation] TRIGGER: Always apply to a user request or an existing Video Prompt.}

    \neurcoappcode{INSTRUCTION: Fully address the User Prompt, preserve its core actions and intent, and include only elements explicitly required or clearly implied. Do not add unnecessary competing events.}
\end{neurcoapppromptbox}

\begin{neurcoapppromptbox}{TIDE rule 2: duration and scenes}
    \raggedright
    \neurcoappcode{- [vista.duration_scene_budget] TRIGGER: The source explicitly specifies a duration, a single scene or take, or a multi-scene structure whose time allocation must be preserved.}

    \neurcoappcode{INSTRUCTION: Use a single coherent scene that fits the stated duration. Make the scene's script, events, and visual flow end within that duration; detail the requested action rather than adding transitions or unrelated events.}
\end{neurcoapppromptbox}

\begin{neurcoapppromptbox}{TIDE rule 3: scene components}
    \raggedright
    \neurcoappcode{- [vista.scene_components] TRIGGER: The source omits visual presentation details needed to render a coherent shot, such as environment, framing, or lighting.}

    \neurcoappcode{INSTRUCTION: Specify the central subjects, stable visible traits, initial screen positions, purposeful actions, the visual environment, one complementary camera plan, and lighting or mood when useful. Keep every component subordinate to the core message.}
\end{neurcoapppromptbox}

\begin{neurcoapppromptbox}{TIDE rule 4: motion and physics}
    \raggedright
    \neurcoappcode{- [vista.motion_physics] TRIGGER: The source requests visible subject motion, a physical interaction or force, an object state change, or an explicitly static living subject that needs minimal non-state-changing motion cues.}

    \neurcoappcode{INSTRUCTION: The action timeline is the single highest-priority requirement of the prompt. State the action as the first element, before any setting, character description, or camera detail. If the model's prior bias defaults to a generic motion for the subject type (e.g., walking for a quadruped, swimming for a fish), explicitly forbid that default motion and require the unique action. If the model cannot generate the requested action without artifacts, simplify it to a single clear, continuous motion that the model can reliably generate, and include explicit timing and consistency constraints. Explicitly require that the action is visibly progressing in every single frame with no frozen frames, no long static pauses, no looping/repeating motion, and no still-image output.}

    \neurcoappcode{Translate the requested action into a detailed observable timeline: state initial pose and screen position, direction and path of travel, continuous changes in speed or orientation, physical contact and force, relevant environmental or material response, and a visible end state in order. Bind every phase to the same subject and keep the event budget feasible for the short clip.}

    \neurcoappcode{If the subject is in a resting/static state, require at least one continuous, visible motion cue throughout the clip (e.g., slow camera push-in, breathing, blinking, ear twitch) that does not alter the subject's pose or position; explicitly state the exact count and timing of each micro-action (e.g., 'exactly one blink in the final 1-2 seconds, no blinking before that'), and forbid any premature, repeated, or extra micro-actions. Explicitly require that these motion cues are clearly visible and not merely implied, and state that the subject must not appear frozen, statue-like, or as a still image. When the subject is stationary or the motion is minimal, explicitly require that the entire background is static with no moving objects (e.g., cars, people, animals, swaying trees) unless the source explicitly requests them. Provide a positive description of the background content (e.g., 'softly blurred garden with a brick wall, plants, and sky') and explicitly forbid the appearance of any moving elements.}

    \neurcoappcode{If the action involves spatial displacement (e.g., walking, running, flying), require that the subject's position changes relative to the background, with visible parallax or background shift, and include environmental interaction (e.g., footprints, grass bending, water ripples) to ground the motion physically. Critically, when the subject is supposed to move across the frame, explicitly require that the subject's body translates through the frame (i.e., its center of mass moves horizontally or vertically relative to the static background) and that each leg movement results in a corresponding forward displacement. Explicitly forbid the subject from performing leg movements that do not translate the body (e.g., 'walking in place', 'moonwalking', 'sliding' without actual position change).}

    \neurcoappcode{When the source specifies a direction of travel relative to the frame (e.g., 'left to right', 'toward the camera', 'across the screen'), explicitly define the camera-subject geometry: state whether the subject is seen in side profile (path parallel to the camera plane) or frontal (path toward/away from the camera), and require that the camera movement matches that geometry (e.g., lateral tracking for side profile, static or push-in for frontal). Explicitly forbid the opposite interpretation (e.g., if lateral movement is requested, forbid the subject approaching the lens; if frontal movement is requested, forbid side-scrolling).}

    \neurcoappcode{When the source specifies a directional force (e.g., a breeze, wind, or push), explicitly state that the force comes from a single, fixed source (e.g., 'a single off-screen fan at screen-left') and require that the resulting motion (e.g., hair, clothing, leaves) flows continuously in the specified direction for the entire clip, never reversing, oscillating, or becoming chaotic; explicitly forbid the motion from switching to the opposite direction or fanning out symmetrically at any point, and require that the final frame still shows the motion in the specified direction. Ensure transitions between action phases are gradual and physically plausible, with no abrupt jumps in speed, position, or posture.}

    \neurcoappcode{When the source specifies a camera movement or angle, explicitly restate that camera movement type and forbid any zoom, push-in, pull-out, or field-of-view change unless the source explicitly requests it; require that the subject's on-screen size remains roughly constant when the camera is purely translational. When the action involves an object interaction (e.g., lifting, gripping, drinking), explicitly state the object's weight and require that it remains in continuous contact with the subject or a supporting surface at all times, with no floating or teleporting. When the action involves a specific tool, vehicle, or equipment, explicitly name the exact equipment type and require that it is a single unbroken physical piece of the correct type, with no visually similar but incorrect alternatives.}

    \neurcoappcode{When the action involves consumption, transfer, breaking, or any transformation of an object's state (e.g., eating, drinking, pouring, filling, biting, breaking), explicitly describe the intermediate states of the object (e.g., whole, bitten, partially consumed, core, cracked, shattered) and require that the object's state changes gradually and continuously across frames, with no abrupt disappearance, appearance, or teleportation of parts. State the initial and final object state explicitly (e.g., 'plates start full and end partially emptied').}

    \neurcoappcode{Critically, when the action creates, builds, or transforms an object or pattern (e.g., drawing, pouring, assembling, cooking), explicitly require that the initial frame shows the pre-action state (e.g., blank surface, empty container, unassembled parts) and that the final state (e.g., completed pattern, filled container, assembled structure) is only fully visible in the final frame; forbid the final state from appearing at the start or being present before the action completes. Add micro-movements to the subject (e.g., breathing, blinking, ear twitches, muscle contractions, weight shifts) that are visible and continuous throughout the clip, to counter the frozen-frame or still-image look.}

    \neurcoappcode{Ensure the action has a clear non-looping arc: a defined start state, a middle progression, and a distinct end state (e.g., stopping, arriving, completing the motion, partially consumed object), with no repetition of the same frames or cycle. If a trade-off is necessary, state the priority: the action must be visibly progressing and the object state must visibly change, even if other visual elements are simplified.}

    \neurcoappcode{For process-oriented actions (e.g., building, weaving, assembling, cooking, drawing), explicitly enumerate the sequential stages of the process (e.g., 'first X, then Y, then Z') and require that each stage is visibly completed before the next begins, with the subject continuously engaged in the action (e.g., manipulating materials, moving limbs) and the intermediate state of the object/structure remaining visible and attached throughout. Forbid the subject from performing unrelated locomotion (e.g., walking away) or remaining static while the object changes state. If the action is time-compressed, specify that the compression is achieved by accelerating the same continuous motion, not by cutting or skipping stages.}

    \neurcoappcode{For living subjects in motion, describe the motion in biomechanical terms (e.g., four-beat rhythm, full extension and retraction, visible suspension phase, natural head bob) rather than generic 'steady' or 'consistent' speed, and require that the subject's anatomy, markings, body shape, and the density of any generated effects remain stable and continuous from frame to frame, with no morphing, flickering, warping, or duplication, so the clip reads as a single continuous take.}

    \neurcoappcode{When the source specifies a demeanor, mood, or emotional state (e.g., 'menacing', 'calm', 'elegant', 'sad', 'smiling', 'cheerful'), explicitly restate that demeanor as a hard constraint and require that the subject's facial expressions and body language remain consistent with it throughout the clip. Define the expression mechanically where possible: specify the exact configuration of facial features that constitutes the expression (e.g., 'the upper teeth form a fixed upward-curved crescent; the lower jaw remains slightly below, keeping a narrow smile-like mouth gap no wider than a finger-width; never a large dark oral cavity'). Explicitly forbid any action or expression that contradicts the requested demeanor (e.g., if 'calm' is requested, forbid smiling, laughing, or exaggerated gestures; if 'menacing' is requested, forbid cheerful or welcoming behavior; if 'smiling' is requested, forbid yawning, screaming, snarling, grimacing, or opening the mouth wide). State that the subject must not perform any action that is not explicitly listed in the prompt, even if physically plausible, when such an action would undermine the requested mood or narrative.}

    \neurcoappcode{When the action involves a state change of an object (e.g., opening, closing, filling, emptying), explicitly require that the initial frame shows the object in its pre-action state (e.g., closed, empty, whole) and that the object's state changes gradually and continuously across frames, with the final state (e.g., open, full, broken) only fully visible at the end of the action; forbid the final state from appearing in the first frame or before the action begins. Add a temporal breakdown to the prompt when the action has multiple distinct phases (e.g., '0-2s: lift the closed fan, 2-4s: open the fan, 4-7s: wave the fan, 7-10s: rest'), and require that each phase is visibly completed before the next begins, with no skipping or merging of stages.}

    \neurcoappcode{When the action involves a sequence of distinct sub-actions (e.g., 'scoop feed, extend hand, goat eats from palm, then lower bucket'), explicitly require that the initial frame shows the first sub-action's starting state (e.g., bucket held at waist, hand empty) and that each subsequent sub-action is visibly initiated and completed in order, with no sub-action skipped or replaced by a different action (e.g., pouring feed instead of hand-feeding); forbid the final state of any sub-action (e.g., bucket on ground, goats eating from bucket) from appearing before that sub-action is reached in the sequence.}

    \neurcoappcode{When the action is a sequence of subtle micro-actions (e.g., blinking, breathing, slight head turn), explicitly state the exact count and timing of each micro-action (e.g., 'exactly one blink in the final 1-2 seconds, no blinking before that'), and forbid any premature, repeated, or extra micro-actions that would break the intended choreography.}

    \neurcoappcode{If the action involves a hand or limb interacting with an object (e.g., adjusting glasses), explicitly require that the limb remains a coherent, anatomically correct shape in every frame, with no distortion, blurring, or teleportation during entry or exit, and that the final pose matches the prompt's end state (e.g., 'hand lowered to side', not 'hand pointing up'). Establish a priority hierarchy: the action progression and object state change are mandatory; secondary visual details (e.g., background movement, texture details) are enhancements that must not compromise the primary action.}
\end{neurcoapppromptbox}

\begin{neurcoapppromptbox}{TIDE rule 5: exit and re-entry}
    \raggedright
    \neurcoappcode{- [vista.frame_exit_reentry] TRIGGER: A requested subject leaves the frame and later returns, crosses a frame boundary, or temporarily becomes occluded during a continuous action.}

    \neurcoappcode{INSTRUCTION: State the exit edge or occluder, preserve the subject's appearance and direction of travel while off-screen, and specify the physically consistent re-entry edge, timing, and resumed motion. Do not invent a re-entry when the source action ends off-screen.}
\end{neurcoapppromptbox}

\begin{neurcoapppromptbox}{TIDE rule 6: identity and count}
    \raggedright
    \neurcoappcode{- [vista.identity_count_guard] TRIGGER: The source explicitly requires an exact entity count, multiple persistent subjects, a specifically named person or individual, a specific species or breed, or explicitly emphasizes one persistent object as unique. Ordinary singular articles such as 'a' or 'an' do not activate this rule by themselves.}

    \neurcoappcode{INSTRUCTION: State the exact count of each subject type as the first mandatory sentence and repeat it in the final clause. Give each persistent subject or object a stable role, appearance cue (e.g., color, markings, body shape, fur texture), and separate screen position when needed for tracking.}

    \neurcoappcode{When a specific species, breed, or type is named, explicitly restate that exact identity and its distinguishing visual features, requiring that the generated subjects match those features exactly. Additionally, explicitly name the most visually similar incorrect alternative species/breed and require that the generated subjects do NOT exhibit that alternative's distinguishing features.}

    \neurcoappcode{When the source names a real or fictional individual with distinctive recognizable features, explicitly require that the subject's facial features, hairstyle, body proportions, and other distinctive physical characteristics match that individual's recognizable appearance, and explicitly forbid the subject from being rendered as a generic, idealized, or different person; name the most likely incorrect substitution and require that the subject does NOT resemble that alternative.}

    \neurcoappcode{Explicitly forbid the subject from being rendered as a human wearing a mask, costume, or prosthetic of that species, and forbid any animal head or face from appearing on a human body or vice versa; require that the subject is a full-bodied, anatomically coherent animal of the exact named species with natural fur, skin, and body proportions. Preserve those identities and counts through every action phase, explicitly requiring that the same subjects remain present in every frame with no duplication, disappearance, appearance from nowhere, morphing into another entity, face swapping, mask swaps, costume changes, or any transformation of one character into another.}

    \neurcoappcode{When multiple subjects are present, explicitly require that each subject's body outline, head, ears, and face remain distinct and fully visible, with no subject hidden under another or merging into a single mass.}

    \neurcoappcode{When the source specifies a spatial relationship between subjects (e.g., distance, separation, or containment), explicitly state that the subjects must maintain that relationship throughout the clip, never touching, overlapping, clipping through, or entering each other's space, and forbid any subject from approaching closer than the specified distance.}

    \neurcoappcode{When the source specifies a positional layout (e.g., 'third from the left', 'in the center', 'on the right side'), explicitly restate that layout as a hard spatial constraint and require that the camera framing keeps every subject fully visible in that exact arrangement for the entire clip; explicitly forbid any camera zoom, push-in, pull-out, pan, tilt, or reframing that would crop, obscure, or reposition any subject relative to the specified layout.}

    \neurcoappcode{When a single object is specified, explicitly state that it remains one unbroken physical piece with no splitting, duplication, or morphing during any contact, grip, or motion phase; explicitly require that exactly one such object exists in the entire video, and that no similar-looking objects, decorations, or themed items appear anywhere in the scene, including on other objects or in the background.}

    \neurcoappcode{When the source specifies a single object that must carry or contain multiple subjects (e.g., one bicycle with two riders, one vehicle with multiple passengers), explicitly state that the object is a single unbroken physical piece and that all subjects are already positioned on or in it in the first frame, remaining there in every frame until the final frame; explicitly forbid the generation of multiple instances of that object, forbid any subject from appearing on a separate object, and forbid any subject from entering the frame from the side or from behind an obstacle.}

    \neurcoappcode{If occlusion occurs, specify that the occluded subject re-emerges continuously from the same direction. Add explicit negative constraints against count drift, missing subjects, identity switching, overlapping bodies, extra subjects appearing beyond the specified count, species substitution, generic-subject substitution, and duplicate objects. Require that the subject count and identity be explicitly verified at the start, middle, and end of the clip, and that no new subjects enter the frame from off-screen or from behind existing subjects.}
\end{neurcoapppromptbox}

\begin{neurcoapppromptbox}{TIDE rule 7: visual-only delivery}
    \raggedright
    \neurcoappcode{- [vista.visual_only_delivery] TRIGGER: Audio, dialogue, narration, captions, subtitles, or visible text are not explicitly requested.}

    \neurcoappcode{INSTRUCTION: Keep the prompt visual-only. Add no dialogue, narration, voice-over, music, sound effects, captions, subtitles, logos, or other visible text.}
\end{neurcoapppromptbox}


\end{document}